\documentclass[11pt,twoside,a4paper]{report}

\usepackage[DETI,newLogo,final]{uaThesis}
\def\ThesisYear{2018}
\usepackage{natbib} % various citation commands

\usepackage{pdflscape}
\usepackage{graphicx}
\usepackage{subfigure} % for images with a), b), etc
\graphicspath{{images/}}
\usepackage[font=footnotesize]{caption}

\usepackage[T1]{fontenc}
\usepackage[utf8]{inputenc}
\usepackage[english]{babel}

\usepackage{amsmath}

\usepackage[hidelinks]{hyperref}
\usepackage[ruled]{algorithm2e}
\usepackage{multicol}
\usepackage{multirow}
\usepackage{url}
\usepackage{algorithmicx}
\usepackage[flushleft]{threeparttable}
\usepackage[T1]{fontenc}
\usepackage[flushleft]{threeparttable}
\usepackage{enumerate}
\usepackage{breakcites} % break long citation in new line
\usepackage[bottom]{footmisc} % put footnote under figures
\usepackage{centernot}
\usepackage{float}
\usepackage{algpseudocode}
\usepackage{soul}
\usepackage{wrapfig}
\usepackage{multirow}
\usepackage{color}
\definecolor{red}{rgb}{1.00,0.00,0.00}
\definecolor{blue}{rgb}{0.00,0.00,1.00}
\definecolor{green}{rgb}{0.4,1.00,0.0}
\definecolor{yellow}{rgb}{0.5,0.5,0.0}
\definecolor{dgreen}{rgb}{0.0,0.5,0.0}

\newcommand{\cblue}[1] {\textcolor{blue}{#1}}

\usepackage{caption}
\usepackage[nolist,nohyperlinks]{acronym}
\usepackage{algorithm2e}
\usepackage{algpseudocode}

\usepackage{acronym}
\acrodef{ROS}{Robot Operating System}
\acrodef{TID}{Tracked Object Identification}
\acrodef{VID}{Object View Identification}
\acrodef{FID}{Extracted Feature Identification}
\acrodef{OID}{Anchored Object Identification}
\acrodef{CN}{Category Name}
\acrodef{CID}{Category Identification}
\acrodef{OD}{Object Detection}
\acrodef{ORK}{Object Recognition Kitchen}
\acrodef{OC}{Object Conceptualizer}
\acrodef{OT}{Object Tracker}
\acrodef{TS}{Tabletop Segmenter}
\acrodef{OR}{Object Recognizer}
\acrodef{OA}{Object Anchoring}
\acrodef{FE}{Feature Extractor}
\acrodef{ST}{Skeleton Tracker}
\acrodef{GR}{Gesture Recognizer}
\acrodef{UI}{Interface Manager}
\acrodef{SMem}{Semantic Memory}
\acrodef{PMem}{Perceptual Memory}
\acrodef{PCL}{Point Cloud Library}
\acrodef{3D}{3D}

\usepackage{fancyhdr}
\usepackage{xintexpr}
\usepackage[toc,page]{appendix}

\begin{document}

%%%%%%%%%%%%%%%%%%%%%%%%%%%%%%%%%%%%%%%%%%%%%%%%%%%%%%%%%%%%%%%%%%%%%%%%%%%%%%%%

%%%%%%%%%%%%%%%%%%%%%%%%%%%%%
%%%%%% Chapter name in the header
\renewcommand{\chaptermark}[1]{\markboth{#1}{}}

%
% Cover page (use only one of the first two \TitlePage)
%

% First alternative, with a figure
\TitlePage
  %\GRID  % for debugging ONLY
  \HEADER{\BAR%
          \FIG{\includegraphics[height=60mm]{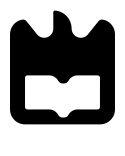}}} % the \FIG{} is optional
         {\ThesisYear}

  \TITLE{Seyed Hamidreza Mohades Kasaei}
        {Interactive Open-Ended Learning for 3D Object Recognition}
        \vspace{5mm}
  \TITLE{}
        {Aprendizagem  contínua interativa para Reconh- ecimento de objetos 3D \\ \vspace{15cm} \small The original version of this thesis is also available at the repository of the university:  \href{https://ria.ua.pt/handle/10773/26803}{https://ria.ua.pt/handle/10773/26803}}

\EndTitlePage
\titlepage\ \endtitlepage % empty page

%
% Initial thesis pages
%

\TitlePage
  \HEADER{}{\ThesisYear}
  \TITLE{Seyed Hamidreza Mohades Kasaei}
        {Interactive Open-Ended Learning for 3D Object Recognition}
         \vspace{5mm}
  \TITLE{}
        {Aprendizagem  contínua interativa para Reconh- ecimento de objetos 3D}

  \vspace*{15mm}
  \TEXT{}
       {Disserta\c c\~ao apresentada \`a Universidade de Aveiro para cumprimento dos requesitos
        necess\'arios \`a obten\c c\~ao do grau de Doutor em Engenharia Informática,
        realizada sob a orienta\c c\~ao cient\'\i fica de Lu\'{i}s Filipe de Seabra Lopes e Ana Maria Perfeito Tom\'{e},
        professores do Departamento de Electr\'onica, Telecomunica\c c\~oes e Inform\'{a}tica da Universidade de Aveiro}
\EndTitlePage
\titlepage\ \endtitlepage % empty page

\TitlePage
  \vspace*{55mm}
  \TEXT{\textbf{The jury\newline}}
       {}
  \TEXT{presidente~/~president}
       {\textbf{Professor Doutor Helmuth Robert Malonek}\newline {\small
        Professor Catedr\'atico da Universidade de Aveiro}}
  \vspace*{5mm}
  \TEXT{vogais~/~examiners committee}
     {\textbf{Doutor Lu\'{i}s Filipe Barbosa de Almeida Alexandre}\newline {\small
    	Professor Catedr\'atico da Universidade da Beira Interior}}
      \vspace*{5mm}
  \TEXT{}
       {\textbf{Doutor Alexandre Jos\'{e} Malheiro Bernardino}\newline {\small
        Professor Associado, Instituto Superior T\'{e}cnico, Universidade de Lisboa}}
          \vspace*{5mm}
	\TEXT{}
       {\textbf{Doutor Paulo Martins de Carvalho}\newline {\small
        Professor Associado, Universidade do Minho}}
        \vspace*{5mm}
  \TEXT{}
       {\textbf{Doutor Lu\'{i}s Filipe de Seabra Lopes (orientador)}\newline {\small
		Professor Associado, Universidade de Aveiro }}
  \vspace*{5mm}

  \TEXT{}
       {\textbf{Doutora Ana Raquel Ferreira de Almeida Sebasti\~ao}\newline {\small
        Investigadora Doutorada, Universidade de Aveiro}}

\EndTitlePage
%\titlepage\ \endtitlepage % empty page
\newpage
\newpage
\TitlePage
  \vspace*{55mm}
  \TEXT{\textbf{Acknowledgements}}
       {As I finish this thesis I have the pleasure of recounting the people that have shaped my experience. My most sincere thanks go to my supervisors Prof. Lu\'{i}s Seabra Lopes and Prof. Ana Maria Tom\'{e} for their support, guidance, and encouragement during the work on this dissertation. Their efforts in shaping my Ph.D. were absolutely invaluable.

Two colleagues have contributed to the part of the work in this thesis. Miguel Oliveira contributed to the development of the gesture recognizer module, an extended version of which became part of this thesis. Gi Hyun Lim contributed to the development of software for the memory system. I was fortunate to have Miguel and Gi Hyun as my colleagues, in a new country and a new lab. Their technical knowledge and their patience with my questions have helped me greatly. I am extremely thankful to Vahid Mokhtari and Juil Sock for their friendship, guidance, stimulating conversations and help when I needed. I am lucky to have you as a friend. You guys have been amazing, and I am so glad to met you. 

I would also like to thank the RACE — Robustness by Autonomous Competence Enhancement — project partners for their efforts in the integration and the demonstrations, and especially to the Technical Aspects of Multi-modal Systems (TAMS) group, University of Hamburg, for making the PR2 robot available to the project. My time at RACE project as a researcher has been like no other in my life.

I have had the privilege of collaborating with another professor at Imperial College London, from whom I have learned a set of great ideas. Prof. Tae-Kyun Kim has an enthusiasm for research and is an excellent source of encouragement. I am grateful for having had the opportunity to immerse myself in such an intellectually stimulating environment. 

I want to thank my parents, my brothers and my sister for their constant encouragement, strength, and support. This thesis, in many ways, is a result of the dedication of my parents to the education of their children. Mere words cannot suffice to express my gratitude. Fatemeh, your patience, strength and uncommon ability to empathize has made this journey possible for me. You have suffered with me and for me. I find myself very fortunate to have you by my side. Thank you.

Finally, I want to thank the Portuguese Science Foundation (FCT) for the Ph.D. scholarship grant. I am very thankful to Institute of Electronics and Informatics Engineering of Aveiro (IEETA), my research institution, for providing the financial support for attending several conference events. I would also like to thank the open-source community for sharing and spreading the knowledge, for all to use.

}
  \TEXT{}
       {}
\EndTitlePage
\titlepage\ \endtitlepage % empty page

\newpage

\TitlePage
\vspace*{25mm}
\TEXT{\textbf{Palavras-chave}}
{Percepção visual, aprendizagem aberta de categorias, arquiteturas de
aprendizagem, interação homem-robô, robótica.}
\vspace{10mm}
\TEXT{\textbf{Resumo}}
{As abordagens atuais de aprendizagem e reconhecimento de categorias de objetos são tipicamente pensadas para ambientes estáticos, nos quais é viável separar o treino
(off-line) e a utilização do conhecimento aprendido (on-line). Em tais cenários, o conhecimento é estático, no sentido em que a representação das categorias não muda após a fase de treino. No entanto, para migrar um robô para um novo ambiente torna-se muitas vezes necessário redesenhar completamente a base de conhecimento.
\\
A tese contribui em várias frentes para a investigação em aprendizagem e reconhecimento de categorias de objetos 3D. Para lidar com as mencionadas limitações, olhamos para a cognição humana, em particular para o fato de o ser humano aprender incessantemente a reconhecer categorias de objetos. Essa capacidade de refinar e extender o conhecimento com base na experiência acumulada facilita a adaptação a novos ambientes. Inspirados por essa capacidade, procuramos criar uma arquitetura cognitiva para percepção de objetos e aprendizagem perceptual capaz de aprender categorias de objetos 3D de maneira aberta. Neste contexto, o conjunto de categorias a serem aprendidas é inicialmente desconhecido e as instâncias a usar no treino são gradualmente extraídas das observações do agente, em vez de estarem disponíveis desde o início do processo. Assim, esta arquitetura fornece capacidades de percepção que permitirão que os robôs aprendam categorias de forma incremental com base nas experiências acumuladas e raciocinar sobre execução de tarefas complexas. A arquitetura integra detecção, seguimento, ensino, aprendizagem e reconhecimento de categorias de objetos.
\\
Uma parte importante deste trabalho centra-se na representação de objetos, a qual deve
ser fiável e calculável em tempo real, para permitir que o robô interaja fisicamente
com os objetos no seu ambiente. Nós abordamos o problema da representação, propondo um novo descritor global de objetos 3D designado Global Orthographic Object Descriptor (GOOD). Este descritor distingue-se de outras representações globais no facto de ser rápido de calcular, robusto contra variações na pose, variações na densidade de amostragem e ruído. Propomos ainda uma modificação da técnica de Latent Dirichlet Allocation para aprender característica semânticas (tópicos) com base em co-ocorrências
de características locais. 
\\
O problema central nesta tese é a aprendizagem aberta para reconhecimento de categorias de objetos 3D é. Foram exploradas abordagens, quer baseadas em instâncias, quer baseada em modelos, para a aprendizagem incremental e aberta de categorias. Finalmente, uma nova metodologia de avaliação experimental, que leva em conta a natureza aberta da
aprendizagem de categorias em cenários multi-contexto, é proposta e utilizada.
Foi realizada avaliação experimental sistemática, em múltiplos cenários experimentais,
das várias abordagens propostas. Os resultados experimentais mostram que o sistema proposto é capaz de interagir com utilizadores humanos, aprender novas categorias de objetos ao longo do tempo e realizar tarefas complexas. As contribuições apresentadas nesta tese foram totalmente implementados e avaliados em diferentes conjuntos de dados, quer de objetos, quer de cenas, e avaliados empiricamente em diferentes  plataformas robóticas.
}
\EndTitlePage
\titlepage\ \endtitlepage % empty page

\TitlePage
\vspace*{25mm}
\TEXT{\textbf{Keywords}}
{3D object perception, open-ended learning of object categories, architectures of
learning, human-robot interaction, robotic.}
\vspace{10mm}

  \TEXT{\textbf{Abstract}}
	{%To deploy a robot in a human-centric environment, it is important that the robot is able to continuously acquire and update object categories while working in the environment.
 Current object category learning and recognition approaches are typically designed for static environments in which it is viable to separate the training (off-line) and testing (on-line) phases. In such scenarios, the learned object category models are static, in the sense that the representation of the known categories does not change after the training stage. However, to migrate a robot to a new environment one must often completely redesign and remodel the knowledge-base that it is running with. 
%Therefore, these systems are unable to adapt to dynamic environments. 
%This leads to several shortcomings such as the inability to detect/recognize new or unknown categories. 

The thesis contributes in several important ways to the research area of 3D object category learning and recognition.
To cope with the mentioned limitations, we look at human cognition, in particular at the fact that human beings learn to recognize object categories ceaselessly over time. This ability to refine and extend knowledge from the set of accumulated experiences facilitates the adaptation to new environments. Inspired by this capability, we seek to create a cognitive object perception and perceptual learning architecture that can learn 3D object categories in an open-ended fashion. In this context, ``open-ended'' implies that the set of categories to be learned is not known in advance, and the training instances are extracted from actual experiences of a robot, and thus become gradually available, rather than being available since the beginning of the learning process. 
%A 3D object perception and perceptual learning architecture developed for a complex artificial cognitive agent working in a restaurant scenario.
This architecture provides perception capabilities that will allow robots to incrementally learn object categories from the set of accumulated experiences and reason about how to perform complex tasks. This framework integrates detection, tracking, teaching, learning and recognition of objects. %The system learns in an incremental and open-ended way from user-mediated experiences.

An important part of this work is concerned with the object representation.
This is one of the most challenging problems in robotics because it must provide reliable information in real-time to enable the robot to physically interact with the objects in its environment. We have first tackled the problem of object representation, by proposing a new global object descriptor named Global Orthographic Object Descriptor (GOOD).% designed to be robust, descriptive and efficient to compute and use. 
This descriptor distinguishes itself from alternative 3D global object representations in that it is very fast to compute, robust against variations in pose and sampling density, and copes well with noisy sensor data. We also propose an extension of Latent Dirichlet Allocation to learn structural semantic features (i.e. topics) from local feature co-occurrences for each object category independently. %Although the model we developed has been intended to be used for object category learning and recognition, it is a novel probabilistic model that can be used in the fields of computer vision and machine learning.  

Open-ended learning for 3D object category recognition is the core problem in this thesis.  Both instance-based and model-based approaches were explored for incrementally scaling-up to larger sets of categories. Finally, a novel experimental evaluation methodology, that takes into account the open-ended nature of object category learning in multi-context scenarios, is proposed and applied. An extensive set of systematic experiments, in multiple experimental settings, was carried out to thoroughly evaluate the described learning approaches. 
Experimental results show that the proposed system is able to interact with human users, learn new object categories over time, as well as perform complex tasks.
The contributions presented in this thesis have been fully implemented and evaluated on different standard object and scene datasets and empirically evaluated on different robotic platforms. 
%The results indicate that all approaches were able to incrementally acquire new categories. 

}
%       {Nowadays, it is usual to evaluate a work \ldots}
\EndTitlePage
\titlepage\ \endtitlepage % empty page

%\TitlePage
 % \vspace*{55mm}
  %\TEXT{\textbf{Abstract}}
	%{\input{chapters/abstract_p2}}
%       {Nowadays, it is usual to evaluate a work \ldots}
%\EndTitlePage
%\titlepage\ \endtitlepage % empty page

%
% Tables of contents, of figures, ...
%

%\cleardoublepage
%\listofalgorithms

\fancyhf{} % clear the headers

\pagenumbering{roman}
\phantomsection
\fancyhead[R]{\emph{Contents}}
\tableofcontents
\clearpage\fancyhead[R]{}
\cleardoublepage

\pagenumbering{roman}
\fancyhead[R]{\emph{List of Figures}}
\listoffigures
\clearpage\fancyhead[R]{}
\cleardoublepage
\pagenumbering{roman}

\listoftables
\clearpage\fancyhead[R]{}
\cleardoublepage

\fancyhead[R]{%
   % We want italics
   \itshape
   % The chapter number only if it's greater than 0
   \ifnum\value{chapter}>0 \chaptername\ \thechapter. \fi
   % The chapter title
   \leftmark}
\fancyfoot[C]{\thepage}

\fancypagestyle{plain}{
  \renewcommand{\headrulewidth}{0pt}
  \fancyhf{}
  \fancyfoot[C]{\thepage}
}

\def\AddVMargin#1{\setbox0=\hbox{#1}%
                  \dimen0=\ht0\advance\dimen0 by 2pt\ht0=\dimen0%
                  \dimen0=\dp0\advance\dimen0 by 2pt\dp0=\dimen0%
                  \box0}   % add extra vertical space above and below the argument (#1)
\def\Header#1#2{\setbox1=\hbox{#1}\setbox2=\hbox{#2}%
           \ifdim\wd1>\wd2\dimen0=\wd1\else\dimen0=\wd2\fi%
           \AddVMargin{\parbox{\dimen0}{\centering #1\\#2}}} % put #1 on top #2

%% Front pages (title, jury, abstract, toc)

\cleardoublepage
\pagenumbering{arabic}

%%%%%%%%%%%%%%%%%%%%%%%%%%%%%
%%%%%%%%%%%%%%%%%%%%%%%%%%%%%%%%%%%%%%%%%%%%%%%%%%%%%%%%%%%%%%%%%%%%%%%%%%%%%%%%
\chapter{Introduction}
\label{chapter_1}
Service robots are appearing more and more in our daily life. They are extremely useful because they can help elders or people with motor impairments to achieve independence in everyday tasks like delivering objects, e.g. serving a coffee / meal or cleaning tables \citep{ciocarlie2014towards}. One of the primary challenges of service robotics is the adaptation of robots to new tasks in changing environments, where they interact with non-expert users. Elderly, injured, and disabled people have consistently attributed a high priority to object manipulation tasks \citep{jain2010assistive}. Object manipulation tasks consist of two phases: the first is the perception of the object and the second is the planning and execution of arm or body motions which grasp the object and carry out the manipulation task. These two phases are closely related: object perception provides information to update the model of the environment, while planning uses this world model information to generate sequences of arm movements and grasp actions for the robot. Therefore object perception is one crucial component of a service robot besides capabilities like manipulation or navigation. 

In addition, assistive robots must perform the tasks in reasonable time. It is also expected that the competence of the robot increases over time, that is, robots must robustly adapt to new environments by being capable of handling new objects. However, it is not feasible to assume that one can pre-program all necessary object categories for assistive robots. Instead, robots should learn autonomously from novel experiences, supported in the feedback from human teachers. In order to incrementally adapt to new environments, an autonomous assistive robot must have the ability to process visual information and conduct learning and recognition tasks in a concurrent and open-ended fashion.
 
To cope with these issues, we explore how robots could learn incrementally 
from their own experiences as well as from interaction with humans. This thesis is a product of efforts in this direction.

%^^^^^^^^^^^^^^^^^^^^^^^^^^^^^^^^^^^^^^^^^^^^^^^^^^^^^^^^^^^^^^^^
\section {Motivation}
\label{motivation}

Several state-of-the-art assistive robots use traditional object category learning and recognition approaches \citep{leroux2013armen,beetz2011robotic}. 	
These classical approaches are often designed for static environments in which it is viable to separate the training (off-line) and testing (on-line) phases. In these cases, the world model is static, in the sense that the representation of the known categories does not change after the training stage. Although such systems have been shown to be useful in a variety of real world scenarios, they are unable to
adapt to dynamic environments~\citep{Jeong2012}. Therefore, most robots lack the ability to learn new objects from past experiences. To migrate a robot to a new environment one must often completely re-generate the knowledge-base that it is running with. 

In open-ended domains, it is not viable to hand-code all possible behaviours and to anticipate all possible exceptions. One of the challenging tasks in open-ended domains is object category learning and recognition because of the very large number of objects present in household environments and, moreover, due to the infinite variety in the appearance of those objects. Given this, the appropriate strategy to solve the problem is to make robots capable of learning on site, rather than to exhaustively program them before deployment. One of the early works on open-ended object category learning and recognition was carried out having in mind applications in language acquisition. The authors characterize their approach as follows: 

\begin{itemize}
  \item[] ``\emph{The learning approach is open-ended in that there is no set of words and meanings defined in advance, and new words and meanings are acquired incrementally through interaction with a human instructor.}'' (\cite{chauhan2011}; see also \cite{Seabra2007}; \cite{Lopes2008})
\end{itemize}

In other words, ``open-ended'' means that the robot does not know in advance which object categories it will have to learn, which observations will be available, and when they will be available to support the learning.

Towards this goal, cognitive robotics looks at human cognition as a source of inspiration for developing automatic perception capabilities that will allow robots to, incrementally learn object categories from the set of accumulated experiences and reason about how to perform complex tasks. In particular, humans learn to recognize object categories ceaselessly over time. This ability to refine knowledge from the set of accumulated experiences facilitates the adaptation to new environments. Inspired by such abilities, this thesis proposes different approaches towards 3D object category learning and recognition in an interactive and open-ended manner. This is necessary for service robots, not only to perform tasks in a reasonable amount of time and in an appropriate manner, but also to robustly adapt to new environments by handling new objects. In particular, we propose a set of perception capabilities that will allow robots to, incrementally learn object categories from the set of accumulated experiences and reason about how to perform complex tasks. To achieve these goals, it is critical to detect, track and recognize objects in the environment as well as to conceptualize experiences and learn novel object categories in an open-ended manner, based on human-robot interaction. 
%The central contribution of this thesis is to study and develop interactive open-ended learning approaches for 3D object recognition system to be utilized in autonomous service robots. 

This PhD project puts forward the development of a real-time system for acquiring and recognizing object categories in open-ended manner based on human-robot interaction. 
This work focuses on learning and recognizing table-top objects, which can be manipulated by the robot. In this context, many interesting and challenging issues regarding interactive open-ended 3D visual object category learning and recognition are addressed. 

\section {Objectives}
\label{objectives}

The main objective of this research was to study, design and develop an interactive open-ended 3D
object category learning and recognition system to be utilized in autonomous service robots. This kind of perception system comprises a significant number of software modules, which must be closely coupled in their structure and functionality. The software modules of the proposed system are including \emph{Pre-Processing}, \emph{Object Detection}, \emph{Feature Extraction}, \emph{Object Representation}, \emph{Object Conceptualization}, \emph{Object Recognition}, \emph{Perceptual Memory} and \emph{Human-Robot Interaction}. The developed perception and perceptual learning capabilities target objects in table-top scenes, e.g. in a restaurant environment. 
%These capabilities are fully integrated in the RACE: Robustness by Autonomous Competence Enhancement architecture \citep{RACE2013,Hertzberg2014projrep} (i.e., Fig.~\ref{fig:completerace}) and are running on the PR2 robot used by the project. 
Every module must be evaluated based on real or simulated data individually. The following specific objectives will be pursued:

\begin{itemize}
\item Design and develop a real-time modular 3D object recognition system for acquiring and recognizing object categories in open-ended manner based on human-robot interaction. 

\item Acquire a deep understanding about \emph{how to automatically detect, recognize and conceptualize objects in 3D unorganized scenes in open-ended manner}.  

\item Design and develop human-robot interaction capabilities for naming objects and
scenes, and for providing corrective feedback for learning. %All the aspects of the human-robot interaction are discussed in \cblue{Chapter 3}.

\item Acquire a deep understanding of point cloud processing and 3D shape descriptors.
This objective led to proposing a new object descriptor named Global Orthographic Object Descriptor (GOOD) as described in Chapter \ref{chapter_5}.

\item Investigate and use hierarchical object representation technique for facilitating object category learning and optimizing recognition as well as memory usage.
%The relevant contributions are detailed in \cblue{Chapter 6}.

\item Consider instance-based and model-based learning mechanisms to incrementally learn object categories from the set of accumulated experiences.

\item Design and develop a ``\emph{Simulated Teacher}'' to evaluate system performance in a systematic and comprehensive way in different scenarios, including multi-context scenarios.
\end{itemize}

%^^^^^^^^^^^^^^^^^^^^^^^^^^^^^^^^^^^^^^^^^^^^^^^^^^^^^^^^^^^^^^^^
%^^^^^^^^^^^^^^^^^^^^^^^^^^^^^^^^^^^^^^^^^^^^^^^^^^^^^^^^^^^^^^^^
\section {Research Context}
\label{research_context}

\begin{figure}[!b]
  \begin{center}
\includegraphics[width=1\linewidth, trim=0cm 0cm 0cm 0cm, clip=true]{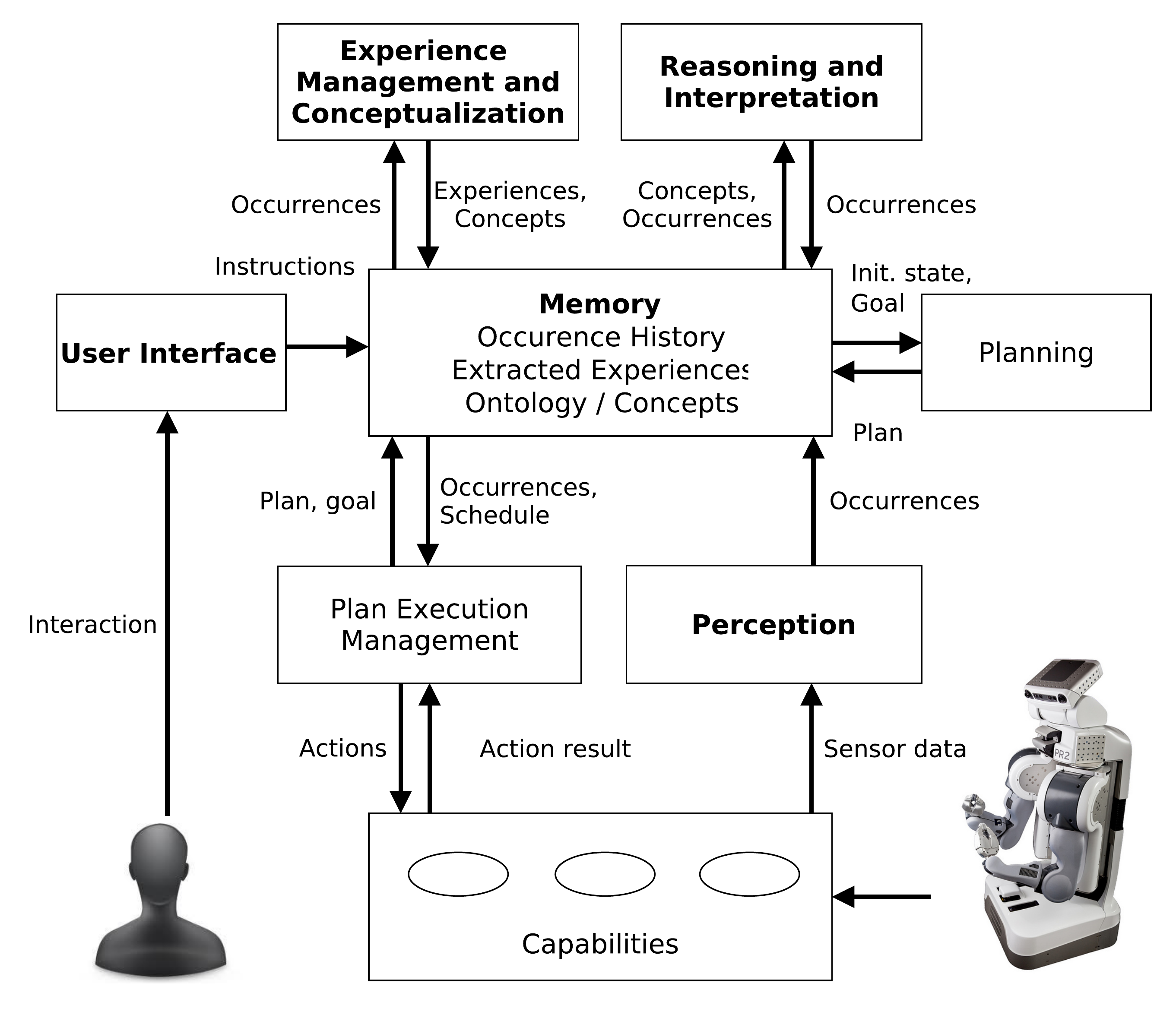}\\ 
  \end{center}
  \vspace{-5mm}
\caption{A high-level overviews of the RACE architecture}
\label{fig:completerace}
\end{figure}

This work started in the framework of the European project, RACE: Robustness by Autonomous Competence Enhancement \citep{RACE2013,Hertzberg2014projrep}. \emph{The overall aim of this project is to develop an artificial cognitive system, embodied by a service robot, able to build a high-level understanding of the world it acts in by storing and exploiting appropriate memories of its experiences\footnote{\url{http://project-race.eu/}}}. The RACE project assumed that versatility and competence enhancement can be obtained by learning from experiences. The project focused on acquiring and conceptualizing experiences about objects \citep{oliveira20153d}, scene layouts \citep{dubba2014grounding} and activities \citep{mokhtari2017learning} as a means to enhance robot competence over time thus achieving robustness. Stimuli for learning can be collected, either autonomously by robots, or when they receive appropriate feedback from users. The functional components of the RACE architecture are represented by boxes in Fig.~\ref{fig:completerace}. 

Each component may contain one or more modules, which are implemented as nodes (or nodelets) over the \emph{Robot Operating System} (ROS) \citep{Cousins2010a}. The \emph{Reasoning and Interpretation} component includes a temporal reasoner, a spatial reasoner and a description logics reasoner. \emph{Perception} contains several modules for symbolic proprioception and exteroception, which generate occurrences. The \emph{Experience Management and  Conceptualization} component pre-processes occurrences, extracts relevant experiences, uses them to create new concepts and stores these in the \emph{Memory} component. The \emph{User Interface} component receives instructions from the user and relays them to the \emph{Planning} component. Planning is carried out using SHOP2, a Hierarchical Task Network planner  \citep{Nau2003}. The produced plans are executed by the \emph{Plan Execution Management} component. It outputs a plan which is collected by the \emph{Plan Execution and Management} component. Finally, actions are conveyed to the robot actuators.

In this work, we study the topic of 3D object category learning and recognition in open-ended robotic domains and evaluate the proposed object recognition system in the
framework of the RACE project. In this context, ``open-ended'' implies that visual information arrives continuously, and learning and recognition are performed in an any time basis. The open-ended object learning system must update its models ceaselessly over time with limited computational resources. This type of perception system must comprise a significant number of software modules, which must be closely coupled in their structure and functionality. These modules might handle large size data, which they may need to store or transfer to other modules. 

In this thesis, two different cognitive perception architectures were explored, where the key difference between these architectures is the learning and recognition approach used. We evaluate the proposed object perception system in the framework of the RACE project. Later, the proposed methodology was also evaluated on standard datasets and using a robotic arm platform provided by the University of Aveiro.

%^^^^^^^^^^^^^^^^^^^^^^^^^^^^^^^^^^^^^^^^^^^^^^^^^^^^^^^^^^^^^^^^
\section {Publications}
\label{sec:publication}

The work presented in this thesis spawned a series of publications presented at major conferences and top journals in the field. Below is an excerpt from this list: 
%\cred{(note: I will add the full list of authors.)\\}
\begin{itemize}
\item \textbf{Journals}
\begin{enumerate}
	\item \textbf{\small {Kasaei, S. Hamidreza}}, et al. ``Local-LDA: Open-Ended Learning of Latent Topics for 3D Object Recognition''. IEEE Transactions on Pattern Analysis and Machine Intelligence (TPAMI) (2019). %\{Q1\}

		\item \textbf{\small {Kasaei, S. Hamidreza}}, et al. ``Towards Lifelong Assistive Robotics: A Tight Coupling between Object Perception and Manipulation''.  Neurocomputing Journal 291 (2018): 151-166. %\{Q1\}
			
	\item \textbf{\small {Kasaei, S. Hamidreza}}, et al. ``GOOD: A Global Orthographic Object Descriptor for 3D Object Recognition and Manipulation.'' Pattern Recognition Letters 83 (2016): 312-320. %\{Q1\}
		
	\item Oliveira Miguel, Lu{\'i}s Seabra Lopes, Gi Hyun Lim, \textbf{\small {Kasaei, S. Hamidreza}}, Ana Maria Tom\'{e}, ``3D object perception and perceptual learning in the RACE project.'' Robotics and Autonomous Systems 75 (2016): 614-626. %\{Q1\}
		
		\item \textbf{\small {Kasaei, S. Hamidreza}}, et al. ``Interactive Open-Ended Learning for 3D Object Recognition: An Approach and Experiments." Journal of Intelligent \& Robotic Systems 80, no. 3 (2015): 537-553. %\{Q2\}

%		\item Miguel Oliveira, Lu{\'i}s Seabra Lopes, Gi Hyun Lim, S. Hamidreza Kasaei, Ana Maria Tom\'{e}, Aneesh Chauhan ``3D Object Perception and Perceptual Learning in the RACE Project''
		\item Joachim Hertzberg, Jianwei Zhang, Liwei Zhang, Sebastian Rockel, Bernd Neumann, Jos Lehmann, Krishna S. R. Dubba, Anthony G. Cohn, Alessandro Saffiotti, Federico Pecora, Masoumeh Mansouri, Stefan Konecny, Martin Gunther, Sebastian Stock, Luis Seabra Lopes, Miguel Oliveira, Gi Hyun Li, \textbf{\small {Kasaei, S. Hamidreza}}, Vahid Mokhtari, Lothar Hotz, Wilfried Bohlken. ``The RACE Project~--~Robustness by Autonomous Competence Enhancement.'' Kunstliche Intelligenz 28, no. 4 (2014): 297-304.

	\end{enumerate}	
	\item \textbf{Conferences}
		\begin{enumerate}	
				
		\item \textbf{\small {Kasaei, S. Hamidreza}}, et al. ``Interactive Open-Ended Object, Affordance and Grasp Learning for Robotic Manipulation''. 2019 IEEE/RSJ International Conference on Robotics and Automation (ICRA 2019).
				
	\item \textbf{\small {Kasaei, S. Hamidreza}}, et al. ``Coping with Context Change in Open-Ended Object Recognition without Explicit Context Information''. 2018 IEEE/RSJ International Conference on Intelligent Robots and Systems (IROS 2018).
			
	\item \textbf{\small {Kasaei, S. Hamidreza}}, et al. ``Perceiving, Learning, and Recognizing 3D Objects: An Approach to Cognitive Service Robots.'' 2018 Thirty-Second AAAI Conference on Artificial Intelligence (AAAI 2018).
			
	\item \textbf{\small {Kasaei, S. Hamidreza}}, et al. ``Hierarchical Object Representation for Open-Ended Object Category Learning and Recognition.'' 2016 Advances in Neural Information Processing Systems (NIPS 2016), pp. 1948-1956.

	\item \textbf{\small {Kasaei, S. Hamidreza}}, et al. ``An Orthographic Descriptor for 3D Object Learning and Recognition''. 2016 IEEE/RSJ International Conference on Intelligent Robots and Systems (IROS 2016), pp. 4158-4163.

	\item Nima Shafii, \textbf{\small {Kasaei, S. Hamidreza}}, et al. ``Learning to Grasp Familiar Objects using Object View Recognition and Template Matching''. 2016 IEEE/RSJ International Conference on Intelligent Robots and Systems (IROS 2016), pp. 2895-2900.

\item \textbf{\small {Kasaei, S. Hamidreza}}, et al. ``Object Learning and Grasping Capabilities for Robotic Home Assistants''. RoboCup-2016: Robot Soccer World Cup, Lecture Notes in Computer Science, vol. 9776, Springer, 2016.

		\item \textbf{\small {Kasaei, S. Hamidreza}}, et al. ``Concurrent 3D Object Category Learning and Recognition based on Topic Modelling and Human Feedback''. 	2016 IEEE International Conference on Autonomous Robot Systems and Competitions (ICARSC 2016), pp. 329-334.

		\item Shafii, Nima, \textbf{S. Hamidreza Kasaei}, et al. ``A Learning Approach for Robotic Grasp Selection in Open-Ended Domains''. 2016 IEEE International Conference on Autonomous Robot Systems and Competitions (ICARSC 2016), pp. 112-117.

		\item \textbf{\small {Kasaei, S. Hamidreza}}, et al. ``An Adaptive Object Perception System based on Environment Exploration and Bayesian Learning''. 2015 IEEE International Conference on Autonomous Robot Systems and Competitions (ICARSC 2015), pp. 221-226.
						
		\item Oliveira Miguel, Lu{\'i}s Seabra Lopes, Gi Hyun Lim, \textbf{S. Hamidreza Kasaei}, Angel D. Sappa, Ana Maria Tom\'{e}, ``Concurrent Learning of Visual Codebooks and Object Categories in Open-ended Domains''. 2015 IEEE/RSJ International Conference on Intelligent Robots and Systems (IROS 2015), pp. 2488-2495.

		\item Lim, Gi Hyun, Miguel Oliveira, \textbf{S. Hamidreza Kasaei}, Lu{\'i}s Seabra Lopes, "Hierarchical Nearest Neighbor Graphs for Building Perceptual Hierarchies", 2015 International Conference on Neural Information Processing (ICONIP 2015). Springer, pp. 646-655.

		\item \textbf{\small {Kasaei, S. Hamidreza}}, et al. ``An Interactive Open-Ended Learning Approach for 3D Object Recognition'', 2014 IEEE International Conference on Autonomous Robot Systems and Competitions (ICARSC 2014), pp. 47-52.

		\item Oliveira Miguel, Gi Hyun Lim, Lu{\'i}s Seabra Lopes, \textbf{S. Hamidreza Kasaei}, Ana Maria Tom\'{e}, Aneesh Chauhan, ``A Perceptual Memory System for Grounding Semantic Representation in Intelligent Service Robot''. 2014 IEEE/RSJ International Conference on Intelligent Robots and Systems (IROS 2014), pp. 2216-2223.

		\item Lim, Gi Hyun, Miguel Oliveira, Vahid Mokhtari, \textbf{S. Hamidreza Kasaei}, Lu{\'i}s Seabra Lopes, Ana Maria Tom\'{e}, Aneesh Chauhan,``Interactive Teaching and Experience Extraction for Learning about Objects and Robot Activities''. 2014 The 23rd IEEE International Symposium on Robot and Human Interactive Communication (RO-MAN 2014), pp. 153-160.

	    \item Krishna S.R. Dubba, Miguel Oliveira, Gi Hyun Lim, \textbf{S.Hamidreza Kasaei}, Lu{\'i}s Seabra Lopes, Ana Maria Tom\'{e}, Anthony G. Cohn and David C. Hogg, ``Grounding Language in Perception for Scene Conceptualization in Autonomous Robots'', Proceedings of Artificial Intelligence spring symposium on qualitative representations for robots, (AAAI 2014), AI Access Foundation, pp. 26-33.

		\item \textbf{\small {Kasaei, S. Hamidreza}}, et al. ``On-Line Evaluation of Open-Ended Object Recognition System'', Proceedings of the 20$^{th}$ Portuguese Conference on Pattern Recognition (RecPad2014), Covilha, Portugal, 2014.

		\end{enumerate}	
			\item \textbf{Workshops}
		\begin{enumerate}
		\item Sock  Juil, \textbf{\small {Kasaei, S. Hamidreza}}, et al. ``Multi View 6D Object Pose Estimation and Camera Motion Planning using RGBD Images''. 2017 IEEE International Conference on Computer Vision Workshop (ICCV). 2017.
		
		\item \textbf{\small {Kasaei, S. Hamidreza}}, et al. ``An Object Perception Framework for Open-Ended Object Conceptualization from Experiences''. In
NIPS, workshop of Continual Learning and Deep Networks, Barcelona, Spain, 2016.

		\item \textbf{\small {Kasaei, S. Hamidreza}}, et al. ``An Instance-Based
Approach to 3D Object Recognition in Open-Ended Robotic Domains''. In
Robotics: Science and Systems (RSS), workshop of RGB-D: Advanced Reasoning with Depth Cameras, Berlin, Germany, 2013.	
		\end{enumerate}	
\end{itemize}
%^^^^^^^^^^^^^^^^^^^^^^^^^^^^^^^^^^^^^^^^^^^^^^^^^^^^^^^^^^^^^^^^
\section {Thesis Outline}
\label{thesis_outline_and_contributions}

This thesis is structured in nine chapters. The first being this introduction. In {Chapter~\ref{chapter_2}}, we focus on the development of a 3D object perception and perceptual learning architectures designed for complex artificial cognitive agents. 
{Chapter~\ref{chapter_3}} is dedicated to the gathering object experiences in both supervised and unsupervised manner. In particular, we propose automatic perception capabilities that will allow robots to automatically detect multiple objects in a crowded scene. {Chapter~\ref{chapter_4}} is devoted to object representations. We present a new object descriptor named Global Orthographic Object Descriptor (GOOD), designed to be robust, descriptive and efficient to compute and use. Furthermore, we propose an extension of Latent Dirichlet Allocation to learn structural semantic features (i.e., topics) from low-level feature co-occurrences for each object category independently. Although the model we developed has been intended to be used for object category learning and recognition, it is a novel probabilistic model that can be used in the fields of computer vision and machine learning.

%An overview of the developed system is presented in \cblue{Chapter~2}. Then the next chapters present the main components of the system including an overview about the work related to the respective component, the actual method, and a discussion about results and experiments.  

In {Chapter~\ref{chapter_5}}, we approach object category learning and recognition from a long-term perspective and with emphasis on open-endedness, i.e., not assuming a pre-defined set of categories. {Chapter~\ref{chapter_6}} is dedicated to classical evaluation of all the proposed representation, learning and recognition approaches. To examine the performance of the proposed approaches, several sets of 10-fold cross validation experiments were carried out. {Chapter~\ref{chapter_7}} begins by a discussion about an open-ended evaluation protocols. A novel protocol for evaluating open-ended learning approaches in multi-context scenarios is then proposed. Afterwards, we report and discuss the results of the open-ended experiments that were carried out in both classic single context and multi-context settings. Profiling and demonstration of the developed system is the topic of {Chapter~\ref{chapter_8}}. Finally, in {Chapter~\ref{chapter_9}}, the conclusions are presented and future research directions are discussed.

\cleardoublepage
\chapter{Architecture of Object Perception and Perceptual Learning System}
\label{chapter_2}
This chapter proposes two cognitive architectures designed to create a proper coupling between perception and action for service robots. This is necessary for service robots, not only to perform manipulation tasks in a reasonable amount of time and in an appropriate manner, but also to robustly adapt to new environments by handling new objects. In particular, these cognitive architectures provide perception capabilities that will allow robots to, incrementally learn object categories from the set of accumulated experiences and reason about how to perform complex tasks. To achieve these goals, it is critical to detect, track and recognize objects in the environment as well as to conceptualize experiences and learn novel object categories in an open-ended manner, based on human-robot interaction. Therefore, the following key aspects will have to be taken into consideration:

\begin{itemize}
	\item \textbf{Perception} is used to perceive the world. An agent may sense the world through different modalities. The perception system provides important information that the robot has to use for interacting with users and environments. For instance, a robot needs to know which kinds of objects exist in a scene and where they are, to interact with users and environment.
	
	\item \textbf{Memory} is used to store content about the agent's beliefs, goals, and knowledge; Learning is closely related to memory in human cognition. In the cognitive science literature, the existence of multiple memory systems is widely accepted. Most recent literature converges on five major memory systems \citep{Tulving2005, Tulving1991}: \emph{procedural memory}, for sensory-motor skills; \emph{perceptual representation memory}, mainly for the identification of objects; \emph{working memory}, to support basic cognitive activity; \emph{semantic memory}, mainly for spatial and relational information; and \emph{episodic memory}, for specific past happenings, enabling ``mental time travel''.
	
	\item \textbf{Interaction and Communication} is one of the effective way for an agent to obtain knowledge from a human user/teacher. Therefore, 
	a communication interface that facilitates language transfer from a human user to the robotic agent is another important aspect that a cognitive architecture should support \citep{langley2009cognitive}. Interaction capabilities are mainly developed to enable human users to teach new object categories and instruct the robot to perform complex tasks.
	
	\item \textbf{Learning} mechanisms that allow incremental and open-ended learning. 
	A cognitive robot must update its models over time with limited computational resources. Moreover, open-ended systems should involve experience management to prevent the accumulation of examples. Otherwise, the memory consumption and the required time to both update the models and recognize new objects would increase exponentially.
	
	\item \textbf{Recognition and Categorization} is essential for a robotic agent to be considered intelligent \citep{langley2009cognitive}. It must have capabilities to make decisions and select among alternatives. Recognition is closely related to categorization, which involves the assignment of objects to known concepts or categories. An ideal cognitive architecture should incorporate some way to improve its decisions through learning.
	
\item \textbf{Planning and Execution}: a cognitive architecture must be able to generate plans and solve problems to achieve adaptability in
novel situations. Furthermore, it must be able to execute skills and actions in the environment. In some frameworks, this happens in a completely reactive manner, with the agent selecting one or more primitive actions on each decision cycle, executing them, and repeating the process on the next cycle. This approach is associated with closed-loop strategies for execution, since the agent can also sense the environment on each time step \citep{langley2009cognitive}.   
\end{itemize}	

\begin{figure}[!t]
\centering
\begin{centering}
\begin{tabular}{cc}
\hspace{-3mm}
	 \includegraphics[width=0.465\linewidth, trim=0cm 0cm 0cm 0cm, clip=true]{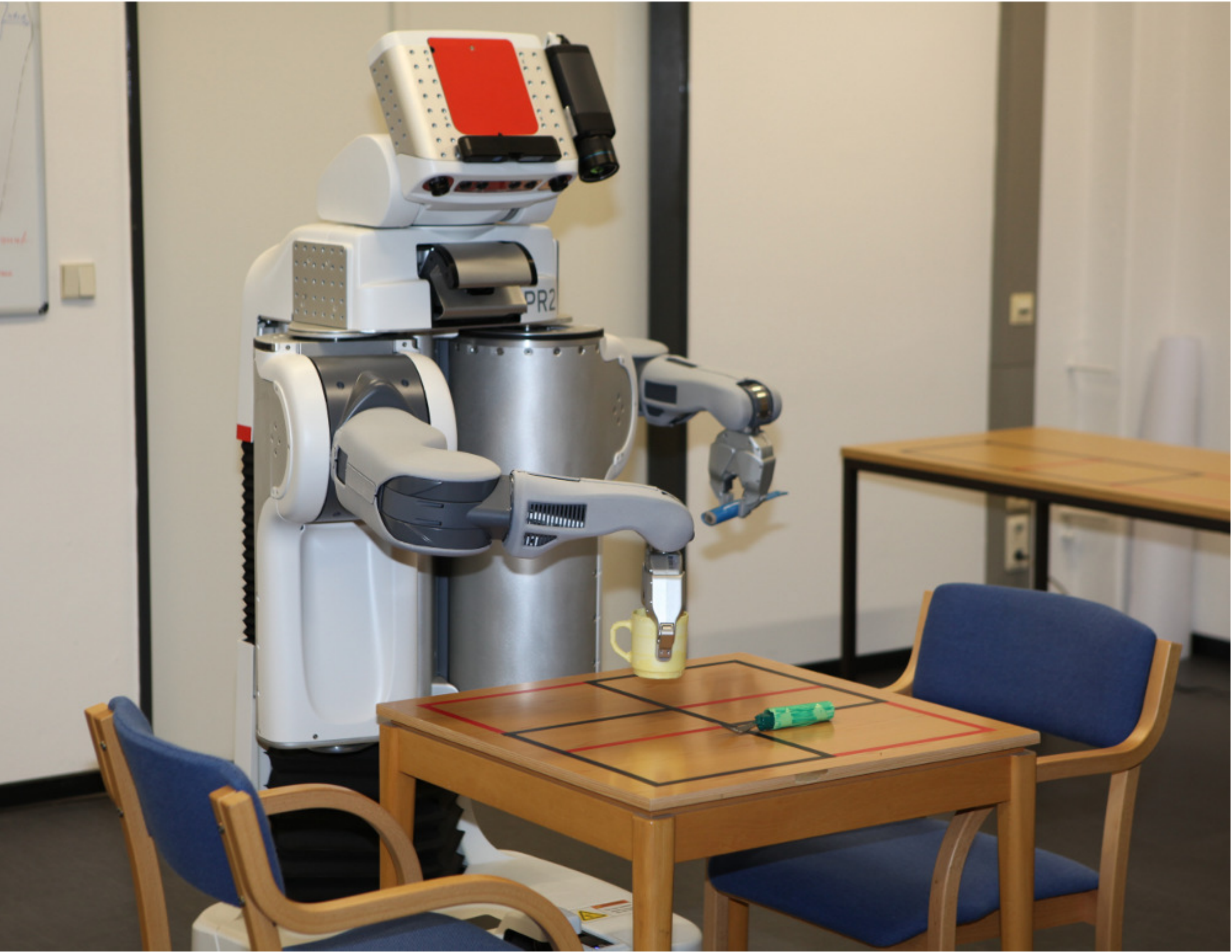}&
	\includegraphics[width=0.515\linewidth, trim=0cm 0.25cm 0cm 0cm, clip=true]{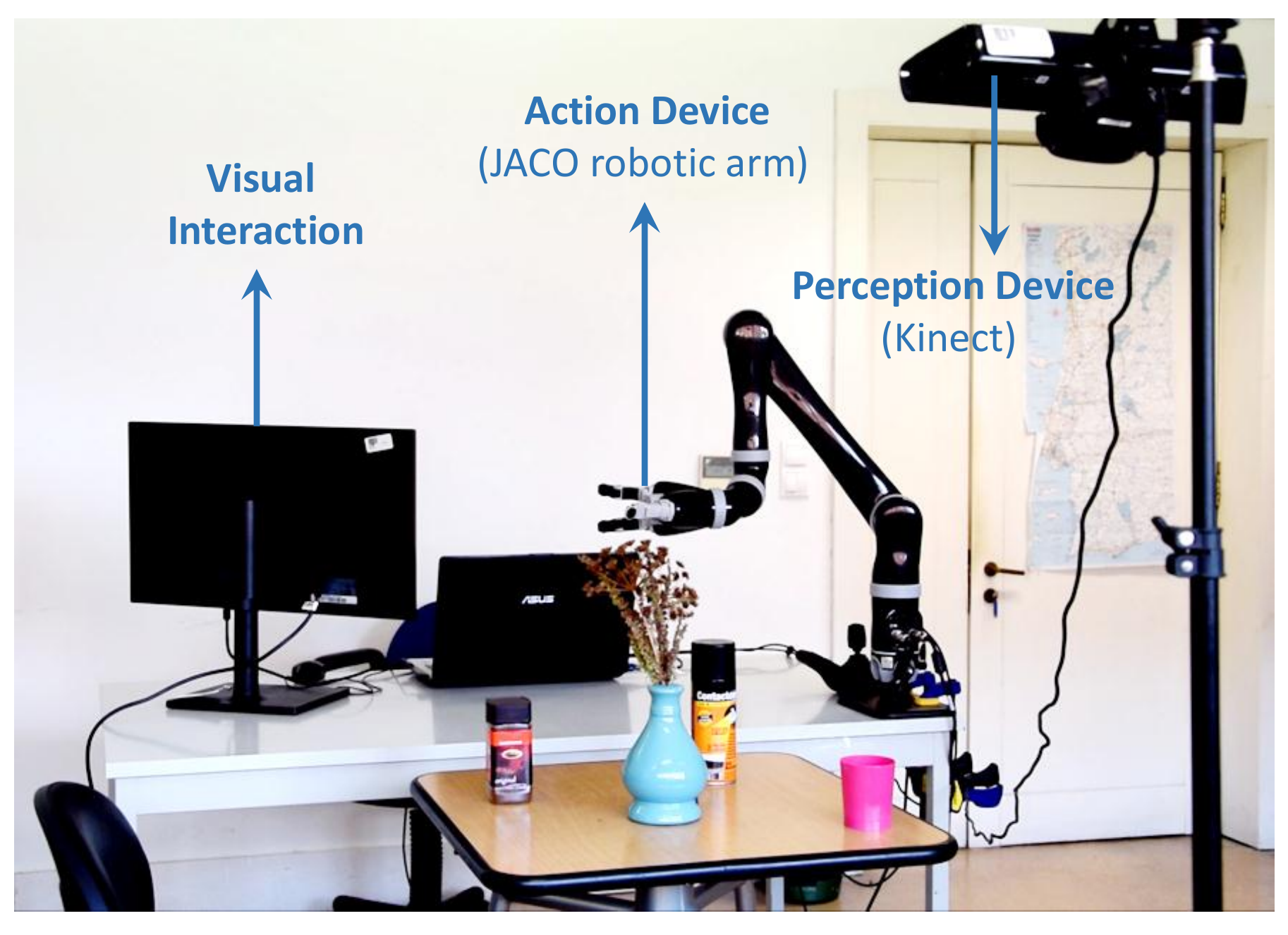}\\
\end{tabular}
\vspace{-2mm}
\end{centering}
\caption{The proposed system being tested on two robotic platforms: \emph{(left)} a PR2 service robot used by the RACE project; \emph{(right)} the physical architecture of the platform provided by the University of Aveiro includes a Kinect and a JACO robotic-arm as the primary sensory-motor embodiments for perceiving and acting upon its environment. }
\label{fig:platforms}
\vspace{-10pt}
\end{figure}

This kind of system must comprise a significant number of software modules, which must be closely coupled in their structure and functionality \citep{Jeong2012}. Three main design options address the key computational issues involved in processing and storing perception data. First, a lightweight, NoSQL database, is used to implement the perceptual memory. Second, a threadbased approach with zero copy transport of messages is used in implementing the modules. Finally, a multiplexing scheme, for the processing of the different objects in the scene, enables parallelization. This way, the system is capable of real time object detection, tracking and recognition. The developed perception and perceptual learning capabilities target objects in table-top scenes, e.g. in a restaurant environment. These capabilities are fully integrated in both cognitive architectures and are running on the PR2 robot used by the RACE project \citep{Hertzberg2014projrep}, as depicted in Fig.~\ref{fig:platforms} \emph{(left)}, and on a robotic-arm platform provided by the University of Aveiro as shown in Fig.~\ref{fig:platforms} \emph{(right)}. It is worth mentioning that two colleagues have contributed to the part of the work in this thesis. Miguel Oliveira contributed to the development of the gesture recognizer module, an extended version of which became part of this thesis. Gi Hyun Lim contributed to the development of software for the memory system. Most material in this chapter has already been published in \citep{Hertzberg2014projrep, Kasaei2014, oliveira20153d, kasaei2016object, shafii2016learning,kasaei2017Neurocomputing}.

The remaining part of this chapter is organized as follows. In the next section, related works are discussed. Memory and cognitive architecture issues are discussed  in Section~\ref{dual_memory}, leading to the choice of a dual memory approach and to the development of a perceptual memory system. Two 3D object perception architectures and their computational issues are then discussed. Finally, in Section~\ref{summary_chapter2}, the summary is presented.

%^^^^^^^^^^^^^^^^^^^^^^^^^^^^^^^^^^^^^^^^^^^^^^^^^^^^^^^^^^^^^^^^
%^^^^^^^^^^^^^^^^^^^^^^^^^^^^^^^^^^^^^^^^^^^^^^^^^^^^^^^^^^^^^^^^
\section {Related Work}
\label{related_work_chapter2}

Although an exhaustive survey of service robotics as well as object perception techniques is beyond the scope of this chapter, representative works will be reviewed in this section.

As robots are expected to increasingly interact and collaborate closely with humans, robotics researchers need to look at human cognition as a source of inspiration. Learning is closely related to memory in human cognition.  Biological findings about memory and learning have served as inspiration for the development of computational models and applications. \cite{Wood2011} present a thorough review and discussion on memory systems in animals as well as artificial agents, having in mind further developments in artificial intelligence and cognitive science.

Over the past decade, several researches have been conducted to develop cognitive architectures for autonomous robots. Most of the state-of-the-art cognitive architectures like SOAR \citep{laird2012soar}, DIARC \citep{scheutz2013novel} and ACT-R \citep{anderson1997act} use classical object category learning and recognition approaches (i.e., offline training and online testing are two separated phases), where open-ended object category learning is generally ignored \citep {leroux2013armen}. Therefore, they work well for specific tasks where there are limited and predictable numbers of objects and fail at any other assignment. Unlike our approach, the perceptual knowledge of these cognitive architectures are static, in the sense that the representation of the known categories does not change after the training stage. Therefore, these robots are unable to adapt to dynamic environments \citep{Jeong2012}\citep{laird2012cognitive}. This leads to several shortcomings such as the inability to detect/recognize new or unknown categories.  To cope with these issues, several cognitive robotics groups have started to explore how robots could learn incrementally  from their own experiences as well as from interaction with humans.

In the ARMEN project, Leroux et al. proposed a mobile assistive robotics approach providing advanced functions to help maintaining elderly or disabled people at home \citep {leroux2013armen}. Similar to our system, this project involves object manipulation, knowledge representation and object recognition. The authors also developed an interface to facilitate the communication between the user and the robot. Jain et al. presented an assistive mobile manipulator named EL-E that can autonomously pick objects from a flat surface and deliver them to the users \citep{jain2010assistive}. They used a multi-step control policy that is not suitable to achieve real time performance. In our approach, we can achieve real-time performance through the use of ROS nodelets and multiplexing mechanisms. Furthermore, in \citep{jain2010assistive}, the user provides the location of the object to be grasped to the robot by briefly illuminating a location with a laser pointer. In our system, objects are detected and recognized autonomously. Therefore it is enough for the user to specify the category of the object to be picked up.

In another work \citep{srinivasa2008robotic}, a multi-robot assistive system, consisting of a Segway mobile robot with a tray and a stationary Barrett WAM robotic arm, was developed. The Segway robot navigates through the environment and collects empty mugs from people. 
Then, it delivers the mugs to a predefined position near the Barrett arm.  Afterwards, the arm detects and manipulates the mugs from the tray and loads them into a dishwasher rack. This work is similar to ours in that it integrates perception and motion planning for pick and place operations. However there are some differences: their
vision system is designed for detecting a single object type (mugs), while our perception system not only tracks the pose of different types of objects but also recognizes their categories. Furthermore, because there is a single object type (i. e. mug), they computed the set of grasp points off-line. In our approach, grasping must handle a variety of objects never seen before.  

In the RACE project (Robustness by Autonomous Competence Enhancement), a PR2 robot demonstrated effective capabilities in a restaurant scenario including the ability to serve a coffee, set a table for a meal and clear a table \citep{Hertzberg2014projrep} \citep{Rockel20132}. The aim of RACE was to develop a cognitive system, embodied by a
service robot, which enabled the robot to build a high-level understanding of the world by storing and exploiting appropriate memories of its experiences. Other examples of assistive robot platforms that have demonstrated coupling perception and action include TUM Rosie robot \citep{beetz2011robotic}, HERB \citep{srinivasa2010herb} and ARMAR-III \citep{vahrenkamp2010integrated}. 

Willow Garage developed the \acf{ORK}\footnote{http://wg-perception.github.io/object\_recognition\_core}, a 3D object recognition system built on top of the Ecto framework\footnote{http://plasmodic.github.io/ecto/}. Ecto organizes computation as a directed acyclic graph, which implies important limitations in the architecture of the perception system. Moreover, in ORK, training/learning and detection/recognition are two separate stages. Such approach is not suitable for developing open-ended learning agents. In contrast, our system allows for concurrent or interleaved learning and recognition, and real-time performance is achieved through nodelets and multiplexing.

\section {The Dual Memory System Approach of RACE}
\label{dual_memory}
Arguably, robots that interact closely with non-expert users should 
be: \emph{animate}, meaning that they react appropriately to different events, based on a tight coupling of perception and action; \emph{adaptive}, to cope with changing users, tasks and environments, which requires reasoning and learning capabilities; and \emph{accessible}, that is, they should be easy to command and instruct, and they should also be able to explain their beliefs, motivations and intentions~\citep{Lopes2001}.

\begin{figure}[!b]
\centering
\begin{centering}
\begin{tabular}{cc}
	 \includegraphics[width=0.41\linewidth, trim=0cm 0cm 0cm 0cm, clip=true]{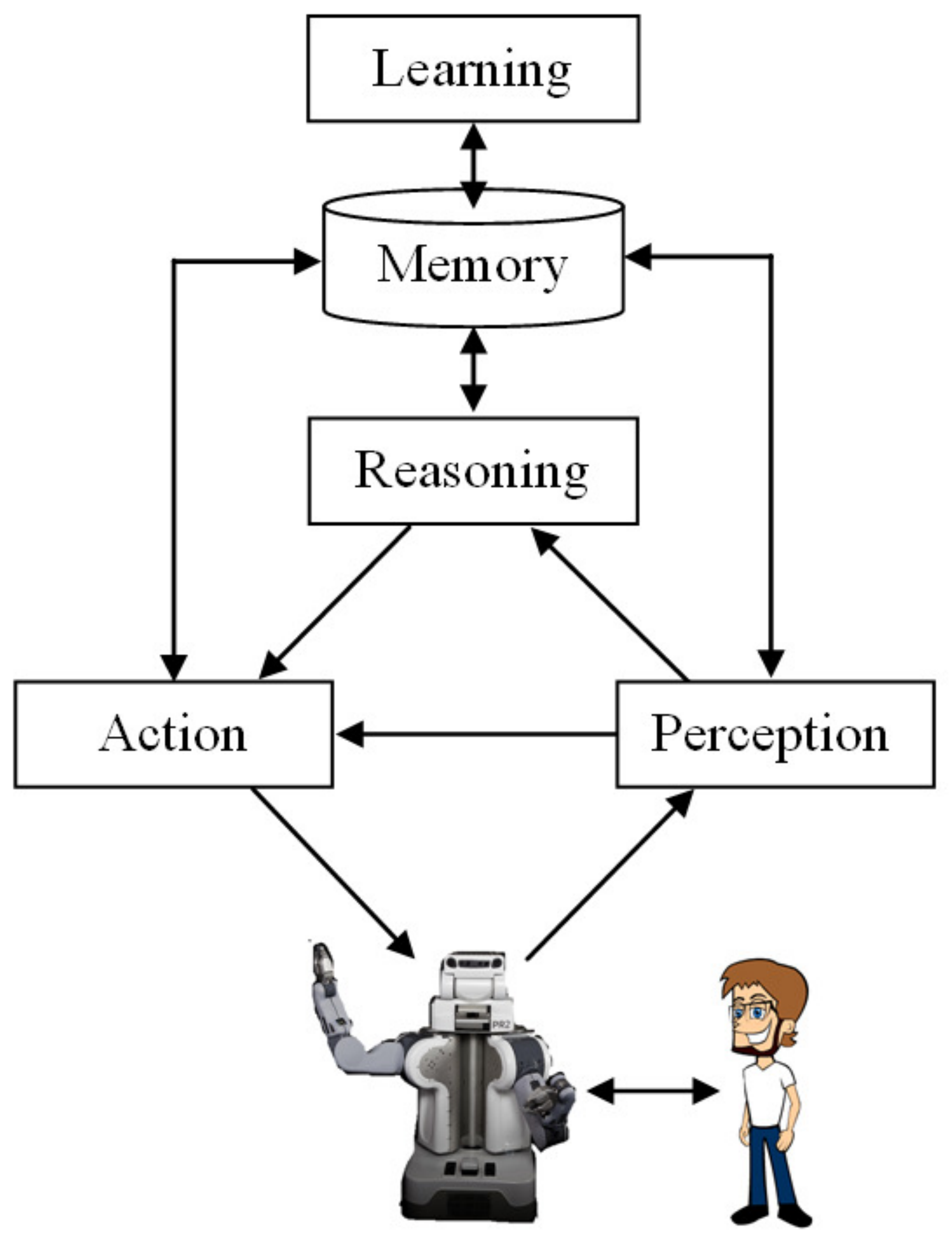}&\quad\quad\quad\quad\quad
	\includegraphics[width=0.35\linewidth, trim=0cm 0cm 0cm 0cm, clip=true]{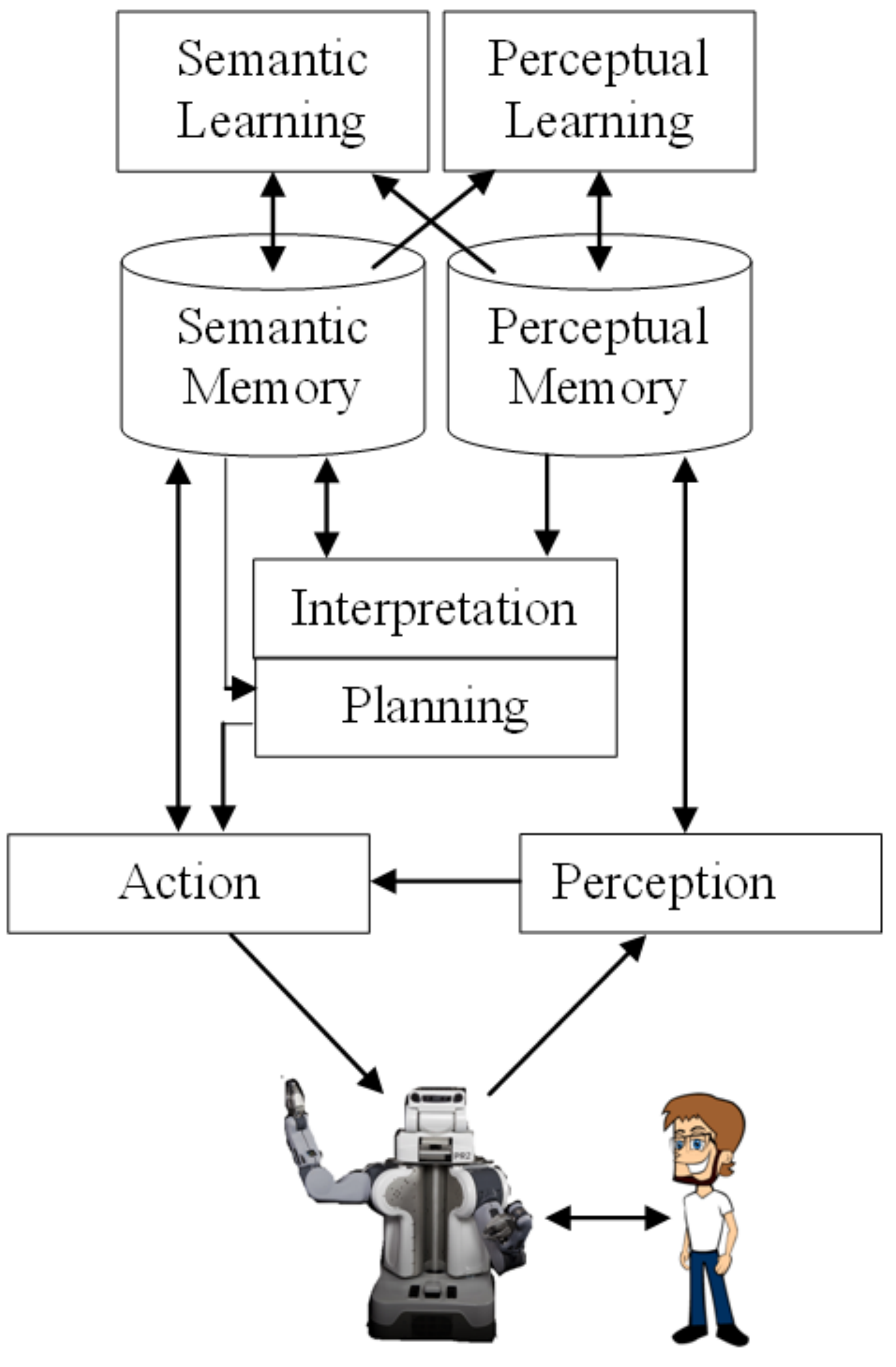}\\
\end{tabular}
\end{centering}
\caption{Abstract cognitive architectures for hybrid reactive-deliberative 
robots: (\emph{left}) with a single memory system; (\emph{right}) with a dual memory system.}
\label{fig:plot_luis}
\vspace{-10pt}
\end{figure}

In an abstract architecture for intelligent robots, as shown in Fig.~\ref{fig:plot_luis} (\emph{left}), a \emph{Perception} component processes all momentary information coming from sensors, including sensors that capture the actions and utterances of the user. A \emph{Reasoning} component updates the world model and determines plans to achieve goals. An \emph{Action} component reactively dispatches and monitors the execution of actions, taking into account the current plans and goals. Action processing ranges from low-level control to high-level execution management. Finally, a \emph{Learning} component, which typically runs in the \emph{background}, analyzes the trace of \emph{foreground} activities recorded in a \emph{Memory} component and extracts and conceptualizes possibly interesting \emph{experiences}. The resulting \emph{conceptualizations} are stored back in memory. Each component in such abstract architecture decomposes into a set of software modules, possibly distributed across multiple computers.

The reasoning component manipulates primarily semantic representations of the current world state, goals and plans, that is, representations that are symbolic and relational in nature. In RACE, where case studies were carried out in a restaurant environment, semantic representations describe tables, chairs, table-top objects, guests, the robot, etc., the categories of these objects, the relations between them, and the actions and events that change these relations. The action component includes multiple modules that control the robot actuators based on a tight coupling with perception. In addition, the action component carries out high-level execution management, which consists of reactively dispatching and monitoring the execution of actions, taking into account the current plans and goals. Like reasoning, execution management primarily manipulates semantic representations. The semantic information flowing between reasoning, execution management and memory is typically of small size, and its processing tends to be slow \citep{RACE2013}.

One of the challenges in a project like RACE was to combine and store semantic and perceptual representations. Standard \emph{SQL} databases do not cope well neither with semantic data nor with perception data, as both tend to be partially unstructured and/or of variable size. This suggests that modern \emph{NoSQL} databases \citep{sahib2013} should be used. Semantic data represents the world in terms of instances, categories and relations between them. A semantic representation of the state of the world can be simply a set of \emph{subject-predicate-object triples}. A special kind of database, the \emph{triplestore}, which shares some features with both \emph{SQL} and graph databases, is especially optimized to store information in the form of a set of triples. Triplestores are clearly one of the database types to take into account when developing memory systems for robots. An RDF triplestore was in fact the choice for the initial memory component in the RACE architecture \citep{RACE2013}.  The contents of this memory system, which is used as \emph{blackboard} for all processes, is semantic in nature. It keeps track of the evolution of both the internal state of the robot and the events observed in the environment.

Access to the triplestore is granted via a \ac{ROS} node that provides database query and write services for all other nodes (an interface node). Information exchange is performed using either publisher / subscriber or client / server mechanisms. 
\ac{ROS} communications are a robust framework \citep{Zaman2013}. 
However, when the size of the messages is large (e.g., when passing 3D point clouds), the communication between processes is slow.  
In the case of perception related data, its large size implies large \ac{ROS} messages to be passed between the database interface node and the other nodes. This is a major constraint, especially considering that, unlike semantic data, perceptual data flows
continuously at the sensor output frequency. Using a database interface node creates a bottleneck for accessing the database, since it handles access requests in a first in, first out basis.

Moreover, although triplestores are well suited for storing semantic information, they can hardly be considered suited for storing perception data. In fact, the perception modules will primarily process numeric information organized in structures like vectors and matrices, possibly grouped in sets. For instance the raw perception data about an object, after detection, can be a 3D point cloud, which is a set of points described by their 3D coordinates and possibly RGB information. Based on the
point cloud, shape features can be extracted, and the object can be represented by a set of local shape features, where each of them can be a 2D shape histogram. To ensure timely reaction to events in the environment, perception modules run continuously at the frame rate of the used sensors. Although raw data tends to be massive (high-dimensional), the perception modules must run fast, and whatever memory support they use, must also be lightweight.

In the context of RACE, to accommodate semantic and perceptual information in the same database, the only option would be to replace the triplestore with a more generic kind of database. However, we would loose the special features of triplestores, which are optimized for storing triples. In alternative, two different databases can be used, one for semantic information, and the other for perceptual information. The second alternative, which seems more promissing, allows to use databases that are well suited for the kinds of data that each will store. In RACE, we converged to the second option \citep{Hertzberg2014projrep, dubba2014grounding, oliveira20153d}.

Figure~\ref{fig:plot_luis} (\emph{right}) shows an abstract architecture diagram in which we make explicit the dual memory approach. In what concerns reasoning, we make explicit both interpretation and planning capabilities.  One of the most basic interpretation capabilities is anchoring, i.e., connecting object symbols used in the semantic memory to the perception of those objects that is recorded in the perceptual memory. Interpretation also includes computing spatial relations between objects to keep an updated relational model of the scene around the robot. In turn, this scene model can be taken into account for anchoring.

The perceptual memory contains, not only object perception data, but also object category knowledge, in the form of perceptual categories that enable to recognize instances of those categories. These perceptual categories are learned in an open-ended fashion with user mediation \citep{chauhan2013towards, lim2014interactive}. The perceptual learning 
component primarily uses data from perceptual memory (e.g. shape features of objects) 
as well as from the semantic memory (e.g. teaching instructions from the user). In RACE, the implementation of the perceptual memory  was carried out using  a flexible and scalable \emph{NoSQL} database which operates in memory (see the next section for details).

It is worth emphasizing that, although our design choices were guided primarily by engineering criteria, we converged to a solution that is biologically and cognitively plausible. In fact, as previously pointed out, human memory is not a single monolithic system, but rather a combination of several memory subsystems specialized for storing different types of information and supporting different functionalities \citep{Tulving2005,Tulving1991}. In particular, our perceptual memory resembles the so-called \emph{Perceptual Representation Memory System}, used in human cognition for enhancing the identification of objects as structured physical-perceptual entities, a process referred to as \emph{perceptual priming} \citep{Tulving1991}. Another key distinction in cognitive science is between processes that are fast, automatic and unconscious, and processes that are slow, deliberative and conscious \citep{Evans2008}. Our dual semantic/perceptual memory approach is also in line with these findings.

In the following sections, we focus on extending the perceptual capabilities of the initial RACE architecture~\citep{Hertzberg2014projrep}. In particular, two different system architectures are presented as depicted in Fig.~\ref{fig:perceptionsystem} and Fig.~\ref{fig:OverallSystemArchitecture}. Both architectures are reusable frameworks and developed over ROS \citep{quigley2009ros}. Each software module is organized into a \ac{ROS} package and will typically correspond to a node or a nodelet\footnote{http://wiki.ros.org/nodelet} at runtime. Information exchange is performed using standard ROS mechanisms (i.e., either publish / subscribe or server / client). Therefore, any new module can be easily added to the system. Each of these architectures will be discussed in detail in the following sections. 

%\section{Overall System Architecture}
%\label{overall_system_architecture}
\section{ RACE Perception System}
\label{RACE_overall_system_architecture}

\begin{figure}[!t]
 \centering
 \begin{centering}
 \begin{tabular}{c}
 \includegraphics[width=0.95\linewidth, trim=0cm 0cm 0.0cm 0.0cm,clip=true]{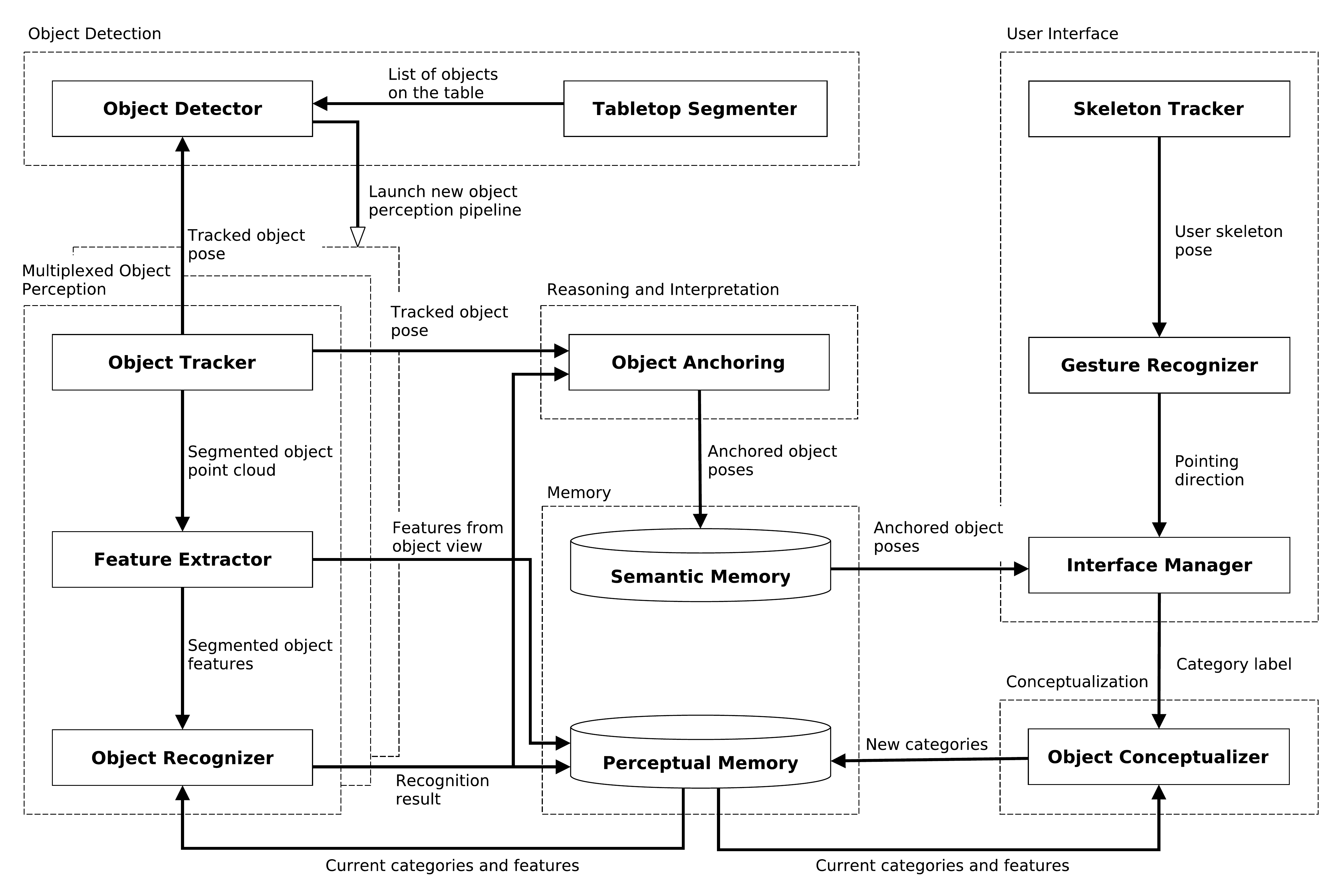}\\
 \end{tabular}
 \end{centering}
  \vspace{-10pt}
 \caption{Architecture of the developed object perception and perceptual learning system (RACE).}
 \label{fig:perceptionsystem}
 \vspace{-10pt}
 \end{figure}

The overall RACE perception and perceptual learning system is depicted in Fig.~\ref{fig:perceptionsystem}.
From a global perspective, the RACE perception system is composed of six functional components: \emph{Object Detection}, \emph{Multiplexed Object Perception},  \emph{User Interface}, \emph{Reasoning and Interpretation}, \emph{Memory} and \emph{Conceptualization}. 
The implementation of the \emph{Perceptual Memory} was carried out using \emph{LevelDB}, a lightweight, flexible and scalable \emph{NoSQL} database developed by \emph{Google}\footnote{https://code.google.com/p/leveldb/}. \emph{LevelDB} is a key-value storage database that provides an ordered mapping from string keys to string values. In addition, \emph{LevelDB} operates in memory and is copied to the file system  asynchronously. This significantly improves its access speed.

\begin{wrapfigure}{r}{0.5\textwidth}
\vspace{-5mm}
  \begin{center}
   \includegraphics[width=0.9\linewidth]{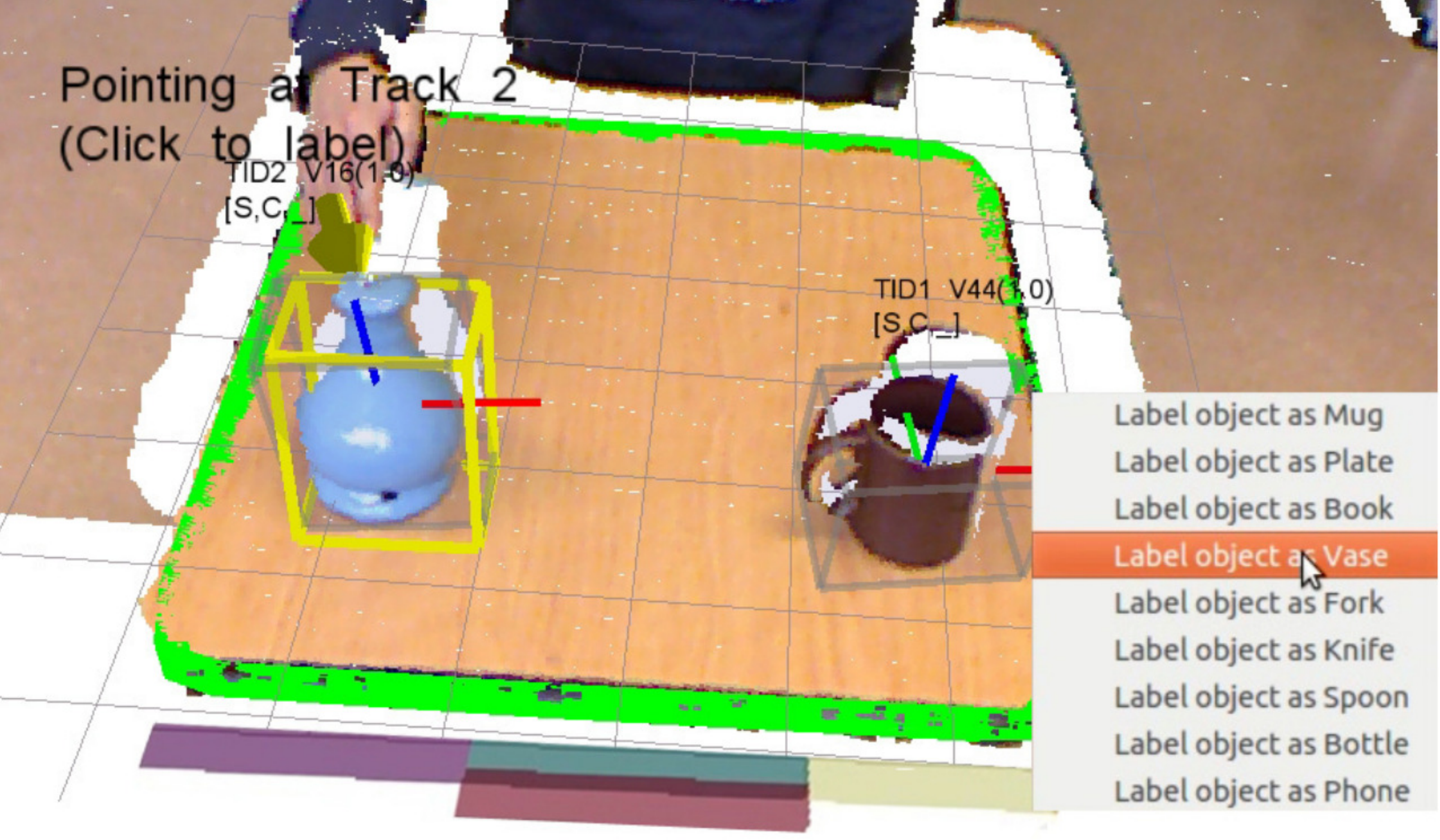}
  \end{center} 
  \caption{Visualization of the tracking, pointing, and user labelling graphical interfaces.}
  \label{fig:detection_tracking}
\end{wrapfigure}

In the RACE system, an RGB-D sensor is used for the perception of both the user and the table-top scene. The starting point for the perception of the table-top scene is \emph{Table-Top Segmentation (TTS)}\footnote{http://wiki.ros.org/tabletop\_object\_detector}, which uses a hierarchical clustering procedure to isolate (partial) point clouds of the objects. The \emph{Object Detection (OD)} module periodically requests the current list of objects from  \emph{TTS}. Then,  \emph{OD} will check if any of those objects is already being tracked. To do this,  \emph{OD} matches the point clouds of all objects on the table with the bounding boxes of all objects currently being tracked. The percentages of points of the tabletop objects that lie inside the bounding boxes of the tracked objects are computed. A large percentage indicates that the tracked object and the segmented object are the same. Point clouds that cannot be matched with any of the tracked bounding boxes are assumed to represent new objects just added to the scene. \emph{OD} will assign a new identifier (\emph{track-id}) to
each newly detected object. Also for each new object,  \emph{OD} will launch an object perception pipeline which contains three modules: \emph{Object Tracking}, \emph{Feature Extraction} and \emph{Object Recognition}. Figure \ref{fig:detection_tracking} shows a situation where two objects are segmented and tracked, i.e., they have bounding boxes around them.

 \emph{Object Tracking (OT)} is responsible for keeping track of the target object over time while it remains visible. Tracking is an essential base for anchoring. On initialization,  \emph{OT} receives the point cloud of the detected object and computes a bounding box for that point cloud, the centre of which defines the pose of the object. A particle filter approach is then used\footnote{http://www.willowgarage.com/blog/2012/01/17/tracking-3d-objects-point-cloud-library} to predict the next probable pose of the object. In each cycle,  \emph{OT} sends out the tracked pose of the object both to \emph{OD} and to the Interpretation component. At a lower rate,  \emph{OT} sends the point cloud of the object (i.e., containing the points inside the predicted bounding box) to Feature Extraction.

The \emph{Feature Extraction (FE)} module computes and stores object representations in the perceptual memory. Objects are represented by sets of local shape features computed in certain keypoints. In addition to storing object representations in the perceptual memory,  \emph{FE} also sends them to  \emph{Object Recognition (OR)}. The perceptual categories learned so far and stored in the perceptual memory are used by  \emph{OR} to predict the category of the target object.  \emph{OR} is a low frequency module, which runs at 1 Hz. Accordingly,  \emph{FE} receives object point clouds from  \emph{OT} and sends the extracted representations for recognition at the same frequency. Thus, only  \emph{OT} itself uses object point clouds at the frame rate of the sensor (30 Hz).

Object recognition results are written to the perceptual memory, where the Interpretation component can fetch them to support symbol anchoring. 
Anchoring involves keeping track of objects even when they cannot be visually tracked. Suppose that an object with \emph{track-id}=7 disappears from the visible the scene. Then, after some time, an object becomes visible in the same location and starts being tracked as \emph{track-id}=8. In such case, high-level interpretation may infer that both identifiers refer to the same object, so it will associate both to the same object symbol in the semantic memory. The current implementation is capable of  anchoring symbols that refer to objects only while these remain visible. However, further work in the context of RACE project has been done to enable anchoring object symbols when the visual tracking is lost, but this functionality is out of the scope of this thesis, and will not be further discussed.

In the RACE perception architecture, pointing gesture recognition and verbal teaching of object categories support open-ended category learning. Verbal input is currently provided through interactive markers in \emph{RVIZ}, a 3D visualization tool for ROS. The  \emph{Skeleton Tracker (ST)} module tracks the user skeleton pose over time based on RGB-D data\footnote{http://wiki.ros.org/openni\_tracker}. The skeleton pose information is passed to the  \emph{Gesture Recognition (GR)} module, which computes a pointing direction. Currently, the pointing direction is assumed to be the direction of the right forearm (see an example in Fig.~\ref{fig:detection_tracking}). The pointing direction is then passed to the Interpretation component. Upon receiving verbal input, the \emph{Interpretation} component checks if the received pointing direction intersects the bounding box of any of the objects currently on the table according to the world state recorded in the semantic memory. If that is the case, then a teaching instruction is recorded in the semantic memory, stating that the target object was taught to belong to the given category. Teaching instructions trigger perceptual learning to create and/or update object categories.

\section{Addressing Computational Issues}
\label{subsec:addressing}

In contrast with the reasoning processes supported by the semantic memory, the processes developed around the perceptual memory must run fast to cope with the continuous stream of massive sensor data. As pointed out, one of the reasons for using LevelDB to implement the memory systems is the fact that it operates in RAM. There is, however, the limitation that simultaneous access to LevelDB is only possible by threads within the same process. To comply with this constraint while keeping ROS as the framework for the newly developed modules, we use ROS nodelets\footnote{http://wiki.ros.org/nodelet}. Nodelets, which run as threads of a single process, were designed  to provide a way of concurrently running different modules with zero copy transport between publisher and subscriber calls (as an example, see \cite{Munaro2013}).
The motivation for ROS nodelets comes from systems with high throughput data flows as is common in perception systems. It is not surprising, therefore, that the developers of Point Cloud Library (PCL) and ROS nodelets are the same. In our system, in addition to handling high throughput data  flows, nodelets come handy to implement modules that need  to simultaneously access the perceptual memory (LevelDB).

Another way of optimizing perception is to parallelize computations. In our system, instead of tracking all objects in a single tracking module, there is a tracker for each object. Similar strategy is used for feature extraction and object recognition. In other words, object perception  is designed to be multiplexed. Every time a new object is detected, a corresponding instance of the \emph{object perception pipeline} (see Fig.~\ref{fig:perceptionsystem} and Fig.~\ref{fig:OverallSystemArchitecture}) is launched. Thus there are as many object perception pipelines as the number of currently tracked objects, and each pipeline targets a specific object. Since the modules in an object perception pipeline run as independent nodes/nodelets, they can be distributed to different CPU cores, thus improving the overall computational efficiency of perception. Note that the three modules in the object perception pipeline are traditionally amongst the heaviest in terms of computational requirements. The parallelization is aimed at the hotspot or bottleneck of the computation flow and takes full advantage of modern multi-core machines.  In fact, experiments with a non-multiplexed version of this architecture  show that it cannot run in real-time. 

We can easily configure the perception and perceptual learning modules to be launched with different runtime configurations, that is, using ROS nodelets only, ROS nodes only, or a combination of both.  By default, the object perception pipelines, the perceptual learning module and the perceptual memory run as a set of nodelets of a single process.  When debugging is necessary, we use a configuration where all
modules run as nodes. In this configuration, the modules access the perceptual memory using ROS services provided by a database interface.

\section{Coupling Object Perception with Manipulation}
\label{Neurocomputing_overall_system_architecture}

\begin{figure}[!t]
	\center
	\includegraphics[width=1\textwidth]{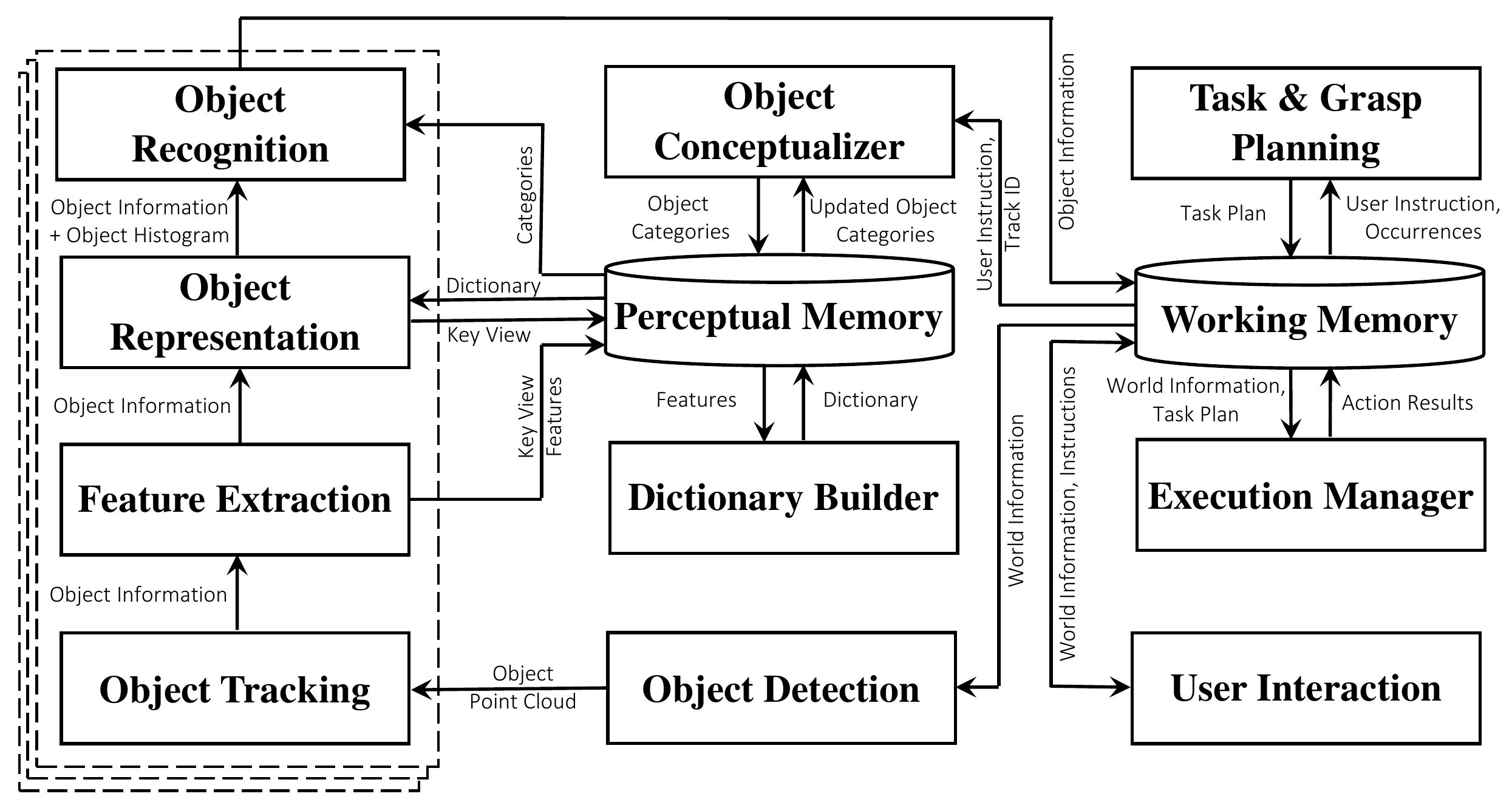}
	\vspace{-6mm}
	\caption{Overall architecture of the proposed system.}
	\label{fig:OverallSystemArchitecture}       % Give a unique label
\end{figure}

Figure \ref {fig:OverallSystemArchitecture} provides an illustration of a cognitive architecture designed to create a tight coupling between perception and manipulation for assistive robots \citep{kasaei2016object}. This architecture is an evolution of the RACE perception architecture described in the section \ref{RACE_overall_system_architecture} \citep{Kasaei2014, oliveira20153d}. Although both architectures support an interactive open-ended learning for 3D object category recognition, their complexity and performance differ depending on the characteristics of modules and the methods used for learning and classification. In the RACE architecture, we mainly use a local 3D shape descriptor for object representation and an instance-based object category learning, while in this architecture, we employ more advanced object representation techniques such as bag-of-words (BoW) and Latent Dirichlet Allocation (LDA), to represent objects in uniform and compact format which is suitable for both instance-based and model-based object category learning. Moreover, in this architecture a proper coupling between object perception and manipulation is provided. As mentioned before, object manipulation tasks consist of two phases: the first is the perception of the object and the second is the planning and execution of arm or body motions which grasp the object and carry out the manipulation task. These two phases are closely related: object perception provides information to update the model of the environment, while planning uses this world model information to generate sequences of arm movements and grasp actions for the robot \citep{kasaei2017Neurocomputing}. 

This cognitive architecture includes two memory systems, namely the \emph{Working Memory} and the \emph{Perceptual Memory}. Both memory systems have been implemented using LevelDB. The \emph{Working Memory} is used for temporarily storing information as well as for communication among different modules. It keeps track of the evolution of both the internal state of the robot and the events observed in the environment (i.e., world model). The object features, dictionary of visual words, object representation data and object category models are stored into \emph{Perceptual Memory}. The goal of \emph{Grasp Planning} is to extract a grasp pose (i.e., a gripper pose relative to the object) either from
above or from the side of the object, using global characteristics of the object. The \emph{Execution Manager} works based on a Finite-State-Machine (FSM) paradigm. It retrieves the task plan and the world model information from \emph{Working Memory} and computes the next action (i.e., a primitive operator) based on the current context. Then, it dispatches the action to the robot platform as well as records success or failure information in the \emph{Working Memory}.

Whenever the robot captures a scene, the first step is preprocessing which includes three filtering procedures, namely distance filtering, a filter to remove 
the robot's body from sensor data, and a downsampling filter for reducing the size of the data.
\emph{Object Detection}, responsible for detecting objects in the scene, launches a new perception pipeline for each detected object. Each pipeline includes \emph{Object Tracking}, \emph{Feature Extraction}, \emph{Object Representation} and \emph{Object Recognition} modules. The \emph{Object Tracking} module estimates the current pose of the object based on a particle filter, which uses shape and color data \citep {Oliveira2014}. 
The \emph{Feature Extraction} module extracts features of the current object view and stores them in the \emph{Perceptual Memory}. Based on the extracted features and on a visual dictionary, the \emph{Object Representation} module describes objects as histograms of visual words/topics and stores them into the \emph{Perceptual Memory}. \emph{User Interaction} is essential for supervised experience gathering. Interaction capabilities are developed to enable human users to teach new object categories and instruct the robot to perform complex tasks \citep{GiHyunLim}.
  
The proposed architecture, as shown in Fig.~\ref{fig:OverallSystemArchitecture}, includes two perceptual learning modules. One of them, the \emph{Dictionary Builder}, is concerned with building a dictionary of visual words for object representation. The dictionary plays a prominent role because it is used for category learning as well as recognition. The second learning module is the \emph{Object Conceptualizer}. Whenever the instructor provides a category label for an object, the \emph{Conceptualizer} retrieves the probabilistic models of the current object categories as well as the representation of the labeled object in order to improve an existing object category model or to create a new category model. In recognition situations, a probabilistic classification rule is used to assign a category label to the detected object. The system is run in two stages. The first stage is dedicated to environment exploration, which will be further discussed in the next chapter. In this stage, unsupervised object discovery is carried out in the environment while the robot operates. The robot seeks to segment the world into ``object'' and ``non-object''. 
Afterwards, a pool of shape features is created by computing local shape features for the extracted objects. The pool of features is then clustered by the \emph{Dictionary Builder} leading to a set of visual words (dictionary). Only the modules directly involved in object discovery and dictionary building are active in this stage. The second stage corresponds to the normal operation of the robot, with object category learning, recognition, planning and execution. In the following chapters, the characteristics of each module are explained in detail.

\begin{figure}[!b]
	\center
	\includegraphics[width=0.65\textwidth]{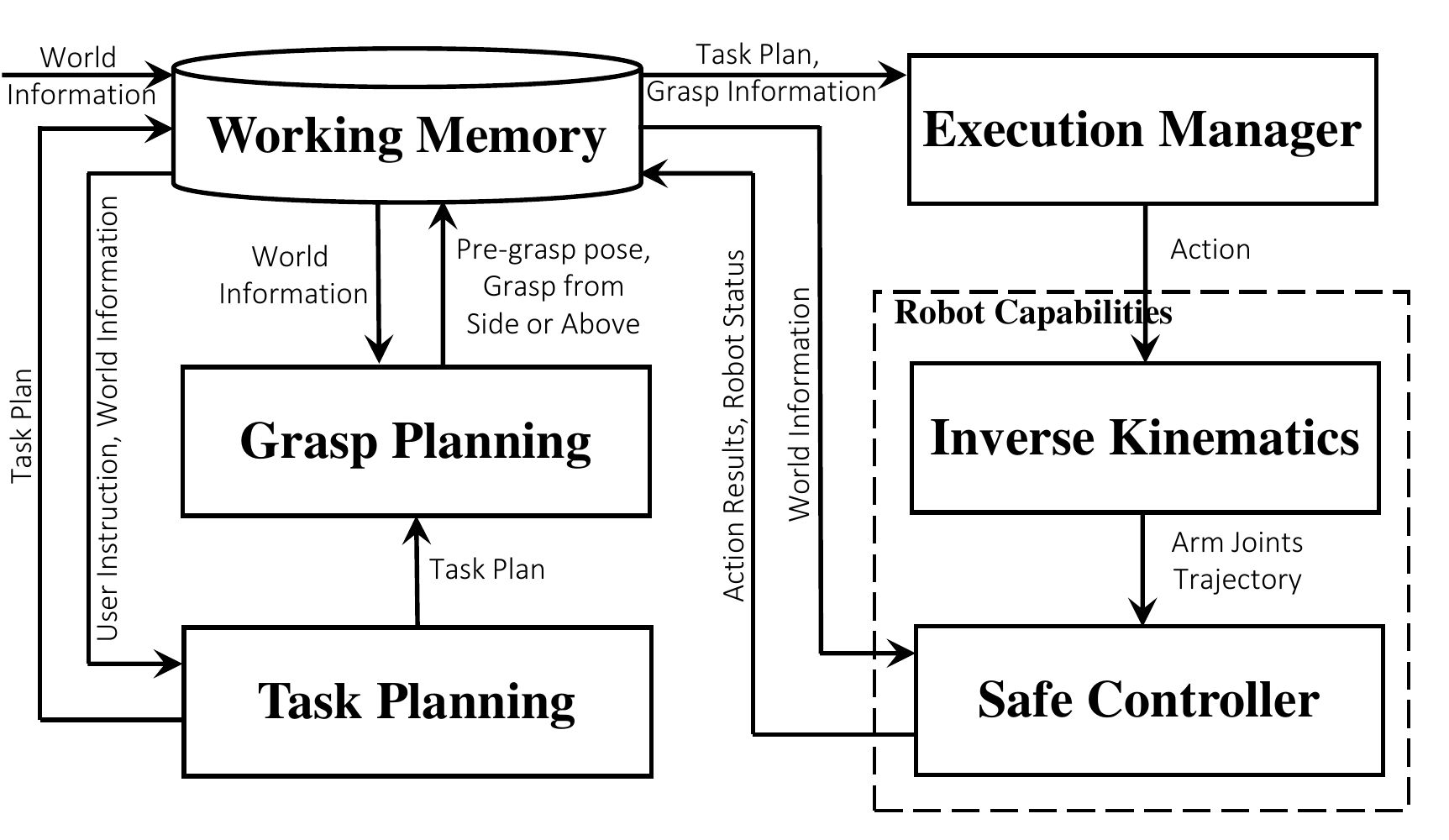}
	\vspace{-4mm}
	\caption{Schematic representation of task planning, grasp planning and execution manager.}
	\label{fig:motion_planning_architecture}       % Give a unique label
\end{figure}

In this framework, task planning is triggered when a user instructs the robot to achieve a task (e.x. \emph{clear\_table}). This is handled by the \emph{User Interaction} module. Figure \ref{fig:motion_planning_architecture} shows a schematic representation of the planning and execution framework. The current state of the system, including world model information, global characteristics of the object of interest (i.e., overall shape, main axis, center of bounding box) and robot pose is retrieved from the working memory. Then, a task plan would be generated. A plan is a sequence of primitive operators to be performed to achieve the given goal. It should be noted that \emph{Task Planning} is not in the scope of this thesis. Previously, we showed how to conceptualize successfully executed task plans and how to use these conceptualized experiences for task planning \citep{mokhtari2016experience}. In the present work, a predefined task
plan is used. In order to be executed, a task plan must be complemented with end-effector poses. A pose is represented as a tuple $G~=~(x, y, z, roll, pitch, yaw)$, specified relative to the base reference frame of the robot.
\begin{figure}[!b]
\centering
\begin{centering}
\begin{tabular}{ccc}
	\includegraphics[scale=0.095]{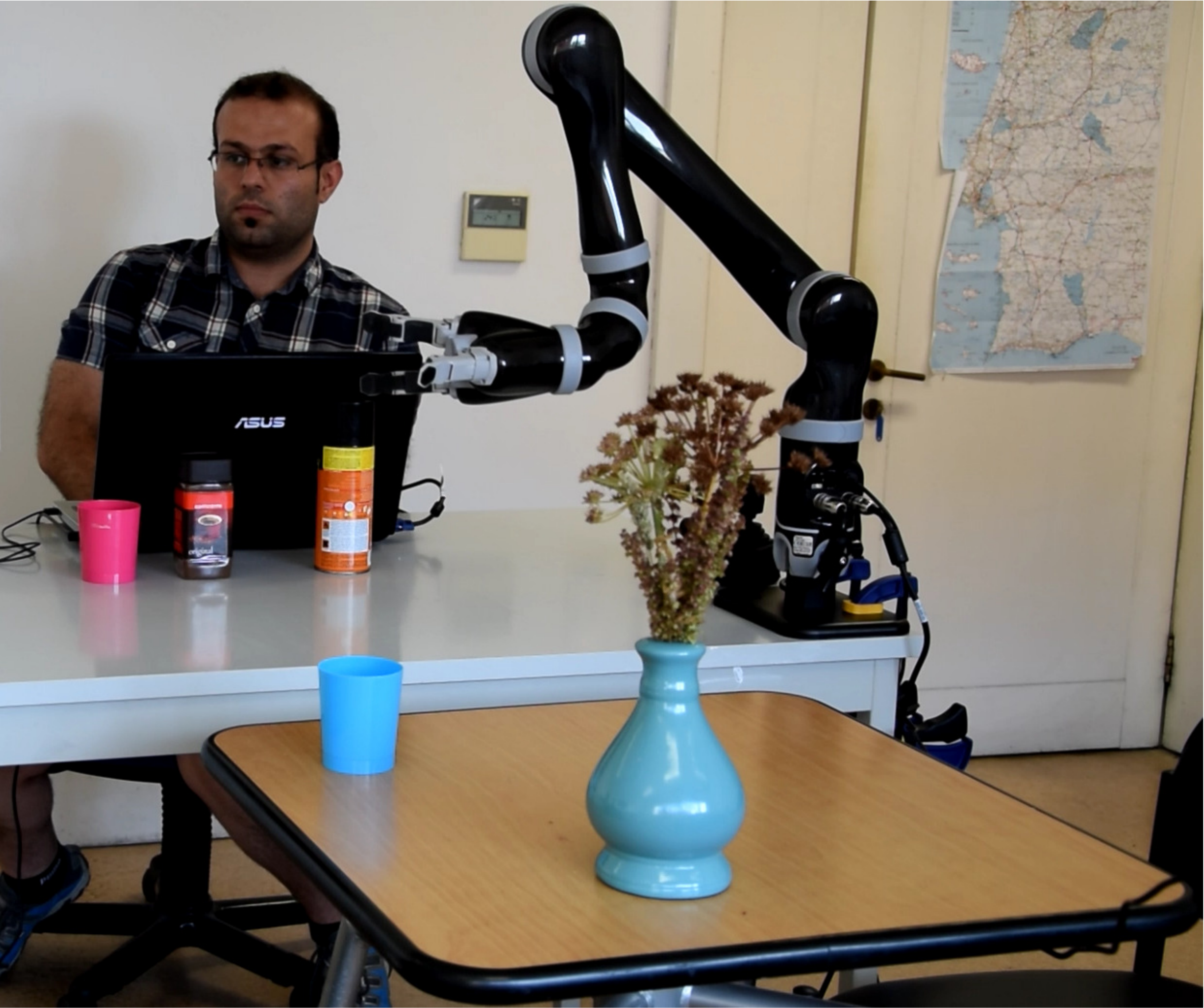}&
	\includegraphics[scale=0.095]{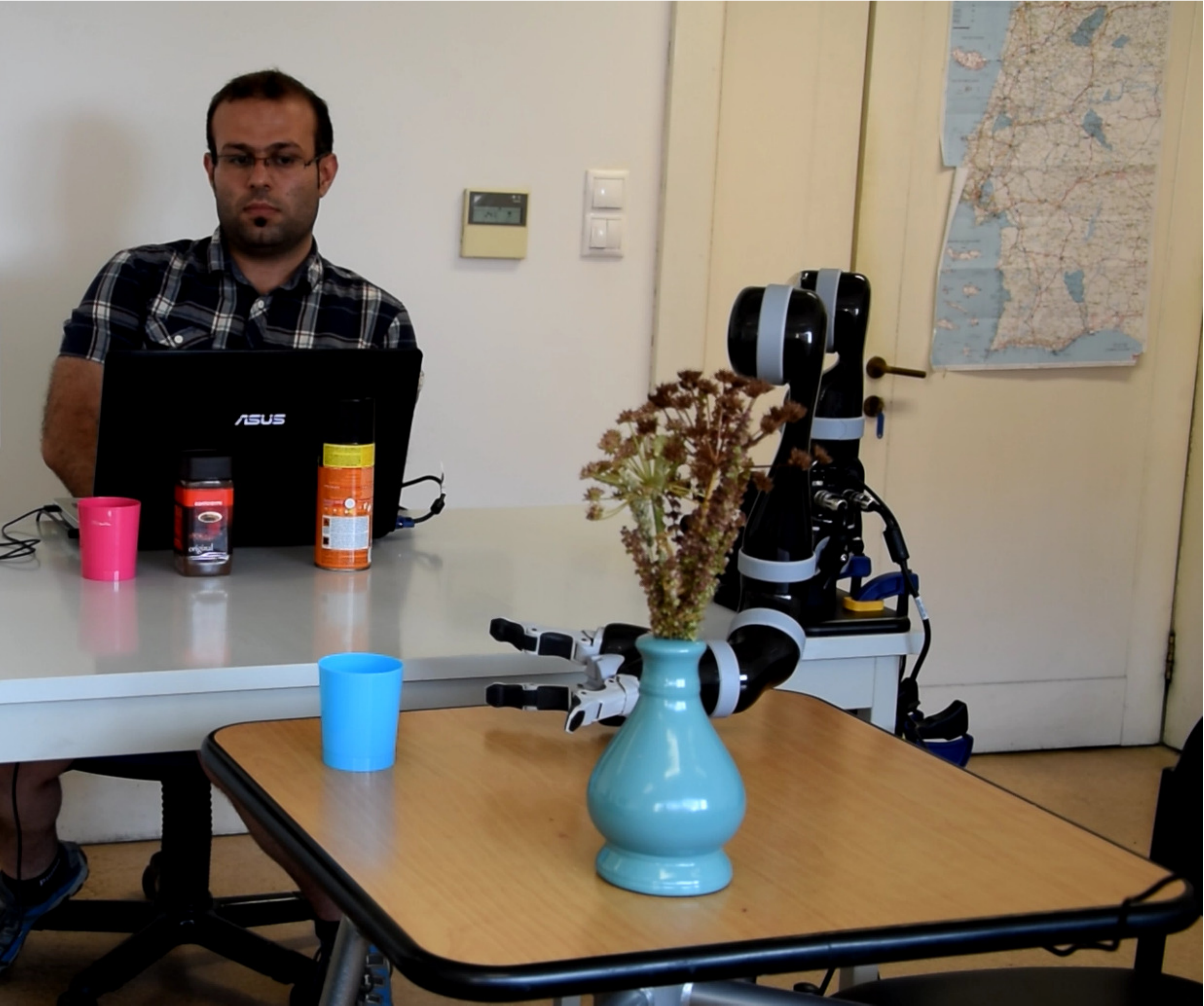}&
	\includegraphics[scale=0.095]{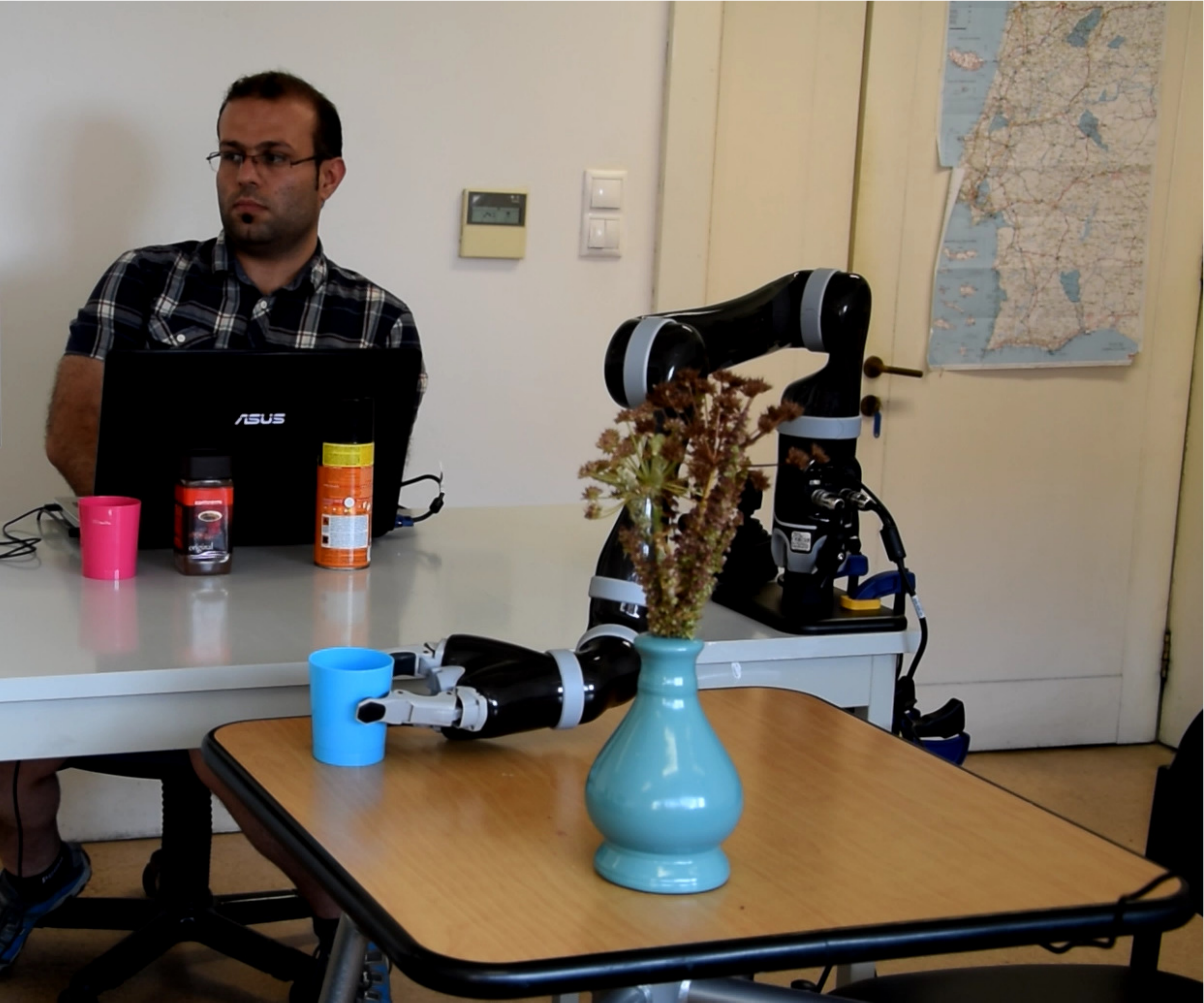}\\
	(\textit{a}) & (\textit{b})& (\textit{c}) \\
		\includegraphics[scale=0.095]{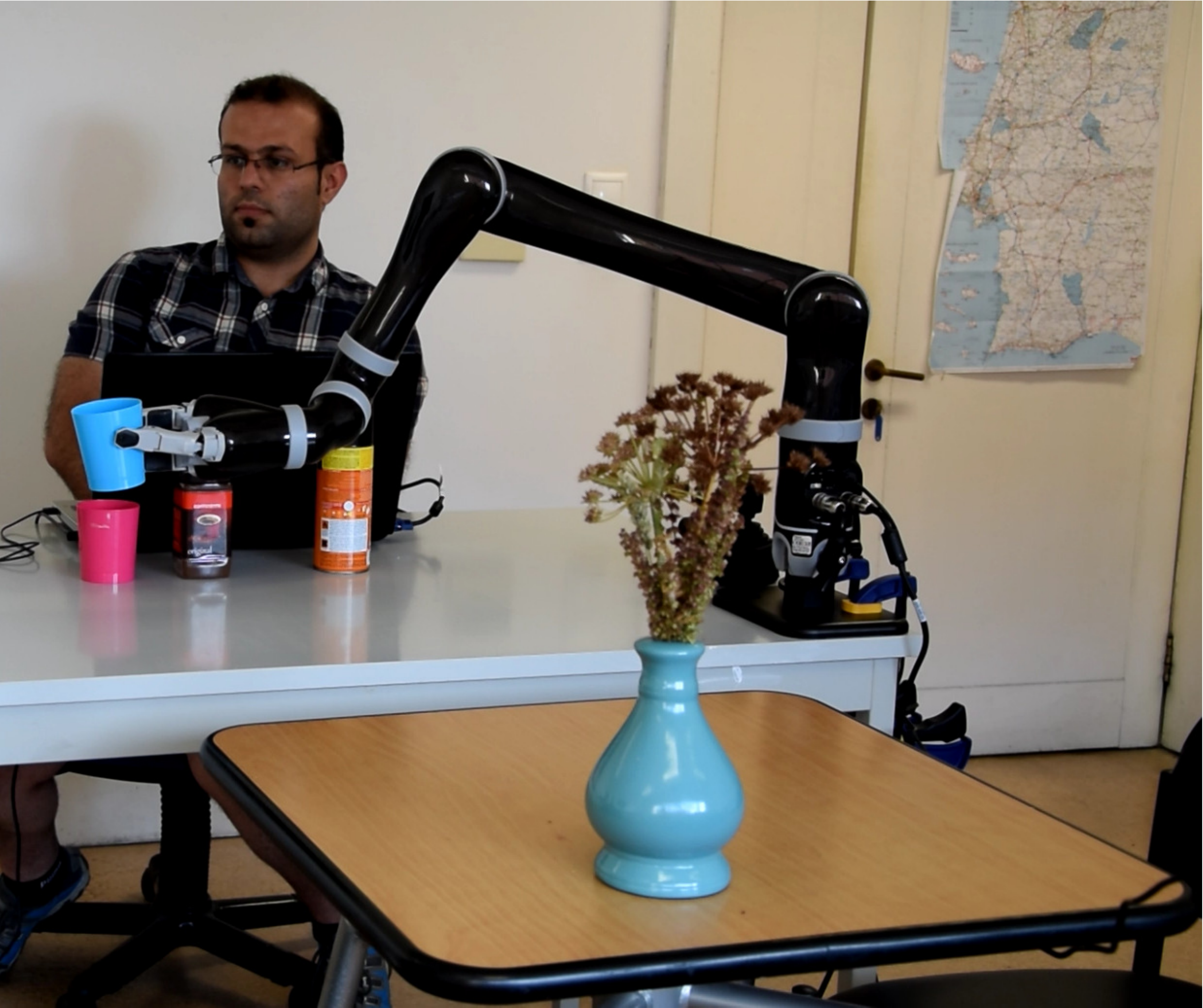}&
	\includegraphics[scale=0.095]{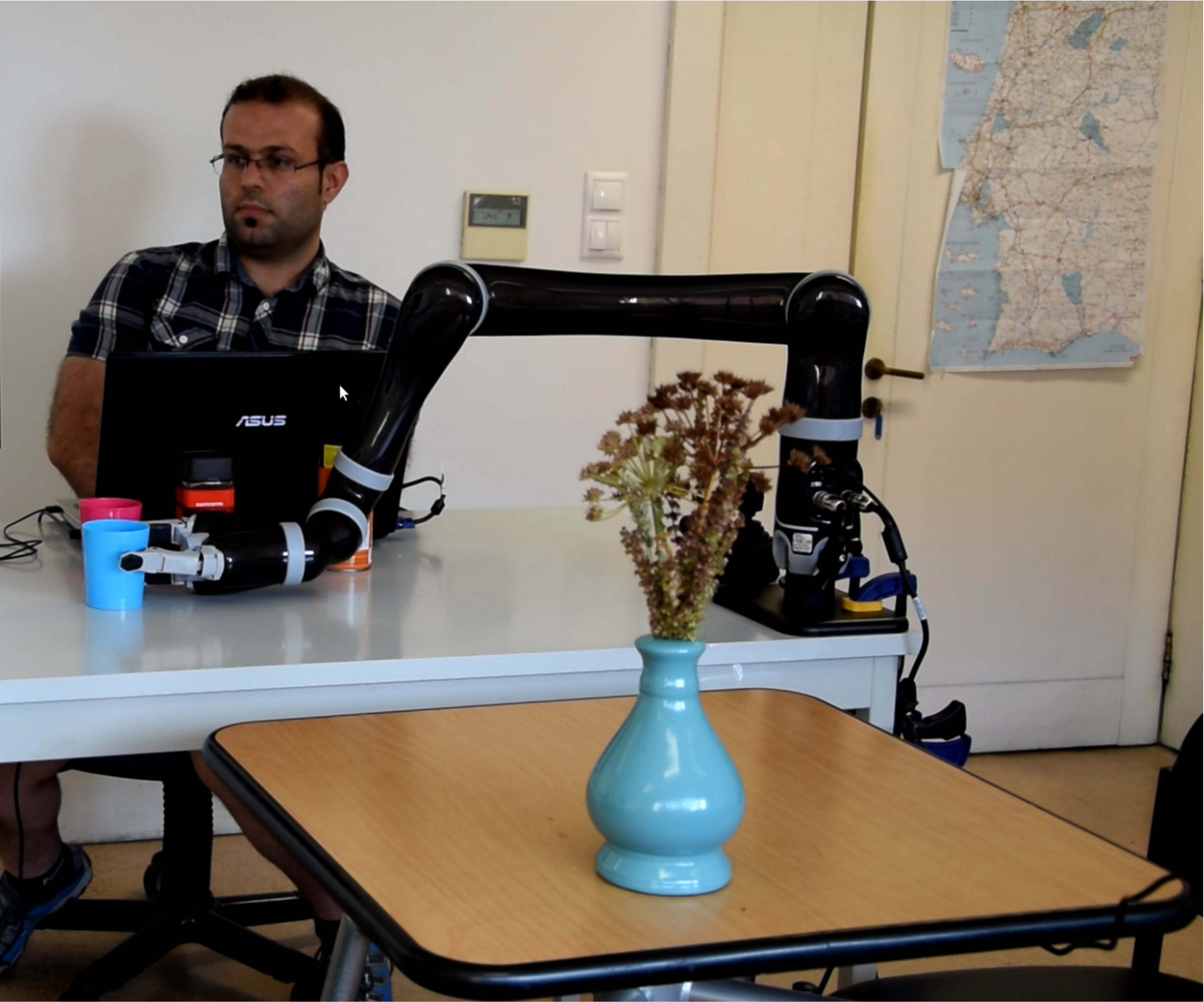}&
	\includegraphics[scale=0.095]{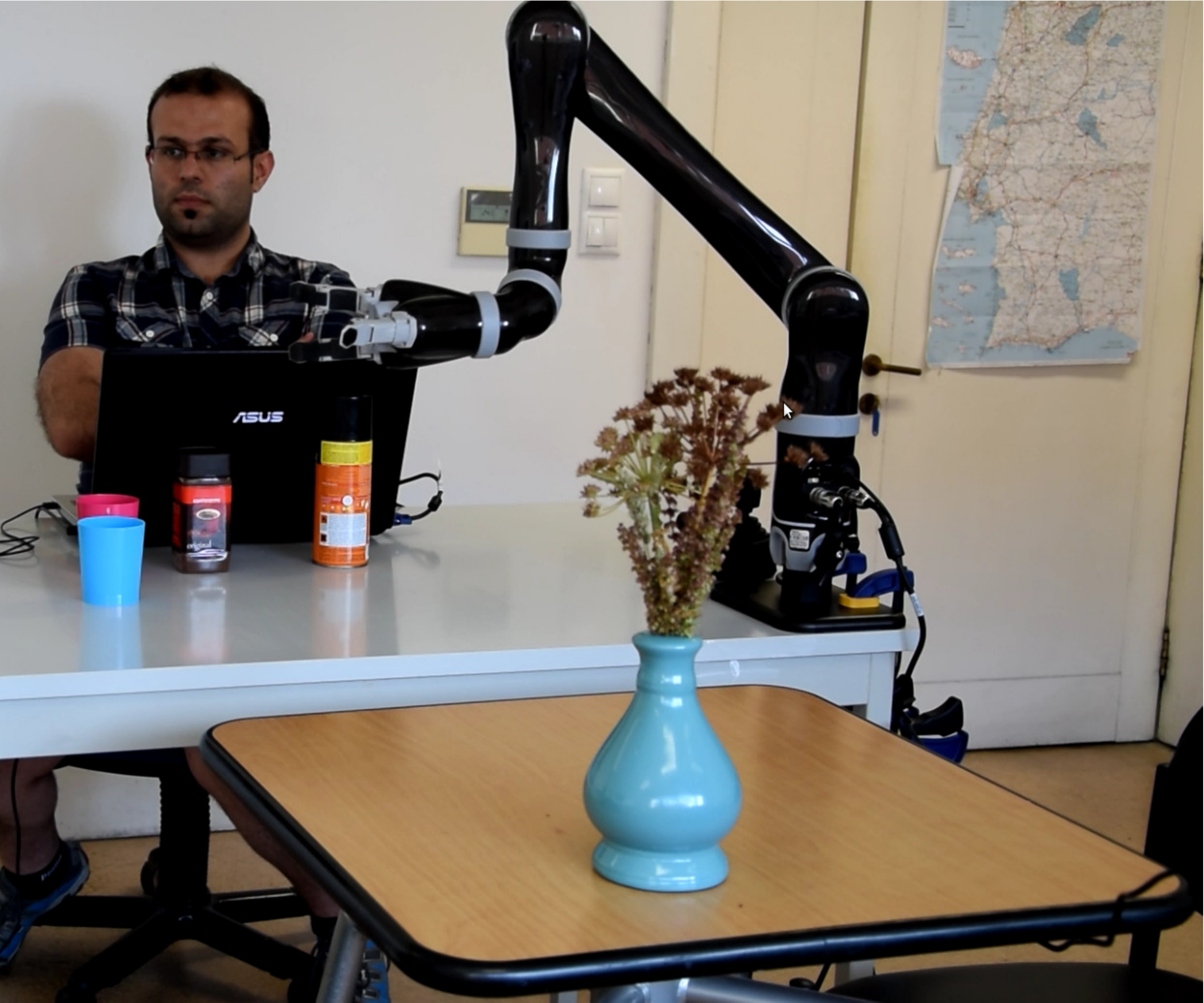}\\
	(\textit{d}) & (\textit{e})& (\textit{f}) 
\end{tabular}
\end{centering}
\caption{Sequence of snapshots showing the JACO robotic arm performing a constrained pick and place task to clean the table; In this task, the orientation of the grasped object must be kept consistent throughout the plan; (\emph{a}) the JACO robotic arm goes to the initial pose and extracts object (i.e., '\emph{PlasticCup}') pose and shape properties; (\emph{b}) a side grasp is selected and the robot goes to pre-grasp position; (\emph{c}) the robot approaches and grasps the \emph{PlasticCup}; (\emph{d}) picking up the \emph{PlasticCup} and moving it to the side; (\emph{e}) place the object and (\emph{f}) goes back to the initial position.}
\label{fig:pick_place}
\end{figure}
The \emph{Grasp Planning} module receives the task plan and chooses a grasp point either from above or from the side as well as a pre-grasp pose using the world model information and global characteristics of the object.
In the current setup, the pre-grasp pose is placed at a fixed distance ($d_{pre-grasp} = 0.15~m$) behind or above the center of bounding box of the object. The intuition behind this assumption is that many domestic objects are graspable by aligning grippers with the principal axes of the object \citep{ciocarlie2014towards} \citep{stuckler2013efficient}. Afterwards, the \emph{Execution Manager} retrieves the plan and grasp information from the \emph{Working Memory}. The \emph{Execution Manager} uses a Fine State Machine to reactively execute the plan. The actions are dispatched to the \emph{Robot Capabilities} module. Inverse kinematics and safe controller, integrated from the JACO arm driver, are used to transform a given end-effector pose goal into joint-space goals. It should be noted that, a discussion about the grasp approaches is out of the scope of this thesis. For interested readers, we provide a detailed discussion and evaluation about grasping capabilities in \citep{kasaei2016object}. In another work, we proposed an advanced grasping approach to learn how to grasp familiar objects using interactive object view labeling and kinesthetic grasp teaching \citep{shafii2016learning}. 

Whenever the object is grasped, the height of the robot's end-effector relative to the robot's base is recorded into \emph{Working Memory} and it is used as the desired height for placing the grasped object. The \emph{Execution Manager} computes a new trajectory to navigate the robot's end-effector to the placing area and sends out the action. After executing each action, the current state of the robot is updated in the \emph{Working Memory}. Since world model information is updated by different modules (i.e., \emph{Object Detection}, \emph{Execution Manager} and etc.), the \emph{Execution Manager} can abort execution when an unpredictable situation happens along expected execution path such as new obstacles move into the planned path of the robot arm. It should be noted that an orientation constraint on the end-effector is used to grasp and move an object parallel to the support plane. In addition, objects outside of the arm's workspace are not considered. Figure \ref{fig:pick_place} illustrates the result of a constrained pick and place plan executed on the robot.

\section {Summary}
\label{summary_chapter2}

In this chapter, we have presented two system architectures designed
to enhance a proper perception for a complex artificial cognitive agent working in a restaurant scenario. In particular, we provide a thorough description of the developed system, starting with motivations, cognitive considerations and then architecture design. Both architectures integrate detection, tracking, learning and recognition of tabletop objects. Interaction capabilities were also developed to enable a human user to take the role of instructor and teach new object categories. Thus, the system learns in an incremental and open-ended way from user-mediated experiences. 
In RACE architecture, based on the analysis of memory requirements for storing both semantic and perceptual data, a dual memory approach, comprising a semantic memory and a perceptual memory, was adopted. The second architecture enables robots to adapt to different environments and reason out how to behave in response to the request of a complex task such as \emph{clear\_table}. We have also tried to make the proposed architectures easy to integrate on other robotic systems. In the next chapter, we will discuss the process of gathering object experiences in both supervised and unsupervised manner.

\cleardoublepage
%\chapter{Gathering Object Experiences}
\chapter{Object Perception and Experience Extraction}
\label{chapter_3}

An autonomous robot typically uses a perception system to perceive the world. The perception system provides a set of important information that the robot has to use for interacting with users and environments. A robot needs to know which kinds of objects exist in a scene and where they are. Therefore, both object detection and recognition are important and difficult tasks due to the dynamic nature of the world.
Some robots mainly use a RGB camera to perceive the world. RGB data is not suitable for acquiring complete information of the world due to the fact of encoding 3D world by 2D images \citep{Philipona6789645}. Moreover, environmental changes such as light, shadows and reflections complicate 2D detection approaches. To cope with these limitations, several constraints are usually considered (e.g. uniform colored backgrounds or less clutter). These kinds of constraints obviously reduce the applicability of the entire system to work robustly in a real world environment. 

Following the recent release of inexpensive 3D sensing devices such as Microsoft Kinect\footnote{http://www.xbox.com/en-US/kinect} and ASUS Xtion\footnote{http://www.asus.com/Multimedia/Xtion\_PRO/}, which record RGB and depth information, 3D object detection has become a widespread research topic. We believe involving spatial depth information facilitates the detection of objects in domestic environments. In 3D space, objects are more likely to be correctly detected due to their spatial and geometric properties. Nevertheless RGB-D sensors have some drawbacks such as the inability of detecting transparent objects and distorted sensor readings of reflective and dark surfaces.

Real 3D scenes generally consist of several objects present in a scene. It is a challenging task to robustly detect multiple objects in a domestic environment due to severe occlusions and clutter. The environment can be highly \textbf{crowded} and \textbf{cluttered}. Clutter is seen when points that do not belong to the target object are included in the segmentation. \textbf{Occlusion} and \textbf{self-occlusion} can lead to only a part of an object being visible in certain views. Moreover, some captured data can be inaccurate or missing due to \textbf{sensor noise}.

Throughout this chapter, we try to impose as few constraints as possible for the object detection. We assume objects are situated on a planar surface, as this is the common pose of objects in domestic environments and transparent objects like glasses are not considered; but we do not consider any other assumptions about the object appearance. In this chapter, we will propose automatic perception capabilities that will allow robots to, (\emph{i}) automatically detect multiple objects in a crowded scene; (\emph{ii}) incrementally accumulate object experiences in both supervised and unsupervised manner; and (\emph{iii}) Next-Best-View (NBV) prediction algorithm to predict the next best camera pose for object detection by rendering virtual scenes based on current object hypotheses.

All the work presented in this chapter has appeared in conference and journal papers \citep{Hertzberg2014projrep,Kasaei2014,oliveira20153d,kasaei2016object,kasaei2017Neurocomputing,JuilICCVW2017,kasaeiAAAI2018}. The remainder of the chapter is organized as follows: the first section is dedicated to the related work. In section~\ref{sec:preprocessing}, the pre-processing of raw sensory data is explained. Section \ref{sec:objectDetection} describes in detail the object detection methodologies. Unsupervised and supervised experience gathering are explained in sections~\ref{sec:unsupervisedExperienceGathering} and section~\ref{sec:supervisedExperienceGathering} respectively. An approach for online object model construction will be discussed in section~\ref{model_construction} followed by the Next Best View prediction approach. Summary is then presented in section~\ref{sec:summary_chapter3}.

%^^^^^^^^^^^^^^^^^^^^^^^^^^^^^^^^^^^^^^^^^^^^^^^^^^^^^^^^^^^^^^^^
%^^^^^^^^^^^^^^^^^^^^^^^^^^^^^^^^^^^^^^^^^^^^^^^^^^^^^^^^^^^^^^^^
\section{Related Work}
\label{sec:related_work}

Many approaches have been developed for gathering online object experiences based on both supervised and unsupervised algorithms \citep{kirstein2012,oliveira20153d}. In supervised methods, an object experience is collected when an instructor provides a category label for an object. These methods provide an opportunity to collect interactively object experiences (e.g., visual observations) for learning. \cite {kirstein2012} proposed a lifelong learning approach for interactive learning of multiple categories based on vector quantization and an user interface. Similar to our work, \cite {chauhan2011} approached the problem of object experience gathering and category learning with a focus on open-ended learning and human-robot interaction. They used RGB data whereas we used depth data. Therefore, their object detection and representation approaches are completely different from our approach. \cite{collet2014herbdisc} proposed a graph-based approach for lifelong robotic object discovery. Similar to our approach, they used a set of constraints to explore the environment and to detect object candidates from raw RGB-D data streams. In contrast, their system does not interactively acquire more data to learn and recognize the object. \cite{steels2000aibo} used the notion of \emph{``language game''} to
develop a social learning framework through which a robot can learn its first words. A teacher points to objects and provides their names. The robot uses color histograms and
an instance-based learning method to learn word meanings. The mediator can also ask questions and provide feedback on the robot’s answers. They conclude that social interaction must be used to help the learner focus on what needs to be learned in the context of communication. We take inspiration from this work, since our system also employs the three core instructions \emph{teach}, \emph{ask}, \emph{correct}. \cite{lopes2007many} also developed a vocabulary acquisition and category learning system that integrates the user as instructor. The user can provide the names of objects through pointing and verbal teaching actions. The user can also ask questions about the categories of objects under shared attention and, if appropriate, provide corrective feedback. In this work the teaching was limited to object names.

Object detection and pose estimation are also crucial for robotics applications and recently attracted attention of the research community \citep{JuilICCVW2017,Tejani2014}. Many researchers participated in public challenges such as Amazon picking challenge\footnote{https://www.amazonrobotics.com} to solve multiple objects detection and pose estimation in a realistic scenario. This shows that object detection research is moving towards more realistic robotics environment. Traditional approaches have mostly been well studied in known object scenarios. For instance, \cite{Doumanoglou2016} used sparse autoencoder to represent each patch and classified using random forest. \cite{Tejani2014} used LINEMOD \citep{Hinterstoisser2011} to represent each patch and also used random forest and Hough voting to generate hypotheses. In these approaches, the robot is expected to encounter the same objects it was trained on. 

A more realistic and challenging setting is when the objects that the robot will encounter were not learned in advance. Towards this goal, \cite{mason2014unsupervised} proposed a robotic system for unsupervised object and class discovery, in which object candidates are first discovered from several scenes, and then grouped into classes in an unsupervised fashion. Similar to our work, this approach first segments the world into ``object'' and ``non-object'' components, and then performs data association between the objects. \cite{kang2011discovering} presented an unsupervised object discovery approach based on combining a hierarchical over-segmentation with visual information. \cite{kang2011discovering} and \cite{karpathy2013object} proposed methods for discovering object models from 3D point clouds of indoor environments. Similar to our approach, the focus of these works is on identifying portions of a scene that could correspond to objects (i.e., scene segmentation) for the purposes of object recognition, semantic understanding or robotic manipulation.

A robot operating in human environments may frequently encounter with a pile of objects such as a pile of toys in the living room, tidying up a messy dinning table, and multiple unused objects stacked in a box in the garage. Object detection and object pose estimation in such environments are challenging tasks due to severe occlusions and clutter. Several strategies, including active exploration \citep{Doumanoglou2016,mauro2014unified} and interaction with piles \citep{katz2013clearing, katz2014perceiving} have been proposed to overcome pile segmentation issues. 

In active perception scenarios, whenever the observer fails to recover the poses of objects from the current view point, the observer will estimate the next view position and capture a new scene image from that position to improve the knowledge of the environment. This will reduce the object detection and pose estimation uncertainty. Towards this end, \cite{mauro2014unified} proposed a unified framework for content-aware next best view selection based on several quality features such as density, uncertainty, 2D and 3D saliency. Using these features, they computed a view importance factor for a given scene. Unlike this approach, we first segment a given scene into object hypotheses. Then, the next best view is predicted based on the properties of those object hypotheses.
In another work, \cite{biasotti2013sketch} approached the problem of defining the representative views for a single 3D object based on visual complexity. They proposed a new method for measuring the viewpoint complexity based on entropy. Their approach revealed that it is possible to retrieve and to cluster similar viewpoints. \cite{Doumanoglou2016} uses class entropy of samples stored in the leaf nodes of Hough forest to estimate the Next-Best-View which could reduce the uncertainty of the class of detected objects. 

Some researchers have recently adopted deep learning algorithms for next best view prediction in active object perception \citep{wu20153d, Johns2016}. For instance, \cite{wu20153d} proposed a deep network namely 3D ShapeNets to represent a geometric shape as a probabilistic distribution of binary variables on a 3D voxel grid. As they pointed out, training a deep network for next best view prediction requires a large collection of 3D objects to provide accurate representations and typically involves long training times. Moreover, unlike our approach, these kinds of approaches are mainly suitable for isolated objects and become brittle and unreliable in crowded scenarios.

Object manipulation is also useful for a robot to discover and segment objects in cluttered environments \citep{katz2013clearing,katz2014perceiving}. \cite{van2014probabilistic}  presented a part-based probabilistic approach
for interactive segmentation. They tried to minimize human intervention in the sense that the robot learns from the effects of its actions, rather than human-given labels. In another work, \cite{gupta2012using} explored manipulation-aided perception and grasping in the context of sorting small objects on a tabletop. They presented a pipeline that combines perception and manipulation to accurately sort the bricks by color and size. This topic is out of the scope of this thesis and an interesting topic for further research.
%^^^^^^^^^^^^^^^^^^^^^^^^^^^^^^^^^^^^^^^^^^^^^^^^^^^^^^^^^^^^^^^^
%^^^^^^^^^^^^^^^^^^^^^^^^^^^^^^^^^^^^^^^^^^^^^^^^^^^^^^^^^^^^^^^^
\section{Pre-Processing}
\label{sec:preprocessing}
%^^^^^^^^^^^^^^^^^^^^^^^^^^^^^^^^^^^^^^^^^^^^^^^^^^^^^^^^^^^^^^^^
%^^^^^^^^^^^^^^^^^^^^^^^^^^^^^^^^^^^^^^^^^^^^^^^^^^^^^^^^^^^^^^^^
\begin{figure}[!b]
\begin{tabular}[width=\textwidth]{c c}
\hspace{-0.5cm}
 \includegraphics[width=0.65\textwidth]{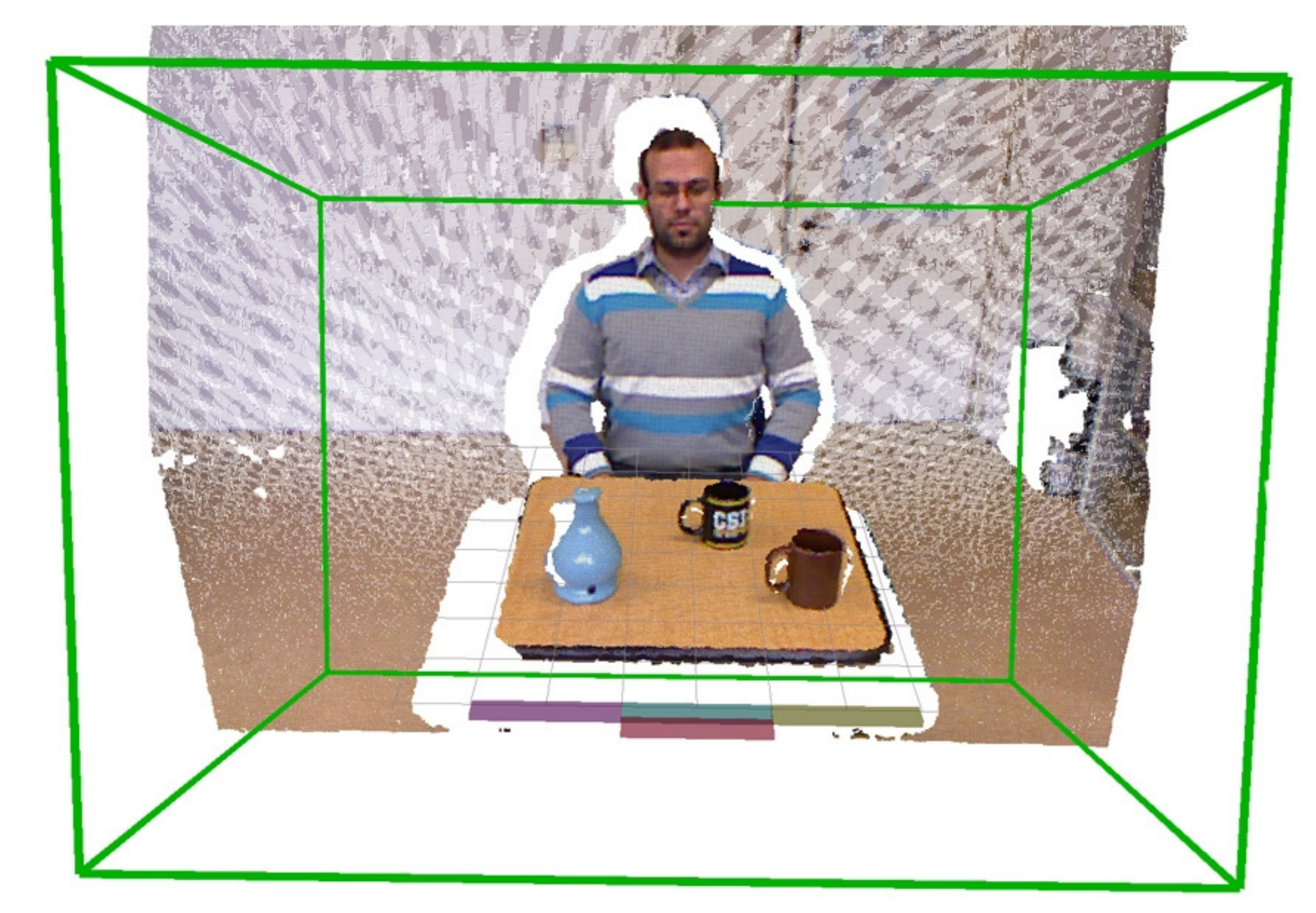} & \hspace{-0.4cm}
  \includegraphics[width=0.38\textwidth]{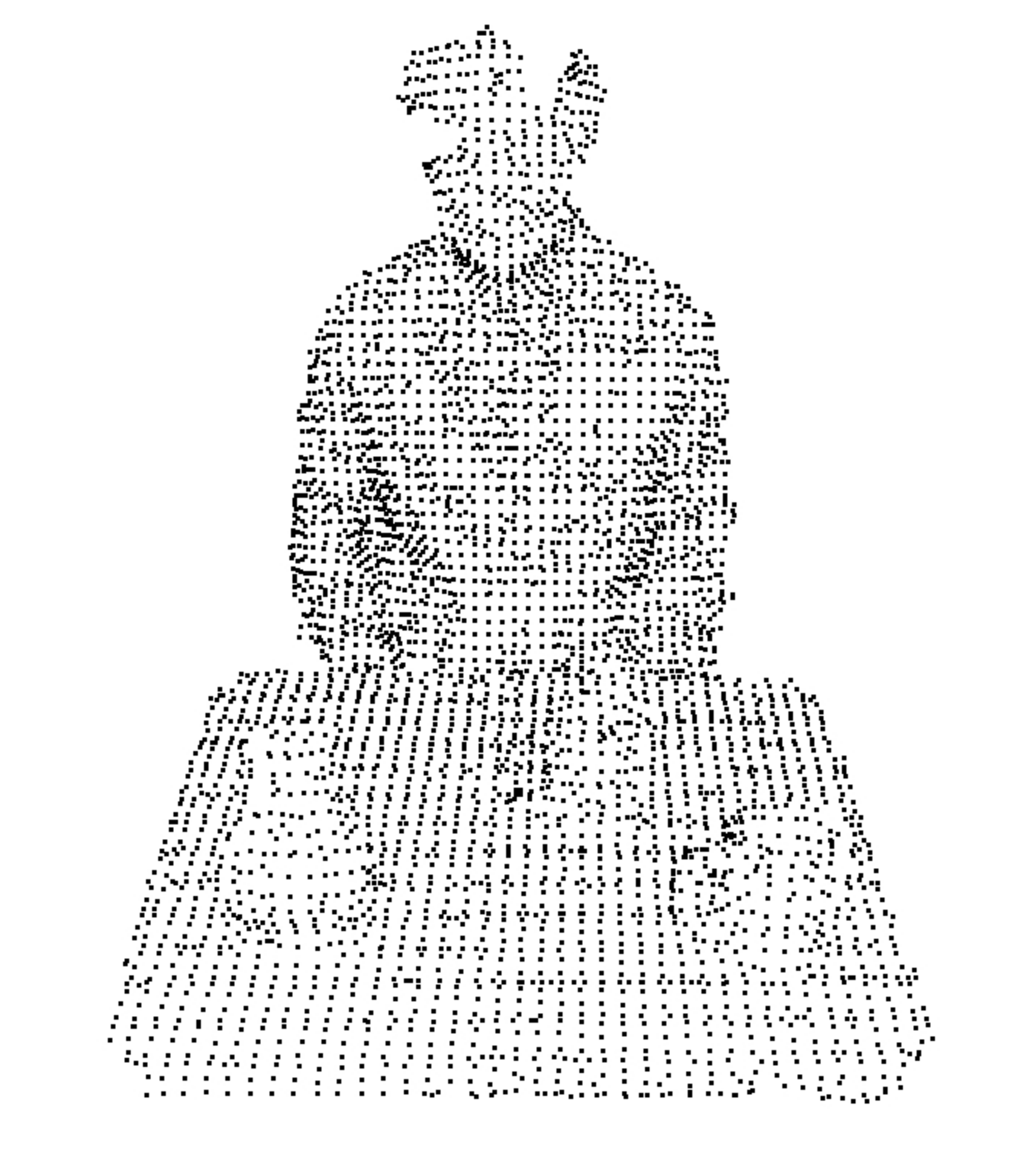} \\ 
\end{tabular}
% figure caption is below the figure
\caption{An example of the pre-processing functions: \emph{(left) } an example of the distance filtering using a cubic volume.
graphical perspective projection of the original point cloud and distance filtering; \emph{(right)} an example of the down sampling using a voxel grid filter.}
\label{fig:2}       % Give a unique label
\end{figure}
Processing massive point clouds is one of the main challenges of 3D perception systems. 
In dense 3D point cloud data, considering all points is computationally too expensive, and real-time processing is not possible. The key idea for fast processing of massive point clouds is to use mechanisms for removing unnecessary or irrelevant data. To accomplish this, we use two separate filters that discard vast quantities of unnecessary 3D points.  The first filter uses a cubic volume to define the region of interest. Experimental results provided by \cite{khoshelham2011accuracy} have shown that the reliable and useful data are located in a cubic area with almost two meters length on each side. In our current setup, we use a table which is approximately one meter away from the camera / robot. Using this information, we define the size of the cubic volume to include a typical table in front of the robot. We also filter out undefined or unrepresentable (NaN) points. Figure \ref{fig:2} shows an example of this process. In Fig.~\ref{fig:2}\emph{ (left)}, the complete point cloud is shown, along with the cube. This filter enables a significant reduction of the number of points.

\begin{figure}[!b]
\center
\begin{tabular}[width=\textwidth]{c c}
 \includegraphics[width=0.48\textwidth]{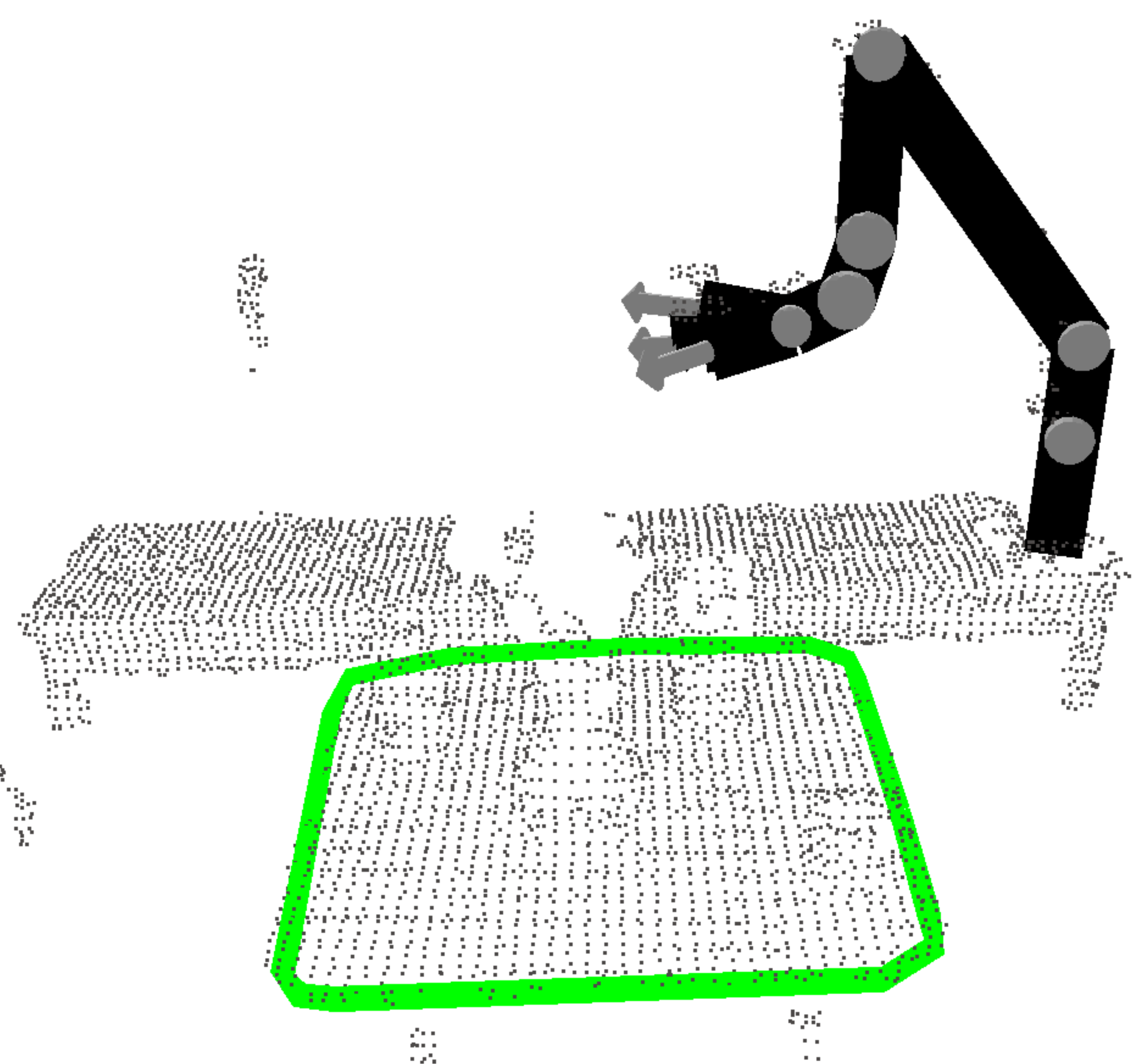}&
 \includegraphics[width=0.48\textwidth]{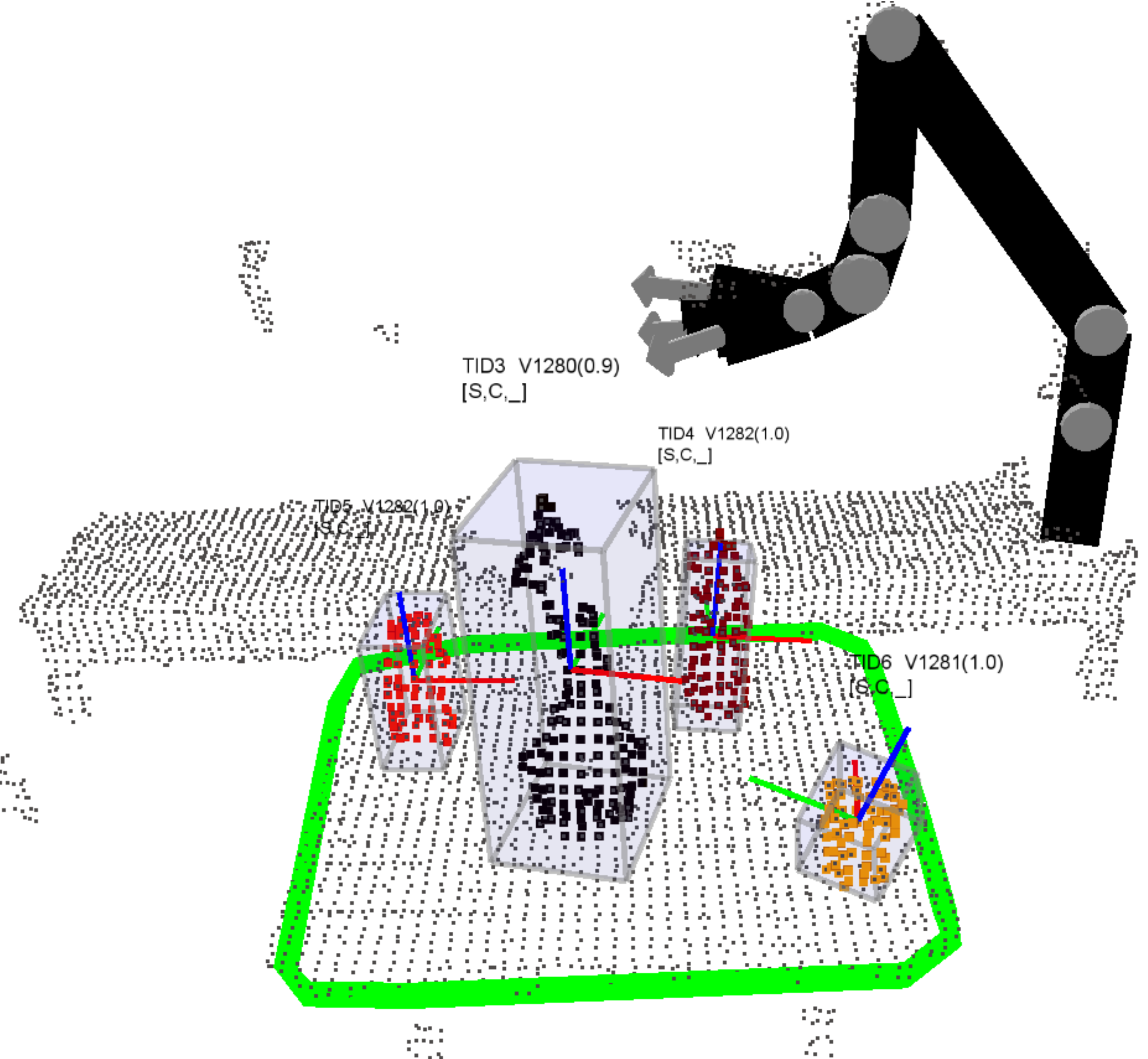} \\ 
\end{tabular}
% figure caption is below the figure
\caption{ An example of the planar surface detection and the \emph{Object Detection}: \emph{(left)} result of the table detection; the position of the arm joints are used to filter out the points corresponding to robot's body from the original point cloud. \emph{(right)} The object candidates are shown by different bounding boxes and colors. The red, green and blue lines represent the local reference frame of the objects. }

\label{fig:3}       % Give a unique label
\end{figure}

The second filter reduces the spatial resolution of the point cloud, since our approach does not require the full resolution of the sensor. To do this, the point cloud is down sampled using a voxelized grid approach\footnote{http://pointclouds.org/documentation/tutorials/voxel\textunderscore grid.php}. The advantage of this, apart from the fact that the number of points is further reduced, is that the spatial distribution of 3D points becomes uniform. In Fig.~\ref{fig:2} \emph{(right)}, the filtered point cloud is displayed. Furthermore, as shown in Fig.~\ref{fig:3}, the points corresponding to the body of the robot are filtered out from the original point cloud by retrieving the knowledge of the positions of the arm joints relative to the camera pose from the working memory. 

% For one-column wide figures use
%
%^^^^^^^^^^^^^^^^^^^^^^^^^^^^^^^^^^^^^^^^^^^^^^^^^^^^^^^^^^^^^^^^
%^^^^^^^^^^^^^^^^^^^^^^^^^^^^^^^^^^^^^^^^^^^^^^^^^^^^^^^^^^^^^^^^
\section { Object Detection and Tracking}
\label{sec:objectDetection}
%^^^^^^^^^^^^^^^^^^^^^^^^^^^^^^^^^^^^^^^^^^^^^^^^^^^^^^^^^^^^^^^^%^^^^^^^^^^^^^^^^^^^^^^^^^^^^^^^^^^^^^^^^^^^^^^^^^^^^^^^^^^^^^^^^
After pre-processing, the next step is to find objects in the scene. Our approach assumes that objects are placed on top of a planar surface (e.g., a table) in order to be detected. The planar surface is detected by finding the dominant plane in the point cloud. This is done using the RANSAC algorithm \citep{fischler}. The algorithm starts by generating plane hypotheses based on three unique non-collinear points. For each plane hypothesis, distances from all points in the point cloud to the plane are computed. The plane hypotheses are then scored based on counting the number
of points whose distance to the plane falls below a user-specified threshold, $\tau$. The RANSAC algorithm is repeated for a certain number of iterations, $N$. In the current implementation, $\tau = 20 mm$ and $N = 200$. An example of the proposed table detector algorithm is illustrated in Fig.~\ref{fig:3} \emph{(left)}. With the table detected, it is now possible to extract the points which lie directly above it. The mechanism we use to do this is called extraction of polygonal prisms\footnote {http://docs.pointclouds.org/1.0.0/classpcl\textunderscore1\textunderscore1\textunderscore extract\textunderscore polygonal\textunderscore prism\textunderscore data.html}. After this, we have a point cloud where all the objects that are on top of the table are included. Therefore, the extracted points are assumed to potentially belong to objects. The obtained point cloud is then segmented into individual clusters using the Euclidean Cluster Extraction algorithm\footnote{http://www.pointclouds.org/documentation/tutorials/cluster\textunderscore extraction.php}.
Each small group of points will be treated as an object candidate. An example of the object detector module is shown in Fig.~\ref{fig:3}. \emph{(right)}, where four objects are on top of the table. In Fig.~\ref{fig:3} \emph{(right)}, the segmented object point clouds are shown. Note that the point clouds of each object have different colors, meaning that they have been correctly segmented.
\noindent The object detection uses a size constraint, $C_{\operatorname{size}}$, to detect objects which can be manipulated by the robot.

% For one-column wide figures use
\begin{figure}[!b]
\begin{tabular}[width=0.9\textwidth]{c}
 \includegraphics[width=\textwidth]{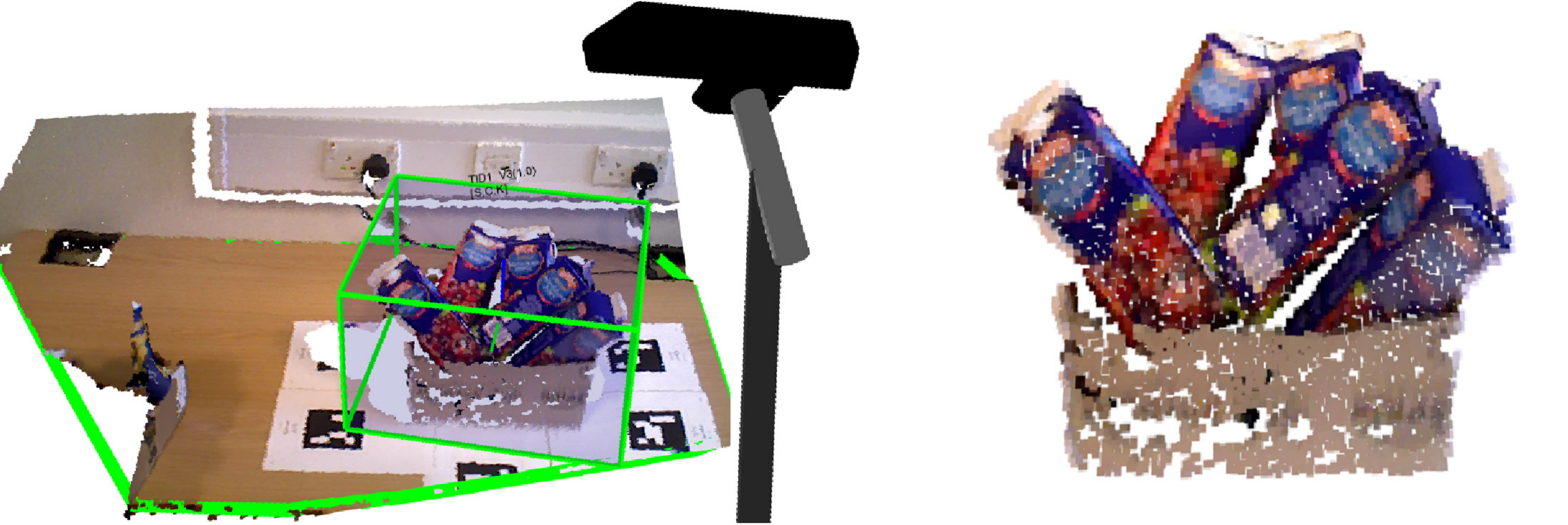}
\end{tabular}
% figure caption is below the figure
\caption{ Euclidean clusterings algorithm is not enough for such crowed scenes as shown in the (\emph{left}) image. (\emph{right}) the extracted region is dispatched to a hierarchical clustering procedure to detect multiple object candidates.}
 
\label{fig:pile}       % Give a unique label
\end{figure}
\begin{figure}[!t]
\center
\begin{tabular}[width=1\textwidth]{c}
 \includegraphics[width=0.8\textwidth,trim= 0cm 0cm 0cm 0cm,clip=true]{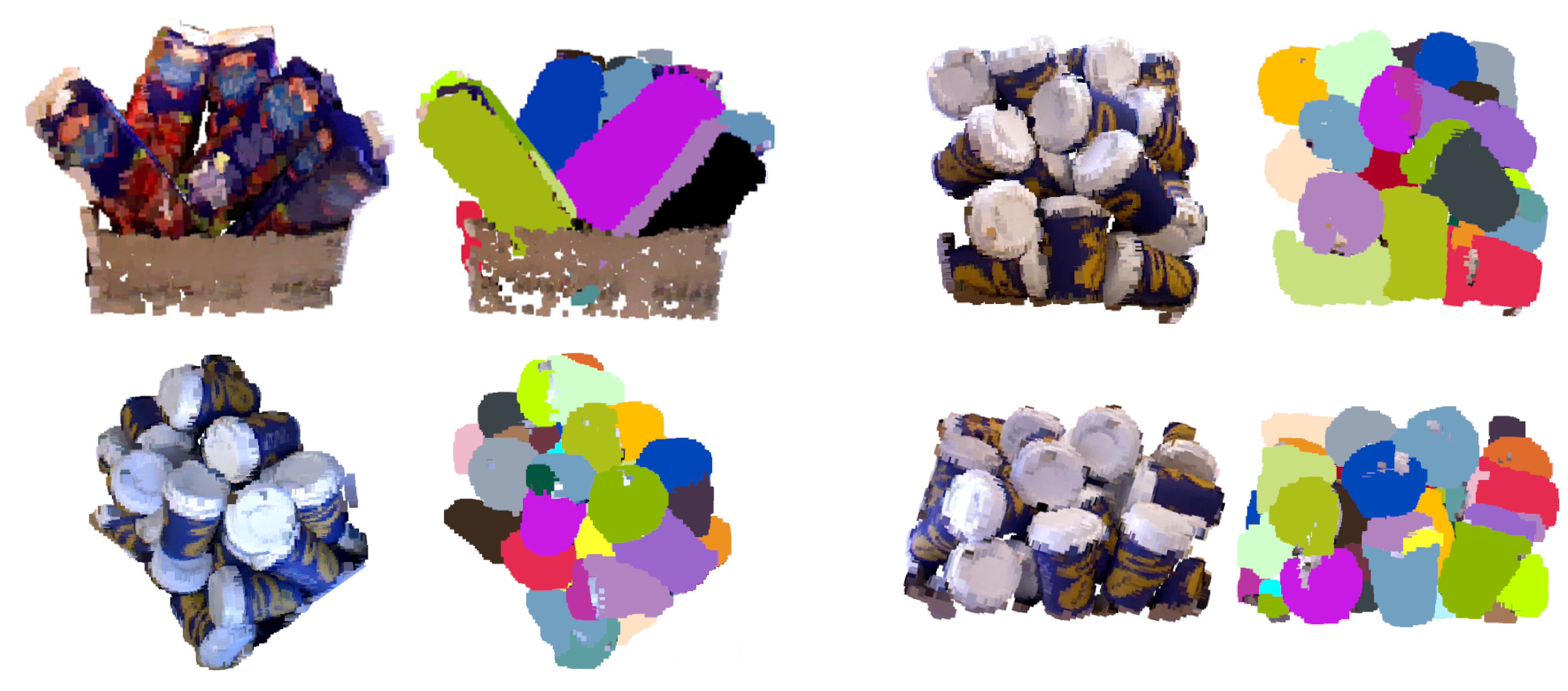}
\end{tabular}
\vspace{-2mm}
% figure caption is below the figure
\caption{Four complex scenes from Imperial College London bin-picking dataset and their corresponding segmentation results.}
\vspace{-2mm}
\label{fig:two_complex_scene}       % Give a unique label
\end{figure}
%^^^^^^^^^^^^^^^^^^^^^^^^^^^^^^^^^^^^^^^^^^^^^^^^^^^^^^^^^^^^^^^^
%^^^^^^^^^^^^^^^^^^^^^^^^^^^^^^^^^^^^^^^^^^^^^^^^^^^^^^^^^^^^^^^^
\begin{figure}[!t]
\center
\begin{tabular}[width=1\textwidth]{ccc}
\includegraphics[width=0.3\textwidth,trim= 0cm 0cm 0cm 0cm,clip=true]{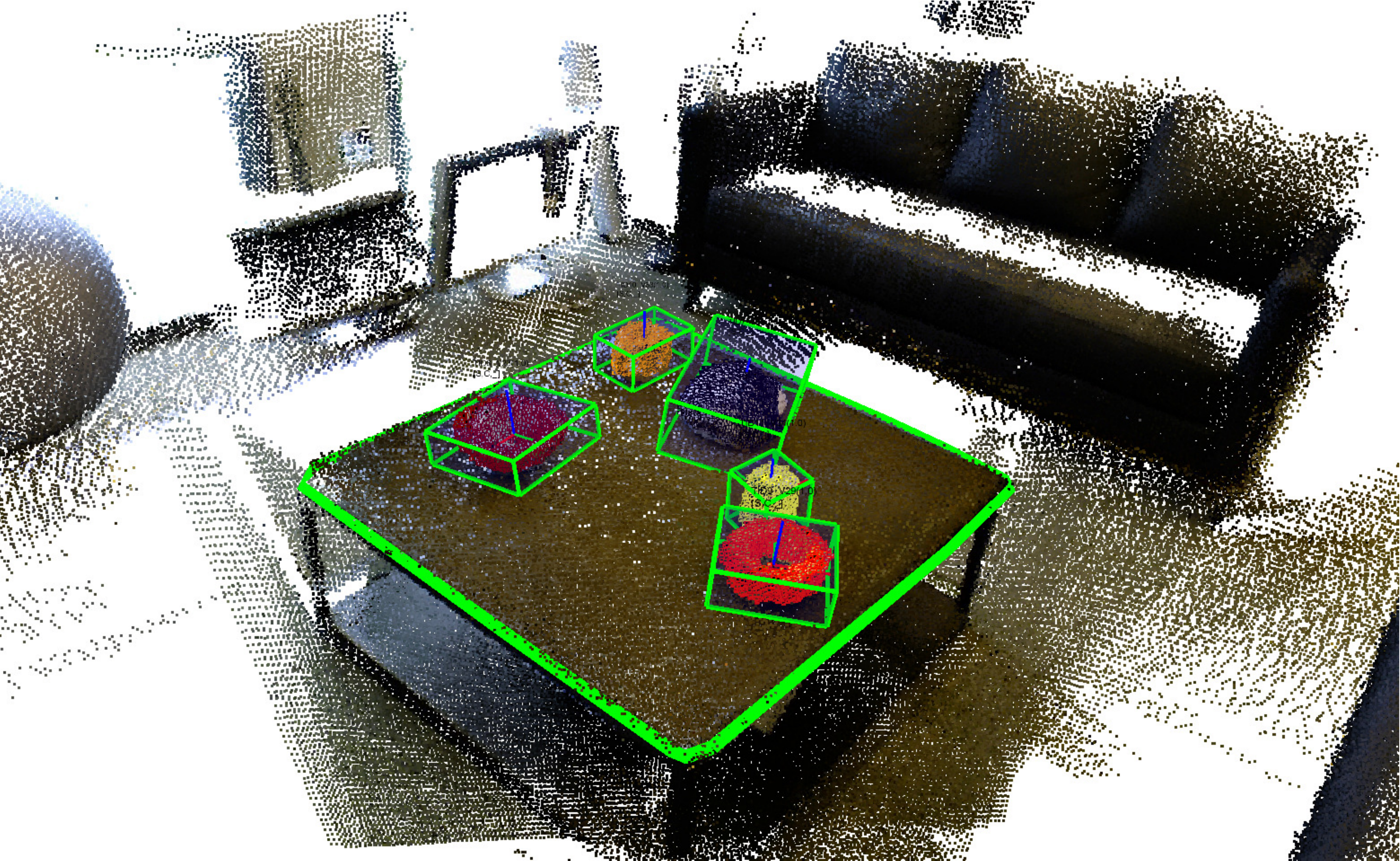}&
\includegraphics[width=0.3\textwidth,trim= 0cm 0cm 0cm 0cm,clip=true]{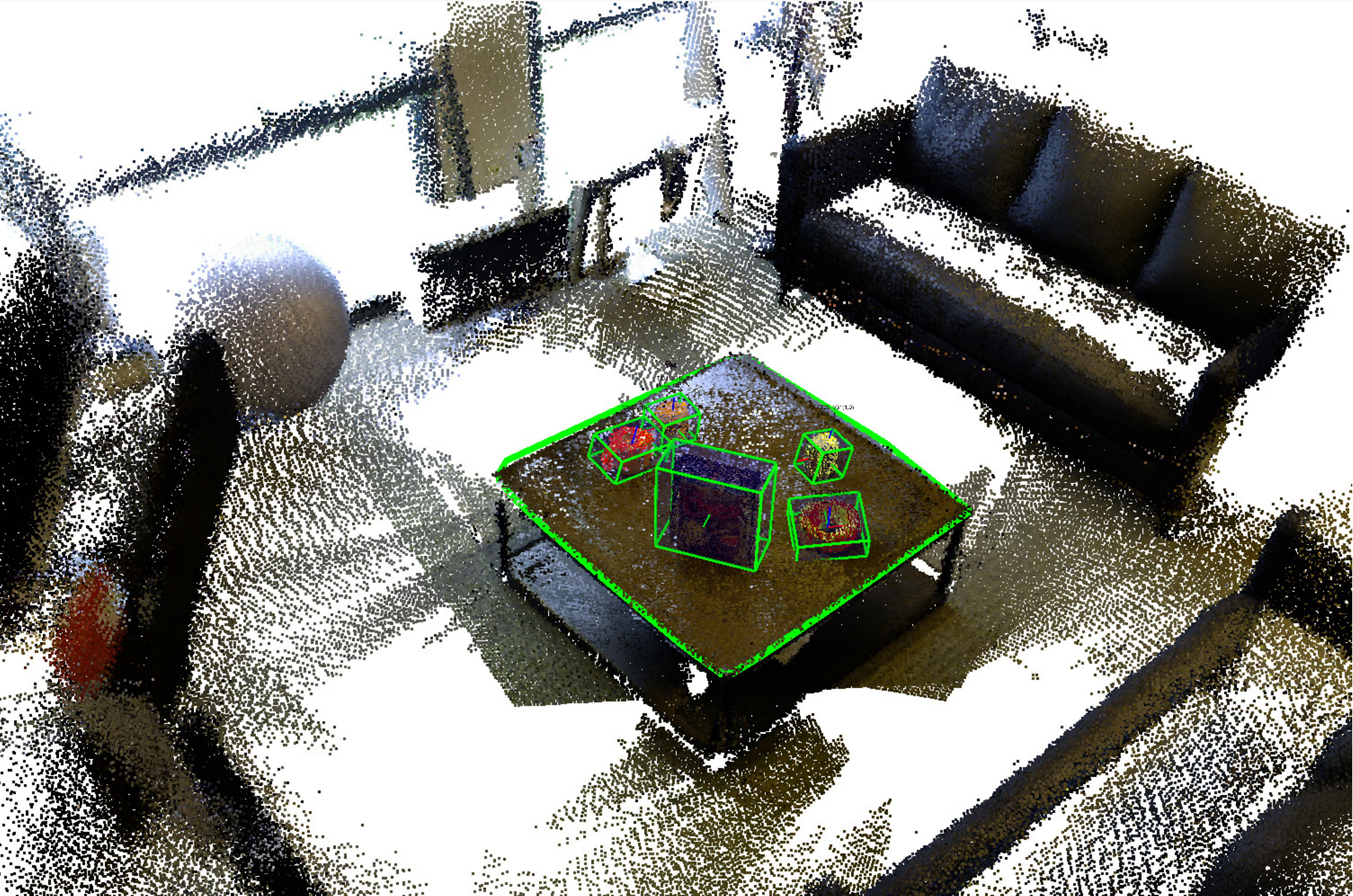}&
\includegraphics[width=0.3\textwidth,trim= 0cm 0cm 0cm 0cm,clip=true]{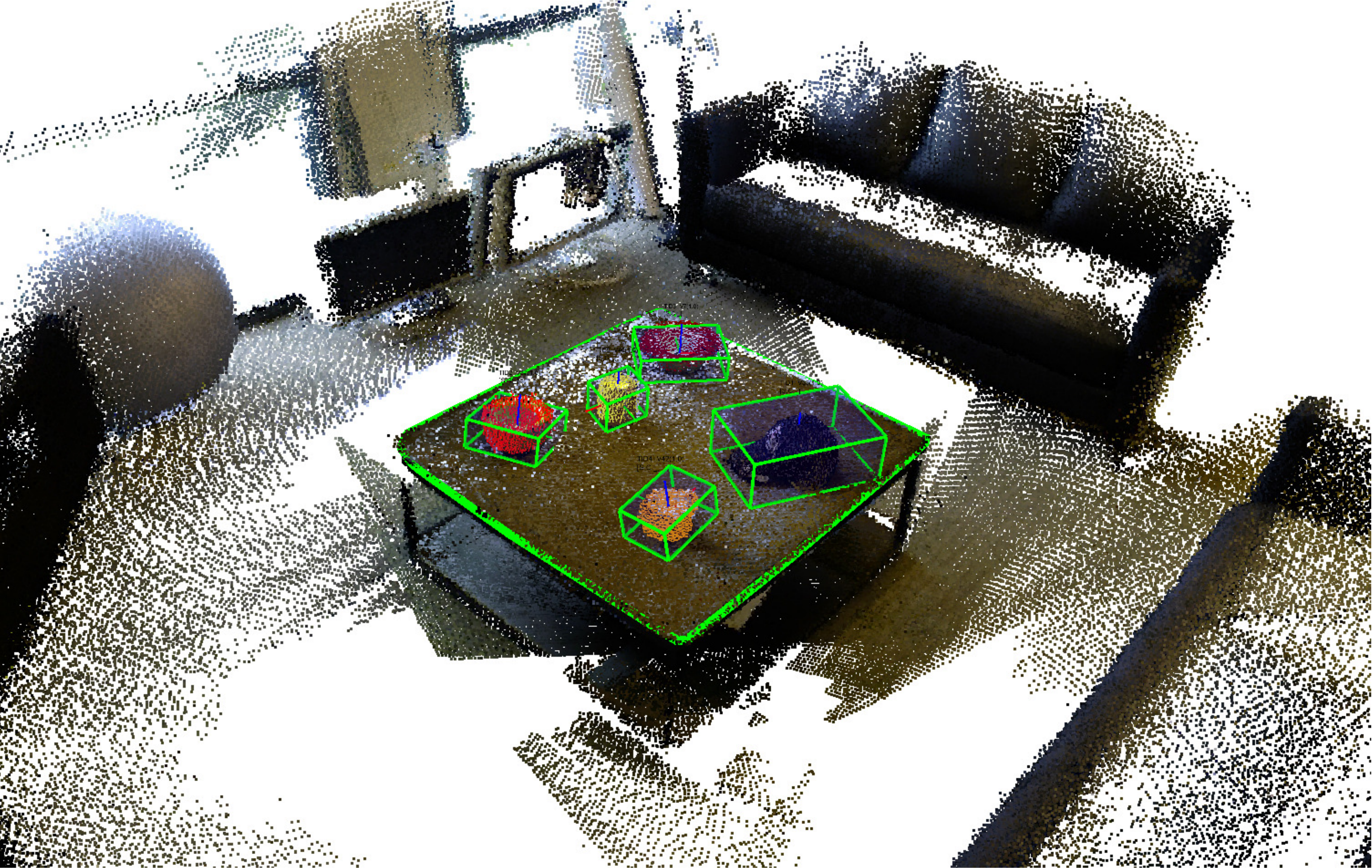}\\
\includegraphics[width=0.3\textwidth,trim= 0cm 0cm 0cm 0cm,clip=true]{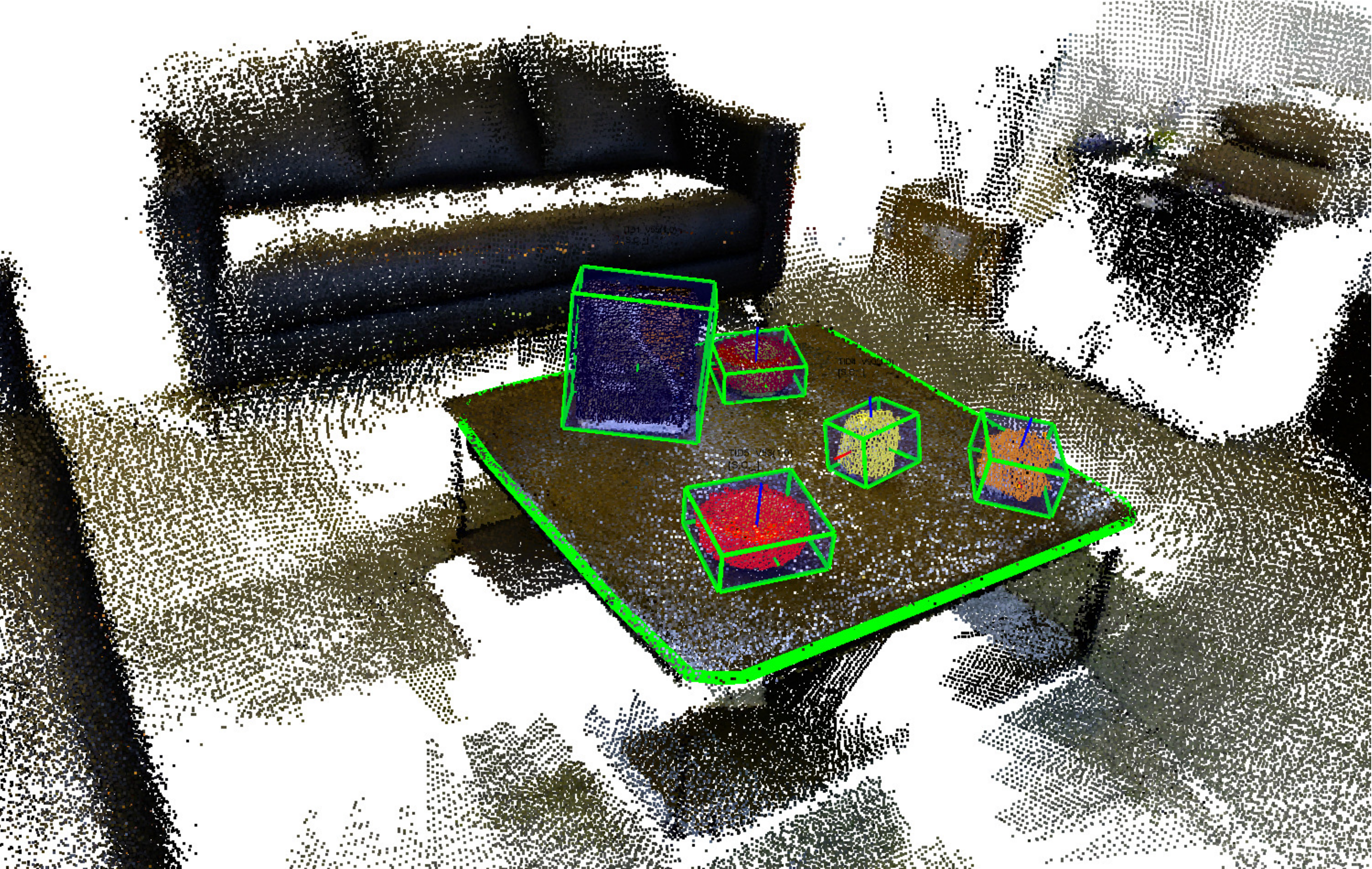}&
\includegraphics[width=0.3\textwidth,trim= 0cm 0cm 0cm 0cm,clip=true]{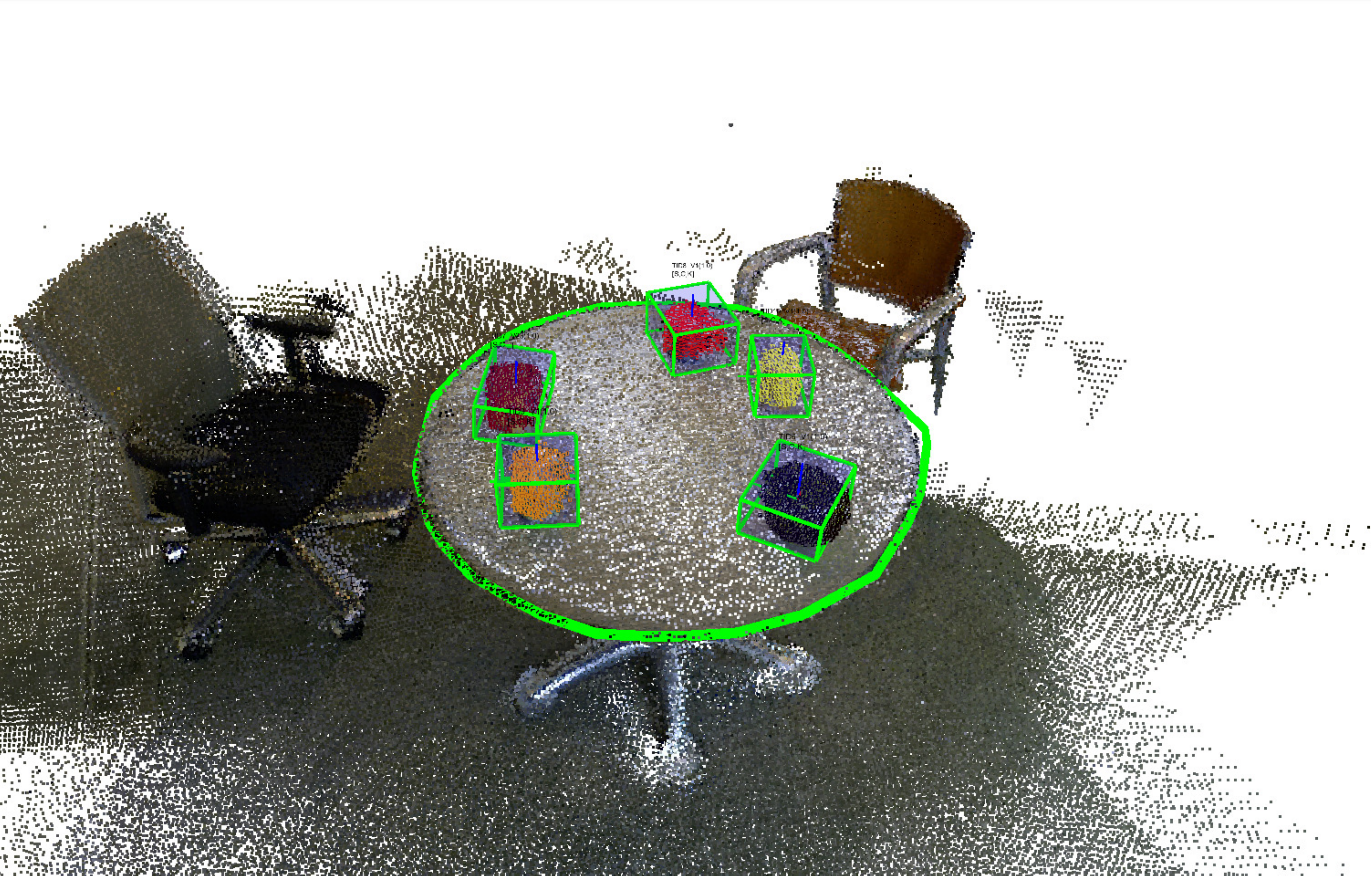}&
\includegraphics[width=0.3\textwidth,trim= 0cm 0cm 0cm 0cm,clip=true]{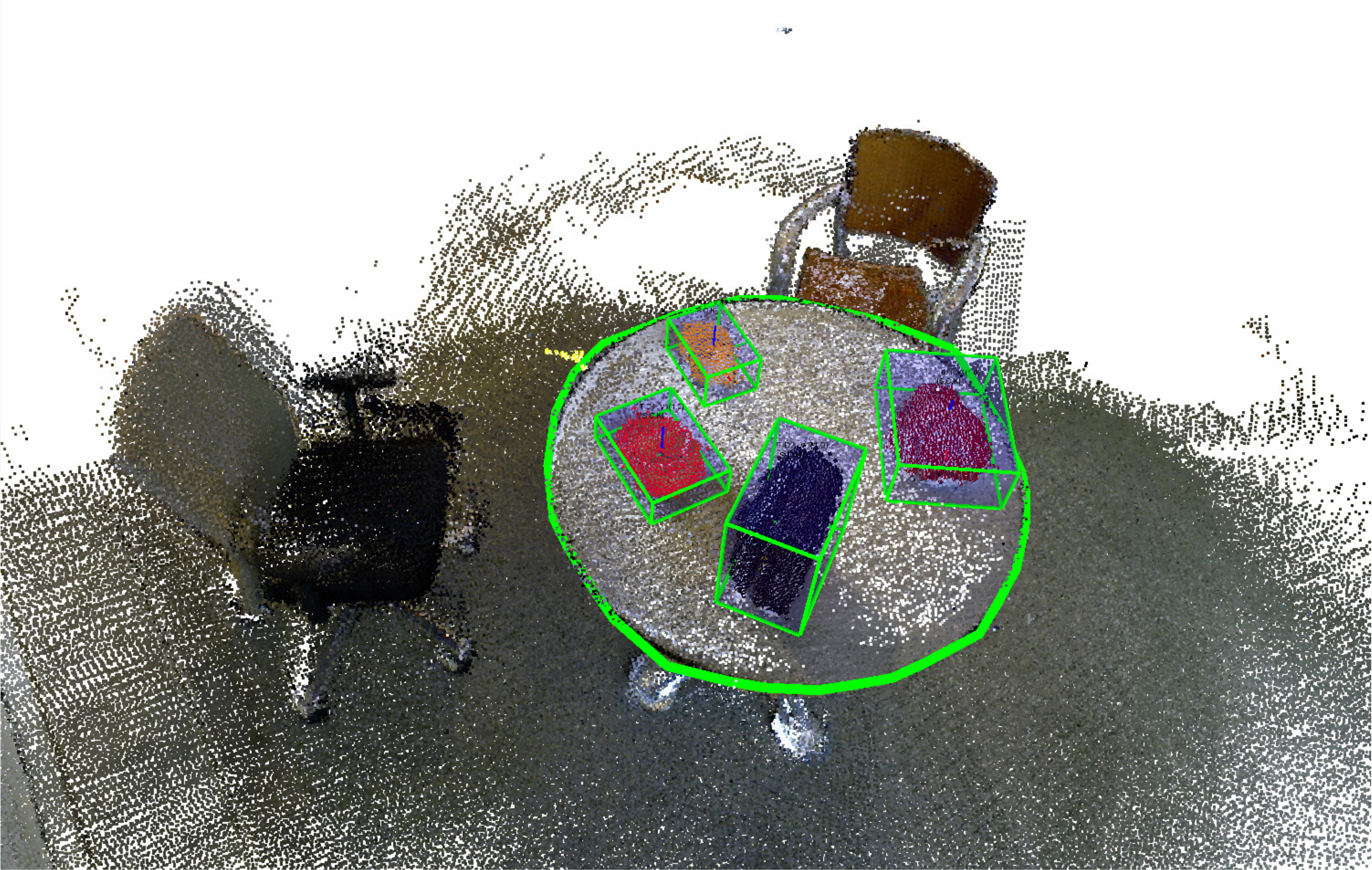}\\
\includegraphics[width=0.3\textwidth,trim= 0cm 0cm 0cm 0cm,clip=true]{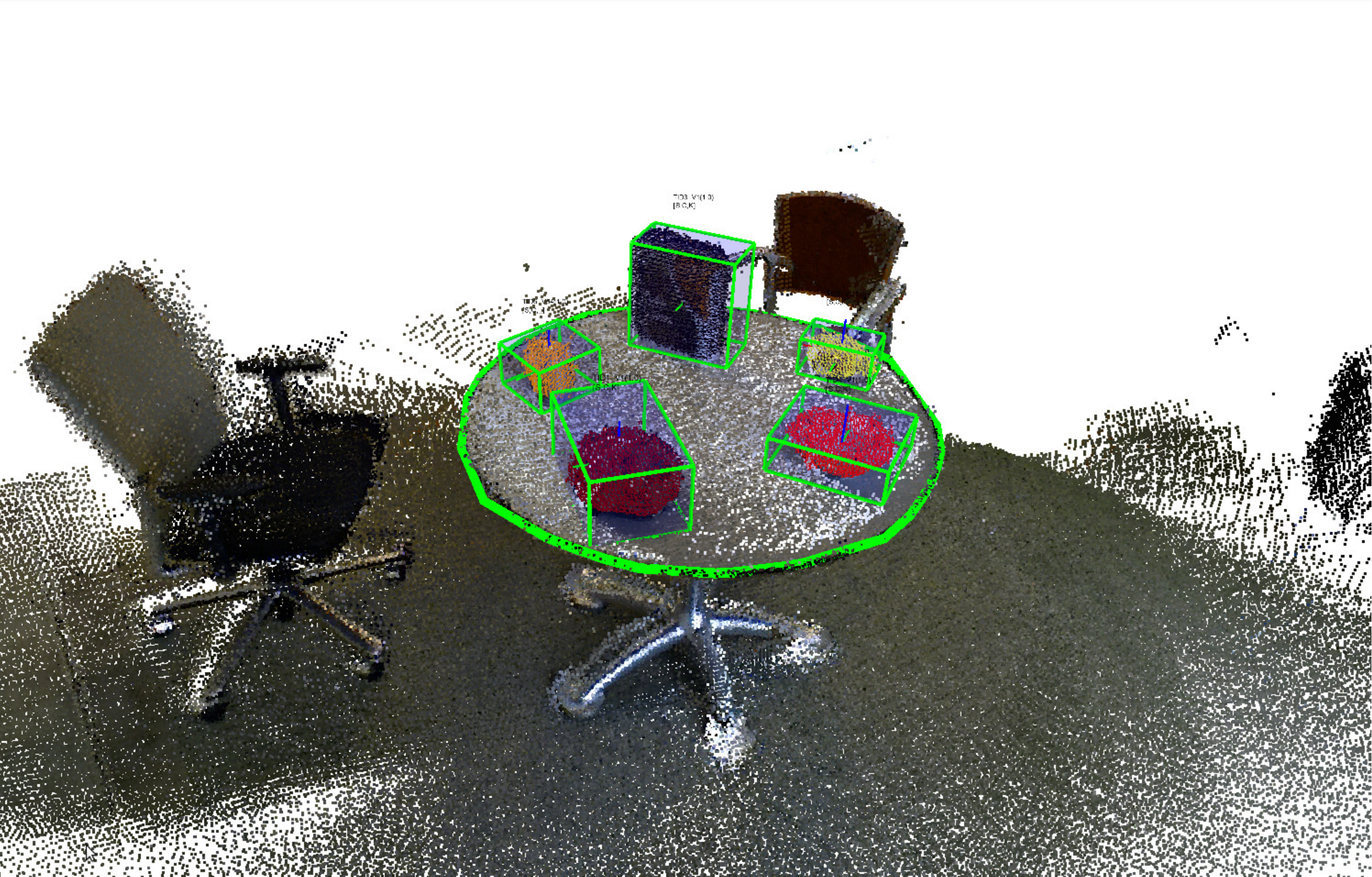}&
\includegraphics[width=0.3\textwidth,trim= 0cm 0cm 0cm 0cm,clip=true]{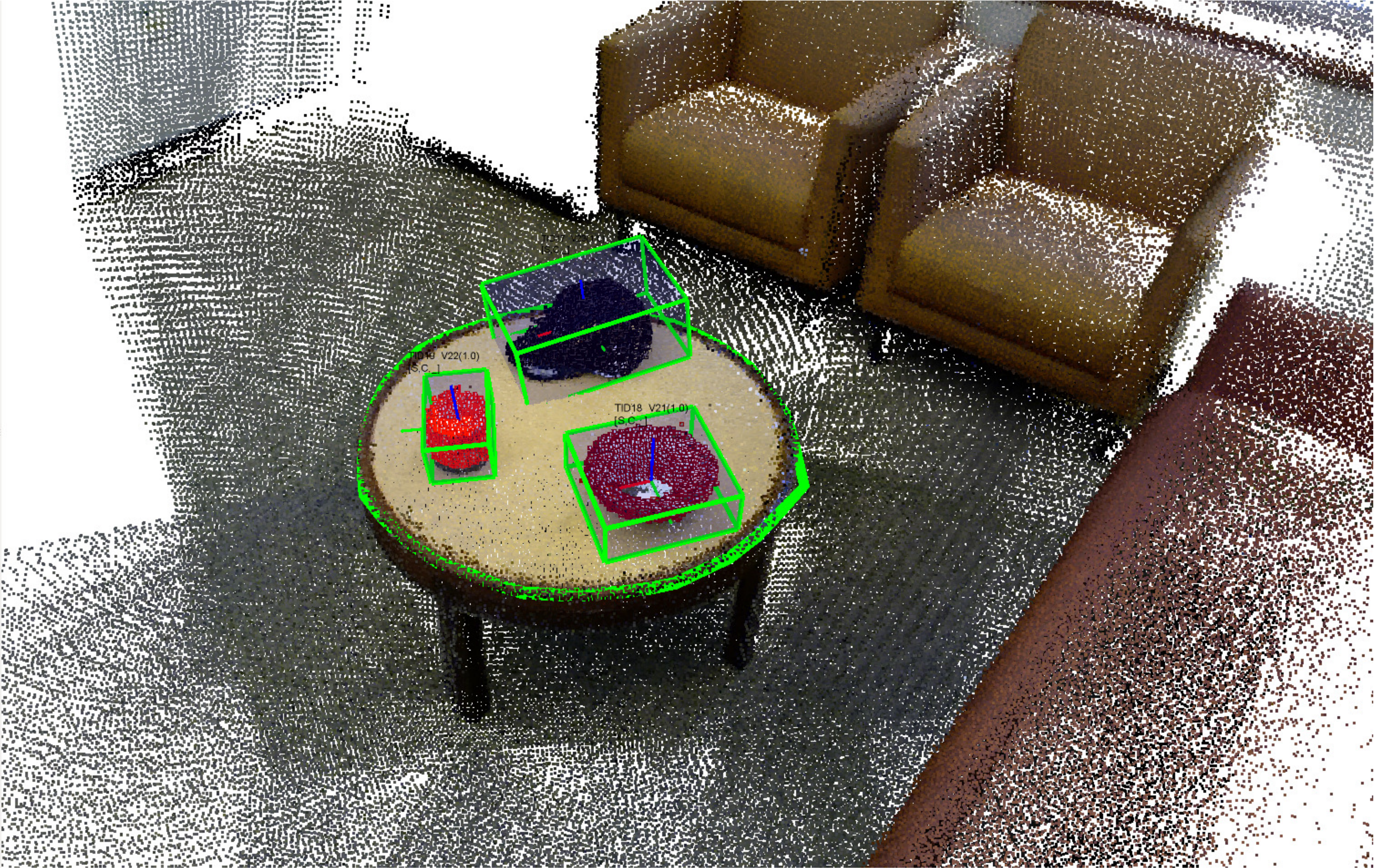}&
\includegraphics[width=0.3\textwidth,trim= 0cm 0cm 0cm 0cm,clip=true]{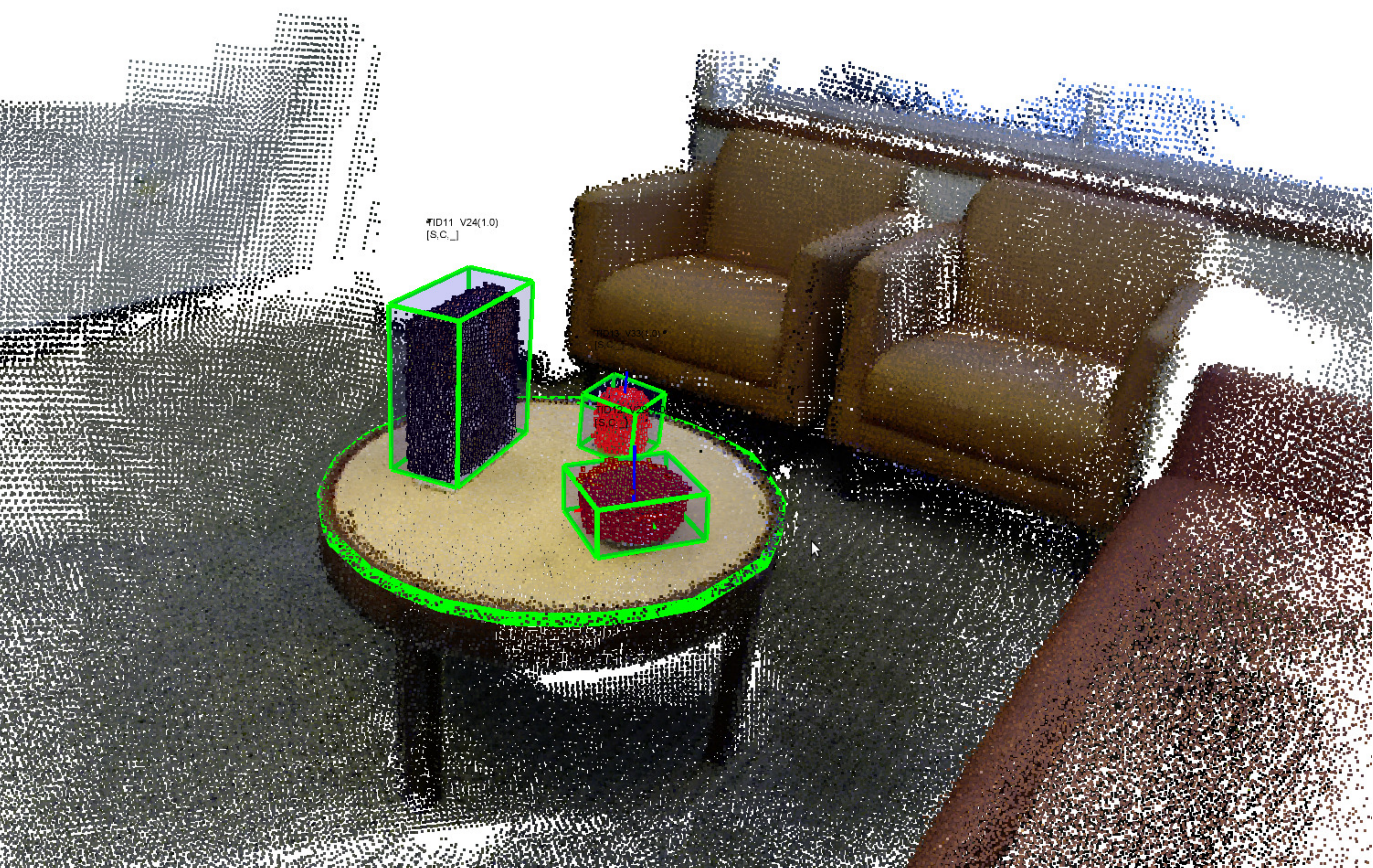}\\
\includegraphics[width=0.3\textwidth,trim= 0cm 0cm 0cm 0cm,clip=true]{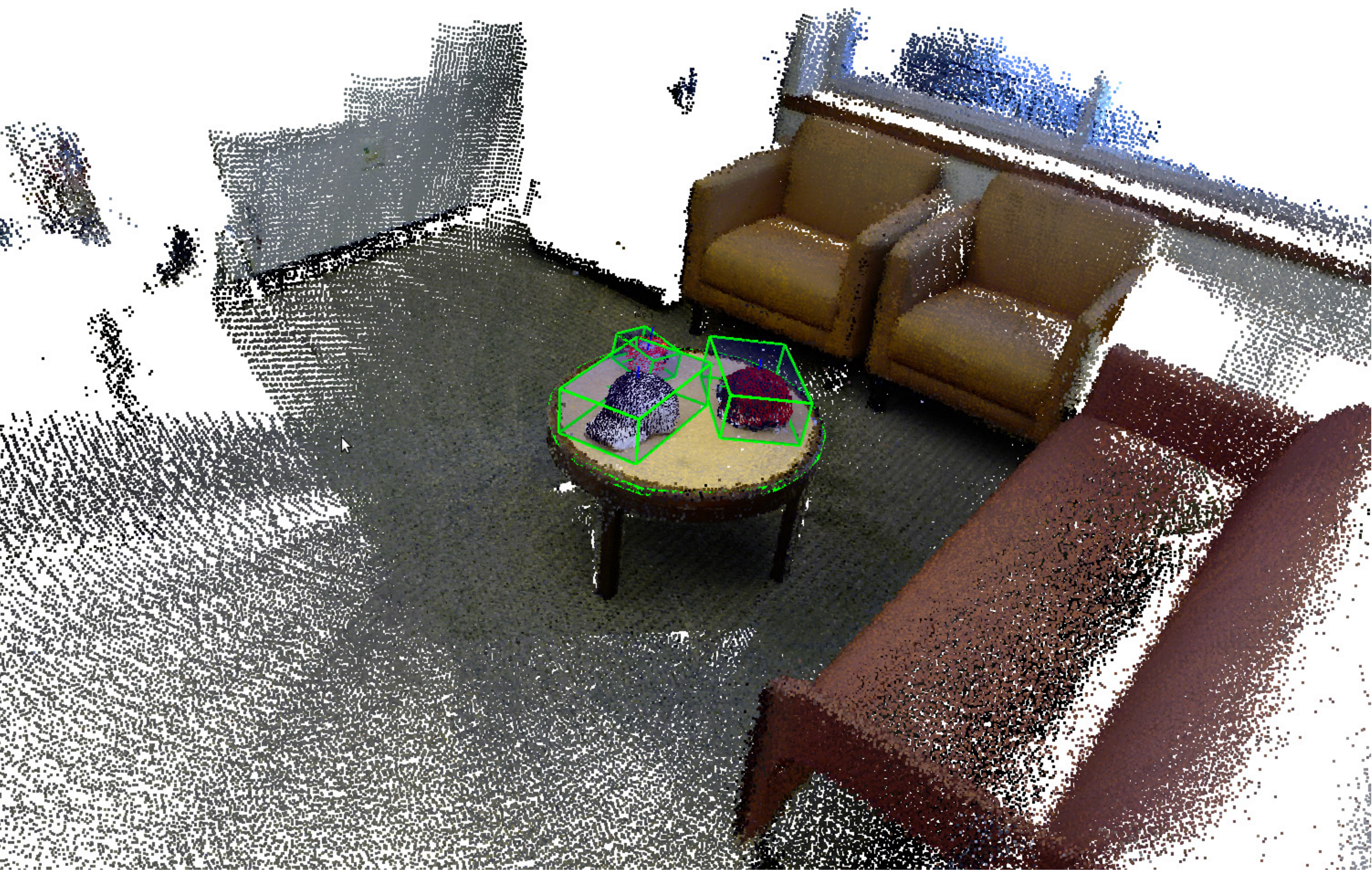}&
\includegraphics[width=0.3\textwidth,trim= 0cm 0cm 0cm 0cm,clip=true]{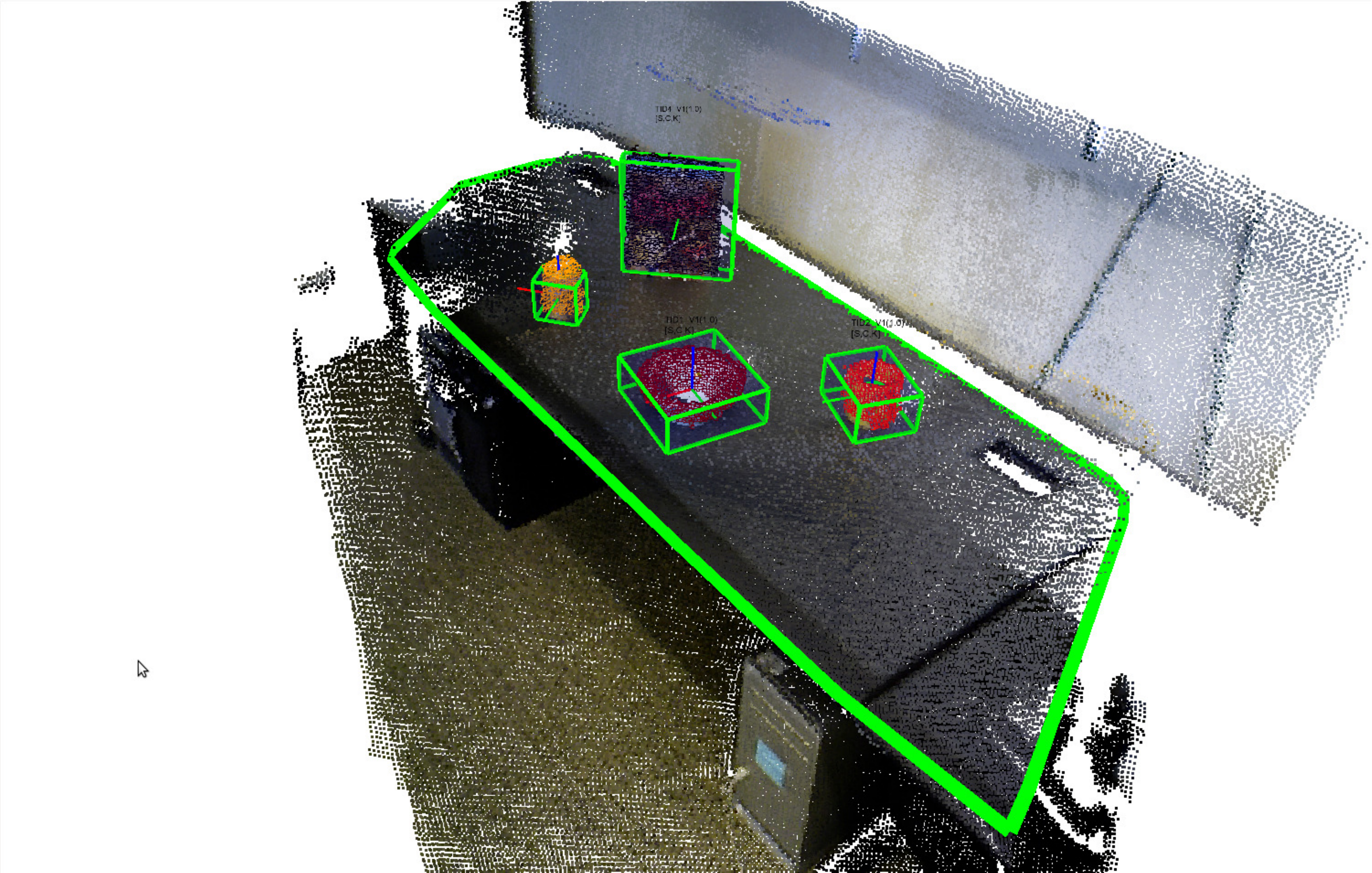}&
\includegraphics[width=0.3\textwidth,trim= 0cm 0cm 0cm 0cm,clip=true]{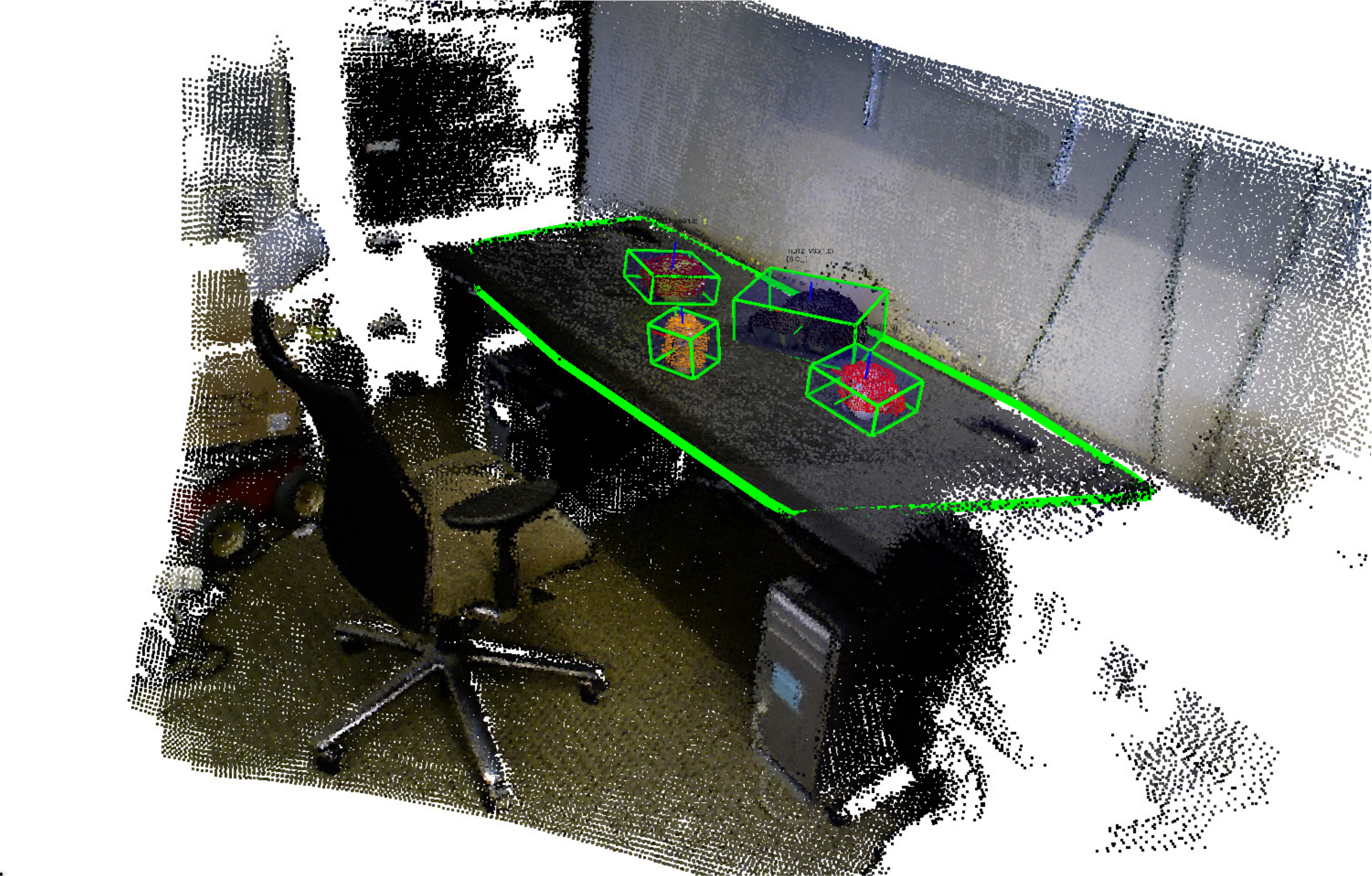}\\
\end{tabular}
% figure caption is below the figure
\caption{Qualitative results on Washington RGB-D Scenes Dataset v2.}
\label{fig:washington_scenes}       % Give a unique label
\end{figure}
In case of multiple objects touching each other (e.g., in a pile of objects or a messy dinning table), the Euclidean cluster extraction algorithm is not enough to appropriately detect object candidates  and further processing is required (see Fig.~\ref{fig:pile}). It is a challenging task to robustly detect multiple objects stacked in a pile or in a box, which are often found in a domestic environment.  In our current setup, a hierarchical clustering procedure is presented to segment the extracted point cloud using geometric, surface normal data and color. A region of the given point cloud is considered as an object candidate whenever points inside the region are continuous in both the orientation of surface normals and the depth values. The depth continuity between every point and its neighbors is computed. If the distance between points is lower than a threshold, then the two points belong to the same region. A region growing segmentation algorithm\footnote{http://pointclouds.org/documentation/tutorials/region\_growing\_segmentation.php} is also applied on medium-size hypotheses. The purpose of this algorithm is to merge the points that are close enough in terms of the smoothness and color constraints. 

Each cluster of points will be treated as an object candidate namely $\textbf{O}_i$, where $i \in \{1 , ~\dots~ , K\}$. It should be noted that the number of clusters, $K$, is not pre-defined and varies for different viewpoints. Fig.~\ref{fig:two_complex_scene} illustrates the results of the segmentation process in four different pile scenarios.

As discussed in the previous chapter, the \emph{Object Detection} module assigns a new \emph{TrackID} to each newly detected object and launches an object perception pipeline for the object. Finally, the object detection pushes the segmented object candidate into the respective pipeline for subsequent processing steps. The \emph{Object Tracking} module is responsible for keeping track of the target object over time while it remains visible. The object tracker works based on a particle filter \citep{Schulz,Hertzberg2014projrep} which uses geometric information as well as color and surface normal data to predict the next probable pose of the object. It receives the point cloud of the detected object and computes an oriented bounding box aligned with the point cloud's principal axes. The center of the bounding box is considered as the position of the object. The module sends out the tracked object information to the \emph{Feature Extraction} module. The detail of object's reference frame construction and feature extraction will be presented in the next chapter. We qualitatively report the performance of object detection on different scene datasets including Washington RGBD Scenes Dataset v2 \citep{lai2014unsupervised} and Imperial College London Bin-Picking Datasets \citep{Doumanoglou2016}. Washington RGB-D Scenes Dataset v2 dataset consists of $14$ scenes containing a subset of the objects in the RGB-D Object Dataset. As depicted in Fig.~\ref{fig:washington_scenes}, all objects have been correctly detected in all scenes.

%^^^^^^^^^^^^^^^^^^^^^^^^^^^^^^^^^^^^^^^^^^^^^^^^^^^^^^^^^^^^^^^^
%^^^^^^^^^^^^^^^^^^^^^^^^^^^^^^^^^^^^^^^^^^^^^^^^^^^^^^^^^^^^^^^^
\section { Unsupervised Experience Gathering}
\label{sec:unsupervisedExperienceGathering}

Gathering object experiences by exploration has the advantage of not requiring any human annotation of individual objects. Non goal-directed exploration provides chances to discover new objects. In general, object exploration is a challenging task because of the dynamic nature of the world and ill-definition of the objects \citep {collet2014herbdisc}. 

\begin{table}[!b]
\centering
\vspace{-5mm}
\caption{List of used constraints with a short description for each one.}
\resizebox{\columnwidth}{!}{
\begin{tabular}[l]{|l| l| c|}
%\\[-9.5pt]\hline\\[-6.5pt]
\hline
\textbf{Constraints} &  \textbf{Description} & \textbf{Section}\\
\hline
~ $C_{\operatorname{table}}$: \emph{``is this candidate on a table?''} & The target object candidate is placed on top of a table. & \ref{sec:objectDetection}
\\
\hline
              
~ $C_{\operatorname{track}}$: \emph{``is this candidate being tracked?''}  & Storing all object views while the object is static would lead to unnecessary  & \\ & accumulation of highly redundant data. This  constraint is used to infer that & \ref{sec:unsupervisedExperienceGathering}\\ & the segmented object is already being tracked or not. &
\\
\hline

~ $C_{\operatorname{size}}$: \emph{``is this candidate manipulatable?''} & Reject large object candidate %that its width does not fit into the gripper
&\ref{sec:objectDetection}\\
\hline

~ $C_{\operatorname{instructor}}$: \emph{``is this candidate part of the instructor's body? } & Reject candidates that are belong to the user's body & \ref{sec:preprocessing}\\
\hline

~ $C_{\operatorname{robot}}$: \emph{``is this candidate part of the robot's body?''} & Reject candidates that are belong to the robot's body
&\ref{sec:preprocessing}\\
\hline

~ $C_{\operatorname{edge}}$: \emph{``is this candidate near to the edge of the table?''} & Reject candidates that are near to the edge of the table &\ref{sec:unsupervisedExperienceGathering}\\
\hline
~ $C_{\operatorname{key\_view}}$: \emph{``is this candidate a key view?''} & For representing an object, only object views that are marked as 
 key-views &\\ & are stored in the database. An object view is selected as a key view whenever & \\ & the tracking of an object is initialized, or when it becomes static again after & \ref{sec:unsupervisedExperienceGathering}\\ & being moved. In case the hands are detected near the object, storing key views &\\ & is postponed until the hands are withdrawn. &
\\
\hline
\end{tabular}
\label{tbl:constraints}}
\end{table}

Since a system of boolean equations can represent any expression or any algorithm, it is particularly well suited for encoding the world and object candidates. Similar to Collet's work \citep{collet2014herbdisc}, we use boolean algebra. A set of boolean constraints, C, was then defined based on which boolean expressions, $\psi$, were established to encode object candidates for the process of constructing a pool of object candidates as well as for interactive object category learning and recognition (see Table~\ref{tbl:constraints}). The definition of \emph{``object''} in the stage of unsupervised experience gathering (i.e., exploration stage) is more general than in the normal operation stage (see equations \ref{boolean_expression_object_exploration} and \ref{boolean_expression_object_detection}). 
In both cases, we assume that interesting objects are on horizontal planar surfaces (i.e., tables) and the robot seeks to detect tabletop objects (i.e., $C_{\operatorname{table}}$). 
Due to memory size concerns, a representation of an object should only contain distinctive views. Key object views are selected by the \emph{Object Tracking} module.
A view which is different from the current view may appear after the object is moved (i.e., the pose of the object relative to the sensor changes). An object view is selected as a key view (i.e., $C_{key\_view}$) whenever the tracking of an object is initialized ($C_{\operatorname{track}}$), or when it becomes static again after being moved and the user's hands are far away from the object \citep{GiHyunLim}. Therefore, the $C_{key\_view}$ constraint is used to optimize memory usage and computation while keeping potentially relevant and distinctive information. Moreover, $C_{\operatorname{instructor}}~$ and $~C_{\operatorname{robot}}$ are used to filter out object candidates which are part of the instructor's body or robot's body. Accordingly, the resulting object candidates are less noisy and include only data corresponding to the environment: 
\begin{figure}[!t]
 \vspace {-1mm}
 \center 
 \includegraphics[width = 0.9\linewidth]{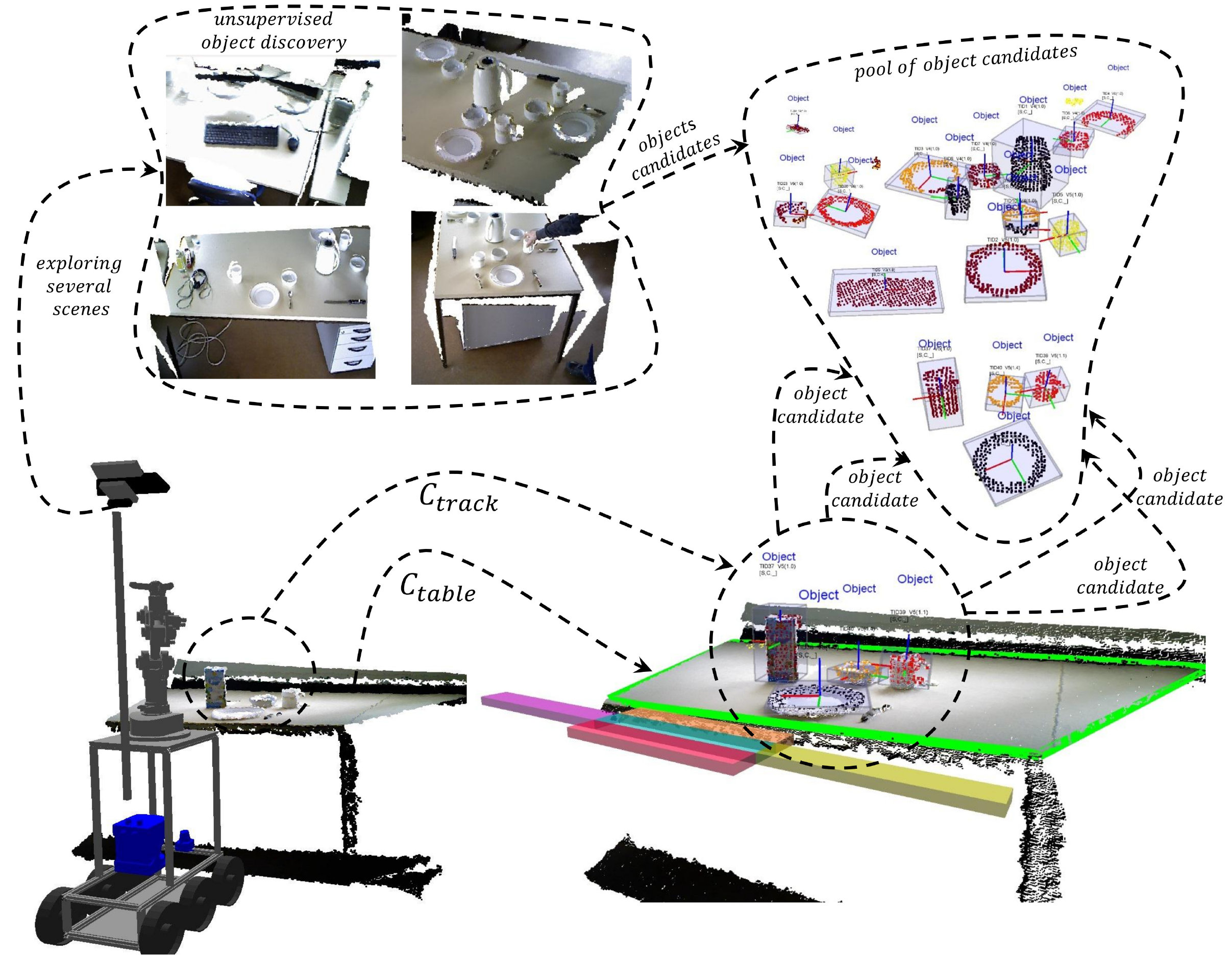}
% figure caption is below the figure
\vspace{-2mm}
\caption{Unsupervised experience gathering: (\emph{left}) the robot moves through an office to extract tabletop objects; (\emph{right}) the captured scenes are processed to produce a pool of object candidates.}

\label{fig:dictionary}       % Give a unique label
\end{figure}
\begin{equation}
\psi_{\operatorname{exploration}} = C_{\operatorname{table}}~\wedge~C_{\operatorname{track}}~\wedge ~C_{\operatorname{key\_view}}~\wedge~ \neg~(C_{\operatorname{instructor}}~\vee~C_{\operatorname{robot}}),
	\label {boolean_expression_object_exploration}
\end{equation}
\vspace{-3mm}
\begin{equation}
	\begin {split}
\psi_{\operatorname{detection}} = C_{\operatorname{table}}~\wedge ~C_{\operatorname{track}}~\wedge~C_{\operatorname{size}}~\wedge~ \neg~(C_{\operatorname{instructor}}~\vee~C_{\operatorname{robot}}~\vee~C_{\operatorname{edge}}),
	\end {split}
	\label {boolean_expression_object_detection}
\end{equation}

\noindent The object detection expression uses a size constraint, $C_{\operatorname{size}}$, to detect objects which can be manipulated by the robot. Moreover, a $C_{\operatorname{edge}}$ constraint is considered to filter out the segmented point clouds that are too close to the edge of the table due to the safety concern. 

For unsupervised experience gathering, every cluster that satisfies the exploration expression, $\psi_{\operatorname{exploration}}$, is selected. The output of object exploration is a pool of object candidates that can be used for the process of constructing the dictionary of visual words. The subject of visual word dictionary construction will be discussed in detail in the next chapter. In the context of the RACE project \citep{Hertzberg2014projrep}, the University of Osnabruck provided us with a rosbag collected by one of their robots while exploring an office environment. The exploration stage was run on this rosbag. A video of this exploration is available at: {\small \href{http://youtu.be/MwX3J6aoAX0}{\cblue{http://youtu.be/MwX3J6aoAX0}}}. 

%^^^^^^^^^^^^^^^^^^^^^^^^^^^^^^^^^^^^^^^^^^^^^^^^^^^^^^^^^^^^^^^^
%^^^^^^^^^^^^^^^^^^^^^^^^^^^^^^^^^^^^^^^^^^^^^^^^^^^^^^^^^^^^^^^^
\section { Supervised Experience Gathering}
\label{sec:supervisedExperienceGathering}
Human-robot interaction is essential for supervised experience gathering, i.e., for humans to teach robots how to perform different tasks. Particularly, open-ended object category learning  will be faster and more robust if it is able to learn new categories using the feedback of human users. In this section, a user interface for supervised experience gathering is presented. The interface is used not only for teaching new object categories in situations where the robot encounters with new objects but also for providing corrective feedback in case there is a misclassification \citep {chauhan2013towards,GiHyunLim}. Moreover, interaction capabilities were developed to enable human users to instruct the robot to perform complex tasks such as \emph{clear\_table} and \emph{serve\_a\_meal} (out of the scope of this chapter).

\begin{figure}[!b]
\center
\begin{tabular}[width=1\textwidth]{c}
\begin{tabular}[width=1\textwidth]{cc}
\hspace{-20pt}
 \includegraphics[scale=1.25, trim=0.25cm 0.0cm 0cm 2.6cm,clip=true]{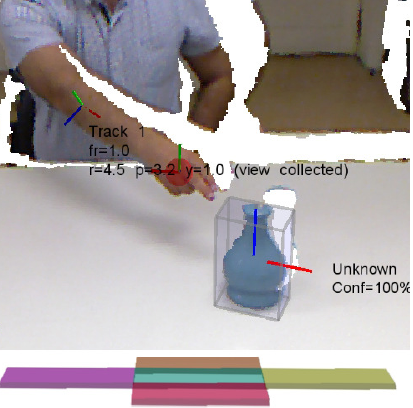} & \quad\quad\quad
 \includegraphics[scale=1.25, trim= 0.35cm 0.0cm 0cm 2.5cm,clip=true]{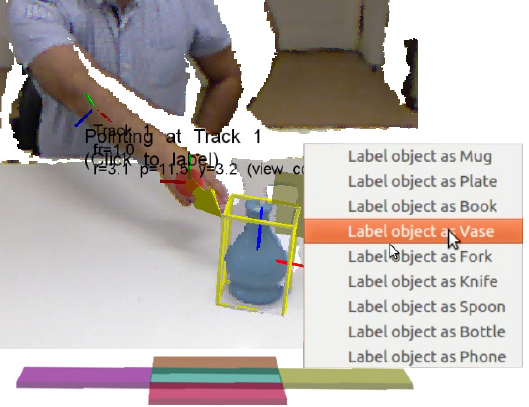}\\
 \emph{(a)} & \emph{(b)} \\
 \includegraphics[scale=0.25, trim= 1.cm 0.07cm 0cm 0cm,clip=true]{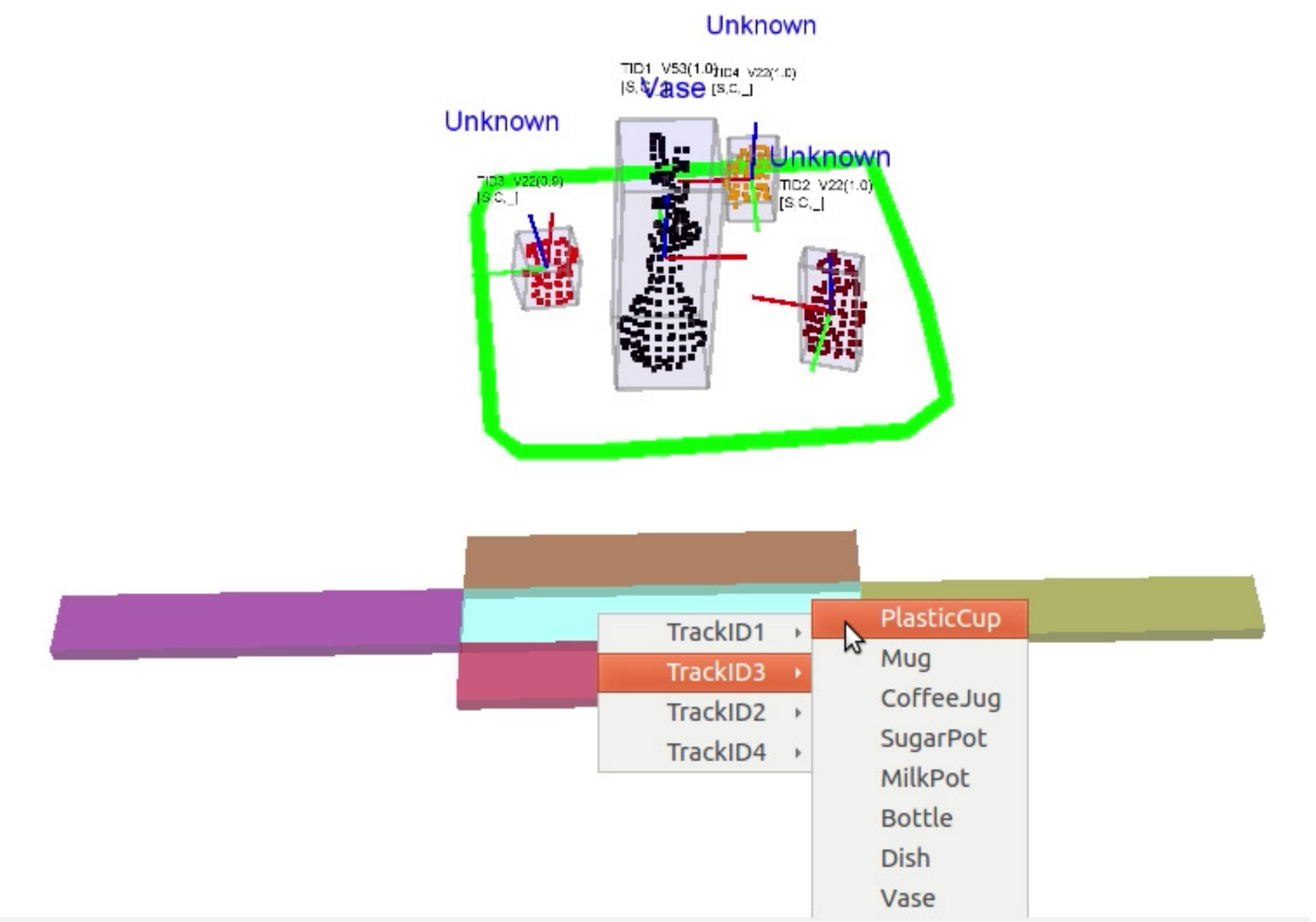}&
 \includegraphics[scale=0.25, trim= 1.0cm 0.07cm 0.0cm 0.0cm,clip=true]{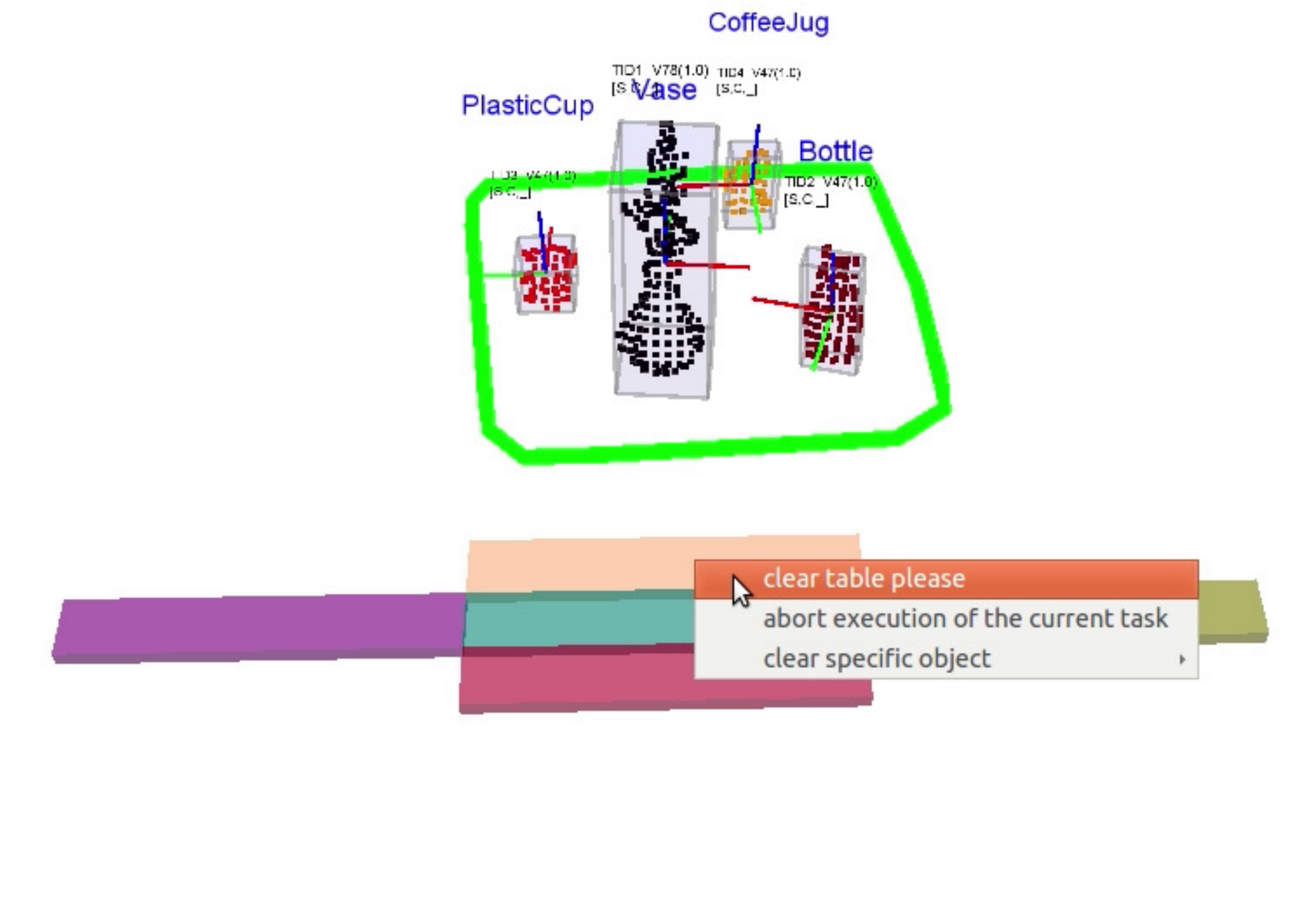}\\
 \emph{(c)} & \emph{(d)} \\
\end{tabular}\\
\end{tabular}
\caption{ A 3D visualization of an object labelling event: (\emph{a}) instructor pointing to an object; (\emph{b}) associating a label to the object that is currently being pointed; (\emph{c}) labelling object categories by associating a label to a TrackID; (\emph{d}) instructing the robot to perform the \emph{clear\_table} task;
}
\label{fig:user_interface}       % Give a unique label
\end{figure}
The \emph{User Interaction} module provides a graphical menu to facilitate the collection of supervised object experiences. Two alternative interactions mechanisms are supported: gesture recognition or the usage of a graphical menu interface. In the first case the instructor points to an object and then selects the desired label from a menu that appears in front of the table in the 3D visualization of the scene. The \emph{Skeleton Tracker} (ST) module tracks the user skeleton pose over time based on RGB-D data\footnote{http://wiki.ros.org/openni\_tracker}. The skeleton pose information is passed to the Gesture Recognition (GR) module, which computes a pointing direction. Currently, the pointing direction is assumed to be the direction of the right forearm. To produce a valid object category label, the pointing gesture must co-occur with the object labeling via the menu. The \emph{User Interaction} module contains a global state of the scene. Upon receiving category label input, it checks if the received pointing direction intersects the bounding box of any of the objects and decides if an object is being pointed at by the instructor. If that is the case, teaching instructions trigger perceptual learning to create and / or update object categories. 

An example of object labelling is depicted in Fig.~\ref{fig:user_interface}. The instructor puts a \emph{`Vase'} on the table. Tracking is initialized with TrackID 1. The gray bonding box signals the pose of the object as estimated by the tracker. TrackID 1 is classified as \emph{`Unknown'} because vases are not yet known to the system; the instructor points at the object. The system detects the pointing gesture and the corresponding menu is activated. The instructor labels the object as \emph{`Vase'}, which activates the \emph{Object Conceptualizer} (category learning) module.

As an alternative, the instructor can select the category label for an object based on its \emph{TrackID}. This capability is currently provided through interactive markers in RVIZ, a 3D visualization tool for ROS. In addition, the \emph{User Interaction} module provides a task menu that is used to instruct the robot to perform a task or to abort the current task. Further details on the supervised object experience gathering developed for the RACE project are available in \citep{GiHyunLim}. 

%^^^^^^^^^^^^^^^^^^^^^^^^^^^^^^^^^^^^^^^^^^^^^^^^^^^^^^^^^^^^^^^^
%^^^^^^^^^^^^^^^^^^^^^^^^^^^^^^^^^^^^^^^^^^^^^^^^^^^^^^^^^^^^^^^^
\section {Online Object Model Construction}
\label{model_construction}
In this section, we develop an approach to autonomously construct models of unknown objects. This capability is necessary for cognitive robots, since it will allow robots to actively investigate their environments and learn about objects in an unsupervised and incremental way. Online construction of full surface models of objects is not only useful for improving object recognition performance by collecting several views, but also can be used for manipulation purposes.

As shown in Fig.~\ref{fig:object_construction}, our approach enables a robot to move around an object and build an increasingly complete 3D model of the object by extracting object's points from different perspectives and aligning them together by considering the tracked object pose and robot pose as well as geometrical and visual information. In such scenario, tracking the target object is necessary since many objects in everyday environments exhibit rotational symmetries or lack a distinctive geometry for matching. As stated by \cite{krainin2010manipulator}, without pose information, Iterative Closest Point (ICP) based approaches are unable to recover the proper transformations because of the ambiguity in surface matching. Towards this goal, we employ a Kalman filter that uses depth and visual information for keeping track of robot motion and the target object while it remains visible over time. Afterwards, the extracted object views are united using an incremental ICP~\footnote{http://pointclouds.org/documentation/tutorials/pairwise\_incremental\_registration.php\#pairwise-incremental-registration} approach~\citep{pomerleau2013comparing} that incorporates both tracking and appearance information. 

It should be noted that this approach cannot provide information about regions of the target object that were not visible throughout the exploration procedure. Moreover, since the robot localization is out of the scope of this work, we use noisy ground truth information and showed that this approach can compensate the noise of robot motion and generate proper models of household objects. A video showing the robot exploring an environment for constructing a full model of an \emph{Amita juice box} is available at: {\href{https://youtu.be/CuBS2L2q5NU}{\cblue{https://youtu.be/CuBS2L2q5NU}}
\begin{figure}[!t]
\centering
\begin{centering}
\begin{tabular}{ccc}
\hspace{-1mm}
	\includegraphics[scale=0.12,trim= 0cm 1cm 0cm 0cm,clip=true]{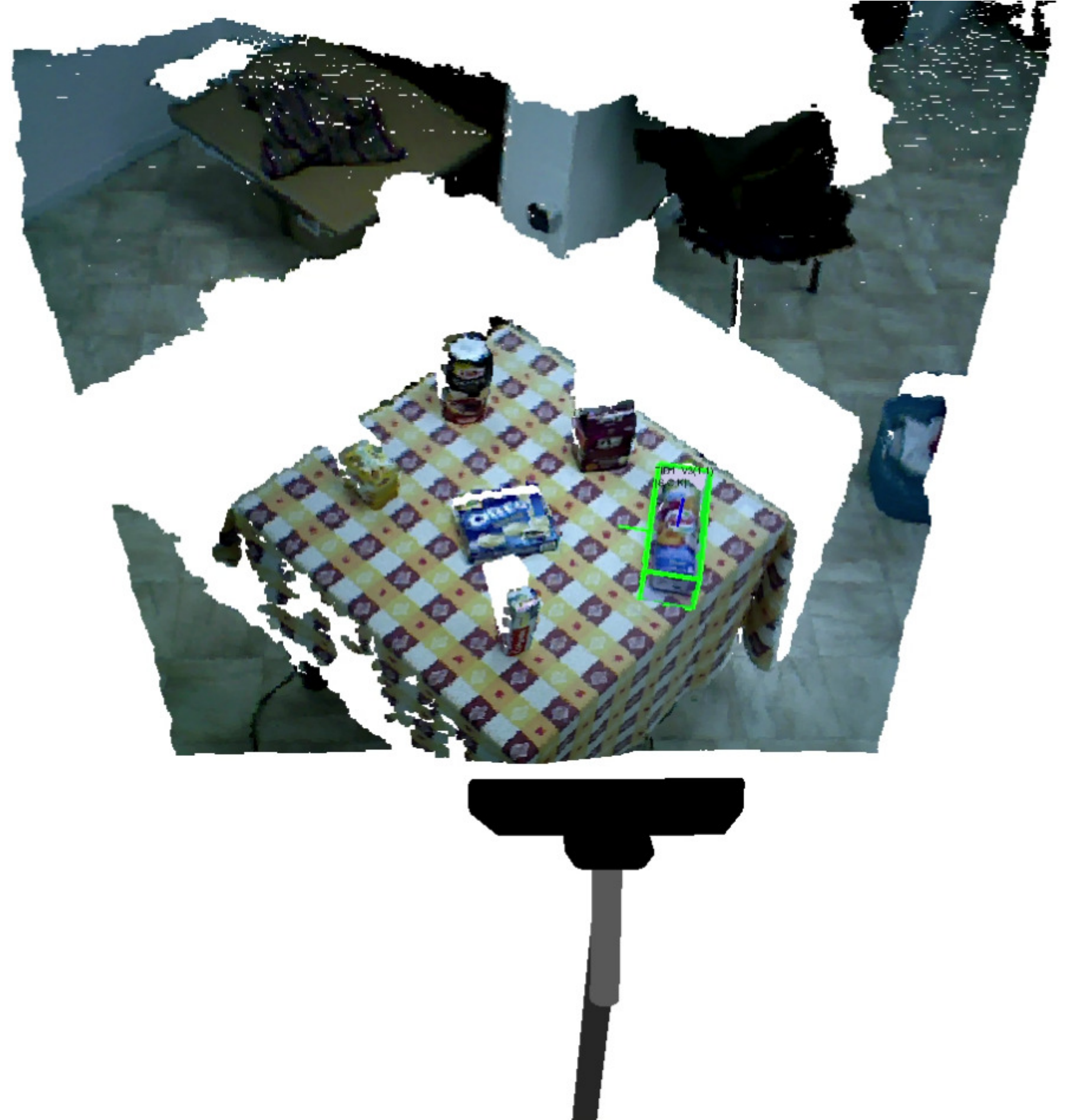}&
	\includegraphics[scale=0.12,trim= 0cm 1cm 0cm 0cm,clip=true]{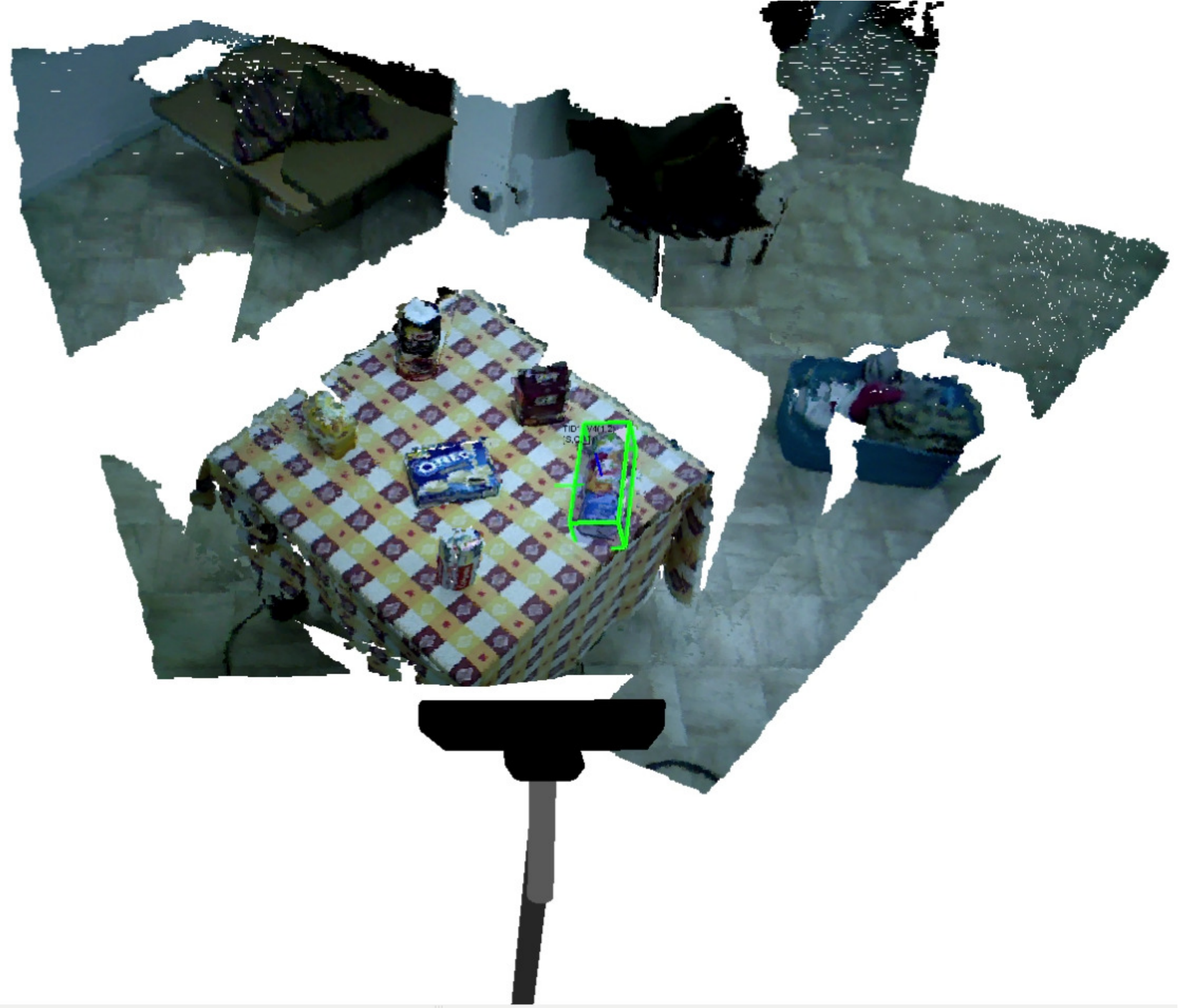}&
	\includegraphics[scale=0.12,trim= 0cm 1cm 3cm 0cm,clip=true]{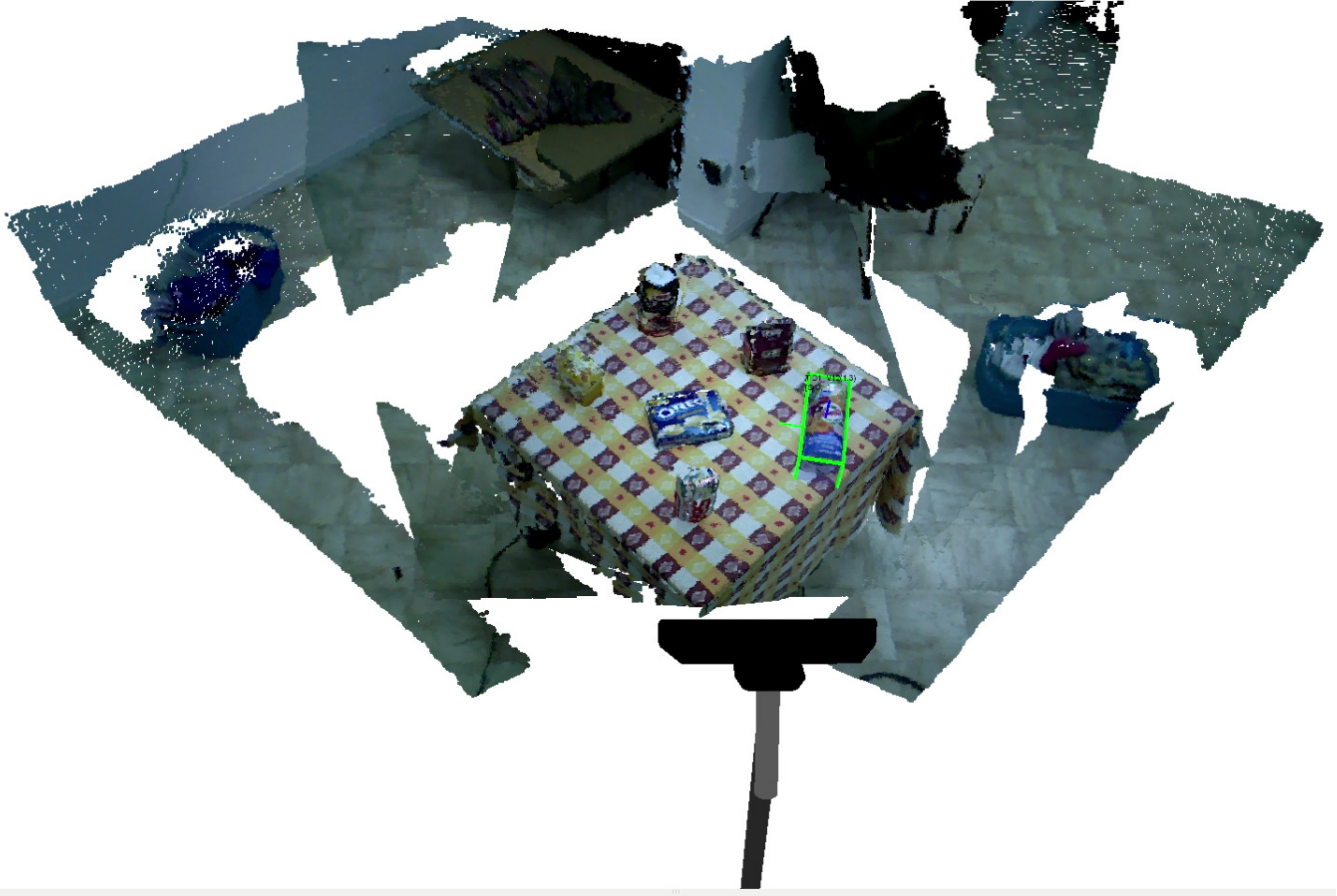}\\
\multicolumn{2}{c}{\includegraphics[scale=0.4,trim= 0cm 0cm 0cm 0cm,clip=true]{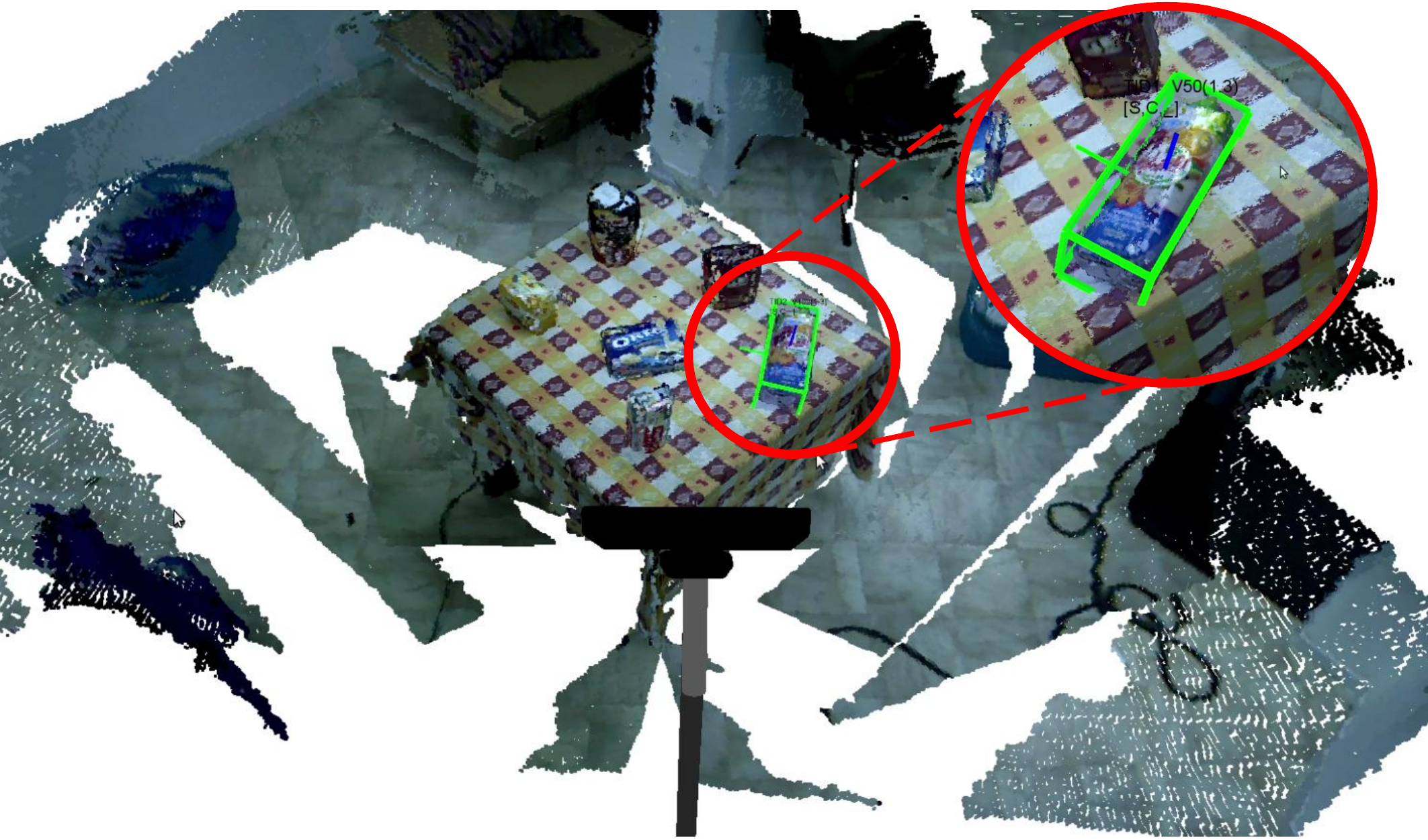}}&
	\includegraphics[scale=0.14,trim= 0cm 0cm 0cm 0cm,clip=true]{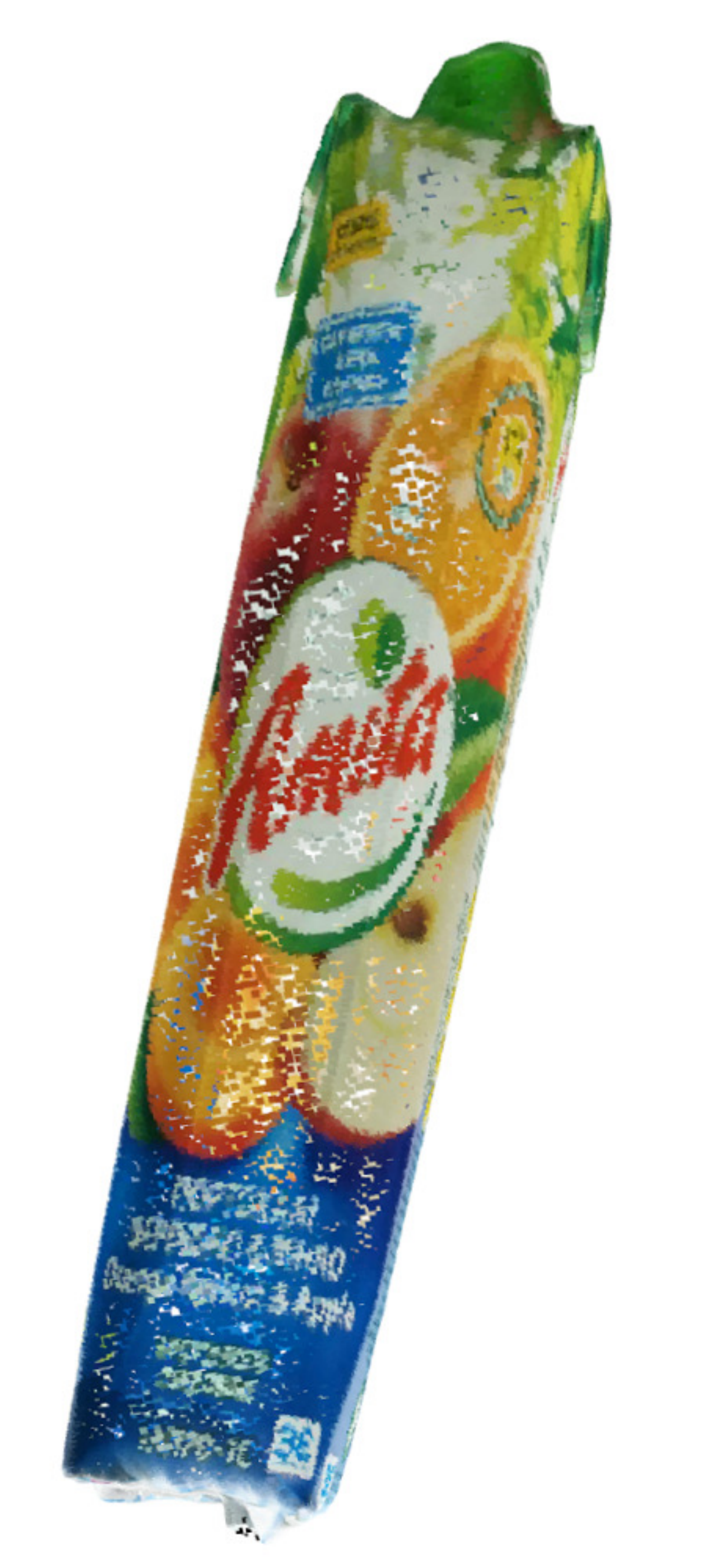}
	
\end{tabular}
\end{centering}
\caption{\small Constructing a full model of an \emph{Amita} object: (\emph{top row}): incremental construction of a 3D model for the tracked \emph{Amita} object by accumulating three viewpoints; (\emph{bottom row}): \emph{Amita} model constructed by merging 7 view points. The green bounding boxes and its reference frame showing the 6D pose of the object.}
\label{fig:object_construction}
\end{figure}

%^^^^^^^^^^^^^^^^^^^^^^^^^^^^^^^^^^^^^^^^^^^^^^^^^^^^^^^^^^^^^^^^
%^^^^^^^^^^^^^^^^^^^^^^^^^^^^^^^^^^^^^^^^^^^^^^^^^^^^^^^^^^^^^^^^
\section{Next-Best-View Prediction}
\label{section:NBV_prediction}

The ability to predict the Next-Best-View (NBV) point is important for mobile robots performing tasks in everyday environments. In active scenarios, whenever the robot fails to detect or manipulate objects from the current view point, it is able to predict the next best view position, goes there and captures a new scene to improve the knowledge of the environment. This may increase the object detection and manipulation performance (see Fig.~\ref{fig:nbv_fig}). Towards this end, we proposed an entropy-based NBV prediction algorithm by rendering the scene using the current object hypotheses\footnote{This work was done in collaboration with Imperial Computer Vision and Learning Lab at Imperial College London \citep{JuilICCVW2017}}.
\begin{figure}[!t]
\centering
\begin{centering}
\begin{tabular}{cc}
	\includegraphics[scale=0.32,trim= 1cm 0.15cm 2.3cm 0.5cm,clip=true]{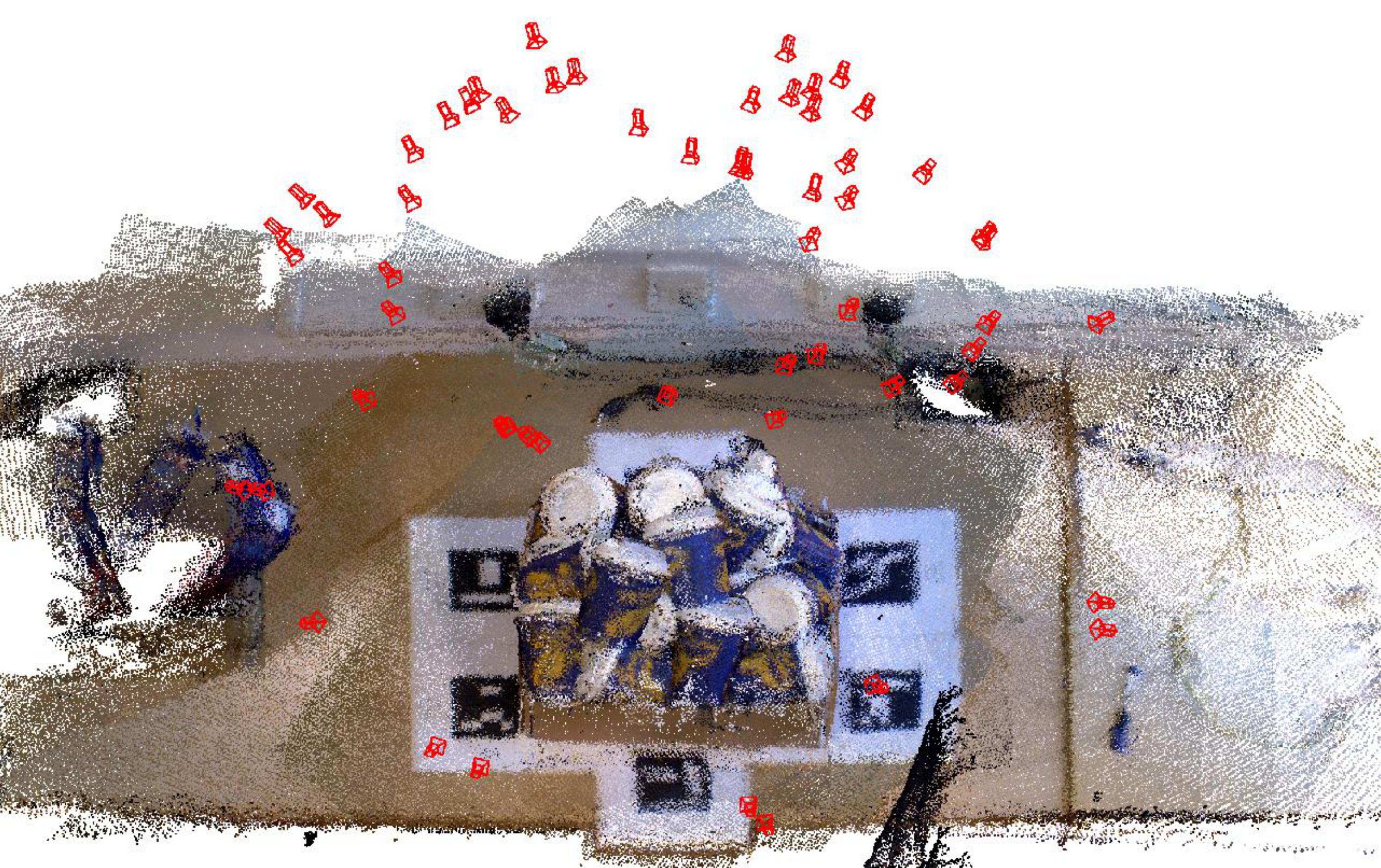} &\hspace{0.2cm}
		\includegraphics[scale=0.22,trim= 0cm 0cm 0cm 0cm,clip=true]{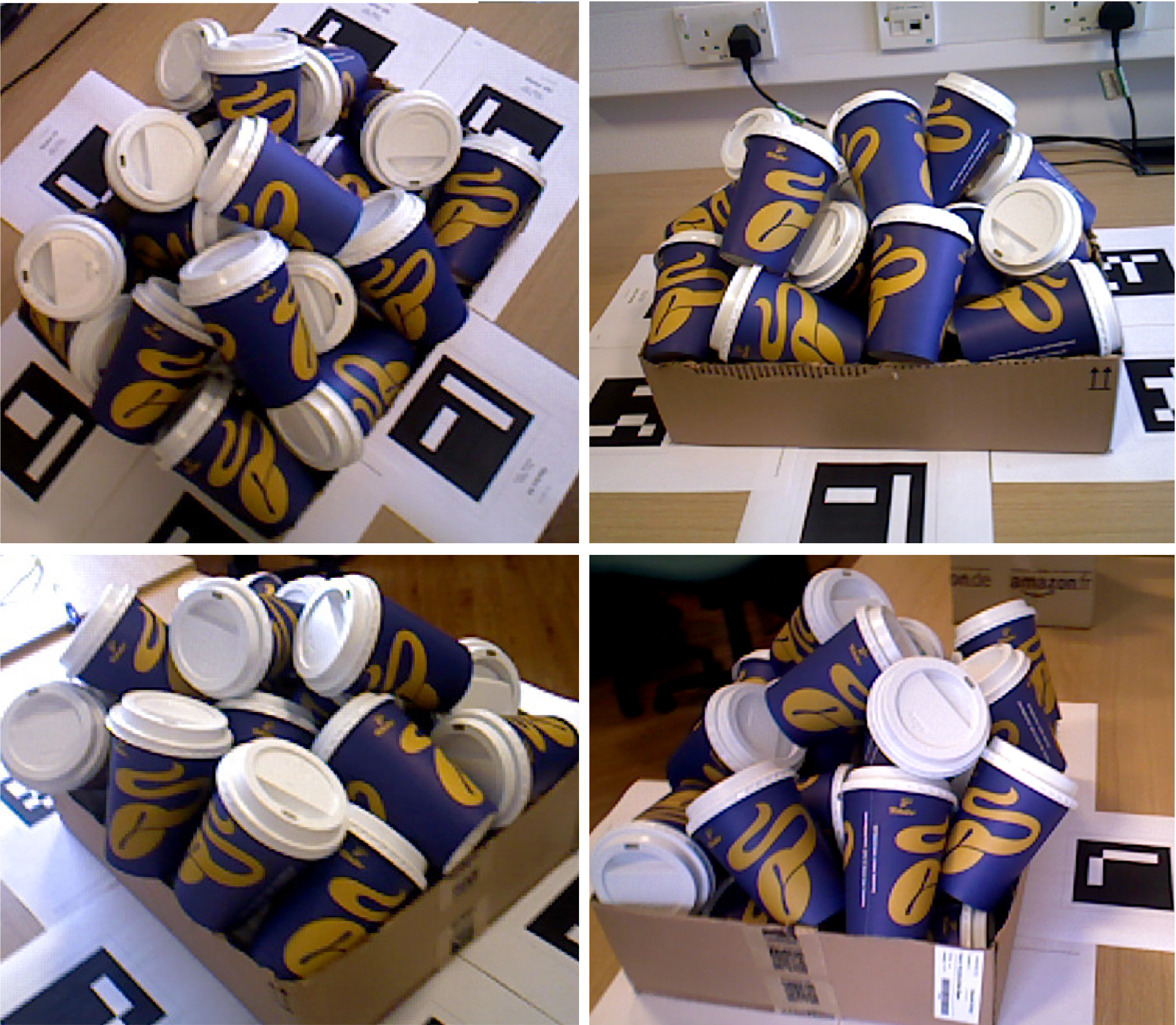}
\end{tabular}
\end{centering}
\vspace{-1mm}
\caption{(\emph{left}) Accumulated view and camera pose of different view of a pile of coffee cup scenario; (\emph{right}) images from four
different views. }
\label{fig:nbv_fig}
\end{figure}

In this step, the robot captures a point cloud of the scene and computes a list of object hypotheses containing both objects' 6D pose and recognized label (i.e., object recognition and pose estimation will be discussed in the next chapters). The inputs to the NBV prediction module are: 

\begin {itemize}
\item the constructed 3D models of the objects; 
\item the point cloud of the scene; 
\item a set of 6D object hypotheses $\textbf{P}=\{ h_1,\dots, h_n\}$; 
\item and a possible set of viewing poses, $\textbf{V} =\{ v_1,\dots, v_m\}$ where each viewing poses represents the camera rotation and translation in 3D space. 
\end {itemize}
The given scene is first segmented and the obtained clusters are then used to compute viewpoint entropy for the given scene. There are various methods for computing the viewpoint entropy. In general, the number of visible voxels or points is used as an indicator of area for entropy computation. This measure is not good enough since it only considers the coverage objective. We note the following: on one hand, observing a large portion of an object (a big cluster) helps the system to recognize the object better; on the other hand, wider scene coverage (visibility objective) causes the system to detect more object candidates. Therefore, we propose a new formulation for viewpoint entropy calculation that takes into account both the visibility (i.e., given by the number of visible points in the scene) and saliency (i.e., given by the size of each cluster; since observing a large portion of an object can potentially reduce the object recognition and pose estimation uncertainty) objectives.  The viewpoint entropy of a given scene is computed as follows:

\begin{equation}
\label{eq:entropy}
H = - \sum\limits_{i=1}^{K}\frac{A_{i}}{S}\log \frac{A_{i}}{S},
\end{equation}

\noindent where, $K$ is the number of clusters, $A_i$ is the area of the $i^{th}$ cluster and $S$ is the total area of the given scene.
Before actually moving the camera, we aim to predict the NBV from the camera pose list, $\textbf{V}$. For this purpose, first, we have to predict what can be observed from each pose in $\textbf{V}$ by taking a ``\emph{virtual point cloud}''. Towards this goal, based on the given set of 6D object hypothesises, the full model of the objects are first added to the current scene (see Fig.~\ref{fig:nbv} \emph{b}). Afterwards, for each possible camera pose, a virtual point cloud is rendered based on depth buffering and orthogonal projection methods (see Fig.~\ref{fig:nbv} \emph{c} and \emph{d}). Then, the viewpoint entropy is calculated for each rendered view as before.

\begin{figure}[!t]
\hspace{-4mm}
	\begin{tabular}{cccc}
	\includegraphics[width=0.21\linewidth]{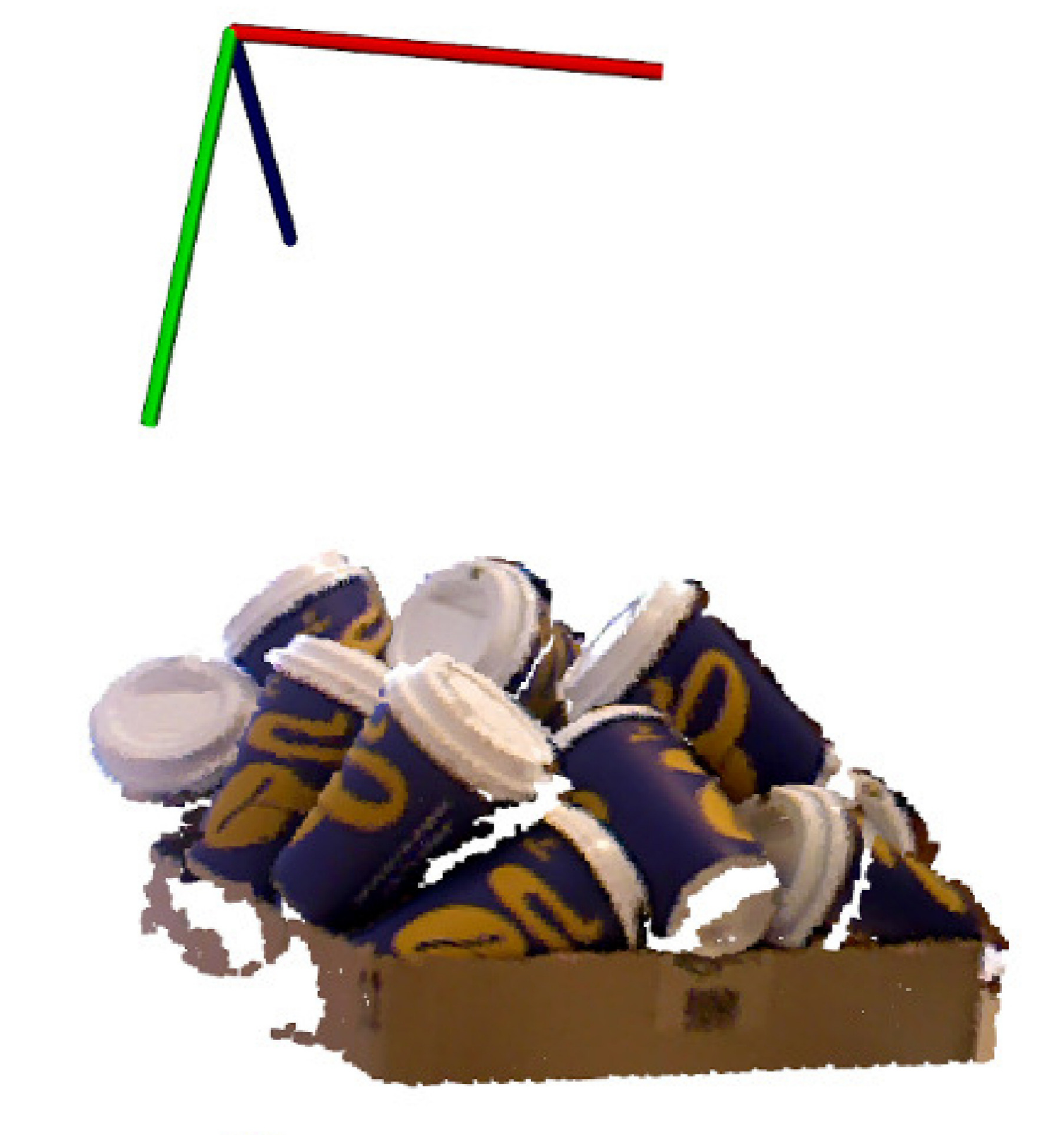} &
	\includegraphics[width=0.24\linewidth]{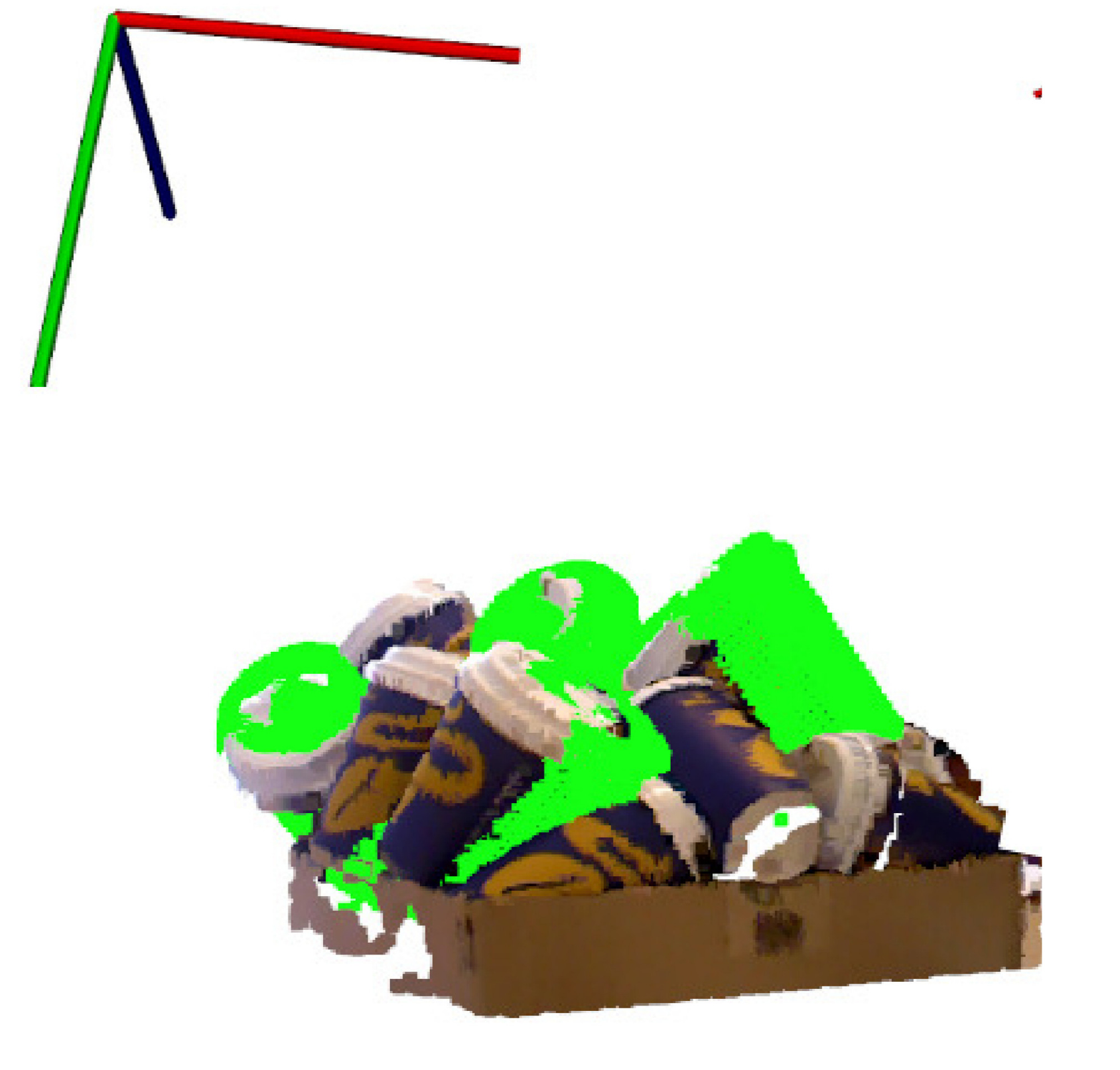} &
	\includegraphics[width=0.24\linewidth]{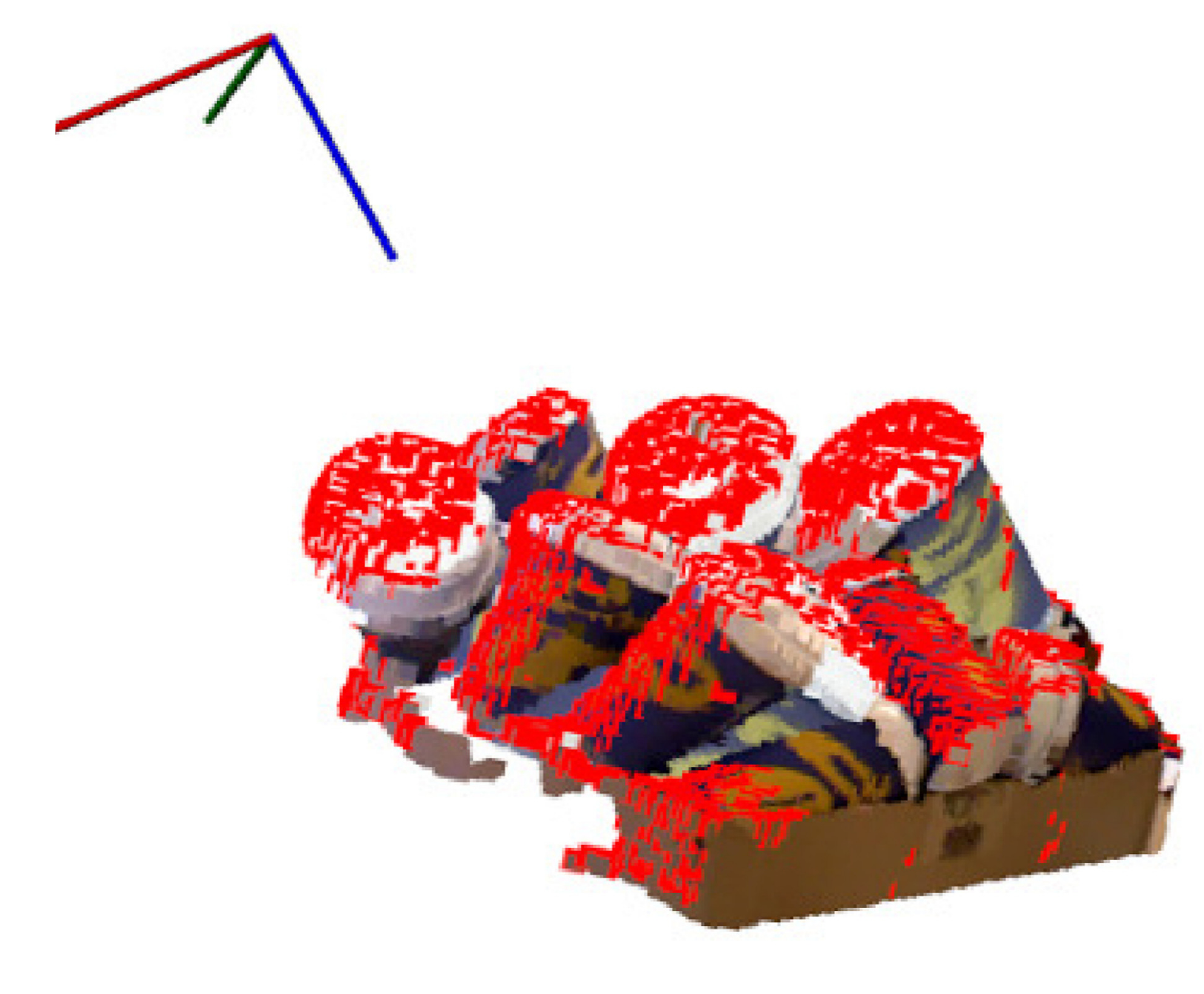} &
	\includegraphics[width=0.24\linewidth]{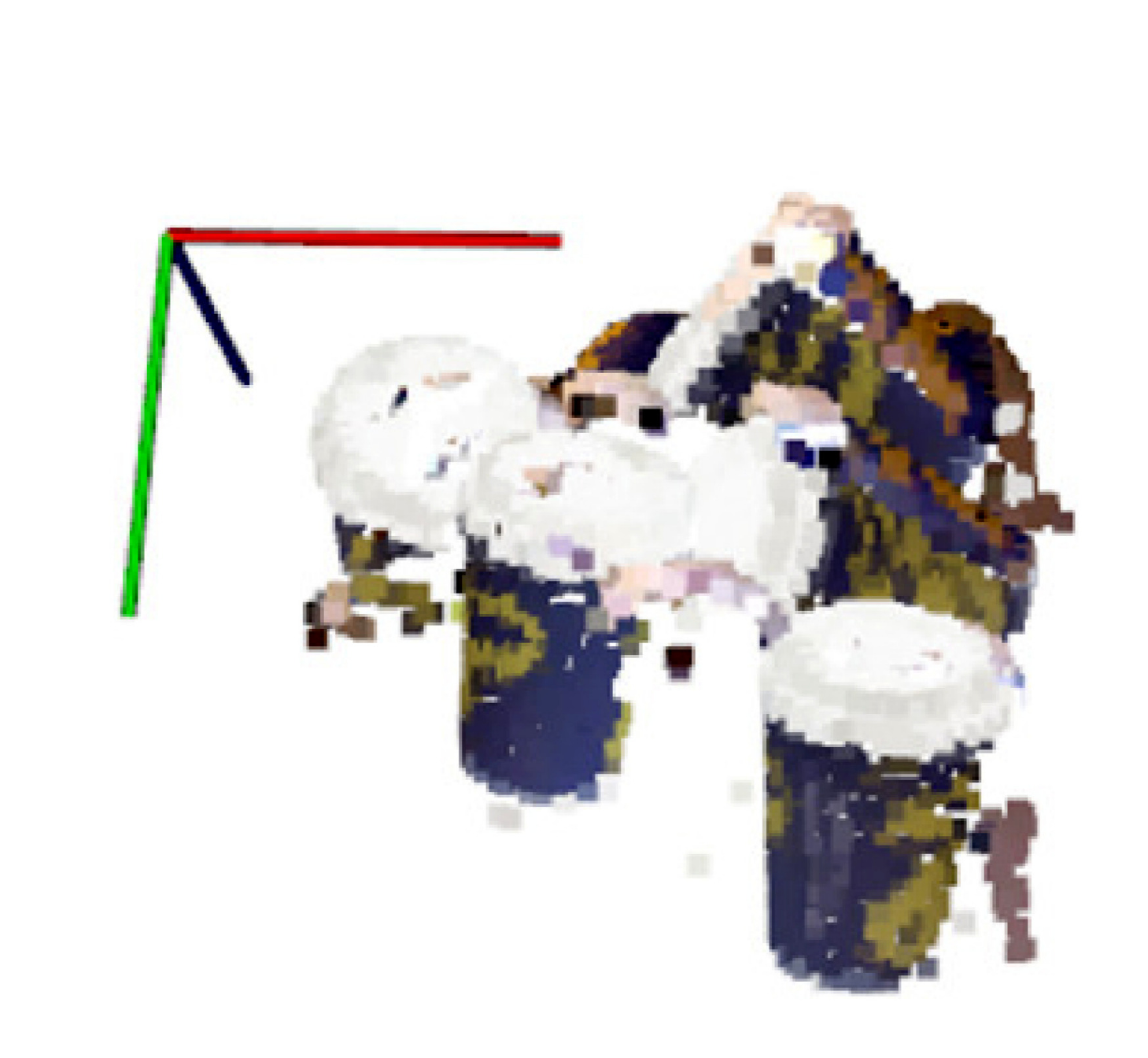} \\
	(\emph{a}) & (\emph{b}) & (\emph{c}) & (\emph{d}) \\
	\end{tabular}
	\caption{ Rendering virtual point cloud for next best view prediction: (\emph{a}) original point cloud; (\emph{b}) the full model of detected
objects are added to the scene (i.e., corresponding points are highlighted by green color); (\emph{c}) the visible points from the virtual camera pose
are highlighted by red color; (\emph{d}) the rendered virtual point cloud. The reference frame represents the camera pose of the acquired view.}
	\label{fig:nbv}       % Give a unique label
\end{figure}

In general, choosing the view with the minimum view-entropy as the next camera position has two problems. Firstly, in a real system, the cost of moving the camera too far could also be considered. Secondly, view entropy estimation becomes less reliable if the virtual (rendering) view is far from the current position. To alleviate this issue, we apply weights to the view entropy value calculated for each view candidate by Gaussian distribution.

\begin{equation}
\begin{split}
\label{eq:gaussian}
H_{v_i}^{w} = w_{v_i}H_{v_i} : ~
 w_{v_i} = \frac{1}{{\sigma \sqrt {2\pi } }}e^{{{ - ||{{\textbf{v}_i} - {\textbf{v}_c} }||^2 } \mathord{\left/ {\vphantom {{ - \left( {x - \mu } \right)^2 } {2\sigma ^2 }}} \right. \kern-\nulldelimiterspace} {2\sigma ^2 }}},
\end{split}
\end{equation}

\noindent where $\sigma$ is a smoothness parameter which restrict the movement of the camera, $w_{vi}$ is the weight applied to view entropy for $v_i$, $v_c$ is the current camera pose, $H_{v_i}$ is the view entropy of the view $v_i$ and $H_{v_i}^{w}$ is the weighted view entropy of the view $v_i$. Although a set of viewpoints which are close to each other may have good attributes, obtaining a sequence of similar viewpoints would not help to detect new objects which are visible from different viewpoints. To encourage exploratory behaviour, the following equation is introduced where viewpoints with higher entropy have a higher chance of being chosen :
\begin{equation}
\label{eq:view_selection_equation}
p(v^{\operatorname{Next}}=v_i)={H_{v_i}^{w}}~/~{\displaystyle \sum_{n=1}^{m}H_{v_n}^{w}}.
\end{equation}

%^^^^^^^^^^^^^^^^^^^^^^^^^^^^^^^^^^^^^^^^^^^^^^^^^^^^^^^^^^^^^^^^
%^^^^^^^^^^^^^^^^^^^^^^^^^^^^^^^^^^^^^^^^^^^^^^^^^^^^^^^^^^^^^^^^
\begin{figure}[!b]
\centering
\begin{centering}
\begin{tabular}{cc}
	\includegraphics[scale=0.28,trim= 0cm 0cm 0.0cm 0cm,clip=true]{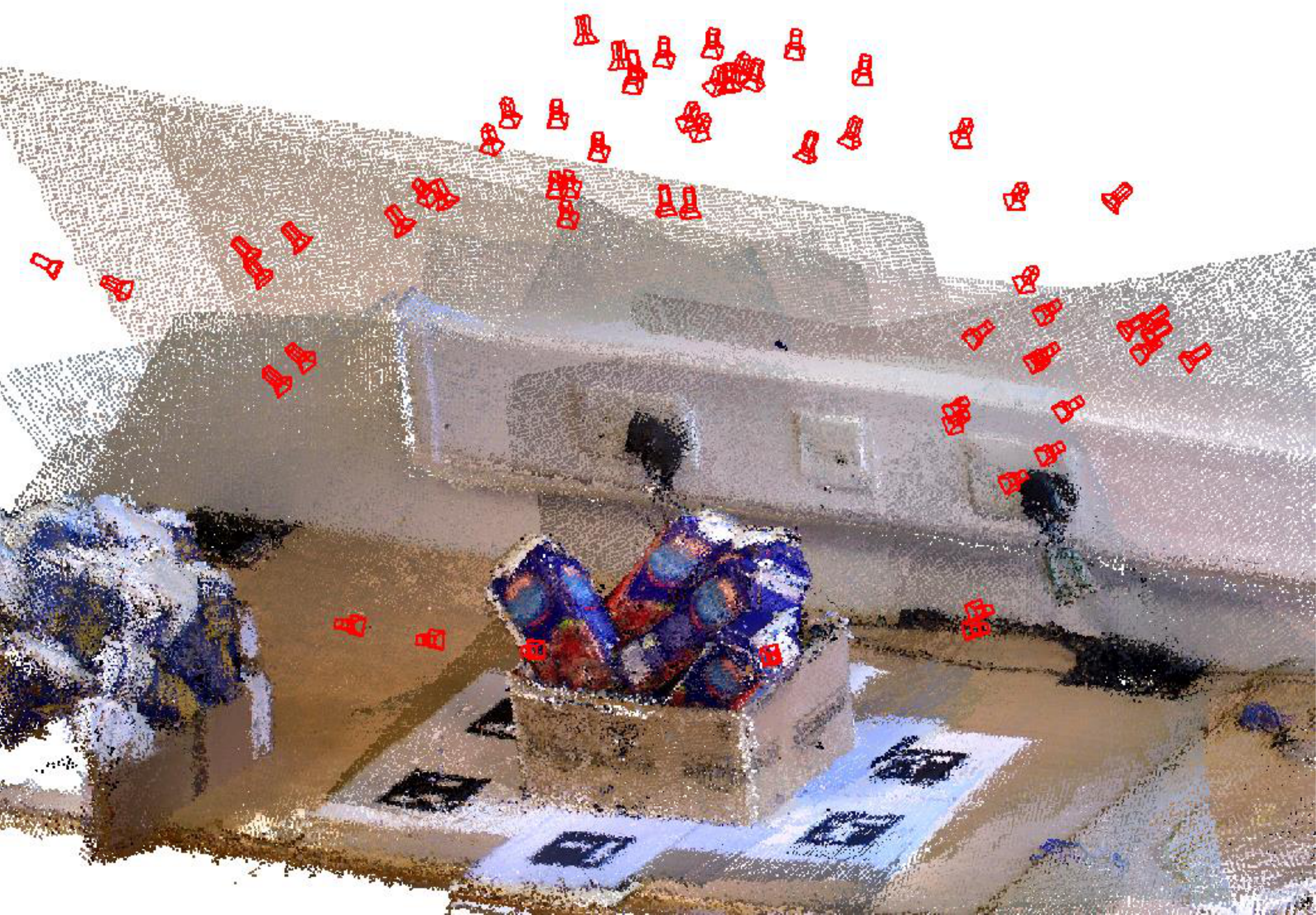}&\hspace{0.2cm}
	\includegraphics[scale=0.24,trim= 0cm 0cm 0.1cm 0cm,clip=true]{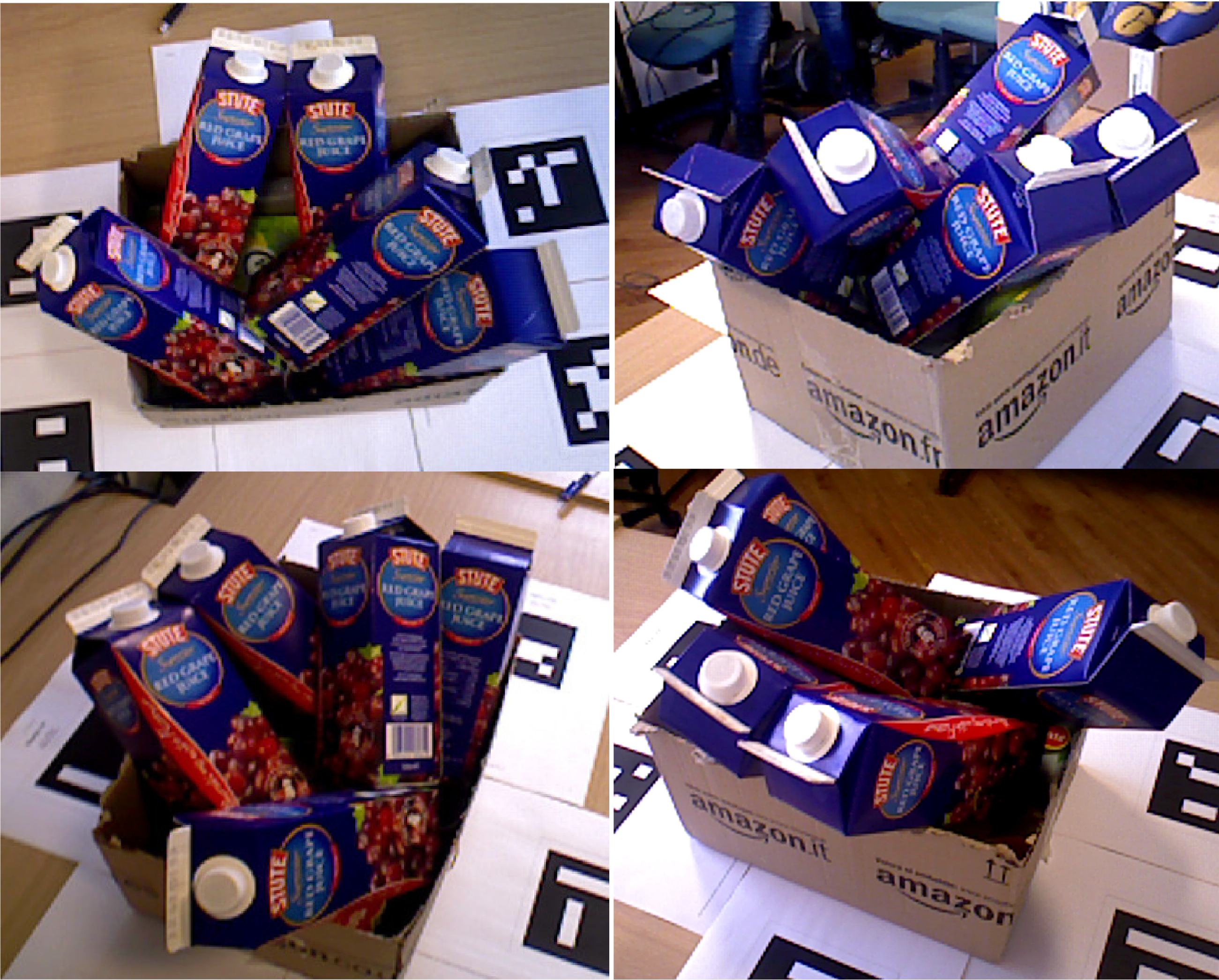}
\end{tabular}
\end{centering}
\caption{(\emph{left}) Accumulated view and camera pose of different views of a pile of juice box scenario; (\emph{right}) images from four different views. }
\label{fig:IC_dataset_bin_picking}
\end{figure}
\subsection{Evaluation of Next-Best-View Prediction} 
We test the proposed NBV prediction on the bin-picking dataset \citep{Doumanoglou2016}, which is one of the few datasets that contain multiple objects in highly crowded scenes. The two scenarios of the dataset are depicted in Fig.~\ref{fig:nbv_fig} and Fig.~\ref{fig:IC_dataset_bin_picking}. The coffee-cup scenario contains $59$ different views of the scene with 15 cups in a pile. The Juice box scenario contains $5$ juice boxes and also has $59$ views. 
Ground truth view point entropy (i.e., highlighted by the red lines in Fig.~\ref {fig:NBV_entropy}) is calculated based on the proposed viewpoint entropy and ground truth object positions provided by the dataset. In this evaluation, an object pose estimator based on sparse auto-encoder \citep{Doumanoglou2016} is first used to generate multiple object hypotheses. The proposed method for rendering \emph{virtual point cloud} is then used. Afterwards, an entropy for the rendered viewpoints is calculated.

\begin{figure}[!b]
\vspace{-8mm}
\hspace{-0.5cm}
\center
\begin{tabular}{c}
	\includegraphics[width=\linewidth,trim= 4.5cm 0cm 4.5cm 0cm,clip=true]{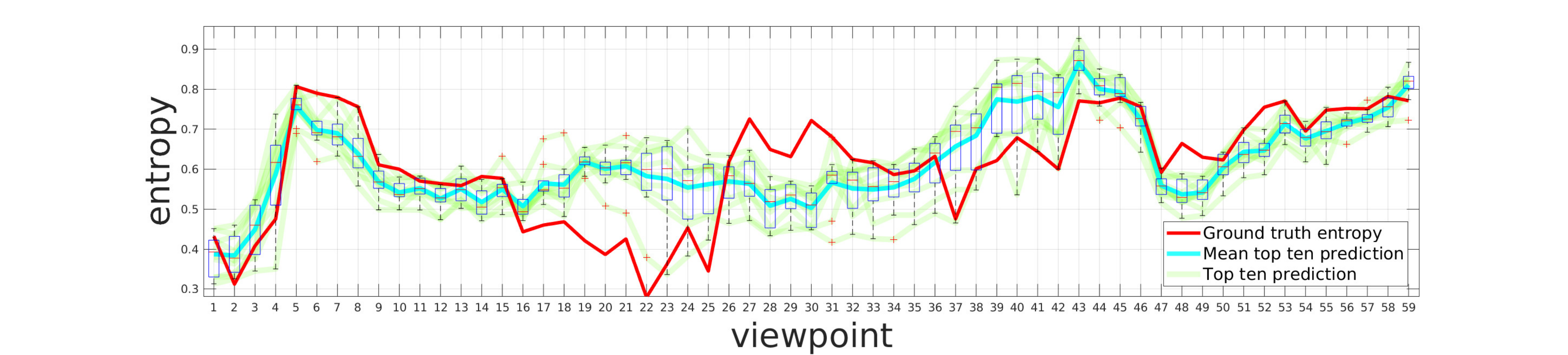}\vspace{-6mm}\\
\includegraphics[width=\linewidth,trim= 4.5cm 0cm 4.5cm 0cm,clip=true]{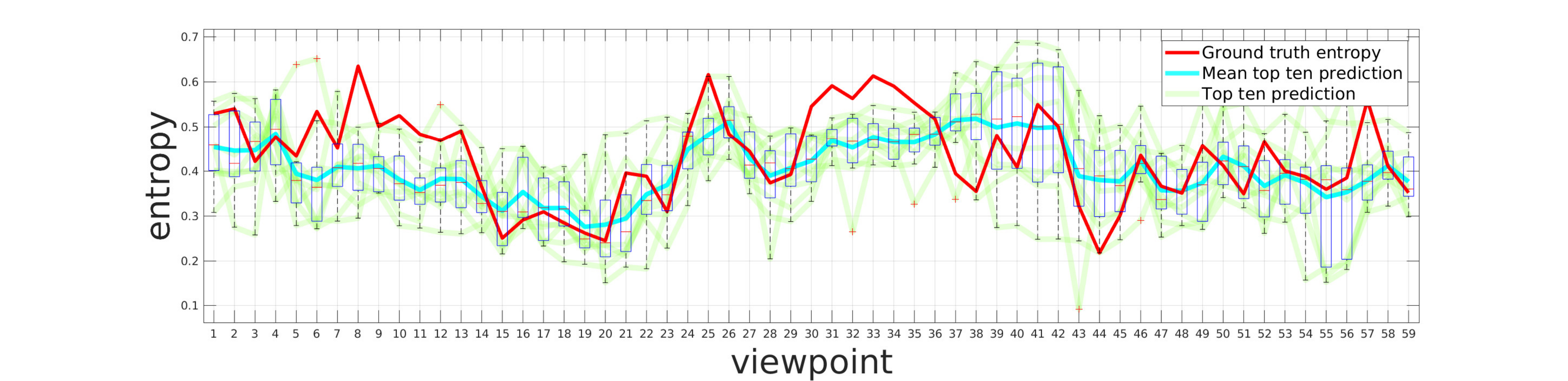}
\end{tabular}
\vspace{-0mm}
\caption{Viewpoint entropy for the bin-picking dataset: (\emph{top}) coffee-cup and (\emph{bottom}) juice-box scenarios.}
\label{fig:NBV_entropy}
\end{figure}

Top-ten views, in terms of predicting view point entropy, are selected based on the square error to the ground truth. Boxplot for the selected views in both scenarios are depicted in Fig.~\ref {fig:NBV_entropy}, which displays the range of variation. The NBV algorithm works well in both scenarios since the standard deviation (SD) of the view entropy is small (i.e., SD for the coffee-cup was $0.105$ and for the juice-box was $0.067$) and the mean is near to the ground truth in both scenarios. The mean squared error (MSE) was $5.22$ and $3.95$ for the coffee-cup and the juice-box scenarios respectively. Note that the coffee cup scenario is much more complex as it has more objects and many of them are occluded in different views. In contrast, objects in the juice box scenario are visible in most of the views.

\begin{figure}[!t]
\centering
\begin{centering}
\begin{tabular}{cc}
	\includegraphics[scale=0.17,trim= 0cm 0cm 0.1cm 0cm,clip=true]{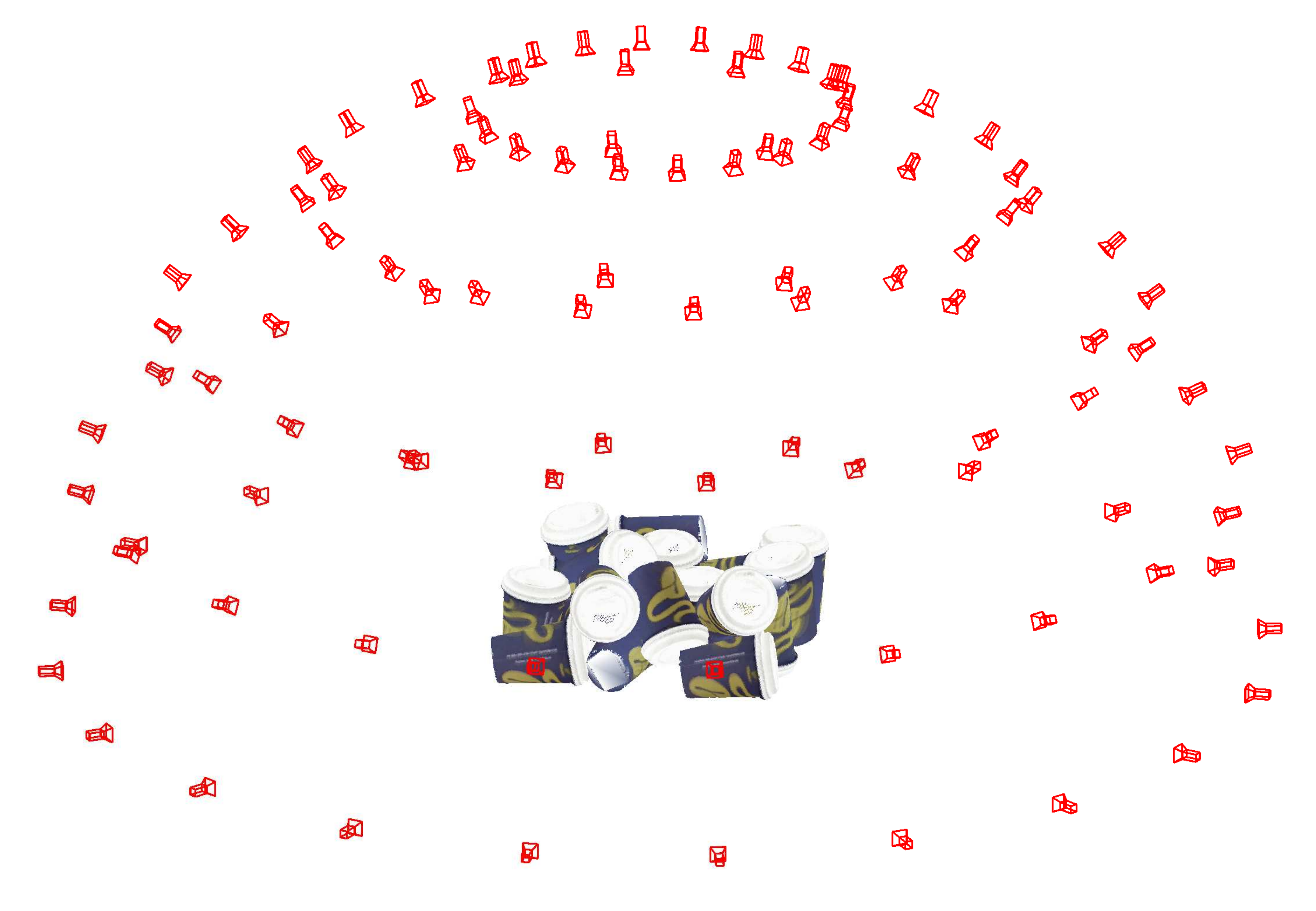} &
	\includegraphics[scale=0.9,trim = 0cm 0cm 0cm 0cm,clip=true]{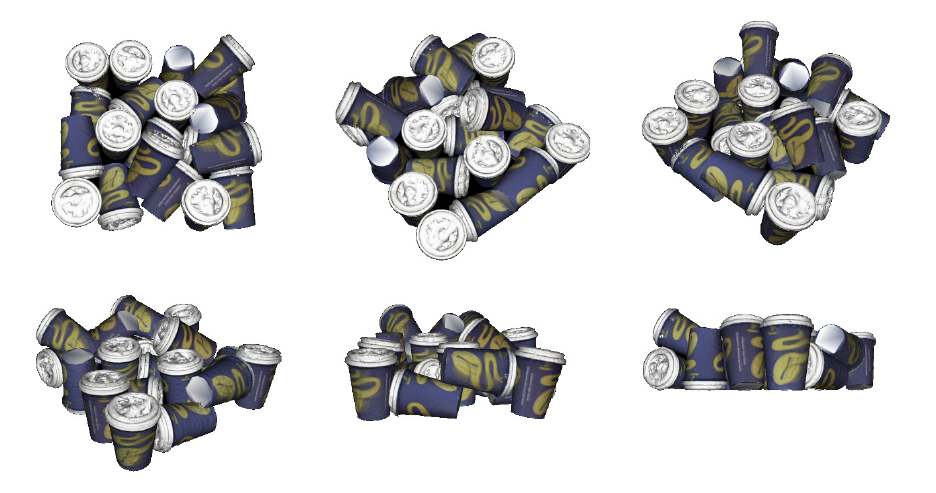} 
\end{tabular}
\end{centering}
\caption{Synthetic dataset contains 20 coffee cups and 100 views: (\emph{left}) accumulated view and camera pose; (\emph{right}) six different views.}
\label{fig:IC_dataset_synthetic}
\end{figure}

For additional details and comparison with furthest view and random view selecting strategies, we refer
the reader to our previous works on 6D object pose estimation~\citep{JuilICCVW2017}}. It is worth to mention that furthest view strategy, as the name implies, selects the furthest possible view from the current camera position. Random selection and furthest possible view are two popular choices in the literature for comparison.

%^^^^^^^^^^^^^^^^^^^^^^^^^^^^^^^^^^^^^^^^^^^^^^^^^^^^^^^^^^^^^^^^
%^^^^^^^^^^^^^^^^^^^^^^^^^^^^^^^^^^^^^^^^^^^^^^^^^^^^^^^^^^^^^^^^
\subsection {Correlation between Visibility and Viewpoint Entropy}
To verify the correlation between object detection performance and the proposed viewpoint entropy, a synthetic dataset is built for the following reasons: (\emph{i}) more dense and even sampling of camera viewpoint can be obtained; (\emph{ii}) perfect knowledge on object ground truth, camera pose and calibration parameters are known. Towards this end, $20$ object models are randomly thrown into a virtual box using MuJoCo physics engine \citep{todorov2012mujoco}.
As depicted in Fig.~\ref{fig:IC_dataset_synthetic} (\emph{right}), RGBD scenes are rendered at $100$ evenly sampled viewpoints around the upper hemisphere. The dataset is publicly available at \cblue{\href{https://goo.gl/BSr2mU}{https://goo.gl/BSr2mU}}

\begin{figure}[!h]
\center
\begin{tabular}{c}
	\includegraphics[width=0.8\linewidth,trim= 0cm 0cm 0cm 0cm,clip=true]{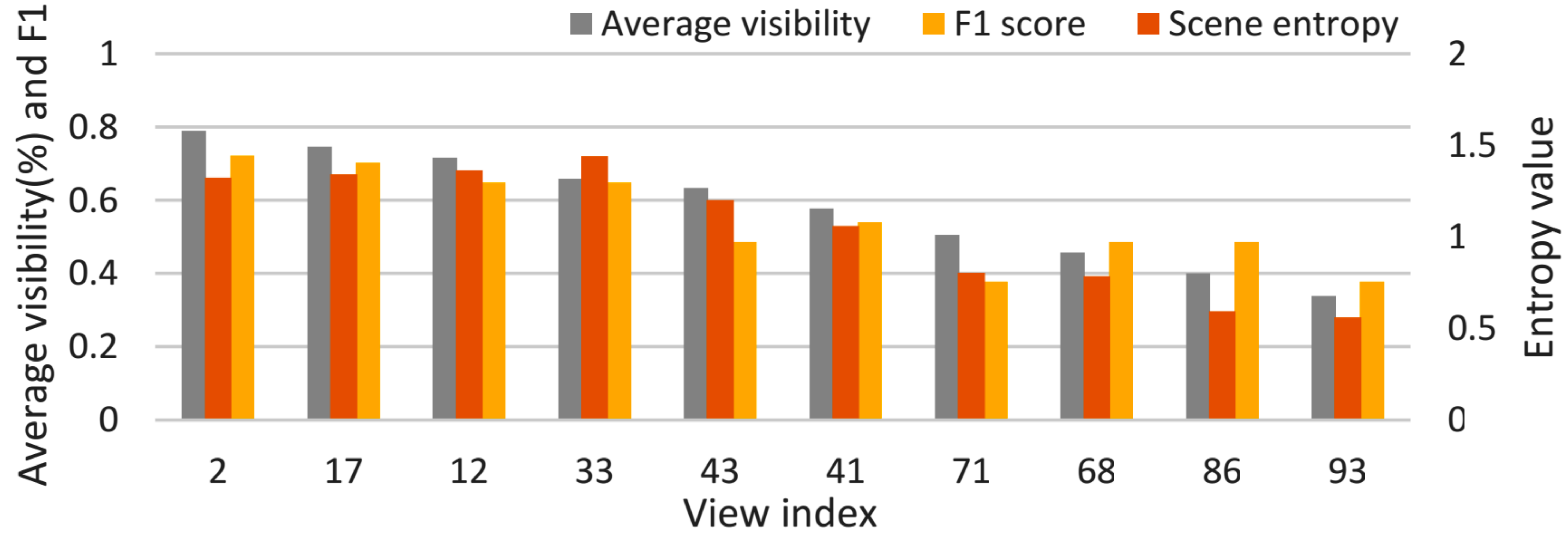}
\end{tabular}
\vspace{-2mm}
\caption{Graph showing the correlation between visibility, viewpoint entropy and detection performance.}
\label{fig:correlation}
\end{figure}

For each object, the ratio of visible pixels to the total number of pixels if the object were not occluded is calculated and these values of every objects in a scene are averaged to quantify the average visibility score for each viewpoint. Object detector proposed by \cite{Doumanoglou2016} is used to obtain the F1-score for each viewpoint and the proposed view entropy is also used to calculate viewpoint entropy. Results are plotted in Fig.~\ref{fig:correlation}. The view indices are ordered in descending average visibility score and the graph shows the F1 score and viewpoint entropy decreases along with the visibility of the viewpoint. The viewpoint entropy and F1 score are positively correlated, with a correlation coefficient of $0.6644$ for the dataset.

\vspace{2mm}

%^^^^^^^^^^^^^^^^^^^^^^^^^^^^^^^^^^^^^^^^^^^^^^^^^^^^^^^^^^^^^^^^
%^^^^^^^^^^^^^^^^^^^^^^^^^^^^^^^^^^^^^^^^^^^^^^^^^^^^^^^^^^^^^^^^
\section {Summary}
\label{sec:summary_chapter3}

In this chapter, several approaches for supervised and unsupervised gathering of object experiences have been introduced. In particular, we first proposed perception capabilities that will allow robots to automatically detect multiple objects in a crowded scene. Furthermore, we discussed how to collect object experiences interactively and incrementally in both supervised and unsupervised manner. We also introduced view entropy formulation, which can be used to predict the Next-Best-View in an environment where robot movement is costly and the scene is complex. This approach was tested on several datasets as shown in Fig.~\ref{fig:3} and Fig.~\ref{fig:two_complex_scene}. Results showed that 3D information is very helpful for object detection purposes. We have also shown that the proposed NBV prediction approach works well in different scenarios. In the next chapter, we present several approaches for 3D object representation.

\newpage
\newpage
\cleardoublepage
\chapter{Object Representation}
\label{chapter_4}
One of the primary goals in service robotics is to develop perception capabilities that will allow robots to robustly 	interact with the environment by manipulating objects. For this purpose, a robot must reliably recognize the object. Furthermore, in order to interact with human users, this process of object recognition cannot take more than a fraction of a second.

Although many object recognition methods for both 2D and 3D data have been proposed~\citep{andreopoulos201350}, recognizing 3D objects in the presence of noise and variable point cloud resolution is still a challenging task. However, 3D data contains more information about the spatial positioning of objects, which in turn eases the process of object segmentation. Moreover, depth data is more robust than RGB data to the effects of illumination and shadows \citep{regazzoni2014rgb}. Therefore, 3D data can be employed to describe the surface of the objects based on geometric properties and 2D data can also be used to distinguish objects that have same geometric properties with different texture (e.x. a Coke can from a Diet Coke can). 

As described in Chapter \ref{chapter_2}, a 3D object recognition system is composed of several software modules such as \emph{Object Detection}, \emph{Object Representation}, \emph{Object Recognition} and \emph{Perceptual Memory}. \emph{Object Detection} is responsible for detecting all objects in a scene. \emph{Object representation} is concerned with the calculation of a set of features for the detected object, which are sent to the \emph{Object Recognition}. The category of an object is recognized by comparing its description against the descriptions of known objects of different categories (stored in the \emph{Perceptual Memory}). Among these modules, \emph{Object Representation} plays a prominent role because the output of this module is used for learning as well as recognition. {Therefore, the \emph{Object Representation} module  must provide reliable information in real-time to enable the robot to physically interact with the objects in its environment. Moreover, the representation of an object should contain enough information enabling to recognize the same or similar objects seen from different perspectives.} 

Existing 3D object representation approaches are based on either hand-crafted descriptions or machine learning approaches. Global descriptors encode the entire 3D object, while local descriptors represent a small area around a specific keypoint in the object. Generally, global descriptors are increasingly used in the context of 3D object recognition, object manipulation, as well as geometric categorization \citep{aldoma2012point}. These must be efficient in terms of computation time as well as memory, to facilitate real-time performance. For example, Ensemble of Shape Functions (ESF) \citep{wohlkinger2011ensemble}, Global Fast Point Feature Histogram (GFPFH) 
\citep{rusu2009detecting}, Viewpoint Feature Histogram (VFH) \citep{rusu2010fast} and Global Radius-based Surface Descriptor (GRSD) \citep{marton2010hierarchical}, are global descriptors.

{Object representation based on local features tends to handle occlusion and clutter better when compared to global features. Examples in this category include Signature of Histograms of Orientations (SHOT) \citep{tombari2010unique}, Spin Images (SI) \citep{johnson1999using}, Fast Point Feature Histogram (FPFH) \citep{rusu2009fast} and Hierarchical Kernel Descriptors \citep{bo2011object}.  However, comparing 3D object views based on their local features tends to be computationally expensive \citep {aldoma2012point}. To address this problem, machine learning approaches such as Bag of Words (BoW) \citep{csurka2004visual} and Latent Dirichlet Allocation (LDA) \citep{blei2003latent} can be used for representing objects in a uniform, highly compact and distinctive way. 

In open-ended learning, the learning agent extracts training instances from its online experiences. Thus, training instances become gradually available over time, rather than being completely available at the beginning of the learning process. Classical object recognition approaches do not perform well in these scenarios because they are often designed for static environments, i.e., training (offline) and testing (online) are two separated phases. If limited training data is used, this might lead to non-discriminative object representations and, as a consequence, to poor object recognition performance. Therefore, building a discriminative object representation is a challenging step to improve object recognition performance. Most of the recent approaches use hand-crafted features. These approaches may be not the best option for open-ended domains and require to be designed separately for each application.  Topic modeling~\citep{blei2003latent} is suitable for open-ended learning because it provides, not only short object descriptions (i.e., optimizing memory), but also enables efficient processing of large collections. It is a type of generative latent structure model that represents the underlying structure of data as topics. 

In this chapter, we first present a new global 3D shape descriptor named GOOD (i.e., Global Orthographic Object Descriptor). GOOD provides an appropriate trade-off between descriptiveness, computation time and memory usage. The descriptor is designed to be scale and pose invariant, informative and stable, with the objective of supporting  accurate 3D object recognition. Afterwards, a set of object representation approaches based on local shape features is presented and discussed. The main motivation of this part is to explore and compute object representations at different level of abstraction (see Fig.~\ref{fig:general_over_view}). We propose a novel object representation approach for describing the shape of an object. In particular,  we propose an extension of Latent Dirichlet Allocation (LDA) to learn structural semantic features (i.e., topics) from low-level feature co-occurrences for each category independently. Topics in each category are discovered in an unsupervised fashion and are updated in an open-ended manner using new object views.}

Throughout this chapter we assume that an object has already been segmented from the point cloud of the scene, and we focus on detailing the object representation approaches. All object representation approaches are computed directly from a segmented 3D point cloud and require neither triangulation of the object's points nor surface meshing. The contributions presented in this chapter are the following: (\emph{i}) a novel global object descriptor computed using a local reference frame, that provides a good trade-off between descriptiveness, computation time and memory usage; (\emph{ii}) design a new sign disambiguation method to compute a unique and unambiguous complete local reference frame, from the eigenvectors obtained through Principal Component Analysis of the segmented point cloud of the object and (\emph{iii}) propose an extension of Latent Dirichlet Allocation to learn topics for each category incrementally and independently. The work presented in this chapter spawned a series of publications presented at major conferences \citep{kasaei2016orthographic,kasaei2016hierarchical,Oliveira2015,Oliveira2014} and top journals in the field \citep{GOODKasaei2016,kasaei2017Neurocomputing,kasaei2015interactive,oliveira20153d}.

The remainder of this chapter is organized as follows. In section \ref{related_work}, we discuss related works. Section \ref{GOOD_shape_description} describes the GOOD object descriptor. The methodology of object representation based on local features is presented in section \ref{sec:local_features_new}. Finally, summary is presented and future research is discussed.

%^^^^^^^^^^^^^^^^^^^^^^^^^^^^^^^^^^^^^^^^^^^^^^^^^^^^^^^^^^^^^^^^
%^^^^^^^^^^^^^^^^^^^^^^^^^^^^^^^^^^^^^^^^^^^^^^^^^^^^^^^^^^^^^^^^

\section{Related Work}
\label{related_work}

Three-dimensional shape description has been under investigation for a long time in various research fields, such as pattern recognition, computer graphics and robotics. Although an exhaustive survey of 3D shape descriptors is beyond the scope of this chapter, we will review the main efforts. 

\subsection{Hand-Crafted Descriptors}
As previously mentioned, some descriptors use a Reference Frame (RF) to compute a pose invariant description. Therefore, this property can be used to categorize 3D shape descriptors into three categories including (\emph{i}) shape descriptors without a common reference; 
(\emph{ii}) shape descriptors computed relative to a reference axis; (\emph{iii}) shape descriptors computed relative to a RF.  Most of the shape descriptors of the first category use certain statistic features or geometric properties of the points on the surface like depth value, curvature and surface normal  to generate a description. For instance, the Shape Distributions descriptor \citep{osada2002shape} represents an object as a shape distribution sampled from a shape function measuring global geometric properties of the object. The Extended Gaussian Images (EGI) descriptor \citep{horn1984extended} is based on the distribution of surface normals on the Gaussian sphere. Descriptiveness of the EGI depends on the shape of the object and it is not suitable for non-convex object. \cite{wohlkinger2011ensemble} introduced a global shape descriptor called Ensemble of Shape Function (ESF) that does not require the use of normals to describe the object. The characteristic properties of the object are represented using an ensemble of ten 64-bin histograms of angle, point distance, and area shape functions. The descriptiveness of the above shape descriptors is limited because the 3D spatial information either is not taken into account or is discarded during the description process. 

In contrast, the shape descriptors in the second and third category encode the spatial information of the objects' points using a Reference Frame (RF). \cite{chen20073d} proposed a Local Surface Patch (LSP) descriptor that encodes the shape of objects by accumulating points in particular bins along the two dimensions that are the shape index value and the cosine of the angle between the surface normals. Point Feature Histogram~(PFH) \citep{rusu2008towards} can be used as a local or global shape descriptor. PFH represents the relative orientation of normals, as well as distances, between point pairs. For each point $p$, k-neighbourhood points are selected based on a sphere centred at $p$ with radius $r$. Afterwards, a surface normal for each point is estimated. Subsequently, four features are calculated for every pair of points using their surface normals, positions and angular variations. In a later work \citep{rusu2009fast}, in order to improve the robustness of PFH in case of point density variations, the distance between point pairs is excluded from the histogram of PFH. The computation complexity of a PFH is $O(n^2)$, where $n$ is the number of points in the point cloud. Fast Point Feature Histogram (FPFH) \citep{rusu2009fast} is an extension of PFH. FPFH estimates the sets of values only between every point and its k nearest neighbours. This is different from PFH, where all pairs of points in the support region are considered. Therefore, the computational complexity is reduced to $O (k.n)$. FPFH is a scale and pose invariant descriptor which is not suitable for grasping. Viewpoint Feature Histogram (VFH) \citep{rusu2010fast} is another extension of PFH that encodes both geometry and viewpoint information, allowing simultaneous recognition of the object and its pose. VFH computes the same angular features as PFH. Additionally, it computes other statistics between the central viewpoint direction and the normals estimated at each point. The VFH shape descriptor produces a single histogram that encodes the geometry of the whole object and its viewpoint. Because of the global nature of VFH, the computational complexity of VFH is $O(n)$. 

Invariance to the pose of an object is a critical property of object descriptors too. A number of 3D shape descriptors achieve pose invariance using either a reference axis only or a complete object reference frame. For example, Spin-Images \citep {johnson1999using} use the surface normal in a vertex as a reference axis. The Spin-Image \citep {johnson1999using} encodes the surrounding shape in a keypoint by projecting the surface points to the tangent plane of the keypoint. Each projected point is represented by a pair $(\alpha, \beta)$, where $\alpha$ is the distance to the surface normal, i.e., the radius, and $\beta$ is the perpendicular distance from the point to the tangent plane. A histogram is formed by counting the occurrences of different discretized distance pairs. Spin-images have been successfully used in many applications, but one limitation of this descriptor is that it is not scale invariant. \cite{dinh2006multi} proposed multi-resolution pyramids of spin images to improve the discrimination of the original spin image and speed up the matching process. Some variants of the spin image shape descriptor were also presented, such as Tri-Spin-Image descriptor (TriSI) \citep{guo2013trisi} and color spin image \citep{pasqualotto2013combining}.

Similar to the spin-image, the 3D Shape Context (3DSC) uses the surface normal at a given point as its RF \citep{frome2004recognizing}. The 3DSC descriptor is calculated by counting the weighted number of points falling into each bin of a spheric grid centred on the given point and its north pole oriented with the surface normal. The spheric grid is constructed based on dividing the support area into bins by logarithmically spaced boundaries along the radial dimension and equally spaced boundaries in the azimuth and elevation dimensions. Whenever only an axis is used as a reference frame, there is an uncertainty in the rotation around the axis that should be handled for generating a robust and repeatable description. In order to eliminate this issue, several descriptors (e.g. 3D Shape Context) proposed to compute multiple descriptions for different possible rotations of the object. Since this kind of solutions greatly increase the computational cost in terms of both execution time as well as memory usage, they are not satisfactory solutions.

\cite{zhong2009intrinsic} proposed a shape descriptor named Intrinsic Shape Signatures (ISS) based on a LRF computed from the eigenvectors of the scatter matrix of the point cloud of the object and describing the point distribution in the spherical angular space. Similar to Zhong's work, \cite{mian2010repeatability} introduced a RF computed with eigenvectors of the covariance matrix of the object's points. In both cases, although the eigenvectors define the principal directions of the data, their sign is not defined unambiguously. Therefore, different descriptors can be generated for the object. As highlighted before, they are neither computationally efficient nor repeatable. \cite{pang2015fast} proposed a multi-view 3D object recognition approach in which each object is projected into 46 projection planes uniformly distributed around a sphere. In contrast we just compute three principal orthographic projections. Their object representation is clearly not efficient for real time application like robotics. In order to achieve true rotation invariance, \cite{tombari2010unique} proposed a 3D shape descriptor named Signature of Histograms of OrienTations (SHOT). They first apply a sign disambiguation technique to the eigenvectors of the scatter matrix of the object and constructed a unique and unambiguous RF. The object's points are then aligned with the RF. Similar to 3DSC, an approach based on spherical coordinate is used to generate a SHOT description for the given object. 3D object descriptors that use the spherical coordinates system suffer from the singularity issue at the poles, because bins at the poles are significantly smaller than bins around the equator.

\cite{aldoma2012point} reviewed properties, advantages and disadvantages of several state-of-the-art 3D shape descriptors available from the Point Cloud Library (PCL) to develop 3D object recognition and pose estimation systems. They also proposed two pipelines for object recognition systems using local and global 3D shape descriptors from PCL. These descriptors have been popular among the robotic community. A summary of the mentioned descriptors is reported in Table~\ref{table:summary_descriptors}.

The GOOD descriptor differs from all of the global descriptors listed in Table~\ref{table:summary_descriptors} as it is simultaneously unique, unambiguous, and robust to noise and varying low-level point cloud density. Besides, our approach can be used not only for object recognition but also for object manipulation. It is worth mentioning that GOOD has been recently integrated into the PCL and will appear in the next release (i.e., PCL 1.9.x).  

\begin{table}[!t]
\begin{center}
    \caption {Summary of the mentioned hand-crafted descriptors.}
\resizebox{\columnwidth}{!}{
\begin{tabular}{|c|c|c|c|c|c|c|}
\hline
\textbf{No.}& \textbf{Descriptor} & \textbf{Reference Frame} & \textbf{Type of Feature} & \textbf{Reference}\\
\hline
1 & Shape Distributions (SD) & no & global & \cite{osada2002shape}\\
\hline
2 & Extended Gaussian Images (EGI) & no & global & \cite{horn1984extended}\\
\hline
3 & Ensemble of Shape Function (ESF) & no & global & \cite{wohlkinger2011ensemble} \\
\hline
4 & Local Surface Patch (LSP) & yes & local & \cite{osada2002shape}\\
\hline
5 & Point Feature Histogram (PFH) & yes & local & \cite{rusu2008towards} \\
\hline
6 & Fast Point Feature Histogram (FPFH) & yes & local & \cite{rusu2009fast} \\
\hline
7 & Global Fast Point Feature Histogram (GFPFH) & yes & global & \cite{rusu2009detecting} \\
\hline
8 & Viewpoint Feature Histogram (VFH) & yes & global & \cite{rusu2010fast}\\
\hline
9 & Global Radius-based Surface Descriptor (GRSD) & no & global & \cite{marton2010hierarchical} \\
\hline
10 & Spin-Images (SI) & yes & local & \cite {johnson1999using}\\
\hline
11 & Tri-Spin-Image (TriSI)  & yes & local & \cite{guo2013trisi} \\
\hline
12 & Color Spin-Image (CSI) & yes & local & \cite{pasqualotto2013combining} \\
\hline
13 & 3D Shape Context (3DSC) & yes & local & \cite{frome2004recognizing}  \\
\hline
14 & Intrinsic Shape Signatures (ISS) & yes & local & \cite{zhong2009intrinsic}\\
\hline
15 & Signature of Histograms of OrienTations (SHOT) & yes & local & \cite{tombari2010unique} \\
\hline
16 & Global Orthographic Object Descriptor (GOOD) & yes & global & \cite{kasaei2016orthographic,GOODKasaei2016}\\
\hline

\end{tabular}}
\label{table:summary_descriptors}
\end{center}
\end{table}

%%%%%%%%%%%%%%%%%%%%%%%%%%%%%%%%%%%%%%%%%%%%%%%%%%%%%%%%%%%%%%%%%%%%%%%%%%%%%%%%%%%%%%%%%%%%%%
\subsection{Machine Learning Approaches for Object Representation}

Comparing 3D objects based on their representative sets of local features is computationally expansive. To address this limitation, several machine learning approaches have been introduced for object representation in which objects are described as a histogram of local shape features. In particular, Bag-of-Words (BoW) and topic modeling~\citep{blei2003latent} techniques are suitable for open-ended learning because, these approaches provide not only short object descriptions (i.e., optimizing memory) but also enable efficient processing of large collections. \cite{csurka2004visual} extended the BoW methods from Natural Language Processing (NLP) domain to the object recognition domain. In text-classification, BoW is a methodology where documents are represented as a histogram of words disregarding grammar. \cite {yeh2009fast} integrated the bag-of-words methodology to propose an efficient method for concurrent object localization and recognition. In \citep {Islam2011}, an object classification approach was proposed, in which object representation was based on SIFT, SURF and color histograms.  All these features were compacted into a histogram of visual words for optimizing the recognition process, as well as memory usage. In this case, the authors used a naive Bayes classifier in the recognition stage. \cite{kasaei2015adaptive} proposed an approach for object representation in which objects are described by histograms of features. 

The codebooks used in the BoW models are usually constructed offline by feeding a sample set of features (the codebook training set) to a clustering algorithm, e.g., K-means or hierarchical K-means. Because clustering is done offline, the representativeness of this training set, i.e., how well the features in the set describe the space of features that the robot will find when operating online, becomes critical to the performance of the system. This is most noticeable in open-ended domains because the categories to be learned are not known a priori and may yield feature patterns which were not included in the codebook training set. 
In other words, the learning agent extracts training instances from its online experiences in open-ended learning. Thus, training instances become gradually available over time, rather than being completely available at the beginning of the learning process. Classical object recognition approaches do not perform well in these scenarios because they are often designed for static environments, i.e., training (offline) and testing (online) are two separated phases. If limited training data is used, this might lead to non-discriminative object representations and, as a consequence, to poor object recognition performance. A long-term process of building and updating a discriminative object representation is an important contribution to improving object recognition performance. 

On the other hand, the approaches that represent objects based on sets of local features or histogram of visual words completely discard structural information, i.e., information related to the co-occurrence of local object features. 
To handle these limitations, several kinds of research have been proposed to concurrently learn both the object category models and the underlying representation used to encode those category models. \cite{riesenhuber1999hierarchical} proposed a hierarchical approach for object recognition consistent with physiological data, in which objects are modelled in a hierarchy of increasingly sophisticated representations. \cite {sivic2005discovering} proposed an approach to discover objects in images using Probabilistic Latent Semantic Indexing (pLSI) modelling \citep {hofmann1999probabilistic}. \cite {blei2003latent} argued that the pLSI is incomplete in that it provides no probabilistic model at the level of documents. They extended the pLSI model calling the approach Latent Dirichlet Allocation (LDA). It is a type of generative latent structure model that represents the underlying structure of data as topics. Similar to pLSI and LDA, we discover topics in an unsupervised fashion. Unlike our approach, pLSI and LDA do not incorporate class information. 

Several works have been presented to incorporate a class label in the generative model. 
\cite{mcauliffe2008supervised} extended LDA and proposed Supervised LDA (sLDA). The sLDA was first used for supervised text prediction. Later, \cite{wang2009simultaneous} extended sLDA to classification problems. Another popular extension of LDA is the classLDA (cLDA) \citep{fei2005bayesian}. Similar to our approach, the only supervision used by sLDA and cLDA is the category label of each training object. However, there are two main differences. First, the learned topics in sLDA and cLDA are shared among all categories, while we propose to learn specific topics per category. Second, the sLDA and cLDA approaches follow a standard train-and-test procedure (i.e., set of classes, train and test data are available in advance), our approach can incrementally update topics using new observations and the set of classes is continuously growing. There are some topic-supervised approaches e.g. Labeled LDA \citep{ramage2009labeled} and semiLDA \citep{wang2007semi} that consider class labels for topics. On the one hand, these approaches need tens of hours of manual annotation. On the other hand, it is hard for a human to provide a specific category label for a 3D local shape feature. 
 
There are some LDA approaches that support incremental learning of object categories. The difference between incremental and open-ended learning is that the set of classes is predefined in incremental learning, while in open-ended learning the set of classes is continuously growing. \cite {banerjee2007topic} proposed online LDA (o-LDA) that is a simple modification of batch collapsed Gibbs sampler. The o-LDA first applies the batch Gibbs sampler to the full dataset and then samples new topics for each newly observed word using information observed so far. \cite{canini2009online} extended o-LDA and proposed incremental Gibbs sampling for LDA (here referred to as I-LDA). The I-LDA does not need a batch initialization phase like o-LDA. In o-LDA and I-LDA, the set of categories is fixed, while in our approach the set of categories is growing. Moreover, o-LDA and I-LDA are used to discover topics shared among all categories, while our approach discovers specific topics per category.
\cite{gao2014learning} and \cite{zhang2016weakly} proposed approaches for fine-grained
image categorization by learning category-specific dictionaries and
a shared dictionary for all the categories. The works of \cite{gao2014learning} and \cite{zhang2016weakly} are similar to ours in that they learned specific representations for each category. However, unlike our approach, the representations of known categories do not change after the training stage.

Some researchers have recently adopted deep learning algorithms for 3D object representation, learning and recognition \citep{li2016fpnn, su2015multi, wu20153d}. These works use a collection of images rendered from different view points to learn a shape representation that aggregates information from input views and provides a compact shape descriptor. As it was pointed by \cite{wu20153d}, training a deep artificial neural network for 3D object representation requires a large collection of 3D objects to provide accurate representations and typically involves long training times.

%^^^^^^^^^^^^^^^^^^^^^^^^^^^^^^^^^^^^^^^^^^^^^^^^^^^^^^^^^^^^^^^^
%^^^^^^^^^^^^^^^^^^^^^^^^^^^^^^^^^^^^^^^^^^^^^^^^^^^^^^^^^^^^^^^^

\section {GOOD: A Global Orthographic Object Descriptor}
\label{GOOD_shape_description}
In this section, a new global 3D shape descriptor named GOOD (i.e., Global Orthographic Object Descriptor) is introduced. GOOD provides an appropriate trade-off between descriptiveness, computation time and memory usage. The descriptor is designed to be scale- and pose-invariant, informative and stable, for ensuring highly accurate 3D object recognition. A novel sign disambiguation method is proposed to compute a unique and repeatable reference frame from the eigenvectors obtained through Principal Component Analysis of the point cloud of the object. Using this reference frame, three principal projections, namely \emph{$XoZ$}, \emph{$XoY$} and \emph{$YoZ$}, are created based on orthographical projection. The space of each projection is partitioned into bins and the number of points falling into each bin is counted. From this, three distribution matrices are obtained for the projected views. Two statistical features, namely \emph{entropy} and \emph{variance} are then calculated for each distribution matrix. The distribution matrices are finally concatenated using the entropy and variance features to decide the order of concatenation.

\subsection{Global Object Reference Frame}
\label{object_reference_frame}

A global object descriptor should be computed in a reference frame that is invariant to translations and rotations and robust to noise. We call it \emph{global object reference frame} to distinguish it from reference frame used for computing local features. Since the repeatability of the global object frame directly affects the descriptiveness of the object representation \citep{mian2010repeatability}, it should be as repeatable and robust as possible to improve the performance of object recognition. In this subsection, we propose a method to compute this RF. For this purpose, the  three principal axes of the target object are firstly determined based on Principal Component Analysis (PCA).
Given a point cloud of an object that contains $m$ points, $\textbf{O}=\{\textbf{p}_1, \dots, \textbf{p}_m\}$, the geometric center of the object is defined as:
\begin{equation} 
	\textbf{c} = \frac{1}{m}\sum_{i=1}^{m} \textbf{p}_i,
	\label {CoM}
\end{equation}
where each $\textbf{p}_i$ is a three dimensional point in the object's point cloud. The normalized covariance matrix, \textbf{C}, of the object is constructed:
\begin{equation}
	\textbf{C} = \frac{1}{m}\sum_{i=1}^{m} (\textbf{p}_i-\textbf{c})(\textbf{p}_i-\textbf{c})^T.
	\label {covariance_matrix}
\end{equation}
Then, eigenvalue decomposition is performed on \textbf{C}:
\begin{equation}
	\textbf{C}\textbf{V} = \textbf{E}\textbf{V},
	\label {covariance_matrix}
\end{equation}
where $\textbf{V} = [\textbf{v}_1, \textbf{v}_2, \textbf{v}_3]$ contains the three eigenvectors,
$\textbf{E} = diag(\lambda_1, \lambda_2, \lambda_3)$ is a diagonal matrix of the corresponding eigenvalues and 
$\lambda_1 \ge \lambda_2 \ge \lambda_3$.
Since the covariance matrix is symmetric positive,  its eigenvalues are positive and the eigenvectors are orthogonal.

\begin{figure}[!t]
\center
\begin{tabular}{cc }
\vspace{-14mm}\multirow{2}{*}{\includegraphics[scale=0.18]{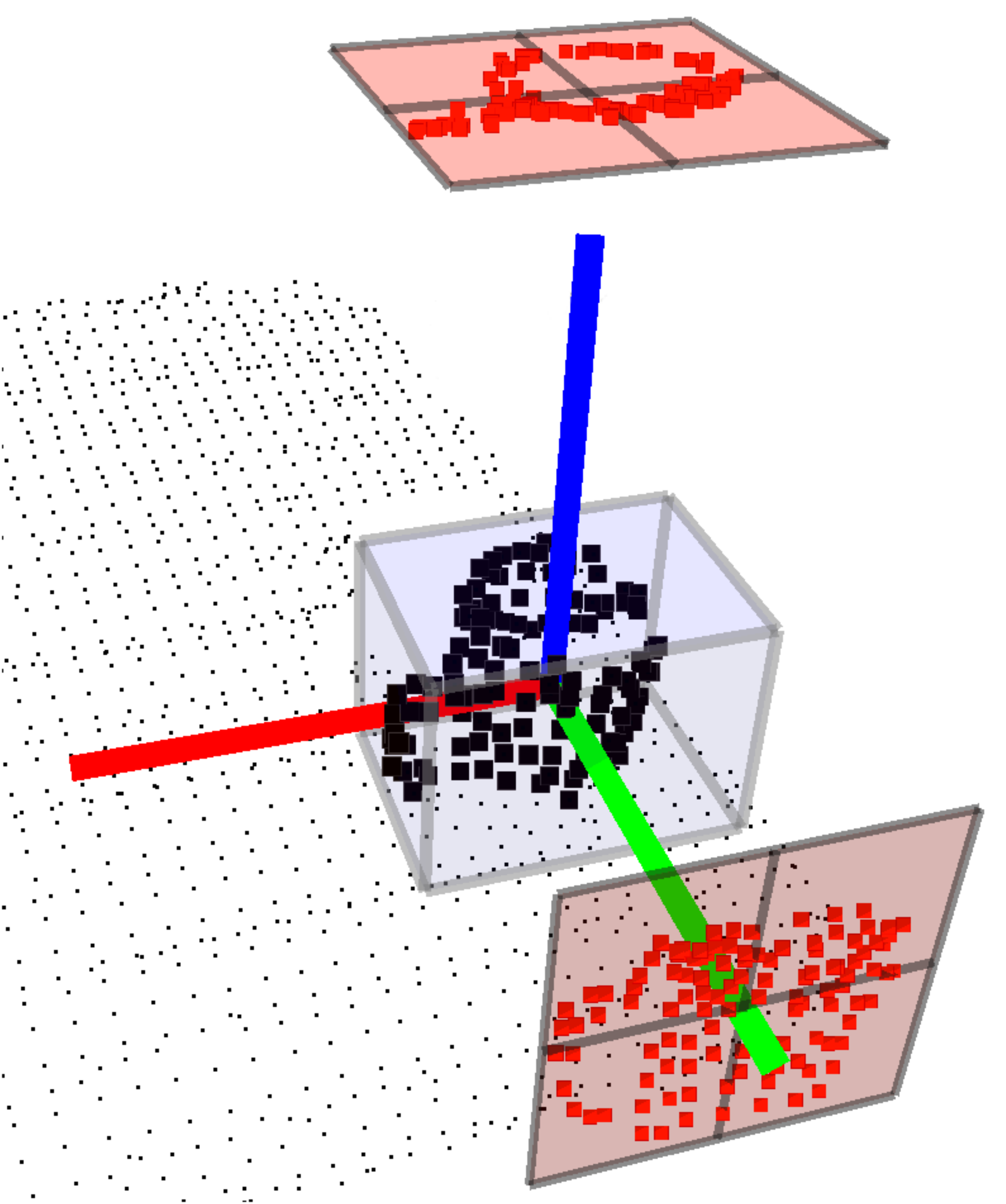} } & \\ 
 & \\ 
  &\quad \quad \includegraphics[width=0.17\textwidth]{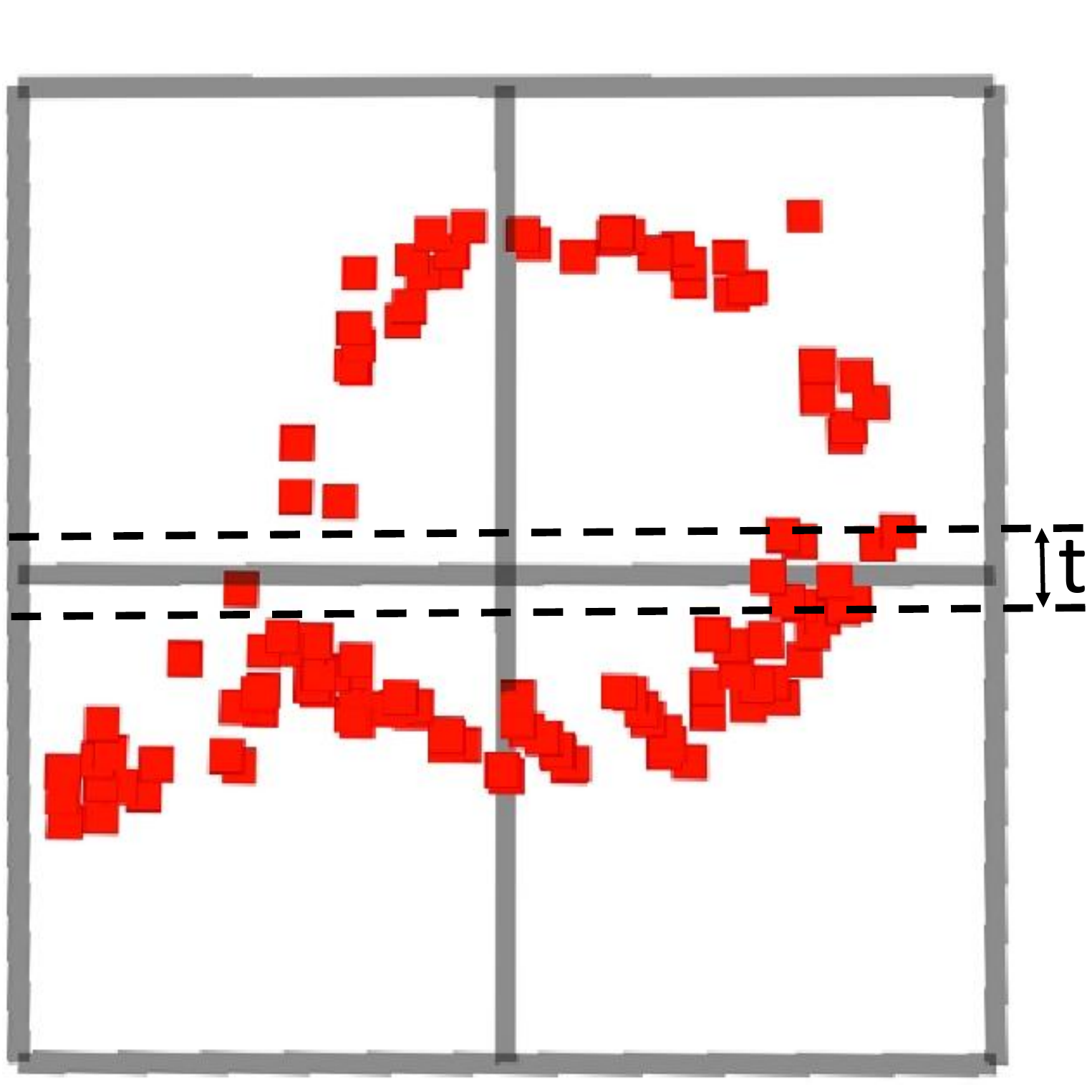} \\
 & \quad \quad \emph{(b)}\\
 & \quad \quad \includegraphics[width=0.17\textwidth]{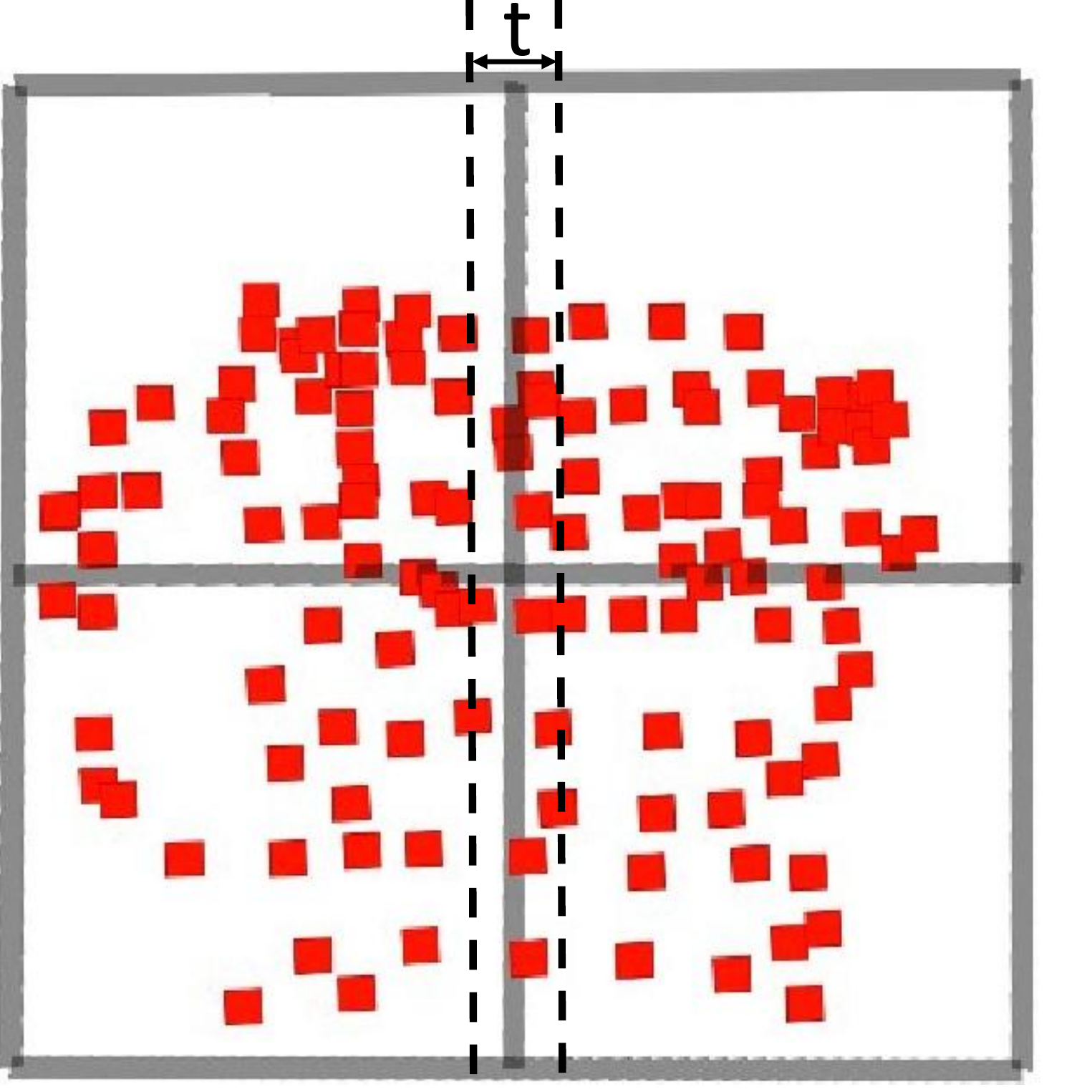} \vspace{2mm}\\
 \emph{(a)} & \quad \quad \emph{(c)}\\

\end{tabular} 
\vspace{-0mm}
\caption{Visualization of sign disambiguation procedure: \emph{(a)} orthographic projection of the object on the $XoZ$ and $XoY$ planes; \emph{(b)}  $XoY$ plane is used to determine the sign of $Y$ axis; \emph{(c)}  $XoZ$ plane is used to determine the sign of $X$ axis. The red, green and blue lines represent the unambiguous $X$, $Y$, $Z$ axes respectively.}
\vspace{-3mm}
	\label{fig:sign_disambiguation}       % Give a unique label
\end{figure}
Eigenvectors define directions which are not unique, i.e., not repeatable across different PCA trials. This is known as the sign ambiguity problem, for which there is no mathematical solution \citep{bro2008resolving}. Since there are two possible directions for each eigenvector, a total of eight reference frames can be created from the same set of eigenvectors. 
A mechanism is needed to transform this reference frame into a unique object reference frame, which will be the same across multiple trials.

We start with a provisional reference frame, in which the first two axes, $X$ and $Y$, are defined by the eigenvectors $\textbf{v}_1$ and $\textbf{v}_2$, respectively. However, regarding the third axis, $Z$, instead of defining it based on $\textbf{v}_3$, we define it based on the cross product $\textbf{v}_1 \times \textbf{v}_2$. This way, because the result of the cross product follows the right-hand rule, the number of alternatives is reduced to four. 
It is now enough to disambiguate the directions of the $X$ and $Y$ axes. So either the directions of $X$ and $Y$ are both changed or both remain unchanged. 

To complete the disambiguation, the object's point cloud, $\textbf{O}$, is transformed to be placed in the provisional reference frame. Then, the number of points that have positive $x$, $N_x^+$, and the number of points that have negative $x$, $N_x^-$, are computed:
\begin{equation}
	 N_x^+  =  |\{i : x_{p_i} > t \}|, \quad  N_x^-  =  |\{i : x_{p_i} < -t \}|,
	\label {sign_x}
\end{equation}
\noindent 
where $|.|$ denotes the number of points of the argument and $t$ is a threshold (e.g. $t = 0.015 m$) that is used to deal with the special case when a point is close to the $YoZ$ plane, and therefore can change from negative to positive $X$ in different trials.
Afterwards, the variable $S_x$ is defined as: 
\begin{equation}
\label{Sx}
S_x = 
\begin{cases}
+1,\quad\quad N_x^+ \ge N_x^-\\
-1,\quad\quad\emph{otherwise}
\end{cases},
\end{equation}
A similar indication, $S_y$, is computed for the $Y$ axis. Finally, the sign of the axes is determined as: 
\begin{equation}
 \emph{s}~ =~ S_x ~.~S_y,
\end{equation}
 %which
where $s$ can be either $-1$ or $+1$. In case of $s = -1$, the directions of $X$ and $Y$ must be changed, otherwise not. 
Therefore, the final RF $(X,Y,Z)$ will be defined by $( \texttt{s}\mathbf{v}_1,\texttt{s}\mathbf{v}_2, \mathbf{v}_1 \times\mathbf{v}_2)$.
An illustrative example of the sign disambiguation procedure is provided in Fig.~\ref{fig:sign_disambiguation}.

%%%%%%%%%%%%%%%%%%%%%%%%%%%%%%%%%%%%%%%%%%%%%%%%%%%%%%%%%%%%%%%%%%%%%%%%%%%%%%%%%%%%%%
%%%%%%%%%%%%%%%%%%%%%%%%%%%%%%%%%%%%%%%%%%%%%%%%%%%%%%%%%%%%%%%%%%%%%%%%%%%%%%%%%%%%%%

\subsection{Computing the Object Descriptor}
\label{shape_description}

This section describes the computation of the proposed object descriptor, GOOD, in the obtained RF centered in the geometric center of the object. The descripor consists of a concatenation of the orthographic projections of the object on the three orthogonal planes, $XoY$, $YoZ$ and $XoZ$. Each projection is described by a distribution matrix. To ensure correct comparison between different object shapes, the number of bins in the distribution matrices must be the same and the bins should be of equal size. Therefore, each distribution matrix must be computed from a square area in the projection plane centered on the object's center, and this square area must have the same dimensions for the three projections. The side length of these square areas, $l$, is determined by the largest edge length of a tight-fitting axis-aligned bounding box (AABB) of the object. The dimensions of the AABB are obtained by computing the minimum and maximum coordinate values along each axis. With this setup, the number of bins, $n$, is the only parameter that must be specified to compute GOOD. For each projection, the $l \times l$ projection area is divided into $n \times n$ square bins. Finally, a distribution matrix $\textbf{M}_{n \times n}$ is obtained by counting the number of points falling into each bin.

\begin{figure}[!t]
\hspace{-8mm}
\begin{tabular}{ccc}
		\includegraphics[scale=0.17]{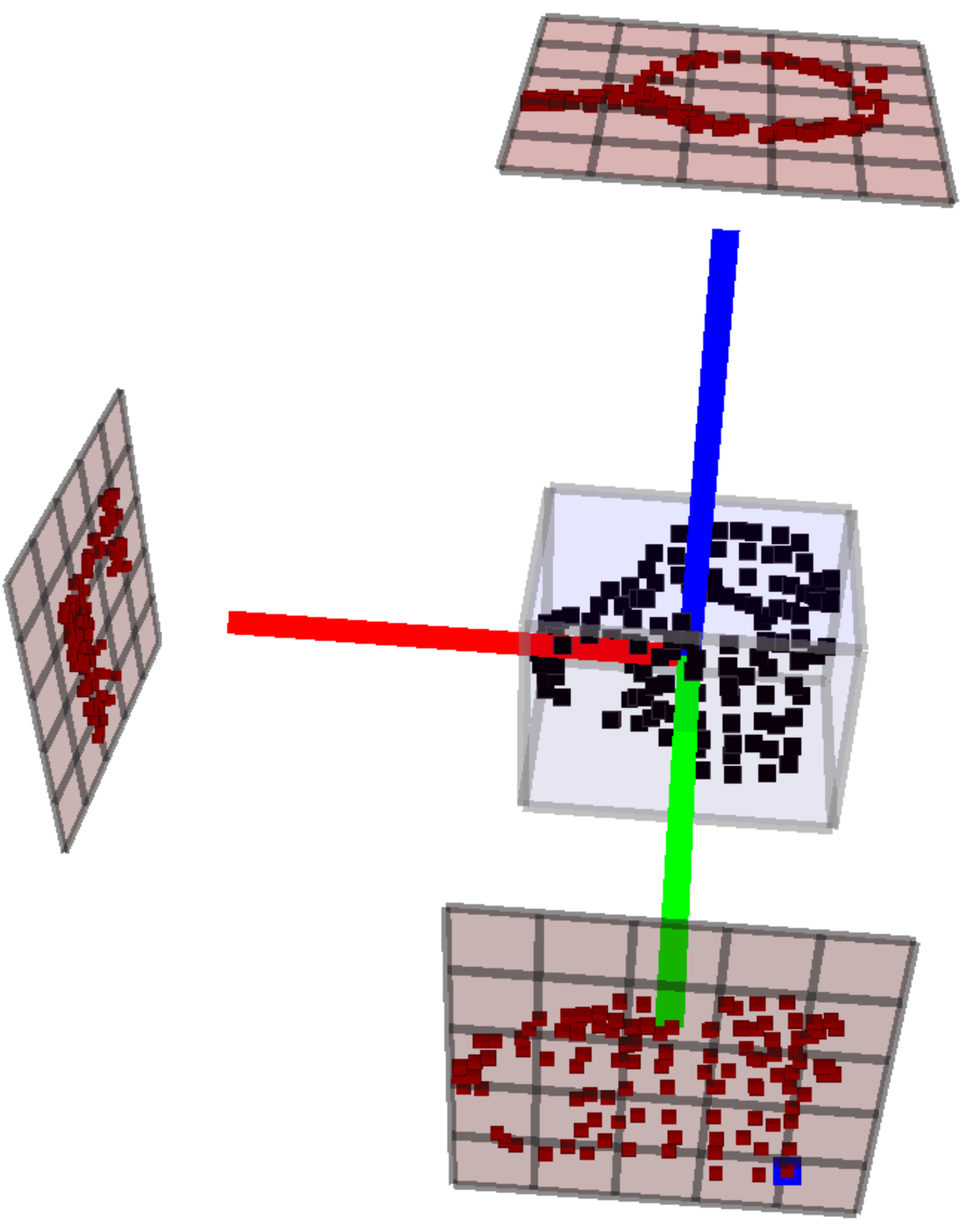}&
		\begin{tabular}{ccc}
			\hspace{-2mm}\includegraphics[scale=0.25]{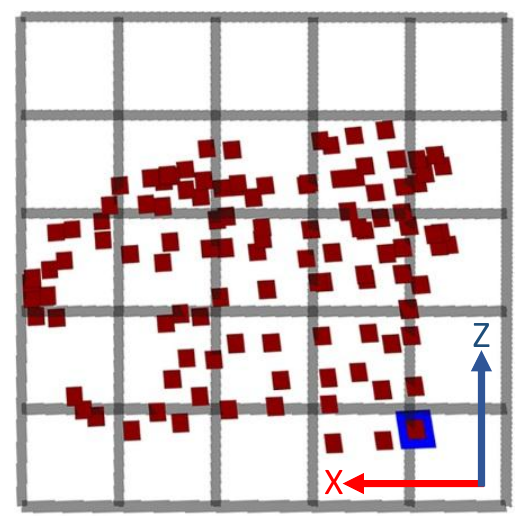}&
			\hspace{-2mm}\includegraphics[scale=0.25]{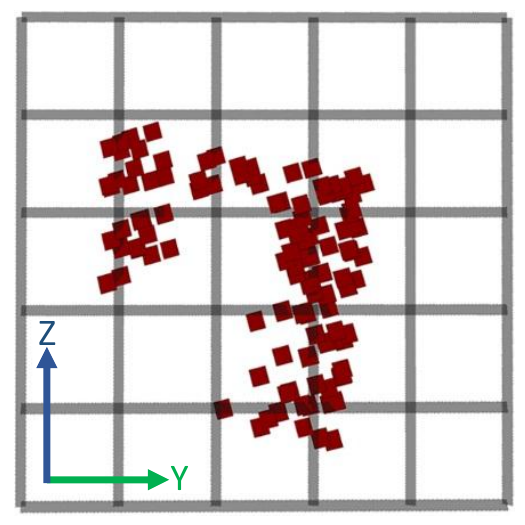}&
			\hspace{-2mm}\includegraphics[scale=0.25]{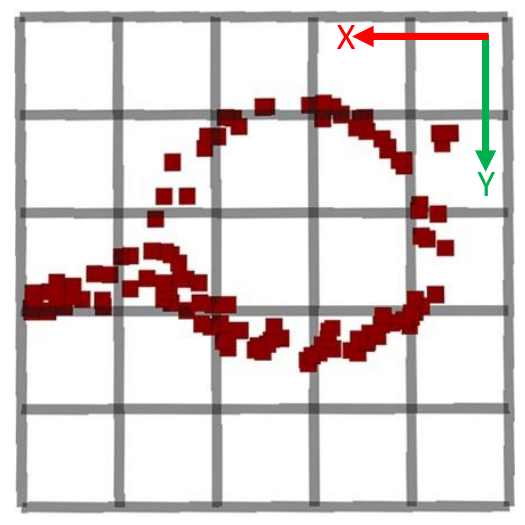} \\
			\hspace{-2mm}(\emph{b}) & \hspace{-2mm}(\emph{c}) & \hspace{-2mm}(\emph{d})\\
			\hspace{-2mm}\includegraphics[scale=0.12]{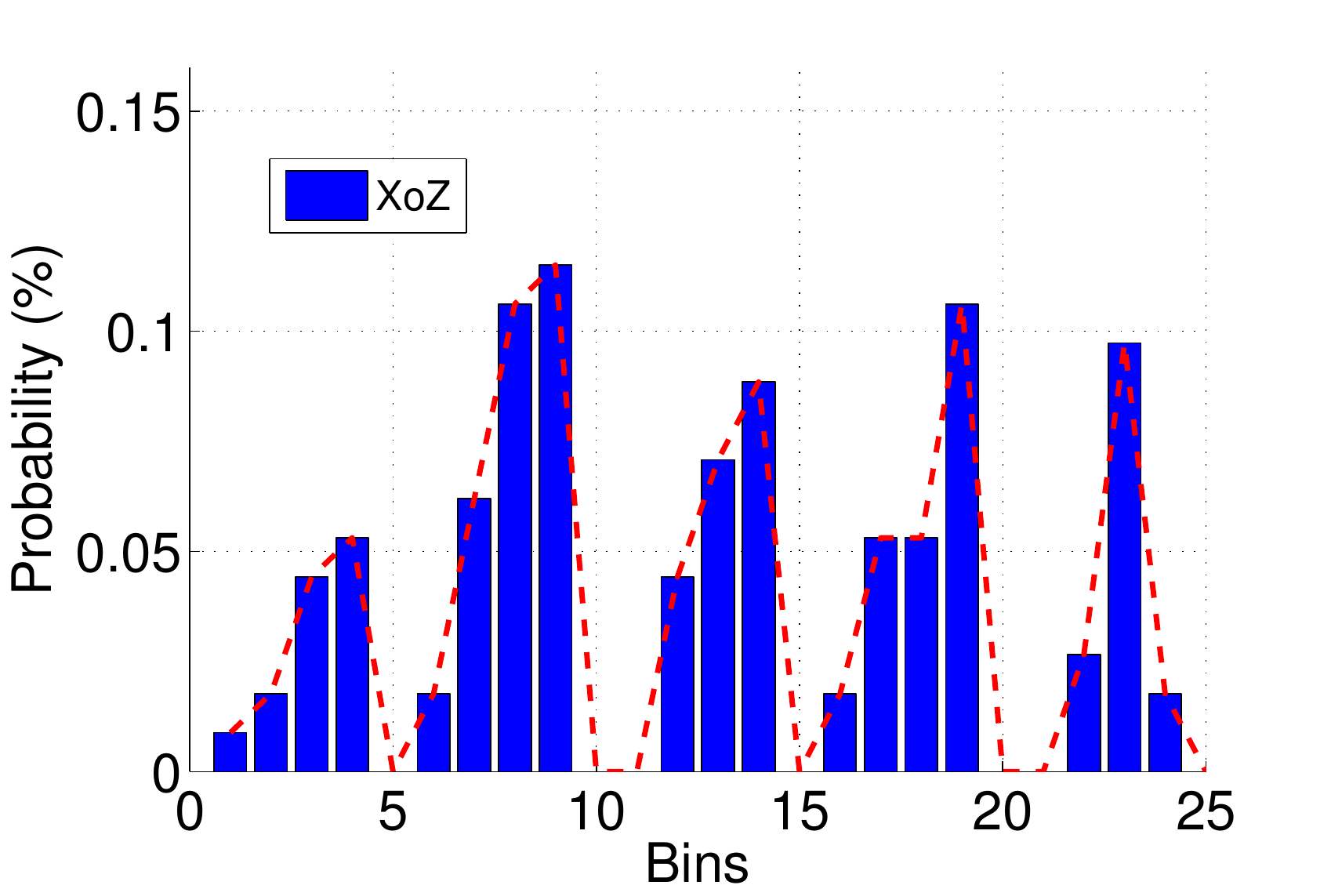}&
			\hspace{-2mm}\includegraphics[scale=0.12]{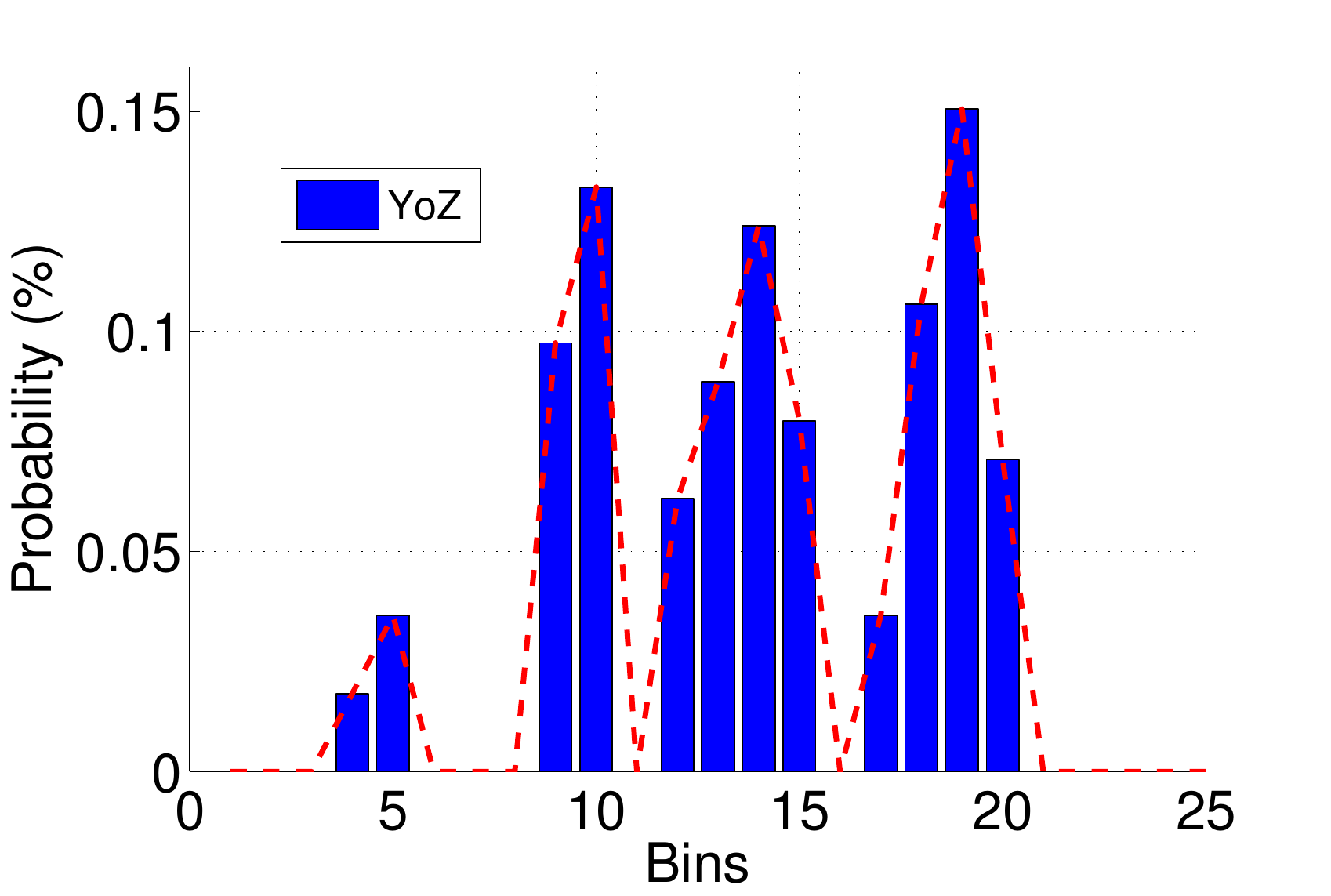}&
			\hspace{-2mm}\includegraphics[scale=0.12]{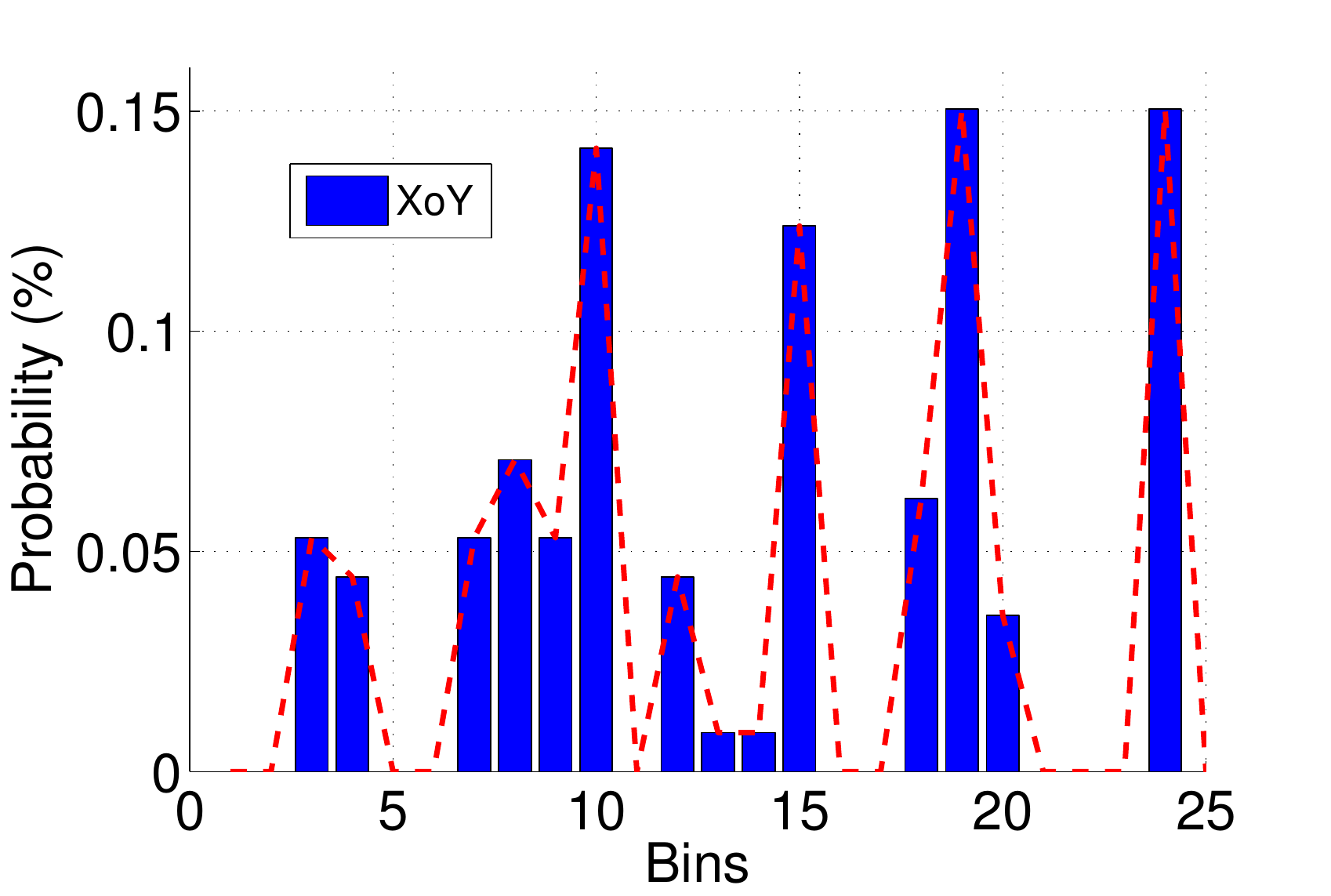} \\
			\vspace{-40mm}
			\hspace{-2mm}(\emph{e}) & \hspace{-2mm}(\emph{f}) & \hspace{-2mm}(\emph{g})\vspace{70mm}
		\end {tabular}	
	\vspace{-30mm}
	&\hspace{-9mm}\includegraphics[scale=0.13]{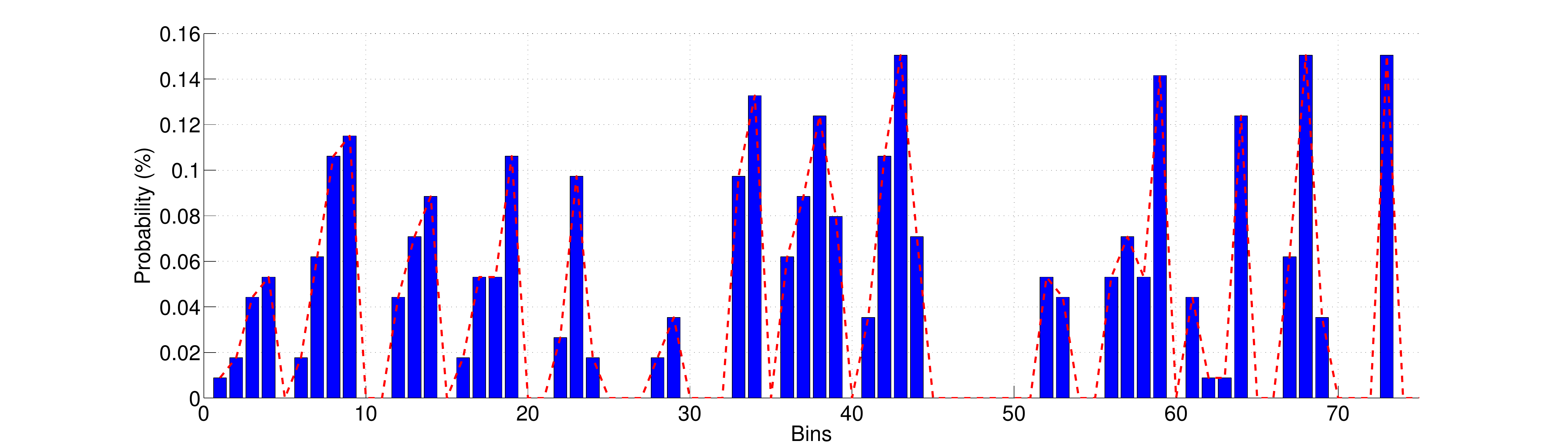} \vspace{-4mm}\\

	{\vspace{-10mm}(a)} & &  \hspace{-6mm}(h) \vspace{40mm}\vspace{-30mm}\\
\end{tabular}
\caption{An illustrative example of producing a GOOD shape description for a mug, using $n = 5$: \emph{(a)} The mug object and its bounding box, reference frame and three projected views; $XoZ$ \emph{(b)}, $YoZ$ \emph{(c)} and $XoY$  \emph{(d)} projections are created. Each projection is partitioned into bins, the number of points falling into each bin is counted and three distribution matrices are obtained for the projected views; afterwards, each distribution matrix is converted to a distribution vector, (i.e., \emph{(e)}, \emph{(f)} and \emph{(g)}) and the two statistic measures, \emph{entropy} and \emph{variance}, are then calculated for each distribution vector; \emph{(h)} the distribution vectors are consequently concatenated to form a single description for the given object. The ordering of the three distribution vectors is first by decreasing values of entropy. Afterwards the second and third vectors are sorted by increasing values of variance. }
	\label{fig:complete_example}       % Give a unique label
	\vspace{-4mm}
\end{figure}

For each projected point $\rho = (\alpha, \beta) \in R^2$, where $\alpha$ is the perpendicular distance to the horizontal axis and $\beta$ is the perpendicular distance to the vertical axis, a row, $r(\rho)\in \{0,\dots,n-1 \}$, and a column, $c(\rho)\in \{0,\dots,n-1 \}$, are associated as follows:
\begin{equation}
	r(\rho)= \Big\lfloor \frac {\alpha+\frac {l}{2}}{\frac{l+\epsilon}{n}}\Big\rfloor = \Big\lfloor n \frac {\alpha+\frac {l}{2}}{l+\epsilon}\Big\rfloor,
	\label {row_index}
\end{equation}
\begin{equation}
	c(\rho)= 
\Big\lfloor \frac {\beta+\frac {l}{2}}{\frac{l+\epsilon}{n}}\Big\rfloor = \Big\lfloor n \frac {\beta+\frac {l}{2}}{l+\epsilon}\Big\rfloor,
	\label {column_index}
\end{equation}
\noindent
where $\epsilon$ is a very small value used to deal with the special cases when a point is projected onto the upper bound of the projection area, and $\lfloor x\rfloor $ returns the largest integer not greater than $x$. Note that the projected view is shifted to right and top by $\frac {l}{2}$ (i.e., $\alpha+\frac {l}{2}$ and $\beta+\frac {l}{2}$). Furthermore, to achieve invariance to point cloud density, $\textbf{M}$ is normalized such that the sum of all bins is equal to 1.0 (see Fig.~\ref{fig:complete_example}). The matrix $\textbf{M}$ is called distribution matrix because it represents the 2D spatial distribution of the object's points. 
According to standard practice, this matrix is converted to a vector $\textbf{m}_{1\times n^2} = [M(1,1),M(1,2), \dots, M(n,n)]$.
The three projection vectors will be concatenated producing a histogram vector of dimension $G = 3\times n^2$ which is the final object descriptor, GOOD (i.e., $\textbf{O}^{\textbf{g}} = [h_1 ~ h_2 ~ ... ~h_G]$). Statistical features are used to  decide the order in which the projection vectors will be concatenated.

For the first projection in the descriptor, the one with largest area is preferred. The number of points is not a good indicator of area because all points of the object are represented in the three projections. The number of occupied bins (the ones with a mass greater than 0) could be used as a measure of area. However, this measure tends to be brittle when the boundary of the object is close to boundaries between bins. Therefore, in this work the entropy of the projection is used. Entropy, a measure from Information Theory \citep{cover2012elements}, nicely takes into account both the number of occupied bins and their density. In this work, the entropy of a projection is computed as follows:
\begin{equation}
	H(\textbf{m})= -\sum_{i=1}^{n}\textbf{m}_{i}~\log_2 ~\textbf{m}_{i}, 
	\label {entropy}
\end{equation}
where $\textbf{m}_{i}$ is the mass in bin $i$. The logarithm is taken in base 2 and $0 \log_2 0 = 0$. 
The projection with highest entropy is the one that will appear in the first $n^2$ positions of the descriptor.

The next step is to select, from the remaining two projections, which one should appear in the second part of the descriptor (positions $n^2$ to $2n^2-1$). It is common that these two projections have similar areas, and therefore similar entropies, leading to instability of the decision if it is made based on entropy. Therefore, instead of entropy, we use variance to make this decision. Since the projection matrices are probability mass functions (pmf), the variance is defined as follows:
\begin{equation}
%	H(\textbf{m})= -\sum_{i=1}^{n^2} \textbf{m}_i~\log_2 ~\textbf{m}_i, 
	\sigma^{2}(\textbf{m})= \sum_{i=1}^{n}~(i-\mu_\textbf{m})^2 ~ \textbf{m}_{i}, 
	\label {variance}
\end{equation}
where $\mu_\textbf{m}$ is the expected value (i.e., a weighted average of the possible values of $i$, corresponding to the geometric center of the projection), which is computed as follows:
\begin{equation}
%	H(\textbf{m})= -\sum_{i=1}^{n^2} \textbf{m}_i~\log_2 ~\textbf{m}_i, 
	\mu_\textbf{m}= \sum_{i=1}^{n^2} i~\textbf{m}_{i	 }, 
	\label {mean}
\end{equation}
\noindent

The variance measure, $\sigma^{2}(\textbf{m})$, is used to measure the spread or variability of the spatial distribution of the object's points in the projection vector.
A small variance indicates that the projected points tend to be very close to each other and to the mean of the vector, i.e., the distribution is small and compact. A high variance indicates that the data points in the projection vector are very spread out from the mean.

An illustrative example of the proposed shape descriptor is depicted in Fig.~\ref{fig:complete_example}. In this example, after determining the global object reference frame, a mug is projected onto the three orthogonal planes. Based on the entropy criterion, the $XoZ$ projection (Fig.~\ref{fig:complete_example}b) is selected to appear in the first part of the descriptor. Based on the variance criterion, the $YoZ$ projection (Fig.~\ref{fig:complete_example}c) is selected to appear in the second part of the descriptor. The remaining projection, $XoY$ (Fig.~\ref{fig:complete_example}d), appears in the last part of the descriptor.

%^^^^^^^^^^^^^^^^^^^^^^^^^^^^^^^^^^^^^^^^^^^^^^^^^^^^^^^^^^^^^^^^^^^^
%^^^^^^^^^^^^^^^^^^^^^^^^^^^^^^^^^^^^^^^^^^^^^^^^^^^^^^^^^^^^^^^^^^^^
\subsection {Relevance for Object Manipulation}
In order to grasp an object, it is necessary to know the true dimensions of different parts of the object. Such information is not adequately represented in most shape descriptors (e.g. Viewpoint Feature Histogram \citep{rusu2010fast}). Because GOOD is composed of three orthogonal projections, it is especially rich in terms of information suited for manipulation tasks. We previously showed how to use orthographic projections for object manipulation purposes \citep{kasaei2016object}.

\begin{figure}[!t]
\center
\begin{tabular}{c c }
\includegraphics[scale=0.19]{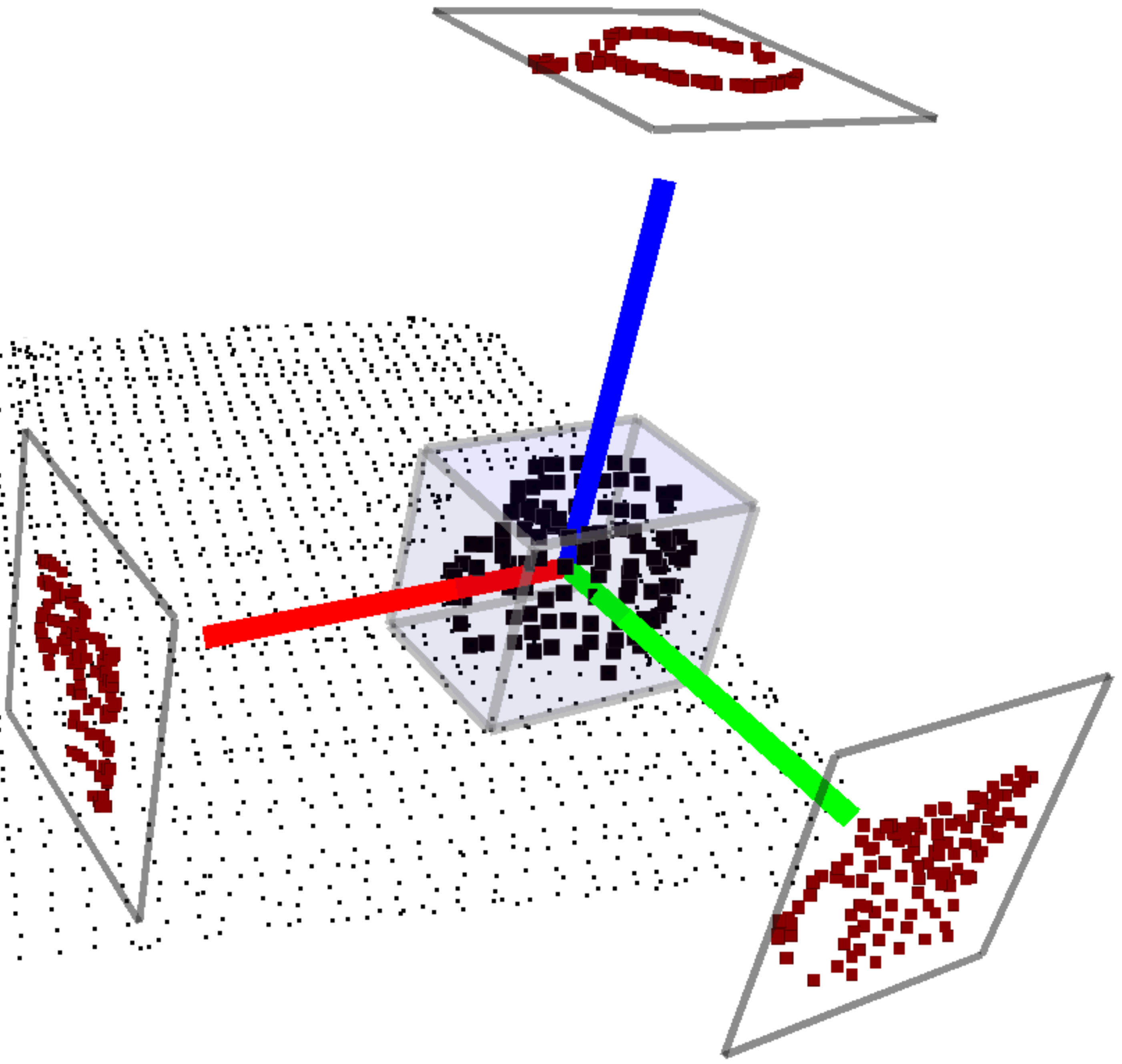}
&\quad\quad \includegraphics[width=0.35\textwidth]{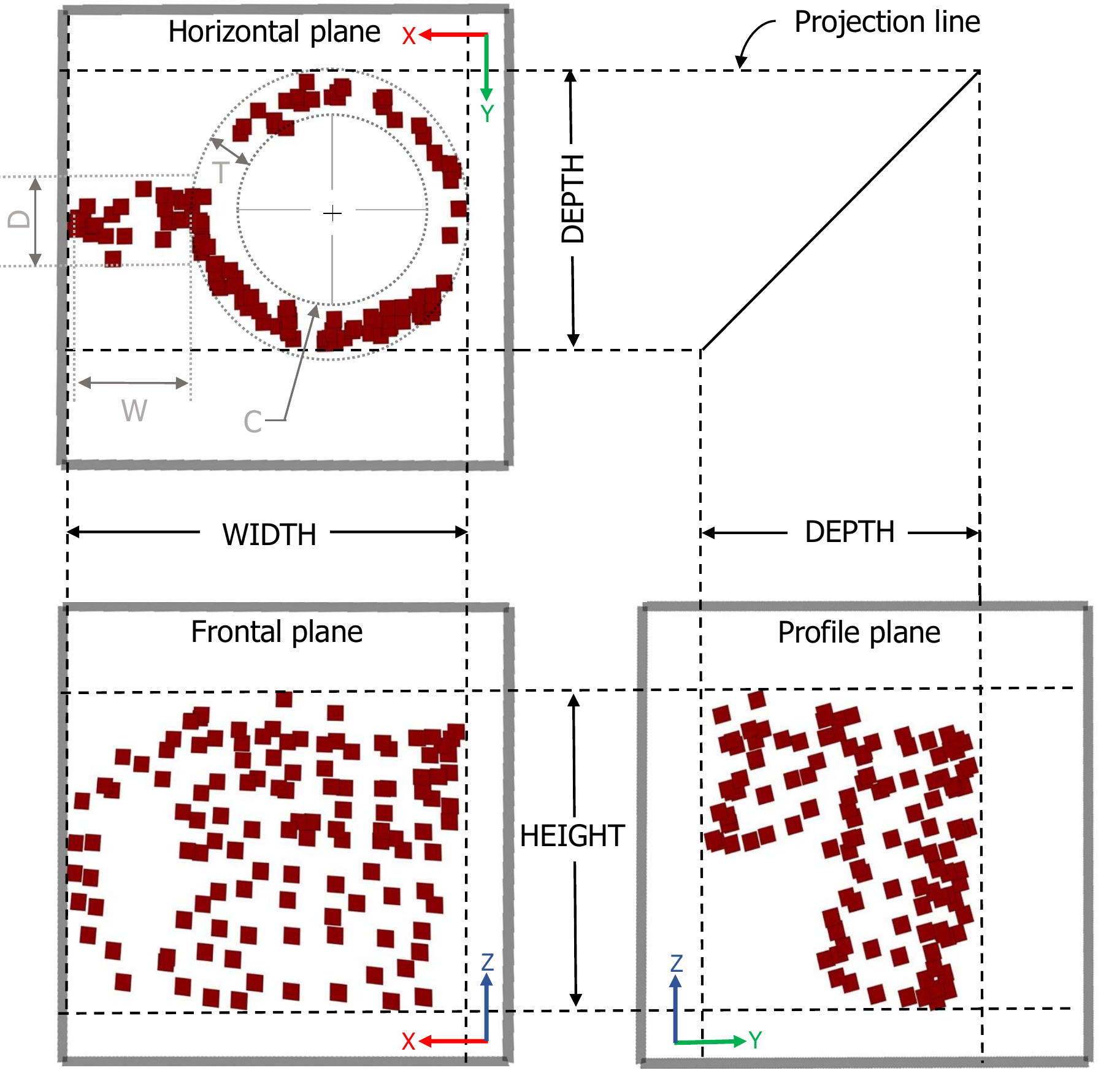} \\
 \emph{(a)} &\quad\quad \emph{(b)}\\
\end{tabular} 
\vspace{-3mm}
\caption{Example of how the projections used to build GOOD can also be used for extracting features relevant for object manipulation (see text): (a) Local reference frame and projections; (b) Projections in multi-view layout.}
\vspace{-3mm}
	\label{fig:object_manipulation}       % Give a unique label
\end{figure}

In Fig.~\ref{fig:object_manipulation}, again we consider the projections of a mug. Here, we adopt a multi-view orthographic projection layout in which there is a central or front view, a top view and a side view. The central view is the one selected based on the entropy criterion and appearing in the first part of the descriptor. The top view contains the projection in the orthogonal plane formed by the horizontal axis of the central projection and the third axis. The side view contains the projection in the orthogonal plane formed by the vertical axis of the central projection and the third axis. The figure shows that projections can be further processed for object manipulation purposes. In the top view, the grey symbols $C$, $W$, $D$ and $T$ represent how the projection can be further processed and some features for manipulation are extracted, namely inner radius ($C$), thickness ($T$), handle length ($W$) and handle thickness ($D$).

%%%%%%%%%%%%%%%%%%%%%%%%%%%%%%%%%%%%%%%%%%%%%%%%%%%%%%%%%%%%%%%%%%%%%%%%%%%%%%%%%%%%%%
%%%%%%%%%%%%%%%%%%%%%%%%%%%%%%%%%%%%%%%%%%%%%%%%%%%%%%%%%%%%%%%%%%%%%%%%%%%%%%%%%%%%%%

\section {Representations based on Local Features}
\label{sec:local_features_new}
Global object representations are widely used in shape retrieval and object recognition techniques, but these features are sensitive to clutter and occlusion. 
To cope with this limitations, local features can be used \citep{mian2006three,bariya20123d}. The main motivation of this section is to explore object representations at different levels of abstraction. We present three object representation methods based on local features. In the first approach, an object is represented by a set of local features. The second approach represents an object based on the \emph{Bag of Visual Words} technique, i.e., as a histogram of visual words (classes of local features). Finally, we propose an extension of \emph{Latent Dirichlet Allocation} (LDA) to learn structural semantic features (i.e., topics) from low-level feature co-occurrences for each category incrementally and independently. 

\begin{figure}[!t]
 \center
 \includegraphics[width=0.88\textwidth]{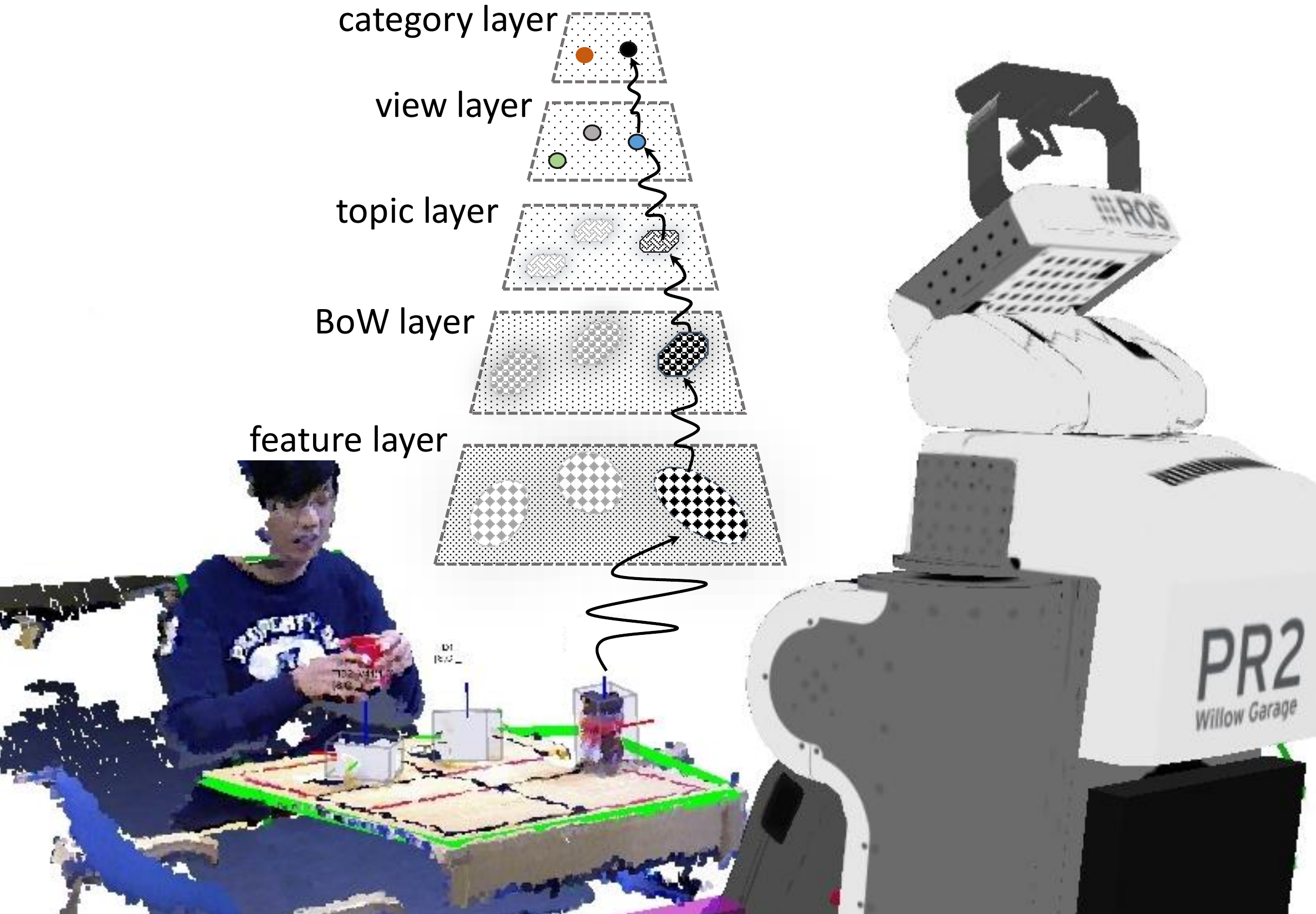}
 \caption{The proposed data processing layers being tested on a service robot.}
\label{fig:general_over_view}       % Give a unique label
\end{figure}

All the approaches are designed to be used by a service robot working in a domestic environment. Fig.~\ref{fig:general_over_view} shows a PR2 robot looking at some objects on the table. Tabletop objects (identified by different bounding boxes) are tracked and processed sequentially through five layers. For instance, to describe an object view using the feature layer, a spin-image shape descriptor \citep{johnson1999using} is used to represent the local shapes of the object in different key points. The object therefore is represented by a set of spin-images. When using the Bag-of-Words (BoW) layer, the given object view is described by a histogram of local shape features, as defined in Bag-of-Words models. In the topic layer, each topic is defined as a discrete distribution over visual words. Depending on how the system is configured, object views are represented in the object view layer as sets of local features, histograms of visual words, or histograms of LDA topics.
Therefore, the content of the object view layer depends on the selected object representation approach. Finally, the category model is updated by incorporating the obtained object representation (category layer). In following subsections, we present and discuss each approach in detail.

\subsection {Object Views Represented by Sets of Spin-images}
\label{sec:local_features}
In this subsection, we adopt an approach to object representation in which objects are described by sets of local shape features called spin-images~\citep{johnson1999using}. The reason why we use spin images rather than other 3D feature descriptors is that the spin-image is pose invariant, and therefore suitable for 3D perception in autonomous robots. Another advantage of the spin-image is that only a repeatable normal - rather than a full reference frame - is required to compute the local descriptor. 

The process of local feature extraction consists of two main phases: extraction of keypoints and computation of spin images. For efficiency reasons, the number of keypoints in an object should be much smaller than the total number of points. For keypoint extraction, first, a voxelized grid approach is used to obtain a smaller set of points. The nearest neighbor point to each voxel center is selected as a keypoint. Afterwards, the spin-image descriptor is used to encode the surrounding shape in
each keypoint using the original point cloud. Figure \ref{fig:feature_extraction} (\emph{a} and \emph{b}) shows an example of the keypoint extraction process. In the second stage, spin-image descriptors are computed for each keypoint in order to describe the shape surrounding the keypoint. A spin-image is a local shape histogram obtained by projecting the 3D surface points onto the tangent plane of the keypoint. The normal vector of the tangent plane is called surface normal. Then, each point is represented by a pair ($\alpha$, $\beta$), where $\alpha$ is the distance to the surface normal of the keypoint, i.e., the radius, and $\beta$ is the perpendicular distance from the point to the tangent plane: 
\begin{figure}[!t] %\label{fig:spin_images}
\begin{tabular}[width=1\textwidth]{c}
\begin{tabular}[width=1\textwidth]{cccc}
 \includegraphics[width=0.18\textwidth]{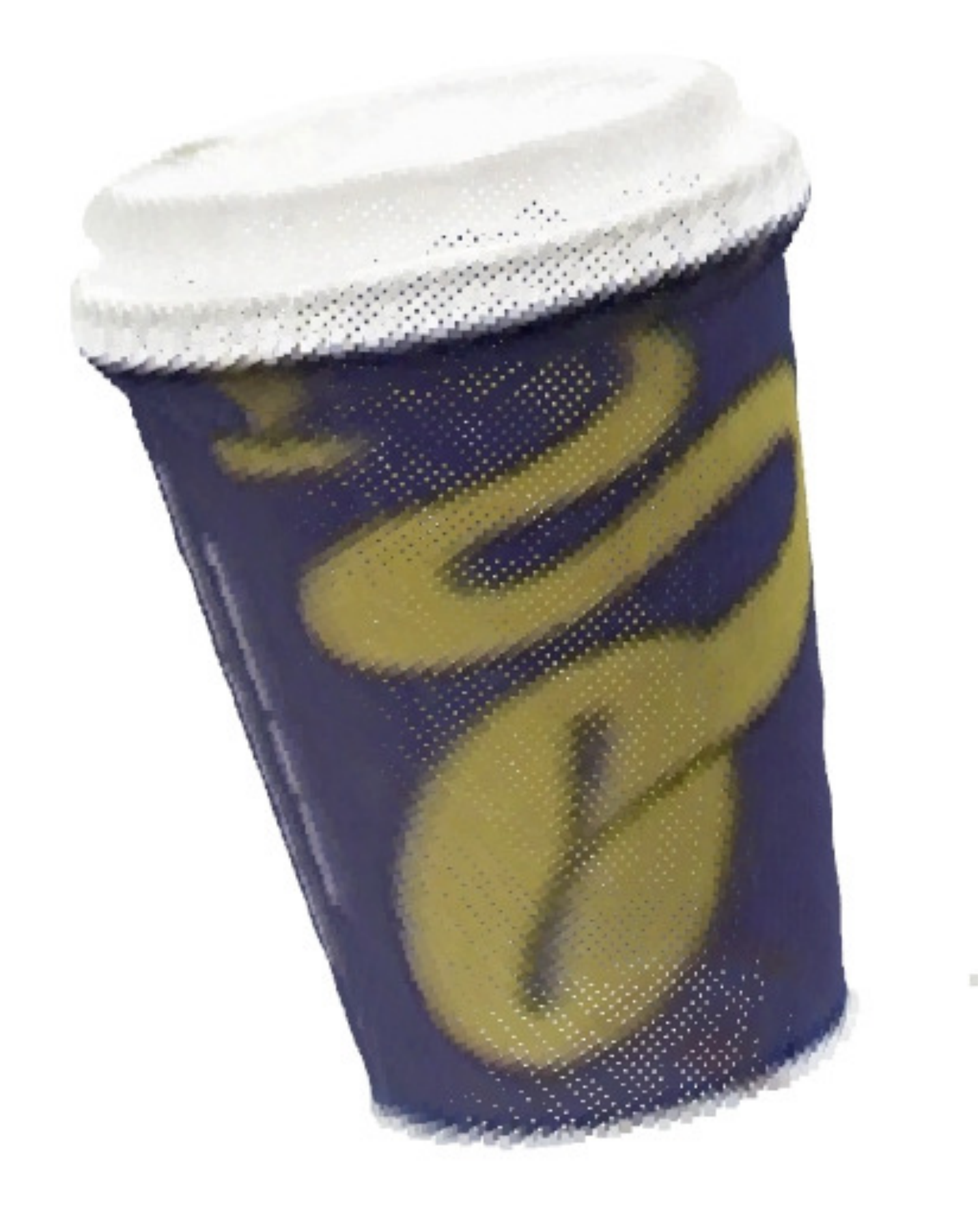} &
 \includegraphics[width=0.18\textwidth]{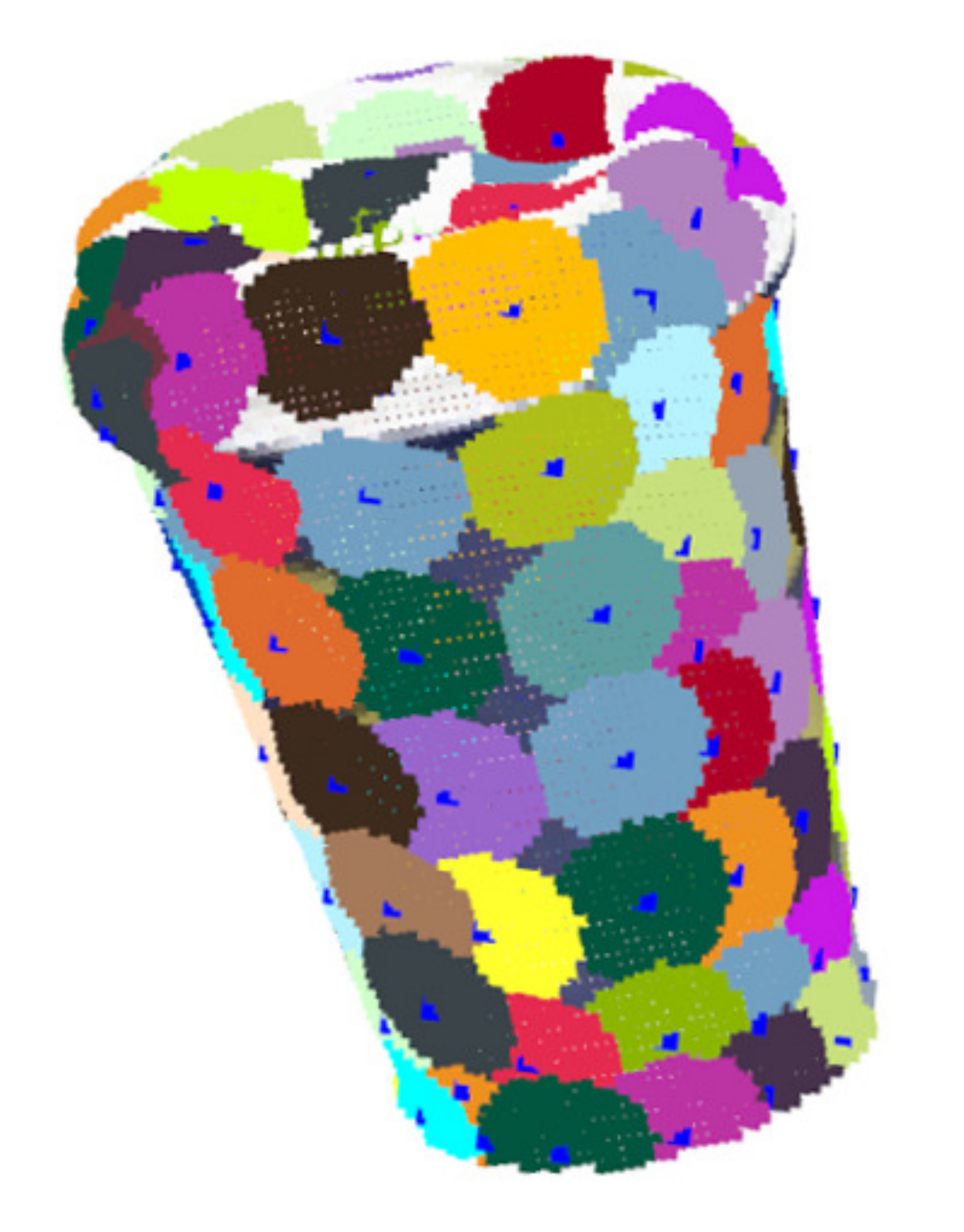}&
 \includegraphics[width=0.28\textwidth]{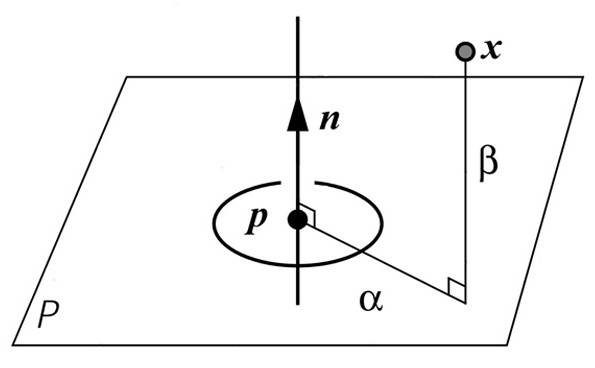} &
  \includegraphics[width=0.22\textwidth]{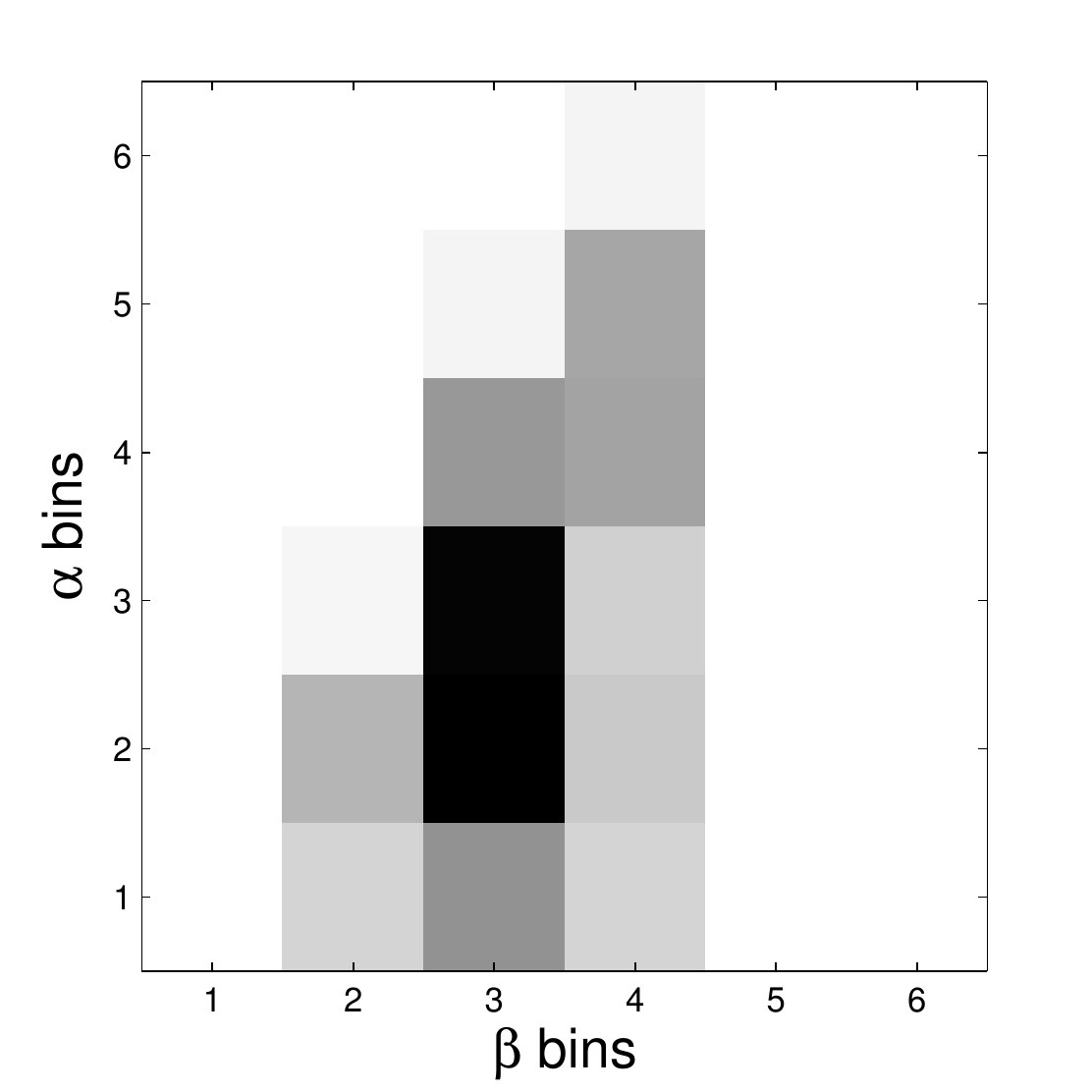} \\ 
 \emph{(a)} & \emph{(b)}&\emph{(c)} & \emph{(d)}\\
\end{tabular}\\
\end{tabular}
% figure caption is below the figure
\caption{Object representation for a coffee cup : \emph{(a)} point cloud of a coffee cup \emph{(b)} key-point selection (blue points); the spin-image descriptor is used to encode the surrounding shape in each key-point (i.e., highlighted by different colours). 
	\emph{(c)} a schematic of how spin-image is computed for a keypoint $p$;
	\emph{(d)} a computed spin image.}
\label{fig:feature_extraction}       % Give a unique label
\end{figure}

\begin{equation}
\alpha = 	\sqrt{ 	\parallel	\textbf{x}-\textbf{p}	\parallel^2 -
	\big(\textbf{n} \cdot (\textbf{x}-\textbf{p})\big)^2},
\end{equation}
\begin{equation}
\beta = \textbf{n} \cdot (\textbf{x}-\textbf{p}),
\end{equation}
\noindent where \textbf{n} is the surface normal for keypoint \textbf{p}. Every spin-image
bin counts, for a given neighborhood of points around the keypoint, the number
of neighbors that fall in a given range of $\alpha$ and $\beta$. The
procedure is illustrated in Fig.~\ref{fig:feature_extraction} (\textit{c}) and
(\textit{d}).

To compute a spin-image, the following parameters must be specified\footnote{http://pointclouds.org}:
\begin{itemize}
\item Image width ($\operatorname{IW}$): defines the resolution of the spin-image, which will be $\operatorname{IW}+1$ bins in the radius dimension and $2 \times \operatorname{IW}+1$ bins in the distance dimension.
\item Support length ($\operatorname{SL}$): determines the amount of space swept out by a spin-image, which will have a radius of $\operatorname{SL}$ and a height of $2\times \operatorname{SL}$.
\item Support angle ($\operatorname{A}$): maximum angle between the surface normal at the keypoint and the surface normal in other points to be included as neighbors.
\end{itemize}

\begin{figure}[!t] %\label{fig:spin_images}
\hspace{-1cm}
\begin{tabular}[width=1\textwidth]{ccccc}
 \includegraphics[width=0.14\textwidth, trim= 0cm 18.5cm 2cm 0cm, clip=true]{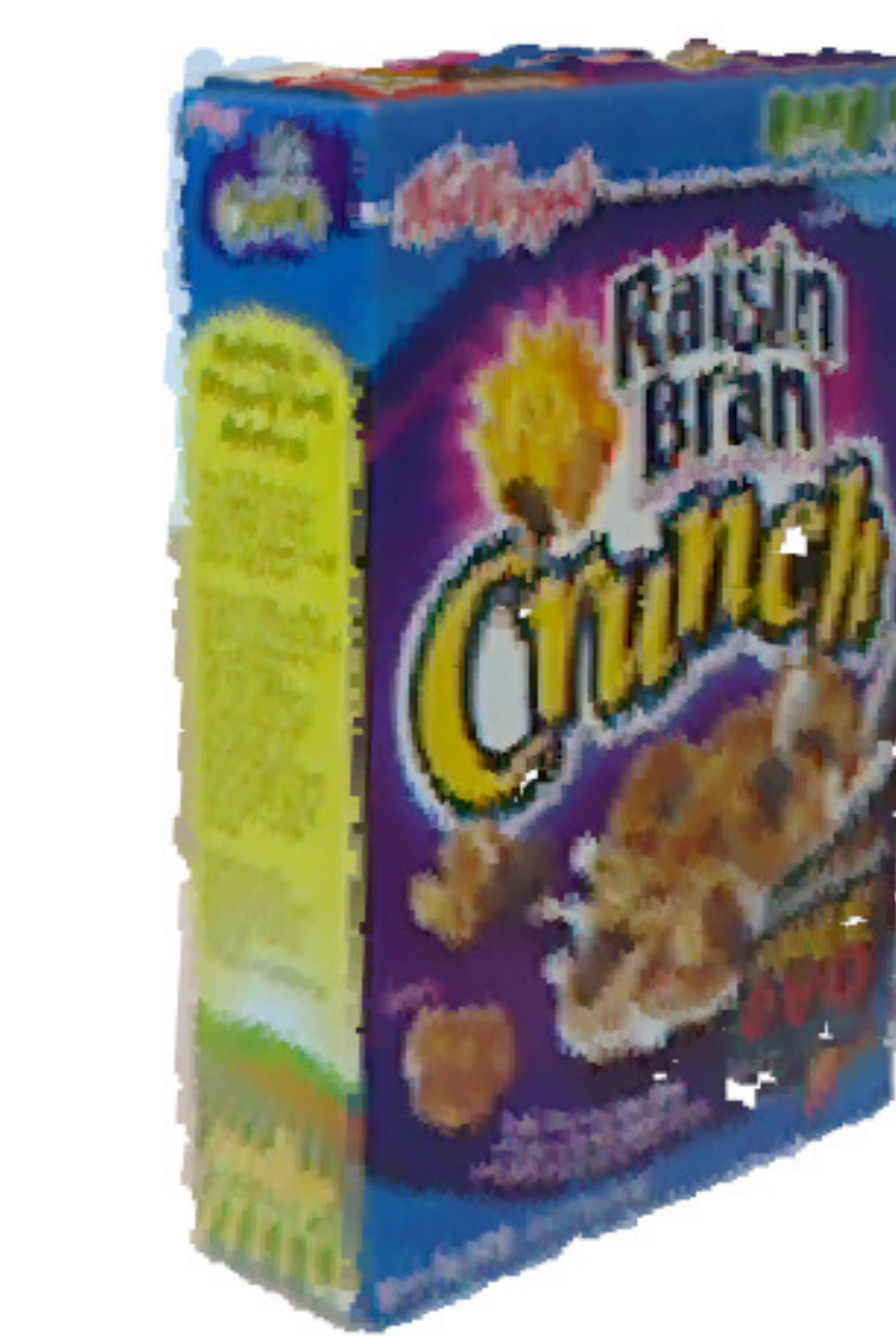} &
  \includegraphics[width=0.15\textwidth, trim= 3cm 18.5cm 3cm 0cm, clip=true]{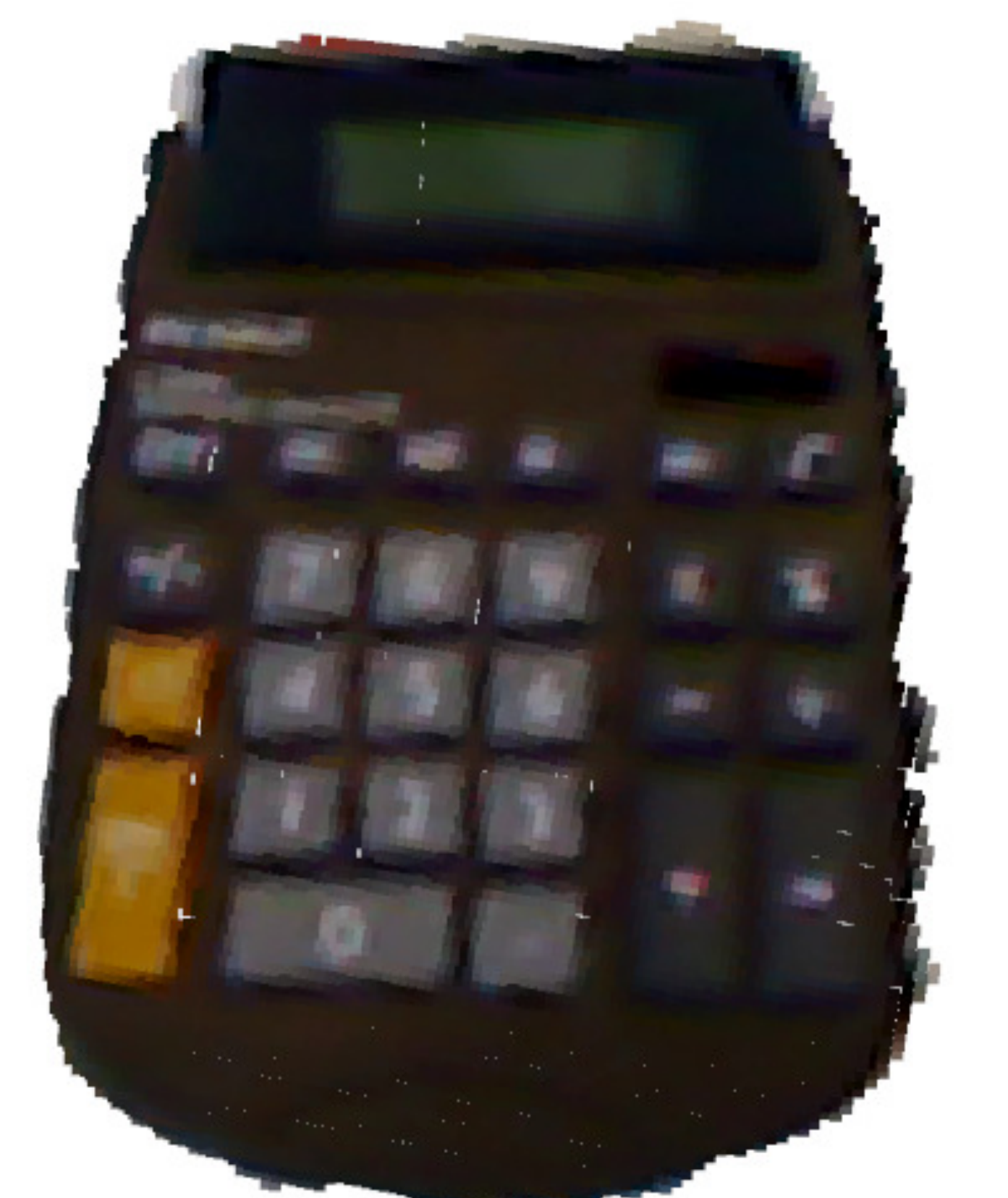} &
 \includegraphics[width=0.17\textwidth, trim= 0cm 18.5cm 0cm 0cm, clip=true]{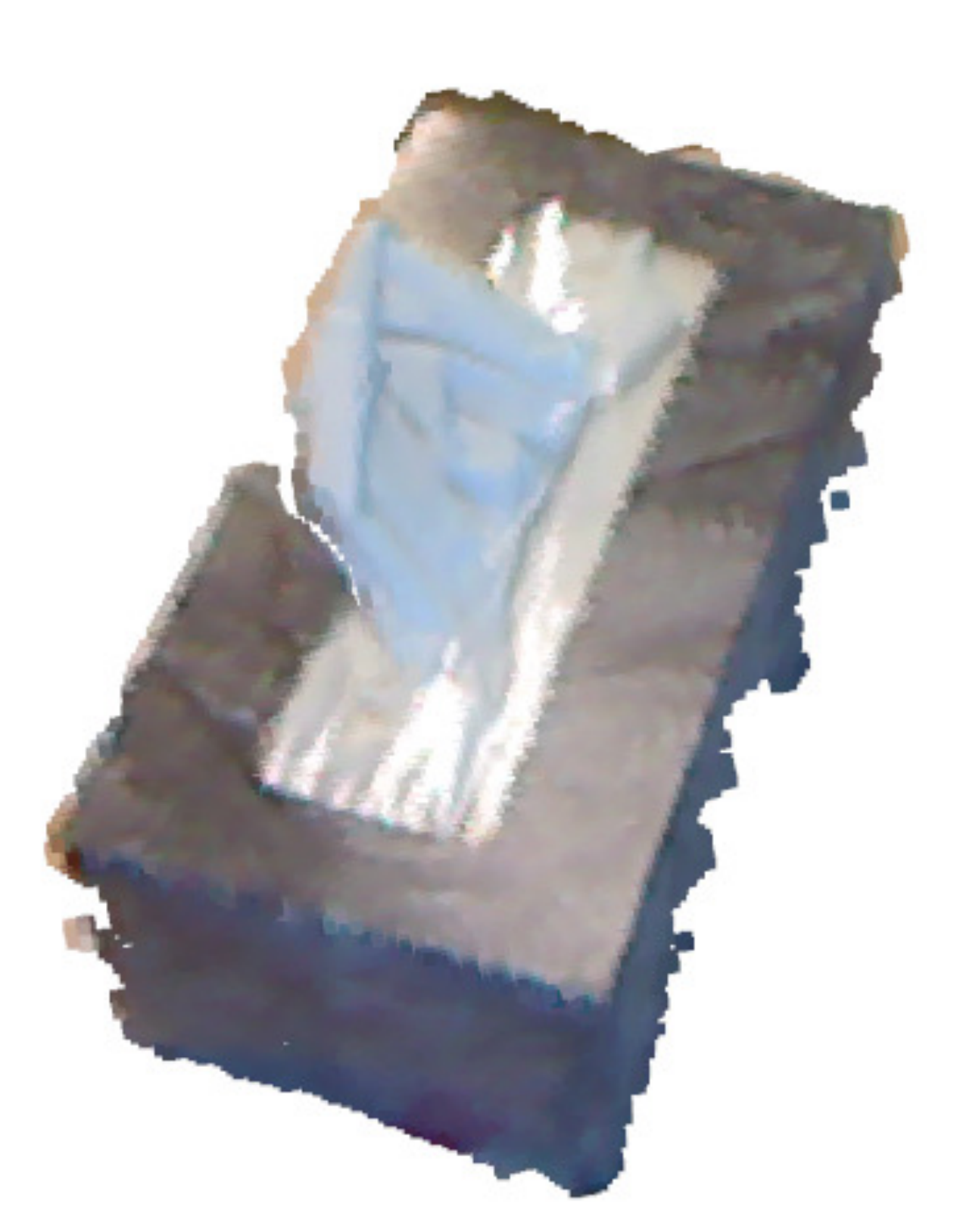}&
  \includegraphics[width=0.21\textwidth, trim= 1cm 17cm 2cm 0cm, clip=true]{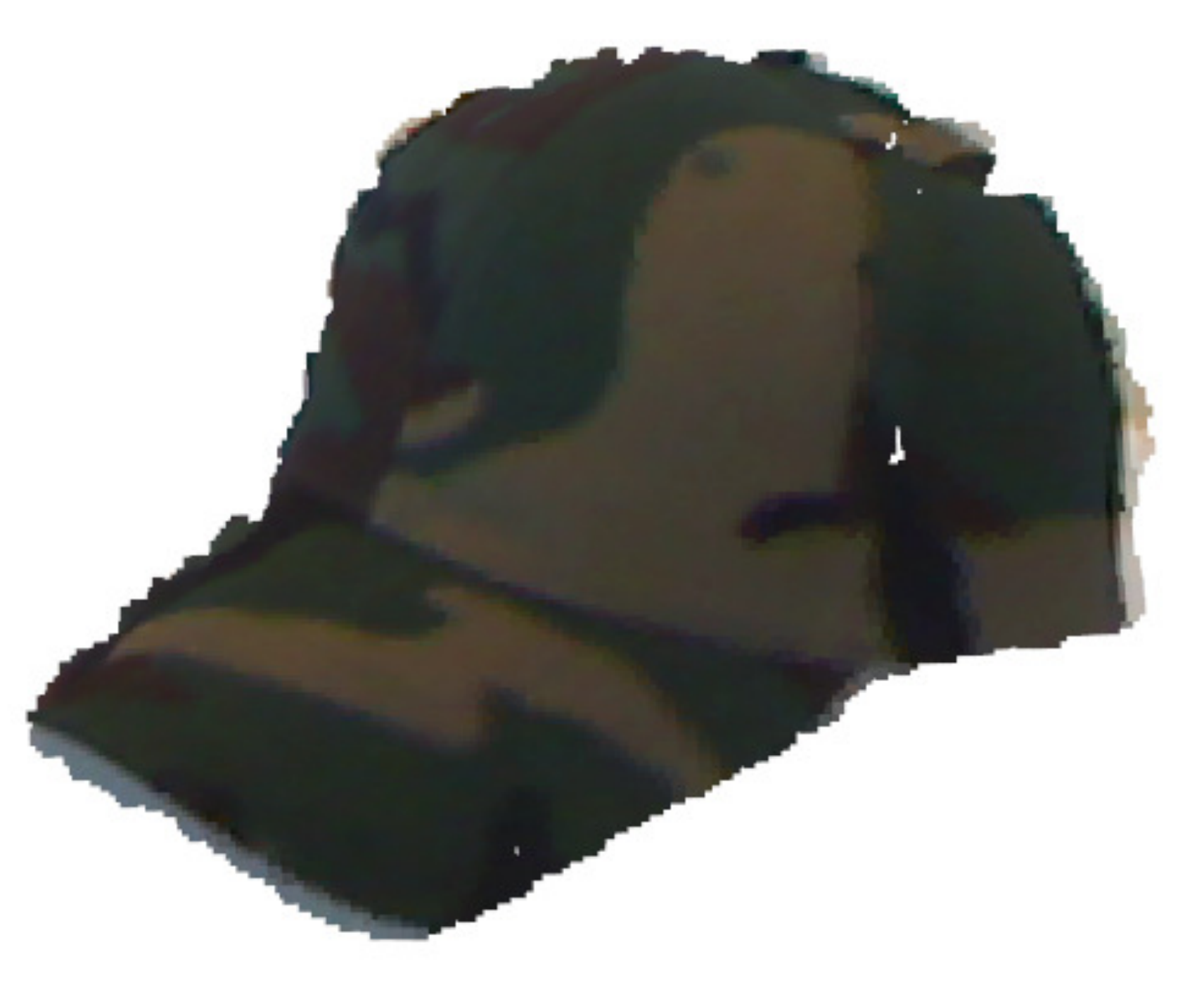}&\hspace{-4mm}
 \includegraphics[width=0.19\textwidth, trim= 0cm 17cm 1cm 0cm, clip=true]{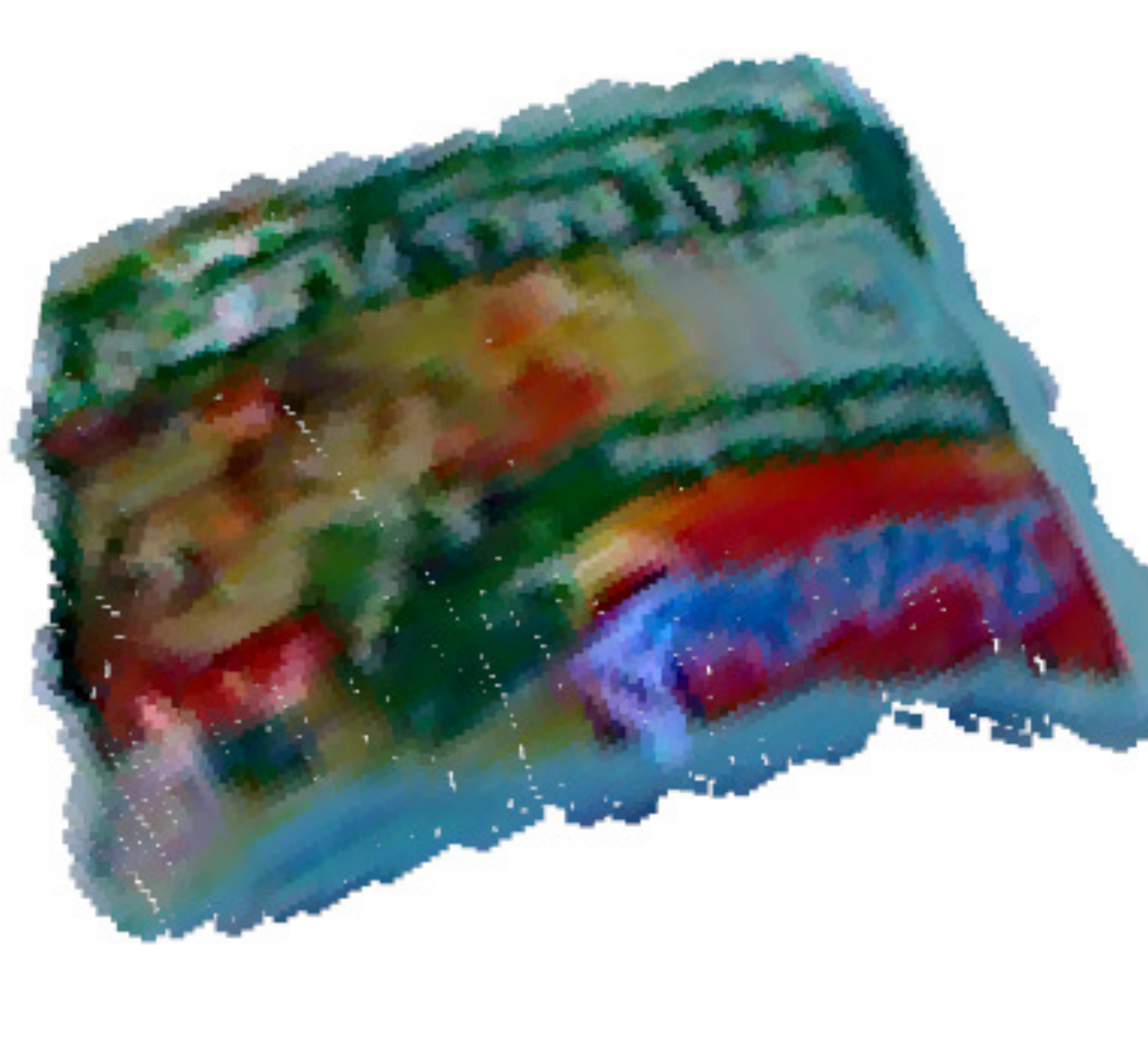}\\
  \includegraphics[width=0.14\textwidth, trim= 0cm 0cm 2cm 18.5cm, clip=true]{Figures/box.pdf} &
    \includegraphics[width=0.16\textwidth, trim= 3cm -3cm 3cm 18.5cm, clip=true]{Figures/calculator.pdf} &
 \includegraphics[width=0.18\textwidth, trim= 0cm 0cm 0cm 18.5cm, clip=true]{Figures/kylenex.pdf}&
  \includegraphics[width=0.22\textwidth, trim= 1cm 0cm 2cm 17cm, clip=true]{Figures/cap.pdf}&\hspace{-4mm}
 \includegraphics[width=0.2\textwidth, trim= 0cm 0cm 0cm 17cm, clip=true]{Figures/noddel.pdf}\\
     \includegraphics[width=0.18\textwidth, trim= 0cm 15cm 0cm 0cm, clip=true]{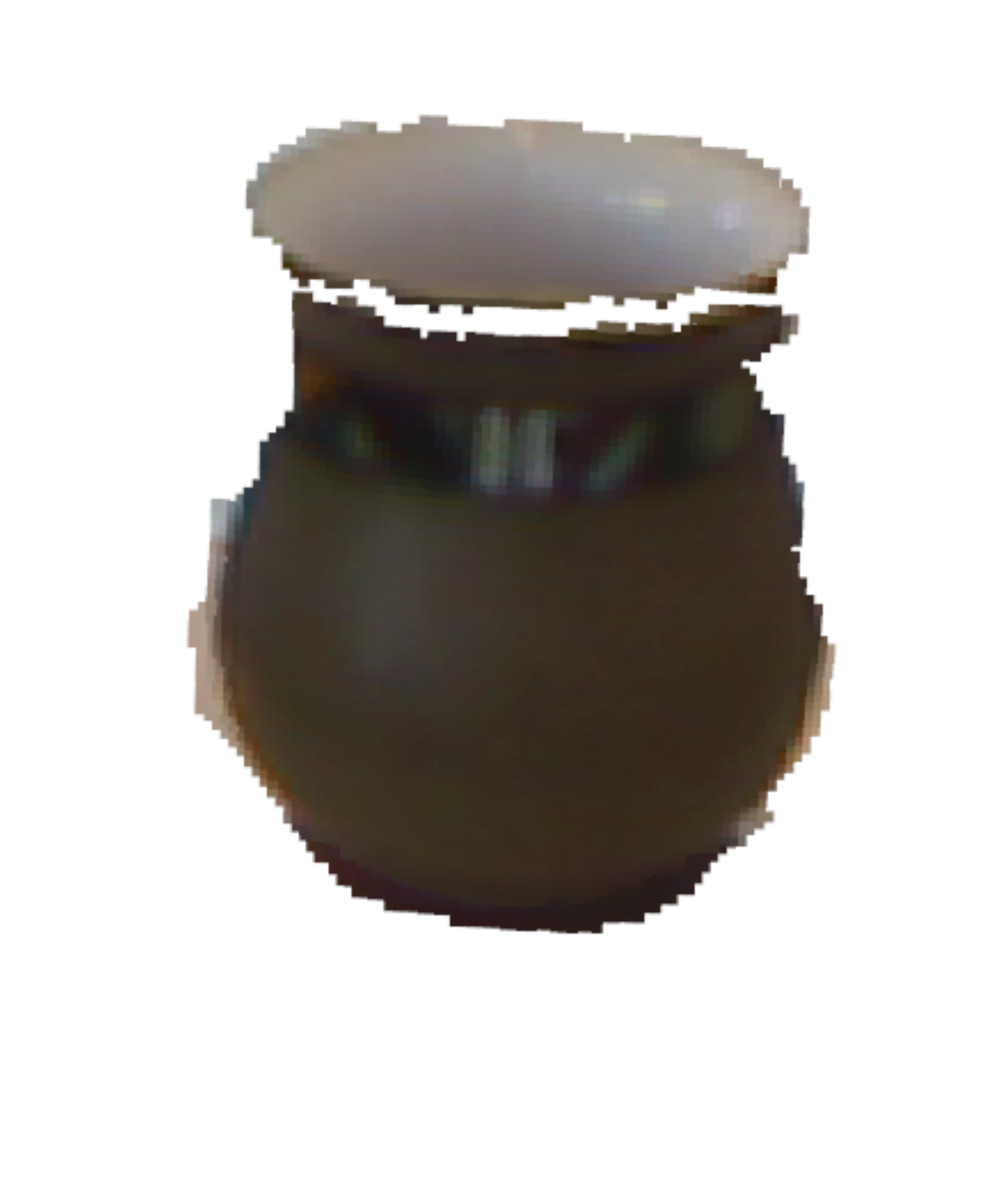} &
  \includegraphics[width=0.18\textwidth, trim= 0cm 15cm 0cm 0cm, clip=true]{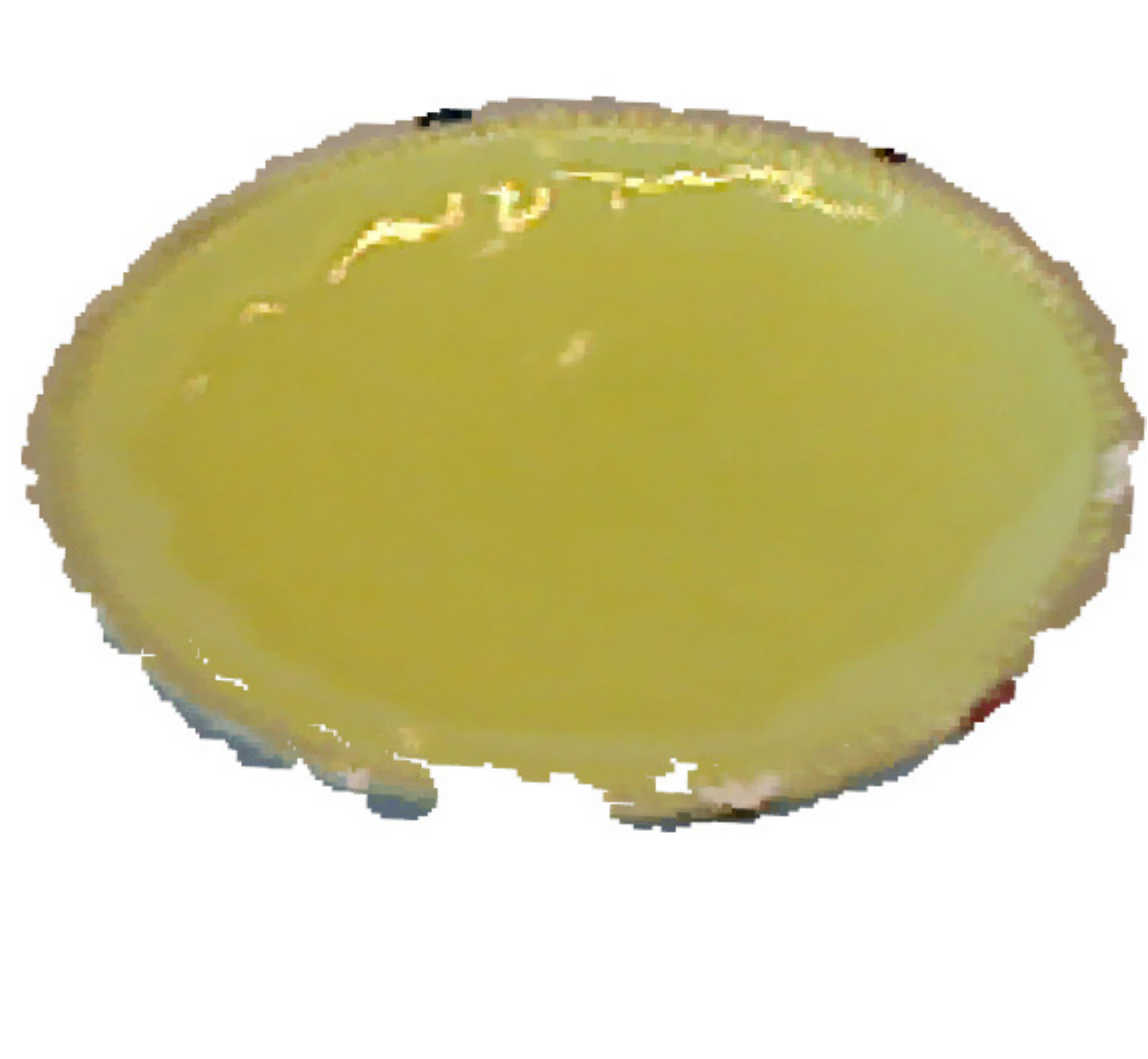} &
 \includegraphics[width=0.18\textwidth, trim= 0cm 16.5cm 1cm 0cm, clip=true]{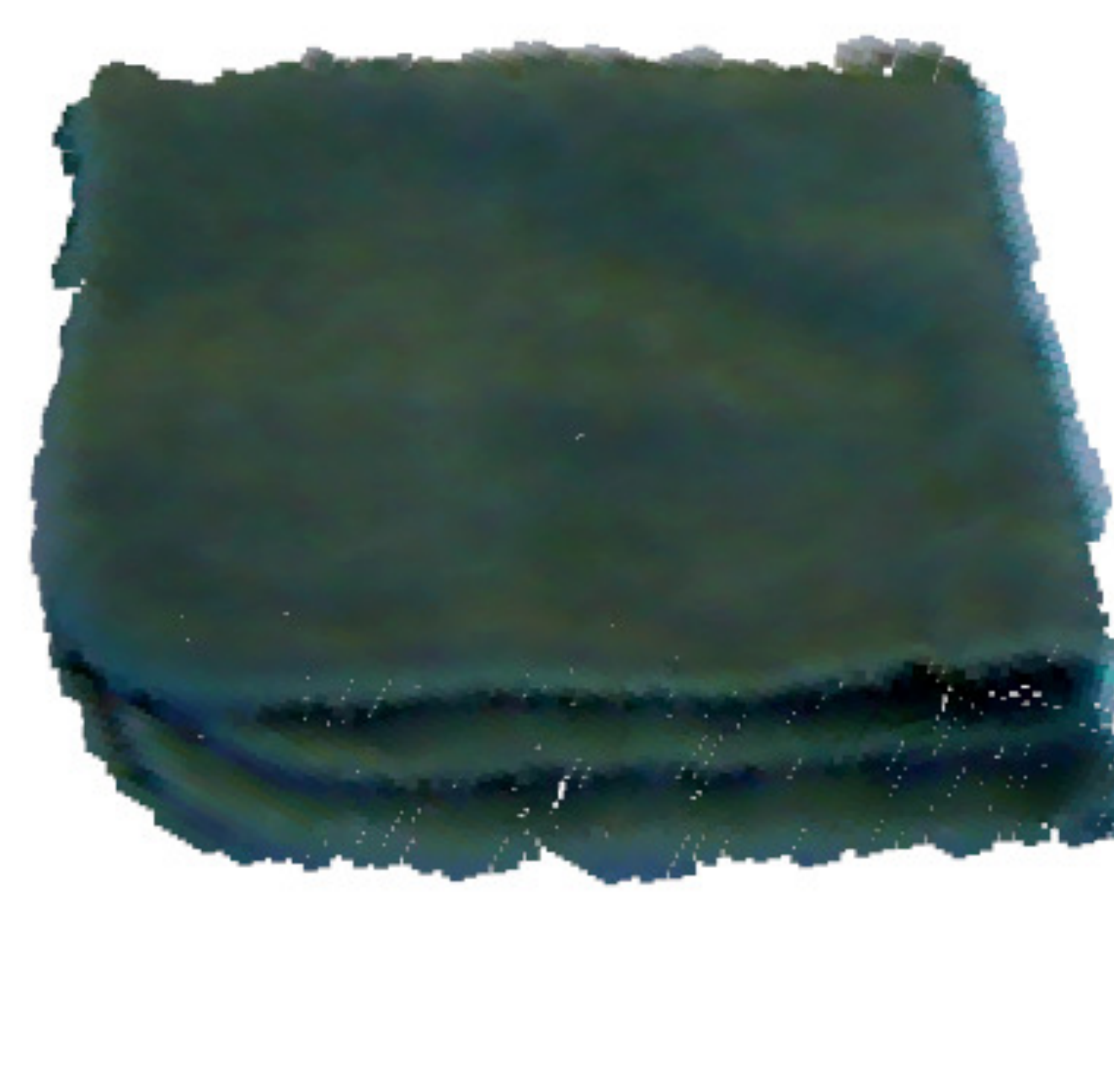}&
 \includegraphics[width=0.14\textwidth, trim= 2cm 17cm 1cm 0cm, clip=true]{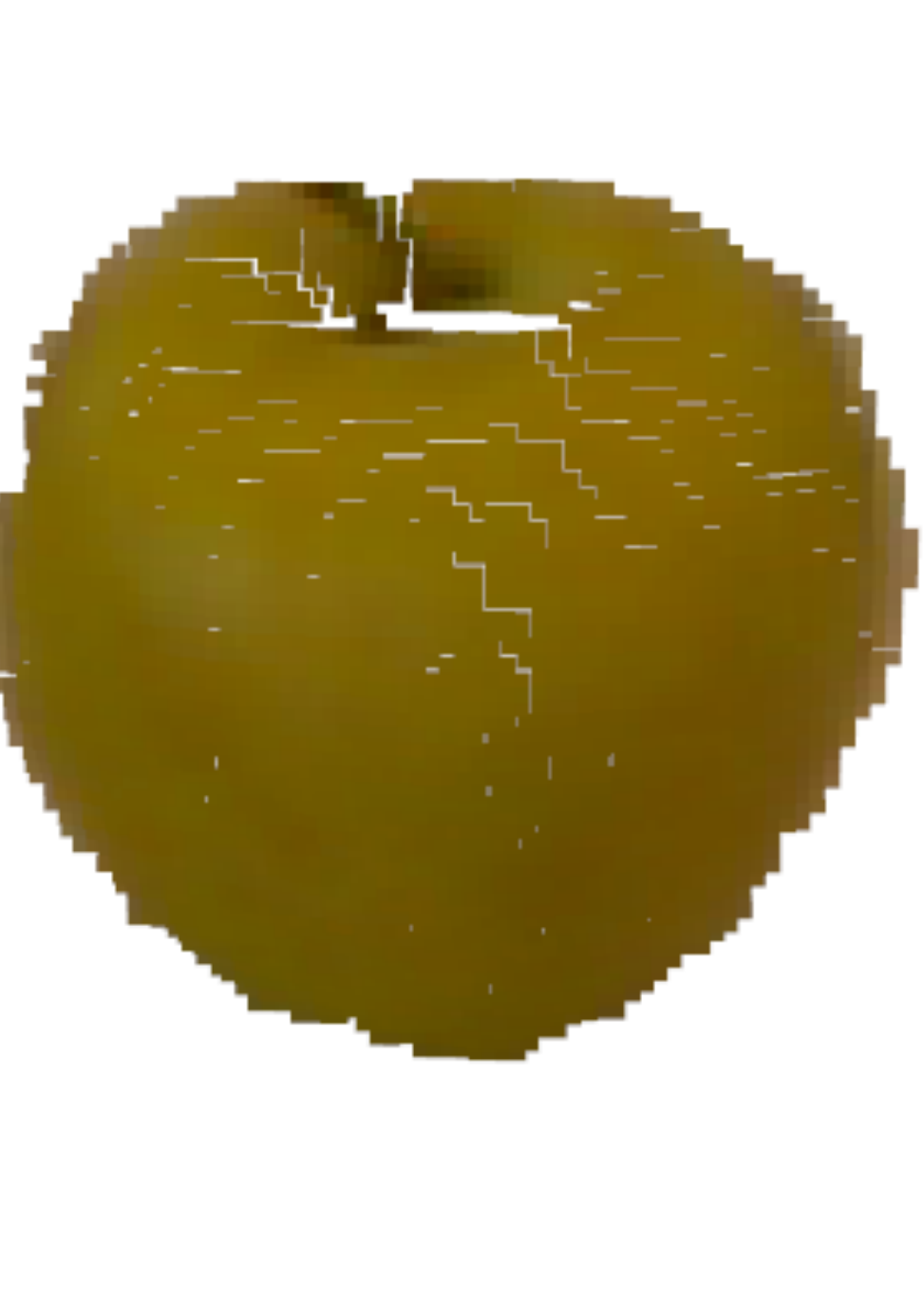}&\hspace{-4mm}
 \includegraphics[width=0.23\textwidth, trim= 0.5cm 14cm 1cm 0cm, clip=true]{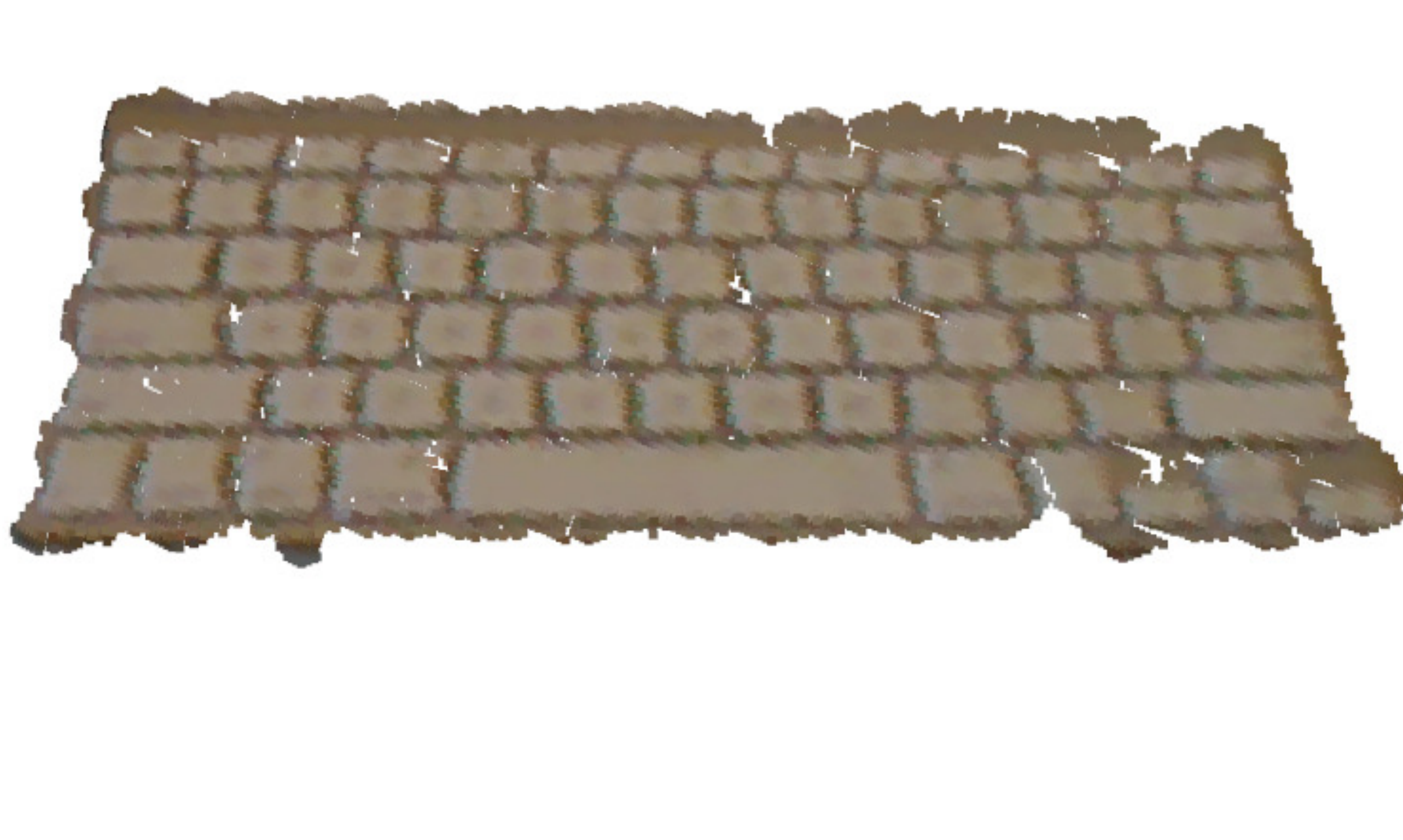}\\
    \includegraphics[width=0.18\textwidth, trim= 0cm -4cm 0cm 18cm, clip=true]{Figures/coffe_cup2.pdf} &
  \includegraphics[width=0.18\textwidth, trim= 0cm -6cm 0cm 15cm, clip=true]{Figures/plate.pdf} &
 \includegraphics[width=0.18\textwidth, trim= 0cm -4cm 1cm 18.5cm, clip=true]{Figures/towel.pdf}&
  \includegraphics[width=0.14\textwidth, trim= 1.5cm -4cm 1cm 20cm, clip=true]{Figures/appel.pdf}&\hspace{-4mm}
  \includegraphics[width=0.23\textwidth, trim= 0.5cm -6cm 1cm 15cm, clip=true]{Figures/keyboard.pdf}\\
  \end{tabular}
\vspace{-1cm}
\caption{Example of keypoints detected by the proposed approach on different object views of the Washington RGB-D dataset \citep{Lai2011}. }
\label{fig:keypoint}       % Give a unique label
\end{figure}

\noindent Finally, each object view, $\textbf{O}$, is described by a set of spin-images,

\begin{equation}
\textbf{O}^{\textbf{f}} = \{s_1,~s_2,~\dots~,~s_N\}
\end{equation}
\noindent where $N$ is the number of key-points. In the next chapter, we will describe how these features are used both for conceptualizing and recognizing object categories. Examples of keypoints, detected by the proposed approach, on a set of object views of the Washington RGB-D dataset \citep{Lai2011} are shown in Fig.~\ref{fig:keypoint}.

%^^^^^^^^^^^^^^^^^^^^^^^^^^^^^^^^^^^^^^^^^^^^^^^^^^^^^^^^^^^^^^^^
%^^^^^^^^^^^^^^^^^^^^^^^^^^^^^^^^^^^^^^^^^^^^^^^^^^^^^^^^^^^^^^^^
\subsection {Object Views Represented by Histograms of Visual of Words}
\label{sec:BoW}

\begin{figure}[!t]
 \center
 \includegraphics[width=\textwidth, trim= 6cm 2.5cm 5cm 1cm, clip=true]{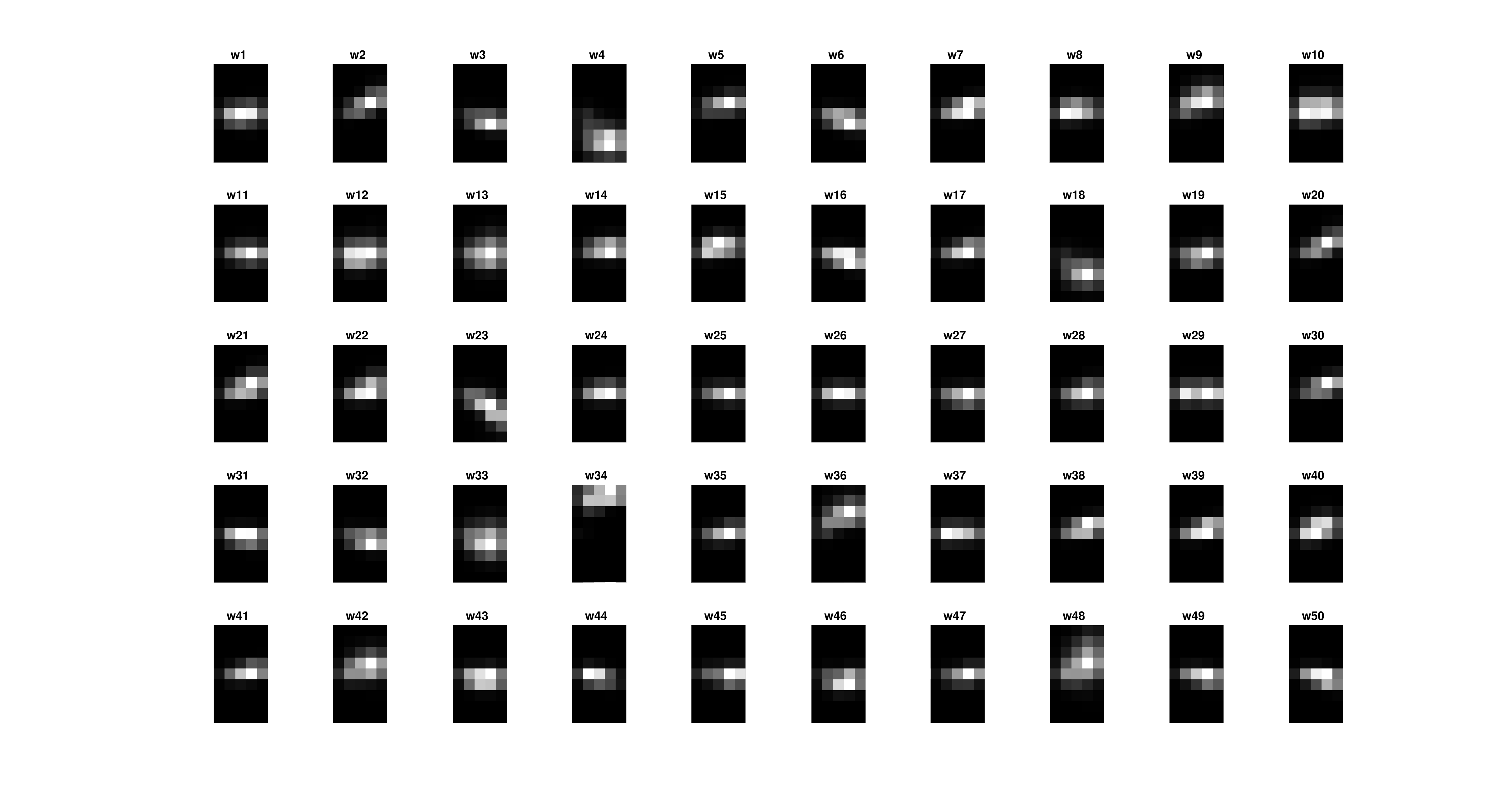}
% figure caption is below the figure
\caption{A visual word dictionary with 50 words: spin-images compiled with IW = $4$ bins and SL = $0.05$m.}
\label{fig:dictionary}       % Give a unique label
%\end{wrapfigure}
\end{figure}

As we discussed earlier, comparing 3D objects by their local features tends to be computationally expensive. To address this problem, a BoW approach can be adopted for object representation, i.e., objects are described by histograms of local shape features. This approach requires a dictionary of visual words. Usually, this dictionary is created off-line through clustering of a given training set. In open-ended learning scenarios, there is not a set of training data available at the beginning of the learning process. To cope with this limitation, we look at human cognition, in particular at the fact that human babies explore their environment in a playful (arbitrary) way \citep{Smith2005}. Therefore, we propose that the robot freely explores several scenes and collects several object experiences. Towards this goal, unsupervised object discovery, as proposed in chapter \ref{chapter_3}, section \ref{sec:unsupervisedExperienceGathering}, is carried out in the environment while the robot operates. Only the modules directly involved in object discovery and dictionary building are active at this stage. The robot seeks to segment the world into ``object'' and ``non-object''. The output of object exploration is a pool of object candidates. It should be noted that, to balance computational efficiency and robustness, a downsampling filter is applied to obtain a smaller set of points distributed over the surface of the object. Subsequently, to construct a pool of features, spin-images are computed for the selected points extracted from the pool of object candidates. We use a PCL function to compute spin-images\footnote {In this work, we computed around 32000 spin-images from the point cloud of the 194 objects.}. Finally, the dictionary is constructed by clustering the features using the k-means algorithm \citep{hartigan1979algorithm}. The centers of $V$ computed clusters are defined as the visual words, $\textbf{w}_i$ ($1\leq i \leq V$). Figure \ref{fig:dictionary} shows a dictionary containing 50 words. In the implementation, we tested different dictionary sizes.

When using the BoW approach, object views (instances) are described in the object view layer by histograms of frequencies of visual words. As depicted in Fig.~\ref{fig:general_over_view}, the input to this layer is the set of spin-image features of an object candidate, $\textbf{O}$, which is computed as described in the previous section. Afterwards, by searching for the nearest neighbor in the dictionary, each spin image is assigned to a visual word. Therefore, the given object is first represented as a set of visual words, $\{{w}_1, {w}_2, ..., {w}_N\}$, where each entry represents one of the $V$ words of the dictionary, and $N$ is the number of features of the given object. The given object is then represented as a histogram of occurrences of visual words:

\begin{equation}
	\textbf{O}^{\textbf{b}} = \operatorname{\textbf{h}} = [h_1, ~ h_2, ~ ...~,~ h_V],
	\label {eg:object_representation}
\end{equation}

\noindent 
where the $i^{th}$ element of $\textbf{h}$ is the count of the number spin-images assigned to a visual word, $\textbf{w}_i$, and $V$ is the size of the dictionary. The obtained histogram is dispatched to the \emph{Object Recognition} module or to the \emph{Object Conceptualization} module. Furthermore, in order to build a more compact and complex object representation, the set of visual words of the object, i.e., $\{{w}_1, {w}_2, ..., {w}_N\}$, is given as input to the \emph{topic layer}.

%^^^^^^^^^^^^^^^^^^^^^^^^^^^^^^^^^^^^^^^^^^^^^^^^^^^^^^^^^^^^^^^^
%^^^^^^^^^^^^^^^^^^^^^^^^^^^^^^^^^^^^^^^^^^^^^^^^^^^^^^^^^^^^^^^^

\subsection {Object Views Represented by Histograms of LDA Topics}
\label{sec:topic_modelling}

In this section, we propose two variant of Latent Dirichlet Allocation (LDA) \citep {blei2003latent}.
In both LDA-based approaches, each object is represented as a finite mixture over a set of latent variables (i.e., topics), which are expected to summarize the word-order (semantic) information. One of these approaches develops local (per category) topics \citep{kasaei2016hierarchical}  whereas the other develops global topics, shared among all categories1, as in standard LDA \citep {blei2003latent}.

\subsubsection {Local LDA}
In the topic layer, a statistical model is used to get structural semantic features from low-level feature co-occurrences. The basic idea is to represent the objects as random mixtures over topics (i.e., latent variables), where each topic is characterized by a distribution over visual words (i.e., observed variables). Both random distributions are defined using multinomial distributions and their parameters are characterized by a Dirichlet distribution. It must be pointed out that we are using shape features rather than semantic properties to encode the statistical structure of object categories~\citep{kim2009adaptation}. 
It is easier to explain the details using an example. We start by selecting a category label, for example \emph{Mug}. To represent a new instance of Mug, a distribution over Mug topics is drawn that will specify which topics should be selected for generating each visual word of the object. According to this distribution, a particular topic is selected out of the mixture of possible topics of the Mug category for generating each visual word in the object. For instance, a \emph{Mug} usually has a handle, and a ``handle'' topic refers to some visual words that frequently occur together in handles. The process of drawing both the topic and visual word is repeated several times to choose a set of visual words that would construct a \emph{Mug}. We use statistical inference techniques for inverting this process to automatically find out a set of topics for each category from a collection of instances.
In other words, we try to learn a model for each category (a set of latent variables) that explains how each object obtains its visual words.

Towards this end, the obtained representation from the BoW layer is presented as input to the \emph{topic layer}. The LDA model contains parameters at three levels: including category-level parameters (i.e., $\alpha$), which are sampled once in the process of generating a category of objects; object-level variables (i.e., $\theta_{o}$),  which are sampled once per object, and word-level variables (i.e., $\textbf{z}_{o,n}$ and $w_{o,n}$), which are sampled every time a feature is extracted. 
The variables $\theta$, $\phi$ and $\textbf{z}$ are latent variables that should be inferred. Assume everything is observed and a category label is selected for each object; i.e., each object belongs to one category. The joint distribution of all hidden and observed variables for a category is defined as follows:
\vspace{-1mm}
\begin{equation}
\label {joint_distribution1}
\footnotesize{
\begin {split}
		p^{(c)}(\textbf{w}, \textbf{z}, \theta, \phi| \alpha , \beta) = 
 \prod_{z=1}^{K} p^{(c)}(\phi_z|\beta)\prod_{o=1}^{|c|}
p^{(c)}(\theta_o|\alpha)\prod_{n=1}^{N}p^{(c)}(\textbf{z}_{o,n}|\theta_o)p^{(c)}(w_{o,n}|\textbf{z}_{o,n},\phi)
\end {split}
},
\end{equation}

\noindent where $\alpha$ and $\beta$ are Dirichlet prior hyper-parameters that affect the sparsity of distributions. Although, $\alpha$ and $\beta$ could be vector-valued, we assume symmetric Dirichlet priors, with $\alpha$ and $\beta$ each having a single value. $K$ is the number of topics, $|c|$ is the number of known objects in the category $c$ and $N$ is the number of words in object $o$. Each $\theta_o$ represents an instance of category $c$ in topic-space as a histogram (i.e., \emph{topic layer}), $\textbf{w}$ represents an object as a vector of visual words, $\{ {w}_1, {w}_2, ..., {w}_N\}$, where each entry represents the index of one of the $V$ words of the dictionary (i.e., \emph{BoW layer}). Next, the object should be described as a histogram of topics, $\textbf{z} = \{ z_1, z_2, ..., z_K\}$. It should be noticed that there is a topic for each word. Therefore, the object is first described as a set of topics $\{ z_1, z_2, ..., z_N\}$, where each element of $\textbf{z}$ represents the index of one of the $K$ topics and then the object is represented as histogram of topics. $\phi$ is a $K \times V$ matrix, which represents word-probability matrix for each topic, where $V$ is the size of the dictionary and $\phi_{i,j}=p^{(c)}(w_i|z_j)$; 
thus, the posterior distribution of the latent variables given the observed data is computed as follows: 

\begin{equation}
\label {posterior}
		p^{(c)}(\textbf{z}, \theta,\phi | \textbf{w}, \alpha, \beta) = \frac{p^{(c)}(\textbf{w}, \textbf{z}, \theta,\phi | \alpha, \beta)}{p^{(c)}(\textbf{w}| \alpha, \beta)}
,\vspace{1mm}
\end{equation}

\noindent As \cite{blei2012probabilistic} showed, the denominator of the equation \ref{posterior} is \emph{intractable} and cannot be computed exactly due to a large number of possible instantiations of the topic structure. \cite{griffiths2004finding} proposed an analytical solution based on collapsed Gibbs sampling \citep{porteous2008fast} that can be used. Since $\theta$ and $\phi$ can be derived from $z_i$, they are integrated out from the sampling procedure. In this work, for each category, an incremental LDA model is created. Whenever a new training instance is presented, the collapsed Gibbs sampling is employed to update the parameters of the model. The collapsed Gibbs sampler is used to estimate the probability of topic $z_i$ being assigned to a word $w_i$, given all other topics assigned to all other words:

\begin{equation}
\label {gibbs_sampler}
\begin{split}
		p^{(c)}(z_i=k|\textbf{z}_{\neg i}, \textbf{w})\propto  
				p^{(c)}(z_i=k|\textbf{z}_{\neg i}) \times p^{(c)}(w_i|\textbf{z}_{\neg i},\textbf{w}_{\neg i})\\				
\propto \frac{ n_{o,k,\neg i}^{(c)} + \alpha}{\sum_{k=1}^{K} [n_{o,k}^{(c)}+ \alpha]-1 } \times \frac{n^{(c)}_{w, k,\neg i}+\beta}{\sum_{w=1}^{V} n^{(c)}_{w,k}+ \beta},\vspace{2mm}\quad\quad
\end{split}
\end{equation}

\noindent where $n^{(c)}_{(.)}$ shows the relative counters of category $c$; $\textbf{z}_{\neg i}$ means all hidden variables expect $z_i$ and $\textbf{z}=\{z_i, \textbf{z}_{\neg i}\}$ and $ n_{o,k}^{(c)}$ is the number of times topic $k$ from category $c$ is assigned to some visual word in object $o$ and $ n^{(c)}_{w,k} $ shows the number of times visual word $w$ is assigned to topic $k$. In addition, the denominator of the $p^{(c)}(z_i=k|\textbf{z}_{\neg i})$ is omitted because it does not depend on $z_i$. After the iterative sampling procedure, the multinomial parameter sets $\theta^{(c)}$ (i.e., object-topic matrix) and $\phi^{(c)}$ (i.e., topic-word matrix) can be estimated using the following equations: 
\begin{figure}[!b]
 \hspace{-8mm}
 \includegraphics[width=1.05\linewidth, trim= 4cm 2cm 4cm 1cm, clip=true]{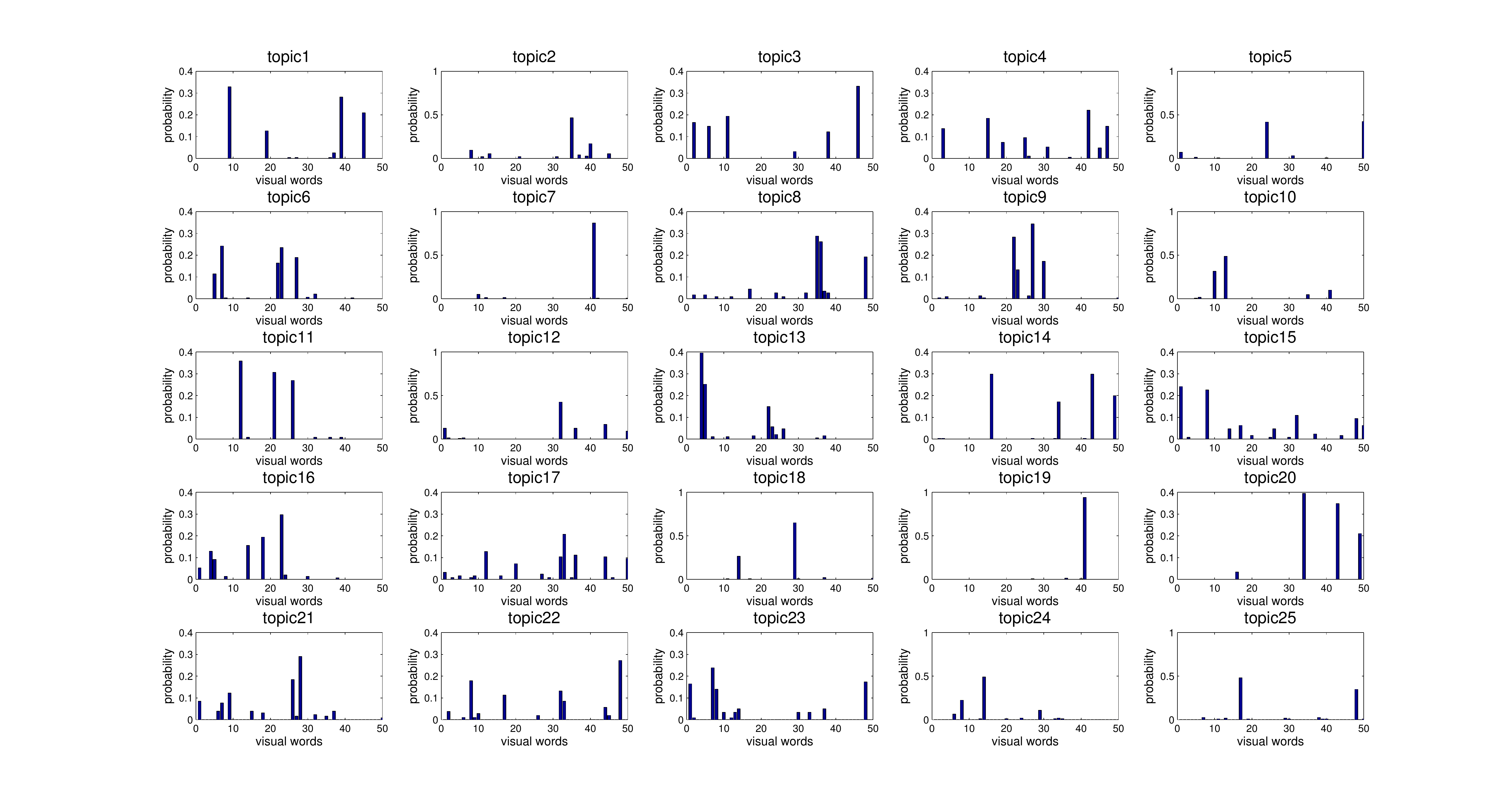}
% figure caption is below the figure
\caption{Visualization of 25 sample topics learned from the Washington RGB-D dataset\citep{Lai2011}.}
\label{fig:topic} 
%\end{wrapfigure}
\end{figure}

\begin{equation}
\label {phi_estimation_chapter4}
\begin {split}
	\theta^{(c)}_{k,o} = \frac{n^{(c)}_{o, k} + \alpha}{n_{o}+ K\alpha}, 	 \quad \\ \quad \phi^{(c)}_{w,k} = \frac{n^{(c)}_{w, k} + \beta}{n^{(c)}_{k}+ V\beta}.
\end {split}
\end{equation}
\noindent where $n^{(c)}_k$ is the number of times a word assigned to topic $k$ in category $c$ and $n_o$ is the number of words in the object $o$.

\begin{figure}[!b]
 \center
 \includegraphics[width=0.99\linewidth, trim = 0.5cm 0cm 0cm 0.5cm ,clip=true]{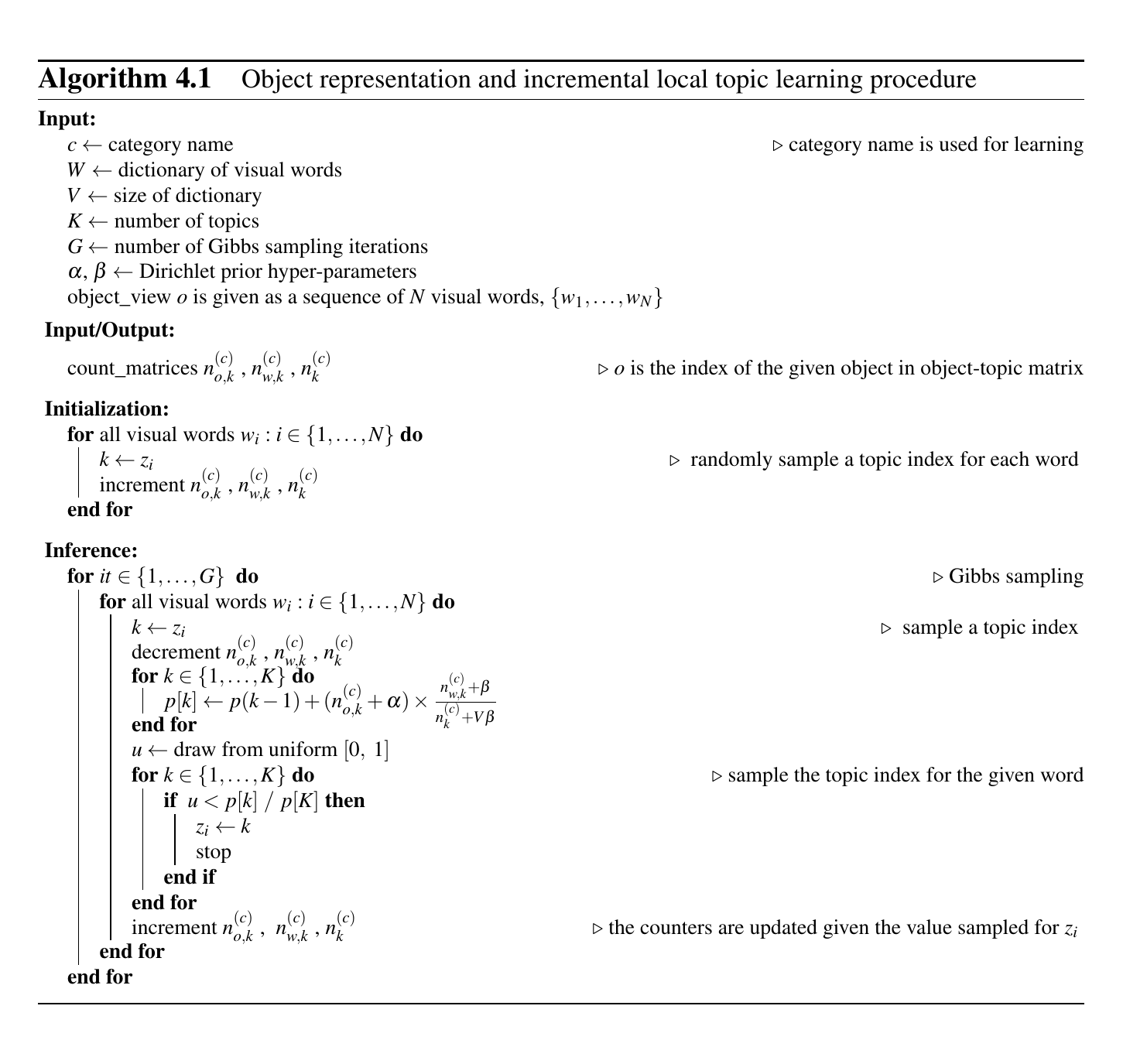}
\label{Gibbs_Sampling_LDA}
\vspace{0mm}
\end{figure}

\begin{figure}[!t]%\label{fig:spin_images}
 \includegraphics[width=0.9\textwidth, trim= 0cm 0cm 0cm 0cm, clip=true]{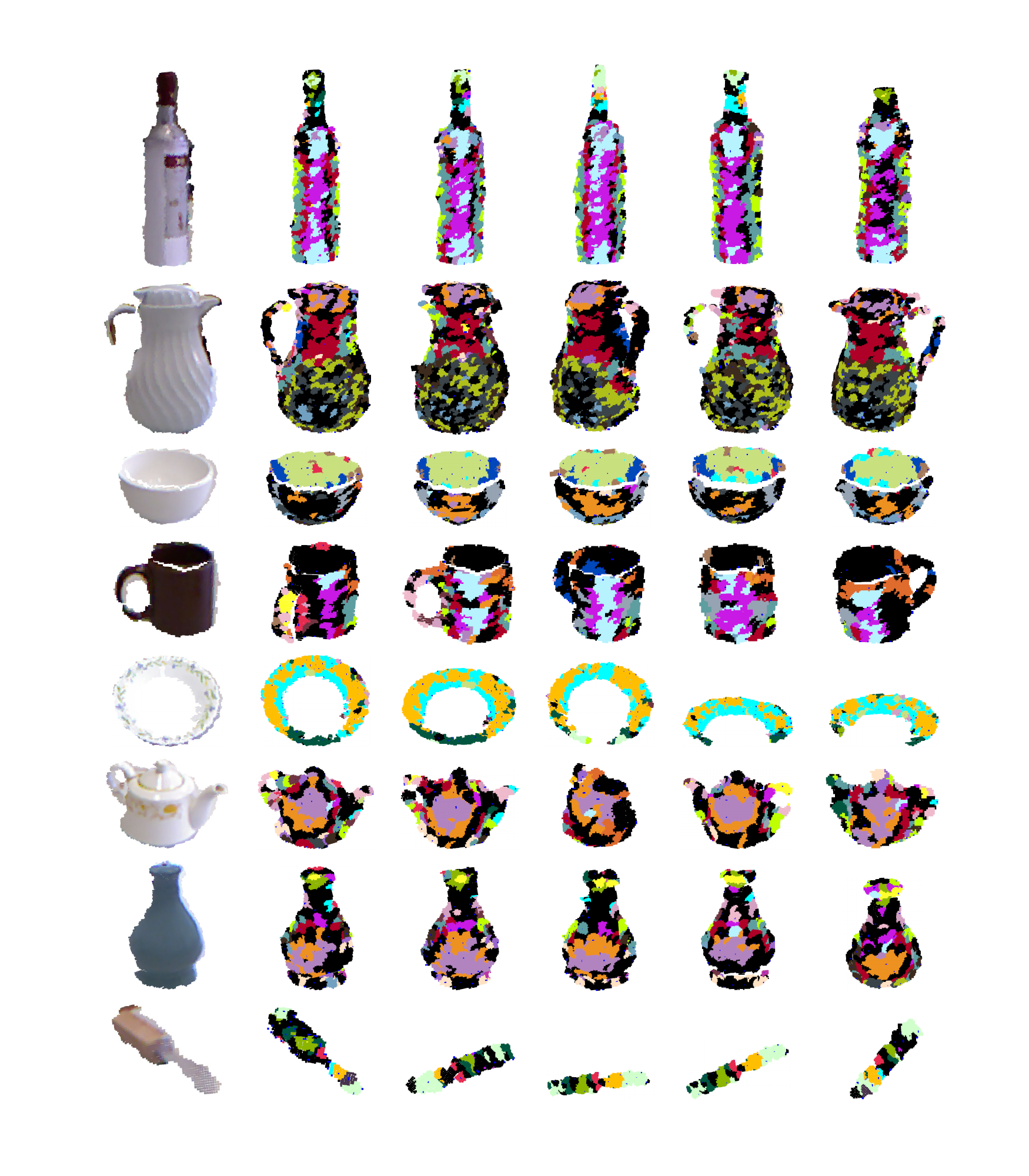} 

\caption{Examples of object-topic representation for eight categories of Restaurant Object Dataset \citep{KasaeiInteractive2015}: each object view is represented as a distribution over topics; each topic is shown with its topmost frequent visual words using a unique color. For instance, three high frequent topics in plate category are shown by yellow, cyan and dark green colors. Note, colors have the different meaning in each category.}
\label{fig:lda}       % Give a unique label
\end{figure}
What happens next depends only on the current state of the system and not on the sequence of previous states. Therefore, whenever a new training instance, $\theta^{(c)}_{o}$, is taught by an instructor (i.e., supervised learning), the collapsed Gibbs sampler is employed to first estimate a topic for each word of the given instance and then update the parameters of the model including $n^{(c)}_k$ and $n^{(c)}_{w, k}$ incrementally (i.e., unsupervised learning). 
Figure \ref{fig:topic} represents the 25 sample topics learned from the Washington RGB-D object dataset \citep{Lai2011}. $\theta$ is the object-topic matrix, where each row represents a training instance in topic-space as a histogram:

\begin{equation}
\label{object_topic_representation}
	\textbf{O}^{\textbf{t}} = [h_1,~ h_2,~ ...~, ~h_K]
	\vspace{1mm}
\end{equation}
\noindent Examples of object-topic representations for eight categories of the Restaurant Object Dataset \citep{KasaeiInteractive2015} are depicted in Fig.~\ref{fig:lda}. The proposed Local LDA procedure is summarized in Algorithm 4.1. 

To assess the dissimilarity between a target object view and stored instances of a certain category $c$, the target object should be described based on the learned topics of the category $\theta^{(c)}_{\textbf{T}}$ as a histogram (Eq.~\ref{phi_estimation_chapter4}). To this end, the target object and a temporal copy of $n^{(c)}_{w,k}$ and $n^{(c)}_k$ counters are first loaded into the Gibbs sampler. Then, a topic for each word of the target object is estimated. Finally, the target object is represented as a topic frequency histogram.

%%%%%%%%%%%%%%%%%%%%%%%%%%%%%%%%%%%%%%%%%%%%%%%%%%%%%%%%%%%%%%%%%%%%%%%%%%%%%%%%%%%
\subsubsection {LDA}
We also implement a modified version of the standard smoothed LDA approach \citep{blei2003latent}. The basic idea is to represent the objects as random mixtures over topics (i.e., latent variables shared among all categories), where each topic is characterized by a distribution over visual words (i.e., observed variables). Both random distributions are {defined} using multinomial distributions and their parameters are characterized by a Dirichlet distribution. As \cite{blei2012probabilistic} showed, the posterior distributions of the latent variables given the observed data cannot be computed exactly due to the large number of possible instantiations of the topic structure. \cite{griffiths2004finding} proposed an analytical solution based on collapsed Gibbs sampling \citep{porteous2008fast} that can be used to solve the inference problem. Similar to the Local LDA approach, an object is first represented as a set of visual words, $\{ w_1, w_2, ..., w_N\}$, where each entry represents one of the $V$ words of the dictionary. Next, the object should be described as a set of topics $\{ z_1, z_2, ..., z_N\}$, where each element of $\textbf{z}$ represents the index of one of the $K$ topics. Towards this goal, the probability of topic $z_j$ being assigned to a word $\textbf{w}$, given all other topics assigned to all other words, i.e., $P(z_{j}=k|\textbf{z}_{\neg j}, \textbf{w})$, is estimated by the collapsed Gibbs sampler \citep{porteous2008fast}. After the iterative {sampling} procedure, the multinomial parameter sets $\theta$ (i.e., object-topic matrix) and $\phi$ (i.e., topic-word matrix) can be estimated using the following equations: 

\begin{equation}
\label {phi_estimation}
	\begin {split}
		\theta_{k,d} = \frac{n_{d, k} + \alpha}{n_{d}+ K\alpha}, \\
	 \quad \phi_{w,k} = \frac{n_{w, k} + \beta}{n_{k}+ V\beta}.
	\end{split}
\end{equation}

\noindent 
where  $n_{d,k} $ is the number of times topic $k$ was assigned to some visual word in object $d$ and $n_d$ is the number of words in the object $d$; $K$ is the number of topics and $n_{w,k} $ shows the number of times visual word $w$ was assigned to topic $k$; $n_k$ is the number of times a word was assigned to topic $k$ and $\alpha$ and $\beta$ are Dirichlet prior hyper-parameters that affect the sparsity of distributions. 
$\phi$ is a $K \times V$ matrix, which represents word-probability matrix for each topic, where $V$ is the size of dictionary, $K$ is the number of topics and $\phi_{i,k}=p(w_i|z_k)$. 

In this approach, whenever a teacher adds a new training instance to a category, the collapsed Gibbs sampler is employed to first estimate a topic for each word of the given instance and then update the $n_k$ and $n_{w, k}$ parameters incrementally (i.e., unsupervised learning). Similar to the Local LDA, $\theta$ is the object-topic matrix, {where} each row represents a training instance in topic-space as a histogram. The obtain histogram can be used as a training instance in instance-based learning or can be used to create or update the probabilistic object category models based on Bayesian learning. To {be classified,} a target object view, $\textbf{T}$, should be first represented as a topic distribution $\textbf{O}^{\textbf{t}}$ (Eq.~\ref{object_topic_representation}). Towards this goal, the target object and a temporal copy of $n_{w, k}$ and $n_k$ are loaded into the Gibbs sampler and then a topic for each word of the target object is estimated. Next, the object is represented as a fixed-size histogram in the topic-space. Then, the dissimilarity between the target object and learned categories can be estimated.

\section{Summary}
\label{conclusion}

This chapter presented a set of object representation approaches for 3D object category learning and recognition. In particular, we presented a new global object descriptor named GOOD (i.e., Global Orthographic Object Descriptor) that provides a good trade-off between descriptiveness, computation time and memory usage, allowing concurrent object recognition and pose estimation. For an object view, GOOD provides a unique and repeatable local reference frame and three principal orthographic projections. GOOD is then calculated with the discretization of the three orthographic projections computed on that reference frame and concatenates them to form a single description for the given object. We release the source code of the GOOD descriptor, to the benefit of the research community, in Point Cloud Library\footnote{http://pointclouds.org/}(PCL version 1.9) and our Github repository\footnote{https://github.com/SeyedHamidreza/GOOD\_descriptor}.

Furthermore, we presented a set of object representation approaches based on local features to enhance a 3D object category learning and recognition. In this part, we mainly focused on representing objects based on local shape features, BoW and topic-based representations to construct powerful object descriptions. In particular, we propose an extension of Latent Dirichlet Allocation to learn structural semantic features (i.e., topics) from low-level feature co-occurrences for each category independently. In this approach, for optimizing the recognition process and memory usage, each object view was described as a random mixture over a set of latent topics, and each topic was defined as a discrete distribution over visual words. Topics in each category are discovered in an unsupervised fashion and are updated incrementally using new object views. In the next chapter, we will propose and discuss a set of open-ended object category learning and recognition approaches.

\cleardoublepage
\chapter{Open-Ended Object Category Learning and Recognition}
\label{chapter_5}

Nowadays robots utilize 3D computer vision algorithms to perform complex tasks such as object recognition and manipulation. 
Although, many problems have already been understood and solved successfully, many issues still remain. Open-ended object recognition is one of these issues waiting for many improvements. 

In particular, most robots lack the ability to learn new objects from past experiences. To migrate a robot to a new environment one must often completely re-generate the knowledge base that it is running with. Since in open-ended domains the set of categories to be learned is not predefined, it is not feasible to assume that one can pre-program all necessary object categories for assistive robots. Instead, robots should learn autonomously from novel experiences, supported in the feedback from human teachers. This way, it is expected that the competence of the robot increases over time. 

Several state-of-the-art assistive robots use traditional object category learning and recognition approaches \citep  {leroux2013armen,beetz2011robotic,vahrenkamp2010integrated}. These classical approaches are often designed for static environments in which it is viable to separate the training (off-line) and testing (on-line) phases. In these cases, the world model is static, in the sense that the representation of the known categories does not change after the training stage. Therefore, these robots are unable to adapt to dynamic environments \citep{Jeong2012}. This leads to several shortcomings such as the inability to detect/recognize previously unseen categories.  To cope with these issues, several cognitive robotics groups have started to explore how robots could learn incrementally from their own experiences as well as from interaction with humans \citep{Smith2005,chauhan2011,he2008imorl}. In order to incrementally adapt to new environments, an autonomous assistive robot must have the ability to process visual information and conduct learning and recognition tasks in a concurrent and interleaved fashion.

We approach object category learning and recognition from a long-term perspective and with emphasis on open-endedness, i.e., not assuming a pre-defined set of categories. Learning methods used in most of the classical object recognition systems are not designed for open-ended domains, since those methods do not support an incremental update of the robot's knowledge based on new experiences. In contrast, open-ended learning approaches can, not only incrementally update the acquired knowledge (category models), but also extend the set of categories over time, which is suitable for real-world scenarios. For example, if the robot does not know how a \emph{`Mug'} looks like, it may ask the user to show one. Such situation provides an opportunity to collect training instances from actual experiences of the robot and the system can incrementally update its knowledge rather than retraining from scratch when a new instance is added or a new category is introduced.

In this chapter, the subject of online 3D object category learning and recognition in open-ended robotic domains is investigated. The characteristics of open-ended learning and
the possible role of human-robot interaction in that context are analyzed. Two main open-ended learning approaches are investigated, namely instance-based learning and model-based
learning. Instance-based learning is explored with both variable size representations (sets of local features) and fixed size representations (GOOD, histograms of local features and histograms of LDA topics). The notion of intra-category distance is proposed as a reference to decide if an object belongs to one of the known categories or if it belongs to an unknown category. For model-based learning, the Bayesian learning approach is explored with fixed size representations

%In particular, several 3D object category learning and recognition approaches are proposed. The contributions presented here are the following: \emph{(i)} defining a new distance function for estimating dissimilarity between two sets of local shape features (variable size representations) that can be used in instance-based learning approaches; \emph{(ii)} proposing a learning approach to incrementally learn probabilistic models of object categories to achieve adaptability. The second contribution uses a Naive Bayes learning method to compute category models from the observed views of instances of the categories.

Parts of the work presented herein have been published in various international workshops \citep{kasaeiRSSinstance,kasaeiobjectNIPS}, conferences \citep{Kasaei2014,Oliveira2014,GiHyunLim,dubba2014grounding,kasaei2015adaptive,kasaei2016concurrent} as well as journal articles \citep{kasaei2015interactive,oliveira20153d,Hertzberg2014projrep,kasaei2017Neurocomputing}. The remainder of this chapter is organized as follows. In section~\ref{sec:related_work}, we discuss related works. Afterwards, the characteristics of open-ended learning are presented in section~\ref{sec:characteristics_of_open-ended_learning}. The methodology of instance-based learning approach is described in section~\ref{sec:instance_based_learning}. Model-based object category learning and recognition are then explained in section~\ref{sec:model_based_learning}. Finally, conclusions are presented and future research is discussed.

%^^^^^^^^^^^^^^^^^^^^^^^^^^^^^^^^^^^^^^^^^^^^^^^^^^^^^^^^^^^^^^^^
%^^^^^^^^^^^^^^^^^^^^^^^^^^^^^^^^^^^^^^^^^^^^^^^^^^^^^^^^^^^^^^^^
\section {Related Work}
\label{sec:related_work}

Object category learning and recognition has become an active subject in several research communities such as computer vision and robotics because of its potential applications. Over the past five decades, different object category learning and recognition approaches have been proposed \citep{andreopoulos201350}. 

\cite{Collet} described an object recognition and pose estimation system based on one-step learning. In this case, the system is decomposed into an off-line training stage and an on-line recognition stage. In the training stage, for every object, a set of images are captured from different viewpoints. Then, SIFT features are extracted for each image and stored in a database. During the recognition stage, SIFT features are computed for the current view and matched against the training models. The authors use a Best Bin First algorithm \citep{Beis} for matching. \cite{Kootstra} proposed an active perception system for recognizing objects that are placed in cluttered and uncontrolled environments. They used a mobile robot that explores the objects by circling around them and capturing data. They also used SIFT descriptors for learning and recognition tasks. \cite{Liu} developed a system based on the bag-of-words technique to optimize memory usage and the recognition process. The authors investigated the problem of efficient partial 3D shape retrieval. First, a Monte-Carlo method to select interest points is proposed, and then, the spin-image descriptors are used to encode the geometry around the interest points. In the recognition stage, they proposed to use a dissimilarity measure based on the asymmetric Kullback-Leibler divergence.

Willow Garage started a project named Object Recognition Kitchen (ORK)\footnote{http://wg\textunderscore perception.github.io/object\textunderscore recognition\textunderscore core/}, a 3D object recognition system built on top of the Ecto framework\footnote{http://plasmodic.github.io/ecto/}. ORK was designed to run simultaneously several traditional object recognition techniques, so that these can be combined for example using a voting scheme. Ecto is a C++ / Python computation graph framework, which can organize the computation modules in a directed acyclic graph. In ORK, the training and recognition are not simultaneous. 

In all the systems described above, training and testing are separate processes, i.e., they do not occur simultaneously. However, in open-ended applications, data is continuously available and the target object categories are not known in advance. In these cases, traditional object recognition approaches are not well suited, because those systems are limited to using off-line data for training and are therefore unable to adapt to new objects and new environments.

There are some approaches which support incremental learning of object categories. In these approaches, the set of classes is predefined and the models of known object categories are enhanced (e.g., augmented, improved) over time. \cite{he2008imorl} proposed an incremental multiple-object recognition and localization (IMORL) framework using a multilayer perceptron (MLP) structure as the base learning model. The authors claimed that the proposed framework can incrementally learn from experiences and use such knowledge for object recognition. \cite{Yeh2008} developed novel methods for efficient incremental learning of SVM-based visual category classifiers, and showed that, using their framework, it is possible to adapt the classifiers incrementally.  \cite {Martinez2010} described a fast and scalable perception system for object recognition and pose estimation. The authors employed the RANSAC and Levenberg Marquardt algorithms to segment objects and represented them based on SIFT descriptors. 

In the last decade, various research groups have made substantial progress towards the development of learning approaches which support life-long object category learning \citep{Oliveira2016614,kasaeiNips2016,kasaei2015adaptive}. These methods assume that the set of categories to be learned is not known in advance.
For instance, zero-shot, low-shot, and open-ended learning approaches have recently received significant attention from the machine learning and computer vision communities \citep{akata2014good,hariharan2017low,kasaeiNips2016}. {Zero-shot learning} approaches aim to recognize instances of object categories that have not been included during the training phase. This type of learning approaches, in addition to visual features, use textual descriptions of object categories to train classifiers. In other words, they first project object categories into a semantic space learned using large text corpora, and then predict an initial model for new categories by image synthesis methods \citep{akata2014good}. Although, this type of approaches could be used to reduce the number of images required to train a model by incorporating textual information, they are out of the scope of this thesis.

{Low-shot learning} usually consists of two phases including \emph{representation learning} and \emph{few-shot learning for classification purposes}. These approaches mainly assume a large set of training data is available in advance for the representation learning phase and focus on transfer learning and classification tasks. The representation learning phase is commonly based on training a Deep Convolutional Networks (ConvNet). These approaches train a linear classifier in the low-shot learning phase \citep{hariharan2017low}. The learned representation is usually frozen and not changed during the test phase. Such approaches require many training instances, and their learning rate is slow and typically involves long training times. From an applied perspective, collecting massive amounts of labeled data is costly in many domains (e.g. robotics). 

In contrast, {open-ended learning} allows for concurrent or interleaved learning and recognition. In open-ended learning, the learning agent extracts training instances from its on-line experiences. Thus, training instances become gradually available over time, rather than being partially or completely available at the beginning of the learning process \citep{kasaeiNips2016}. Interactive open-ended object category learning and recognition is a key capability in assistive and service robots. This means that a robot should be capable of continuously learning new objects in order to perform different tasks. This type of object category learning and recognition approaches is extremely useful in assistive and service robots since the end user expects that the competence of the robot increases over time. 

\cite{Steels2002} use the notion of ``language game'' to develop a social learning framework through which an AIBO robot can learn its first words. A teacher points to objects and provides their names. The robot uses color histograms and an instance-based learning method to learn word meanings. The teacher can also ask questions and provide feedback on the robot’s answers. The authors show, with concrete robotic experiments, that unsupervised category formation may produce categories that are completely unrelated to the categories that are needed for grounding the words of the used language. They therefore conclude that social interaction must be used to help the learner focus on what needs to be learned in the context of communication.

\cite{Lopes2008} developed a category learning architecture that included a metacognitive processing component. Multiple object representations and multiple classifiers combinations were used. Classifier combinations are based on majority voting and the Dempster-Shafer evidence theory. All learning computations are carried out during the normal execution of the agent, which allows continuous monitoring of the performance of the different classifiers. The measured classification successes of the individual classifiers support an attentional selection mechanism, through which classifier combinations are dynamically reconfigured and a specific classifier is chosen to predict the category of a new object. In another work, the same authors approached the problem of object experience gathering and category learning with a focus on open-ended learning and human-robot interaction \citep{chauhan2011}. Moreover, they considered forgetting rules to optimize memory usage. In this approach, the user can provide the names of objects through pointing and verbal teaching actions. The user can also ask questions about the categories of objects under shared attention and, if appropriate, provide corrective feedback. The authors assume that a long-term category learning process in an artificial agent will eventually reach a breakpoint, i.e., an internal state of the agent in which new categories can no longer be discriminated. They showed a system that starts with an empty vocabulary and can incrementally acquire object categories through the interaction with a human user. In both works, they used RGB data whereas we used depth data. Moreover, their object detection, learning and recognition approaches are completely different from our approach. 

\cite{Iwahashi2010} approach their concept learning problem in an incremental and online manner. Their robotic platform consists of a robotic-arm with multi-fingered gripper (with tactile sensors), a stereo-vision camera and an infra-red sensor. They have developed an online learning approach (LCore) based on a Bayesian framework that allows incremental category learning. The experimental setup involves a human-user describing objects and/or actions in an environment visually shared with the robotic agent. Robot’s participation can be passive (e.g. human user moves the object and describes the action/object) or active (e.g. human user asks the robot to move an object). During active participation, in case the robot makes a mistake, the user verbally corrects the robot. This interaction is carried out until the robot performs the correct action. In a passive interaction or a correct active interaction, the robot stores object features (colour, shape, size and tactile information), object movement configurations and phoneme strings (of each word) identified in a spoken utterance. LCore, using new and previously stored knowledge, calculates joint probabilities over co-occurring speech, visual (shape, size and color descriptors) and tactile information. This process leads to grounding names of concrete categories (e.g. object names) and names of other perceptual characteristics (e.g. words referring to color, shape or size). Additionally, their agent is also able to learn motion concepts (modeled by HMMs) and is able to ground some action words (e.g. move-over, place-on).

\cite{Kirstein} proposed an approach for interactive learning of multiple categories based on vector quantization and a graphical user interface. The instructor could give the names of objects using the graphical user interface. Using only 2D images, they showed the system could successfully learn 5 color categories and 10 shape categories observed in 56 objects. \cite{collet2014herbdisc} proposed a graph-based approach for lifelong robotic object discovery. Similar to our approach, they used a set of constraints to explore the environment and to detect object candidates from raw RGB-D data streams. In contrast, their system does not interactively acquire more data to learn and recognize the object. \cite{skovcaj2016integrated} describe a system with similar goals. \cite{faulhammer2017autonomous} presente a system which allows a mobile robot to autonomously detect, model and recognize objects in everyday environments.

Currently, a popular approach in object recognition is deep learning. It is now clear that if an application has a pre-defined fixed set of object categories and
thousands of examples per category, an effective way to build an object recognition system is to train a deep Convolutional Neural Network (CNN). However, there are several limitations to use CNN in open-ended domains. CNNs are incremental by nature but not open-ended, since the inclusion of novel categories enforces a restructuring in the topology of the network. Moreover, CNN usually need a lot of training data and cannot converge to a proper solution using a small set of training data.

%^^^^^^^^^^^^^^^^^^^^^^^^^^^^^^^^^^^^^^^^^^^^^^^^^^^^^^^^^^^^^^^^
%^^^^^^^^^^^^^^^^^^^^^^^^^^^^^^^^^^^^^^^^^^^^^^^^^^^^^^^^^^^^^^^^ 

\section{Characteristics of Open-Ended Learning}
\label{sec:characteristics_of_open-ended_learning}

One of the primary challenges in service robotics is to allow robots to adapt to open-ended dynamic environments where they need to interact with non-expert users. For robots acting in these domains, it is not viable to hand-code all possible behaviours and to anticipate all possible exceptions. Instead, robots must be supported by life-long learning and adaptation capabilities. Towards this goal, as proposed by \cite{lopes2002towards}, the learning system of the robots should have five basic characteristics:

\begin{itemize}

\item \textbf{Supervised} - to include the human instructor in the learning process. This is an effective way for a robot to obtain knowledge from a human user/teacher.

\item \textbf{On-line} - meaning that the learning procedure takes place while the robot is running.

\item \textbf{Opportunistic} - apart from learning from a batch of labelled training data at predefined times or according to a predefined training schedule, the robot must be prepared to accept a new example when it is observed or becomes available. 

\item \textbf{Incremental} - it is able to adjust the learned model of a certain category when a new instance is taught.

\item \textbf{Concurrent} - it is able to handle multiple learning problems at the same time (e.g., learn object category model as well as learning/improving underlying object representation such as topics).

\end{itemize}
\noindent
As surveyed in the previous section, learning approaches used in classical object category recognition approaches, do not satisfy some of these requirements. For instance, most of these approaches support incremental learning, but are not designed for scaling up to larger sets of categories and do not support open-ended learning. It is worth mentioning that the target set of categories is predefined in standard incremental learning, while in open-ended learning the target set of categories is not known in advance and is continuously growing. In an open-ended scenario, an instructor may teach the robot a set of new instances from previously seen as well as new categories at any arbitrary time. Given the open-ended nature of category learning, different learning methods, presented in this thesis, are designed to support open-ended learning of an arbitrary set of categories. In such approaches, the introduction of new categories can significantly interfere with the existing category descriptions. To cope with this issue, memory management mechanisms, including
salience and forgetting, can be considered. Moreover, as more categories are learned, the classification accuracy first decreases (performance degradation phase), and then starts going up again as more categories are introduced (recovery phase). This is expected since the number of categories known by the system makes the classification task more difficult. However, as the number of learned categories increases, also the number of instances per category increases, which augments the category models and therefore improves performance of the system. Eventually the learning agent reaches to a breakpoint where s/he is no longer able to learn more categories~\citep{chauhan2014grounding}.
\cite{Seabra2007} states these features of an open-ended learning process:

\begin{itemize}
\item \textbf{Evolution:} The classification performance should be improved as the number of examples per category increases while no new categories are introduced. 
Moreover, the prediction accuracy should not fluctuate at every incremental learning step. 

\item \textbf{Recovery:} With the increase in the number of categories, it is expected that the prediction accuracy decreases. The time spent in system evolution until correcting and adjusting all current categories defines recovery. In other words, if the performance drops at a certain learning step, the agent should be able to recover to the previous best performance. Recovery is based on classification errors and corresponding corrections as new instances become available. 

\item \textbf{Breakpoint:} The agent is no longer able to recover and learn more categories.  

\end{itemize}

%However, few researchers have proposed learning approaches supporting open-ended category learning and recognition. 

%^^^^^^^^^^^^^^^^^^^^^^^^^^^^^^^^^^^^^^^^^^^^^^^^^^^^^^^^^^^^^^^^
%^^^^^^^^^^^^^^^^^^^^^^^^^^^^^^^^^^^^^^^^^^^^^^^^^^^^^^^^^^^^^^^^ 

\section {Human-Robot Interaction }
\label{HRI}
In the present work, a human user teaches the robot the names of the objects present in their visually shared environment using the developed interface for Human-Robot Interaction (HRI) (i.e., described in section \ref{sec:supervisedExperienceGathering}). Category names are then grounded by the robot in sensor-based object representations, leading to a vocabulary shared with its teacher. 
%In such scenarios, the introduction of new categories can significantly interfere with the existing category models. As a consequence, by increasing the number of learned categories, the learning performance will evolve with phases of performance degradation followed by recovery, but eventually reach a point where the robot is no longer able to learn any more categories. 
The HRI interface provides a set of actions that the instructor can use for interacting with the robot. For object category learning and recognition purposes, the primary action is the \emph{teach} action which allow a user to provide the category labels of objects, present in a shared scene, to the robot. Another important action is the \emph{correct} action. In particular, whenever the agent could not classify an object correctly, the user has the facility to give correction. Therefore, at the most basic level of interaction, the interface allows the user to perform four main actions:

\begin{itemize}
\item \textbf{Select:} point to the desired object using either the object's \emph{TrackID} or pointing to the object with the arm.

\item \textbf{Teach:} provide the category label of the selected object.

\item \textbf{Ask:} inquire the category label of the selected object, which the agent will predict based on previously learned knowledge.

\item \textbf{Correct:} if the agent could not recognize a given object correctly, the user can teach the correct category.
\end{itemize}

The robot responds to the teacher actions by either running the relevant learning functionalities (i.e., in the cases of teach and correct actions) or performing classification task (i.e., in response to an ask action). In particular, the robot's possible responses include:

\begin{itemize}
\item \textbf{Learning}: \emph{teach} and \emph{correct} actions trigger perceptual learning to create new or modify existing category models, and create association between the category label and its corresponding category description.

\item \textbf{Recognition}: in response to the \emph{ask} action, the results of object recognition are visually available to the user (e.g. user can see which object was recognized correctly) and divergence from expected response leads the user to provide corrective feedback.

\end{itemize}

In the following subsections two object category learning and recognition techniques are presented to support open-ended learning of an arbitrary set of categories. 

%^^^^^^^^^^^^^^^^^^^^^^^^^^^^^^^^^^^^^^^^^^^^^^^^^^^^^^^^^^^^^^^^
%^^^^^^^^^^^^^^^^^^^^^^^^^^^^^^^^^^^^^^^^^^^^^^^^^^^^^^^^^^^^^^^^

\section {Instance-Based Learning}
\label{sec:instance_based_learning} 

As discussed in the previous chapter, either a layered data processing system based on local features or a global object descriptor can be used for object representation purposes. The layer-based data processing system builds an increasingly complex object representation for the given object view. It consists of five layers, including \emph{feature layer}, \emph{BoW layer}, \emph{topic layer}, \emph{object view layer} and \emph{category layer} (see Section~\ref{sec:local_features_new}). In the \emph{feature layer} each object view, $\mathbf{O}$, is described by a set of spin-images. It is worth mentioning that the number of keypoints depends on the size of the object and as a consequence, we have a variable size representation for each object (see Fig.~\ref{fig:keypoint}). In the \emph{BoW layer}, the given object view is represented as a histogram of occurrences of visual words. In order to build more compact and complex object representation in the \emph{topic layer}, the given object is represented as a histogram of topics. 
Object category learning can occur at all layers. 

As proposed in Section~\ref{GOOD_shape_description}, an object can also be represented based on global descriptors. Similar to the BoW and LDA-based approaches, global object descriptors produce a fixed size representation for a given object. 

\begin{figure}[!b]
%\center
\hspace{-5mm}
\begin{tabular}[width=1\textwidth]{cc}
 \includegraphics[width=0.48\linewidth, trim=0.0cm 4cm 0cm 0cm,clip=true]{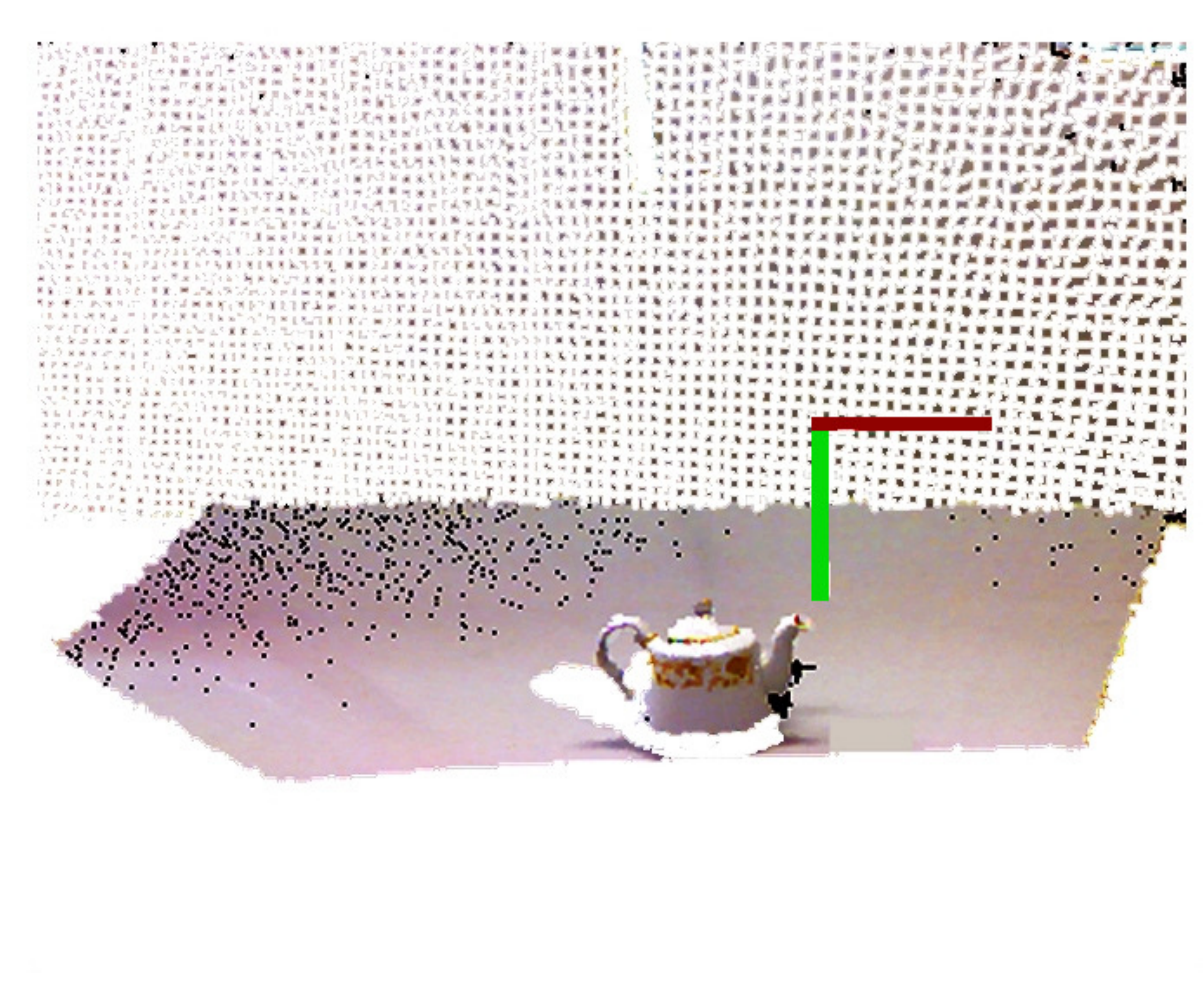} &
 \includegraphics[width=0.53\linewidth, trim= 0cm -0.2cm 0cm 0cm,clip=true]{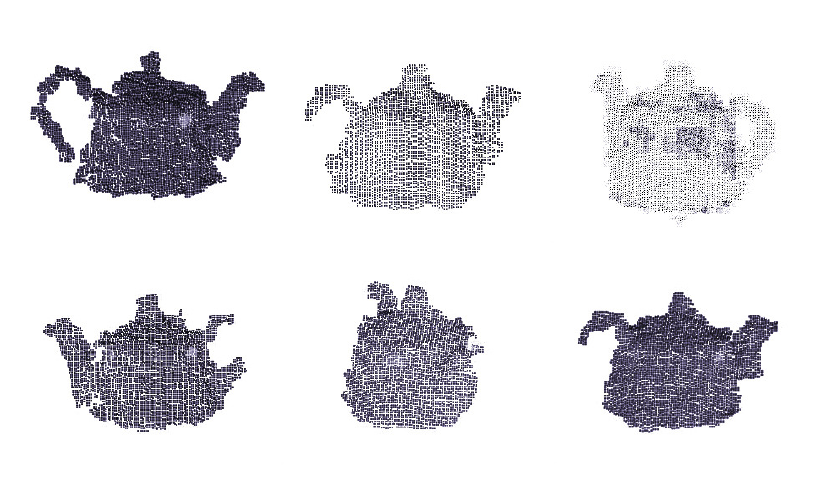}
\end{tabular}
% figure caption is below the figure
\caption{ Object learning: (\emph{left}) example point cloud of a scene containing a teapot; (\emph{right}) six views of the segmented teapot from different perspectives.}
\label{fig:IBL}       % Give a unique label
\end{figure}

%%%%%%%%%%%%%%%%%%%%%%%%%%%%%%%%%%%%%%%%%%%%%%%%%%%%%
\subsection {General Aspects}
\label{general_aspects}
The instance-based approach considers category learning as a process of learning about the instances of the category (see Fig.~\ref{fig:IBL}). In other words, a category is represented by a set of views of instances of the category \citep{daelemans2005memory}:
\begin {equation}
\textbf{C} = \{ \textbf{O}_1,~\dots,~\textbf{O}_n \}
\end  {equation}

\noindent where $\textbf{O}_i$ are the constituent views. As discussed in Section~\ref{HRI}, the user identifies the need for teaching a new category or for providing corrective feedbacks when s/he notices classification errors in robot's responses. As described in chapter~\ref{chapter_3}, the object perception system stores key views of the tracked objects. Therefore, when the user teaches or provides corrective feedback about the category of a target object, the learning module will store one or more views of that object. In an instance-based learning framework, the stored object views are the instances. In particular, new instances are stored in the \emph{Perceptual Memory} in the following situations:

\begin{itemize}
	  \item  When the teacher for the first time teaches a certain category, through a \emph{Teach} or a \emph{Correct} action, an instance-based representation of this new category is created and initialized with the set of key views of the target object collected since object tracking started:
	  
	 % If the user explicitly \emph{teaches} a new category in situations where the robot encounters with a new object, a new object category is formed by storing an initial set of object views:
	   
\begin {equation}
\label{init_cat}
\textbf{C}_1 \leftarrow \{ \textbf{O}_1, \dots, \textbf{O}_1{k_1}\}, 
\end  {equation}
	  	 
  	\noindent where $k_1$ is the number of stored key object views for the first teaching action.
	  	 
\item  In later teaching actions, the target object views are added to the instance-based representation of the category:

\begin {equation}
\label{update_cat}
\textbf{C}_n  \leftarrow \textbf{C}_{n-1} \cup \{ \textbf{O}_{nk}, \dots, \textbf{O}_{nk_n}\}.
\end  {equation}

\noindent where $k_n$ is the number of stored key object views for the $n$-th teaching action.
\end{itemize}

Whenever a new object view is added to a category, the \emph{Object Conceptualizer} module retrieves the current model of the category as well as the representation of the new object views, and creates a new, or updates the existing category. To assess the dissimilarity between a target object and stored instances, a recognition mechanism in the form of a nearest neighbour classifier is usually used. Therefore, each instance-based object learning and recognition approach can be seen as a combination of a particular \emph{object representation}, \emph{similarity measure} and \emph{classification rule}. One advantage that instance-based learning has over other methods of machine learning is its ability to adapt its model to previously unseen data. The disadvantage of this approach is that the computational complexity can grow with the number of training instances. The computational complexity of classifying a single new instance is $O(n)$, where $n$ is number of instances stored in memory. Salience and forgetting mechanisms can be used to bound the memory usage \citep{Lopes2008}. These mechanisms are also useful for reducing the risk of overfitting to noise in the training set. Another advantage of the instance-based approach is that it facilitates incremental learning in an open-ended fashion. 

In the following sub-sections, we present and discuss instance-based object category learning and recognition approaches for both variable size object representations (i.e., set of spin-images) and fixed size object representations such as BoW, LDA and GOOD representations.

%%%%%%%%%%%%%%%%%%%%%%%%%%%%%%%%%%%%%%%%%%%%%%%%%%%%%%%%%%%%%%%%%%%%
\subsection {The ``Unknown Category'' Problem}
\label{sec:unknown_category_problem}

Most of the work in this thesis is focused on maximizing scalability, i.e., the capacity to scale to larger and larger sets of categories while minimizing the losses in other performance metrics. Another problem that is typical of open-ended learning in real-world scenarios is what we call here the ``unknown category'' problem.

In classical learning and classification approaches, there is a set of target categories to be learned and the target objects to be recognized are assumed to belong to one of the target categories. In contrast, in open-ended scenarios, the learning agent must be prepared to handle situations where the target object does not belong to any of the known categories.

To realize that the target object belongs to an unknown category, the agent must reason in terms of distances to the categories taking also into account the size or spread of the
categories. High intra-class variation means that objects from the same category can have very different shapes. For instance, the Cup category contains many types of cups with different shapes (see Fig.\ref{fig:ICD_new}). Even the views of the same object can vary significantly depending on the viewpoint. Therefore, when judging if a given object belongs to a given category, it is not enough to compute the nearest-neighbor distance or the centroid distance. Such distances are only meaningful when compared to the spread of the category. In this work, to address this problem, a measure of category spread called the \emph{Intra-Category Distance} ($\operatorname{ICD}$), is introduced:

\begin{figure}[!t]
\center
 \includegraphics[width=0.65\linewidth, trim=0.0cm 2cm 0cm 0cm,clip=true]{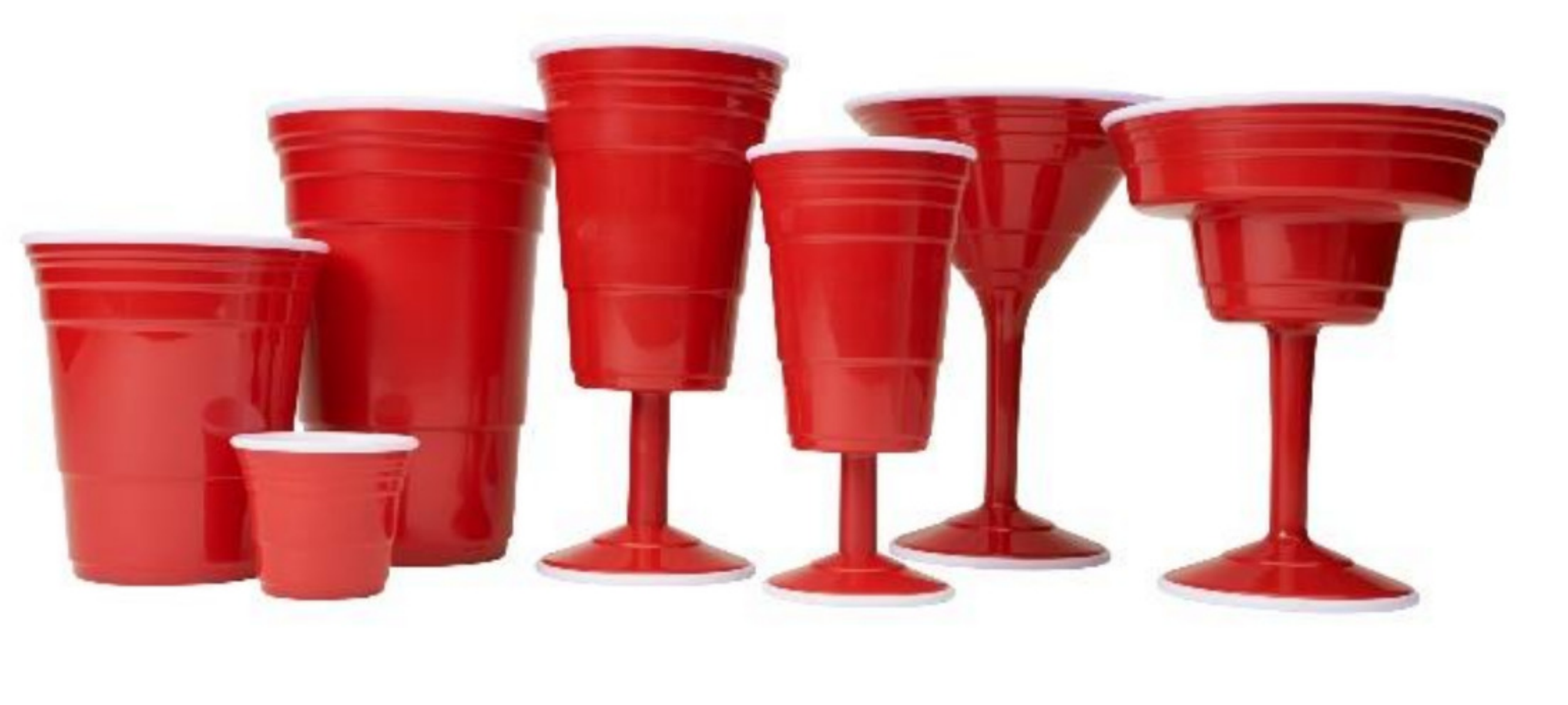}
% figure caption is below the figure
\caption{ An example of intra-class variation among cup instances.}
\label{fig:ICD_new}       % Give a unique label
\end{figure}

\begin{equation}
\operatorname{ICD}(\mathbf{C})=\frac{\displaystyle\sum\limits_{\mathbf{\mathbf{\mathbf{U}}}\in \mathbf{C}}{\,\,\,\displaystyle\sum\limits_{\mathbf{\mathbf{\mathbf{V}}}\in \textbf{C}, \mathbf{\mathbf{U}} \ne \mathbf{\mathbf{V}}}{\ \operatorname{D}\left(\mathbf{\mathbf{U}},\mathbf{\mathbf{V}}\right)}}}{|\textbf{C}| \cdot \left(|\textbf{C}|-1\right)},
\end{equation}

\noindent
where $|\textbf{C}|$ is the total number of instances (i.e., object views)  in the category, $\textbf{U}$ and $\textbf{V}$ are two different known instances of $\textbf{C}$ and $\operatorname{D}(\textbf{U},\textbf{V})$ is a measure of distance between two object views. This way, we represent an object category by (\emph{i}) a set of known instances and (\emph{ii}) the corresponding Intra-Category Distance. Whenever new instances are added to a category, the object conceptualizer retrieves the instances of the object category from memory and the set of features describing the new instances, and then re-computes ICD and stores the new ICD value.

The ``unknown category'' problem is relatively easy to handle when compared to the scalability problem mentioned earlier. Therefore, most of the learning approaches were formulated having in mind scenarios in which, for simplicity, the “unknown category” problem is assumed not to exist. The “unknown category” problem was addressed (based on ICD) only in instance-based learning with a variable size representation (sets of spin images, in the next subsection).

%%%%%%%%%%%%%%%%%%%%%%%%%%%%%%%%%%%%%%%%%%%%%%%%%%%%%%%%%%%%%%%%%%%%
\subsection {Variable Size Representations: Sets of Local Features}
\label{sec:variable_size_representations_sets_of_local_features}

This section presents the instance-based learning approach developed by us for the object perception and perceptual learning system of the RACE project \citep{kasaei2015interactive,oliveira20153d}. It is based on representing objects as sets of spin images and uses ICD to normalize distances and to decide when an object belongs to an unknown category.

\subsubsection {Basic Distance Metric}

Since in this approach the size of an object representation depends on the size of the object (number of points, number of keypoints), we cannot use common distance metrics such as
Euclidean distance. Therefore, we propose the following function to estimate the distance or dissimilarity between two object views (i.e., two sets of spin-images):

%^^^^^^^^^^^^^^^^^^^^^^^^^^^^^^^^^^^^^^^^^^^^^^^^^^^^^^^^^^^^^^^^

\begin{equation}\label{eq:dist}
\operatorname{D}\left(\mathbf{U},\mathbf{V}\right)=\frac{\displaystyle\sum\limits_{i}{\
	\underset{j}{\min}\ \operatorname{d}(\mathbf{u}_i,\mathbf{v}_j)}}{|\mathbf{U}|},
\end{equation}

\noindent where $\mathbf{u}_i$, and $\mathbf{v}_j$, are the spin-images of \textbf{U} and \textbf{V} respectively, $|\mathbf{U}|$ is the number of spin-images in \textbf{U} and $\operatorname{d}(\cdot)$ is the Euclidean distance. It should be noted that D(·) is not symmetric (i.e., $\operatorname{D}\left(\mathbf{U},\mathbf{V}\right)~\ne~\operatorname{D}\left(\mathbf{V},\mathbf{U}\right)$).

\subsection {Category Initialization and Update}

As discussed above, the \emph{teach} and \emph{correct} actions from the user trigger the object conceptualizer module to create a new category or modify an existing category. If the integrated system is used, the instance-based is updated as described in section~\ref{general_aspects}. ICD is initialized when at least three views are available. For experiment based on datasets, the following schema is used. Whenever a new object category is taught, three consecutive object views are stored to initialize the category and to compute the relevant ICD:

\begin {equation}
\textbf{C}_3 \leftarrow \{ \textbf{O}_1,~\textbf{O}_2,~\textbf{O}_3\}, 
\end  {equation}

\begin {equation}
\operatorname{ICD}_3 \leftarrow \operatorname{ICD}(\textbf{C}_3), 
\end  {equation}

If the user provides corrective feedback, or teaches a new instance of a known category, the agent retrieves the current object category from memory, as well as a set
of features describing the labeled object, and updates the category by adding the new instance to the set of instances of the category and re-computing the ICD:

\begin {equation}
\textbf{C}_n  \leftarrow \textbf{C}_{n-1} \cup \{ \textbf{O}_n\}.
\end  {equation}

\begin {equation}
\operatorname{ICD}_n \leftarrow \operatorname{ICD}(\textbf{C}_n). 
\end  {equation}

Such interactive teaching mechanism provides an opportunity to collect a training instance (an experience) for learning in an open-ended manner. In this approach, the set of object categories to be learned is not known in advance. The training instances are extracted from actual experiences of a robot, and thus become gradually available, rather than being available at the beginning of the learning process. Finally, it updates the category model by storing the ICD and representation of the new instance. For the object recognition purpose, we consider two approaches. In the following subsections, the details of each approach are presented.

%^^^^^^^^^^^^^^^^^^^^^^^^^^^^^^^^^^^^^^^^^^^^^^^^^^^^^^^^^^^^^^^^

\subsubsection {Normalized Object-Category Distance: Approach I}	

In this approach, the proposed unsymmetrical distance function (Eq.~\ref{eq:dist}) is used to estimate the dissimilarity between a target object view and a stored instance of a category in memory. In order to define Object-Category Distance, $\operatorname{OCD}(.)$, the minimum distance between the target object, and all the instances of a certain category, $\textbf{C}$, is considered as the $\operatorname{OCD}(.)$:
\begin{equation}
		\operatorname{OCD}(\textbf{T},\textbf{C}) = \underset{\emph{\textbf{\textbf{O}}}\in \textbf{C}}{min} \,\,\operatorname{D}(\textbf{T},\textbf{O}),  
\end{equation}

In object recognition, since some categories are more spread than others, normalizing $\operatorname{OCD}(.)$ by $\operatorname{ICD}(.)$ will help to prevent misclassification. Therefore, the Normalized Object-Category Distance of the target object view $\textbf{T}$ to the category $\textbf{C}$, $\operatorname{NOCD}(\textbf{T},\textbf{C})$, is computed as follows:
\begin{equation}
\operatorname{NOCD}\left(\mathbf{\textbf{T}},C\right)=\frac{{\operatorname{OCD}}(\mathbf{\textbf{T}},\textbf{C})}{\operatorname{ICD}(\textbf{C})}, 
\end{equation}

%^^^^^^^^^^^^^^^^^^^^^^^^^^^^^^^^^^^^^^^^^^^^^^^^^^^^^^^^^^^^^^^^

\subsubsection{Normalized Object-Category Distance: Approach II} 

In this approach, we consider computing the $\operatorname{OCD}(.)$ based on the average distance between the target object, and all the instances of a certain category,~$\textbf{C}$:

\begin{equation}\label{eq:eq2}
\operatorname{OCD}\left(\mathbf{T},\textbf{C}\right)=\frac{\displaystyle\sum\limits_{\mathbf{U}\in \textbf{C}}
{\ ~\operatorname{D}\left(\mathbf{T},\mathbf{U}\right)}}{|\textbf{C}|},
\end{equation} 

In this approach, the Normalized Object-Category Distance, $\operatorname{NOCD}(.)$, between the target object and stored instances of category $\textbf{C}$ is computed by:

\begin{equation}\label{eq:neweq4}
\operatorname{NOCD}\left(\mathbf{T},\textbf{C}\right) =\frac{2  \times \operatorname{OCD}(\mathbf{T},\textbf{C})}{\operatorname{ICD}(\textbf{C})+ \overline{\operatorname{ICD}}},
\end{equation}

\noindent where $\overline{\operatorname{ICD}}$ is the average of the intra-category 
distances of all categories:
\begin{equation}
 \overline{\operatorname{ICD}}= \frac{\sum_{i=1}^{m}\operatorname{ICD}(\textbf{C}_i)}{m}
\end{equation}
\noindent
where $m$ is the number of categories. %The target object is finally classified based on the minimum normalized distance to the known categories (see Eq.~\ref{eq:muli}).  

\subsubsection{Classification rule}

Finally, the target object is classified based on the minimum normalized distance to the known categories:

\begin{equation}\label{eq:muli}
\operatorname{Category}(\mathbf{T}) = \underset{C}{\operatorname{min}} \operatorname{NOCD}\left(\mathbf{\textbf{T}},\textbf{C}\right)
\end{equation}

\begin{figure}[!t]
\center
\begin{tabular}[width=1\textwidth]{ccc}
 \includegraphics[width=0.45\linewidth, trim=0.0cm 0cm 0cm 0cm,clip=true]{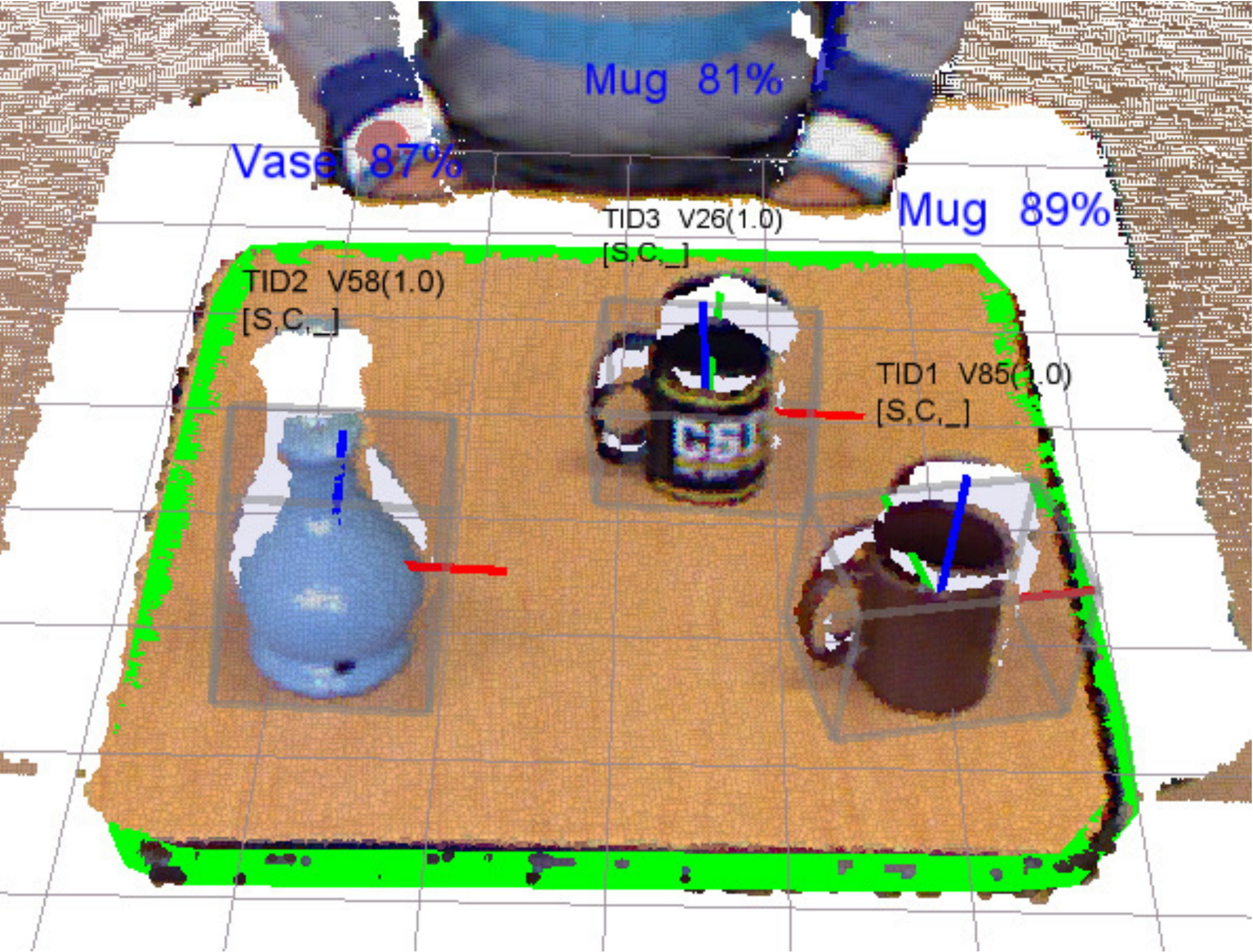} &
 \includegraphics[width=0.48\linewidth, trim= 0cm 0cm 0cm 0cm,clip=true]{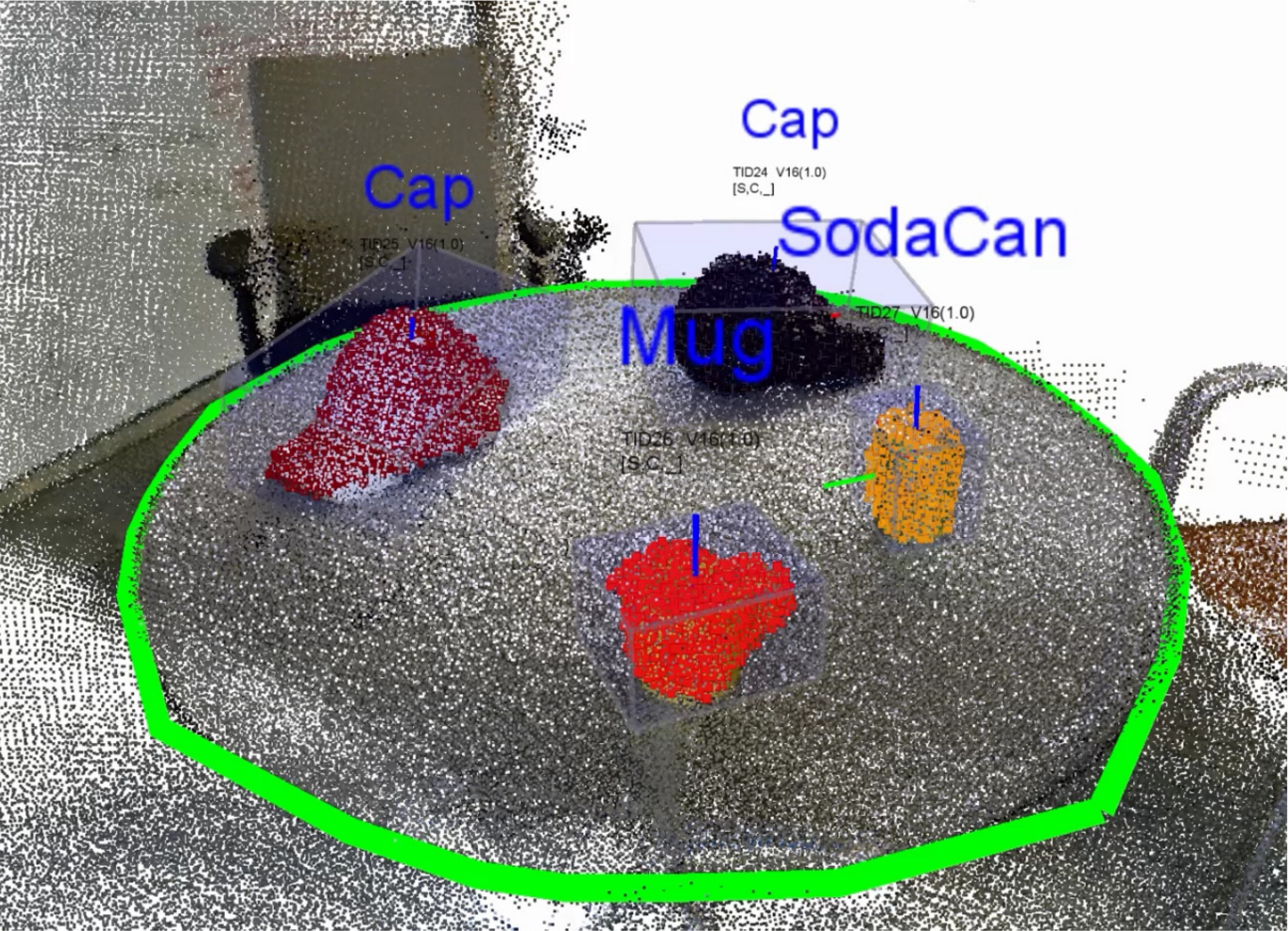}
\end{tabular}
% figure caption is below the figure
\caption{ Two examples of the instance-based 3D object recognition based on: (\emph{left}) approach I and (\emph{right}) approach II.}
\vspace{-3mm}
\label{fig:IBR}       % Give a unique label
\end{figure}

This classification rule can be easily modified to predict an ``\emph{unknown}'' category when the target object is too far from all known categories. This can be done by considering:

\begin{equation}
\operatorname{Category}(\mathbf{T}) =
\begin{cases}
\emph{Unknown}, \, \quad\,\quad\quad\quad\,\,\, ~~~\operatorname{if}\quad\underset{\textbf{C}}{\operatorname{min}} \operatorname{NOCD}\left(\mathbf{\textbf{T}},\textbf{C}\right) > \operatorname{CT}\\
\underset{\textbf{C}}{\operatorname{argmin}} ~\operatorname{NOCD}\left(\mathbf{\textbf{T}},\textbf{C}\right), \,\, ~~~~~~\emph{otherwise}
\end{cases},
\end{equation}
\noindent where the Classification Threshold (CT) is used to decide when the target object is too far from the known categories. Two examples of the instance-based 3D object recognition are shown in Fig.~\ref{fig:IBR}

%%%%%%%%%%%%%%%%%%%%%%%%%%%%%%%%%%%%%%%%%%%%%%%%%%%%%%%%%%%%%%%%%%%%%%%%%%%%%%
%%%%%%%%%%%%%%%%%%%%%%%%%%%%%%%%%%%%%%%%%%%%%%%%%%%%%%%%%%%%%%%%%%%%%%%%%%%%%%
\vspace{3mm}
\subsection {Fixed Size Representations: GOOD, BoW, LDA}
\label{fixed_size_representations}
Fixed size representations (i.e., histograms) are frequently used in various object classification tasks to represent rich information in local/global regions of objects. Global object descriptors such as GOOD and machine-learning techniques based on BoW and topic modelling produce a fixed size representation for a given object. Such representations are \emph{usually} compact, rotation-invariant, translation-invariant and normalized. These properties make these representation methods suitable for performing object recognition task.

In this approach, a purely memory-based learning approach is adopted, in which a category is represented by a set of representations of instances of the category. Similar to the instance-based approaches, teaching and correction by the user lead the agent to add a new instance to the taught category. In this approach a category is initialized using one instance (see Eq.~\ref{init_cat}) and whenever the agent receives a corrective feedback, it adds the object to the set of instances of the category (see Eq.~\ref{update_cat}).

To classify a previously unseen instance, we compute the similarity of the target instance with all the previously seen instances, which have been stored in memory. Finally, the target object is classified based on the minimum distance to the known instances. In this approach, the choice of similarity metric has a great impact on the recognition performance. 
The similarity between two object views can be computed by different standard distance functions. We refer the reader to a comprehensive survey on distance/similarity measures provided by \cite{cha2007comprehensive}. After performing several cross-validation experiments with different representations, we conclude two type of distance functions are suitable to estimate the similarity between two instances. One function properly works with GOOD and BoW representations, and the other one shows good results with LDA-based representations. Both functions are in the form of a bin-to-bin distance function which compares corresponding bins in two object representation vectors.

Mathematically, let P and Q $\in{\rm I\!R^N}$ be two vectors with the same size.
One of the common forms of bin-to-bin distance function is the Minkowski-Form, also known as $L_p$ norm:

\begin{equation}
L_p\operatorname{(P,Q)}= (\sum_{i=1}^{n}|\operatorname{P}_i - \operatorname{Q}_i|^p)^{\frac{1}{p}}
\end{equation}

\noindent where $n$ is the size of representation and  $|.|$ returns the absolute value. The most widely used distances from this family are $L_1$ and $L_2$ forms. The $L_1$ is also known as the Manhattan distance and the $L_2$ distance is also known as the Euclidean distance. Throughout this thesis, we use the $L_2$ norm (i.e., Euclidean distance) with the BoW and GOOD representations. 

Since LDA-based approaches represent a given object in the form of probability distribution, we use a specific type of Kullback–Leibler (KL) divergence to estimate the similarity of two instances. The KL divergence is defined as:
\begin{equation}
\operatorname{KL(P,Q)}= \sum_{i=1}^{n}\operatorname{P}_i \log \frac{\operatorname{P}_i}{\operatorname{Q}_i}
\end{equation}

This form of KL is non-symmetric (i.e., $\operatorname{KL(P,Q)} \ne \operatorname{KL(Q,P)}$).
One of the limitations of KL is that when $\operatorname{P}_i = 0$, it equals infinity and when $\operatorname{Q}_i = 0$, it is undefined. To overcome KL limitations, the Jensen-Shannon (JS) divergence was suggested \citep{pele2009fast}: 

\begin{equation}
\operatorname{JS(P,Q)}= \operatorname{KL(P,M)} + \operatorname{KL(Q,M)}
\end{equation}

\noindent where $\operatorname{M} = \frac{1}{2}(\operatorname{P} + \operatorname{Q})$. One of the specific distance functions that can be derived from  $\operatorname{JS(P,Q)}$ using Taylor extension, is the chi-squared distance \citep{yang2015chi,pele2011distance}. The function is known in the computer vision community as  $\chi^{2}$ distance: 

\begin{equation}
\chi^{2}\operatorname{(P,Q)} = \frac{1}{2}\sum_{i=1}^{n} \frac{(\operatorname{P}_i-\operatorname{Q}_i)^2}{(\operatorname{P}_i+\operatorname{Q}_i)}
\end{equation}

\cite{pele2008linear} and \cite{pele2011distance} showed that the practical results of $\chi^{2}$ and JS are almost identical. Since the computation of $\chi^{2}$ is more efficient than the JS, we use $\chi^{2}$ function to estimate the similarity of two LDA representations.

%^^^^^^^^^^^^^^^^^^^^^^^^^^^^^^^^^^^^^^^^^^^^^^^^^^^^^^^^^^^^^^^^
%^^^^^^^^^^^^^^^^^^^^^^^^^^^^^^^^^^^^^^^^^^^^^^^^^^^^^^^^^^^^^^^^
\section {Model-Based Learning }
\label{sec:model_based_learning}

The model-based learning approaches are often contrasted with instance-based approaches. The instance-based learning considers category learning as a process of learning about the instances of the category while the model-based learning is a process of learning a parametric/non-parametric model from a set of instances of a category. In other words, each category is represented by a single model. The instance-based approach proposes that a new object is compared to all previously seen instances, while the model-based approach proposes that a target object is compared to the model of categories. Therefore, in the case of recognition response, the model-based approach is faster than the instance-based approach. In contrast, instance-based learning approach can recognize objects using small number of experiences, while model-based approaches need more experiences to achieve a good classification results. Therefore, training is very fast in instance based approach while they require more time in recognition phase. Another disadvantage of instance-based learning approach is that they need a large amount of memory to store the instances. 

In this section, we propose an open-ended 3D object category learning approach, which considers category learning as a process of computing a probabilistic model for each object category using the Naive Bayes approach. Similar to the instance-based approach, teaching and correction by the user lead the agent to create a new category or to update an existing category.  In the classification phase, the likelihood between a target object and all object category models is first estimated. Then, the target object is classified based on the maximum likelihood. There are two reasons why Bayesian learning is useful for open-ended learning. One of them is the computational efficiency of the Naive Bayes approach. In fact, this model can be easily updated when new information is available, rather than retrained from scratch. Second, memory usage in instance-based open-ended systems is continuously growing since these systems are constantly storing new object views (instances). Therefore, these systems must resort to experience management methodologies to discard some instances and thus prevent the accumulation of a too large set of experiences~\citep{wilson2000reduction,Lopes2008}. In Bayesian learning, new experiences are used to update category models and then the experiences are forgotten immediately. The category model encodes the information collected so far. Therefore, this approach consumes a much smaller amount of memory when compared to any instance-based approach. This learning approach can be used with all the proposed fixed size object representations, i.e., BoW, LDA and GOOD. In the following sub-section, we present how to acquire and refine object category models based on a naive Bayes learning approach.

%^^^^^^^^^^^^^^^^^^^^^^^^^^^^^^^^^^^^^^^^^^^^^^^^^^^^^^^^^^^^^^^^
\subsection {Category Representation}

Let us assume an object is represented by a n-dimensional vector $\textbf{x} = [x_1,\dots, x_n]$. 
In this approach, an object category, $\textbf{C}_k$, is represented as a tuple: 

\begin{equation}
\textbf{C}_k = \langle~N_k,~\textbf{a}_k,~\operatorname{P}(\textbf{C}_k),~[\operatorname{P}(x_1|\textbf{C}_k),\dots, \operatorname{P}(x_n|\textbf{C}_k)]~\rangle,
\end{equation}

\noindent where $N_k$ is the number of seen instances in category $k$, $\textbf{a}_k$ is a vector of bin accumulators for category $k$. In particular, $a_{ki}$ is the accumulation of $i^{th}$ bin of all instances of category $C_k$ and the length of $\textbf{a}$ is equal to the length of $\textbf{x}$. $\operatorname{P}(\textbf{C}_k)$ is the prior probability of category {$\textbf{C}_k$} and each element of $[\operatorname{P}(x_1|\textbf{C}_k),\dots, \operatorname{P}(x_n|\textbf{C}_k)]$ shows the probability of word/topic occurring in class \emph{$\textbf{C}_k$} or in the case of GOOD descriptor, the $\operatorname{P}(x_i|\textbf{C}_k)$ shows the probability of a point falling into the bin $i$. 
In this work, we consider the probability of each bin independently, regardless of any possible correlations with the other bins (Naive Bayes approach). The $\operatorname{P}(\textbf{C}_k)\operatorname{P}(\textbf{x}|\textbf{C}_k)$ is equivalent to the joint probability model:
\begin{equation}
\operatorname{P}(\textbf{C}_k| {x}_1, \dots, {x}_n) \propto \operatorname{P}(\textbf{C}_k) ~ \operatorname{P}({x}_1, \dots, {x}_n \vert \textbf{C}_k),
\end{equation}

\noindent The joint model can be rewritten using conditional independence assumptions:

\begin{equation}
\nonumber
	\begin {split}
			\operatorname{P}(\textbf{C}_k \vert {x}_1, \dots, {x}_n) 			\propto \operatorname{P}(\textbf{C}_k)~ \operatorname{P}({x}_1 \vert \textbf{C}_k) ~\operatorname{P}({x}_2\vert \textbf{C}_k)\cdots ~
\operatorname{P}({x}_n\vert \textbf{C}_k) \\
	\end {split}
\end{equation}
\begin{equation}
\label{eq2}
        \propto \operatorname{P}(\textbf{C}_k)~ \prod_{i=1}^{n} \operatorname{P}({x}_i \vert \textbf{C}_k),         
\end{equation}

\noindent
where $n$ is the size of the object representation and $P({x}_i |\textbf{C}_k )$ is the
probability of the bin ${x}_i$ in an object of category $k$. 

%%%%%%%%%%%%%%%%%%%%%%%%%%%%%%%%%%%%%%%%%%%%%%%%%%%%%%%%%%%%%%%%%%%%%%%
\subsection {Category Initialization and Update}

Similar to the instance-based approach, the \emph{teach} and \emph{correct} actions by the user lead the robot to create a new category or to modify an existing category. In particular, whenever the user explicitly \emph{teaches} a new category,  the category is initialized using the key views of the target object. For simplicity, the process is formalized below assuming that each teaching action provides a single object view. It would be straightforward to extend this formalization to a certain number of key views of the target object~\footnote{In the simulated teacher experiments, the teach action always sends three views.}.

The new instance, represented as a histogram $\textbf{x}^{\prime} = [x^{\prime}_1,\dots, x^{\prime}_n]$, is added to the taught category $C_k$.
Category initialization involves updating the total number of instances of all known categories, $N$, and initializing category specific parameters, namely the number of instances of the category, $N_k$, and the bin accumulators, $a_{ki}$:

\begin{equation}
\nonumber
		N \leftarrow N+1,
\end{equation}
\begin{equation}
		N_k \leftarrow 1,
\end{equation}
\begin{equation}
\nonumber
		a_{ki} \leftarrow x^{\prime}_{i}, \quad for\quad i=1,~\dots,~n,
\end{equation}

If the user provides corrective feedback for a known category, $C_k$, the category model is updated using that particular instance:

%the agent updates the category using that particular instance. The accumulators $N_k$ and $\textbf{a}_k$ are used to update the category model incrementally.  Specificity, whenever a new instance, which is represented as a histogram $\textbf{x}^{\prime} = [x^{\prime}_1,\dots, x^{\prime}_n]$, is added to the category $k$, the following update is carried out:

\begin{equation}
\nonumber
		N \leftarrow N+1,
\end{equation}
\begin{equation}
		N_k \leftarrow N_k+1,
\end{equation}
\begin{equation}
\nonumber
a_{ki}  \leftarrow  a_{ki} + x^{\prime}_{i},
\end{equation}

Upon each teaching action, the probabilities are updated, namely the the prior probabilities of all existing categories:

\begin{equation}
\operatorname{P}(\textbf{C}_k) =\frac{N_k}{N}, \quad k \in \{1, \dots m\}
\end{equation}
\noindent
where $m$ is the number of known categories up to now and the probability of each bin, ${x}_i$, in the category $k$, $\operatorname{P}({x}_i|\textbf{C}_k)$ is updated as follows:

\begin{equation}
\label{update_p}
\operatorname{P}({x}_i|\textbf{C}_k) =\frac{a_{ki}~+1}{\sum\limits_{j=1}^{n}(a_{kj}+1)},
\end{equation}

\noindent 
Note that the probabilities are estimated with Laplace smoothing, by adding one to every counter, in order to prevent $\operatorname{P}({x}_i|\textbf{C}_k) = 0$.

%%%%%%%%%%%%%%%%%%%%%%%%%%%%%%%%%%%%%%%%%%%%%%%%%%%%%%%%
\subsection {Classification Rule}
 To classify a given object \textbf{O}, which is represented as a n-dimensional histogram, $\textbf{y} = [y_1\dots y_n]$, the posterior probability for each object category is approximated using the Bayes theorem as:
	 
\begin{equation}
\label{posterior_probability}
	\begin {split}
			\operatorname{P}(\textbf{C}_k|\textbf{O}) = \operatorname{P}(\textbf{C}_k|\textbf{y}) = \frac{\operatorname{P}(\textbf{y}|\textbf{C}_k)\operatorname{P}(\textbf{C}_k)}{\operatorname{P}(\textbf{y})}
			\approx \operatorname{P}(\textbf{y}|\textbf{C}_k)\operatorname{P}(\textbf{C}_k),
	\end {split}
\end{equation}

\noindent 
Because the denominator does not depend on $\textbf{C}_k$, equation~\ref {posterior_probability} is re-expressed based on equation~\ref{eq2} and multinomial distribution assumption:

\begin{equation}
			\operatorname{P}(\textbf{y}|\textbf{C}_k)\operatorname{P}(\textbf{C}_k) \approx \operatorname{P}(\textbf{C}_k) \prod_{i=1}^{n} \operatorname{P}({x}_i |\textbf{C}_k)^{y_{i}},
\end{equation}

\noindent
In addition, to avoid underflow problems, the logarithm of the likelihood is computed: 
 \begin{equation}
	\begin {split}
			 \log \operatorname{P}(\textbf{C}_k)	+ \sum_{i=1}^{n} {y_{i}} ~ \log  \operatorname{P}({x}_i |\textbf{C}_k),
	\end {split}
\end{equation}

\begin{figure}[!t]
\center
\begin{tabular}[width=1\textwidth]{ccc}
 \includegraphics[width=0.45\linewidth, trim= 4cm 4cm 4cm 5cm,clip=true]{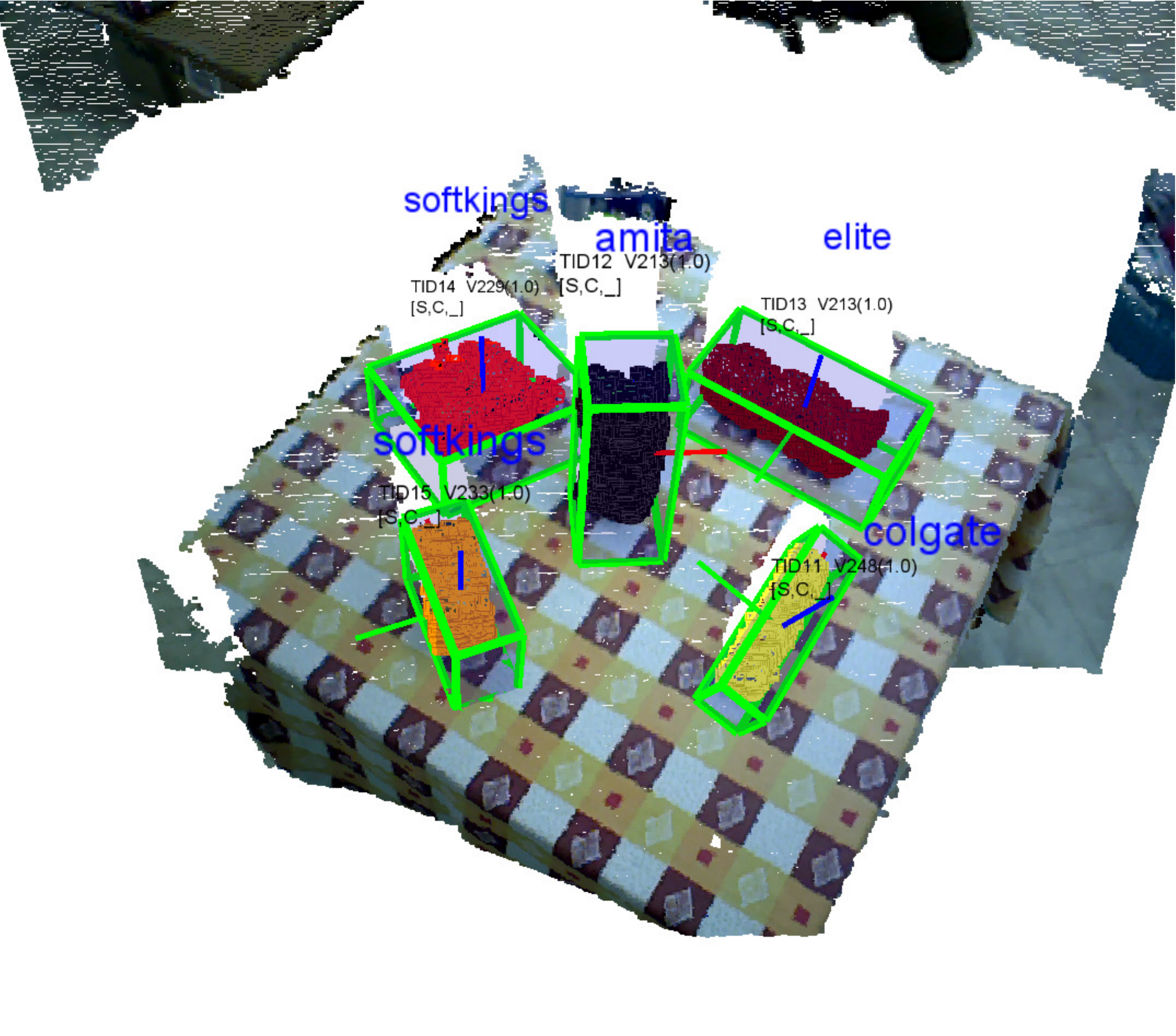} &
 \includegraphics[width=0.48\linewidth, trim=0.0cm 0cm 0cm 0cm,clip=true]{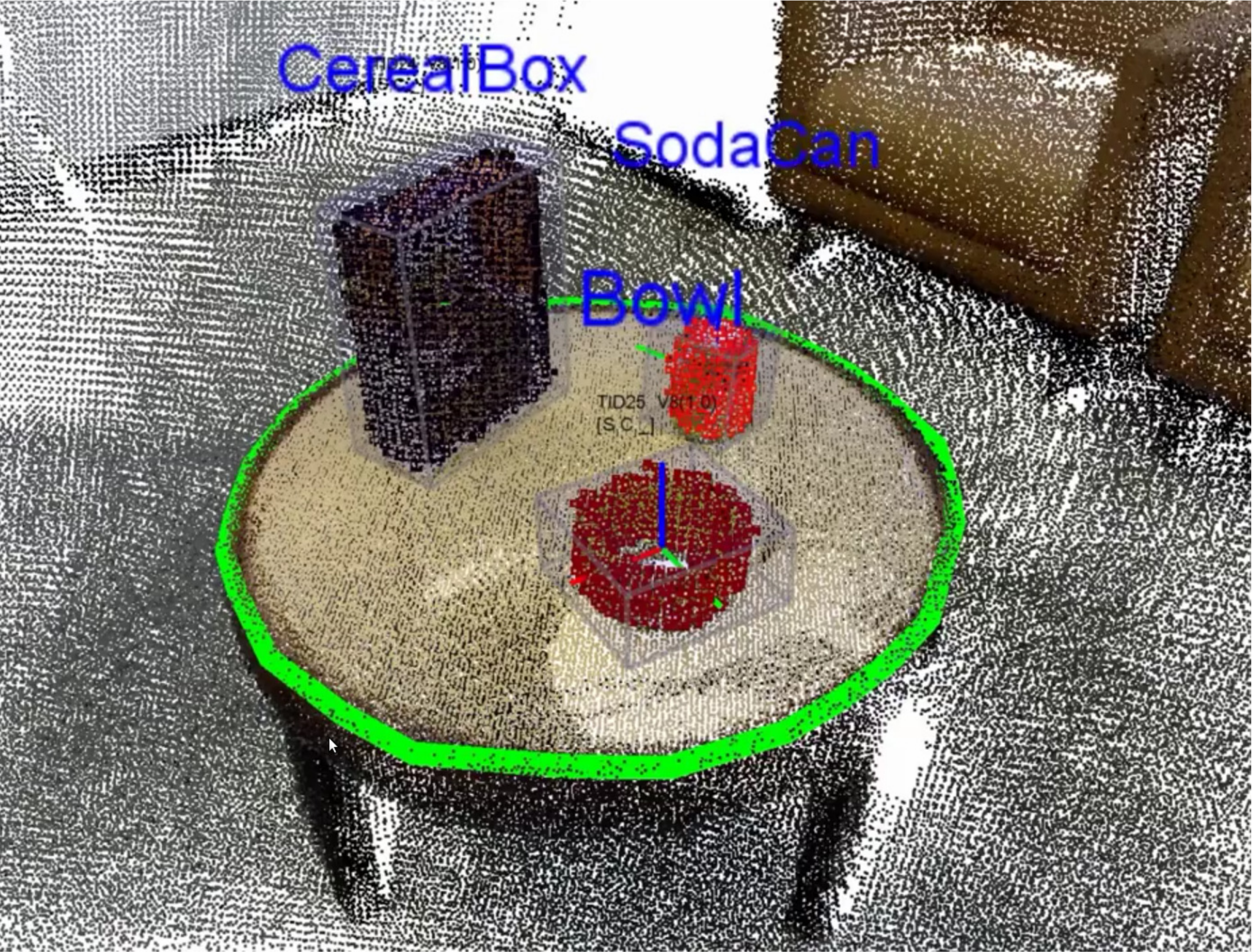}
\end{tabular}
% figure caption is below the figure
\caption{Two examples of object recognition based on the proposed model-based approach: (\emph{left}) with BoW representation; (\emph{right}) with Local LDA representation.}
\label{fig:MBR}       % Give a unique label
\end{figure}

\noindent
The category of the target object $\textbf{O}$ is the one with
highest likelihood:
\begin{equation}\label{eq:muli_proj_5}
\operatorname{Category}(\mathbf{O}) = \underset{{c}_k\in~\textbf{c}}{\operatorname{argmax}}~
\operatorname{P}(\textbf{C}_k|\textbf{O}).
\end{equation}

%^^^^^^^^^^^^^^^^^^^^^^^^^^^^^^^^^^^^^^^^^^^^^^^^^^^^^^^^^^^^^^^^
%^^^^^^^^^^^^^^^^^^^^^^^^^^^^^^^^^^^^^^^^^^^^^^^^^^^^^^^^^^^^^^^^
\section { Summary}
\label{sec:summary}

In this chapter, we first discussed the characteristics of open-ended learning approaches. Afterwards, instance-based learning and model-based learning approaches for grounding 3D object categories were presented in details. In this work, ``\emph{open-ended}'' implies that the set of object categories to be learned is not known in advance. The training instances are extracted from on-line experiences of a robot, and thus become gradually available over time, rather than being completely available at the beginning of the learning process. 

In particular, we presented and discussed a set of instance-based learning approaches for both variable size object representations (i.e., a set of spin-images) and fixed size object representations (i.e., BoW, LDA and GOOD). In these approaches, learning a category involves storing its representative object representations and the classifiers are derived from a combination of a specific instance representation, similarity measure and a classification rule.  In the case of model-based object category learning, a Naive Bayes learning approach was used. In this approach, each object was first represented as a fixed-size histogram (i.e., BoW, LDA and GOOD). Then, a probabilistic object category model was computed for each category. For recognition, a classification rule was used to assign a category label to the given object. 

Each approach has a set of parameters that must be tuned to provide an appropriate balance between recognition performance, memory usage and computation time. We address this point in the next chapter by evaluating the performance of each approach with different configurations using k-fold cross-validation method. This type of evaluation provides a straightforward base for comparing different approaches among themselves and possibly with other approaches described in the literature.

%^^^^^^^^^^^^^^^^^^^^^^^^^^^^^^^^^^^^^^^^^^^^^^^^^^^^^^^^^^^^^^^^
%^^^^^^^^^^^^^^^^^^^^^^^^^^^^^^^^^^^^^^^^^^^^^^^^^^^^^^^^^^^^^^^^

\cleardoublepage
\chapter{Classical Evaluations}
\label{chapter_6}
In this chapter, we present an evaluation of the multi-class learning and classification approaches presented in the previous chapter for object recognition.
There are two main approaches to multi-class classification: in single-label classification, each test object is assigned to one and only one category; and in multi-label classification, a test object can be assigned to one or more categories. In our case, both instance-based and model-based approaches are designed to perform single-label classification by taking advantage of the relation between object representation, similarity measure and classification rule. Each method has a set of parameters, in which its performance strongly depends on the suitability of the value chosen for each parameter. Therefore, in this chapter, we focus on classical evaluation of the proposed approaches for different parameter configurations to determine the best configurations.

To examine the performance of each approach with different configurations, the $k$-fold cross-validation procedure was used. This is one of the most widely used methods for estimating the generalization performance of a learning algorithm. In this case, the $k$ folds are randomly created by dividing the dataset into $k$ equal sized
subsets, where each subset contains examples from all the categories. In each iteration, a single fold is used for testing, and the remaining folds are used as training data. 
For $k$-fold cross-validation, we set $k$ to 10, as is generally recommended in literature. This type of evaluation is not only useful for parameter tuning, but it also provides straightforward results for comparing the different approaches among themselves and possibly with other approaches described in the literature. 

These evaluations are conducted using the datasets described in Section~\ref{datasets}. Evaluation metrics are discussed in Section~\ref{sec:evaluation_metrics}.
A set of experiments was carried out to evaluate the performance of the proposed instance-based object category learning and recognition approaches. The obtained results are reported and discussed in Section \ref{sec:IBL_results}. Then, the proposed model-based approaches are evaluated in Section \ref{sec:model_base_learning_results}. In Section \ref{sec:good_descriptor_evals}, we evaluate in detail the proposed Global Orthographic Object Descriptor (GOOD) concerning {\emph{descriptiveness}}, {\emph{scalability}}, {\emph{robustness}} and {\emph{efficiency}} characteristics. It is worth mentioning that all tests were performed with an i7, 2.40GHz processor, and 16GB RAM. 
Finally, summary and future research are presented and discussed.

%%%%%%%%%%%%%%%%%%%%%%%%%%%%%%%%%%%%%%%%%%%%%%%%%%%%%%%%%%%%%%%%%%%%%%%%%%%
%%%%%%%%%%%%%%%%%%%%%%%%%%%%%%%%%%%%%%%%%%%%%%%%%%%%%%%%%%%%%%%%%%%%%%%%%%%
\section {Datasets}
\label{datasets}
The experimental evaluation of different approaches, reported in this chapter, was carried out on two object images datasets, including the Restaurant Object Dataset  \citep {KasaeiInteractive2015} and the Washington RGB-D Object Dataset \citep{lai2011large}. 
In this chapter, we mainly use the restaurant object dataset for evaluating the performances of the proposed approaches because it has a small number of classes with significant intra-class variation, which is suitable for performing extensive sets of experiments. The Washington RGB-D object dataset is used for evaluating the GOOD descriptor with respect to several characteristics. The following subsections describe these datasets in more detail.

\begin{figure}[!b]
\center
\begin{tabular}[width=1\textwidth]{cc}
 \includegraphics[width=0.3\linewidth, trim=0.0cm 0cm 0cm 0cm,clip=true]{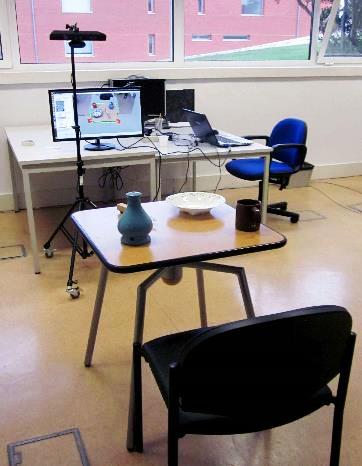} &
 \includegraphics[width=0.65\linewidth, trim= 0cm 0cm 0cm 0cm,clip=true]{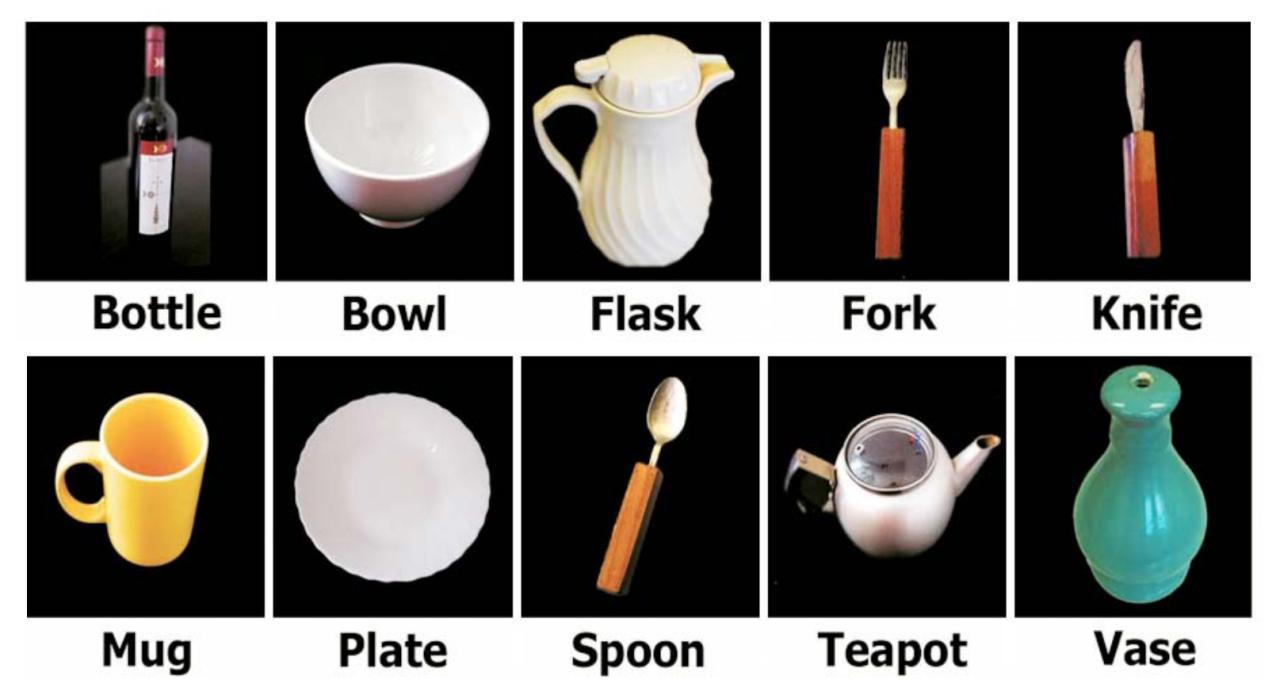}
\end{tabular}
\caption{ Restaurant Object Dataset: (\emph{left}) system setup for collecting dataset; (\emph{right}) objects in the Restaurant Object Dataset. Each object shown here belongs to a different category.}
\label{fig:restaurant_object_dataset_ch6}       % Give a unique label
\end{figure}

\subsection {Restaurant Object Dataset} 
\begin{figure}[!t]
\center
 \includegraphics[width=0.8\linewidth, trim=0.0cm 0cm 0cm 0cm,clip=true]{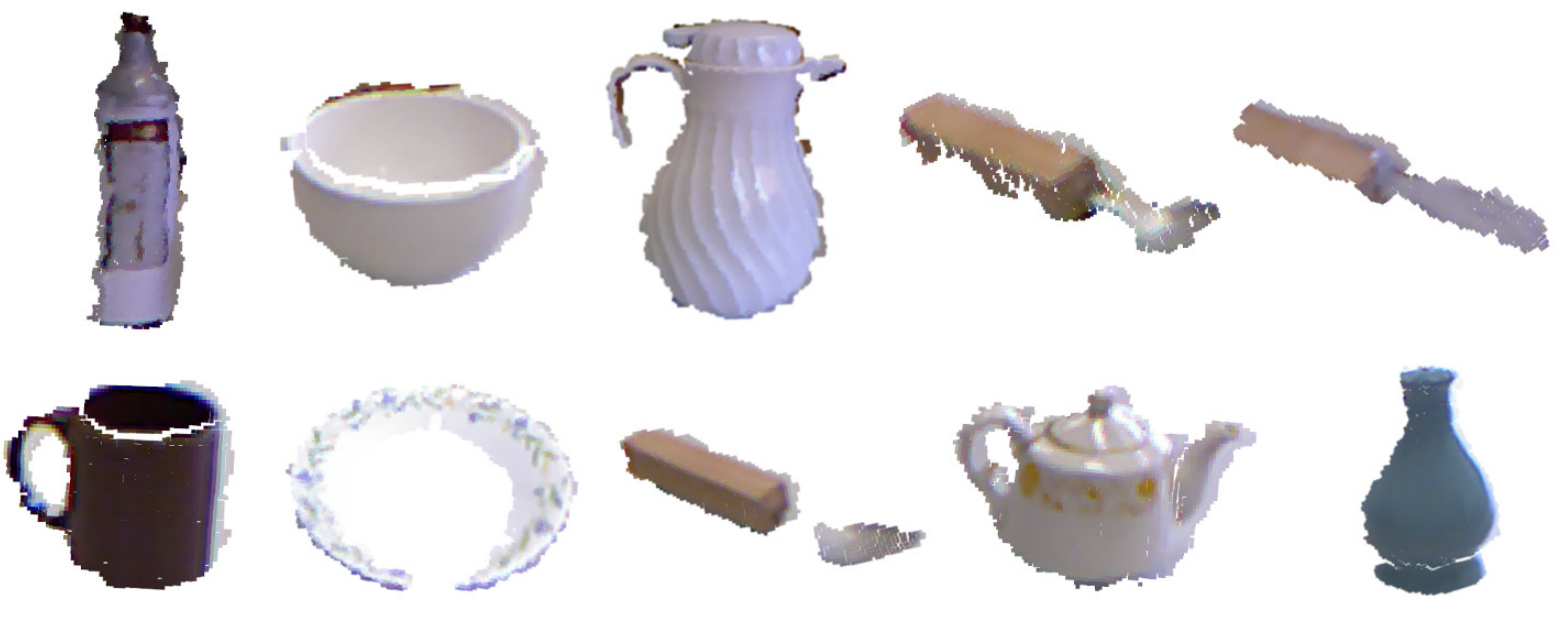} 
 \caption{Sample point clouds of objects in Restaurant object dataset.}
\label{fig:restaurant_object_dataset_ch6_point_clouds}       % Give a unique label
\end{figure}

The Restaurant Object Dataset\footnote{Restaurant Object Dataset is available online at : \url{https://goo.gl/64IXx9}} was created in the framework of the RACE project. It contains 341 views of instances of 10 categories from different perspectives (\emph{Bottle}, \emph{Bowl}, \emph{Flask}, \emph{Fork}, \emph{Knife}, \emph{Mug}, \emph{Plate}, \emph{Spoon}, \emph{Teapot}, and \emph{Vase}) and 30 views of false or unknown objects (e.g. points that belong to the instructor’s hands) \citep{KasaeiInteractive2015}. Each object view contains exactly one object. These object views were extracted from 100 views of table-top scenes by running the proposed object detection module and storing the segmented point clouds. In our setting, a Kinect camera is placed about one meter away from the table. This is the minimum distance required for the Kinect camera to return reliable depth data. The images were recorded with the camera mounted on a tripot at approximately $1.50m$ height and $45$ degrees relative to the table (Fig.~\ref{fig:restaurant_object_dataset_ch6} \emph{left}).
All segmented point clouds, i.e., the object views, were hand-annotated with the respective category labels. Figure~\ref{fig:restaurant_object_dataset_ch6_point_clouds} displays one sample point cloud per category and can give an idea of the type of  objects in this dataset.

\subsection {Washington RGB-D Object Dataset}

\begin{figure}[!h]
\vspace{1mm}
\hspace{-4mm}
 \includegraphics[width=1.05\linewidth, trim=0.0cm 0cm 0cm 0cm,clip=true]{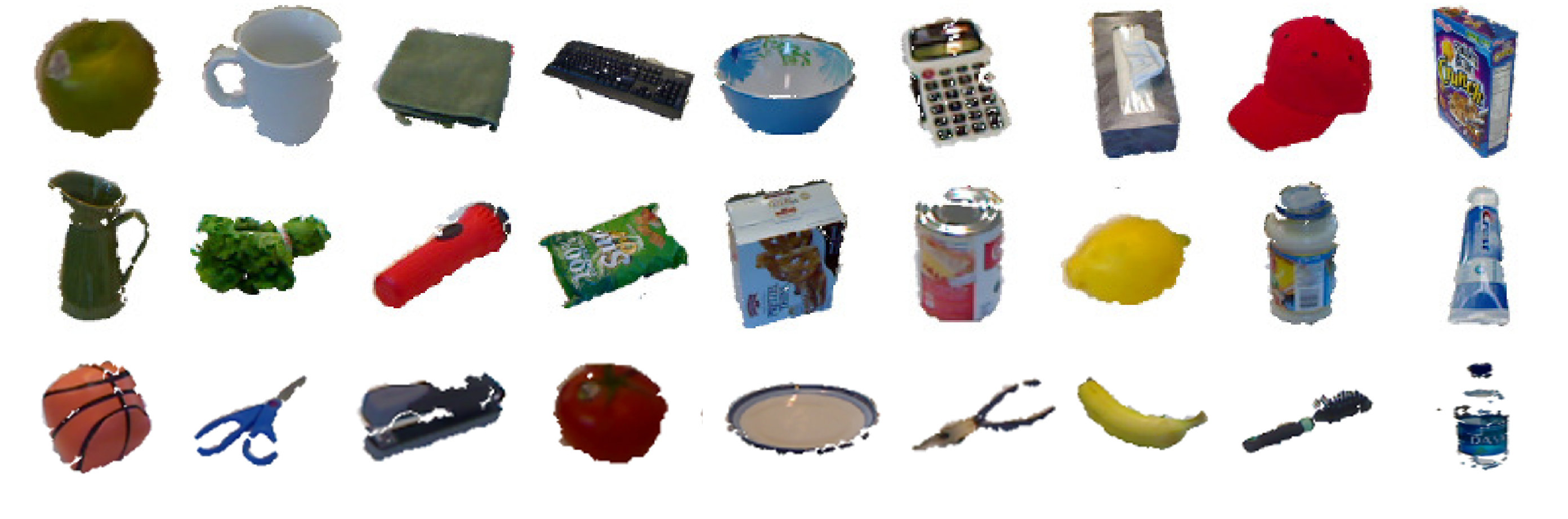} 
 \caption{Sample point clouds of objects in Washington RGB-D Object Dataset.}
\label{fig:RGBD_object_dataset_ch6}       % Give a unique label
\end{figure}

Several experiments were also carried out on one of the largest available RGB-D object
datasets namely the Washington RGB-D Object Dataset \citep{lai2011large}. This dataset is highly popular in the computer vision and robotics communities for evaluating 3D object recognition methods. The Washington RGB-D Object Dataset is a large scale dataset with respect to the number of images. It consists of 250,000 views of 300 common household objects taken from multiple views. The objects are organized into 51 categories arranged using WordNet \citep{miller1995wordnet} hypernym-hyponym relationships (similar to ImageNet \citep{deng2009imagenet}). This dataset was recorded using a Kinect style 3D camera that records synchronized and aligned $640 \times 480$ RGB and depth images. Each object was placed on a turntable and video sequences were captured for one whole rotation. We have excluded the Ball and Binder categories because of high shape similarity to the Apple and Notebook categories respectively.
Since we are using only depth information, it is impossible to distinguish these categories only based on shape features. Therefore, we worked with the remaining 49 categories. Figure~\ref{fig:RGBD_object_dataset_ch6} shows some example objects from the dataset \citep{lai2011large}.

%%%%%%%%%%%%%%%%%%%%%%%%%%%%%%%%%%%%%%%%%%%%%%%%%%%%%%%%%%%%%%%%%%%%%%%%%%%
%%%%%%%%%%%%%%%%%%%%%%%%%%%%%%%%%%%%%%%%%%%%%%%%%%%%%%%%%%%%%%%%%%%%%%%%%%%
\section{Evaluation Metrics}
\label{sec:evaluation_metrics}

In machine learning literature, the main metrics for evaluating supervised multi-class learning algorithms are \emph{accuracy}, micro and macro \emph{precision} (i.e., indices $\mu$ and $m$ respectively) and micro and macro \emph{recall}. In multi-class classification, each test instance gets assigned to one of the $N$ classes leading to an $N \times N$ confusion matrix, $A$. All the mentioned evaluation metrics can be computed from the confusion matrix as:
\begin{itemize}
\item \emph{accuracy} is computed as the fraction of correct predictions to total number of predictions:
	\begin{equation}
	\label{accuracy}
	accuracy =\frac{1}{|A|}\sum_{i=1}^{N} A_{i,i}
	\end{equation}
	
\item $precision_\mu$ is the ratio of number of correct predictions to total number of predictions.
	\begin{equation}
	\label{precision_u}
	precision_\mu =\frac{\sum_{i=1}^{N} tp_i}{\sum_{i=1}^{N} (tp_i + fp_i)},
	\end{equation}

\item $precision_m$ is the average precision for $N$ categories. The precision of each category is computed as the ratio of number of correct predictions to total number of instances that are predicted to belong to that category.
	\begin{equation}
	\label{precision_m}
	precision_m =\frac{1}{N}\sum_{i=1}^{N}\frac{tp_i}{tp_i + fp_i},
	\end{equation}
	  
\item $recall_\mu$: is the ratio of the number of correct predictions to the total number of instances of the known categories.
	\begin{equation}
	\label{recall_u}
	recall_\mu =\frac{\sum_{i=1}^{N} tp_i}{\sum_{i=1}^{N} tp_i+fn_i}.
	\end{equation}

\item $recall_m$ is the average recall across the N categories.

	\begin{equation}
	\label{recall_m}
	recall_m =\frac{1}{N}\sum_{i=1}^{N}\frac{tp_i}{tp_i + fn_i},
	\end{equation}
	  
\end{itemize}

\noindent 
where $|A| = \sum_{i,j} A_{i,j}$ is the total number of predictions, $tp$ and $fp$ are true positive and false positive respectively. False negative and true negative are shown by $fn$ and $tn$ respectively.  It is worth mentioning that the difference between $precision_\mu$ and $precision_m$ (also valid for $recall_\mu$ and $recall_m$) is that $precision_m$ gives equal weight to each category, whereas $precision_\mu$ gives equal weight to each prediction. Furthermore, there are many other common metrics in machine learning literature such as F-measure, kappa statistic, etc \citep{powers2011evaluation}. The focus of this chapter is on evaluating the performance of the proposed approaches. The choice of evaluation metric is based on its suitability. In multi-class learning and recognition, if an approach can not detect unknown objects, then $accuracy$, $precision_\mu$ and $recall_\mu$ will lead to the same results. Therefore, we use \emph{accuracy} as the primary evaluation measure in our experimental setup in this chapter.

%%%%%%%%%%%%%%%%%%%%%%%%%%%%%%%%%%%%%%%%%%%%%%%%%%%%%%%%%%%%%%%
\section {Instance-Based Learning}
\label{sec:IBL_results}
The instance-based learning approach is a baseline approach to evaluate object representations for the purpose of object recognition. Experiments in this section mainly reflect how the different object representations support object recognition. Each approach has a set of parameters that must be tuned to provide a good balance between recognition performance, memory usage and computation time. As mentioned earlier, the restaurant object dataset has a small number of classes with significant intra-class variation. Therefore, we use this dataset to examine the accuracy of each approach using different parameter configurations. Based on the obtained results, we finally select a default value for each parameter of all the approaches.

%%%%%%%%%%%%%%%%%%%%%%%%%%%%%%%%%%%%%%%%%%%%%%%%%%%%%%%%%%%%%%%

\subsection {Sets of Local Features}
\label{sec:IBL_results_spin_images}
For the instance-based approaches with variable size representations, i.e.,  Approach~I and  Approach~II, as discussed in section~\ref{sec:variable_size_representations_sets_of_local_features}, two sets of $24$ experiments were performed for different values of the three system parameters namely the voxel size (VS), which is related to the number of keypoints extracted from each object view, the image width (IW) and the support length (SL) of spin images. Results are presented in Table \ref{table:system_parameters_instance_based_feature_layer}. The accuracy value presented for each parameter value is an average computed over the experiments carried out for all combinations of the other parameters values. One important observation is that, by increasing the SL, the classification performance also increases. This is due to the fact that the support length parameter determines the amount of space swept out by a spin-image, which will have a radius of SL and height of $2$SL. Therefore, selecting a large value for the SL parameter causes the spin image to behave like a global descriptor. To prove this observation, we ran a set of $18$ experiments for each approach. Results are summarized in Table \ref{table:system_parameters_instance_based_feature_layer_global}. 
The parameters that obtained the best average accuracy were selected as the default system parameters. For  Approach~I, the default system configurations are the following: VS~=~$0.05$, IW = $8$ and SL = $0.10$. The accuracy of the proposed system with the default configuration was $81$ percent. For  Approach~II, the following parameters obtained the best average accuracy and were set as the default system parameters: VS~=~$0.04$, IW = $8$ and SL = $0.10$. The accuracy of the proposed system with this configuration was $0.87$. Results show that the overall performance of the recognition system is promising. Spin images are capable of collecting distinctive traits of the local surface patches of each object. For an entire experiment, the average computation time for the first instance-based approach ( Approach~I) was around $1728.60$ seconds while the second approach ( Approach~II) in average took around $1757.20$ seconds. It can be concluded that both approaches require significant computational resources. 
This set of experiments shows the pros and cons of considering a large value for the SL parameter. We conclude that a larger SL adds more information and causes the object representation takes more time to be computed. Since we mainly focus on robotics applications, especially open-ended learning, the computation time is an important factor. Therefore, we prefer to choose a value for the support length parameter in the range of 0.02m to 0.05m in the following approaches.
\begin{table*}[!b]
\begin{center}
\caption {Average object recognition accuracy using the proposed variable size representations by choosing small values for the SL parameter.}
\label{table:system_parameters_instance_based_feature_layer}
\resizebox{0.75\columnwidth}{!}{
\begin{tabular}{ |c|c|c|c|c|c|c|c|c|c|c|c|c|c|c|c|c| }
\hline
\multicolumn{2}{|c|}{Parameters} & \multicolumn{3}{|c|}{VS (m)} &\multicolumn{2}{c}{IW (bins)} &
\multicolumn{4}{|c|}{SL (m)} \\
\hline
\multicolumn{2}{|c|}{Values} & 0.01 & 0.02 & 0.03 & 4 & 8 & 0.02 & 0.03 & 0.04 & 0.05 \\
\hline
\multicolumn{2}{|c|}{ Approach~I}& 21 & \textbf{22} & 21 & \textbf{25} & 17 & 12 & 16 & 22 & \textbf{34}\\\cline{2-11}
\multicolumn{2}{|c|}{ Approach~II} & 34 & 40 & \textbf{41} & \textbf{43} & 34 & 22 & 29 & 46 & \textbf{56}\\
\hline
\end{tabular}
}
\vspace{-4mm}
\end{center}
\end{table*}

\begin{table*}[!b]
\begin{center}
\caption {Average object recognition accuracy using the proposed variable size representations by choosing large values for the SL parameter.}
\label{table:system_parameters_instance_based_feature_layer_global}
\resizebox{0.75\columnwidth}{!}{
\begin{tabular}{ |c|c|c|c|c|c|c|c|c|c|c|c|c|c|c|c|c|c| }
\hline
\multicolumn{2}{|c|}{Parameters} & \multicolumn{3}{|c|}{VS (m)} &\multicolumn{2}{c}{IW (bins)} &
\multicolumn{3}{|c|}{SL (m)} \\
\hline
\multicolumn{2}{|c|}{Values} & 0.03 & 0.04 & 0.05 & 4 & 8 & 0.06 & 0.08 & 0.10 \\
\hline
\multicolumn{2}{|c|}{ Approach~I} & 62 & 64 & \textbf{67} & 61 & \textbf{68} & 53 & 67 & \textbf{73} \\\cline{2-10}
\multicolumn{2}{|c|}{ Approach~II} & 75 & \textbf{76} & 74 & 72 & \textbf{78} & 71 & 77 & \textbf{78} \\
\hline
\end{tabular}
}
\vspace{-4mm}
\end{center}
\end{table*}

%%%%%%%%%%%%%%%%%%%%%%%%%%%%%%%%%%%%%%%%%%%%%%%%%%%%%%%%%%%%%%%%%%%%%
\subsection {Bag of Words}
\label{BOW_results_chapter_5}
\begin{figure*}[!t]
\begin{centering}
\hspace{-4mm}
\begin{tabular}{c}
\includegraphics[width=1.05\linewidth, trim=3.5cm 0.13cm 1.cm 0.5cm,clip=true]{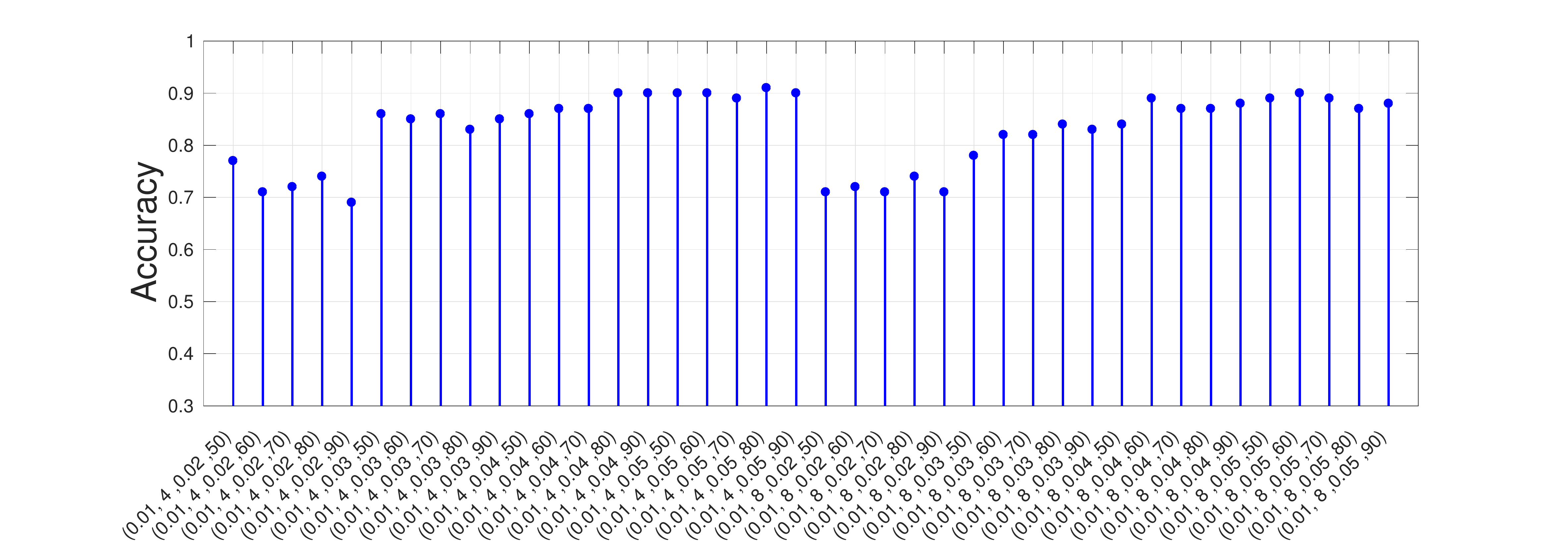}\\
\includegraphics[width=1.05\linewidth, trim=3.5cm 0.13cm 1.cm 0.5cm,clip=true]{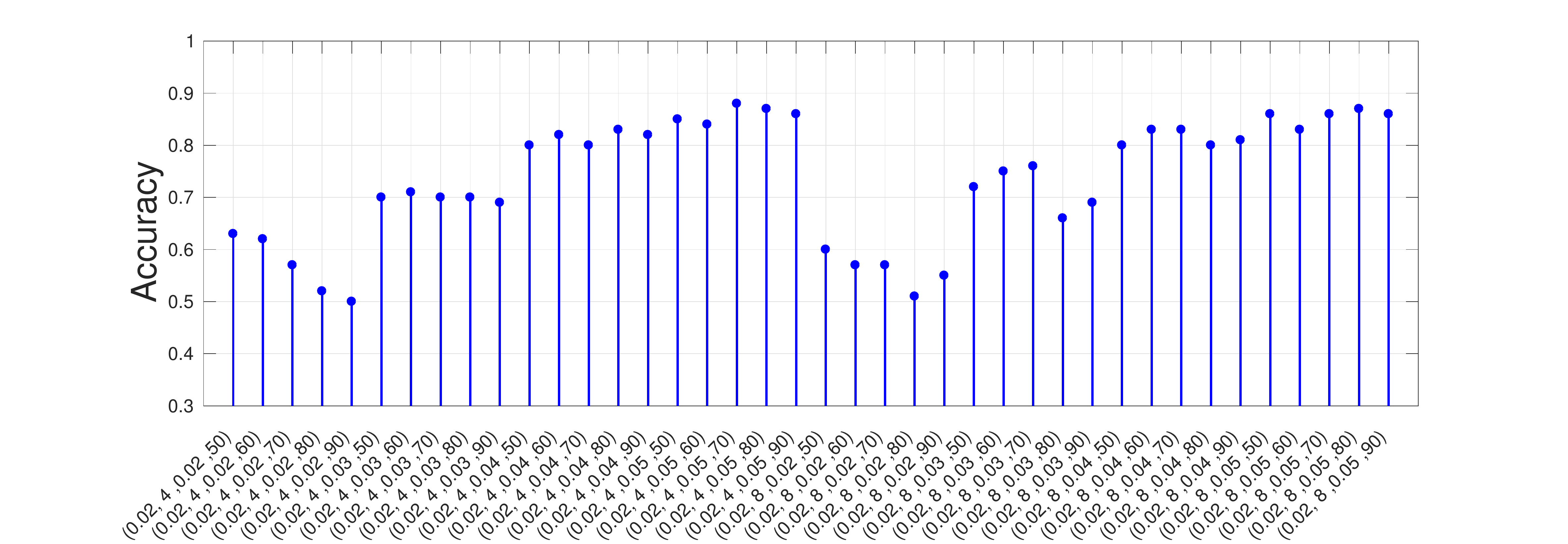}\\
\includegraphics[width=1.05\linewidth, trim=3.5cm 0.13cm 1.cm 0.5cm,clip=true]{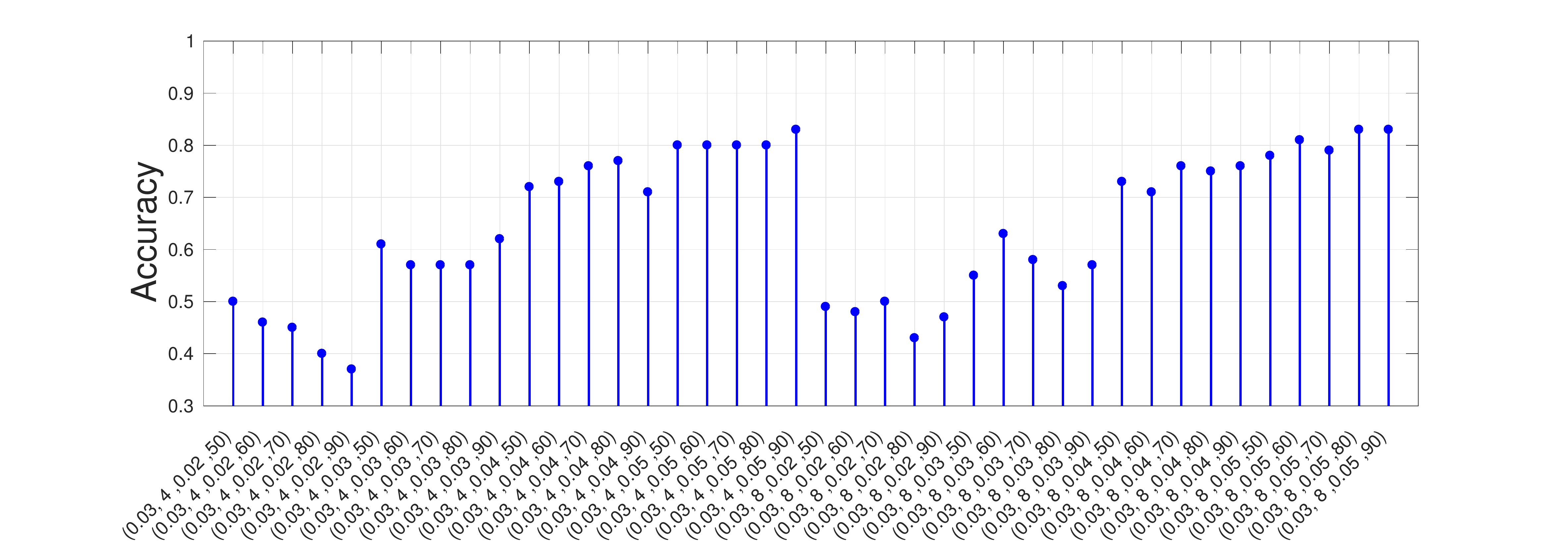}\\
\end{tabular}
\end{centering}
\caption{Object recognition performance for different values of the four parameters of the system in BoW layer; the system parameters are represented as a tuple (VS, IW, SL, DS).}
\label{fig:120exps_instance_based}
\end{figure*}

The descriptiveness of the BoW layer was evaluated with varying dictionary size. All the parameters must be well selected to provide a good balance between recognition performance, memory usage and computation time. Towards this end, a set of 120 experiments has been performed. Experiments were repeated for different values of four parameters of the system, namely the voxel size (VS), the image width (IW) and support length (SL) of spin images and the dictionary size (DS). Results are summarized in Table \ref{table:system_params_bow_layer} and Fig.~\ref{fig:120exps_instance_based}. Based on the obtained results, the default system configuration is set as follow: VS = $0.01$, IW~=~$4$, SL~=~$0.05$ and DS = $70$. The accuracy of this configuration was $89$ percent.

\begin{table*}[!h]
\begin{center}
\caption {Average object recognition performance for different parameters of the BoW representation.}
\vspace{2mm}
\resizebox{\columnwidth}{!}{
\begin{tabular}{ |c|c|c|c|c|c|c|c|c|c|c|c|c|c|c|c|c|c| }
\hline
Parameters & \multicolumn{3}{|c|}{VS (m)} &\multicolumn{2}{c}{IW (bins)} &
\multicolumn{4}{|c|}{SL (m)} &	\multicolumn{5}{|c|}{DS (dictionary size)} \\
\hline
Values & 0.01 & 0.02 & 0.03 & 4 & 8 & 0.02 & 0.03 & 0.04 & 0.05 & 50 & 60 & 70 & 80 & 90 \\
\hline
Avg. Accuracy(\%) & \textbf{83} & 74 & 65 & \textbf{74} & 74 & 58 & 71 & 81 & \textbf{85} & 74 & 74 & \textbf{74} & 73  & 73\\
\hline
\end{tabular}
}
\vspace{-4mm}
\label{table:system_params_bow_layer}
\end{center}
\end{table*}

%%%%%%%%%%%%%%%%%%%%%%%%%%%%%%%%%%%%%%%%%%%%%%%%%%%%%%%%%%%%%%%%%%%%%%%%%%%%%%%%
\subsection {Standard LDA}
\label{open_ended_LDA_instance_based_chapter5}

In the topic layer, we have four new parameters in addition to those of the Bag of Words layer: \emph{Number of Topics} (NT), the Dirichlet prior hyper-parameters $\alpha$ and $\beta$ and \emph{Number of Iterations} for Gibbs sampling. In this thesis, we assumed symmetric Dirichlet prior for both $\alpha$ and $\beta$ parameters. Therefore, a high $\alpha$ value means that each object is likely to contain a mixture of most of the topics, and not a single specific topic. Likewise, a low $\beta$ value means that a topic may contain a mixture of just a few of the words. We have ran several experiments and concluded that $\alpha$ and $\beta$ should be set to $1.0$ and $0.1$ respectively. Moreover, the number of Gibbs sampling iterations was set to $30$. Experiments must be performed for different values of the
remaining five parameters. In this section, the standard LDA approach is evaluated. This is the baseline to which the proposed Local LDA approach must be compared (see the next subsection). A set of 360 experiments was performed with standard LDA for different values of the considered parameters.

%Another set of experiments was carried out to evaluate the performance of LDA approach. Similar to the Local LDA approach, a set of $360$ experiments was performed for different values of the system parameters. For all experiments, we set $\alpha$ and $\beta$ to $1.0$ and $0.1$ respectively. Moreover, the number of Gibbs sampling iterations was set to $30$. 
The obtained results are summarized in Table~\ref{table:system_params_open_ended_lda} which shows the average accuracies obtained with different configurations. 
Similar to the previous approaches, the parameter configuration that obtained the best average accuracy and leads to better efficiency was selected as the default configuration: VS $= 0.02$, IW~$= 4$ and SL $= 0.05$, DS~$= 90$ and K~$= 30$. The performance of the topic layer with LDA (i.e., shared topics among all categories) was carried out using this configuration and the  accuracy of this configuration was $0.88$.
\begin{table*}[!h]
\begin{center}
\caption {Average object recognition performance for different parameters of the LDA approach.}
\vspace{2mm}
\resizebox{\columnwidth}{!}{
\begin{tabular}{ |c|c|c|c|c|c|c|c|c|c|c|c|c|c|c|c|c|c| }
\hline
Parameters & \multicolumn{3}{|c|}{VS (m)} &\multicolumn{2}{c}{IW (bins)} &
\multicolumn{4}{|c|}{SL (m)} &	\multicolumn{5}{|c|}{DS (dictionary size)} &	\multicolumn{3}{|c|}{K (topics)} \\
\hline
Values & 0.01 & 0.02 & 0.03 & 4 & 8 & 0.02 & 0.03 & 0.04 & 0.05 & 50 & 60 & 70 & 80 & 90 & 30 & 40 & 50\\
\hline
Avg. Accuracy(\%) & 78 & \textbf{78} & 75 & \textbf{79} & 79 & 64 & 73 & 78 & \textbf{81} & 81 & 82& 82& 82& \textbf{85}& \textbf{84} & 84 & 81\\
\hline
\end{tabular}
}
\vspace{-4mm}
\label{table:system_params_open_ended_lda}
\end{center}
\end{table*}

%%%%%%%%%%%%%%%%%%%%%%%%%%%%%%%%%%%%%%%%%%%%%%%%%%%%%%%%%%%%%%%%%%%%%%%%%%%%%%%%

\subsection {Local LDA}
\label{topic_layer_instance_based_chapter5}

A similar set of 360 experiments was performed with the proposed Local LDA approach for the same values of the five considered parameters. A summary of the experiments is
reported in Table~\ref{table:system_params}. The parameter configuration that obtained the best average accuracy was selected as the default configuration:  VS~$= ~0.03$, IW~$= 4$ and SL $= 0.05$, DS~$= 90$ and K~$=30$. The accuracy of this configuration was $0.91$. This configuration displays a good balance between recognition performance and
memory usage. 

\begin{table*}[!t]
\begin{center}
\caption {Average object recognition performance for different parameters of the Local LDA representation.}
\vspace{2mm}
\resizebox{\columnwidth}{!}{
\begin{tabular}{ |c|c|c|c|c|c|c|c|c|c|c|c|c|c|c|c|c|c| }
\hline
Parameters & \multicolumn{3}{|c|}{VS (m)} &\multicolumn{2}{c}{IW (bins)} &
\multicolumn{4}{|c|}{SL (m)} &	\multicolumn{5}{|c|}{DS (dictionary size)} &	\multicolumn{3}{|c|}{K (topics)} \\
\hline
Values & 0.01 & 0.02 & 0.03 & 4 & 8 & 0.02 & 0.03 & 0.04 & 0.05 & 50 & 60 & 70 & 80 & 90 & 30 & 40 & 50\\
\hline
Avg. Accuracy(\%) & 81 & 82 & \textbf{86} & \textbf{84} &83 & 78 & 81 & 83 & \textbf{84} & 82 & 82& 83& 85& \textbf{85}& \textbf{88} & 85 & 84\\
\hline
\end{tabular}
}
\vspace{-4mm}
\label{table:system_params}
\end{center}
\end{table*}

%%%%%%%%%%%%%%%%%%%%%%%%%%%%%%%%%%%%%%%%%%%%%%%%%%%%%%%%%%%%%%%%%%%%%%%%%%%%%%%%%%%%
\subsection {GOOD}
\label{GOOD_Representation_IBL}
\begin{wrapfigure}{r}{0.5\textwidth}
\vspace{-10mm}
  \begin{center}
   \includegraphics[width=0.9\linewidth]{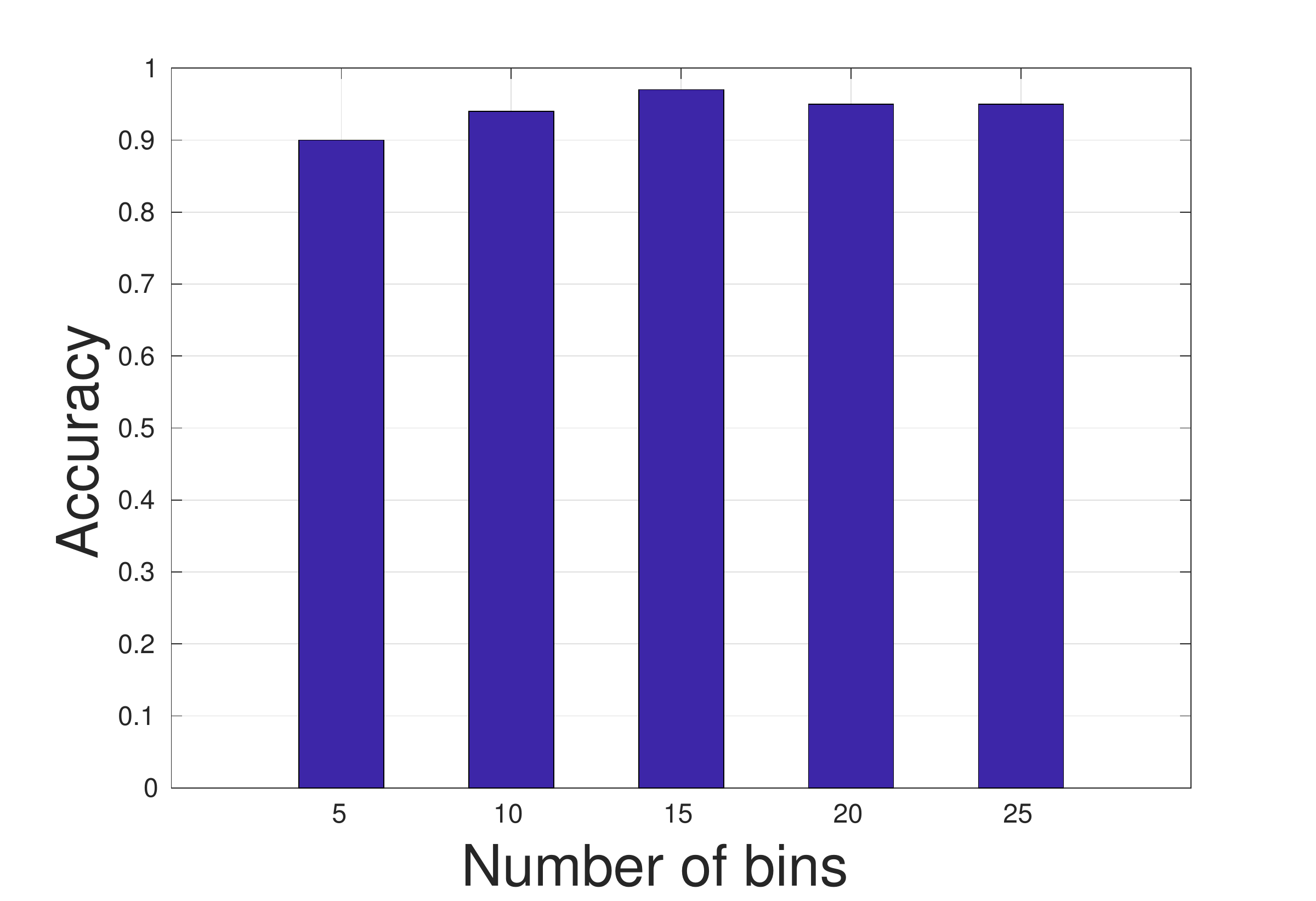}
  \end{center} 
    \vspace{-6mm}
  \caption{Object recognition performance using GOOD feature.}
  \label{fig:good_ibl}
\end{wrapfigure}

A set of experiments was performed to evaluate the performance of instance-based learning using the GOOD descriptor. As described in chapter~\ref{chapter_4}, GOOD has single a parameter namely the \emph{number of bins}, $n$. We performed five experiments for different values of this parameter: $ 5,~10,~15,~20,~25$. Results are summarized in Fig.~\ref{fig:good_ibl}. In these experiments, the best result was obtained with $n = 15$ bins. The accuracy of the proposed system with this configuration was $97$ percent. The experiment time for this approach on average was around $85.10$ seconds.

By comparing the results obtained with the different instance-based approaches using the default configurations, several points can be concluded. One important observation is that the overall performance, concerning classification accuracy and computation time, obtained with GOOD descriptor is clearly better than the best performances obtained with the local-feature based approaches. Experimental results also show that, among approaches based on local-feature, the overall performance of the recognition system based on topic modelling is promising and the proposed Local LDA representation is capable of providing a distinctive representation for the objects. Moreover, it was observed that the discriminative power of the Local LDA representation was better than the other local-feature based approaches. In addition, independent topics for each category provide better representation than shared topics for all categories. Furthermore, it has been observed that the discriminative power of shared topics depends on the order of introduction of categories \citep{AndoTocic}. The accuracy of object recognition based on variable size representation (i.e., sets of spin-images) was not as good as the other approaches. The BoW  and LDA (shared topics) representations obtained an acceptable performance. The local topic representation provided a good balance between memory usage and descriptiveness. The variable size representations were the less compact representations.

%^^^^^^^^^^^^^^^^^^^^^^^^^^^^^^^^^^^^^^^^^^^^^^^^^^^^^^^^^^^^^^^^
%^^^^^^^^^^^^^^^^^^^^^^^^^^^^^^^^^^^^^^^^^^^^^^^^^^^^^^^^^^^^^^^^
\section {Model-Based Learning}
\label{sec:model_base_learning_results}

The performance of the model-based object category learning approach was evaluated using BoW, LDA (standard and local) and GOOD representations. The variable-size representation approaches are not considered since most of the machine learning methods, including the proposed model-based learning approach, take fixed-length vectors as input. In the case of BoW and Local LDA, three parameters (IW, SL and DS) have different effects on both memory usage and recognition accuracy. Dictionary size (DS) determines the number of words in the dictionary. For LDA approaches, an additional parameter, NT is the number of topics. We tried to find a good value for these parameters to obtain a good balance between recognition performance, memory usage, and processing speed. In the case of GOOD, the \emph{number of bins}, is the parameter that has to be set optimally. We describe in detail each set of experiments in the following subsections.

%%%%%%%%%%%%%%%%%%%%%%%%%%%%%%%%%%%%%%%%%%%%%%%%%%%%%%%%%%%%%%%%%%%%%%%%%%%%%%%%

\subsection {Bag of Words}
\label{BoW_NB_chapter6}
\begin{figure}[!t]
\centering
\begin{centering}
\begin{tabular}{c}
\hspace{-7mm}
\includegraphics[width=1.1\linewidth, trim=3.5cm 0.13cm 1.cm 0.5cm,clip=true]{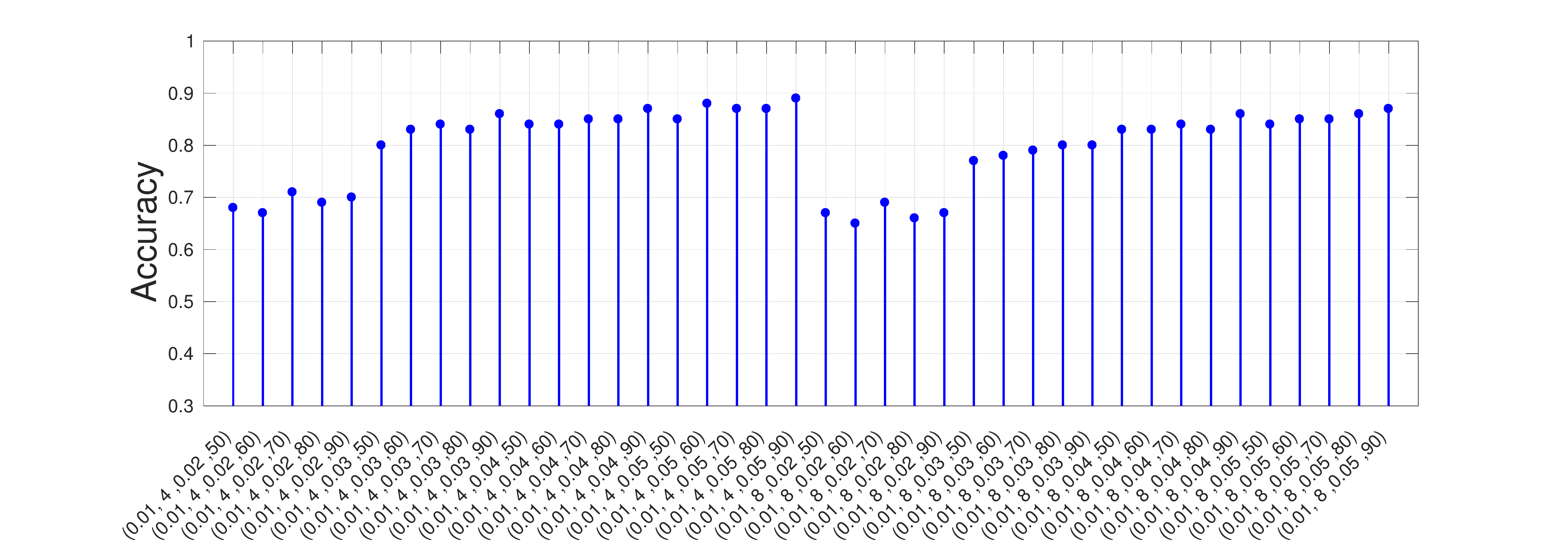}\\\hspace{-7mm}
\includegraphics[width=1.1\linewidth, trim=3.5cm 0.13cm 1.cm 0.5cm,clip=true]{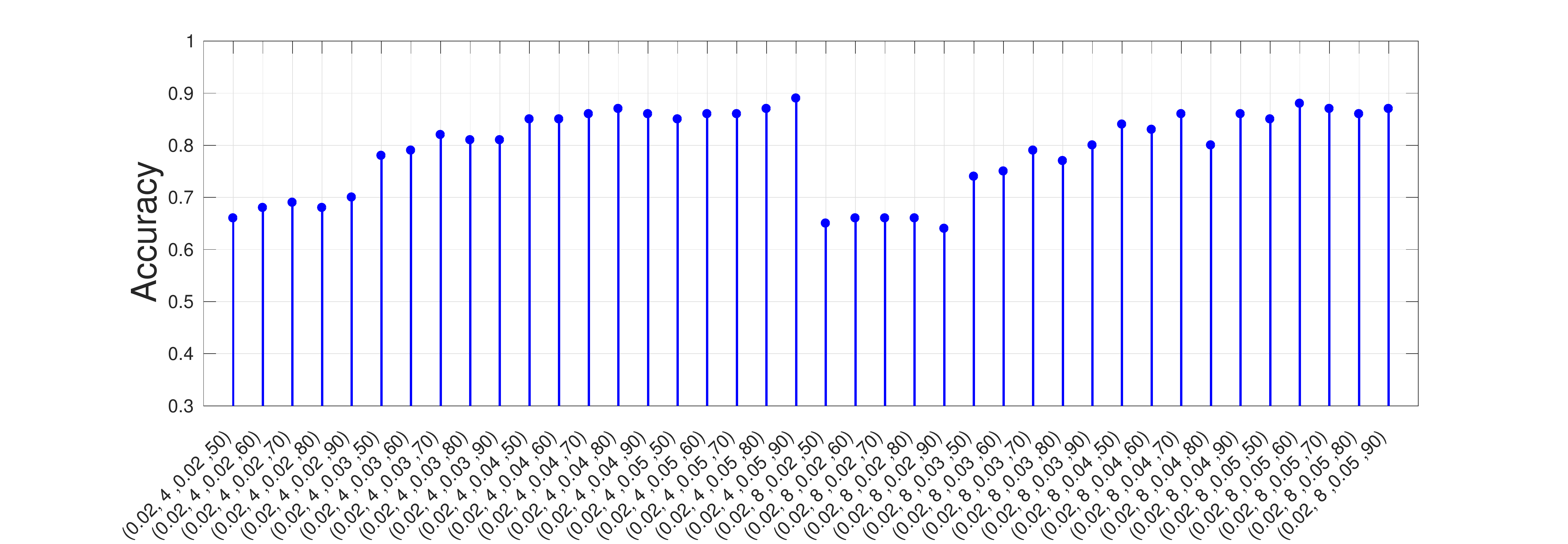}\\\hspace{-7mm}
\includegraphics[width=1.1\linewidth, trim=3.5cm 0.13cm 1.cm 0.5cm,clip=true]{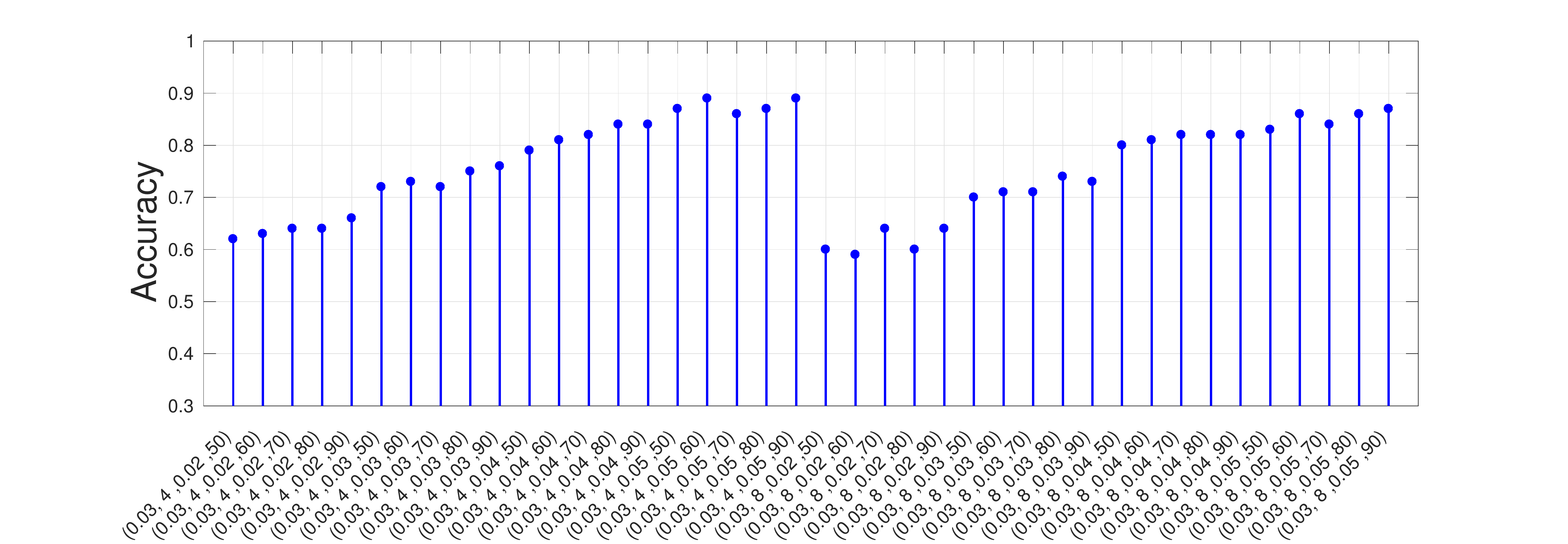}\\
\end{tabular}
\end{centering}
\caption{Object recognition performance for different values of four parameters of the system (i.e., Naive Bayes with BoW representation); the system parameters are represented as a tuple (VS, IW, SL, DS).}
\vspace{-5mm}
\label{fig:120exps}
\end{figure}

A set of 120 experiments was carried out to evaluate the proposed model-based learning approach with BoW representation for different values of the four BoW parameters. The obtained results are summarized in Table \ref{table:system_params_bow_layer_model_based}. The object recognition performance for each system configuration is depicted in figure \ref{fig:120exps} where the system parameters are represented as a tuple (VS, IW, SL, DS). 
The parameters that obtained the best average accuracy were selected as the default system parameters: VS~=~0.01, IW~=~4, SL = 0.05 and DS = 90. The accuracy of the proposed system with this configuration was $89\%$. 

One important observation is that the accuracy of the proposed model-based object recognition approach using BoW is similar to the instance-based approach with BoW (see section~\ref{BOW_results_chapter_5}). Although, the instance-based approach achieves a good description power, it increases computation time, memory usage and sensitivity to outliers. In instance-based approaches, the number of instances directly affects computation time and memory usage. Therefore, we calculate the average time required to complete all the 120 experiments for the proposed instance-based and model-based approaches using BoW. The average computation time of the model-based approach is around $4.3\%$ less than the instance-based approach with BoW representation. In particular, the average computation time for the instance based approach was around $195.60$ seconds while the model-based approach on average took $187.10$ seconds. 
It is worth mentioning that the difference of computation time was significant for such a small dataset and could grow exponentially by increasing the number of instances and categories.

\begin{table}[!h]
\begin{center}
    \caption {Average object recognition performance using BoW representation.}
\resizebox{\columnwidth}{!}{
\begin{tabular}{ |c|c|c|c|c|c|c|c|c|c|c|c|c|c|c| }
\hline
Parameters & \multicolumn{3}{|c|}{VS (m)} &  \multicolumn{2}{|c|}{IW (bins)} &	\multicolumn{4}{|c|}{SL (m)}&\multicolumn{5}{|c|}{DS (dictionary size)}\\
\hline
Values & 0.01 & 0.02 & 0.03 & 4 & 8 &0.02 & 0.03&0.04 & 0.05& 50 & 60 & 70 &80 & 90\\
\hline
Avg. Accuracy(\%) & \textbf{80} & 79 & 76  & \textbf{79} & 77& 66 & 77& 84 &\textbf{86} & 77 & 78 & 79 & 78 & \textbf{80}\\
\hline
\end{tabular}}
\label{table:system_params_bow_layer_model_based}
\vspace{-0mm}
\end{center}
\end{table}

%%%%%%%%%%%%%%%%%%%%%%%%%%%%%%%%%%%%%%%%%%%%%%%%%%%%%%%%%%%%%%%%%%%%%%%%%%%%%%%%%%%%
\subsection {Standard LDA}
\label{naive_bayes_open_ended_topic}

Another round of experiments was carried out for different values of five parameters of the LDA approach. Similar to the previous experiments, we tried to find a good value for these parameters to obtain a good balance between recognition performance, memory usage, and processing speed. A summary of all experiments is reported in Table~\ref{table:system_parameters_model_based_open_ended_LDA}. The parameters that obtained the best average accuracy was selected as the default configuration: VS~$=~0.02$, IW~$=~4$ and SL $= 0.05$, DS $= 90$, K $= 30$. The accuracy of the
proposed system in topic layer with the default configuration was $90$ percent.  It can be noticed that the model-based learning approach with LDA representation performs better than the instance-based approach with LDA representation. We believe that this difference can be explained by the fact that the former approach is not sensitive to the outliers. 
\begin{table*}[!t]
\begin{center}
\vspace{2mm}
\caption {Average object recognition performance using LDA representation.}
\vspace{2mm}
\resizebox{\columnwidth}{!}{
\begin{tabular}{ |c|c|c|c|c|c|c|c|c|c|c|c|c|c|c|c|c|c| }
\hline
Parameters & \multicolumn{3}{|c|}{VS (m)} &\multicolumn{2}{c}{IW (bins)} &
\multicolumn{4}{|c|}{SL (m)} &	\multicolumn{5}{|c|}{DS (dictionary size)} &	\multicolumn{3}{|c|}{K (topics)} \\
\hline
Values & 0.01 & 0.02 & 0.03 & 4 & 8 & 0.02 & 0.03 & 0.04 & 0.05 & 50 & 60 & 70 & 80 & 90 & 30 & 40 & 50\\
\hline
Avg. Accuracy(\%) & 76 & \textbf{77} & 77 & \textbf{80} & 79 & 65 & 72 & 79 & \textbf{83} & 83 & 83& 84& 83& \textbf{85}& \textbf{82} & 82 & 80\\
\hline
\end{tabular}
}
\label{table:system_parameters_model_based_open_ended_LDA}
\end{center}
\end{table*}

The computation time for a complete experiment of model-based learning with LDA representation (including both learning and recognition phases) on average was $198.30$ seconds, which is $1.15$ times less than the instance-based learning with the same object representation. In terms of computation time, it was observed that among the object representation approaches based on local-feature, the BoW with model-based learning approach achieved the best performance, which is around $1.05$ and $1.57$ times less than model-based approach with LDA and Local LDA representations respectively. The underlying reason is that there is a Gibbs sampling procedure in the LDA approaches which takes time to accurately represent the desired distribution. 

%%%%%%%%%%%%%%%%%%%%%%%%%%%%%%%%%%%%%%%%%%%%%%%%%%%%%%%%%%%%%%%%%%%%%%%%%%%%%%%%%%%%
\subsection {Local LDA}
\label{naive_bayes_topic}

Another round of experiments was performed to evaluate the system using the proposed Local LDA representation and model-based object category learning and recognition.
In this case, $360$ experiments were carried out. In these experiments, the parameters that got the best average accuracy were: VS~$=~0.03$, IW~$=~4$ and SL $= 0.05$, DS $= 90$, K $= 30$. This is adapted as the default configuration. The accuracy of the system with this configuration was $93$ percent. Table \ref{table:system_parameters_model_based_topic_layer} provides a detailed summary of the obtained results. 

\begin{table*}[!b]
\begin{center}
\vspace{2mm}
\caption {Average object recognition performance using Local LDA representation.}
\vspace{2mm}
\resizebox{\columnwidth}{!}{
\begin{tabular}{ |c|c|c|c|c|c|c|c|c|c|c|c|c|c|c|c|c|c| }
\hline
Parameters & \multicolumn{3}{|c|}{VS (m)} &\multicolumn{2}{c}{IW (bins)} &
\multicolumn{4}{|c|}{SL (m)} &	\multicolumn{5}{|c|}{DS (dictionary size)} &	\multicolumn{3}{|c|}{K (topics)} \\
\hline
Values & 0.01 & 0.02 & 0.03 & 4 & 8 & 0.02 & 0.03 & 0.04 & 0.05 & 50 & 60 & 70 & 80 & 90 & 30 & 40 & 50\\
\hline
Avg. Accuracy(\%) & 82 & 86 & \textbf{91} & \textbf{87} &84 & 83 & 86 & 88 & \textbf{90} & 84 & 86 & 87 & 89& \textbf{92}& \textbf{92} & 87 & 86\\
\hline
\end{tabular}
}
\label{table:system_parameters_model_based_topic_layer}
\end{center}
\end{table*}

Similar to the previous evaluation, by comparing the obtained results of instance-based and the model-based approaches, the value of model based learning approaches is highlighted. The average computation time of the model-based approach with Local LDA was around $1.18$ times less than the instance-based approach with Local LDA. Specificity, for all experiments, the average computation time for the instance-based approach was $348.50$ seconds while model-based approach on average took $293.80$ seconds. It was observed that the proposed local topic modelling is capable to provide distinctive object representation for recognizing different type of objects. 

%%%%%%%%%%%%%%%%%%%%%%%%%%%%%%%%%%%%%%%%%%%%%%%%%%%%%%%%%%%%%%%%%%%%%%%%%%%%%%%%%%%%%%%%%%%%%%%%%%%%%%%%%%%%%%%%
\subsection{GOOD}
\label{GOOD_Representation_NB} 
\begin{wrapfigure}{r}{0.5\textwidth}
\vspace{-10mm}
  \begin{center}
   \includegraphics[width=0.95\linewidth]{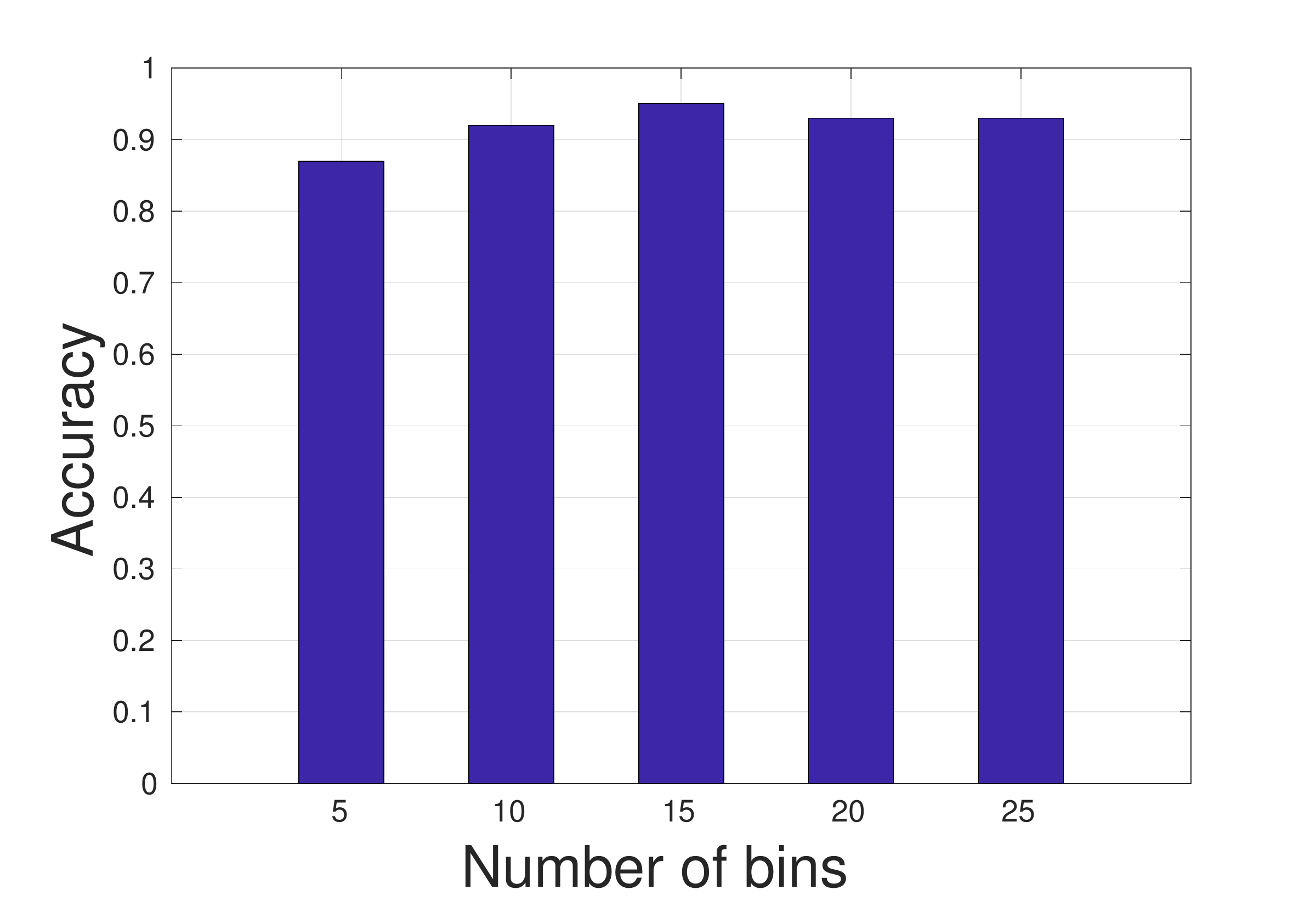}
  \end{center} 
    \vspace{-7mm}
  \caption{Object recognition performance using GOOD feature.}
      \vspace{-3mm}
  \label{fig:good_nb}
\end{wrapfigure}
A set of experiments was carried out to evaluate the proposed model-based learning approach representing object views with the GOOD descriptor. 
In particular, we performed five 10-fold cross-validation experiments with different values for the \emph{number of bins} parameter: $5,~10,~15,~20,$ and $25$. Results are plotted in Fig.~\ref{fig:good_nb}. Although a large number of bins may provide more details about the point distribution, it increases computation time and  memory usage. Therefore, since the difference in accuracy is relatively small, $n$ is set to $15$ bins by default. This is a good balance between recognition performance, memory usage, and processing speed (see additional details in section~\ref{sec:good_descriptor_evals}). The accuracy of the proposed system with this configuration was $95$ percent. It is interesting that instance-based learning with GOOD achieved better accuracy than the model-based approach. We believe that the reason lies in the learning procedure. In the model-based approach, a probabilistic model
for a given category is created by combining information from all the instances of the category. This procedure works fine with all local feature-based object representations. In the case of a global object representation, i.e., GOOD, the presence of some instances with diverging appearances, i.e., totally different from the other instances, will lead to create a poor model for the given category. The instance-based approach with GOOD does not suffer from this drawback and therefore, showed a better performance.

\begin{table*}[!b]
\begin{center}
\caption {Summary of experiments for different object representations and category learning architectures. }
\label{table:summary_chapter6}
\resizebox{0.8\linewidth}{!}{
\begin{tabular}{ |c|c|c|c|c|c|c|c|c|c|c|c|c| }
\hline
\multicolumn{1}{|c|}{Approach} & \multicolumn{2}{|c|}{Representation} &\multicolumn{1}{c}{Accuracy~(\%)} & \multicolumn{1}{|c|}{Exp Time~(s)} & \multicolumn{1}{|c|}{Section}\\
\hline
\multicolumn{1}{|c|}{} & Sets of &  Approach~I & 81 & 1728.60 & \ref{sec:IBL_results_spin_images} \\\cline{3-6}
\multicolumn{1}{|c|}{ } & Spin-Images &  Approach~II & 87 & 1757.20 &  \ref{sec:IBL_results_spin_images}\\\cline{2-6}
\multicolumn{1}{|c|}{Instance-based} & \multicolumn{2}{|c|}{BoW}  & 89 & 195.60 & \ref{BOW_results_chapter_5}  \\\cline{2-6}
\multicolumn{1}{|c|}{} & \multicolumn{2}{|c|}{Local LDA}  & 91 & 348.50 & \ref{topic_layer_instance_based_chapter5}\\\cline{2-6}
\multicolumn{1}{|c|}{} & \multicolumn{2}{|c|}{LDA}  & 88 & 227.50 & \ref{open_ended_LDA_instance_based_chapter5} \\\cline{2-6}
\multicolumn{1}{|c|}{} & \multicolumn{2}{|c|}{GOOD}  & \textbf{97} & 85.10 & \ref{GOOD_Representation_IBL} \\\cline{2-6}
\hline
{}& \multicolumn{2}{|c|}{BoW}  & 89 & 187.10 & \ref{BoW_NB_chapter6}\\\cline{2-6}
Model-based& \multicolumn{2}{|c|}{Local LDA} & 93 & 293.80 & \ref{naive_bayes_topic}\\\cline{2-6}
\multicolumn{1}{|c|}{} & \multicolumn{2}{|c|}{LDA}  & 90 & 198.30 & \ref{naive_bayes_open_ended_topic} \\\cline{2-6}

{}&  \multicolumn{2}{|c|}{GOOD} & 95 & \textbf{83.50} & \ref{GOOD_Representation_NB} \\
\hline
\end{tabular}
}
\vspace{-4mm}
\end{center}
\end{table*}

Although model-based object category learning and recognition with BoW and LDA representations are the two most compact approaches in this evaluation, their computation time and descriptiveness are not as good as the GOOD representation. Overall, instance-based learning with GOOD achieves the best recognition performance, which is 2 percentage points (p.p.) better than GOOD with model-based learning, 4 p.p. better than Local LDA and 7 p.p. and 8 p.p. better than LDA and BoW representations respectively. The BoW and Local LDA led to experiment times 2.2 to 3.5 times higher than the experiment time obtained with GOOD. The underlying reason is that GOOD works directly on 3D point cloud and requires neither computation of local features nor a sampling procedure. According to the evaluations, this approach is competent for robotic applications with strict limits on the computation time requirement. A summary of all evaluations is reported in Table~\ref{table:summary_chapter6}. In the next section, we provide a detailed analysis of the GOOD descriptor.

%%%%%%%%%%%%%%%%%%%%%%%%%%%%%%%%%%%%%%%%%%%%%%%%%%%%%%%%%%%%%%%%%%%%%%%%%%%%%%%%%%%%
%%%%%%%%%%%%%%%%%%%%%%%%%%%%%%%%%%%%%%%%%%%%%%%%%%%%%%%%%%%%%%%%%%%%%%%%%%%%%%%%%%%%
\section {GOOD Descriptor}
\label{sec:good_descriptor_evals}
Several additional experiments were carried out to evaluate the performance of the proposed GOOD descriptor concerning {\emph{descriptiveness}, {\emph{scalability}}, {\emph{robustness}} and {\emph{efficiency}}. In these experiments, we mainly use the Washington RGB-D Object Dataset~\citep{lai2011large}, one of the largest publicly available datasets for object recognition. For some experiments, the Restaurant Object Dataset \citep{KasaeiInteractive2015} is also used (for datasets, see Section~\ref{datasets}). 

In all experiments, the instance-based learning approach is used (see section \ref{fixed_size_representations}). GOOD was compared with four state-of-the-art object descriptors that are available in the Point-Cloud Library\footnote{http://pointclouds.org/}, namely VFH \citep{rusu2010fast}, ESF \citep{wohlkinger2011ensemble}, GFPFH \citep{rusu2009detecting}, from PCL 1.7, and GRSD \citep{marton2010hierarchical}, from PCL 1.8. For all selected descriptors, the default parameters in the respective PCL implementations were used~\citep{aldoma2012}.

\begin{table}[!b]
\begin{center}
\caption {Summary of GOOD descriptiveness experiments.}
\resizebox{0.75\columnwidth}{!}{
\tiny{
\begin{tabular}{|c|c|c|c|c|c|c|c|c|c|c|c|c|c| }
\hline
\textbf{Number of Bins} & \textbf{Descriptor size} & \textbf{Memory} (Kb) &  \textbf{Accuracy~(\%)} \\
\hline
 5   & 75   & 0.3  &  92 \\
\hline
10  & 300  & 1.2  &  93 \\
\hline
 15  & 675  & 2.7  &  \textbf{94} \\
\hline
 20  & 1200 & 4.8  &  93 \\
\hline
 25  & 1875 & 7.5  &  \textbf{94} \\
\hline
 30  & 2700 & 10.8 &  93 \\
\hline
 35  & 3675 & 14.7 &  93 \\
\hline
 40  & 4800 & 19.2 &  93 \\
\hline
 45  & 6075 & 24.3 &  93 \\
\hline
 50 & 7500 & 30.0 &  93 \\
\hline
\end{tabular}}}
\label{table:effect_of_number_of_bins_parameter}
\end{center}
\vspace{-5mm}
\end{table}

\subsection {Descriptiveness}
\label{descriptiveness}

As mentioned above, GOOD has a parameter called \emph{number of bins} that has effect on descriptiveness, efficiency and robustness. Therefore, it must be well selected to provide a good balance between recognition performance, memory usage and computation time. The descriptiveness of the proposed descriptor with respect to varying \emph{number of bins} was evaluated using the Washington RGB-D Object Dataset. For each value of the number of bins, 10-fold cross validation experiments were performed.  

Results are presented in Fig.~\ref{fig:scalability} (\emph{left}) and Table~\ref{table:effect_of_number_of_bins_parameter}. In these experiments, the configurations that obtained the best accuracy figures were 15 and 25 bins.
Although, a large number of bins provides more details about the point distribution, it increases computation time, memory usage and sensitivity to noise. Therefore, since the difference to other configurations is not very large, we prefer to use the first configuration, i.e., $5$ bins which displays a good balance between recognition performance, memory usage, and processing speed. The accuracy of the proposed system with this configuration was 92 percent. It shows that the overall performance of the recognition system is distinctive. Unless otherwise noted, the remaining results are computed using this configuration.

%%%%%%%%%%%%%%%%%%%%%%%%%%%%%%%%%%%%%%%%%%%%%%%%%%%%%%%%%%%%%%%%%%%%%%%%%%%%%%%%%
\subsection {Scalability}

A set of experiments was carried out to evaluate the performance of the proposed descriptor on the Washington RGB-D Object Dataset, concerning its scalability with respect to varying number of categories. 
Results are depicted in Fig.~\ref{fig:scalability} (\emph{center}) and (\emph{right}). One important observation is that the accuracy decreases in all approaches as more categories are covered (Fig.~\ref{fig:scalability}(\emph{center})). This is expected since a higher number of categories tends to make the classification task more difficult. Moreover, it can be concluded from Table~\ref{table:scalability} that when the number of object categories increases (i.e., more than 35 categories), VFH and GOOD (15 bins) descriptors achieve the best accuracy and stable performance regarding varying numbers of categories.

\begin{table}[!b]
\begin{center}
    \caption {Summary of scalability experiments.}
\resizebox{0.7\columnwidth}{!}{
\begin{tabular}{|c|c|c|c|c|c|c|}
\hline
 \multirow{2}{*} {\textbf{Number of categories}} & \multicolumn{6}{|c|}{\textbf{Accuracy~(\%)}} \\
\cline{2-7}
 & GOOD(5bins) & GOOD(15bins) & VFH & ESF & GFPFH & GRSD \\
\hline
5  & 96 & \textbf{98} & \textbf{98} & 97 & 90 & 97\\
\hline
10 & 94 & 95 & \textbf{96} & \textbf{96} & 71 & 92\\
\hline
15 & 95 & 95 & 97 & \textbf{98} & 75 & 80\\
\hline
20 & 95 & 95 & 97 & \textbf{96} & 71 & 81\\
\hline
25 & 95 & 95 & \textbf{97} & \textbf{97} & 61 & 79\\
\hline
30 & 94 & 94 & \textbf{96} & 94 & 61 & 77\\
\hline
35 & 94 & 94 & 94 & \textbf{96} & 59 & 76\\
\hline
40 & 93 & \textbf{94} & \textbf{94} & 93 & 59 & 72\\
\hline
45 & 92 & \textbf{94} & \textbf{94} & 93 & 60 & 69\\
\hline
50 & 92 & \textbf{94} & \textbf{94} & 93 & 59 & 68\\
\hline
\end{tabular}}
\label{table:scalability}
\end{center}
\end{table}

It is clear from Fig.~\ref{fig:scalability} (\emph{right}) that the experiment time of our approach is significantly smaller than VFH, GRSD and GFPFH. Although GOOD, VFH and ESF descriptors obtain an acceptable scalability regarding varying numbers of categories, the scalability of GRSD and GFPFH is very low and their performance drops aggressively when the number of categories increases. Although EFS descriptor achieves better performance than our approach with 5 bins (i.e., GOOD 5bins), the length of EFS (i.e., an inverse indicator of compactness) is around 8.5 times more than our descriptor (see Table~\ref{table:memory_footprint}). For a number of known categories greater than 35, the difference in accuracy between ESF and GOOD with 5 bins, is equal or less than 1\%, whereas GOOD 15 bins is clearly better than the others.

\begin{figure*}[!t]
\centering
\begin{centering}
\begin{tabular}{ccc}
\hspace{-8mm}
	 \includegraphics[width=0.35\linewidth, trim= 0.cm 0cm 0cm 0.0cm,clip=true]{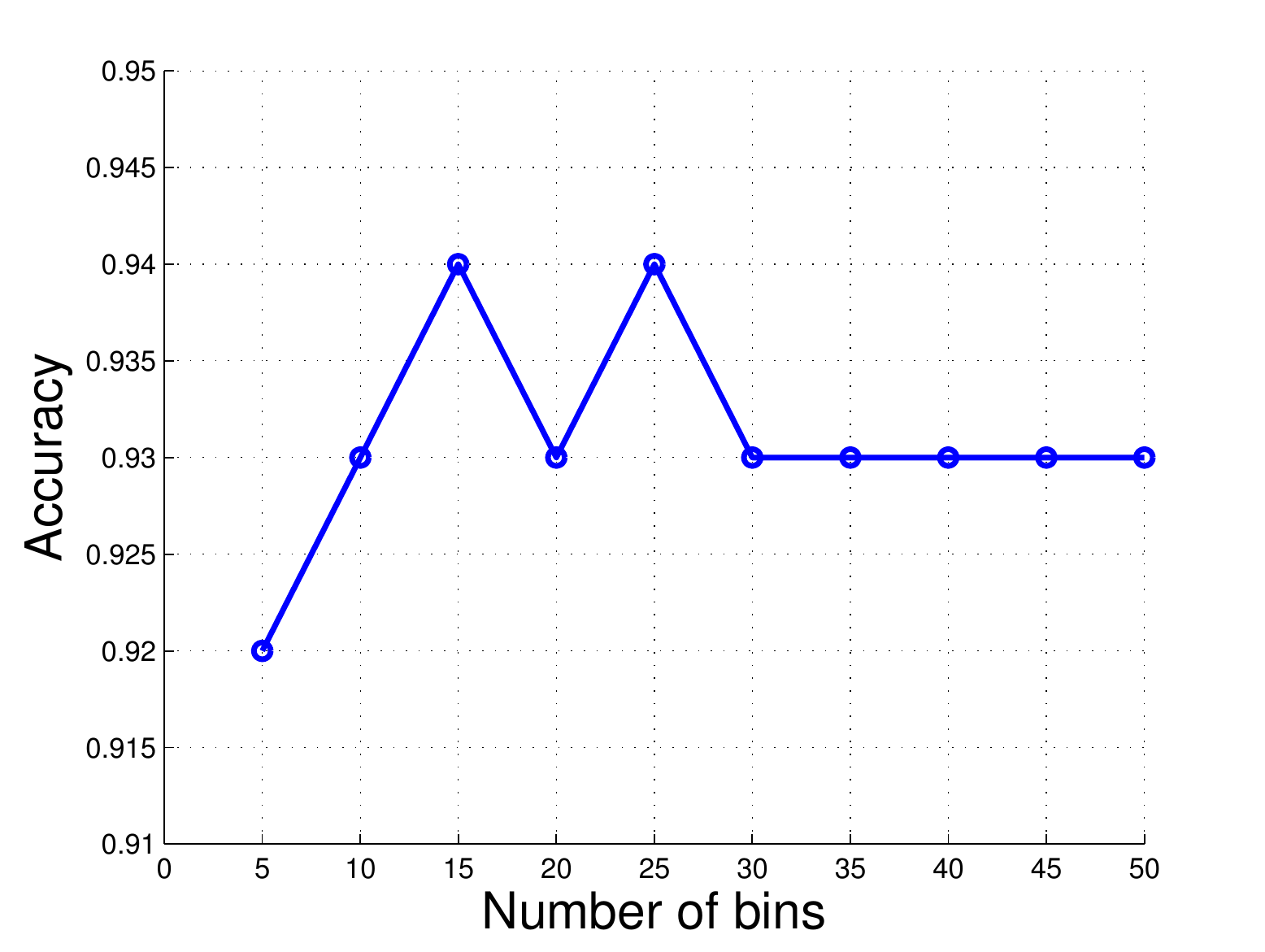} &
\hspace{-5mm}
	 \includegraphics[width=0.35\linewidth, trim= 0.cm 0cm 0cm 0.0cm,clip=true]{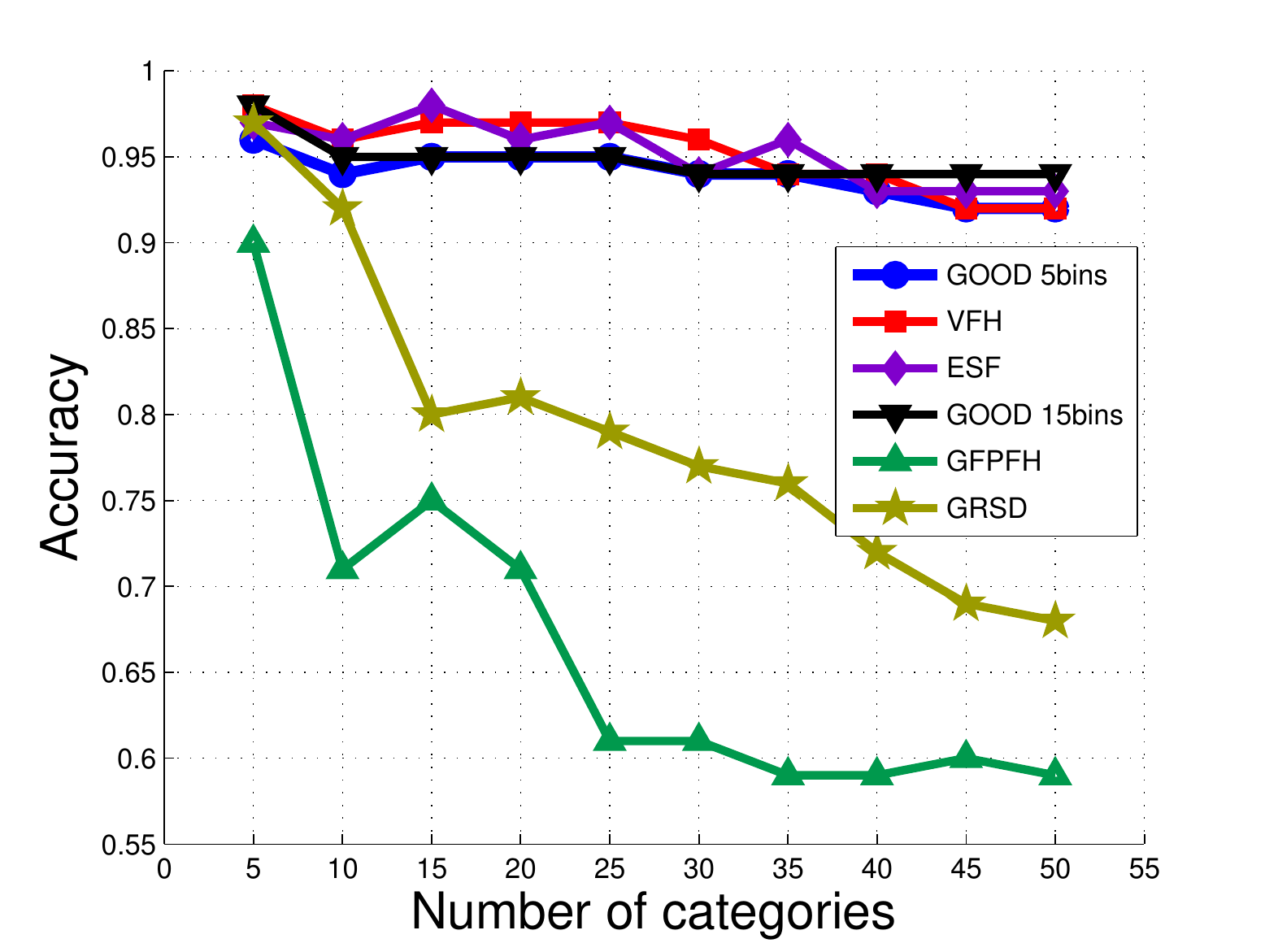}&
	\hspace{-5mm}
	\includegraphics[width=0.35\linewidth]{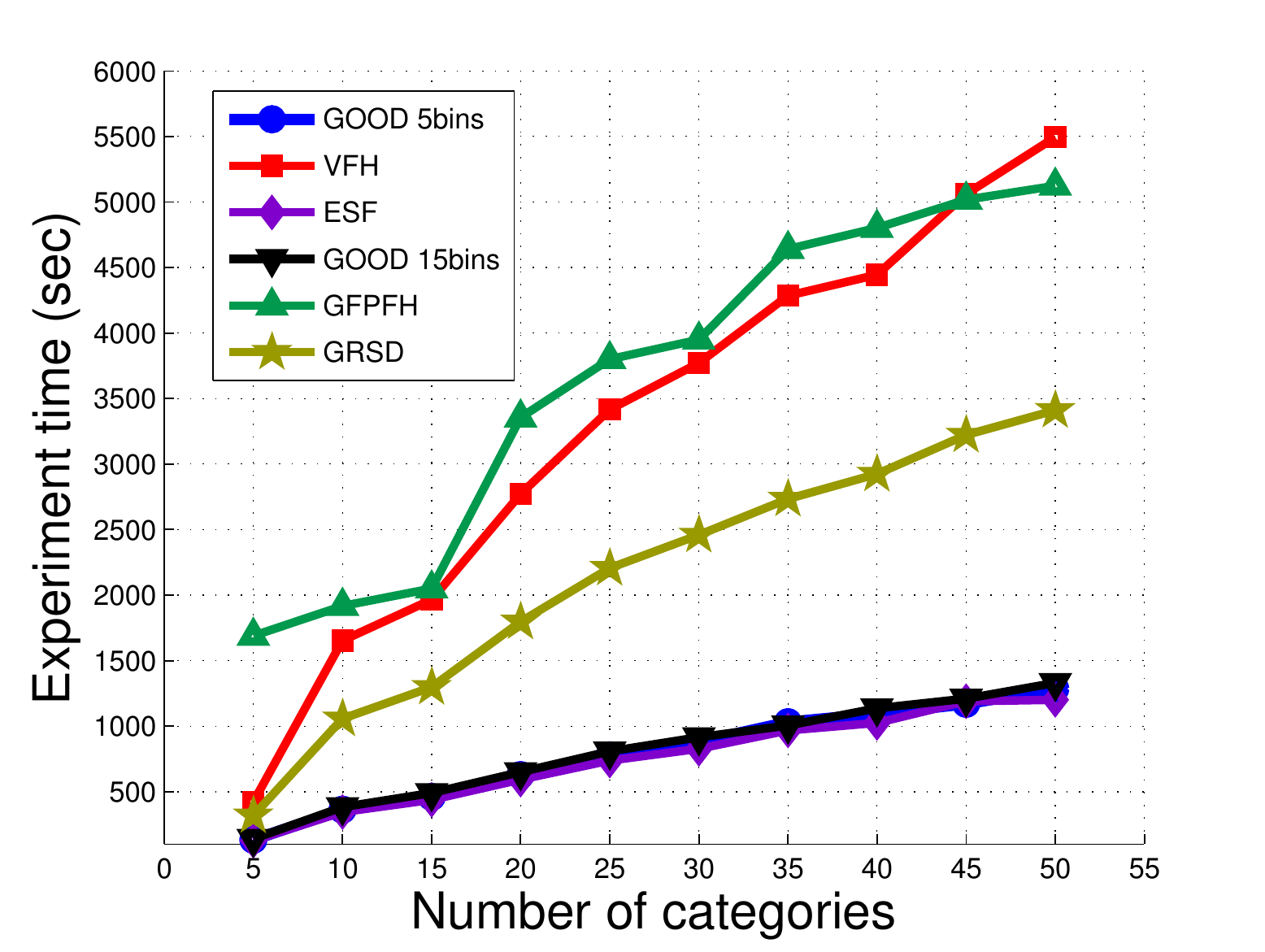}	
	\vspace{-1mm}
\end{tabular}
\end{centering}
\caption{
Object recognition performance in descriptiveness and scalability experiments; (\emph{left}) effect of number of bins on accuracy; (\emph{center}) scalability of the selected descriptors with respect to varying numbers of categories in the dataset as a function of accuracy vs. Number of categories; (\emph{right}) scalability experiment time vs. Number of categories.}
\vspace{-2mm}
\label{fig:scalability}
\end{figure*}
%%%%%%%%%%%%%%%%%%%%%%%%%%%%%%%%%%%%%%%%%%%%%%%%%%%%%%%%%%%%%%%%%%%%%%%%%%%%%%%%%%%%%
\subsection {Robustness}
The robustness of the proposed object descriptor with respect to different levels of Gaussian noise and varying point cloud resolutions was evaluated and compared with other global object descriptors. These experiments were run on the mentioned Restaurant Object Dataset.  This dataset is suitable for such evaluations since it is a small dataset and the objects are extracted from cluttered scenes. Furthermore, the Restaurant Object Dataset contains some occluded or truncated objects, which improves the generalization power of the relevant learnt models.

\begin{figure}[!b]
\vspace{-0mm}
\centering
\begin{centering}
\begin{tabular}{cccc}
	 \includegraphics[width=0.16\linewidth, trim= 0.cm 0cm 0cm 0.0cm,clip=true]{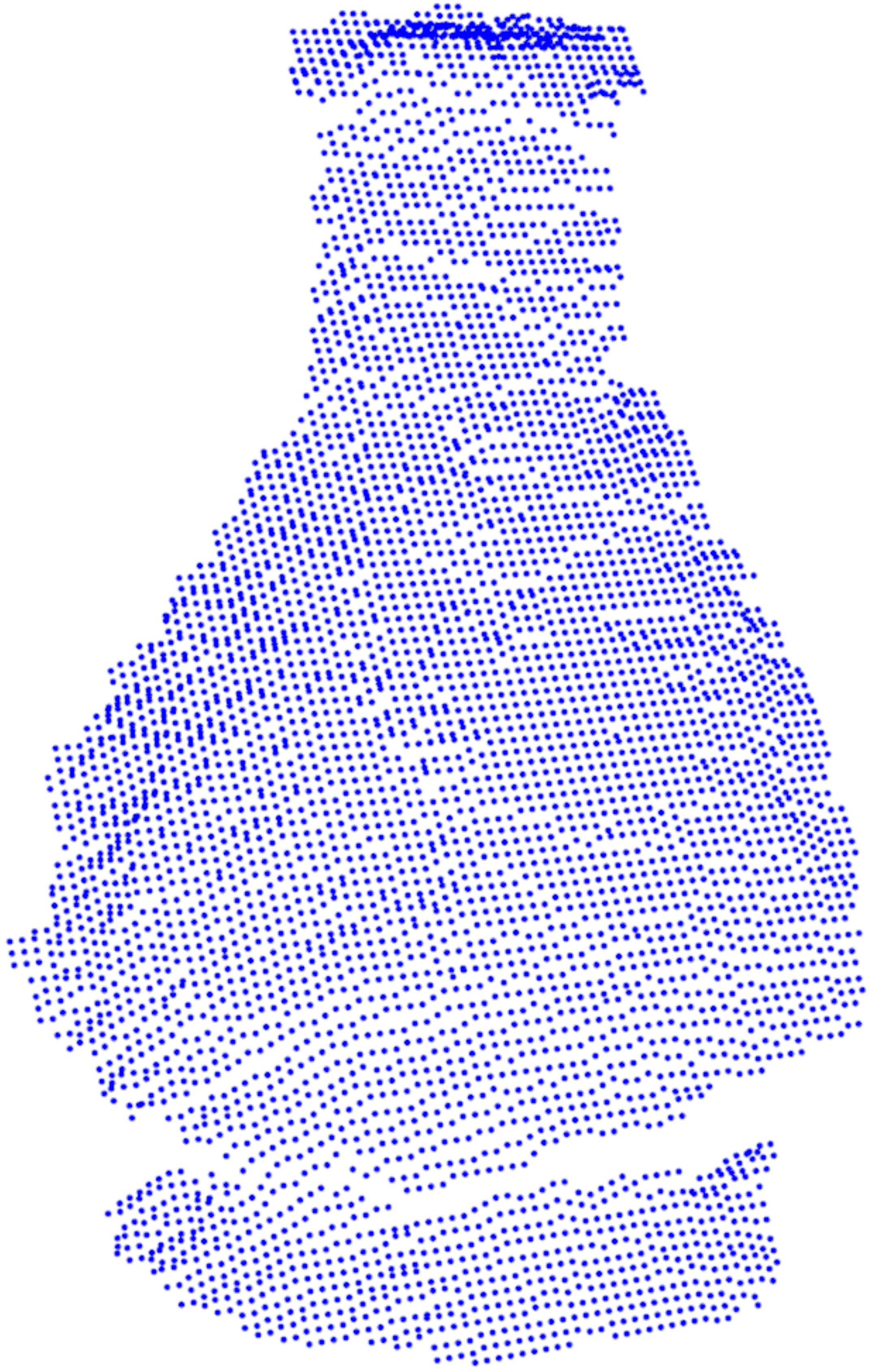}&\quad
	 \includegraphics[width=0.16\linewidth, trim= 0.cm 0cm 0cm 0.0cm,clip=true]{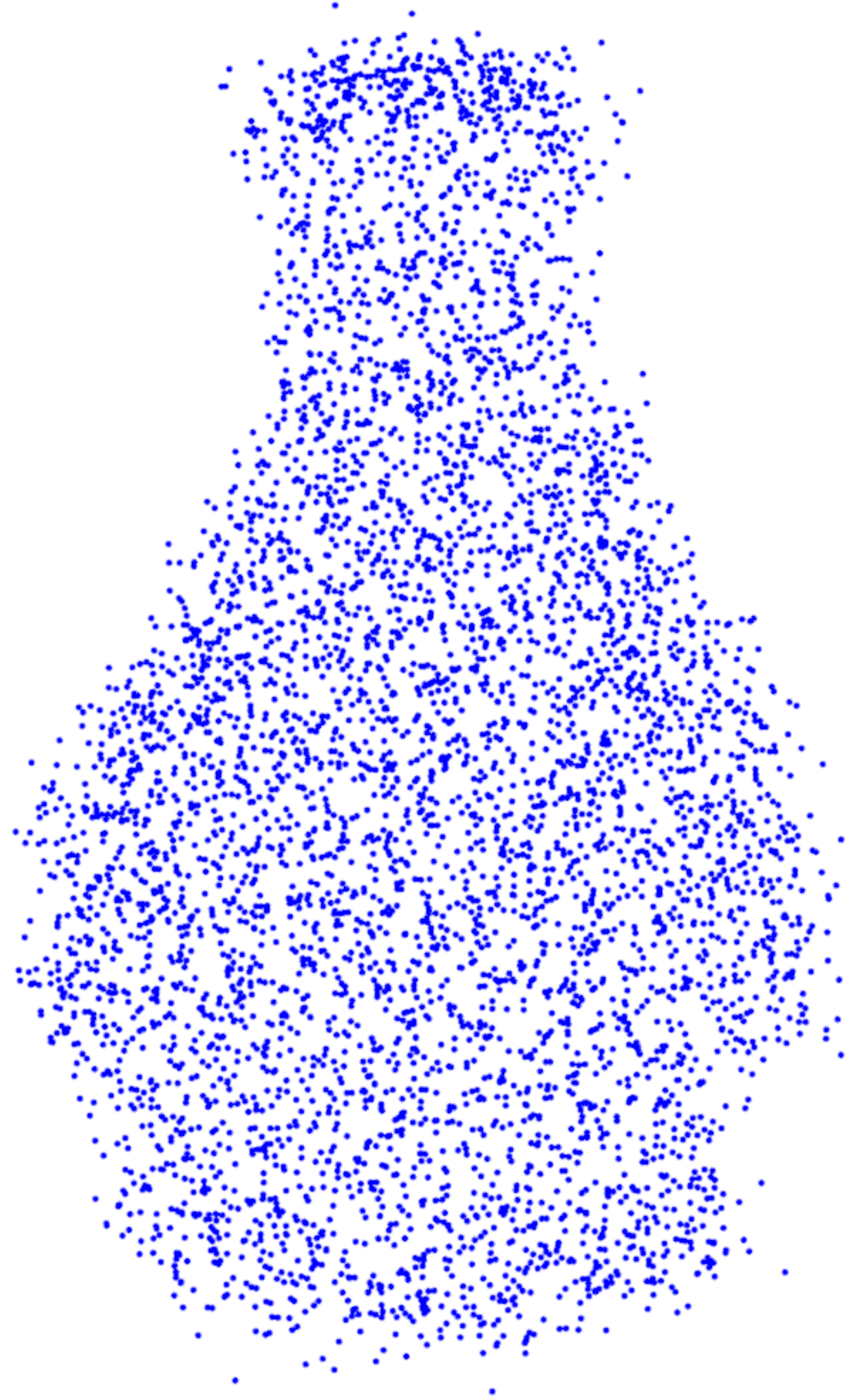}&\quad
	 \includegraphics[width=0.183\linewidth, trim= 0.cm 0cm 0cm 0.0cm,clip=true]{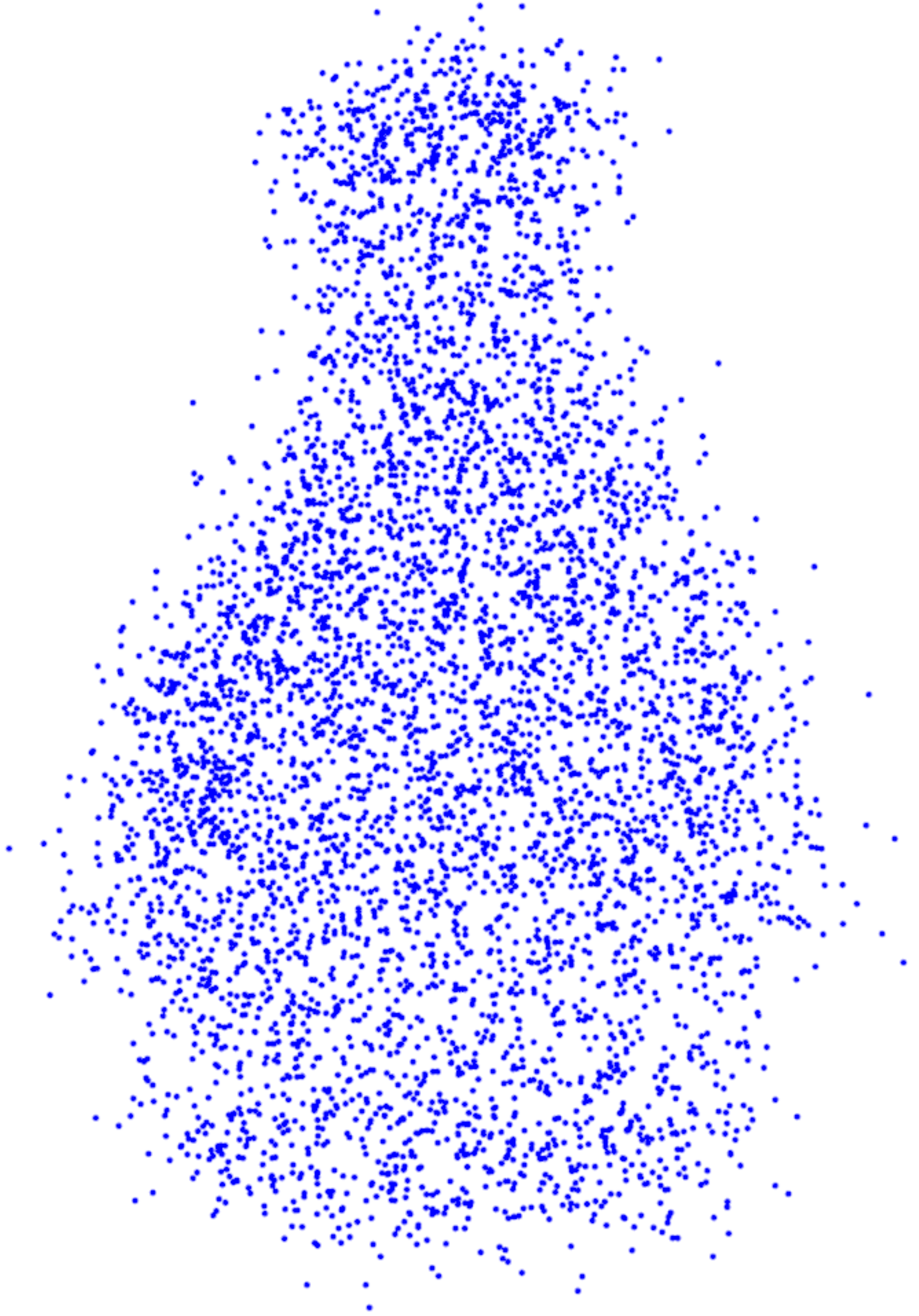}&\quad
 	 \includegraphics[width=0.183\linewidth, trim= 0.cm 0cm 0cm 0.0cm,clip=true]{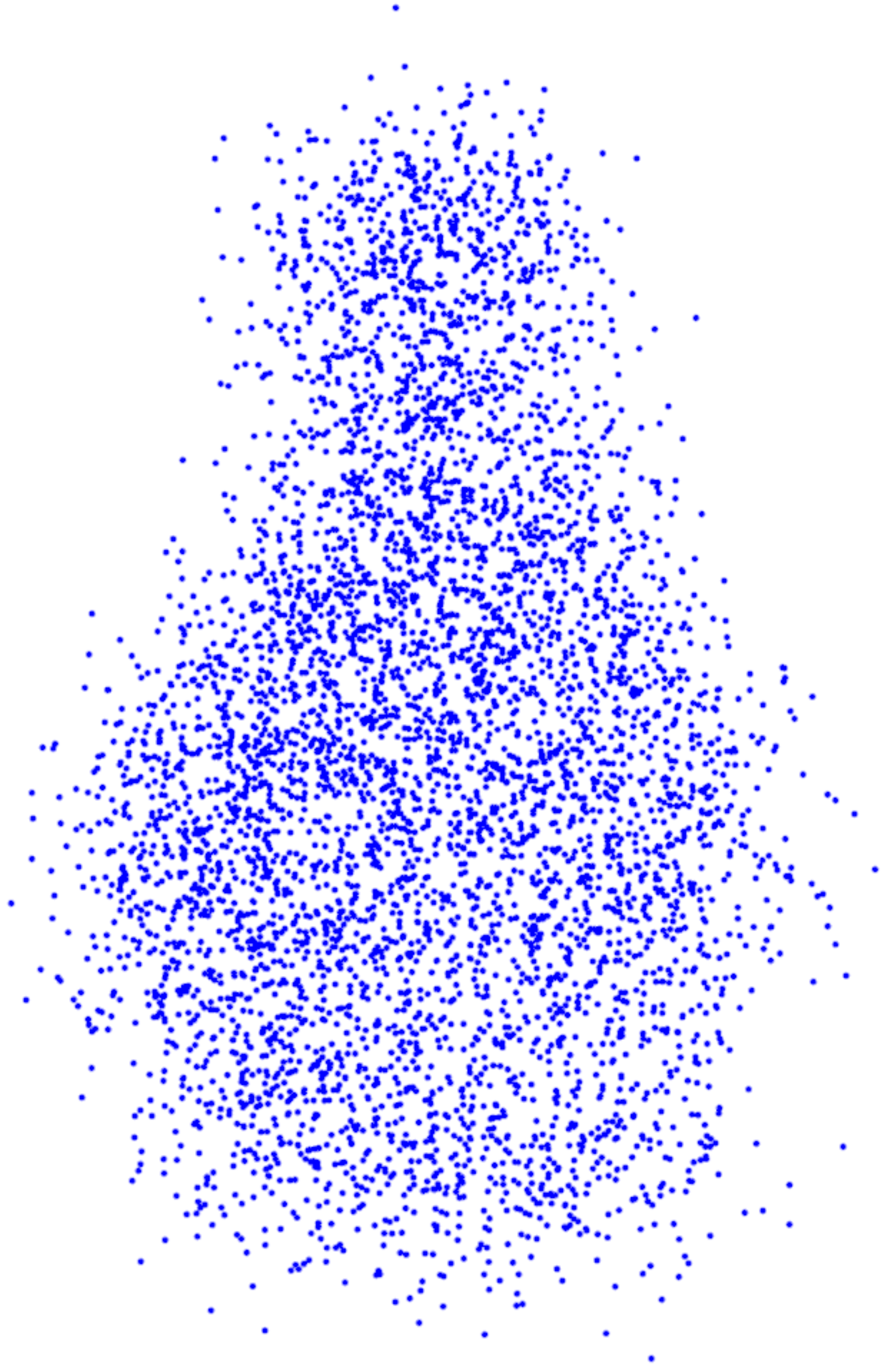}\\
 	 \emph{Original} & \quad $\sigma$ \emph{= 3mm} &\quad $\sigma$ \emph{= 6mm} & \quad $\sigma$ \emph{= 9mm}
\end{tabular}
\end{centering}
\caption{An illustration of a \emph{Vase} object with different levels of Gaussian noise.}
\label{fig:Gaussian_noise}
\end{figure}

%%%%%%%%%%%%%%%%%%%%%%%%%%%%%%%%%%%%%%%%%%%%%%%%%%%%
\subsubsection{Gaussian Noise} 
Ten levels of Gaussian noise with standard deviations from 1 to 10mm were added to the test data. For a given test object, Gaussian noise is independently added to the X, Y and Z-axes. As an example, a \emph{Vase} object with three levels of standard deviation of Gaussian noise ($\sigma = 3mm, \sigma = 6mm, \sigma = 9mm $) is depicted in Fig. \ref{fig:Gaussian_noise}. The robustness results under different levels of noise are presented in Table~\ref{table:Gnoise} and Fig.~\ref{fig:robustness_downsampling}(\emph{left}).
An important observation can be made from Fig.~\ref{fig:robustness_downsampling} and Fig.~\ref{fig:scalability}. Although, GOOD, ESF and VFH achieved a really good performance on noise free data, GOOD outperformed ESF, GFPFH and GRSD descriptors by a large margin under all levels of Gaussian noise. While the performance of VFH was similar to our approach under a low-level noise (i.e., $\sigma \le 6mm$), our shape descriptor outperformed all descriptors under high levels of noise. 

It can be concluded from this observation that GOOD is robust to noise due to using a stable, unique and unambiguous object reference frame. In contrast, since VFH and GFPFH rely on surface normals to calculate their shape descriptions, they are highly sensitive to noise. GRSD employs radial relationships to describe the geometry of points at each voxel cell and ESF uses distances and angles between randomly sampled points to generate a shape description; therefore, GRSD and ESF are also sensitive to noise and their performances decrease rapidly when the standard deviation of the Gaussian noise increases. In addition, GOOD uses three distribution matrices that are constructed based on orthographic projection, therefore less affected by noise (i.e., in each projection one dimension is discarded).

\begin{table}[!t]
\begin{center}
    \caption {Summary of robustness to Gaussian noise experiments.}
    \vspace{2mm}
\resizebox{0.7\columnwidth}{!}{
\begin{tabular}{|c|c|c|c|c|c|}
\hline
 \multirow{2}{*} {\textbf{Gaussian Noise}(mm)} & \multicolumn{5}{|c|}{\textbf{Accuracy~(\%)}} \\
\cline{2-6}
 & GOOD 5bins &  VFH & ESF & GFPFH & GRSD \\
\hline
1  &  94 & \textbf{95 }& 90 & 40 & 66\\
\hline
2 &  \textbf{93} & \textbf{93}  & 74 & 17 & 61\\
\hline
3 & 92 & \textbf{93} & 49 & 10 & 51\\
\hline
4 & 91 & \textbf{94} & 32 & 09 & 35\\
\hline
5 & 89 & \textbf{91} & 24 & 09 & 26\\
\hline
6 & \textbf{85} & \textbf{85} & 23 & 09 & 19\\
\hline
7 & \textbf{83} & 78 & 23 & 09 & 12\\
\hline
8 & \textbf{78 }& 57 & 22 & 09 & 08\\
\hline
9 & \textbf{69} & 42 & 23 & 09 & 10\\
\hline
10 & \textbf{67} & 36 & 22 & 09 & 09\\
\hline
\end{tabular}}
\label{table:Gnoise}
\end{center}
\vspace{-5mm}
\end{table}

\begin{figure*}[!b]
\centering
\begin{centering}
\begin{tabular}{ccc}
	\hspace{-8mm}
	\includegraphics[width=0.36\linewidth, trim= 0.cm 0cm 0cm 0.5cm,clip=true]{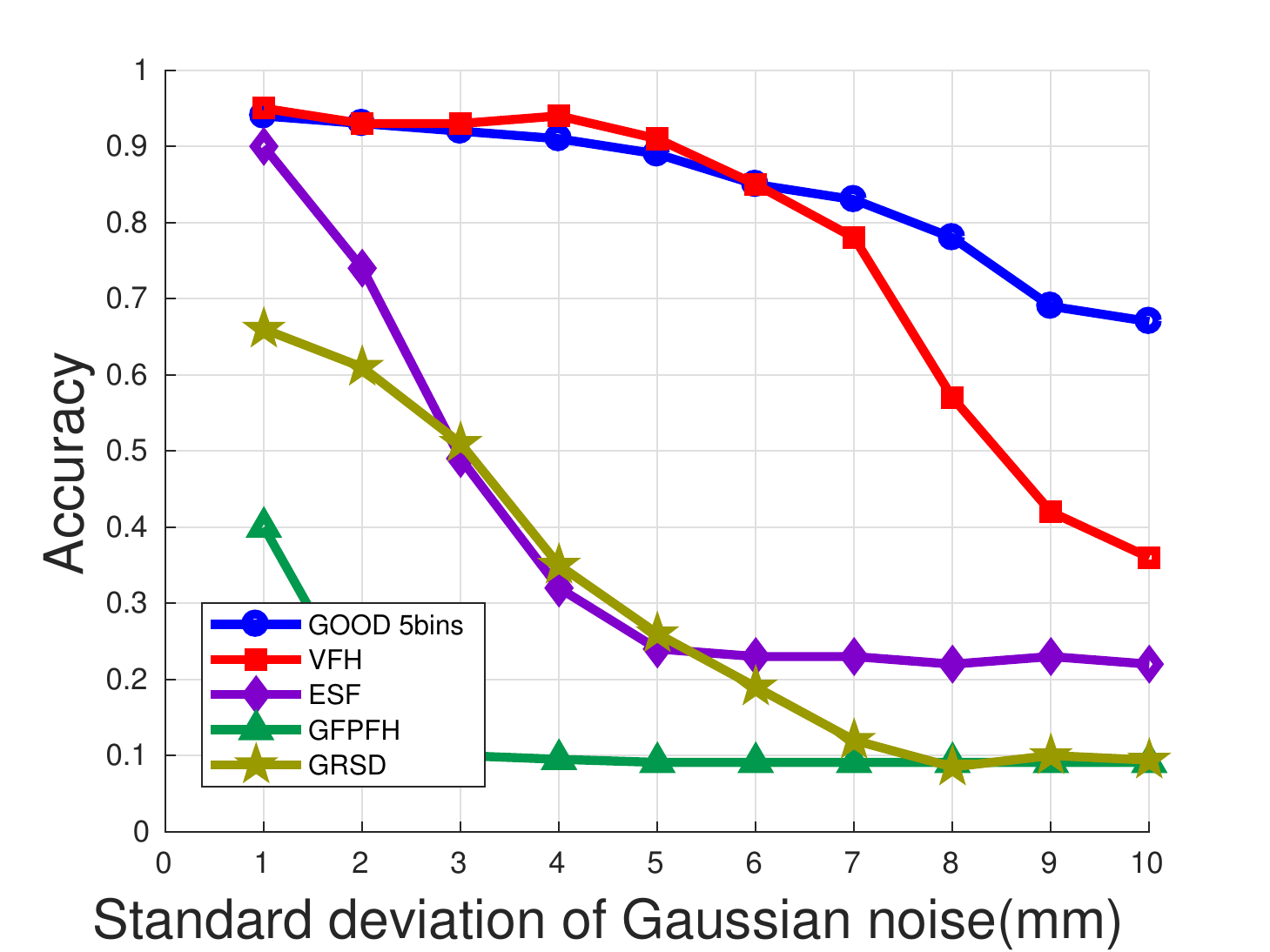}&
	\hspace{-8mm}
	\includegraphics[width=0.36\linewidth, trim= 0.cm 0cm 0cm 0.8cm,clip=true]{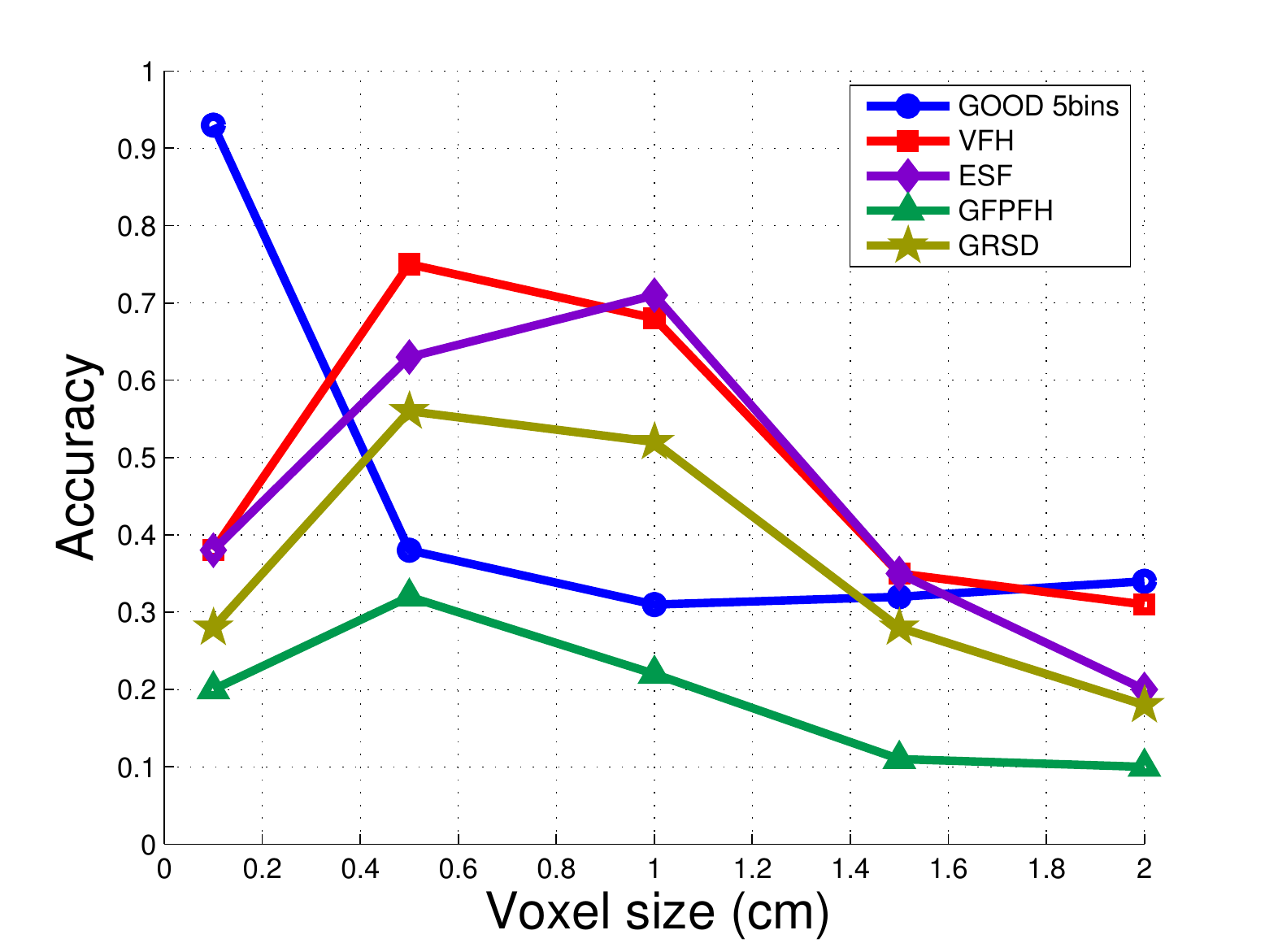}&
	\hspace{-8mm}
	\includegraphics[width=0.36\linewidth, trim= 0.cm 0cm 0cm 0.8cm,clip=true]{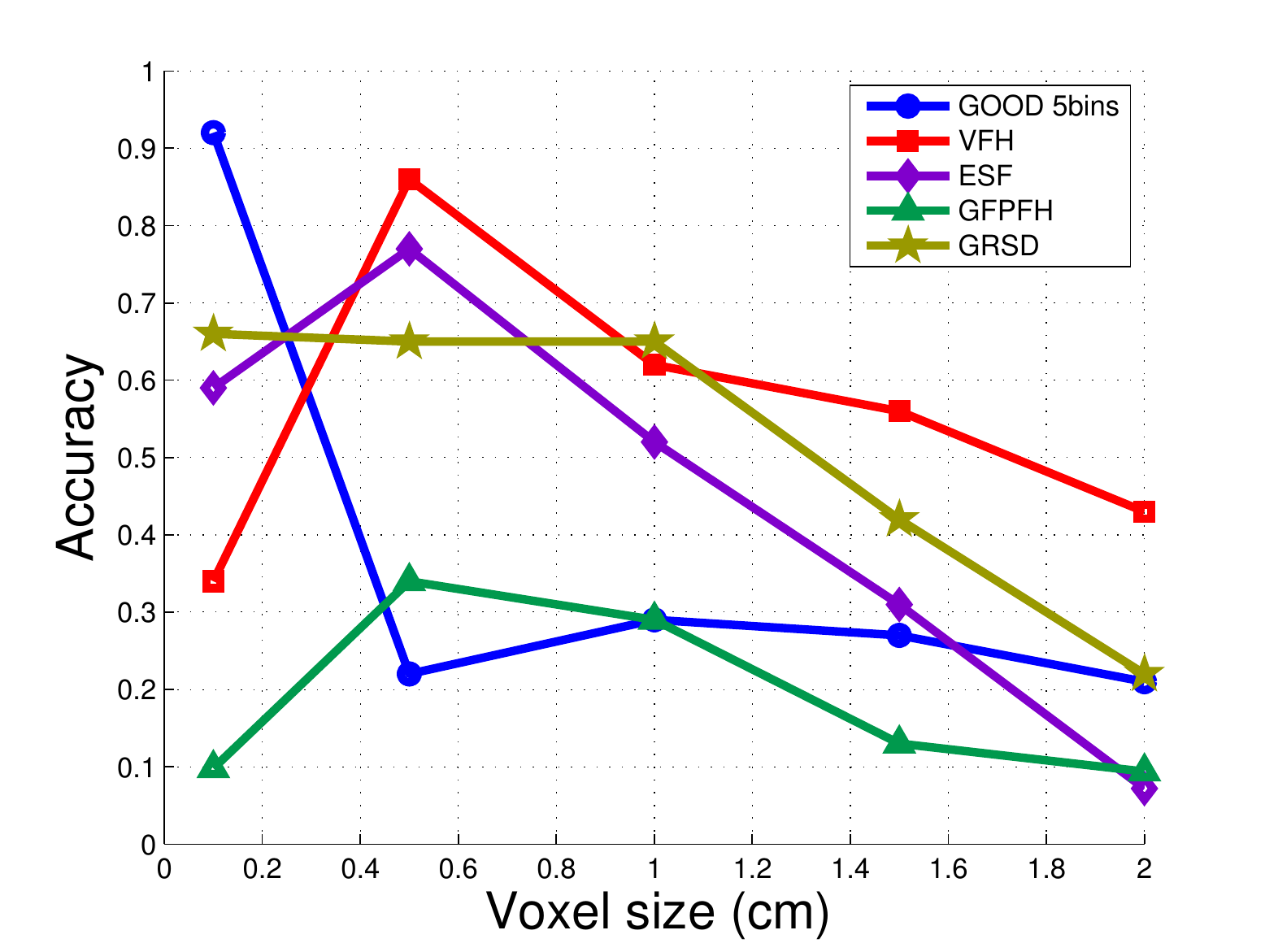}
\end{tabular}
\end{centering}
\caption{The robustness of the selected descriptors to different level of Gaussian noise and varying point cloud density: (\emph{left}) different levels of Gaussian noise applied to the test; (\emph{center}) different levels of downsampling applied to the test data; (\emph{right}) different levels of downsampling applied to the training data.}
\label{fig:robustness_downsampling}
\vspace{-0mm}
\end{figure*}
%%%%%%%%%%%%%%%%%%%%%%%%%%%%%%%%%%%%%%%%%%%%%%%%%%%%%%%%%%%%%%%%%%%%%%%%%%%%%%%%%%%%%%%%%%%
\subsubsection{Varying Point Cloud Density} 
Two sets of experiments were performed to examine the robustness of the proposed descriptor with respect to varying point cloud density. In the first set of experiments, the original density of training objects has been used and the point cloud density of test objects was reduced (downsampling) using a voxelized grid approach\footnote{http://pointclouds.org/documentation/tutorials/voxel\_grid.php}.
The methodology of this kind of downsampling commences with a voxelization of the surface points. This is initiated with a root volume element (voxel) and the eight children voxels in which each internal node has exactly eight children nodes. These are recursively subdivided until all voxels contain at most one point or the minimum voxel size is reached (i.e., the cloud is divided in multiple voxels with the desired resolution). Afterwards all the points that fall into the same voxel will be downsampled with their centroid. In this evaluation, each test object is downsampled using five different voxel sizes: $1, 5, 10, 15,$ and $20$ mm. An example of a \emph{Flask} object with four levels of downsampling is depicted in Fig.~\ref{fig:downsampling}. The robustness results regarding varying point cloud density in test data are presented in Fig.~\ref{fig:robustness_downsampling} (\emph{center}).

From experiments of reducing density of test data (i.e., Fig~\ref{fig:robustness_downsampling} (\emph{left})), it was found that our approach is more robust than the other descriptors concerning low-level downsampling (i.e., $\omega \le 3.5 mm$) and works slightly better than the other in high-level downsampling resolution (i.e., $\omega \ge 18 mm$). In contrast, the performance of VFH, ESF and GRSD were better than GOOD in mid-level downsampling resolution (i.e., $3.5 mm < \omega < 18 $). The performance of GFPFH was very low under all levels of point cloud resolution. Besides, it can be concluded from Fig.~\ref{fig:robustness_downsampling}~(\emph{right}) that when the level of down-sampling decreases, VFH, ESF and GRSD descriptors achieve better performance than GOOD and GFPFH descriptors.

\begin{figure}[!t]
\centering
\begin{centering}
\begin{tabular}{ccccc}
	 \includegraphics[width=0.16\linewidth, trim= 0.cm 0cm 0cm 0.0cm,clip=true]{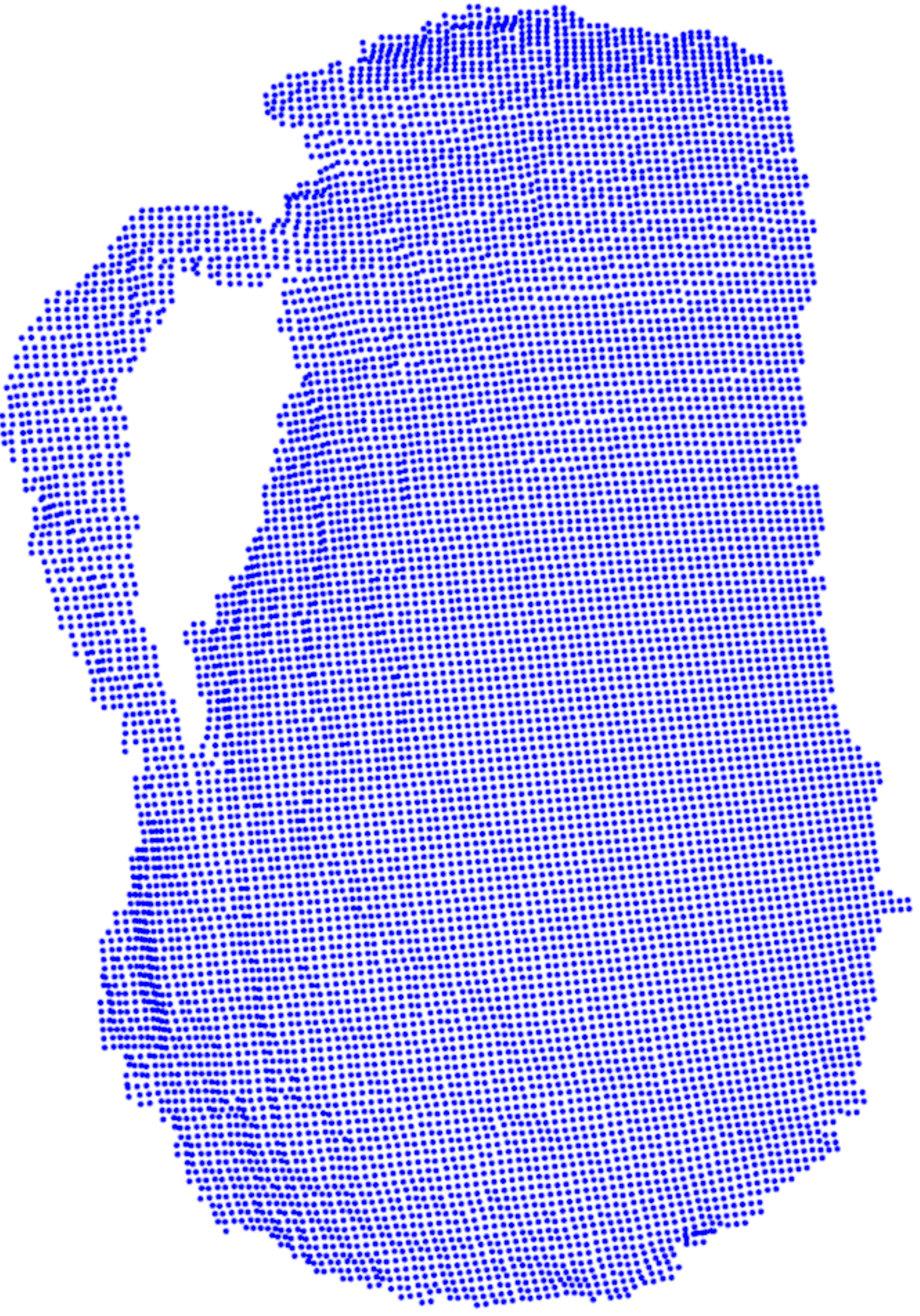}&\quad
	 \includegraphics[width=0.17\linewidth, trim= 0.cm 0cm 0cm 0.0cm,clip=true]{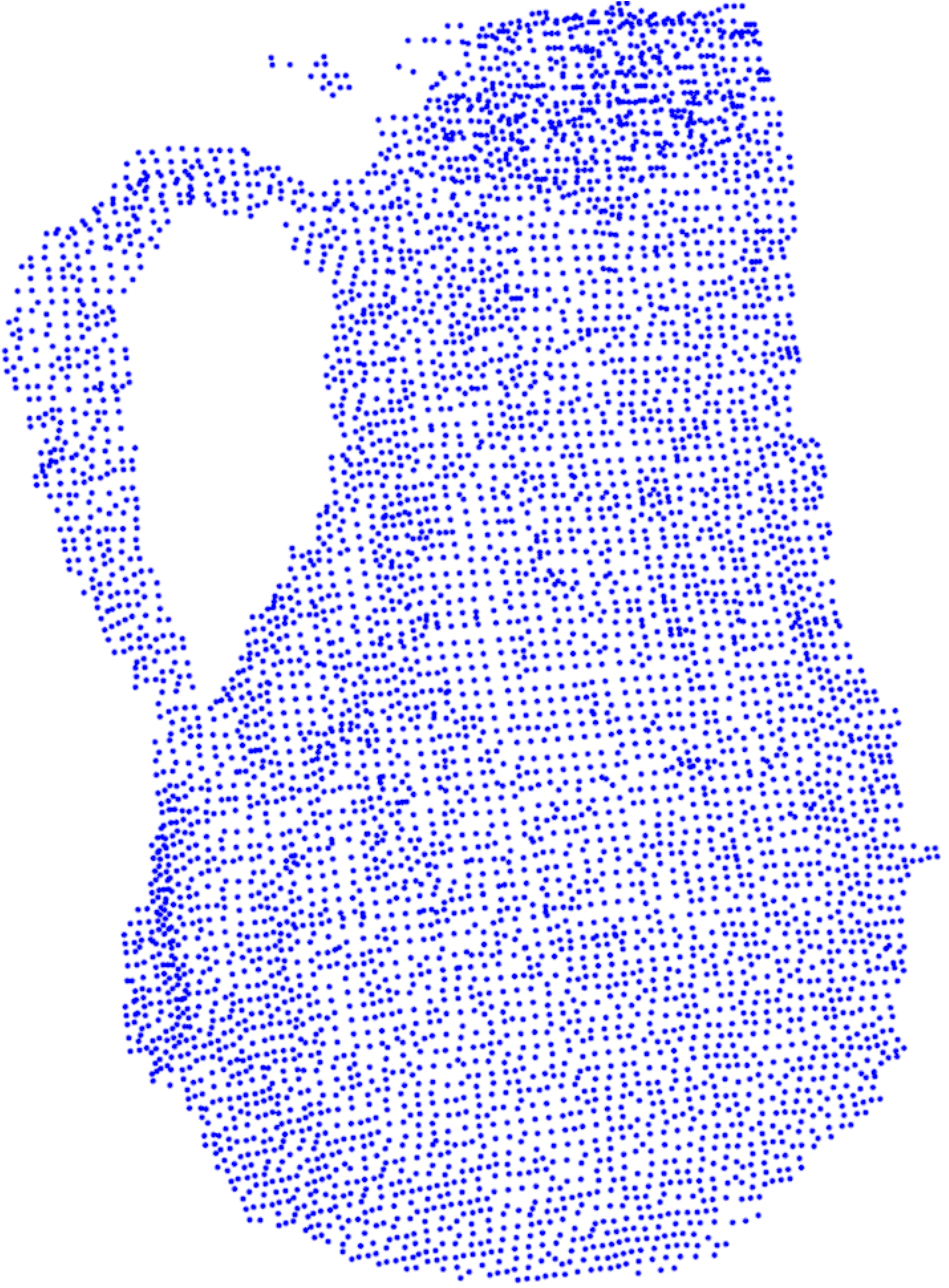}&\quad
	 \includegraphics[width=0.17\linewidth, trim= 0.cm 0cm 0cm 0.0cm,clip=true]{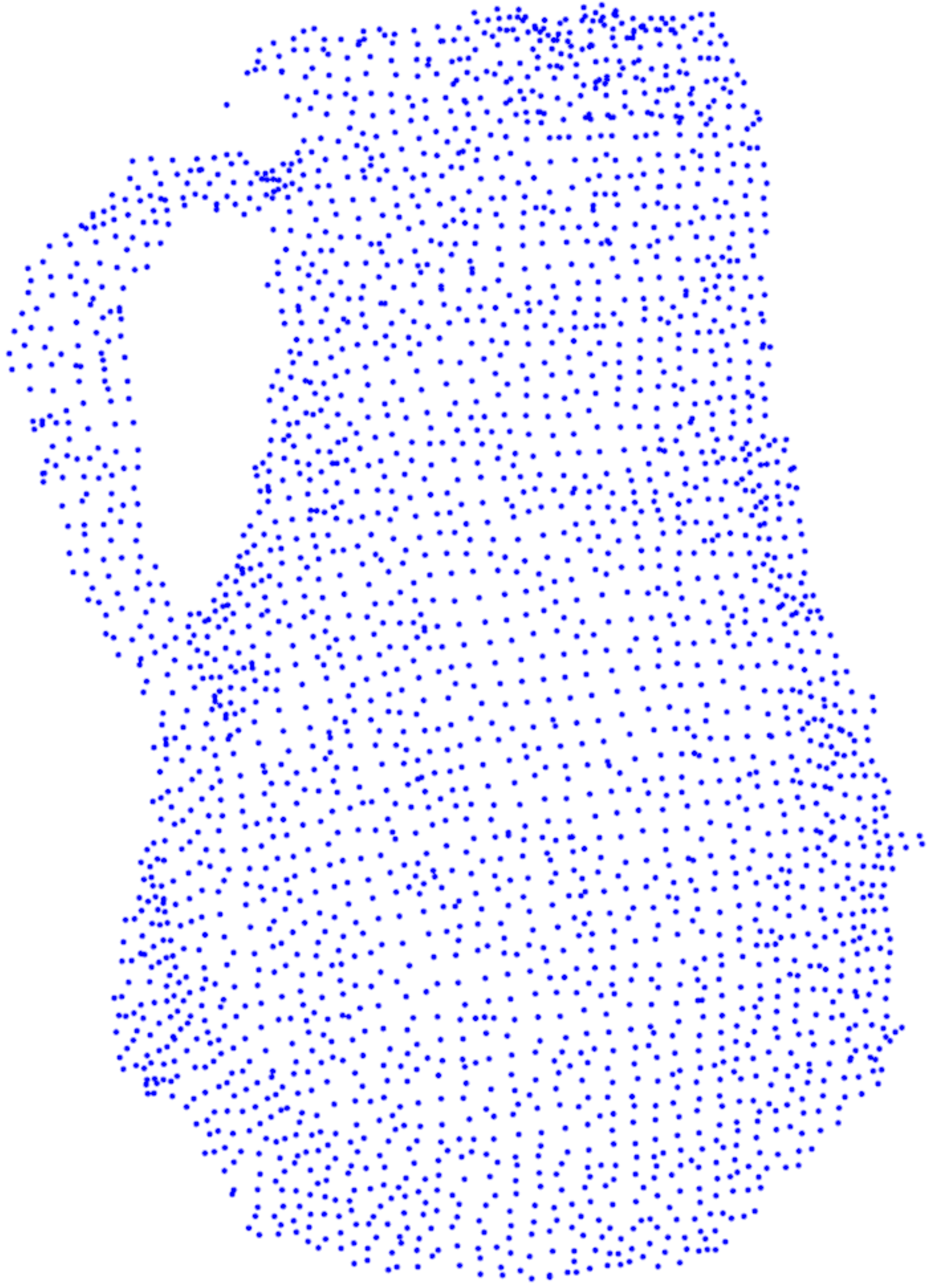}&\quad
	 \includegraphics[width=0.17\linewidth, trim= 0.cm 0cm 0cm 0.0cm,clip=true]{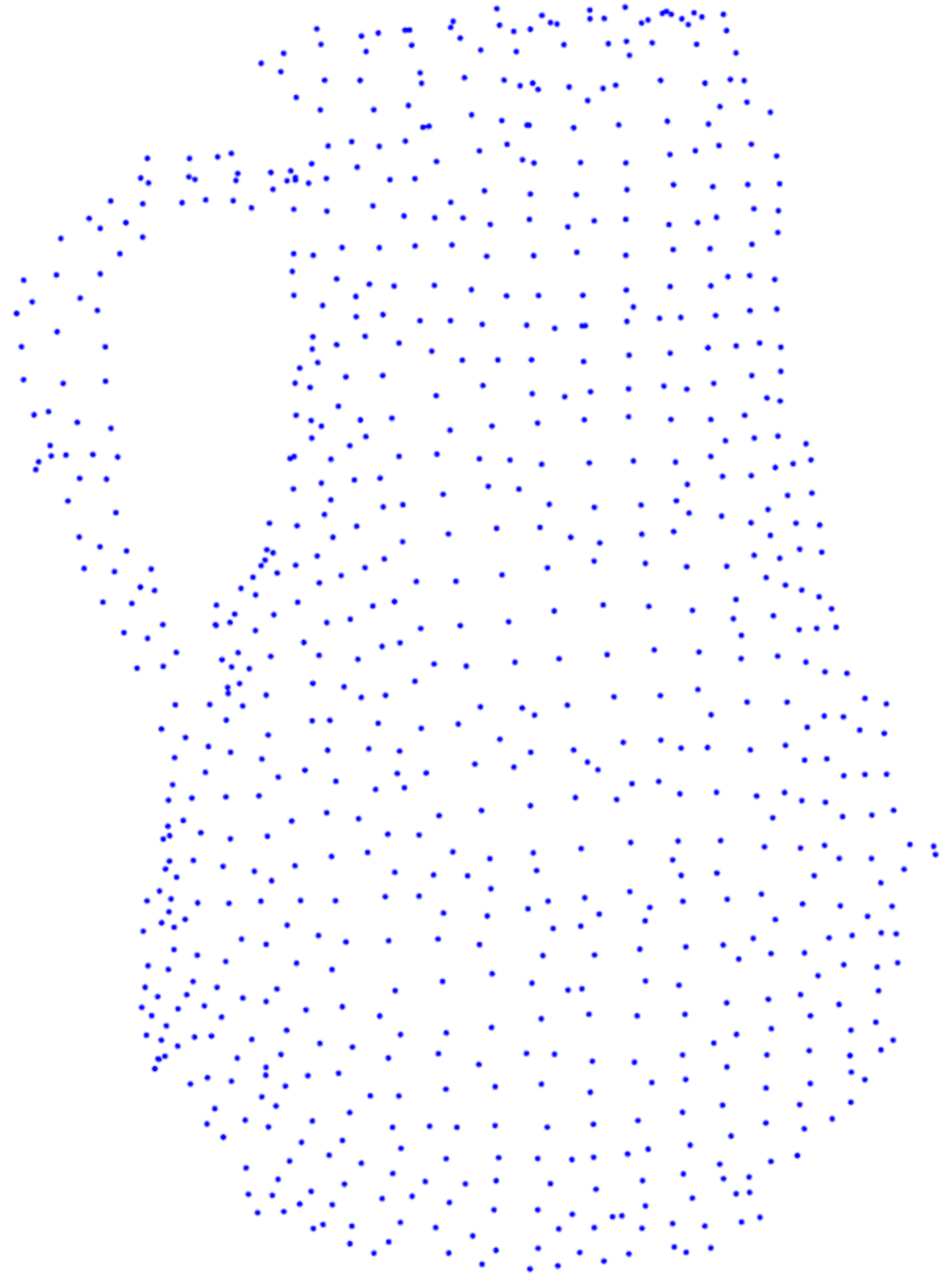}\\
 	 \emph{Original} &\quad \emph{DS = 1mm} &\quad \emph{DS = 5mm} &\quad \emph{DS = 10mm}
\end{tabular}
\end{centering}
\caption{An illustration of a \emph{Flask} object with different levels of downsampling.}
\label{fig:downsampling}
\end{figure}

%%%%%%%%%%%%%%%%%%%%%%%%%%%%%%%%%%%%%%%%%%%%%%%%%%%%%%%%%%%%%%%%%%%%%%%%%%%%%%%%
%%%%%%%%%%%%%%%%%%%%%%%%%%%%%%%%%%%%%%%%%%%%%%%%%%%%%%%%%%%%%%%%%%%%%%%%%%%%%%%%
\subsection {Efficiency}
In this subsection, two evaluations regarding memory footprint (i.e., the amount of main memory that a program uses or references while running) and computation time are provided and discussed.

%%%%%%%%%%%%%%%%%%%%%%%%%%%%%%%%%%%%%%%%%%%%%%%%%%%%%%%%%%%%%%%%%%%%%%%%%%%%%%%%
\begin{table}[!b]
\begin{center}
\caption {Length of selected 3D shape descriptors.}
\resizebox{0.8\columnwidth}{!}{
\tiny{
\begin{tabular}{|c|c|c|c|c| }
\hline
 \textbf{Descriptor} & \textbf{Feature length} (float) & \textbf{Adjustable length} & \textbf{Implementation}\\
\hline
GFPFH & 16 & No &  PCL 1.7 \\
\hline
GRSD & 21  & No &  PCL 1.8\\
\hline
GOOD & 75 & Yes & PCL 1.9 \\
\hline
 VFH & 308 & No &  PCL 1.7\\
\hline
 ESF & 640 & No &  PCL 1.7\\
\hline
\end{tabular}}}
\label{table:memory_footprint}
\end{center}
\end{table}
\subsubsection{Memory Footprint}
\label {memory_footprint}
The length of a descriptor has influence on memory usage and computation time in object recognition (see Fig.~\ref{fig:scalability}). The length of all descriptors used in this evaluation is given in Table~\ref{table:memory_footprint}. Although GFPFH and GRSD are the two most compact shape descriptors (see Table~\ref{table:memory_footprint}), their computation time and descriptive power are not good as depicted in figures~\ref{fig:scalability} and \ref{fig:computationTime}. Our approach is the third most compact descriptor and provides good balance between computation time and descriptiveness with 75 floats. Although, VFH and ESF descriptors achieve a good description power, their lengths are around 4.10 and 8.50 times larger than GOOD with 5 bins and 19 and 40 times larger than GFPFH respectively. ESF is the less compact descriptor compared to all the other descriptors.

%%%%%%%%%%%%%%%%%%%%%%%%%%%%%%%%%%%%%%%%%%%%%%%%%%%%%%%%%%%%%%%%%%%%%%%%%%%%%%%%
\subsubsection{Computation Time}
\label{computation_time}

Several experiments were performed to measure computation time for all descriptors used in this evaluation. Since the number of object's points directly affects the computation time, first, we randomly select 20 objects from the Washington RGB-D dataset. We then calculate the average time required to generate a description for the 20 selected objects. Figure~\ref{fig:computationTime} compares the average computation time of the selected object descriptors. Several observations can be made. GOOD is the most time efficient descriptor. In contrast, GFPFH is the most computationally expensive descriptor.  ESF, VFH and GRSD achieve a medium performance in terms of computation time. GOOD is around 10 times faster than ESF and 44, 50 and 254 times faster than VFH, GRSD and GFPFH descriptors. The underlying reason is that GOOD works directly on 3D point clouds and requires neither triangulation of the object's points nor surface meshing.
According to this evaluation and the memory footprint evaluation (i.e., subsection~\ref{memory_footprint}), our approach is especially well suited for robotic applications with strict limits on the memory footprint and computation time  requirements.
\begin{figure}[!h]
\center
    \begin{tabular}[t]{cc}  
		 \includegraphics[width=0.39\linewidth, trim= 0.cm 0cm 0cm 0.0cm,clip=true]{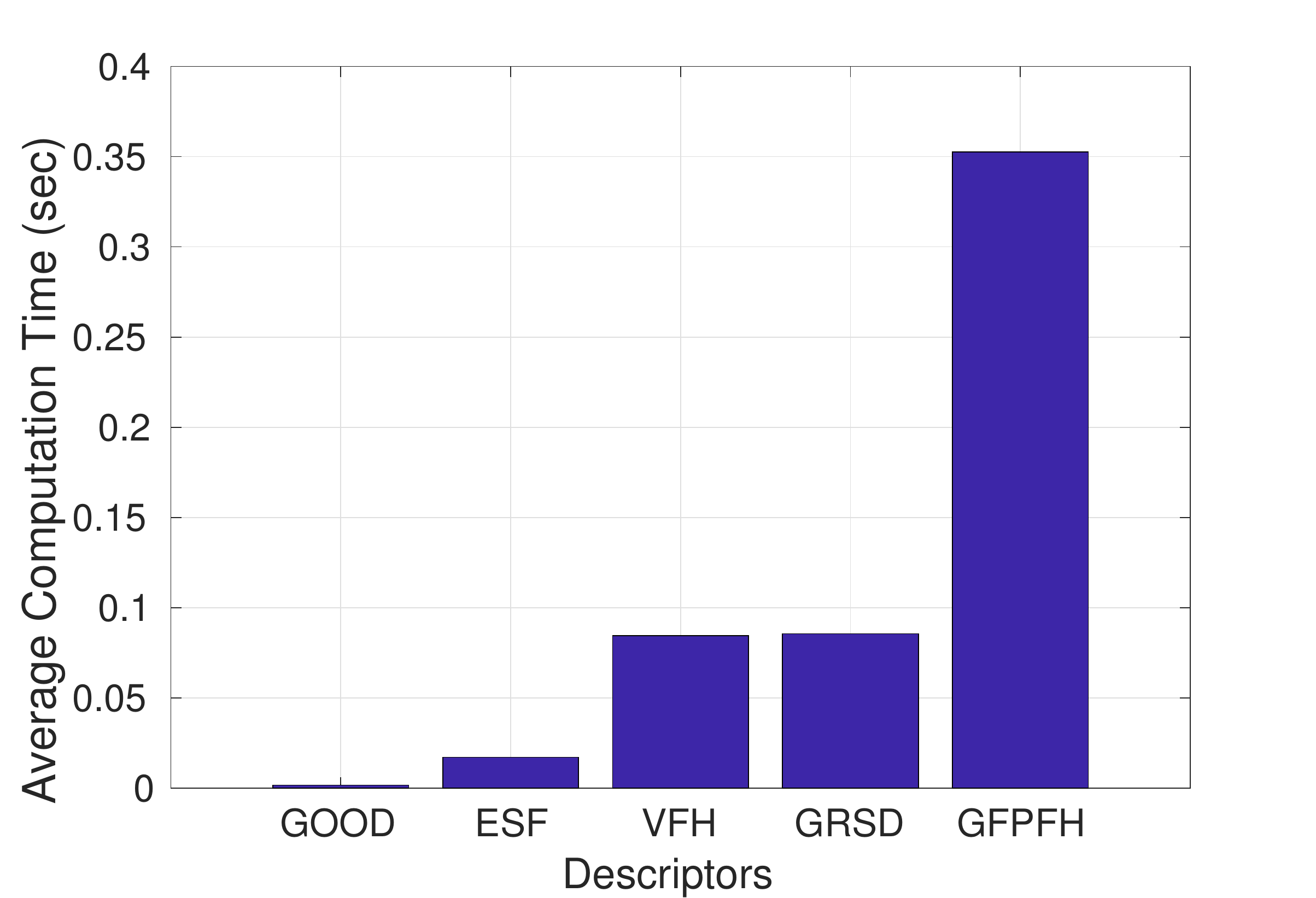} &\quad\quad
		 \includegraphics[width=0.2\linewidth, trim= 0.cm -1.3cm 0cm 0.0cm,clip=true]{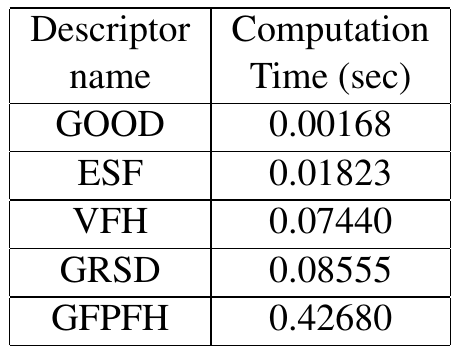}
        \end{tabular}
        \vspace{-2mm}
    \caption{Average computation time of the selected descriptors on 20 randomly selected objects from the RGB-D dataset.}
      	\label{fig:computationTime}
\end{figure}

%^^^^^^^^^^^^^^^^^^^^^^^^^^^^^^^^^^^^^^^^^^^^^^^^^^^^^^^^^^^^^^^^
%^^^^^^^^^^^^^^^^^^^^^^^^^^^^^^^^^^^^^^^^^^^^^^^^^^^^^^^^^^^^^^^^
\section {Summary}

In this chapter, we first focused on the classical evaluation of instance-based (Section \ref{sec:IBL_results}) and model-based (Section~\ref{sec:model_base_learning_results}) object category learning and recognition approaches. The reported results indicate that the overall classification performance obtained with the proposed instance-based learning approach using GOOD feature is better than the best performances achieved with the other approaches. The underlying reason was that GOOD feature encodes the object globally, while the other representations encode the entire object based on a set of local features. Furthermore, it was observed that the model-based learning approach with the GOOD feature provided the best computation time performance. Among the local feature-based approaches, Local LDA achieved the best results. It is due to this point that Local LDA 
transforms objects from bag-of-words space into a local semantic space and used distribution over distribution representation for providing dominant representation. 

After evaluation of all the proposed learning and recognition approaches, a set of experiments was carried out to assess the performance of the GOOD descriptor and compare its performance with other state-of-art descriptors. Experimental results show that the overall classification performance obtained with GOOD is comparable to the best performances obtained with the state-of-the-art Global object descriptors. GOOD outperformed the selected state-of-the-art descriptors (i.e., VFH, ESF, GRSD and GFPFH descriptors), achieving appropriate descriptiveness and significant robustness to Gaussian noise. GOOD was robust to varying low-level point cloud density too. The accuracy of VFH, ESF and GRSD was better than GOOD in the case of varying medium and high point cloud density. In addition, GOOD obtained the best computation time performance. 

The off-line evaluation methodologies (e.g k-fold cross validation, etc.) are not well suited to evaluate open-ended learning systems, because they do not abide to the simultaneous nature of learning and recognition and also those methodologies imply that the set of categories must be predefined. We address this issue in the next chapter by proposing an approach for evaluating open-ended object category learning and recognition approaches in open-ended and multi-context scenarios.

%^^^^^^^^^^^^^^^^^^^^^^^^^^^^^^^^^^^^^^^^^^^^^^^^^^^^^^^^^^^^^^^^
%^^^^^^^^^^^^^^^^^^^^^^^^^^^^^^^^^^^^^^^^^^^^^^^^^^^^^^^^^^^^^^^^

\cleardoublepage
\chapter{Open-Ended Evaluations}
\label{chapter_7}

One of the primary goals in computer vision is to develop reliable capabilities that will allow robots to work in an unconstrained environment by recognizing a large number of object categories. It is still a challenging problem because of ill-definition of objects, large variations in object appearance and {concept drift}. To {deploy} a robot {in} a human-centric environment, it is important that the robot {is} able to continuously {acquire and update} object categories while working in the environment. Therefore, autonomous robots must have the ability to continuously execute learning and recognition in a concurrent and interleaved fashion.  In {an unstructured environment}, an agent must process observations that become gradually available over time{,} and form hypotheses about the environment. If the agent works in a single-context environment, or if the agent can receive/extract explicit information about the current context, one {may consider to} pre-program the agent to use this information for memory management and {for} adapting to the environment. In the first part of this chapter, in order to evaluate and compare the proposed object category learning and recognition approaches in an open-ended manner, the teaching protocol proposed by \cite{Seabra2007} is used. An extensive set of experiments was carried out to evaluate each approach.

The second part of this chapter is dedicated to evaluating how different approaches cope with the effects of context switch. In {a real-world} environment, the context {may} change implicitly and it is not feasible to assume that one can pre-program all the contexts required by {an} agent. Therefore, the agent must have the ability to continuously execute learning and recognition in a concurrent and interleaved fashion {even when changes of context occur without explicit cueing.}  For instance, an intelligent robot working in a human-centric environment needs to learn and remember many different object categories. This is a challenging task since in such environments context may change implicitly and some of the object categories may disappear for some time. As a baseline, the robot must demonstrate a capacity for open-ended learning: that is, the ability to learn new object categories sequentially without forgetting the previously learned object categories. In other words, whenever an agent migrates to a novel context, {some new} object categories should be learned to represent the environment. To achieve adaptability, the agent can either {preserve and} update the current category models, {learning additional} categories {as needed}, or {discard} the categories learned so far and learn new category models {from scratch}. In {unstructured environments}, since the {learning agent may need to go back} to a past context, {discarding} the category models learned in previous {contexts} is not a rational choice (see Fig.~ \ref{fig:context_pr2}). 

%Moreover, it is clear that humans constantly switch between contexts, without re-learning from scratch each time. Several cognitive experiments have been performed showing that animals also {retain} knowledge of past contexts \citep{sissons2009spontaneous, rosas2006context}. Cognitive scientists demonstrated that humans use contextual information to handle large-scale object recognition tasks faster and more accurately \citep{oliva2007role}. {Some authors make a useful distinction between} internal and external contexts \citep{kokinov1997dynamic, snidaro2015context}. \emph{External context} {is} the state of the physical and social environment while \emph{internal context} {is} the current mental (memory) state of the agent.  {Kokinov's dynamic theory of context \citep{kokinov1995dynamic, kokinov1997dynamic} assumes that the internal context influences perception, memory, and reasoning processes. He suggests that the internal context is formed by the interaction between at least three processes: {building new representations based on} perception of the environment; accessing memory traces {therefore} reactivating and possibly modifying old representations; and constructing new representations {based on reasoning}.}  

Having this in mind, we propose an approach for evaluating the adaptability of different open-ended object category learning and recognition methods to context change; %There are several unresolved questions {in addressing} this point: How do agents {represent} past contexts? Can agents reuse previously learned object category models {in new contexts}? {In other words,} whenever a context change {happens}, should the agent build new object category models {from scratch} or {should it} try to reuse and update the previously learned models? To address these points, 
A new teaching protocol, supporting context change, was designed and used for experimental evaluation. A full round of experiments was carried out to assess and compare the proposed methods in depth from the point of view of adaptability to context change. In this chapter, all evaluations are conducted using the Washington RGB-D dataset described in the previous chapter (see Section~\ref{datasets}). %All tests were performed with an i7, 2.40~GHz processor, and 16~GB RAM.

\begin{figure}[t] %\label{fig:spin_images}
\center
\begin{tabular}[width=\textwidth]{cc}
\hspace{-3mm}
 \includegraphics[width=0.5\linewidth, trim= 0cm 0cm 0.0cm 0cm,clip=true]{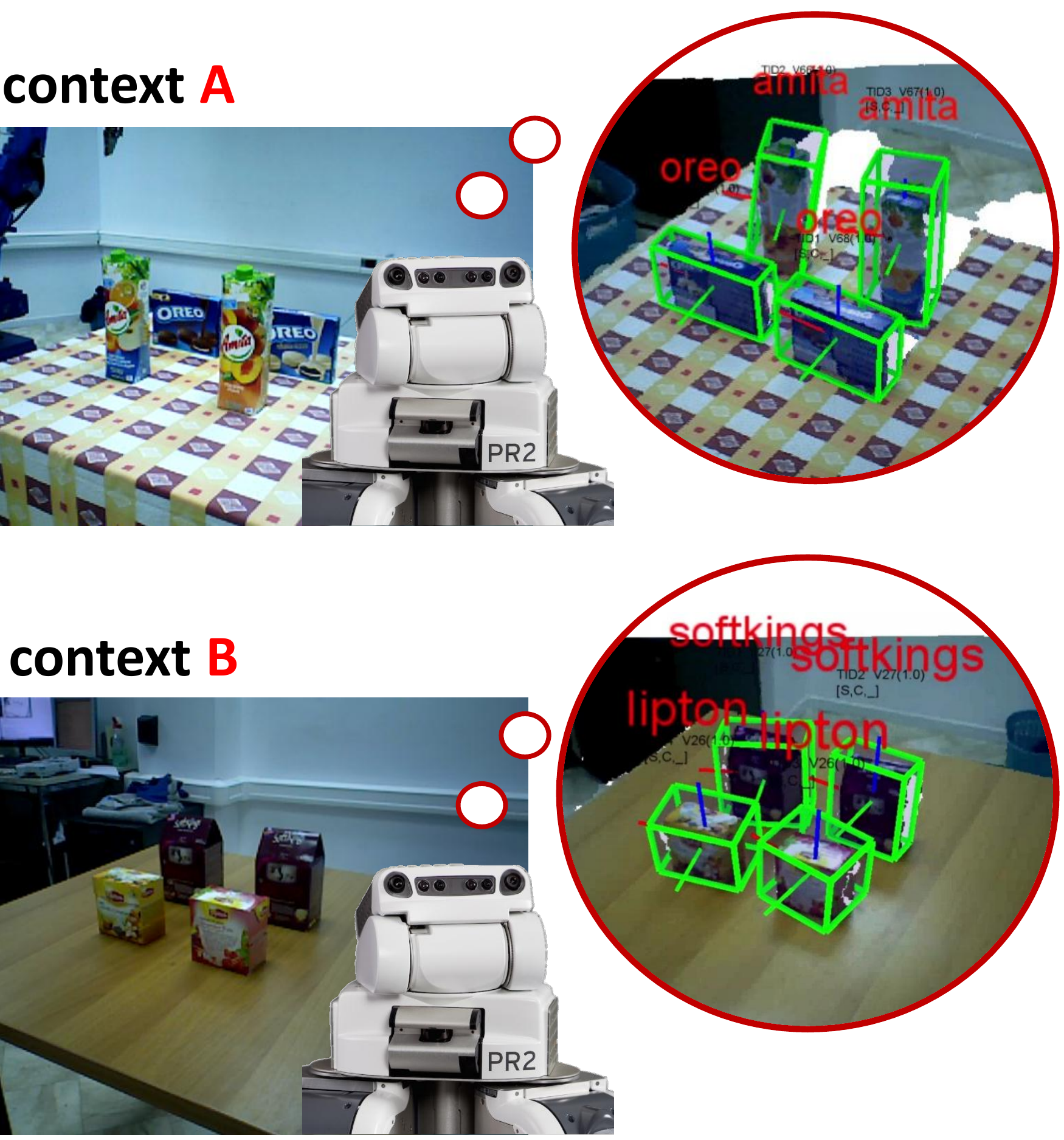} & \hspace{-5mm}
 \includegraphics[width=0.49\linewidth, trim= 0cm 0.0cm 1.0cm 0cm,clip=true]{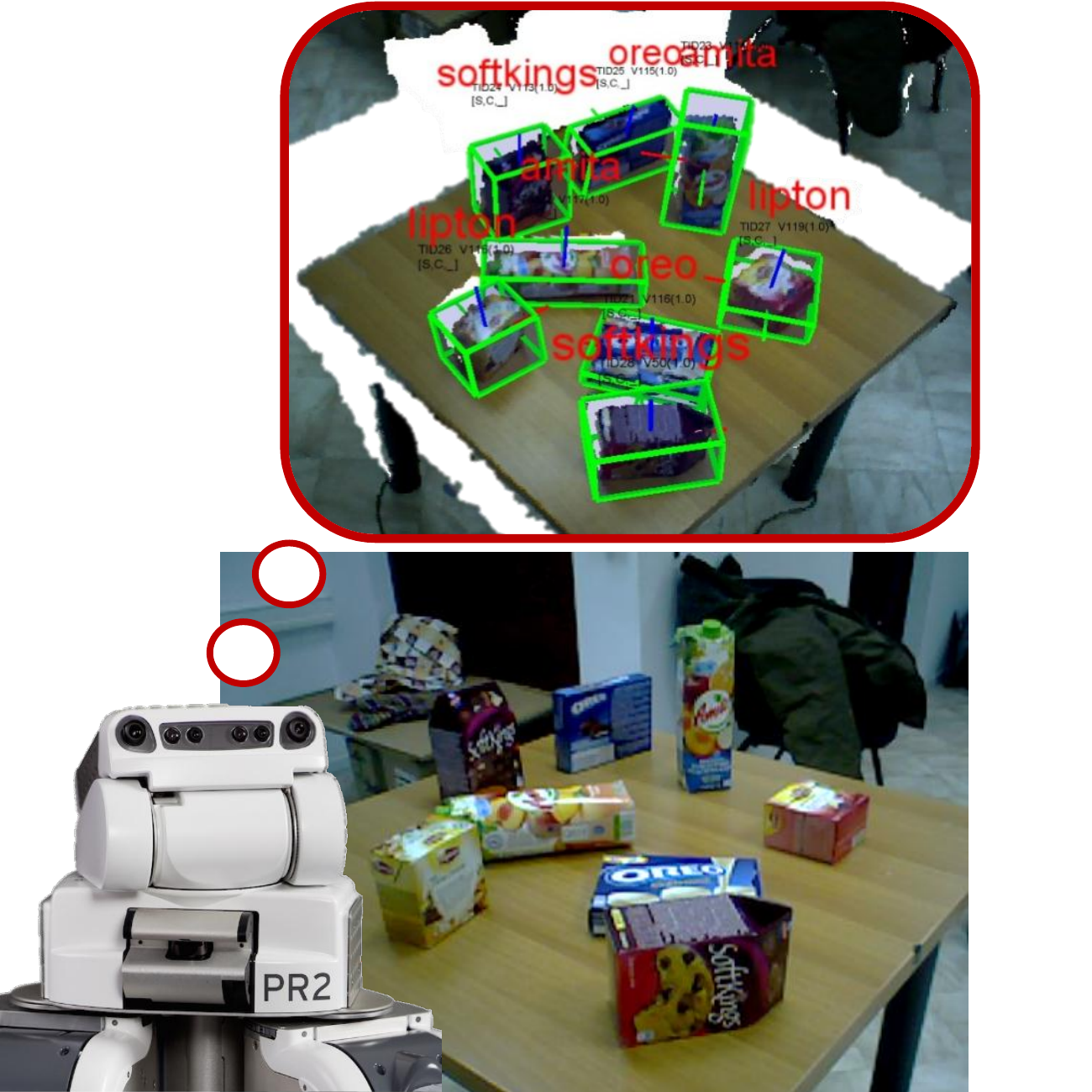}
\end{tabular}\\
\vspace{-2mm}
% figure caption is below the figure
\caption{ This illustration shows that disrupting or erasing the category models learned from the previous contexts is not a rational choice; (\emph{left}) In the \textbf{context A}, the robot learns a set of object categories including \emph{oreo} and \emph{amita} and gains knowledge about \emph{lipton} and \emph{softkings} categories in the \textbf{context B}. (\emph{right}) Later, the robot moves to a new context. In this context, the robot recognizes all objects by using knowledge from previous contexts.}
\vspace{-5mm}
\label{fig:context_pr2}       % Give a unique label
\end{figure}

This chapter presents three main contributions: (\emph{i}) Open-ended evaluation of the proposed object category learning and recognition approaches; (\emph{ii}) An approach for evaluating the adaptability of open-ended object category learning and recognition systems to context change; (\emph{iii}) Evaluation of the adaptability of the proposed object category learning and recognition approaches to context change.  Parts of the work presented herein have been published in the IROS conference \citep{kasaeiIROS2018}.
The remainder of this chapter is organized as follows. In section~\ref{related_work_chapter_7}, we discuss related works. Open-ended evaluation of selected approaches is presented in Section~\ref{sec:open_ended_evaluation_of_selected_approaches}. An evaluation of adaptability to context change is the topic of Section~\ref{sec:evaluation_of_adaptability_to_context_change}. Finally, summary is presented and future research is discussed.

%^^^^^^^^^^^^^^^^^^^^^^^^^^^^^^^^^^^^^^^^^^^^^^^^^^^^^^^^^^^^^^^^
%^^^^^^^^^^^^^^^^^^^^^^^^^^^^^^^^^^^^^^^^^^^^^^^^^^^^^^^^^^^^^^^^

\section {Related Work}
\label{related_work_chapter_7}

In recent years, the role of open-ended learning in {robotics} has been a topic of considerable interest \citep{collet2014herbdisc, oliveira20153d}. In the last decade, various research groups have made substantial progress towards the development of learning approaches which support online, incremental and open-ended category learning \citep{Oliveira2016614,kasaeiNips2016,celikkanat2016learning,kasaei2015adaptive}. Although all the proposed methods have been shown to make progress over the previous one, it is challenging to quantify this progress without a concerted evaluation protocol. Therefore, learning in online and open-ended scenarios calls for new evaluation procedures. Although classical evaluation procedures (holdout, cross-validation, leave-one-out) provide unbiased estimates of the learning performance, they are not suitable for the cases where the problem domain changes over time \citep{Seabra2007,gama2009issues}. 
Accordingly, a well-defined protocol can facilitate the comparison of different approaches as well as the assessment of future improvements. 

%%%%%%%%%%%%%%%%%%%%%%%%%%%%%%%%%%%%%%%%%%%%%%%%%%%%%%%%%%%%%%%%%%%%%%%%%%%%%%%
\subsection{Teaching Protocols for Open-Ended Evaluation}
\label{teaching_protocol_for_open_ended_evaluation}
The well established evaluation methodologies follow the classical train-and-test procedure, i.e., two separate stages, training followed by testing. Training is accomplished offline, and once it is complete the testing is performed. These methodologies are not well suited to evaluate open-ended learning systems, because they do not abide to the simultaneous nature of learning and recognition and because the number of categories must be predefined. 

Teaching protocols for open-ended evaluation of a learning algorithm determines which examples are used for training
the algorithm, and which are used to test the algorithm. Seabra Lopes and Chauhan proposed a teaching protocol to evaluate the ability of an agent to incrementally acquire visual object categories in an open-ended setting \citep{Seabra2007,chauhan2011}. This protocol, which can be followed by a human teacher or by a simulated teacher, is based on a Test-then-Train scheme. It is an elaborate and exhaustive evaluation procedure, where, for every new category introduced to the agent, the average accuracy of the system is calculated by performing classification with all known categories. Towards this end, a teacher repeatedly presents instances of known categories to the agent, checks the agent’s predictions and provides corrective feedback in case of misclassification. This way, the system is trained, and at the same time the recognition performance of the system is continuously estimated. More specifically, the teacher interacts with the learning agent using three basic actions (see sections \ref{sec:supervisedExperienceGathering} and \ref{HRI} for additional details):
\begin{itemize}
\item {\textbf{Teach}}: used for introducing a new object category;
\item \textbf{{Ask}}: used to ask the system what is the category of a given object
view;
\item {\textbf{Correct}}: used for providing corrective feedback in case of
misclassification.
\end{itemize}
\noindent Algorithm~7.1 describes the teaching protocol. This protocol has been used in several recent works \citep{Oliveira2016614,kasaeiNips2016}. 
%\begin{figure}[!t]
%\vspace{-3mm}
%\centering
%\includegraphics[width=0.85\linewidth]{Figures/algorithm_original.pdf}
%\vspace{-2mm}
%\label{fig:simulated_teacher_algorithm_original}
%\end{figure}

\begin{algorithm}[!t]
\DecMargin{5mm} 
\SetAlgoRefName{7.1} 
\caption*{ Teaching protocol for performance evaluation}
\begin{algorithmic}[1]
\State \textbf{Introduce} $Category_1$
\State $n \gets \textit{1}$
\State  \textbf{repeat} 
\State \quad\quad$n\gets \textit{n+1}$ \Comment {{\scriptsize ready for the next category}}
\State \quad\quad\textbf{Introduce} $Category_n$
\State \quad\quad $k\gets \textit{0}$
\State \quad\quad $c \gets \textit{1}$
\State \quad\quad\textbf{repeat} \Comment {{\scriptsize question / correction iteration}}
\State \quad\quad\quad\quad Present a previously unseen instance of $Category_c$
\State \quad\quad\quad\quad \textbf{Ask} the category of this instance
\State \quad\quad\quad\quad If needed, provide \textbf{correct} feedback

\State \quad\quad\quad\quad c {$\gets$ c+1 \textbf{if} $\textit{c < n}$ \textbf{else} ~$1$}
\State \quad\quad\quad\quad \textbf{if}$ k \le 3n$ \textbf{then} \Comment {{\scriptsize sliding window}}
\State \quad\quad\quad\quad\quad  $k\gets \textit{k+1}$
\State \quad\quad\quad\quad $s\gets$ accuracy in last $k$ question/correction iterations
\State \quad\quad\textbf{until (}($s$ > $\tau$ \textbf{and} $k$ >= $n$)
\Comment {{\scriptsize accuracy threshold crossed}}
\State \quad\quad \quad\quad\quad  \textbf{or} (user sees no improvement in accuracy)\textbf{)}
\Comment {{\scriptsize breakpoint reached}}
\State  \textbf{until (}user sees no improvement in accuracy\textbf{)}
\Comment {{\scriptsize breakpoint reached}}
\end{algorithmic}
\end{algorithm}

The \emph{teacher} continuously estimates the recognition performance (accuracy) of the agent using a sliding window of size $3n$ iterations, where $n$ is the number of categories
that have already been introduced. If $k$, the number of iterations since the last time a new category was introduced, is less than $3n$, all results are used. In case accuracy exceeds a given classification threshold ($\tau = 0.67$), the teacher introduces a new object category.

In another work, this protocol has been modified to cope with scenarios in which the learning agent is able to acquire large sets of categories (hundreds or more) \citep{chauhan2015experimental}. The major differences are two fold: in the original protocol, after introducing a new category, all known categories are tested once, while \cite{chauhan2015experimental} proposed that it is enough to test a randomly generated subset of all known categories for introducing a new category; the second difference, with respect to the original protocol, is that, in the original protocol, categories are tested in the sequence in which they were introduced to the agent, while in the modified protocol, categories are tested in a random sequence, which steers to prevent storing more instances for the categories introduced earlier.  

In the field of data-stream learning, i.e., sensor networks, social networks, financial data etc., the data usually became available through the time and can significantly change over time. Given that the number of categories is usually predefined in data-stream learning, the evaluation procedure follows the standard train and test approach. Therefore, the evaluation of these algorithms faces the same issues as that for evaluating open-ended learning algorithms. To handle this issue, some online evaluation approaches have been proposed, such as Prequential evaluation \citep{gama2009issues,gama2013evaluating} and MOA \citep{bifet2010moa}. These approaches also follow a Test-then-Train scheme. However, they do not take into account the impact of learning new categories and the capability of learning algorithms to scale up to larger sets of categories. The existing online evaluation approaches do not take into account the possibility of context change. In this chapter, the mentioned teaching protocol \citep{Seabra2007,chauhan2011} is modified to evaluate the learning agent in scenarios of context change.

%%%%%%%%%%%%%%%%%%%%%%%%%%%%%%%%%%%%%%%%%%%%%%%%%%%%%%%%%%%%%%%%%%%%%%%%%%%%%%%
\subsection{Metrics for Open-Ended Evaluation}

After deciding which teaching protocol is suitable for evaluating an open-ended learning system, one of the unique concerns is how to build a picture of performance over time. Some authors consider the classical measures versus training time. In classical scenarios with a fixed set of categories, such evaluations show a gradual
increase of the classical measures and a convergence to a stable value. The performance evaluation of an open-ended learning system cannot be limited to the classical evaluation metrics (see Section~\ref{sec:evaluation_metrics}). In addition to classical measures such as accuracy, precision and recall, an open-ended learning approach should be evaluated by the other measures like ``\emph{number of learned categories}'' and ``\emph{number of teaching iterations}''. %Some authors considered the classical measures versus training time. In classical scenarios with a fixed set of categories, such evaluations show a gradual increase of the classical measures and a convergence to a stable value. 

In the continuation of their earlier works~\citep{Seabra2007,chauhan2011}, \cite{chauhan2015experimental} proposed additional criteria and measures to evaluate the overall learning performance of the agent during an open-ended experiment. In particular, they suggested the following measures to characterize the quality and coverage of the learned knowledge after an experiment has finished:
{
\renewcommand{\labelitemi}{$\blacksquare$}
\begin{enumerate}[-]
\item \textbf{\emph{Global accuracy}}: This is given as the percentage of correct predictions made during a complete teaching protocol experiment, i.e., in an experiment that reached the breakpoint.
\item \textbf{\emph{Average protocol accuracy}}: The average of all protocol accuracy values, i.e., computed over all the question/correction iterations in a complete teaching protocol experiment;
\item \textbf{\emph{Number of learned categories}} in a complete teaching protocol experiment. 
\end{enumerate}

\noindent In applications where the learning agent can recognize an object as belonging to an ``\emph{unknown}'' category, the accuracy measure can be replaced by some other recognition success measure, e.g. F-measure. Additionally, they considered the following measures to characterize the learning process in terms of memory and time:
\begin{enumerate}[-]
\item \textbf{\emph{Number of question/correction iterations}} during the experiment;
\item \textbf{\emph{Average number of stored instances per category}} in a complete teaching protocol experiment (this measure is applicable to instance-based learning
approaches only).
\end{enumerate}

\cite{Oliveira2015} proposed to organize these evaluation metrics in three groups:

\begin{enumerate}[-]
\item \textbf{\emph{How much does it learn?}} This is measured as the number of categories learned in a complete teaching protocol experiment;
\item \textbf{\emph{How well does it learn?}} They mentioned that classical evaluation metrics such as accuracy, precision and recall are well suited for this metric. They used global accuracy measured as defined above.
\item \textbf{\emph{ How fast does it learn?}} This is measured as the number question/correction iterations in
a teaching protocol experiment up to a certain number of learned categories.
\end{enumerate}
}

%It is worth mentioning that these questions are inspired from the \citep{Seabra2007}, \citep{chauhan2011} and \cite{chauhan2015experimental} works.
 
%%%%%%%%%%%%%%%%%%%%%%%%%%%%%%%%%%%%%%%%%%%%%%%%%%%%%%%%%%%%%%%%%%%%%%%%%%%%%%%
\subsection{Context Change in Open-Ended Learning}

Humans can adapt to different environments dynamically by watching and learning. Learning is closely related to memory in human cognition. 
\cite{yeh2006situated} demonstrated in a series of experiments that human subjects perform better at a variety of cognitive tasks when taking context into account. Without considering contextual information, all possibilities in a classification space must be explored, which scales poorly with large-scale data. 

Several cognitive experiments have been performed showing that animals also {retain} knowledge of past contexts \citep{sissons2009spontaneous, rosas2006context}. Cognitive scientists demonstrated that humans use contextual information to handle large-scale object recognition tasks faster and more accurately \citep{oliva2007role}.

{Some authors make a useful distinction between} internal and external contexts \citep{kokinov1997dynamic, snidaro2015context}. \emph{External context} {is} the state of the physical and social environment while \emph{internal context} {is} the current mental (memory) state of the agent.  {Kokinov's dynamic theory of context \citep{kokinov1995dynamic, kokinov1997dynamic} assumes that the internal context influences perception, memory, and reasoning processes. He suggests that the internal context is formed by the interaction between at least three processes: {building new representations based on} perception of the environment; accessing memory traces {therefore} reactivating and possibly modifying old representations; and constructing new representations {based on reasoning}.}

\cite{kokinov1995dynamic,kokinov1997dynamic} introduced several distinctions between various concepts, e.g. internal versus external context, implicit versus explicit context, and proposed a dynamic approach to context modeling. External context is related to the social and situational dimensions of contexts such as location, time, light and co-location. Internal context can be understood as a mental (memory) state of the agent.  Researchers in both cognitive science \citep{rosas2013context} and computer vision {communities} \citep{galleguillos2010context, oliva2007role} have mainly studied the effects of external context and very rarely the internal context~\citep{snidaro2015context}. Recently, several computer vision approaches have shown that {information about the} external context improves the efficiency of the perceptual tasks such as object detection \citep{mottaghi2014role} and semantic segmentation \citep{mottaghi2014role, shelhamer2016fully}. 

In robotics, the notion of context has grown in prominence over the last decade. Several researchers considered the role of context in object recognition. They used explicit context information, in training and/or in recognition, with different methods for representing the context in terms of relationship among objects in a scene \citep{galleguillos2010context,mottaghi2014role,RuizSarmiento20158805,ruiz2015joint}. Moreover, they showed that a statistical summary of the scene (i.e., global scene representation) provides a rich source of information for contextual inference. For instance, \cite{mottaghi2014role} investigated the role of context for object detection and semantic segmentation. They proposed a new deformable part-based model, which exploits both local {and} global context for object segmentation in {unstructured} environments and showed that this contextual reasoning {was} useful to detect objects at all scales. \cite{nigam2015social} proposed a social context perception {approach} for mobile robots. They considered different {aspects} of the external context, including social environment, physical location, audio (i.e., varied levels of noise) and the time of day, to recognize dining, {studying} and lobby (waiting) contexts. 

Such approaches are at some point dependent on an explicit context cue and may fail when the environment undergoes a change in context without explicit cueing. This is an important limitation since no matter how extensive the training data, an agent might always be confronted with unknown objects in new contexts when operating in everyday environments. Therefore, the agent should be able to deal with implicit context changes in an incremental and open-ended manner. The ability of different learning techniques to cope with context change in the absence of explicit cueing is usually not evaluated. Furthermore, the existing online evaluation approaches do not take into account the possibility of context change. That is precisely one of the focuses of this chapter. In particular, the teaching protocol of \citep{Seabra2007, chauhan2011} is modified to evaluate the learning agent in scenarios of context change. The idea is to emulate the interactions of a {learning and} recognition system with the surrounding environment over long periods of time and evaluate how different approaches cope with the effects of context switch.  For this {purpose}, a simulated teacher was developed to follow the modified teaching protocol and autonomously interact with {the developed learning} agent. After teaching a certain number of categories, the simulated teacher changes the context and {continues teaching and testing} the agent in {the} new context.

%%%%%%%%%%%%%%%%%%%%%%%%%%%%%%%%%%%%%%%%%%%%%%%%%%%%%%%%%%%%%%%%%%%%%%%%%%%%%%%%%%%%%%%%%%%%%

\section {Open-Ended Evaluation of Selected Approaches}
\label{sec:open_ended_evaluation_of_selected_approaches}

This section presents the experimental setup and the results obtained for the main approaches proposed and explored in this thesis. For each approach, the default parameters, as reported in the previous chapter, were used.

\subsection {Simulated Teacher}

Off-line evaluation methodologies do not {comply} with the simultaneous nature of learning and recognition {in autonomous agents. Moreover, they assume} that the set of categories is predefined. Therefore, the mentioned teaching protocol \citep{Seabra2007,chauhan2011} is adopted in this evaluation (Algorithm 7.1; see section~\ref{teaching_protocol_for_open_ended_evaluation}). The idea is to emulate the interactions of a learning agent with the surrounding environment over long periods of time.
{The protocol can be followed by a human teacher. However,} replacing a human teacher with a simulated one makes it possible to conduct systematic, consistent and reproducible experiments for different approaches. It allows the possibility to perform multiple experiments and explore different experimental conditions in a fraction of time a human would take to carry out the same task.
\begin{figure}[!t]
\centering
\includegraphics[width=0.85\linewidth, trim = 0cm 0cm 0cm 0cm, clip=true ]{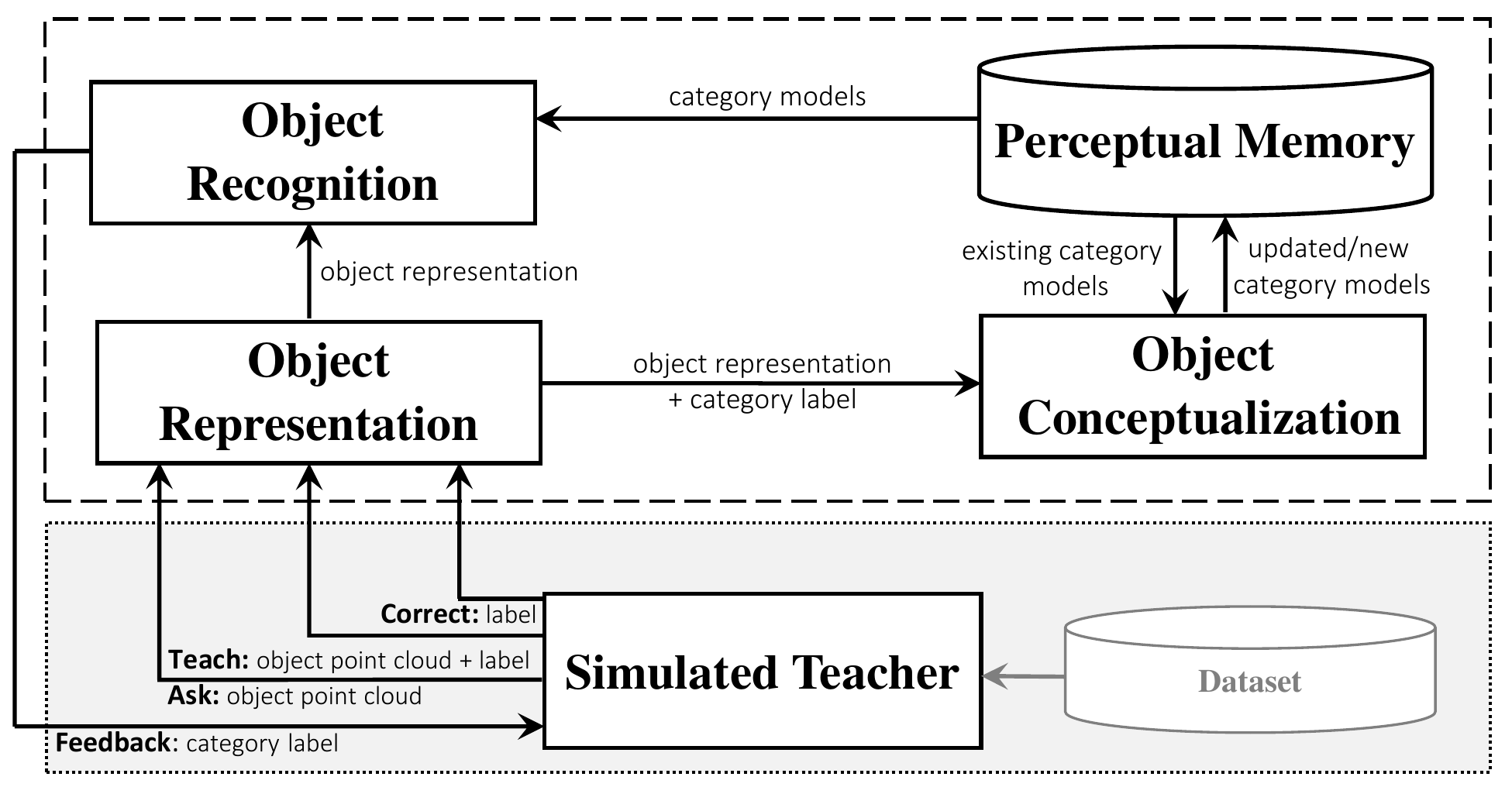}
\caption{{Interaction between the simulated teacher and the learning agent}}
\label{fig:simulated_teacher}
\end{figure}

For this purpose, a simulated teacher, connected to a large database of labelled object views was developed. The overall system architecture is depicted in Fig.~\ref{fig:simulated_teacher}. In this chapter, the Washington RGB-D dataset is used (i.e., see Section~\ref{datasets}).
The idea is that the \emph{simulated teacher} repeatedly picks unseen object views from the currently known categories and presents them to the agent for testing. Inside the learning agent, the object view is recorded in the \emph{Perceptual Memory} if it is marked as a {training sample} (i.e., whenever the teacher uses teach or correct instructions), otherwise it is dispatched to the \emph{Object Recognition} module. 

For introducing a new category, the simulated teacher presents three randomly selected object views. Since the order in which categories are introduced in a teaching protocol experiment may have an effect on the performance of the system, 10 experiments were carried out for each approach and, in each experiment, categories were introduced in random sequences. These sequences were the same for all approaches.

%The \emph{simulated teacher} continuously estimates the recognition performance of the agent using a sliding window of size $3n$ iterations, where $n$ is the number of categories that have already been introduced. If $k$, the number of iterations since the last time a new category was introduced, is less than $3n$, all results are used. In case this performance exceeds a given classification threshold ($\tau = 0.67$),  

In case the agent cannot reach the classification threshold after a certain number of iterations (i.e., 100 iterations), the simulated teacher can infer that the agent
is no longer able to learn more categories and therefore, terminates the experiment. It is possible that an approach learns all existing categories before reaching the breakpoint. In such case, it is no longer possible to continue the protocol and the evaluation process is halted. In the reported results, this is shown by the stopping condition, ``\emph{lack of data}''.

\subsection{Evaluation Metrics} 

When an experiment is carried out, learning performance is evaluated using several measures, including: (\emph{i}) the number of learned categories (NLC) at the end of the experiment, an indicator of \textbf{\emph{how much the system was capable of learning}}; (\emph{ii}) the number of question / correction iterations (QCI) required to learn those categories and the average number of stored instances per category (AIC), indicators of \emph{\textbf{time and memory resources required for learning}}; (\emph{iii}) Global Classification Accuracy (GCA), computed using all predictions in a complete experiment, and the Average Protocol Accuracy (APA),  i.e., average of all accuracy values successively computed during the experiment to control the application of the teaching protocol. GCA and APA are indicators of \emph{\textbf{how well the system learns}}.

%%%%%%%%%%%%%%%%%%%%%%%%%%%%%%%%%%%%%%%%%%%%%%%%%%%%%%%%%%%%%%%%%%%%%%%%%%%%%%%%%%%%%%%%%
\subsection{Evaluation of Instance-Based Approaches}
The first round of experiments was performed to evaluate the proposed learning approaches concerning their scalability with respect to the number of learned categories. 
\begin{figure}[!t]
\resizebox{\linewidth}{!}{
\begin{tabular}{ccc}
	\includegraphics[width=0.36\linewidth,  trim= 0cm -1cm 0.0cm 0cm,clip=true]{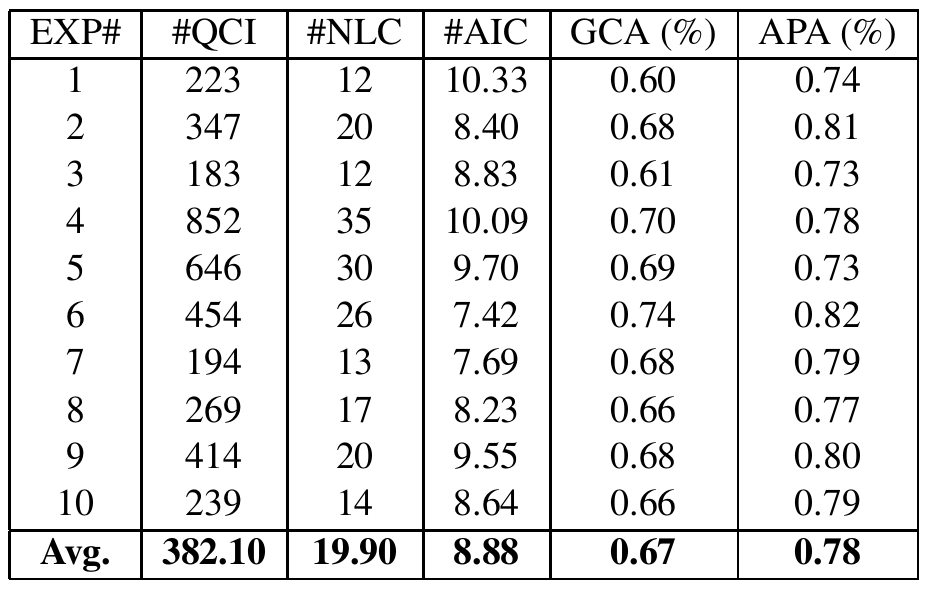}& \hspace{-0mm}
	\includegraphics[width=0.35\linewidth, trim= 0.5cm 0cm 1cm 0.65cm,clip=true]{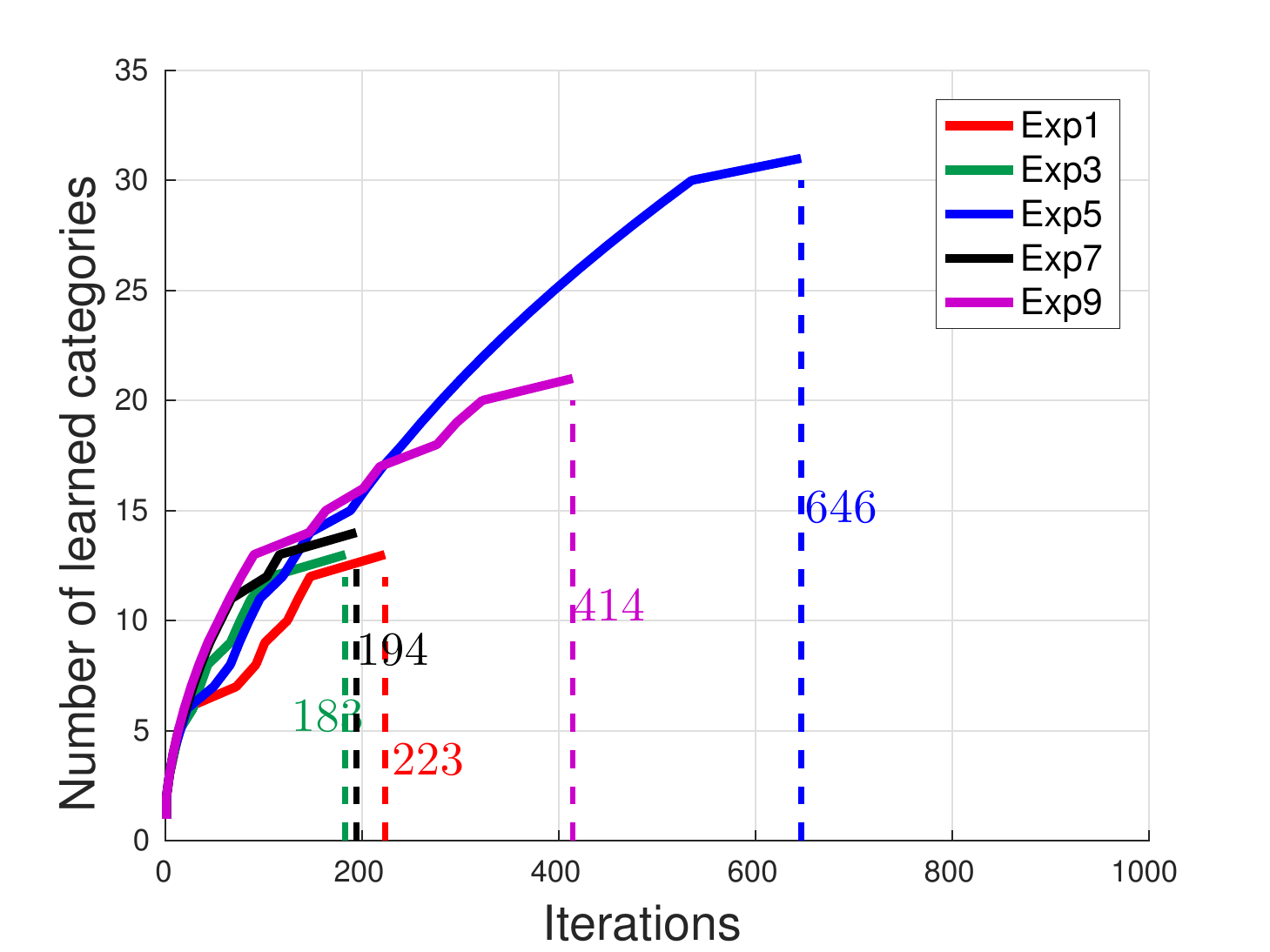}& \hspace{-0mm}
	\includegraphics[width=0.33\linewidth, trim= 0.35cm 0cm 1cm 0.65cm,clip=true]{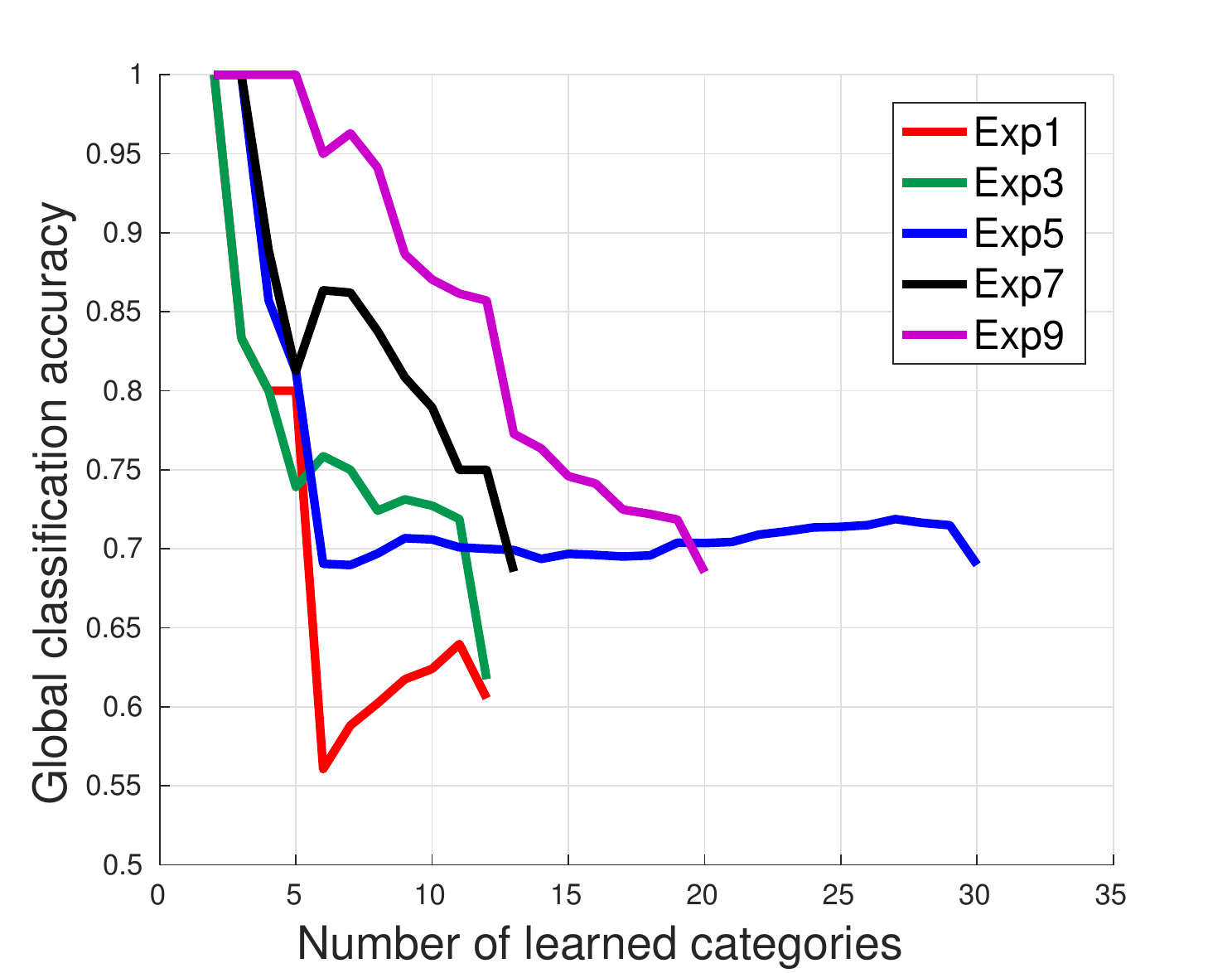}	\vspace{0mm}\\
	 \multicolumn{3}{c}{ (a) Summary of experiments using sets of local features (Approach~II) }\vspace{1mm}
	\\ 
	\includegraphics[width=0.36\linewidth,  trim= 0cm -1cm 0cm 0cm,clip=true]{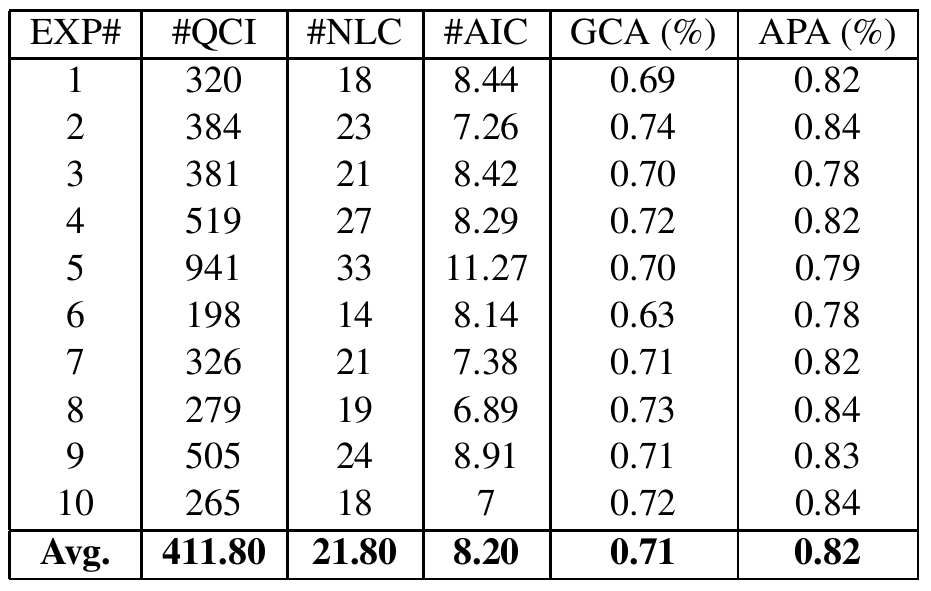}&\hspace{-0mm}
	\includegraphics[width=0.34\linewidth, trim= 0.5cm 0cm 1cm 0.65cm,clip=true]{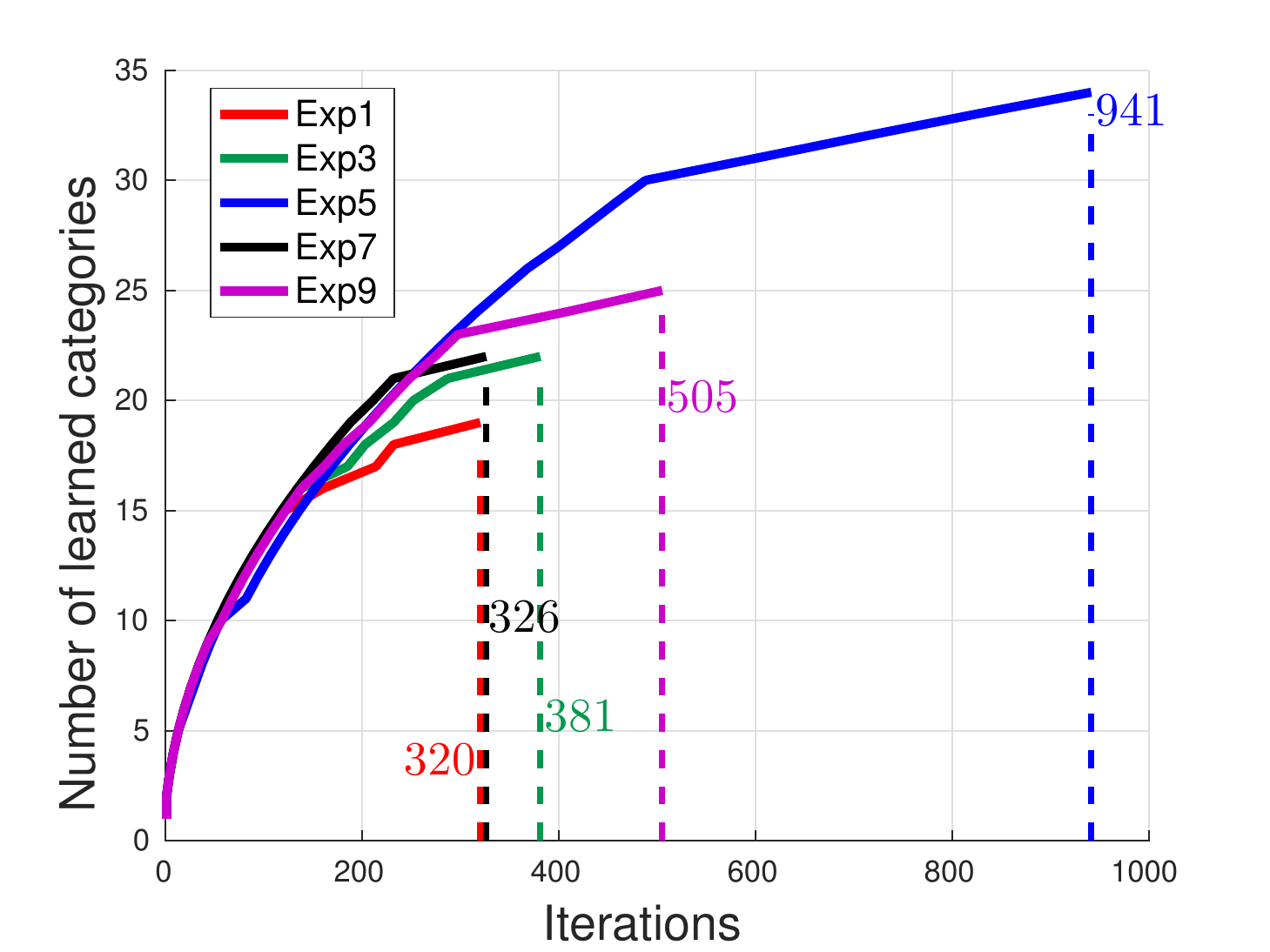}& \hspace{-0mm}
	\includegraphics[width=0.33\linewidth, trim= 0.35cm 0cm 1cm 0.65cm,clip=true]{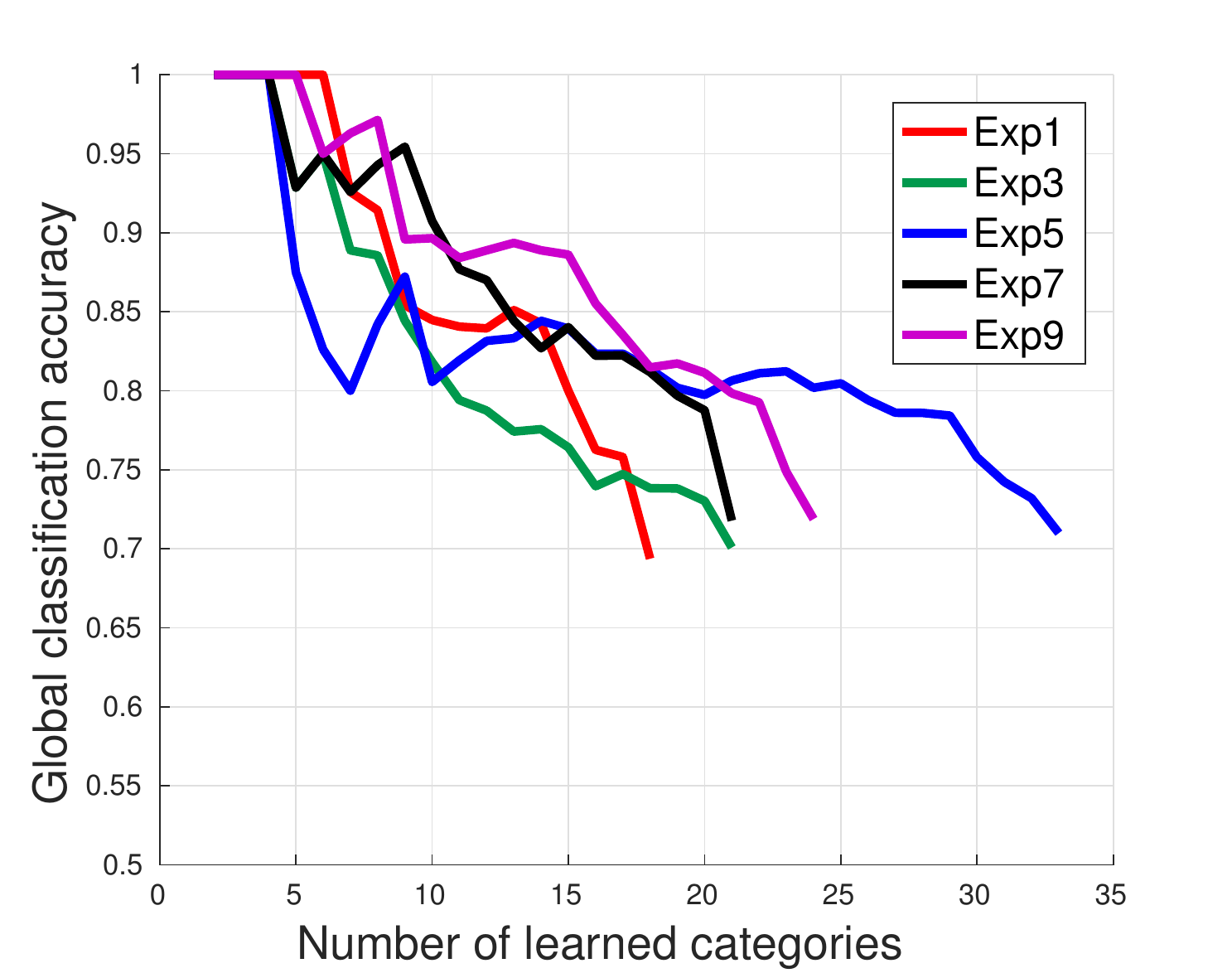}	\vspace{0mm}\\
	\multicolumn{3}{c}{(b) Summary of experiments using BoW}\vspace{1mm}
	\\
	\includegraphics[width=0.36\linewidth,  trim= 0cm -0.7cm 0.0cm 0cm,clip=true]{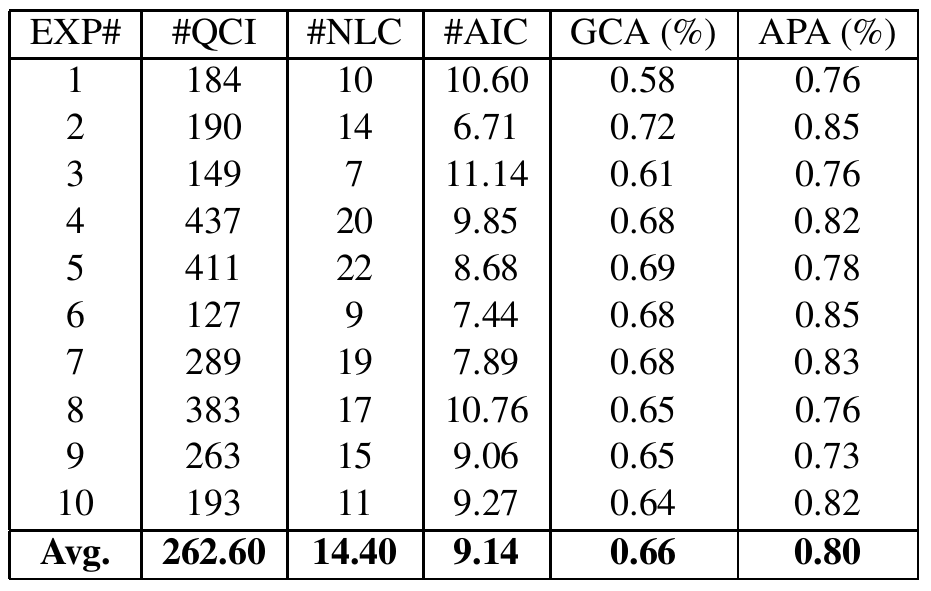}&\hspace{-0mm}
	\includegraphics[width=0.34\linewidth, trim= 0.5cm 0cm 1cm 0.65cm,clip=true]{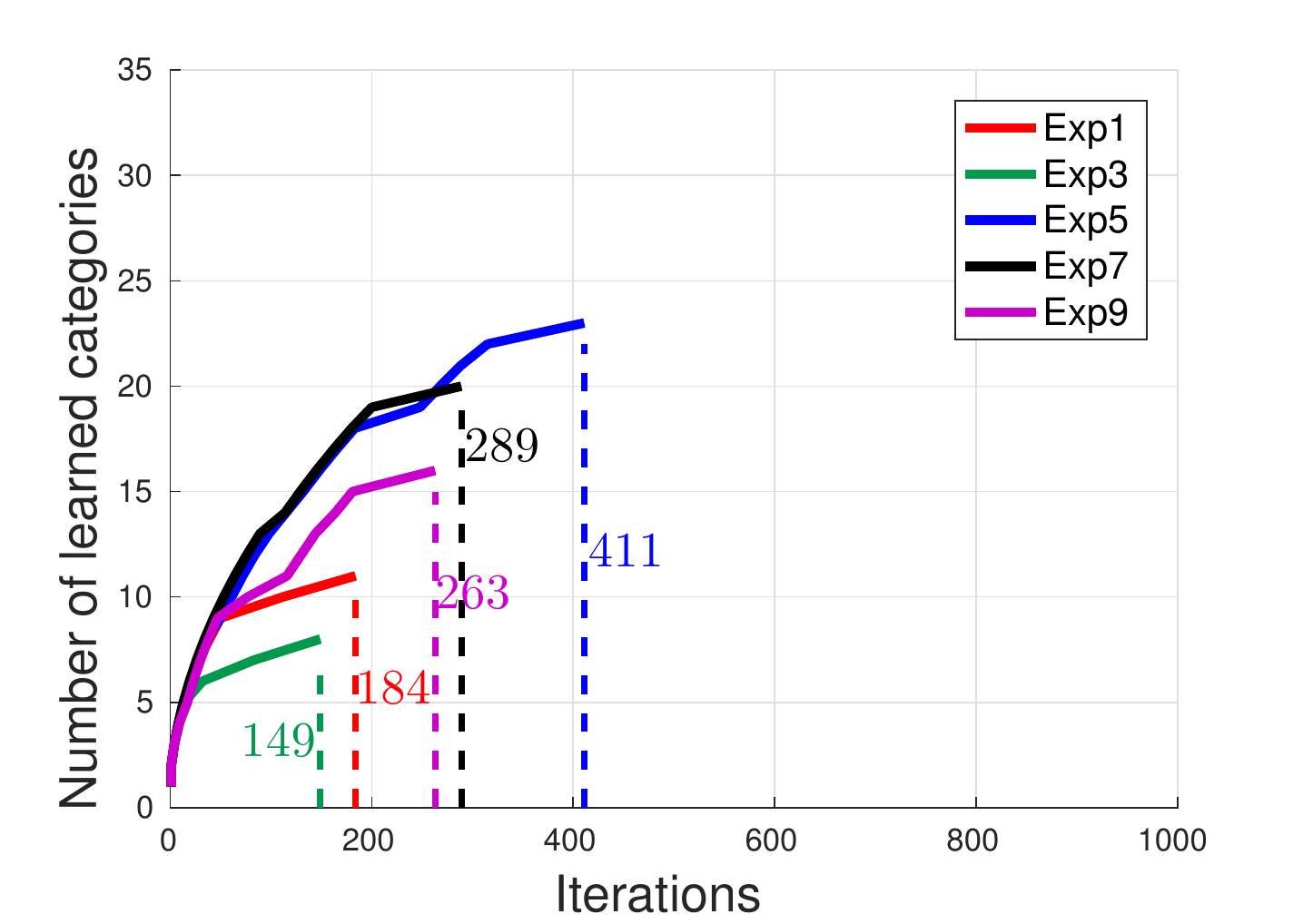}& \hspace{-0mm}
	\includegraphics[width=0.33\linewidth, trim= 0.35cm 0cm 1cm 0.65cm,clip=true]{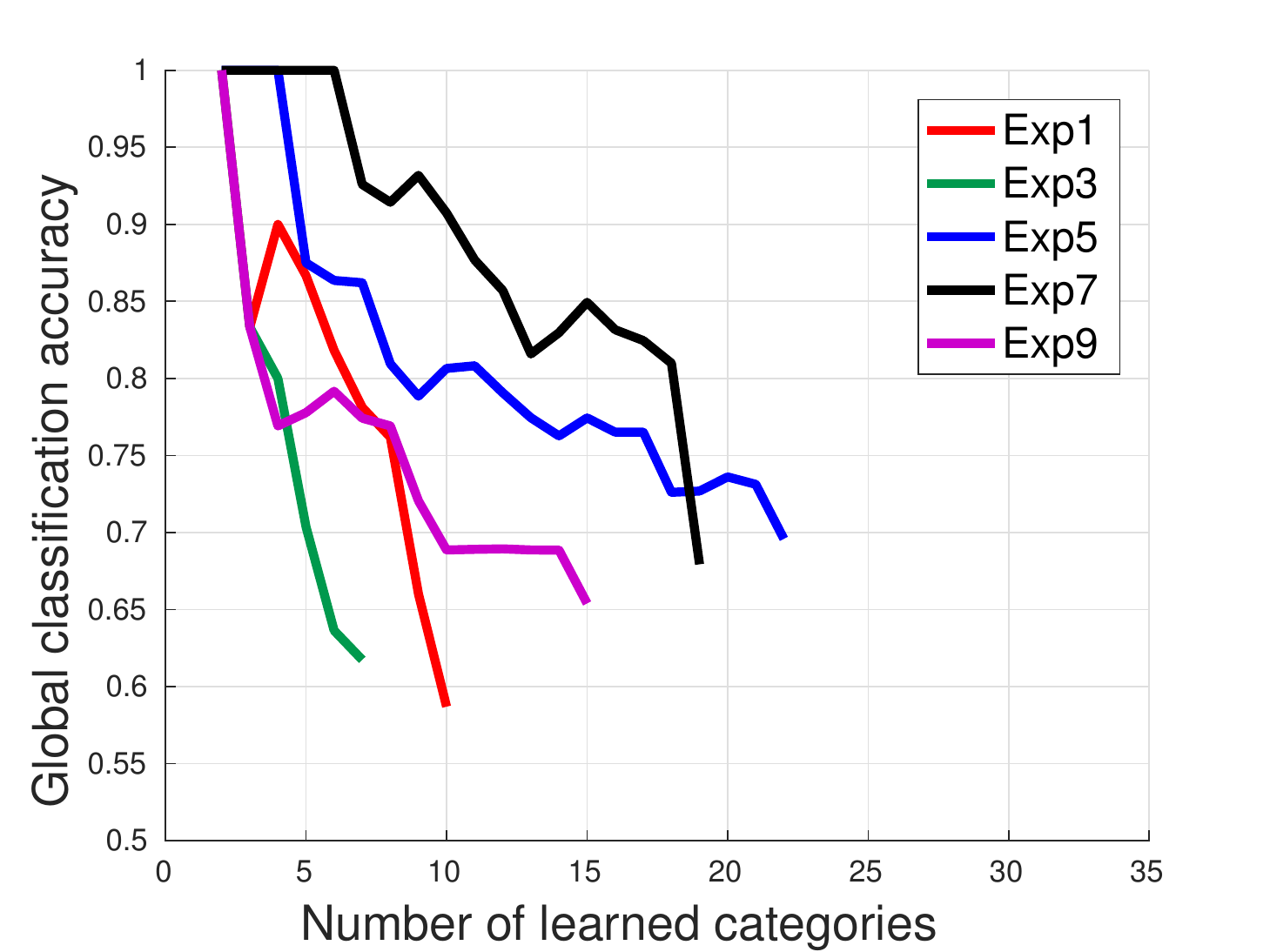}	\vspace{0mm}\\
		 \multicolumn{3}{c}{(c) Summary of experiments using standard LDA}\vspace{1mm}	
	\\
		\includegraphics[width=0.36\linewidth,  trim= 0cm -1cm 0.0cm 0cm,clip=true]{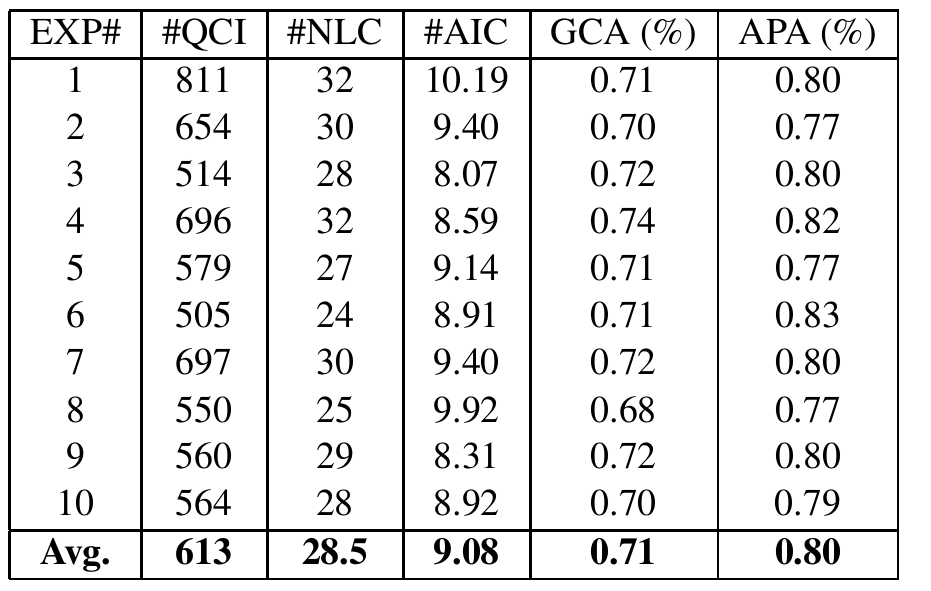}&\hspace{-1mm}
	\includegraphics[width=0.34\linewidth, trim= 0.5cm 0cm 1cm 0.65cm,clip=true]{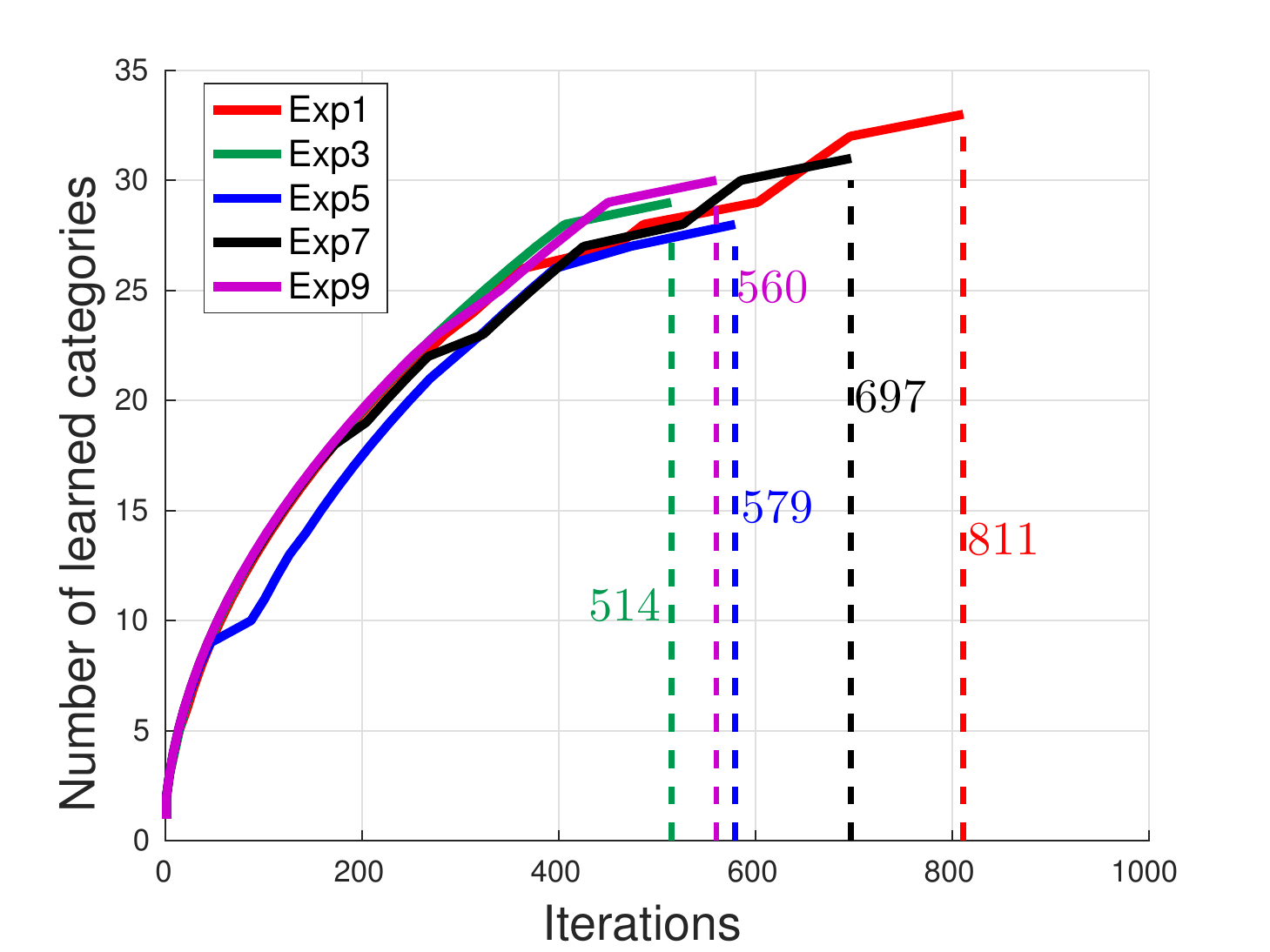}&  \hspace{-3mm}
		\includegraphics[width=0.33\linewidth, trim= 0.35cm 0cm 1cm 0.65cm,clip=true]{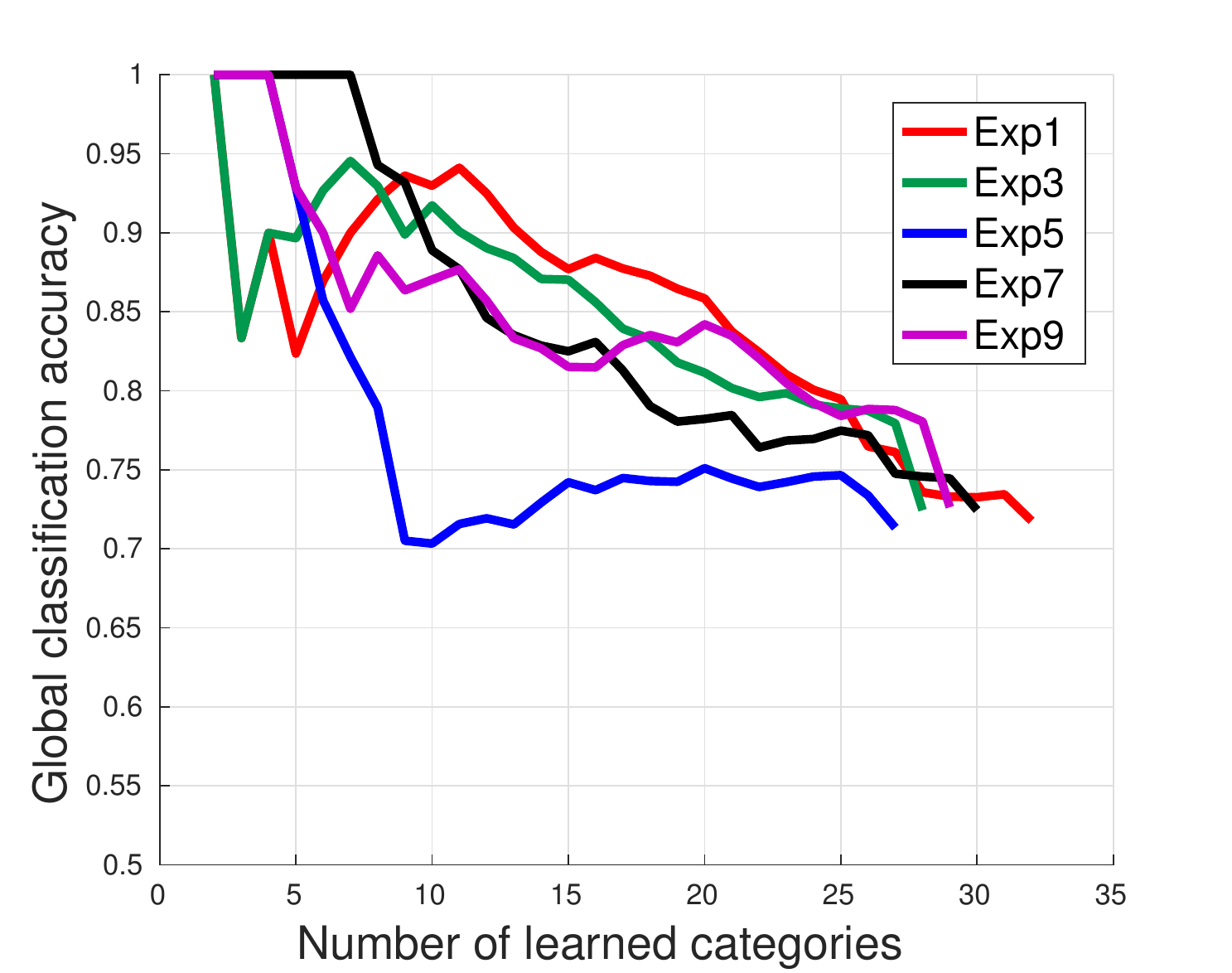}\\
			 \multicolumn{3}{c}{(d) Summary of experiments using Local LDA}\vspace{0mm}	
	\\
		\includegraphics[width=0.35\linewidth,  trim= 0cm -1cm 0.0cm 0cm,clip=true]{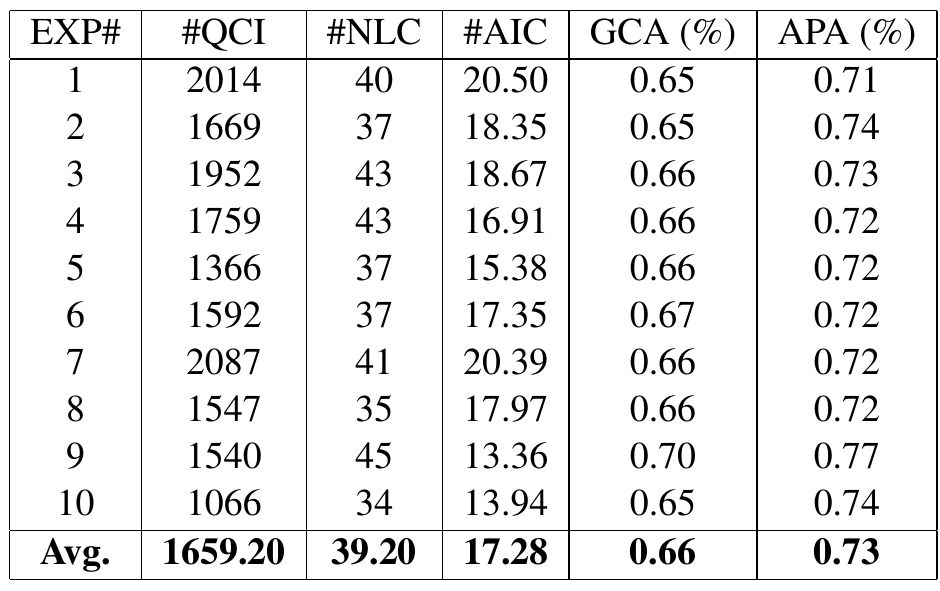}&\hspace{-1mm}
	\includegraphics[width=0.35\linewidth, trim= 0.5cm 0cm 1cm 0.65cm,clip=true]{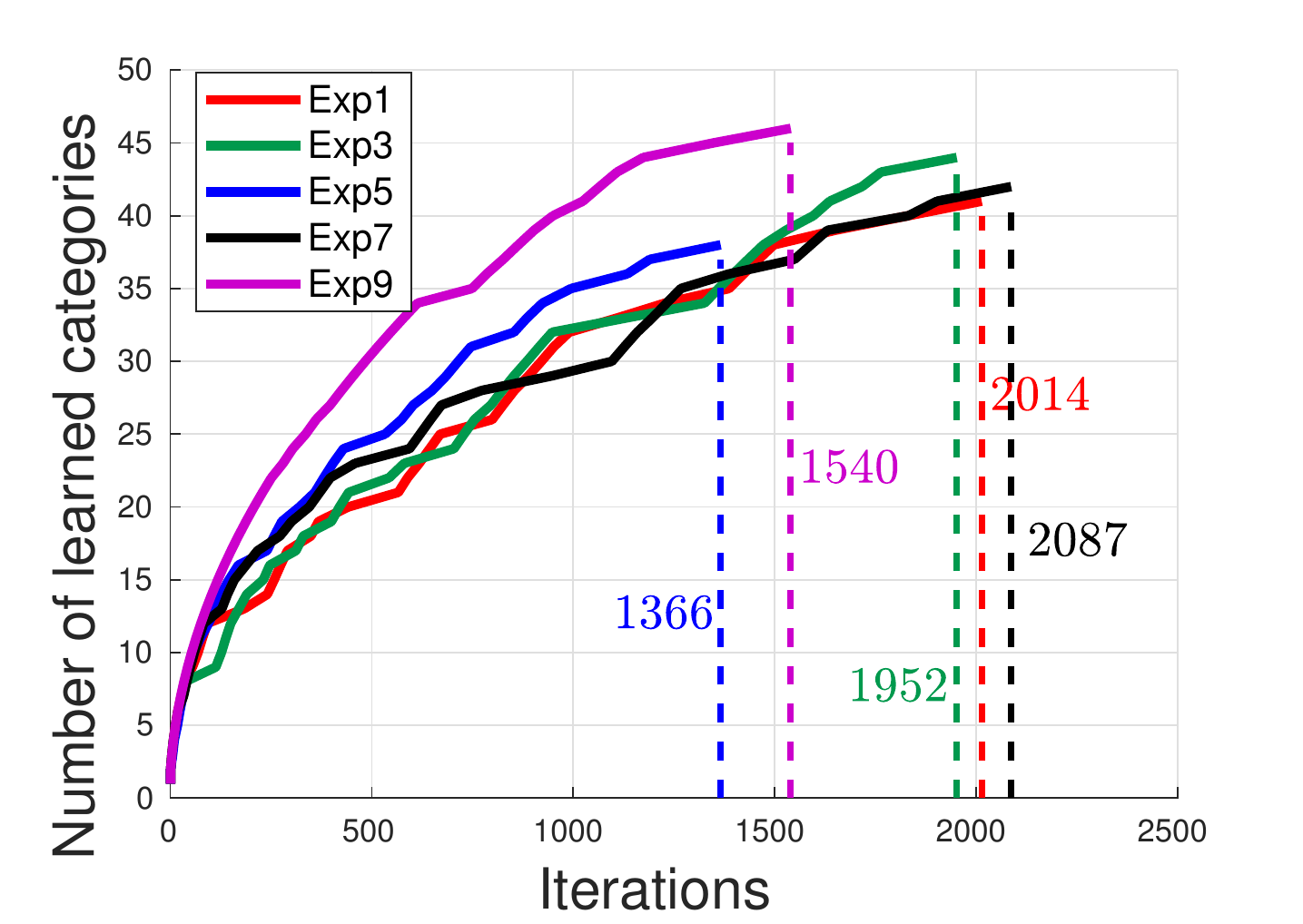}&  \hspace{-3mm}
		\includegraphics[width=0.32\linewidth, trim= 0.35cm 0cm 1cm 0.65cm,clip=true]{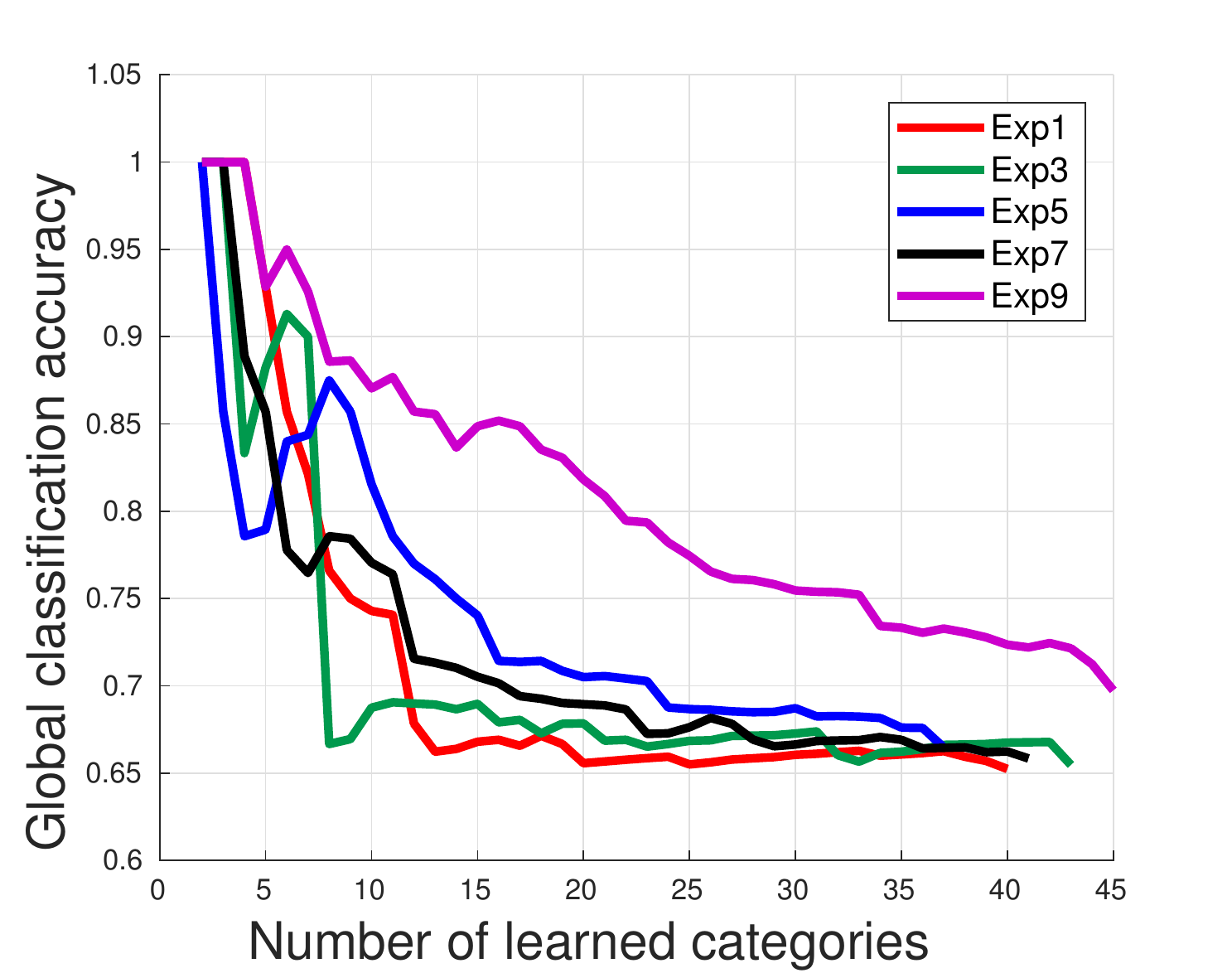}\\
			 \multicolumn{3}{c}{(d) Summary of experiments using GOOD}\vspace{0	mm}	
	\\	
\end{tabular}}
\vspace{-1mm}
\caption{Summary of open-ended evaluations of instance-based approaches without context change.}
\label{fig:open_ended_evaluation}
\end{figure}
The left column in Fig.~\ref{fig:open_ended_evaluation} provides a detailed summary of the obtained results. By comparing all approaches, it is visible that the agent learned (on average) more categories using GOOD than with other approaches. The GOOD approach was able to learn about 39 categories, on average, while the other approaches learned less than 30 categories. In particular, GOOD learned around 11 categories more than Local LDA and 19, 17 and 25 categories more than Approach~II, BoW and standard LDA approaches, respectively. Based on the obtained results, it can be concluded that the agent with the global feature does lead to a better incremental and open-ended performance when compared with the performances of the agent using local-features. Similar results have been reported previously on classical evaluation using instance-based approaches where GOOD led to better category descriptions when compared with the other approaches. It is worth to mention that among the local feature-based approaches,  Local LDA very clearly outperforms the other three approaches. Although, Approach~II and BoW approaches stored fewer instances per category (AIC), on average, than Local LDA, their discriminative power is lower and their performance dropped quickly as the number of categories increased (see the right column of Fig.~\ref{fig:open_ended_evaluation}). It was also observed that the discriminative power of shared topics depends a lot on the order of introduction of categories.

The center column of Fig.~\ref{fig:open_ended_evaluation} illustrates how fast the learning occurred in each of the experiments. It shows the number of question/correction iterations required to learn a certain number of categories. From Fig.~\ref{fig:open_ended_evaluation}, we see that on overage the longest experiments were observed with GOOD and the shortest ones were observed with standard LDA. In the case of standard LDA, the agent on average learned $14.40$ categories using $262.60$ question/correction iterations. GOOD on average continued for $1659.20$ question/correction iterations and the agent was able to learn $39.20$ categories. By comparing all the experiments based on GOOD, it is visible that in the seventh experiment, the number of iterations required to learn 41 object categories was greater than other experiments. The maximum number of categories that the agent could learn with GOOD was 45 categories (i.e., Exp. \#9). This experiment took 1540 question/correction iterations for the agent to acquire these categories. It can be observed that both evaluation measures (i.e., GCA and APA) of this experiment are also higher than the other experiments.  

The right column of the Fig.~\ref{fig:open_ended_evaluation} shows the global classification accuracy obtained by the selected approaches as a function of the number of learned categories. One important observation is that accuracy decreases in all approaches as more categories are introduced. This is expected since a higher number of categories known by the system tends to make the classification task more difficult. BoW and Local LDA achieved the best accuracies with stable performance. One important observation is that
BoW achieved better APA than GOOD and Local LDA approaches. This is expected since BoW learned fewer categories, and it is easier to get better APA in fewer categories.
By comparing all experiments, it is visible that the agent learned more categories when using GOOD while in the case of other metrics, including accuracies (but
this is inversely related to the number of learned categories), memory usage and computation time,  its performance is not as good as Local LDA. The Local LDA provides an appropriate balance between all critical parameters. 

%%%%%%%%%%%%%%%%%%%%%%%%%%%%%%%%%%%%%%%%%%%%%%%%%%%%%%%%%%%%%%%%%%%%%%%%%%%%%%%%%%%%%%%%%%%%%%%%%
\subsection{Evaluation of Model-Based Approaches}

In these experiments, the performance and scalability of the proposed model-based approaches (i.e., naive Bayes with different fixed-size representations), with respect to an increasing number of categories were evaluated. Results are presented in Fig.~\ref{fig:open_ended_evaluation_without_context_NB}. One important observation is that the agent learned all $49$ categories using GOOD and Local LDA in all experiments and all experiments concluded prematurely due to the ``\emph{Lack of data}'', i.e., no more categories available in the dataset  (indicating the potential for learning many more categories). The agent with BoW obtained an acceptable scalability (i.e., the agent on average learned $47.50$ categories) while the scalability of  standard LDA was very low (i.e., on average learned $31$ categories). It should be noted that 8 out of 10 BoW experiments were finished because no more categories were available to be learned (``\emph{lack of data}''). 

\begin{figure*}[!t]
\resizebox{\columnwidth}{!}{
\begin{tabular}{ccc}
	\includegraphics[width=0.37\linewidth,  trim= 0cm -1cm 0.0cm 0cm,clip=true]{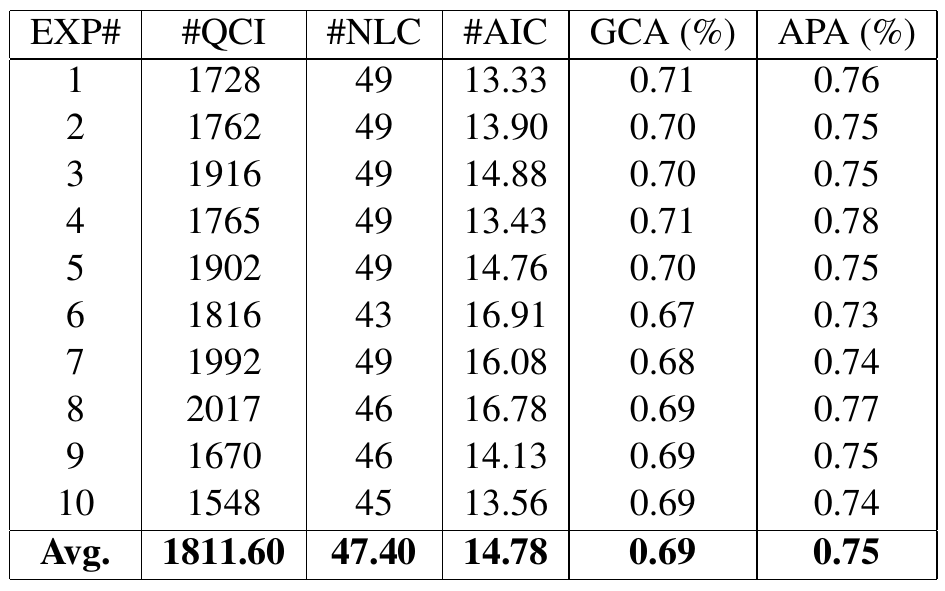}& \hspace{-0mm}
	\includegraphics[width=0.37\linewidth, trim= 0.5cm 0cm 0cm 0.65cm,clip=true]{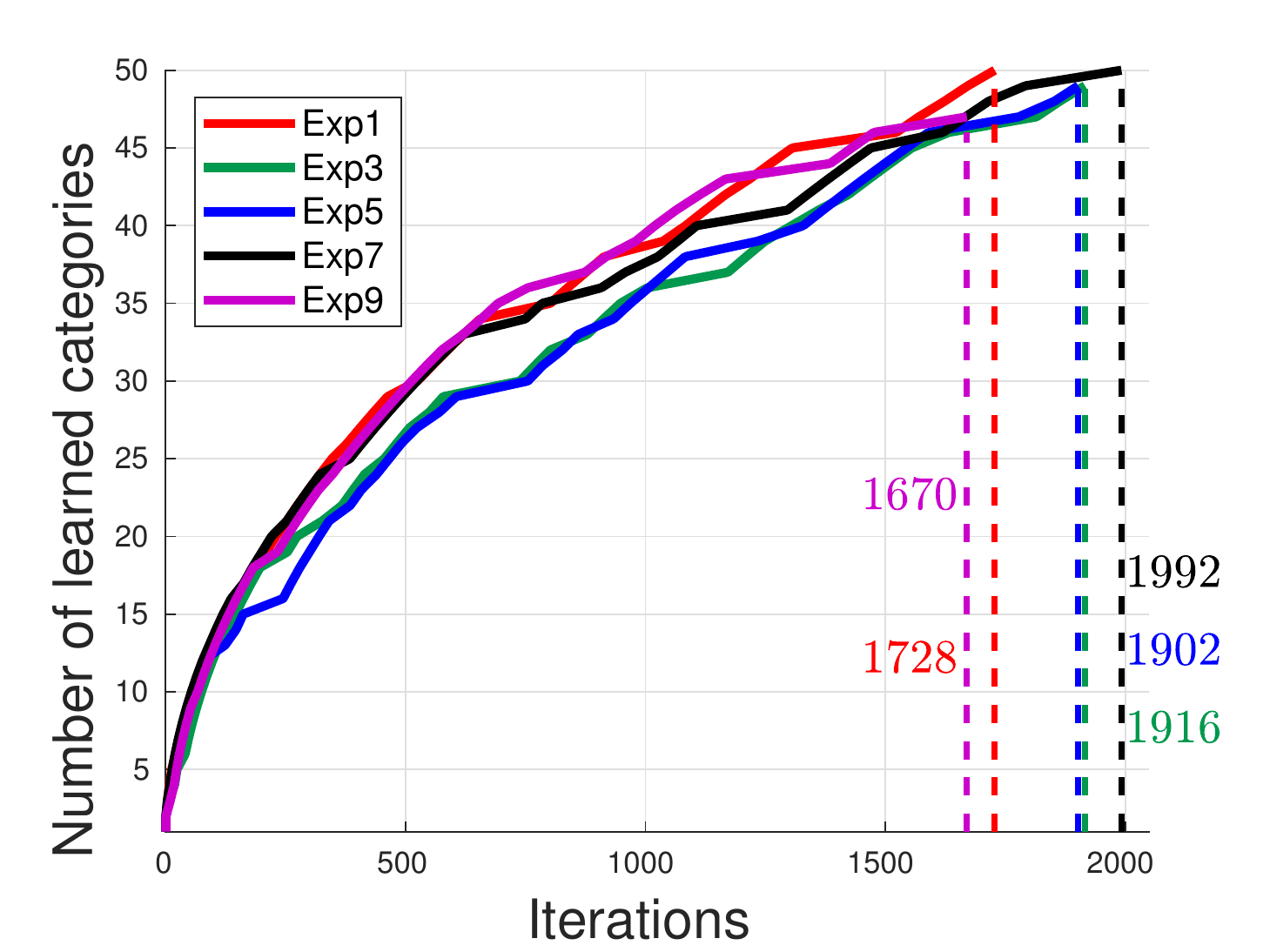}& \hspace{-0mm}
	\includegraphics[width=0.35\linewidth, trim= 0.25cm 0cm 1cm 0.65cm,clip=true]{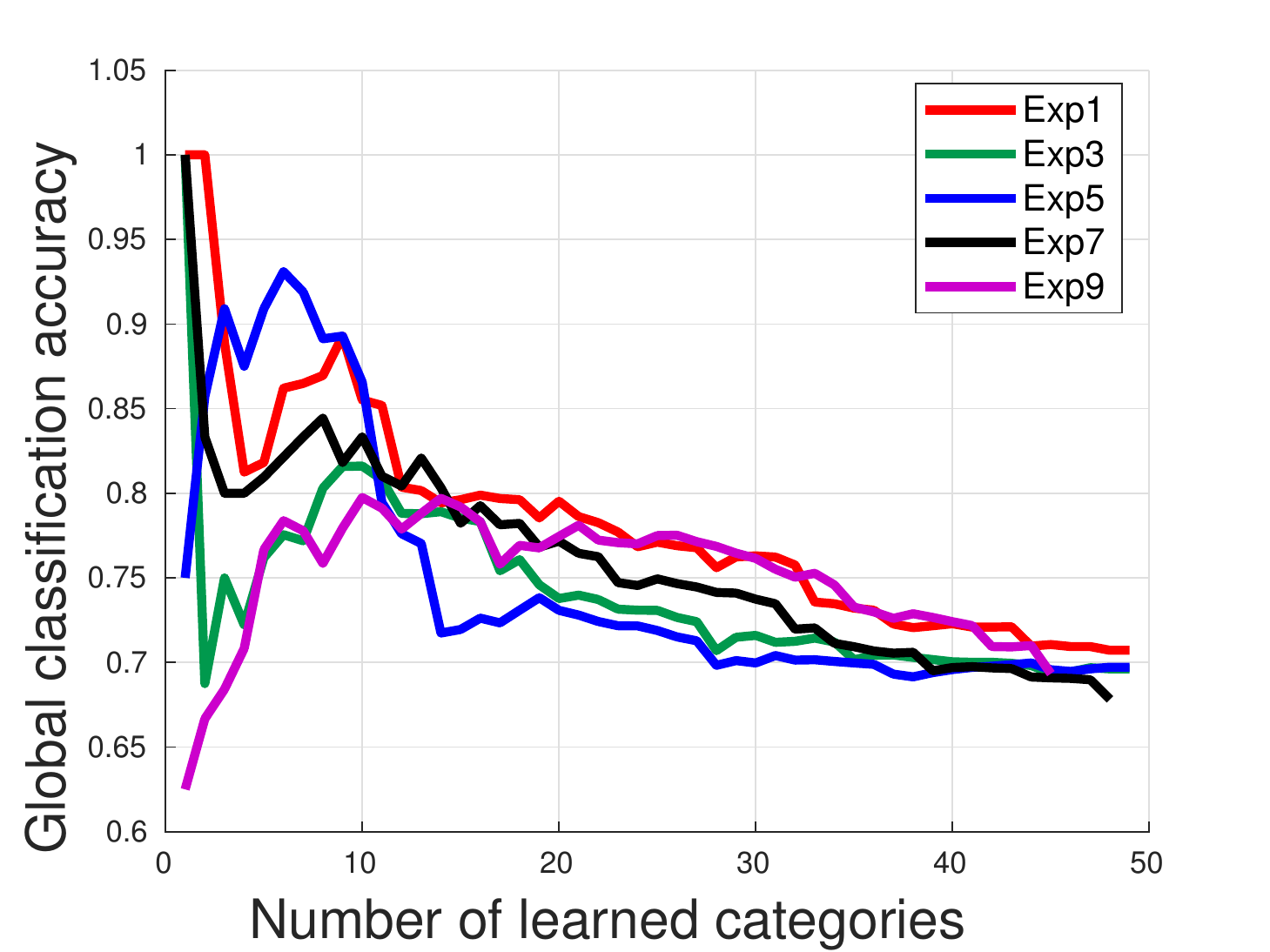}	\vspace{0mm}\\
	 \multicolumn{3}{c}{ (\emph{a}) Summary of experiments using BoW }\vspace{2mm}
	\\ 
	\includegraphics[width=0.37\linewidth,  trim= 0cm -1cm 0cm 0cm,clip=true]{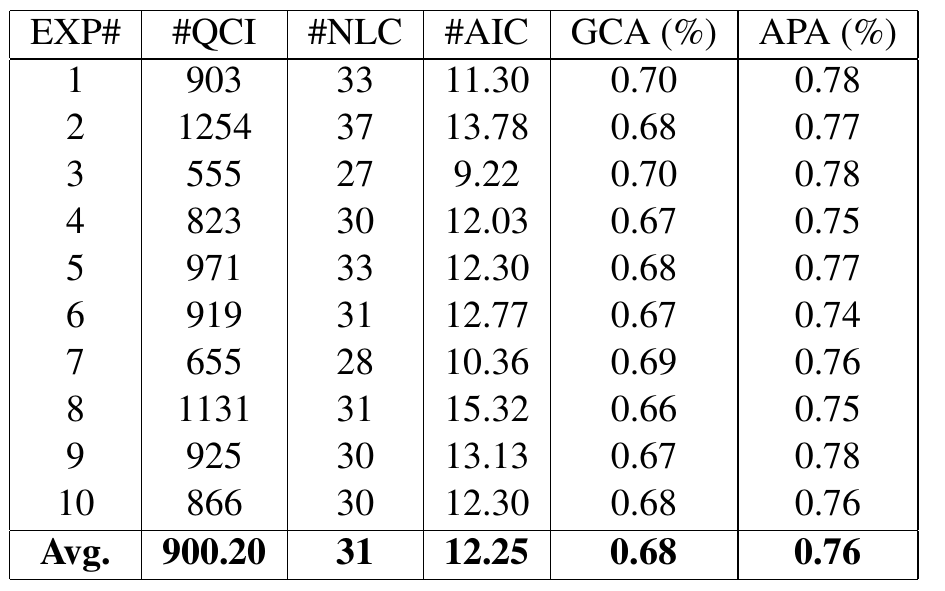}&\hspace{-0mm}
	\includegraphics[width=0.37\linewidth, trim= 0.5cm 0cm 0cm 0.65cm,clip=true]{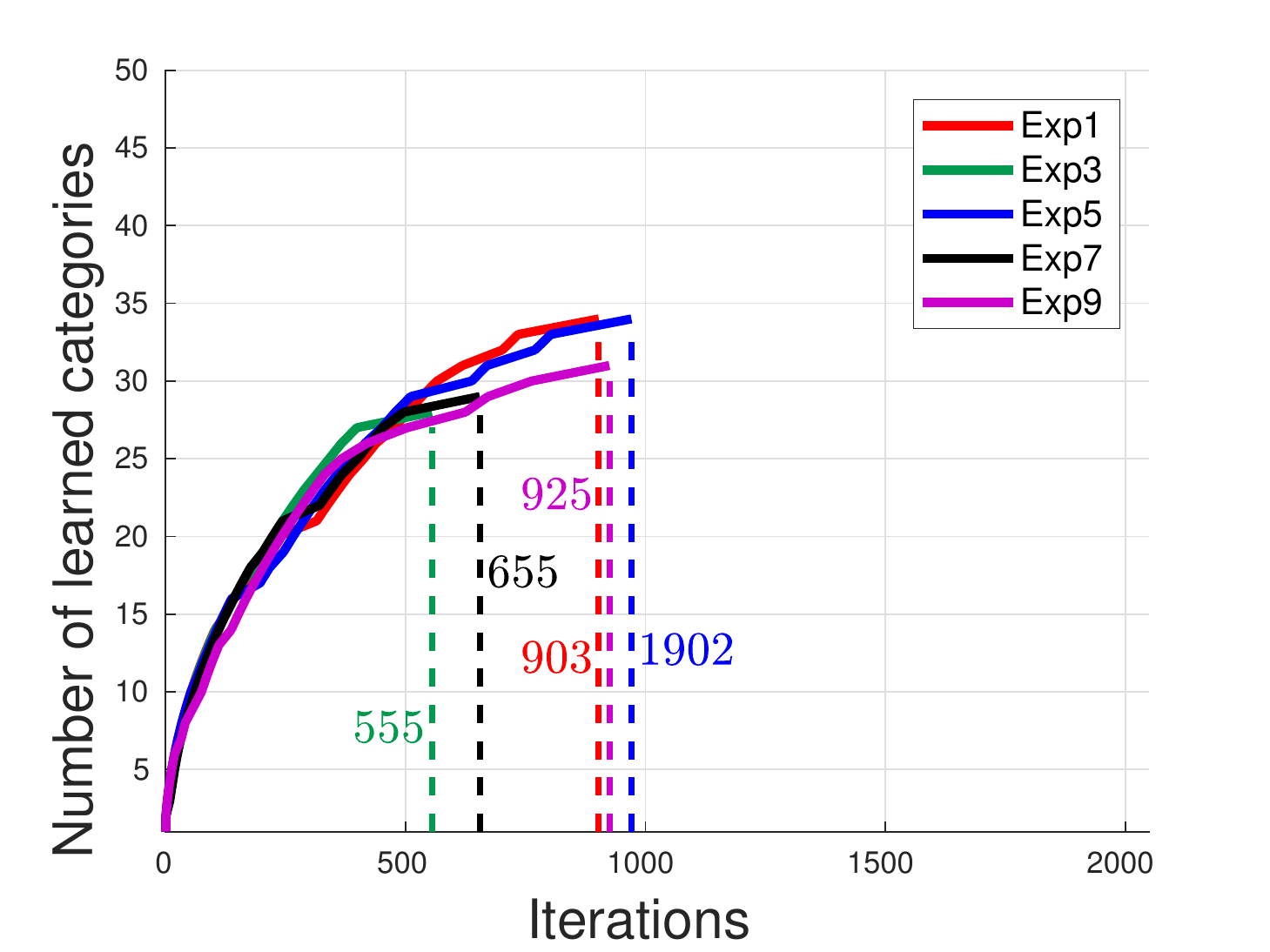}& \hspace{-0mm}
	\includegraphics[width=0.35\linewidth, trim= 0.25cm 0cm 1cm 0.65cm,clip=true]{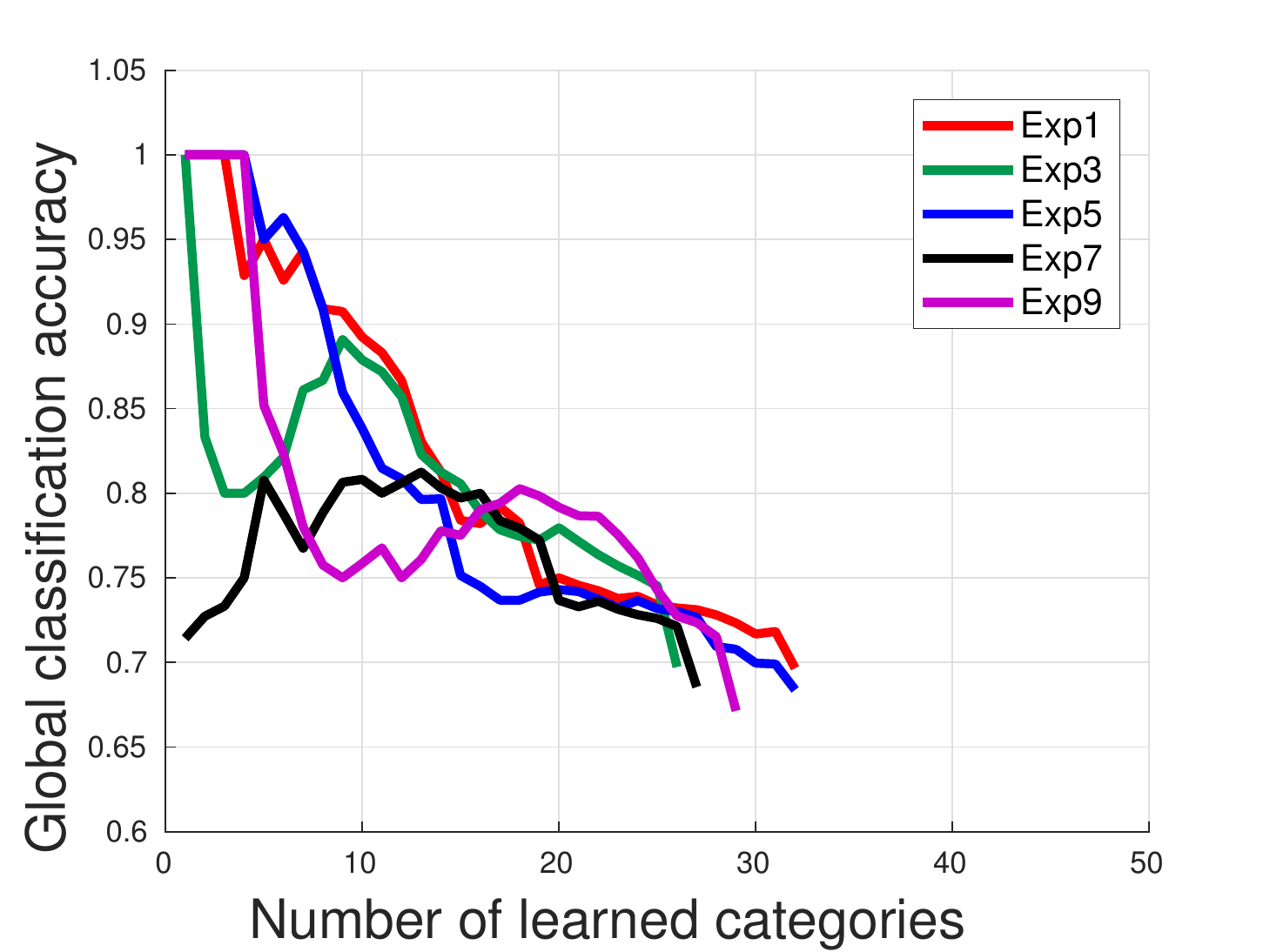}	\vspace{0mm}\\
	\multicolumn{3}{c}{(\emph{b}) Summary of experiments using standard LDA}\vspace{2mm}
	\\
	\includegraphics[width=0.38\linewidth,  trim= 0cm -1cm 0cm 0cm,clip=true]{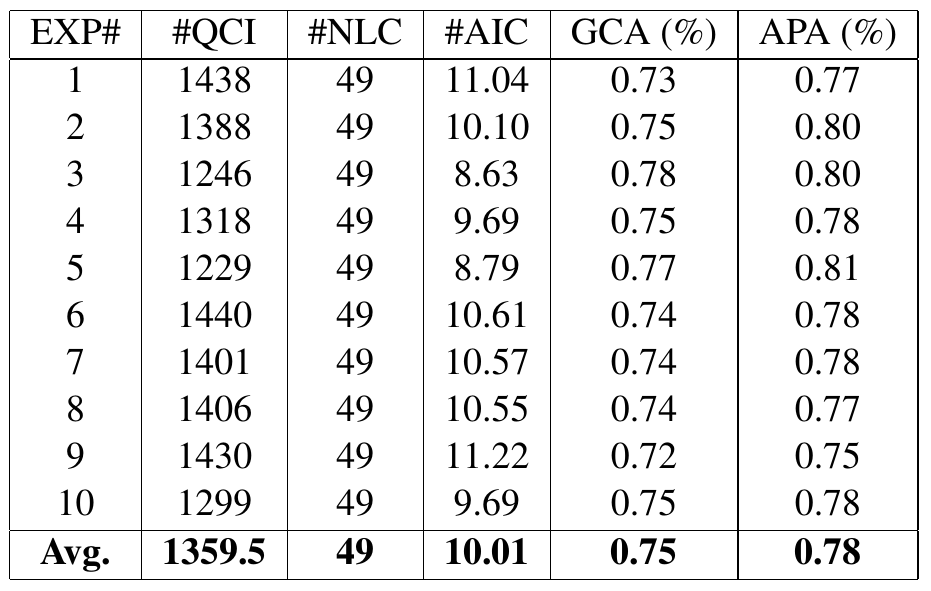}&\hspace{-0mm}
	\includegraphics[width=0.385\linewidth, trim= 0.5cm 0cm 0cm 0.65cm,clip=true]{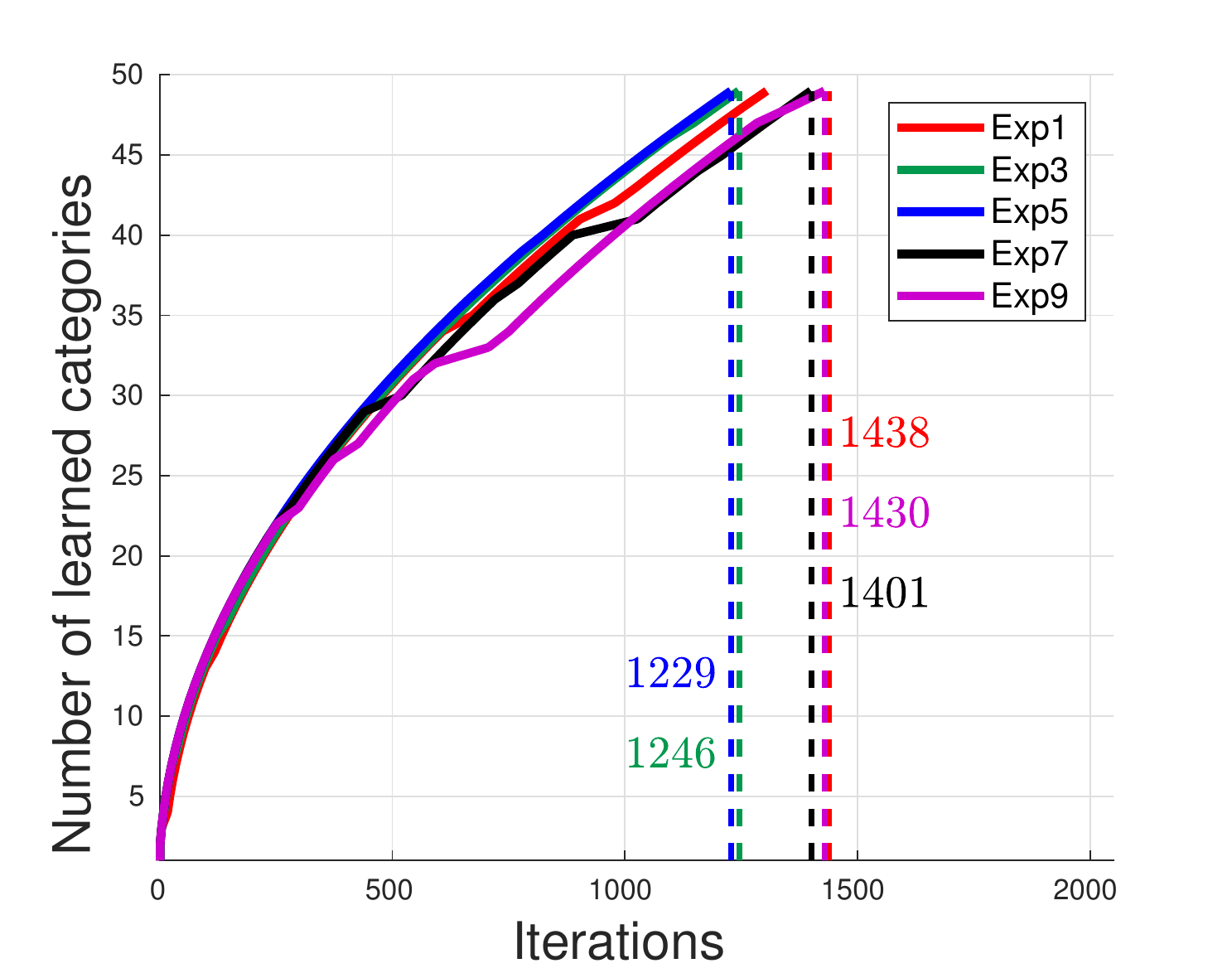}& \hspace{-0mm}
	\includegraphics[width=0.37\linewidth, trim= 0.25cm 0cm 1cm 0.65cm,clip=true]{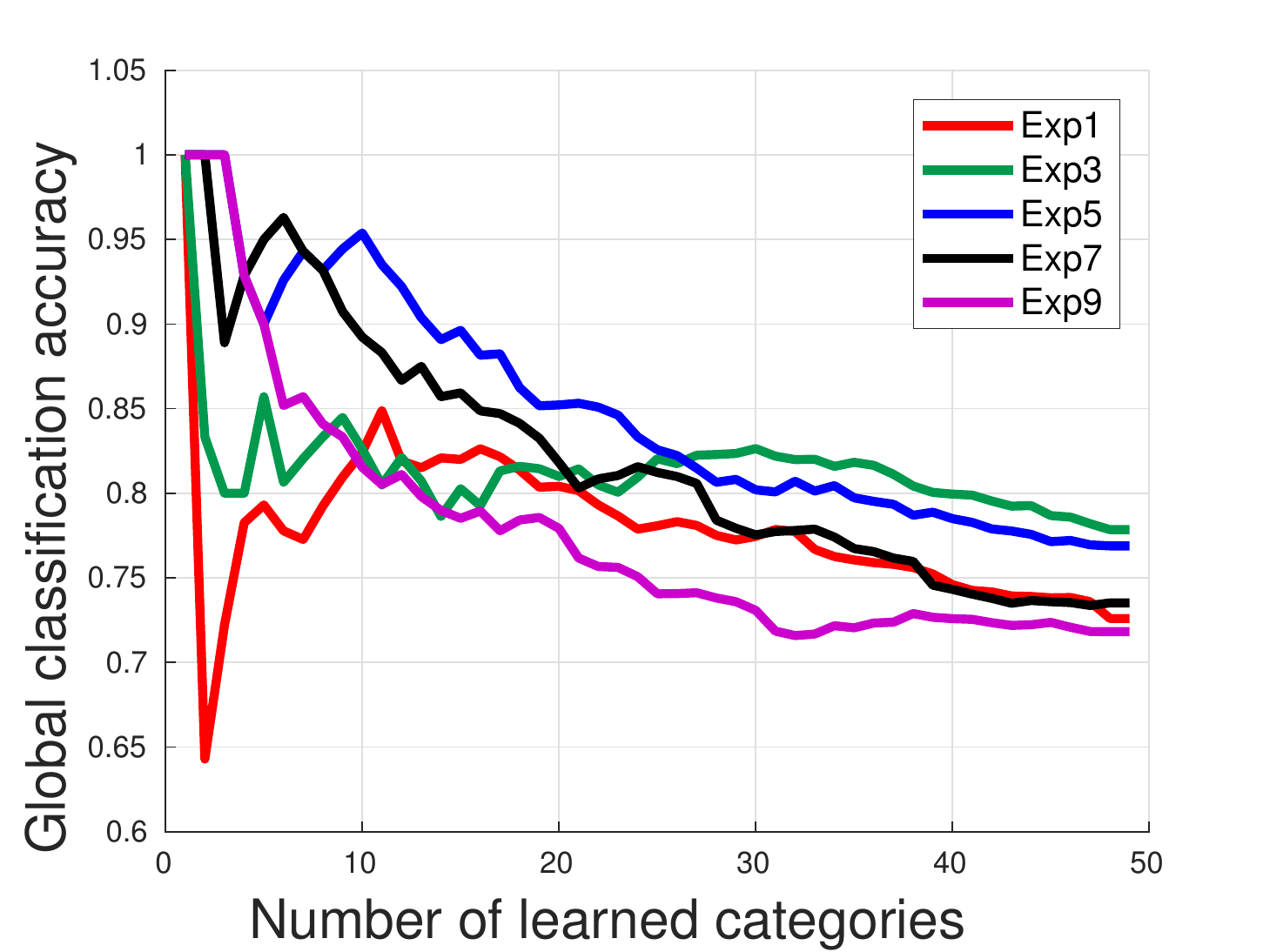}	\vspace{0mm}\\
	\multicolumn{3}{c}{(\emph{c}) Summary of experiments using Local LDA}\vspace{2mm}
	\\
	\includegraphics[width=0.38\linewidth,  trim= 0cm -1cm 0cm 0cm,clip=true]{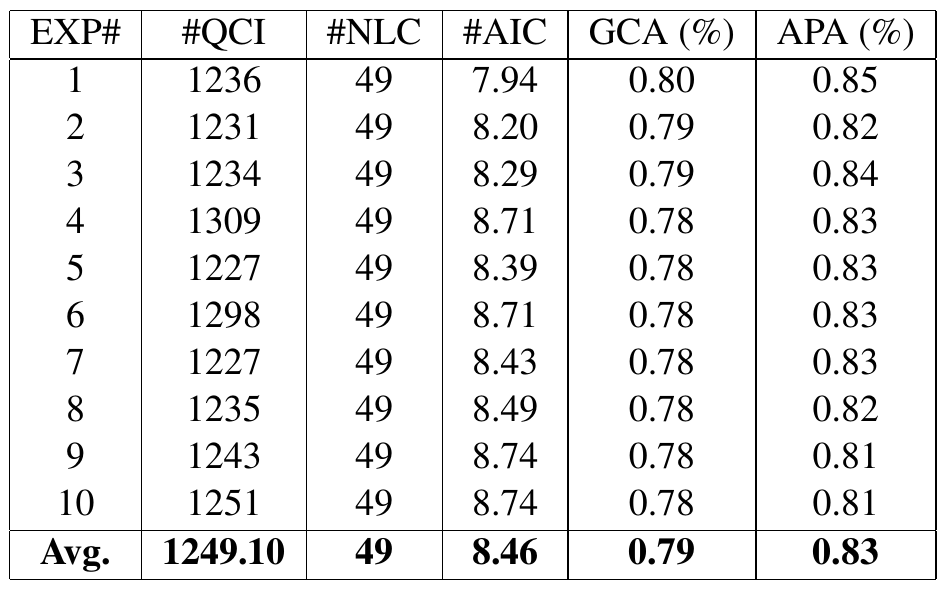}&\hspace{-0mm}
	\includegraphics[width=0.385\linewidth, trim= 0.5cm 0cm 0cm 0.65cm,clip=true]{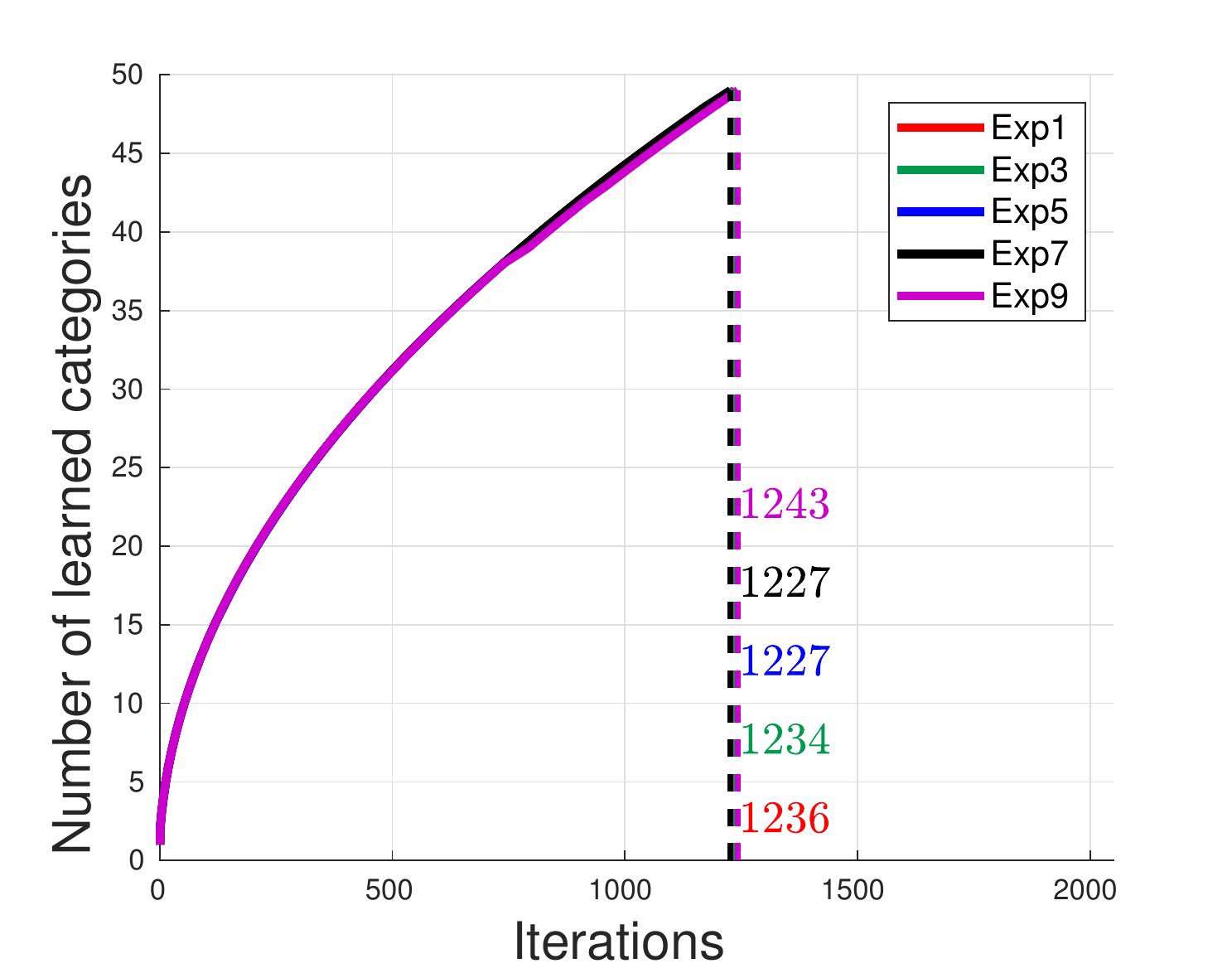}& \hspace{-0mm}
	\includegraphics[width=0.37\linewidth, trim= 0.25cm 0cm 1cm 0.65cm,clip=true]{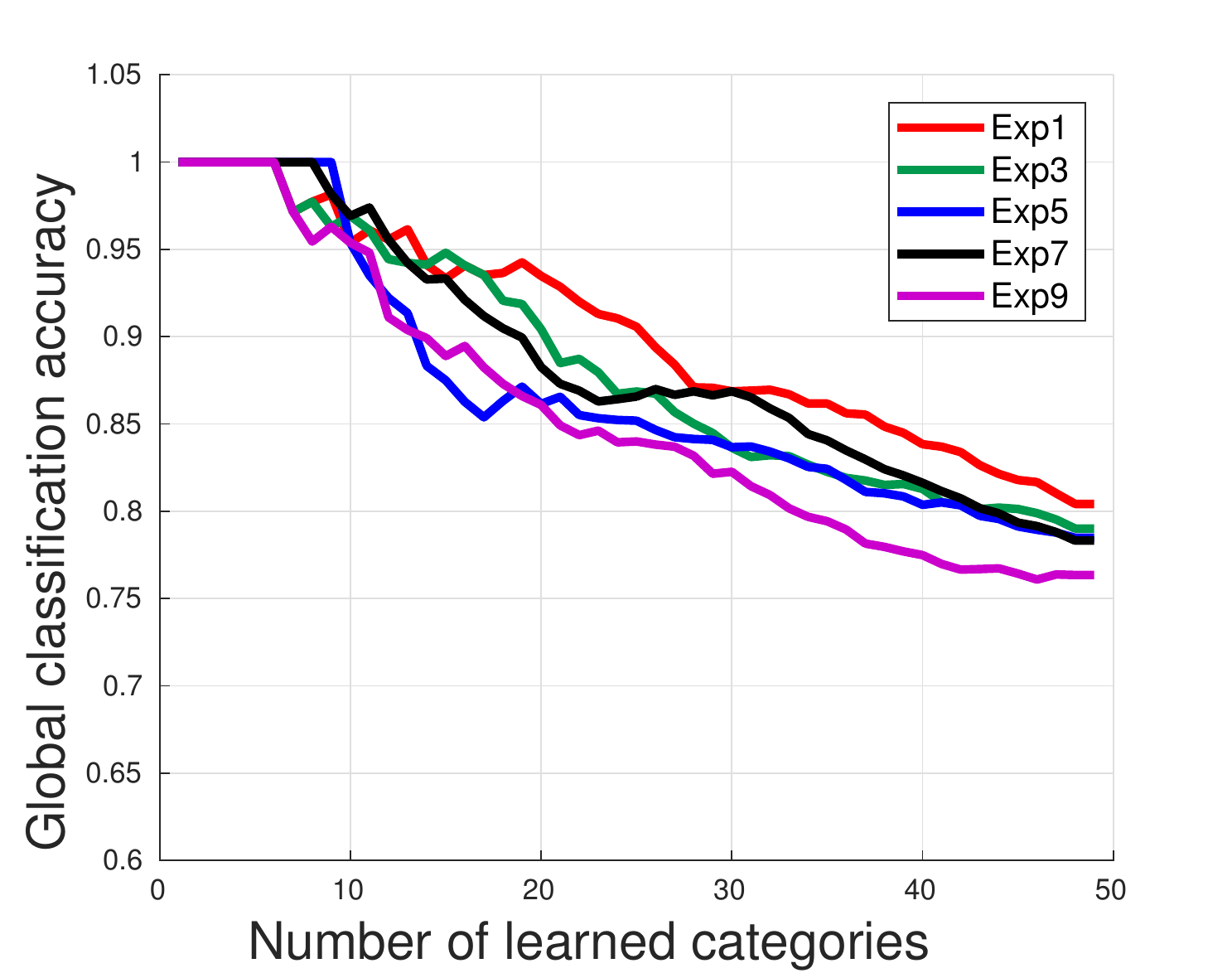}	\vspace{0mm}\\
	\multicolumn{3}{c}{(\emph{d}) Summary of experiments using GOOD}\vspace{2mm}
	\\

\end{tabular}}
\vspace{-2mm}
\caption{Summary of open-ended evaluations of model-based approaches without context change.}
\label{fig:open_ended_evaluation_without_context_NB}
\end{figure*}

The left column in Fig.~\ref{fig:open_ended_evaluation_without_context_NB} provides a detailed summary of the obtained results. %By comparing all approaches, it is clear that the agent with GOOD approach stored fewer instances per category (AIC) than the other approaches.  In particular, the agent with GOOD feature stored on average $8.46$ instances per category which is around $1.55$, $3.79$ and $6.32$ instances less than Local LDA, BoW and LDA approaches respectively. 
The center column of Fig.~\ref{fig:open_ended_evaluation_without_context_NB} shows the number of learned categories as a function of the number of protocol iterations. This gives a measure of \emph{how fast} the learning occurred in each of the experiments. Moreover, it shows the number of \emph{question}/\emph{correction} iterations required to learn a certain number of categories. It can be concluded that the agent with GOOD learned all categories faster than with Local LDA. The agent with BoW and LDA achieved the third and forth places respectively. 

The right column of the Fig.~\ref{fig:open_ended_evaluation_without_context_NB} shows the global classification accuracy obtained by the proposed approaches as a function of the number of learned categories. By comparing all approaches, it is visible that the agent with GOOD achieved the best accuracy (i.e., $79\%$) with stable performance and outperformed the other approaches by a large margin (i.e., 4\% or more). The agent with Local LDA also showed a promising performance and provide a good balance among all parameters.  Although, BoW and standard LDA approaches on average achieved similar accuracies, the discriminative power of standard LDA is lower than BoW and its performance
dropped quickly as the number of categories increased (see the right column of Fig.~\ref{fig:open_ended_evaluation_without_context_NB} (\emph{a} and \emph{b})). 

The average protocol accuracy of the agent with GOOD feature is also considerably higher than the other approaches (i.e., more than $5\%$). It should be noted that these results should be seen in the light of the number of categories learned. For example, BoW and LDA seem to indicate similar average protocol accuracy (APA), however, standard LDA on average reached the breakpoint after the introduction of the 31st category whereas BoW learned around all (49) categories in six experiments and between 43 and 46 categories
in the remaining four experiments.

%%%%%%%%%%%%%%%%%%%%%%%%%%%%%%%%%%%%%%%%%%%%%%%%%%%%%%%%%%%%%%%%%%%%%%%%%%%%%%%%%%%%%%%%%%%%%

\section {Evaluation of Adaptability to Context Change}
\label{sec:evaluation_of_adaptability_to_context_change}
In this section, a methodology is proposed and used for evaluation learning agents in open-ended learning scenarios with context change. This includes a new teaching protocol and an adaptability measure.
%%%%%%%%%%%%%%%%%%%%%%%%%%%%%%%%%%%%%%%%%%%%%%%%%%%%%%%%%%%%%%%%%%%%%%%%%%%%%%%%%%%%%%%%%%%%%

\subsection {Open-Ended Evaluation with Context Change}

In order to evaluate the adaptability to context change, we modified the standard teaching protocol described above to include a change of context. A simulated teacher is developed to follow the teaching protocol and autonomously interact with the system using \emph{teach}, \emph{ask} and \emph{correct} actions. The main idea is to emulate the interactions of an agent with the environment over long periods of time in two different contexts. Towards this end, the object categories in the database and respective object views are randomly assigned to two different contexts, A and B. Then, the teacher starts presenting categories from context A. The simulated teacher repeatedly picks object views of the currently known categories from the context A and presents them to the system for checking whether the system can recognize them. If not, the simulated teacher provides corrective feedback. Whenever the agent learned $\rho$ categories from context A, the simulated teacher changes to the other context, B, and interacts (i.e., teach, ask and correct) with the learning agent using the object categories of context B. The complete process is summarized in Algorithm 7.2.

In this way, the agent begins with zero knowledge and the training instances become gradually available according to the teaching protocol. After context change, the learning agent will recognize object views taking into account all acquired category models from both contexts. Note that, although, a learning agent can have internal mechanisms specifically designed for keeping track of context (explicit context information, either inferred by the learning agent or obtained from an external source), no such mechanism was developed in this work. Therefore, what was evaluated was the potential of the different representations and recognition rules for implicitly coping with context change.
%\begin{figure}[!t]
%\vspace{-3mm}
%\centering
%\includegraphics[width=0.85\linewidth]{Figures/algorithm.pdf}
%\vspace{-2mm}
%\label{fig:simulated_teacher_algorithm}
%\end{figure}

\begin{algorithm}[!t]
\SetAlgoRefName{7.2} 
\caption*{ Teaching protocol with context change}
\label{alg:tp}
\begin{algorithmic}[1]
\State \textbf{Input:} $\rho$ 
\State \footnotesize{\emph{context} $\gets A$}
\State \textbf{Introduce} $\operatorname{category}_1$ from \emph{context}
\State $n \gets \textit{1}$
\State  \textbf{repeat} 
\State \quad\quad\textbf{if} { \emph{context} $ == A$} \textbf{and} $n~>~\rho$ \textbf{then} \Comment {{\scriptsize context switch}}
\State \quad\quad\quad\quad \emph{context} $\gets B$
\State \quad\quad$n\gets \textit{n+1}$ \Comment {{\scriptsize ready for the next category}}
\State \quad\quad\textbf{Introduce} $\operatorname{category}_n$ from \emph{context}
\State \quad\quad $k\gets \textit{0}$
\State \quad\quad $c \gets \textit{1}$
\State \quad\quad\textbf{repeat} \Comment {{\scriptsize question / correction iteration}}
\State \quad\quad\quad\quad \textbf{if} (\textbf{inContext} ($\operatorname{category}_c$, \emph{context})) \textbf{then} 
\State \quad\quad\quad\quad\quad\quad Present a previously unseen instance of $\operatorname{category}_c$
\State \quad\quad\quad\quad\quad\quad \textbf{Ask} the category of this instance
\State \quad\quad\quad\quad\quad\quad If needed, provide \textbf{correct} feedback

\State \quad\quad\quad\quad c {$\gets$ c+1 \textbf{if} $\textit{c < n}$ \textbf{else} ~$1$}

\State \quad\quad\quad\quad \textbf{if} ($k \le 3n$) \textbf{then}  \Comment {{\scriptsize sliding window}}
\State \quad\quad\quad\quad\quad\quad $k\gets \textit{k+1}$
\State \quad\quad\quad\quad $s\gets$ accuracy in last $k$ question/correction iterations
\State \quad\quad\textbf{until (}($s$ > $\tau$ \textbf{and} $k \ge n$) \textbf{or}  \Comment {{\scriptsize accuracy threshold crossed}}
\State \quad\quad\quad\quad\quad (user sees no improvement)\textbf{)} \Comment {{\scriptsize breakpoint reached}}
\State  \textbf{until (}user sees no improvement in accuracy\textbf{)} \Comment {{\scriptsize breakpoint reached}}
\end{algorithmic}
\end{algorithm}

%%%%%%%%%%%%%%%%%%%%%%%%%%%%%%%%%%%%%%%%%%%%%%%%%%%%%%%%%%%%%%%%%%%%%%%%%%%%%%%%%%%%%%%%%%%%%%%%%%%%%
\subsection {Measuring the Adaptability to Context Change}

A new \emph{adaptability} metric is proposed here in order to compare the adaptability of different approaches to context change. This metric is intended to be orthogonal to other
metrics, namely accuracy or number learned categories. Adaptability is a measure of relative performance of the learning agent in the second context, B, when compared
with the performance of the same agent in the first context, A. To carry out this comparison in a controlled way, we define, for each learning approach, the size (i.e., number of
categories) of the first context based on the average number of learned categories of that learning approach in a set of standard teaching protocol experiments i.e., without context
change (see subsection \ref{sec:open_ended_evaluation_of_selected_approaches}). To ensure that all categories in context A are learned, enabling the agent to move to context B, the number of categories of context A is set to be on average 0.75 ALC, where ALC is the average number of learned categories in a round of experiments without context change. In order to converge to that average context size, the context transition point, $\rho$, is generated randomly in the interval $[0.65\operatorname{ALC}~,~ 0.85\operatorname{ALC}]$. for each context change experiment. This way, for each learning approach, the context change happens at the same point (i.e., around 0.75 ALC) with respect to the full capacity of that approach as captured by ALC. This enables an evaluation of adaptability orthogonal to the evaluation of learning capacity. With this setup, adaptability is measured by:

\begin{equation}
\label {adaptability}
\operatorname{Adaptability} =  \frac{\operatorname{ALC}_{2}}{\operatorname{ALC}_{1}},
\end{equation}

\noindent 
where $\operatorname{ALC}_1$ and $\operatorname{ALC}_2$ are the average numbers of categories learned in the first and second contexts respectively.
It should be noted that, if the experiment is finished due to ``\emph{lack of data}'', it is not possible to measure adaptability. In order for the adaptability to be comparable across different experiments, it is essential that all experiments end with the same terminating condition (breakpoint).

%%%%%%%%%%%%%%%%%%%%%%%%%%%%%%%%%%%%%%%%%%%%%%%%%%%%%%%%%%%%%%%%%%%%%%%%%%%%%%%%%%%%%%%%%%%%%%%%%%%%%%%%%%%55
\subsection {Instance-Based Approaches}
\label{sec:instance_based_open_ended_evaluation_with_context_change}

\begin{figure}[!t]
\center
\resizebox{0.995\linewidth}{!}{
\begin{tabular}{ccc}
	\includegraphics[width=0.45\linewidth,  trim= 0cm -0.6cm 0.0cm 0cm,clip=true]{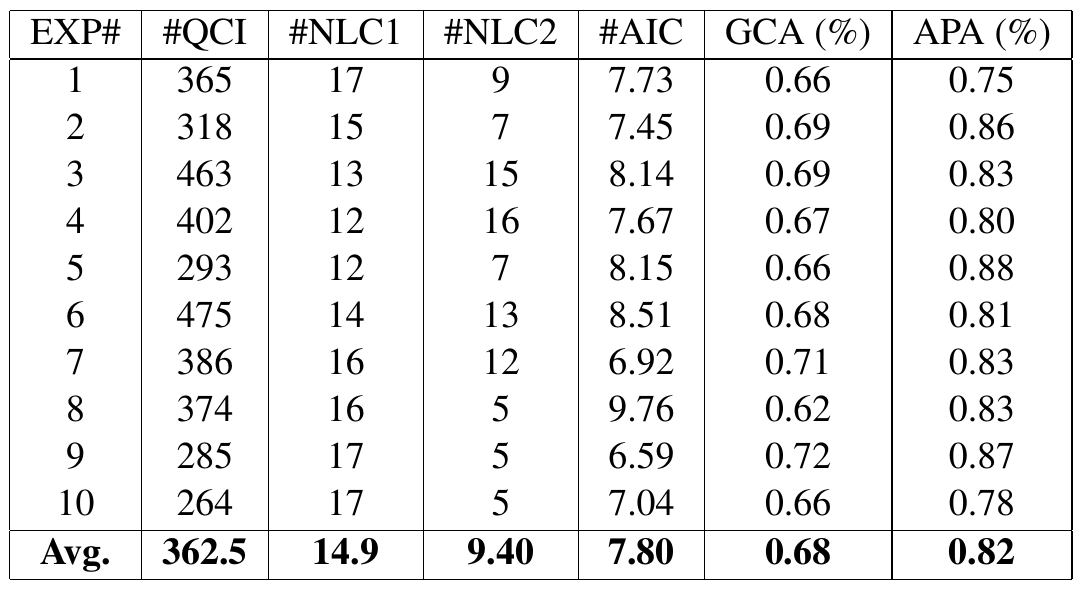}&\hspace{-2mm}
	\includegraphics[width=0.35\linewidth, trim= 0.5cm 0cm 1cm 0.65cm,clip=true]{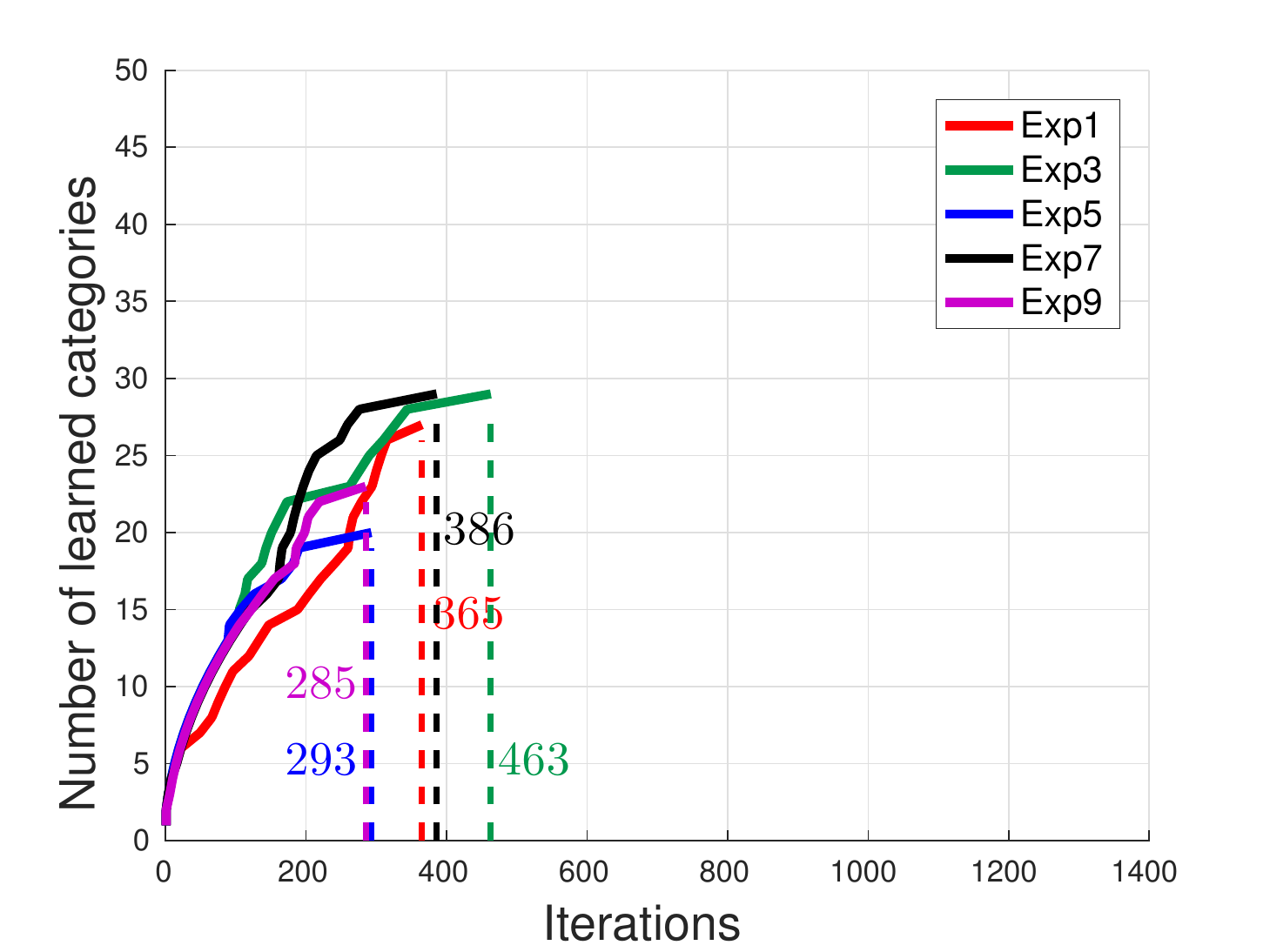}& \hspace{-2mm}
		\includegraphics[width=0.35\linewidth, trim= 0.35cm 0cm 1cm 0.65cm,clip=true]{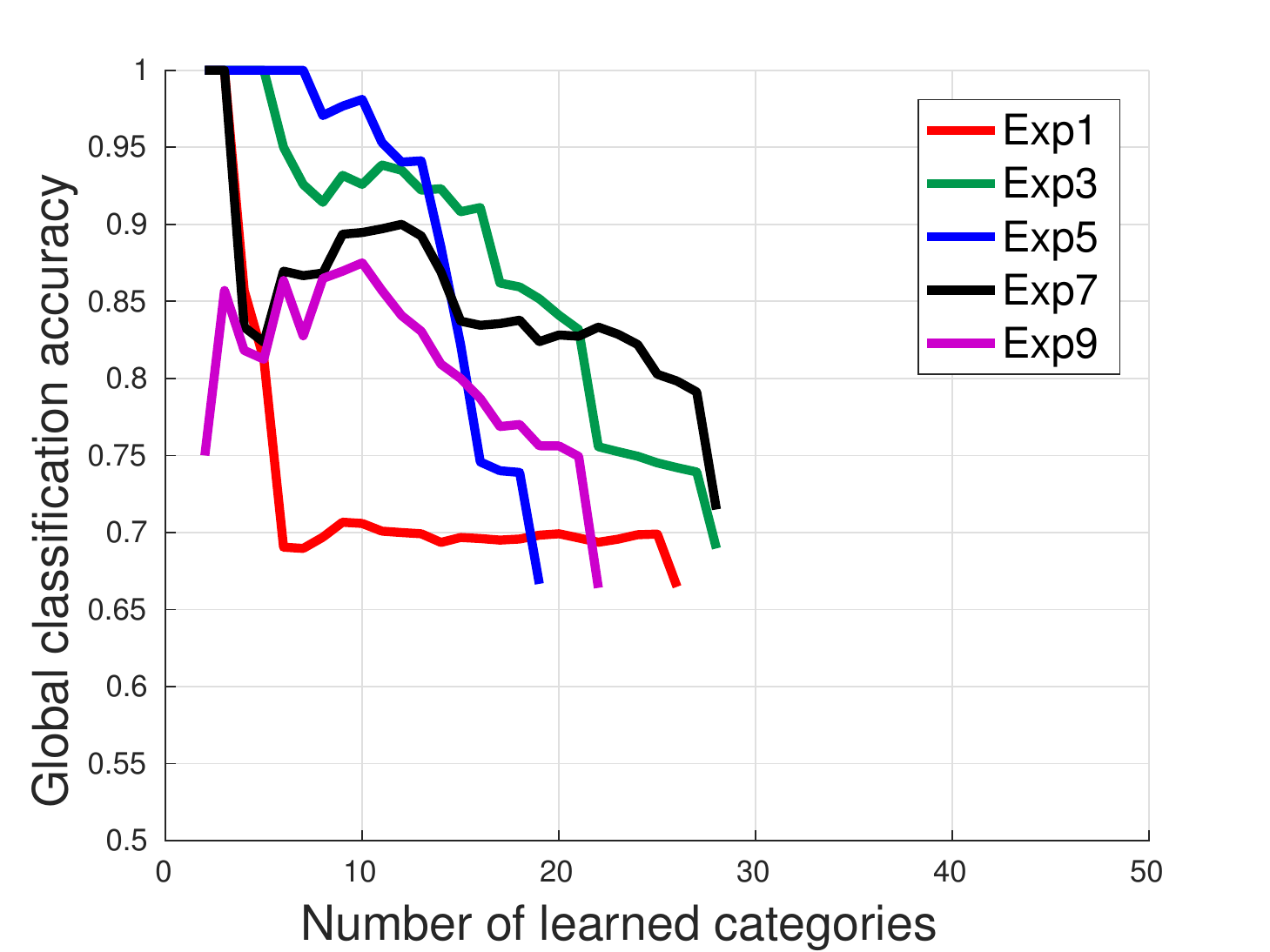}	\vspace{0mm}\\
  \multicolumn{3}{c}{(a) Summary of experiments using sets of local features (Approach~II)} \vspace{5mm}
	\\ 
	\includegraphics[width=0.45\linewidth,  trim= 0cm -0.6cm 0.0cm 0cm,clip=true]{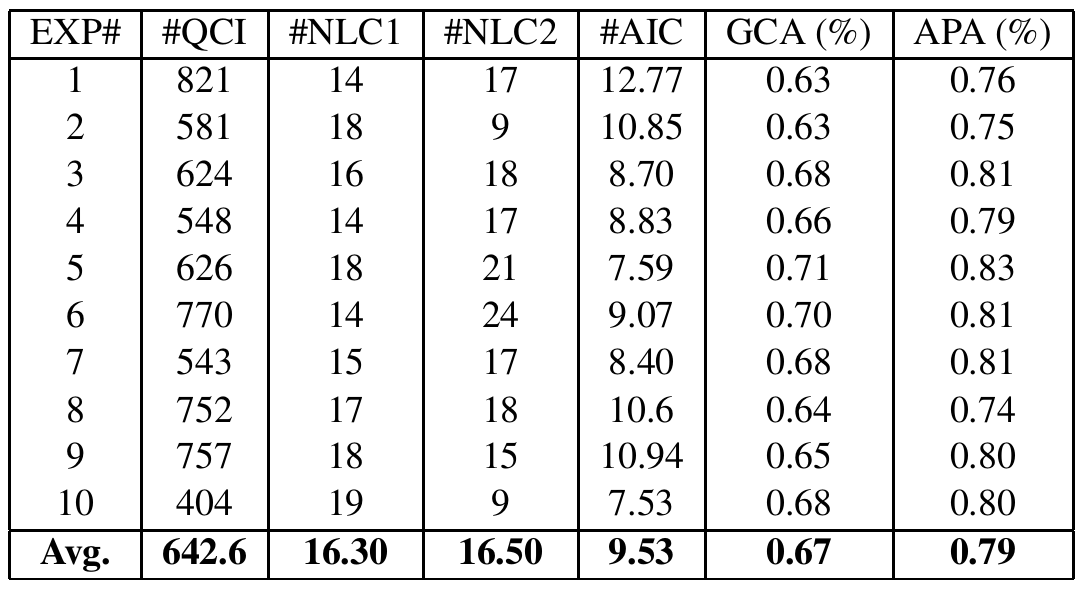}&\hspace{-2mm}
	\includegraphics[width=0.35\linewidth, trim= 0.5cm 0cm 1cm 0.65cm,clip=true]{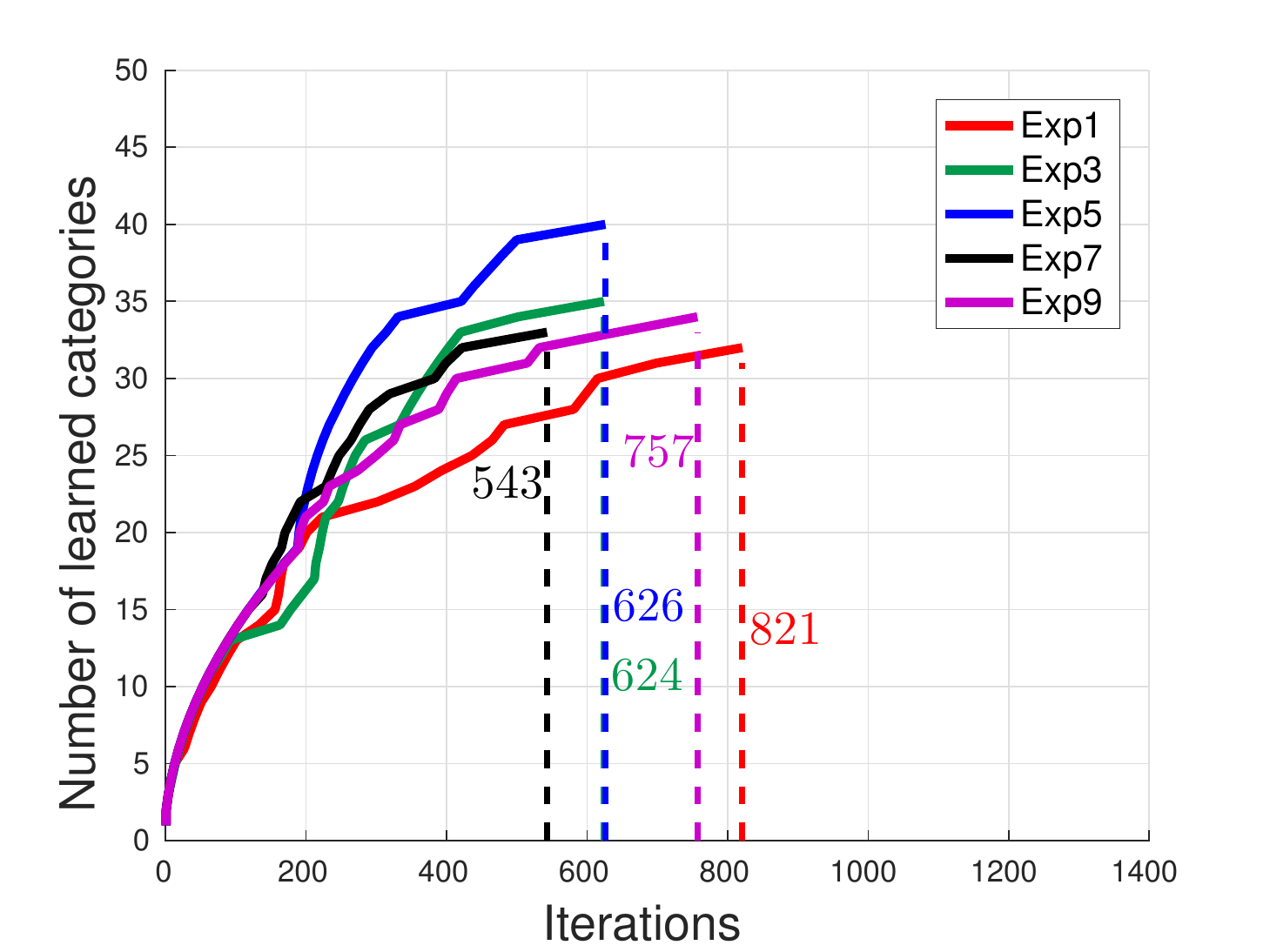}& \hspace{-2mm}
	\includegraphics[width=0.35\linewidth, trim= 0.35cm 0cm 1cm 0.65cm,clip=true]{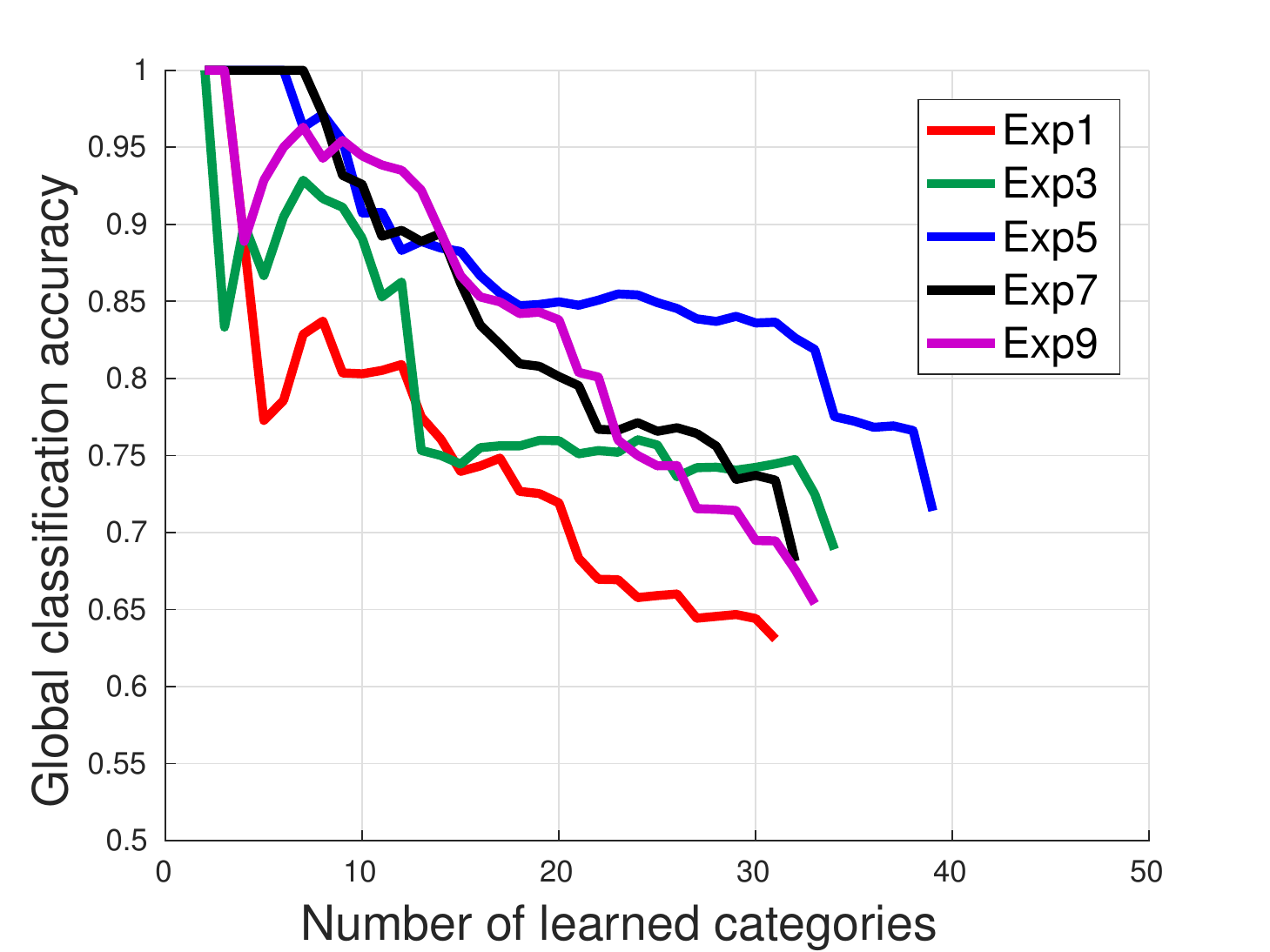}	\vspace{0mm}\\
	 \multicolumn{3}{c}{(b) Summary of experiments using BoW} \vspace{5mm}	
	\\
	\includegraphics[width=0.45\linewidth,  trim= 0cm -0.6cm 0.0cm 0cm,clip=true]{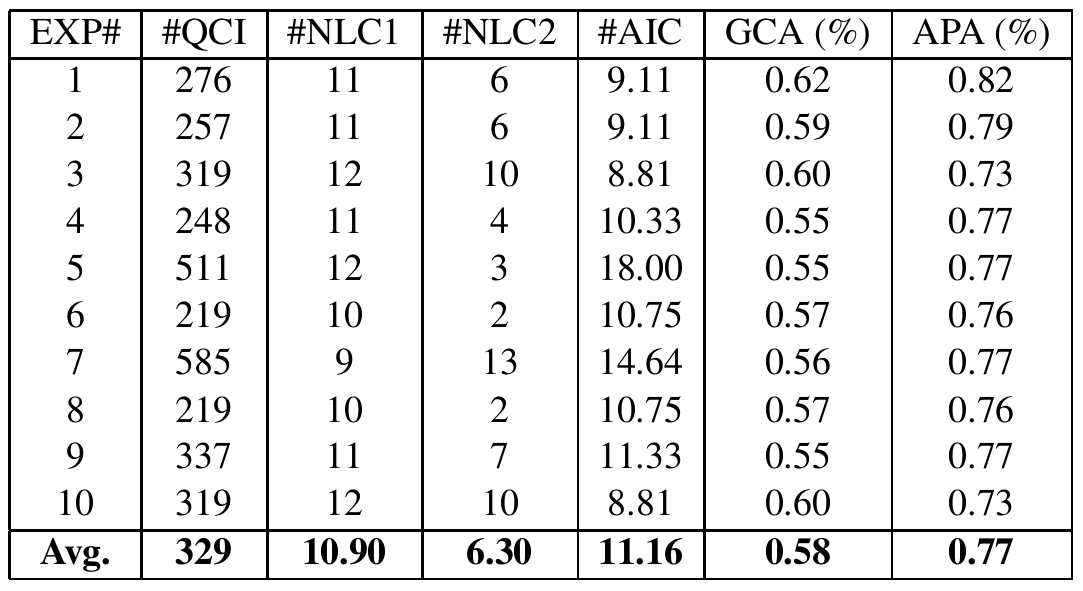}&\hspace{-2mm}
	\includegraphics[width=0.35\linewidth, trim= 0.5cm 0cm 1cm 0.65cm,clip=true]{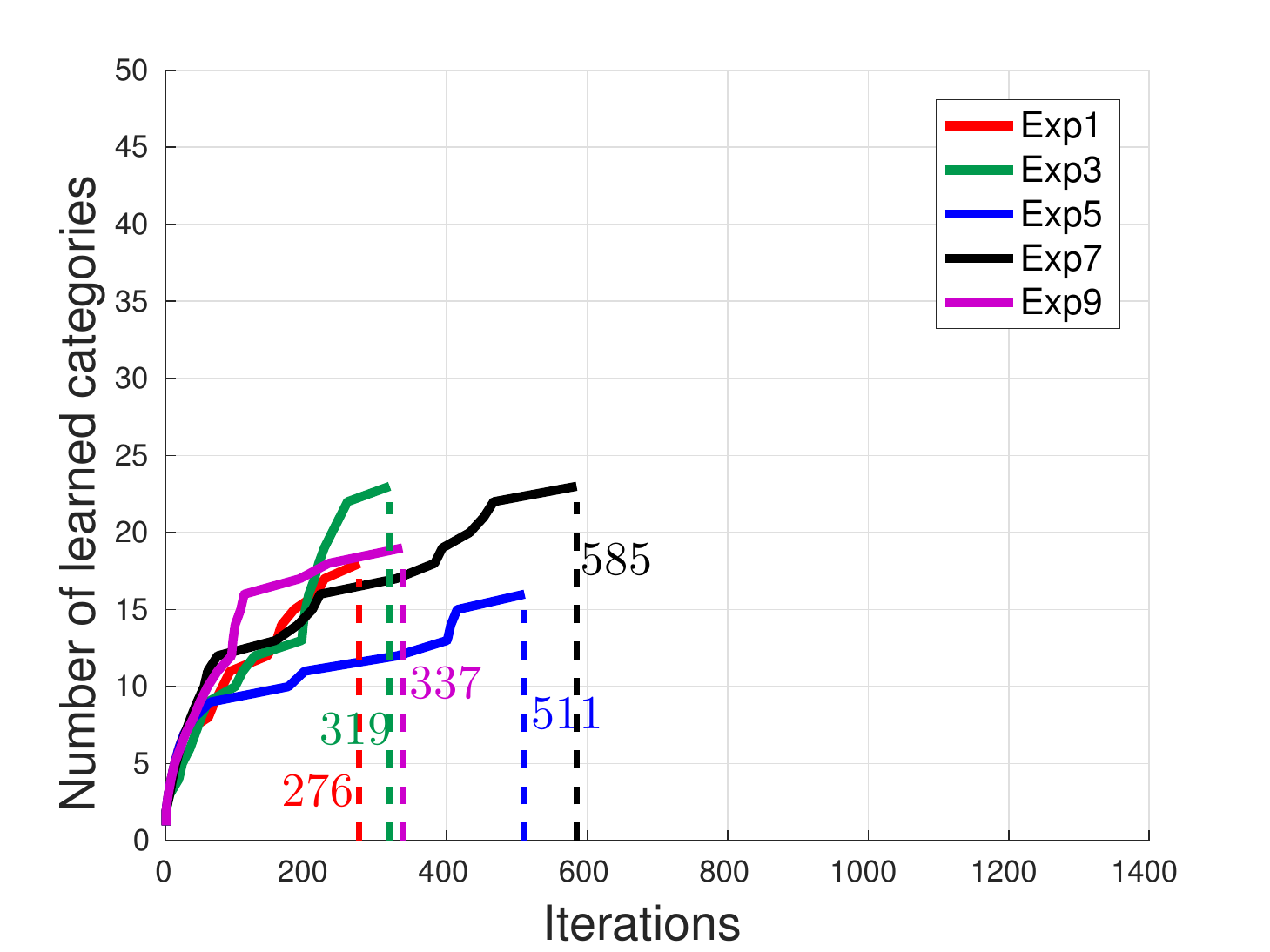}& \hspace{-2mm}
	\includegraphics[width=0.35\linewidth, trim= 0.35cm -0.5cm 1cm 0.65cm,clip=true]{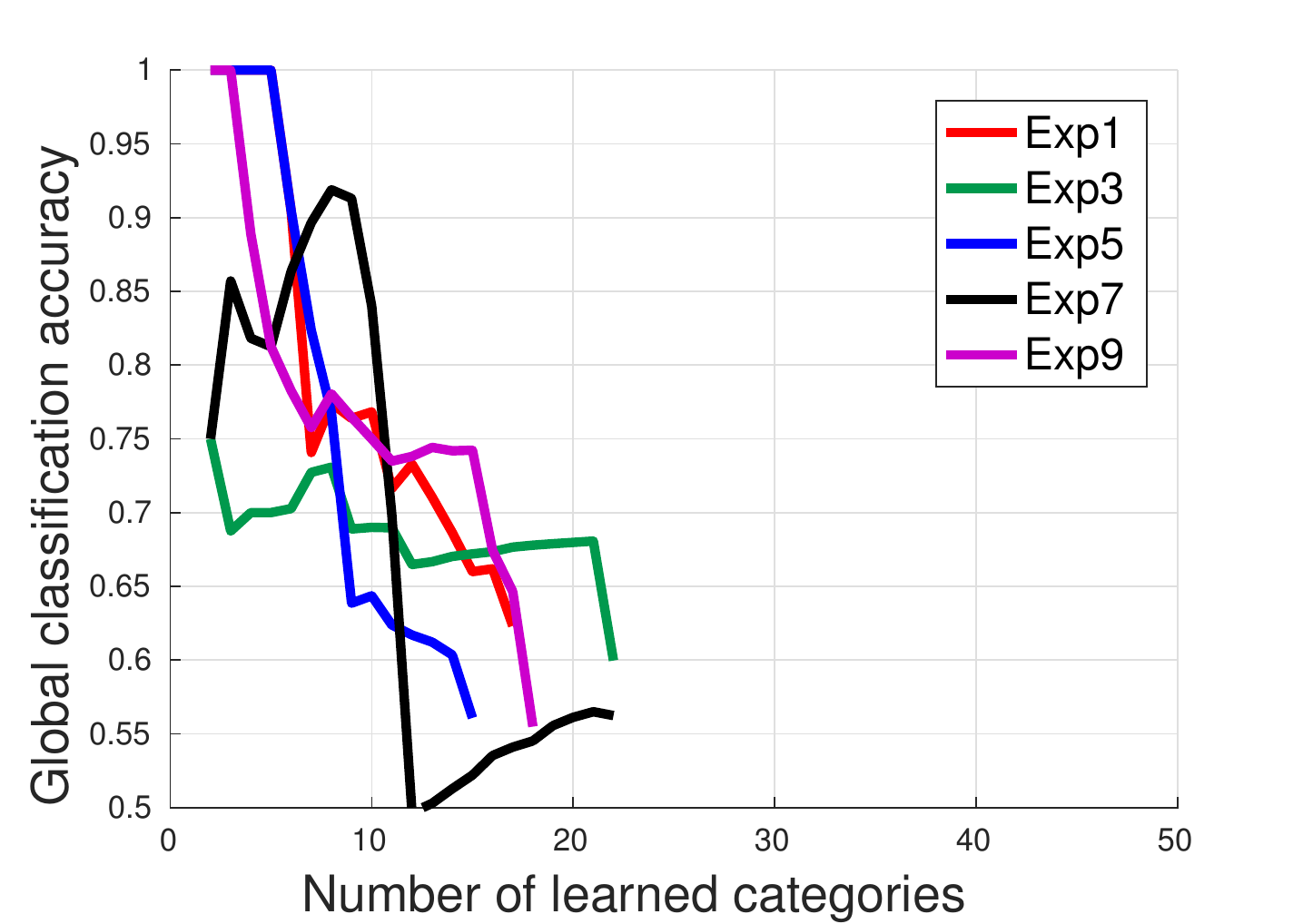}	\vspace{-1mm}\\
	 \multicolumn{3}{c}{(c) Summary of experiments using standard LDA}\vspace{5mm}
	\\
	\includegraphics[width=0.45\linewidth,  trim= 0cm -0.6cm 0.0cm 0cm,clip=true]{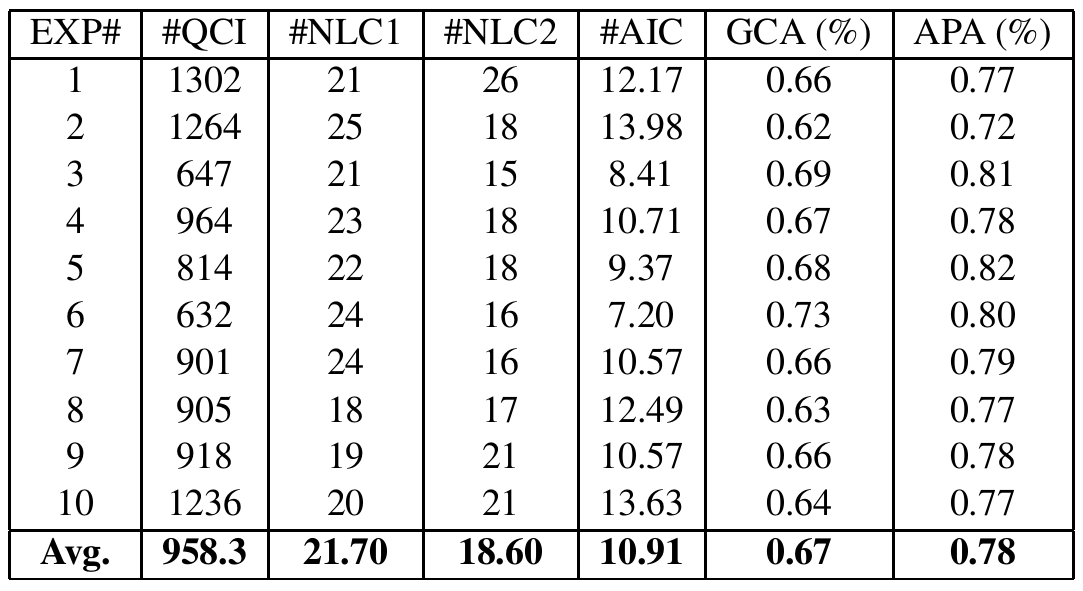}&\hspace{-2mm}
	\includegraphics[width=0.36\linewidth, trim= 0.5cm 0cm 1cm 0.65cm,clip=true]{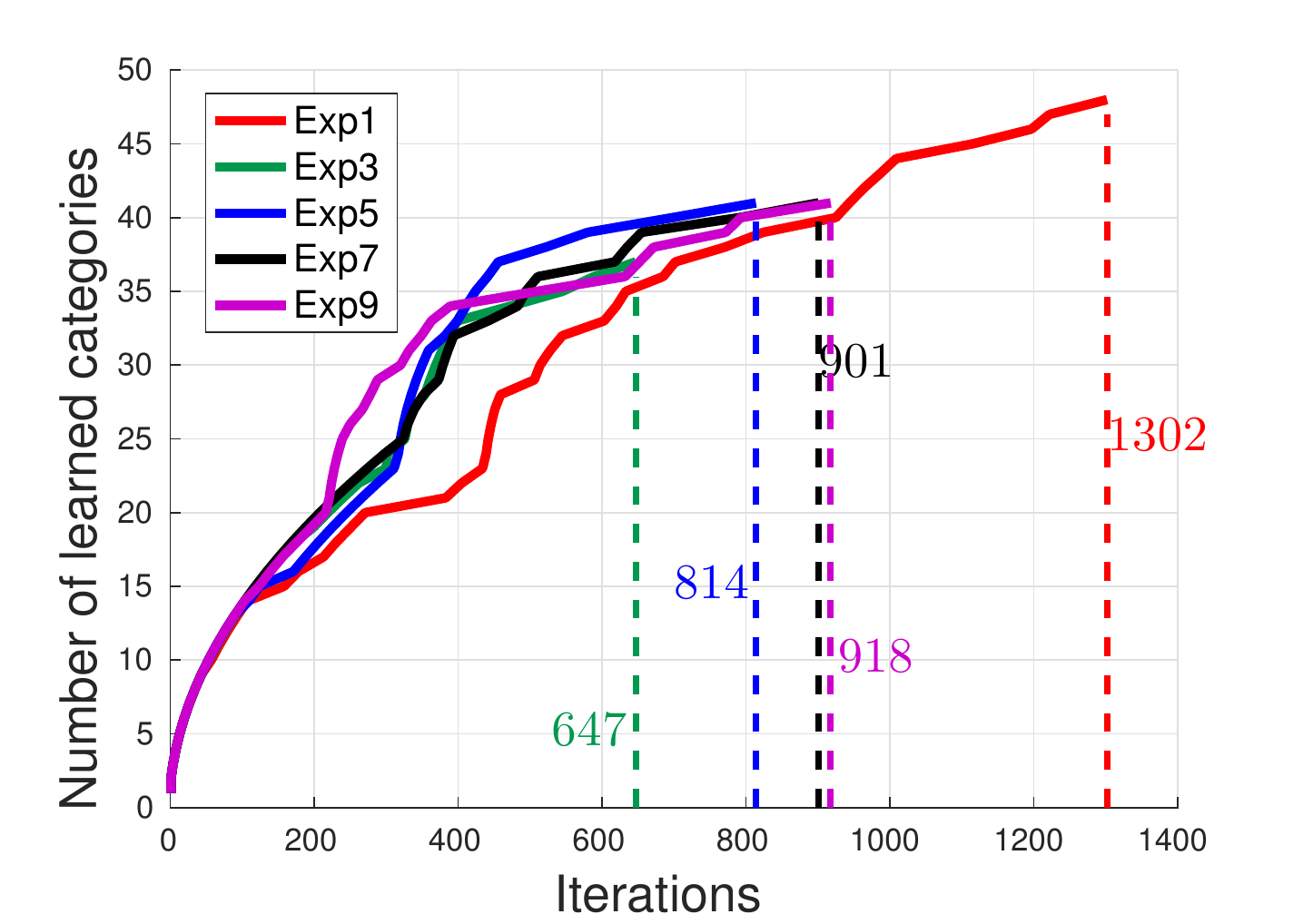}& \hspace{-2mm}
	\includegraphics[width=0.35\linewidth, trim= 0.35cm 0cm 1cm 0.65cm,clip=true]{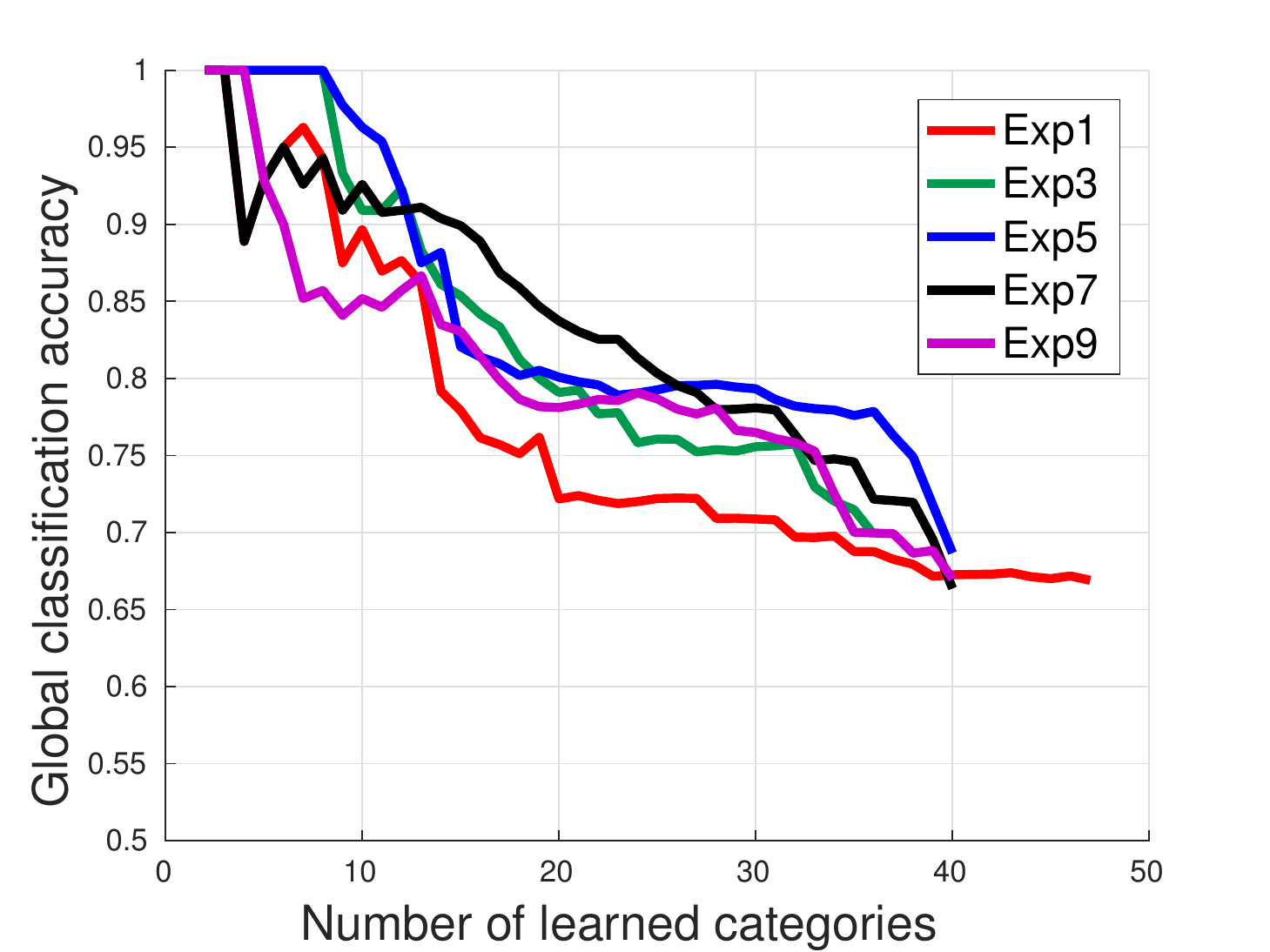} \vspace{0mm}\\
	 \multicolumn{3}{c}{(d) Summary of experiments using Local LDA} \vspace{5mm}
	\\
	\includegraphics[width=0.45\linewidth,  trim= 0cm -0.6cm 0.0cm 0cm,clip=true]{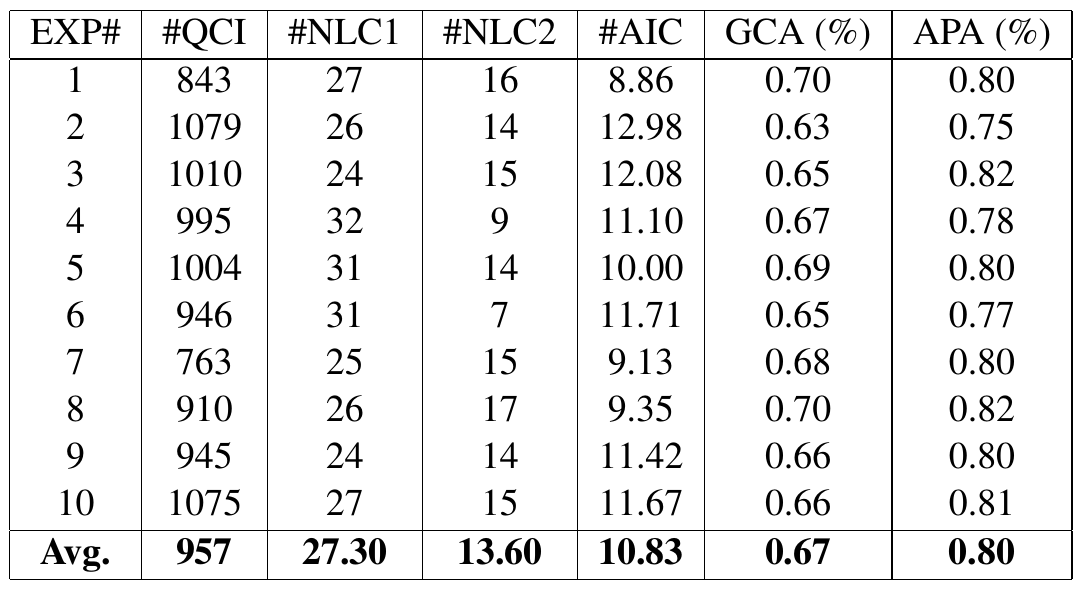}&\hspace{-2mm}
	\includegraphics[width=0.36\linewidth, trim= 0.5cm 0cm 1cm 0.65cm,clip=true]{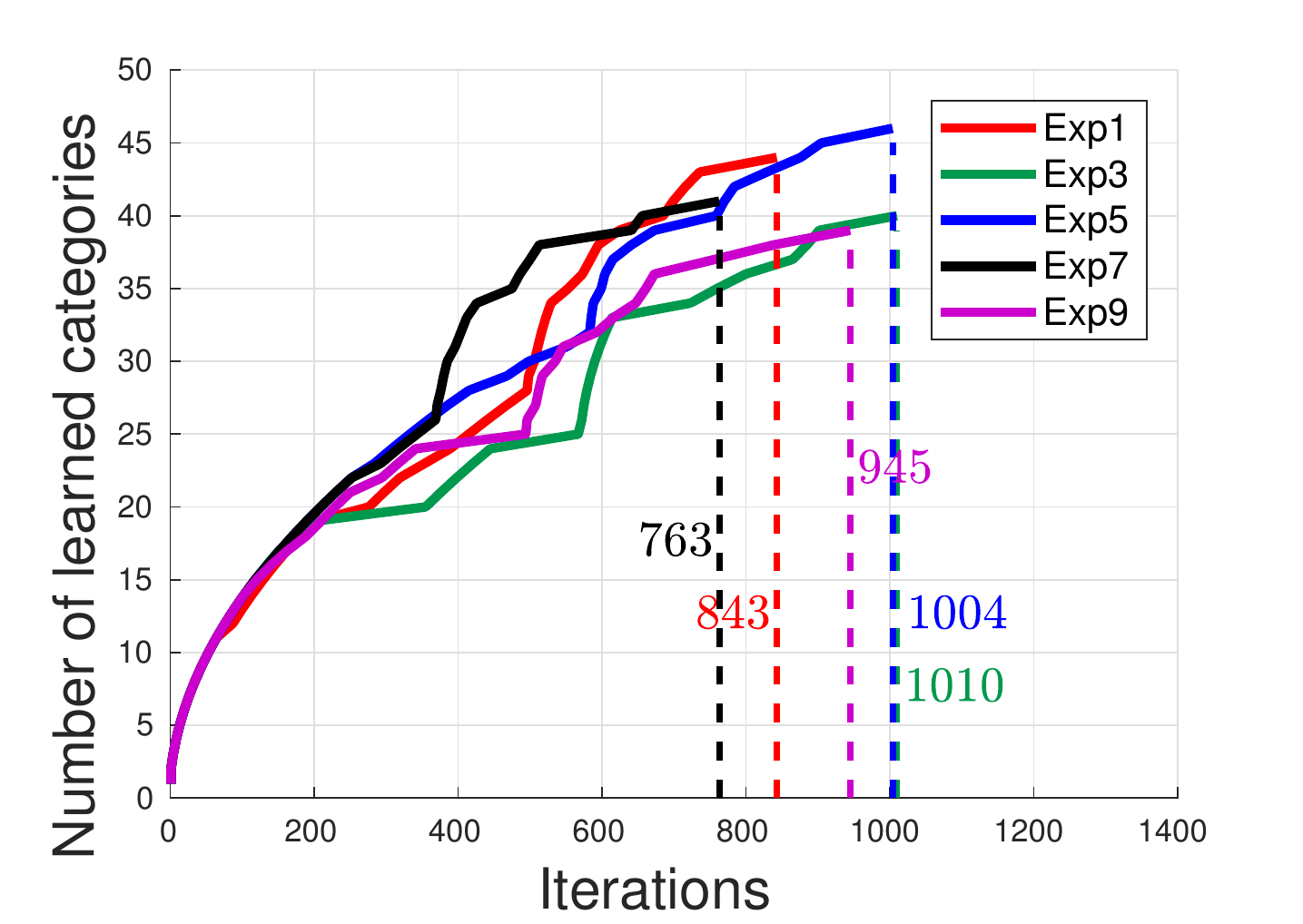}& \hspace{-2mm}
	\includegraphics[width=0.35\linewidth, trim= 0.35cm 0cm 1cm 0.65cm,clip=true]{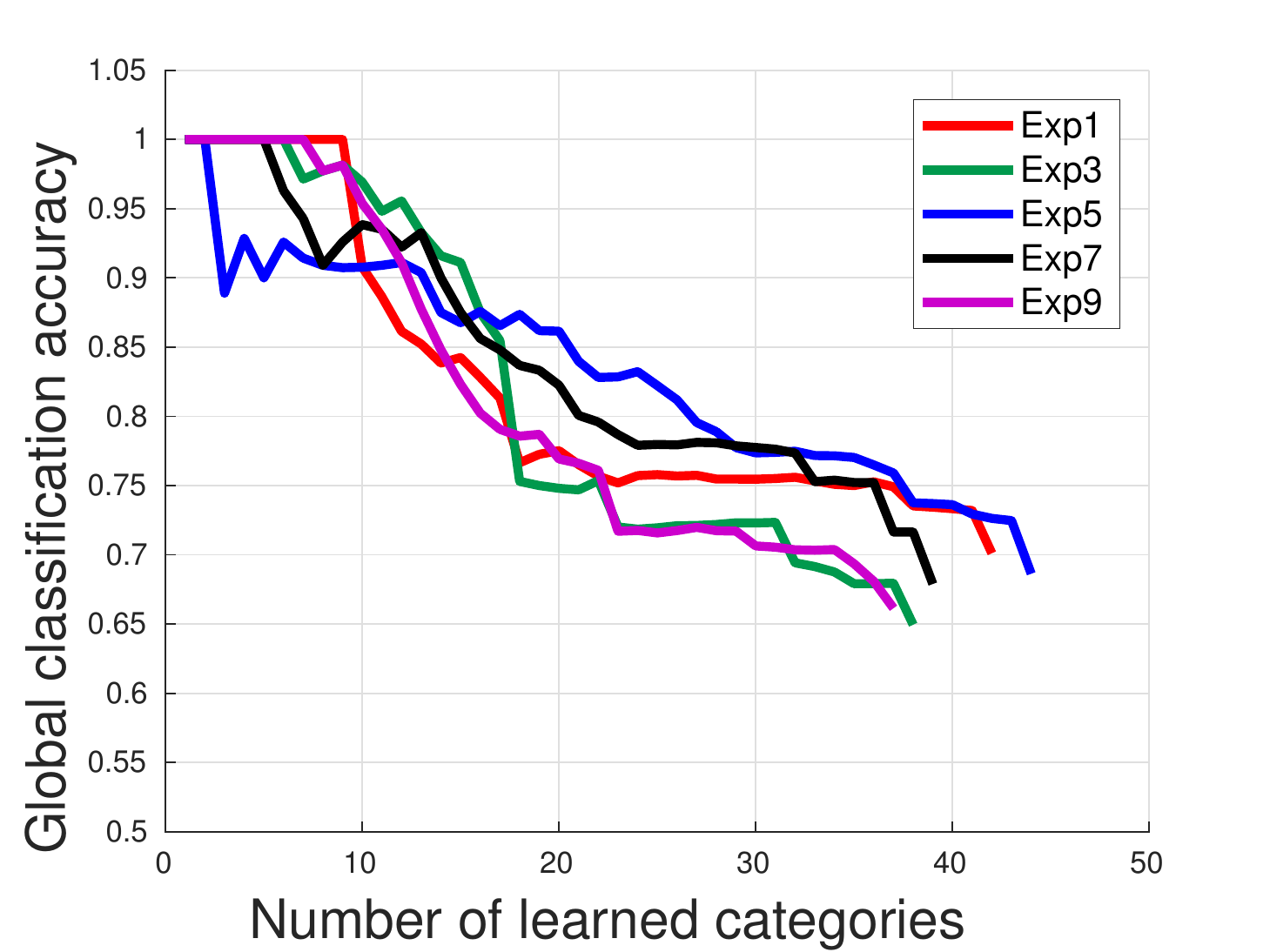} \vspace{0mm}\\
	 \multicolumn{3}{c}{(d) Summary of experiments using GOOD} \vspace{0mm}
	\\
\end{tabular}}
\vspace{-1mm}
\caption{Summary of open-ended evaluations of instance-based approaches with context change.}
\label{fig:open_ended_evaluation_with_context_change}
\end{figure}
\begin{figure}[!t] 
\center
\begin{tabular}[width=1\textwidth]{cc}
 \includegraphics[width=0.4\linewidth, trim= 0.3cm 0cm 0.0cm 0cm,clip=true]{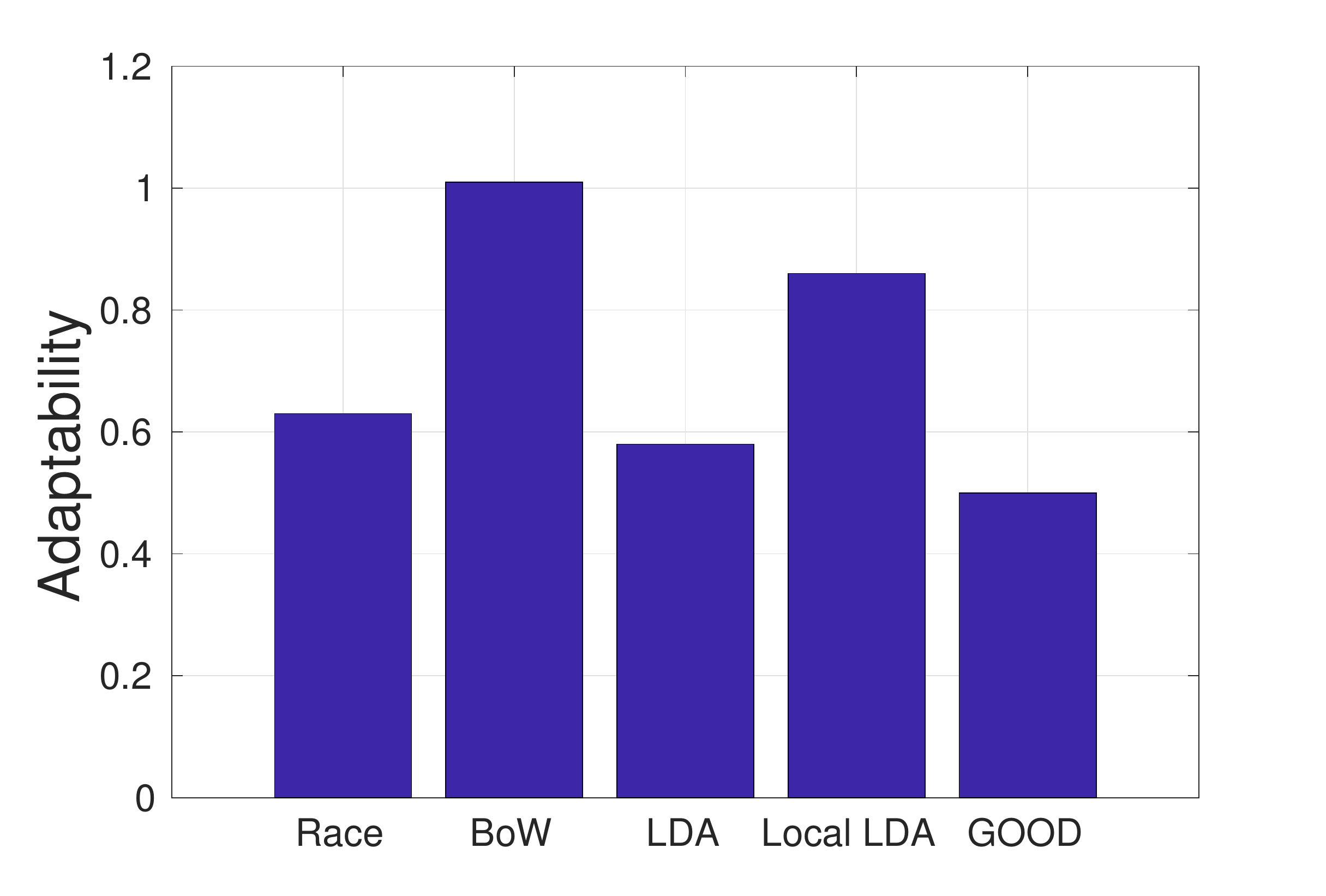} &\hspace{-0mm}
 \includegraphics[width=0.5\linewidth, trim= 0.0cm -1.2cm 0.0cm 0cm,clip=true]{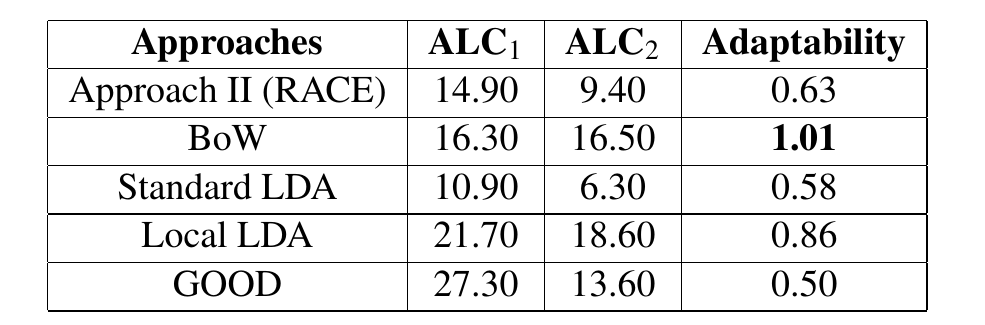}
\end{tabular}
\vspace{-2mm}
\caption{Summary of adaptability of different instance-based approaches.}
\label{fig:adaptability}       % Give a unique label
\end{figure}

%%%%%%%%%%%%%%%%%%%%%%%%%%%%%%%%%%%%%%%%%%%%%%%%%%%%%%%%%%%%%%%%%%%%%%%%%%%%%%%%%%%%%%%%%%%%%%%%%%%%%%%%%
The results obtained with the instance-based approaches are presented in Fig.~\ref{fig:open_ended_evaluation_with_context_change} and Fig.~\ref{fig:adaptability}. 
Several observations can be made based on Fig.~\ref{fig:open_ended_evaluation_with_context_change}. The agent with GOOD and Local LDA representations show a good 
balance among all evaluation criteria. In particular, the best performance, in terms of total number of learned categories in both contexts, was obtained with GOOD, closely followed by the local LDA approach. Approach~II achieved the best GCA and APA accuracies. This point could be explained by the fact that a higher number of categories learned by the other approaches tends to make the classification task more difficult. It is worth to mention that the agent was never able to learn all the existing categories with instance-based approaches.

In terms of number of learned categories in the second context, the agent learned (on average) more categories using Local LDA than with other approaches. The underlying reason is that Local LDA transforms objects from bag-of-words space into a more complex space and used distribution over distribution representation for providing powerful representation. BoW, GOOD and Approach~II achieved the second, third and forth places respectively. The worst one was LDA since the agent needs more data and time to reduce the effects of the topics learned in the first context (left column of Fig.~\ref{fig:open_ended_evaluation_with_context_change}). 

In terms of average number of stored instances per category (AIC), although Approach~II on average stored smallest number of instance per category, its discriminative power is not good. BoW, GOOD and local LDA achieved the second, third and forth places respectively while they provided a good balance between discriminative power and number of question/correction iterations (QCI). LDA was the worst approach in this comparison. The results of adaptability evaluation of all approaches are reported in Fig.~\ref{fig:adaptability}. By comparing all experiments, it is visible that only when the agent used BoW approach, it could learn more categories in the second context than in the first context. Therefore, the most adaptable approach was BoW (adaptability of 1.01), followed by Local LDA (adaptability of 0.86). The Approach~II and standard LDA achieved the third and forth places respectively. The agent with GOOD was the less adaptable approach.

%%%%%%%%%%%%%%%%%%%%%%%%%%%%%%%%%%%%%%%%%%%%%%%%%%%%%%%%%%%%%%%%%%%%%%%%%%%%%%%%%%%%%%%%%%%%%%%%%%%%%%%%%%%%%%%%%
\subsection {Model-Based Approaches}
Another set of experiments in a multi context scenario was carried out to evaluate how different model-based object recognition approaches cope with the effects of the context change (see Fig.~\ref{fig:open_ended_evaluation_with_context_change_NB}). 

\begin{figure*}[!b]
\resizebox{\columnwidth}{!}{
\begin{tabular}{ccc}
	\includegraphics[width=0.44\linewidth,  trim= 0cm -1cm 0.0cm 0cm,clip=true]{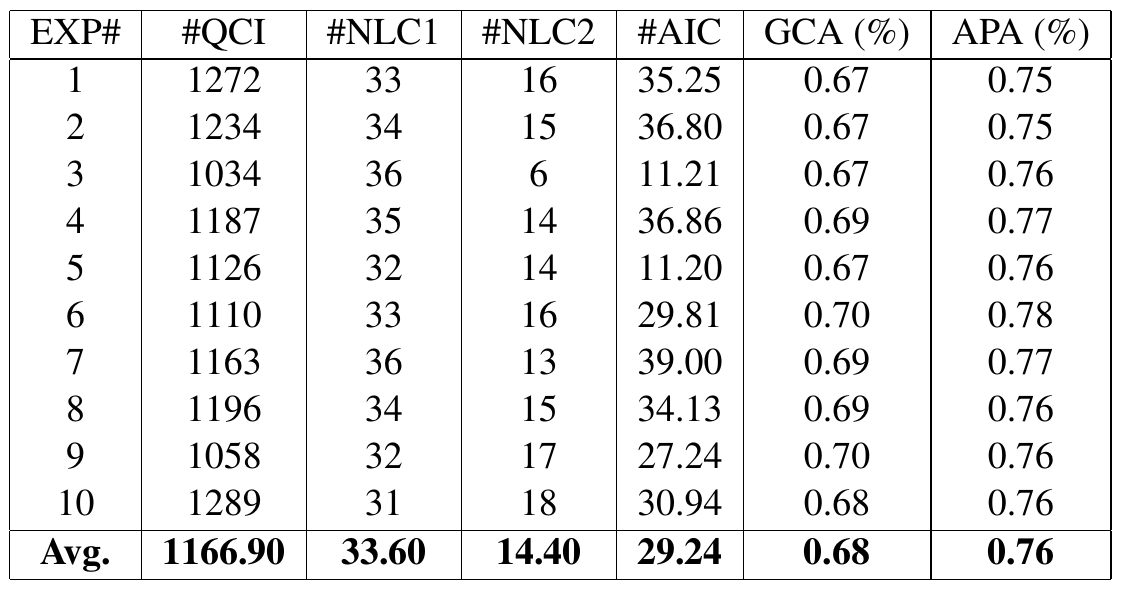}& \hspace{-0mm}
	\includegraphics[width=0.37\linewidth, trim= 0.5cm 0cm 0cm 0.65cm,clip=true]{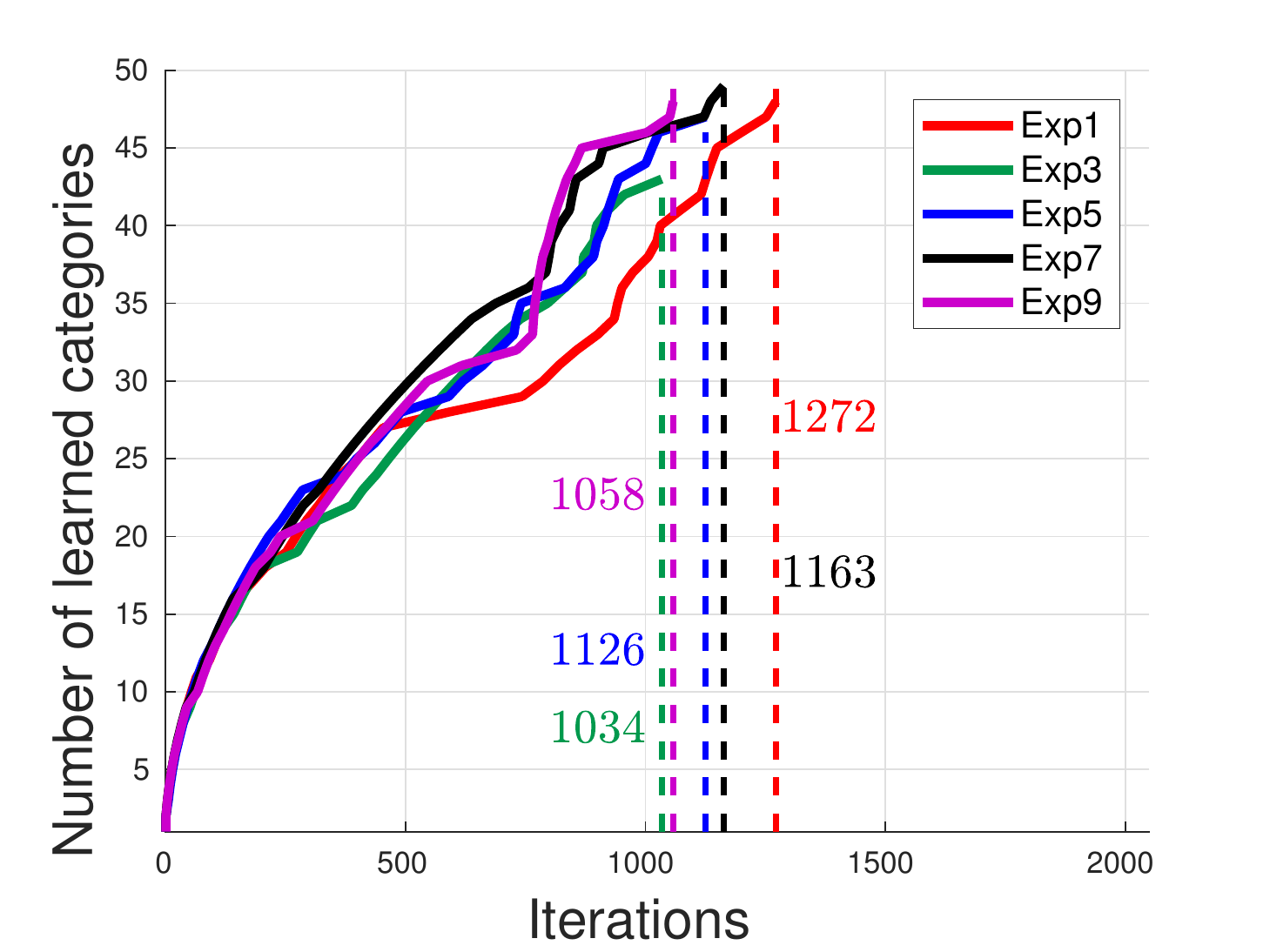}& \hspace{-0mm}
	\includegraphics[width=0.34\linewidth, trim= 0.25cm 0cm 1cm 0.65cm,clip=true]{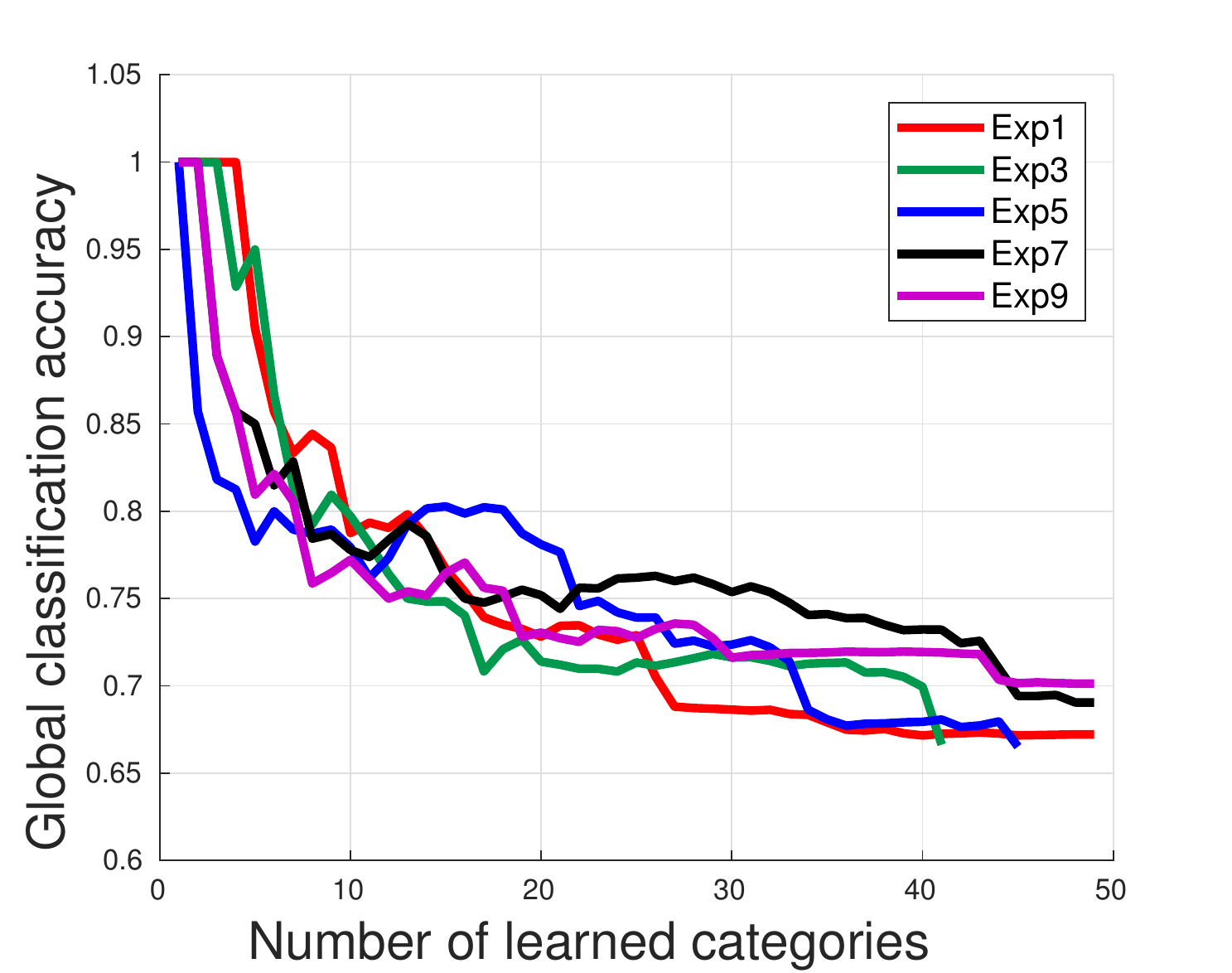}	\vspace{0mm}\\
	 \multicolumn{3}{c}{ (\emph{a}) Summary of experiments using BoW }\vspace{2mm}
	\\ 
	\includegraphics[width=0.45\linewidth,  trim= 0cm -1cm 0cm 0cm,clip=true]{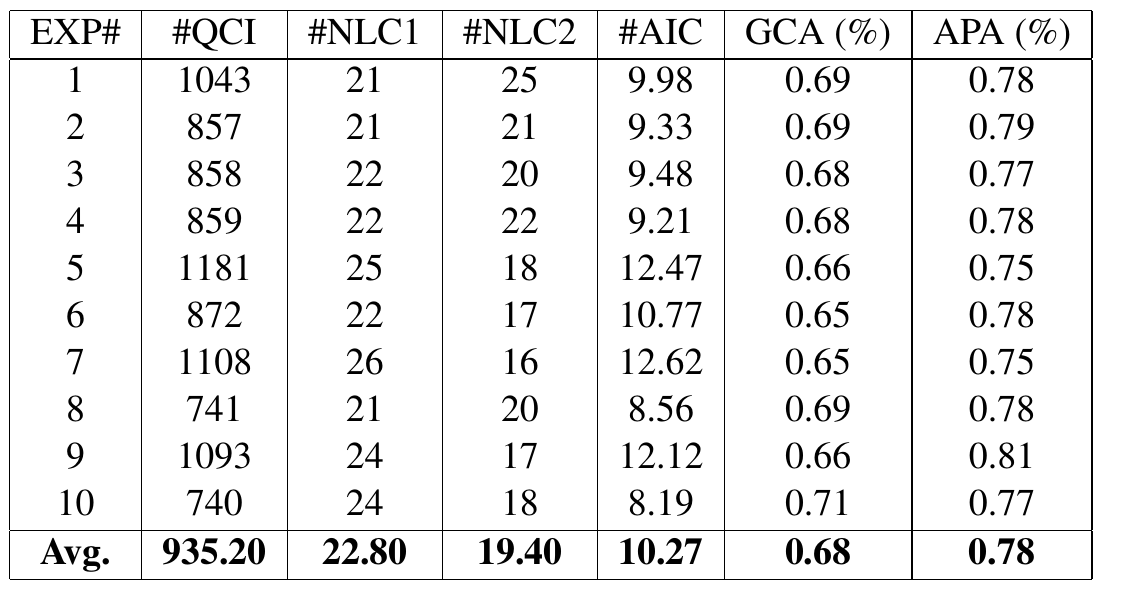}&\hspace{-0mm}
	\includegraphics[width=0.37\linewidth, trim= 0.5cm 0cm 0cm 0.65cm,clip=true]{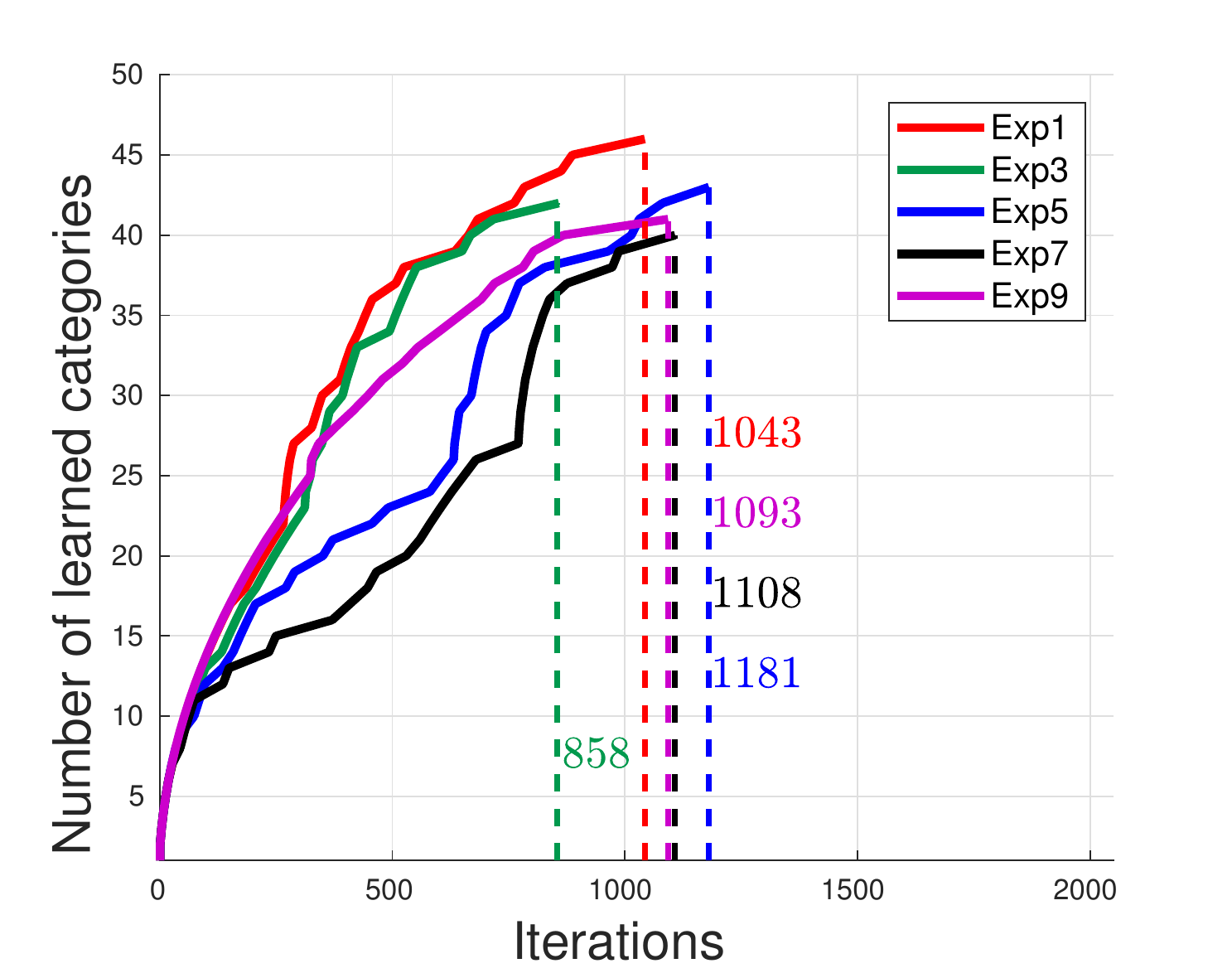}& \hspace{-0mm}
	\includegraphics[width=0.34\linewidth, trim= 0.25cm 0cm 1cm 0.65cm,clip=true]{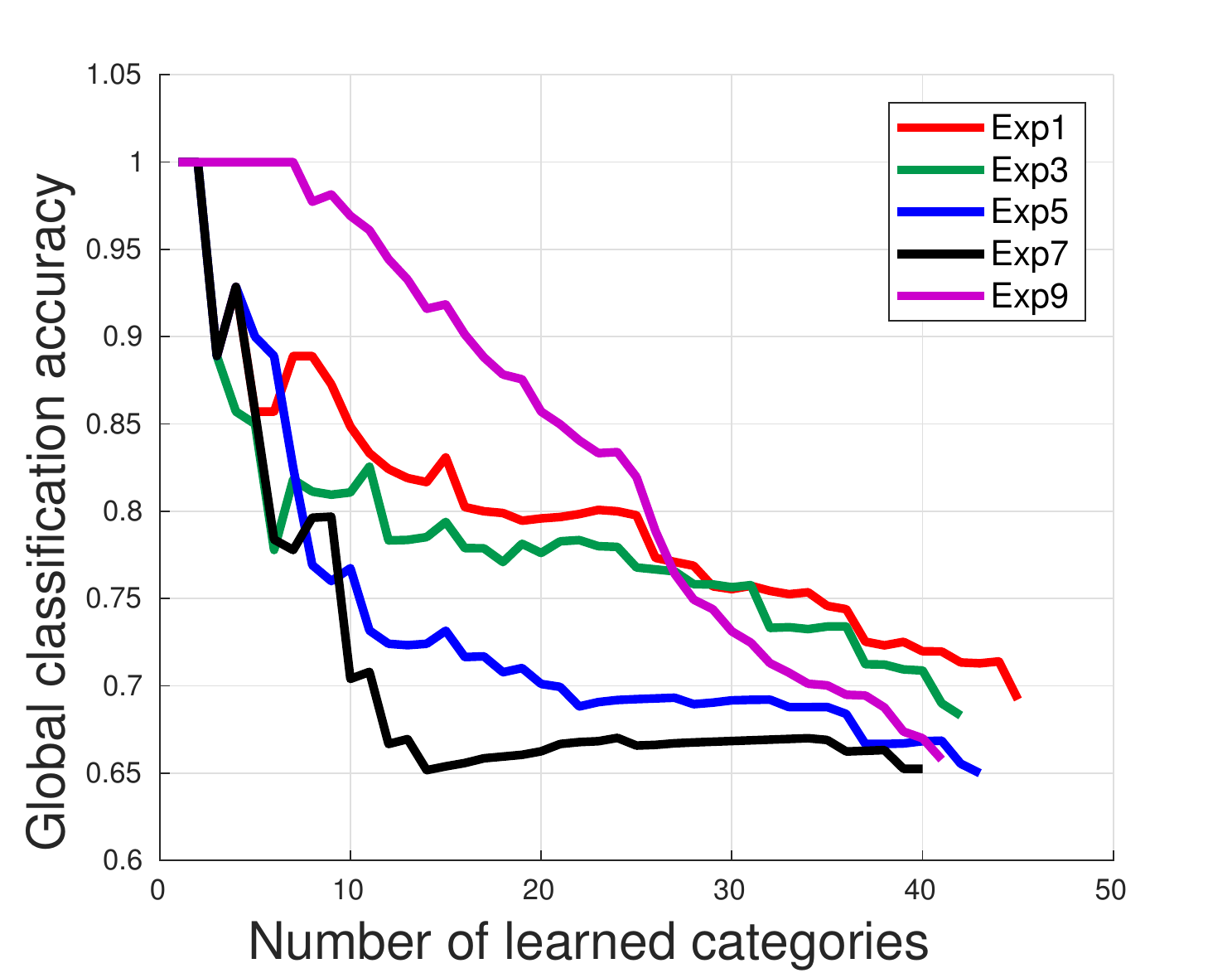}	\vspace{0mm}\\
	\multicolumn{3}{c}{(\emph{b}) Summary of experiments using standard LDA}\vspace{2mm}
	\\
	\includegraphics[width=0.45\linewidth,  trim= 0cm -1cm 0cm 0cm,clip=true]{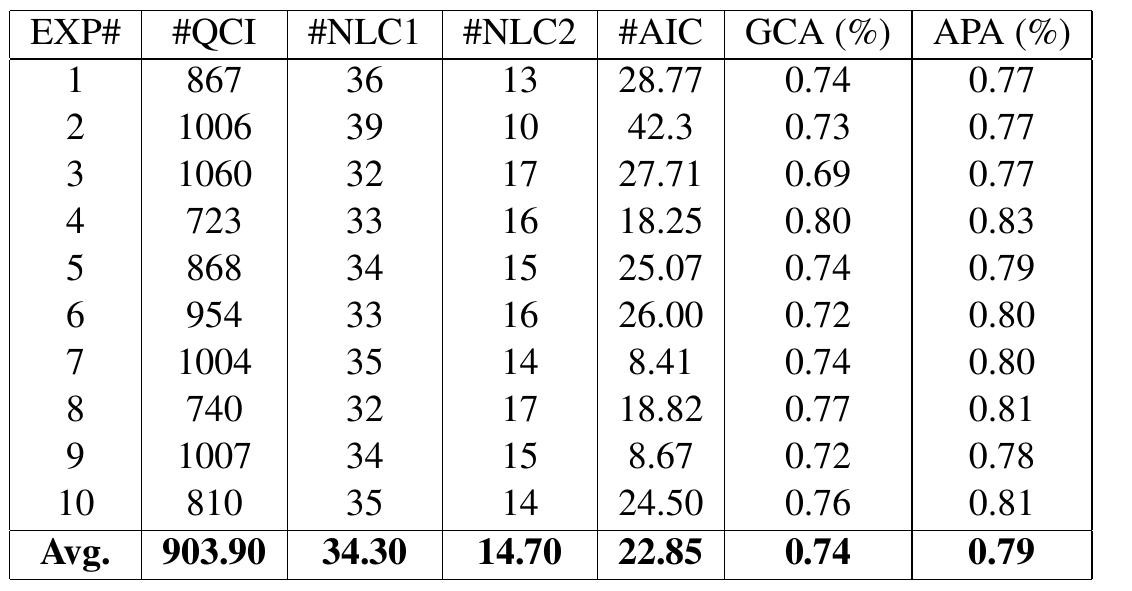}&\hspace{-0mm}
	\includegraphics[width=0.37\linewidth, trim= 0.5cm 0cm 0cm 0.65cm,clip=true]{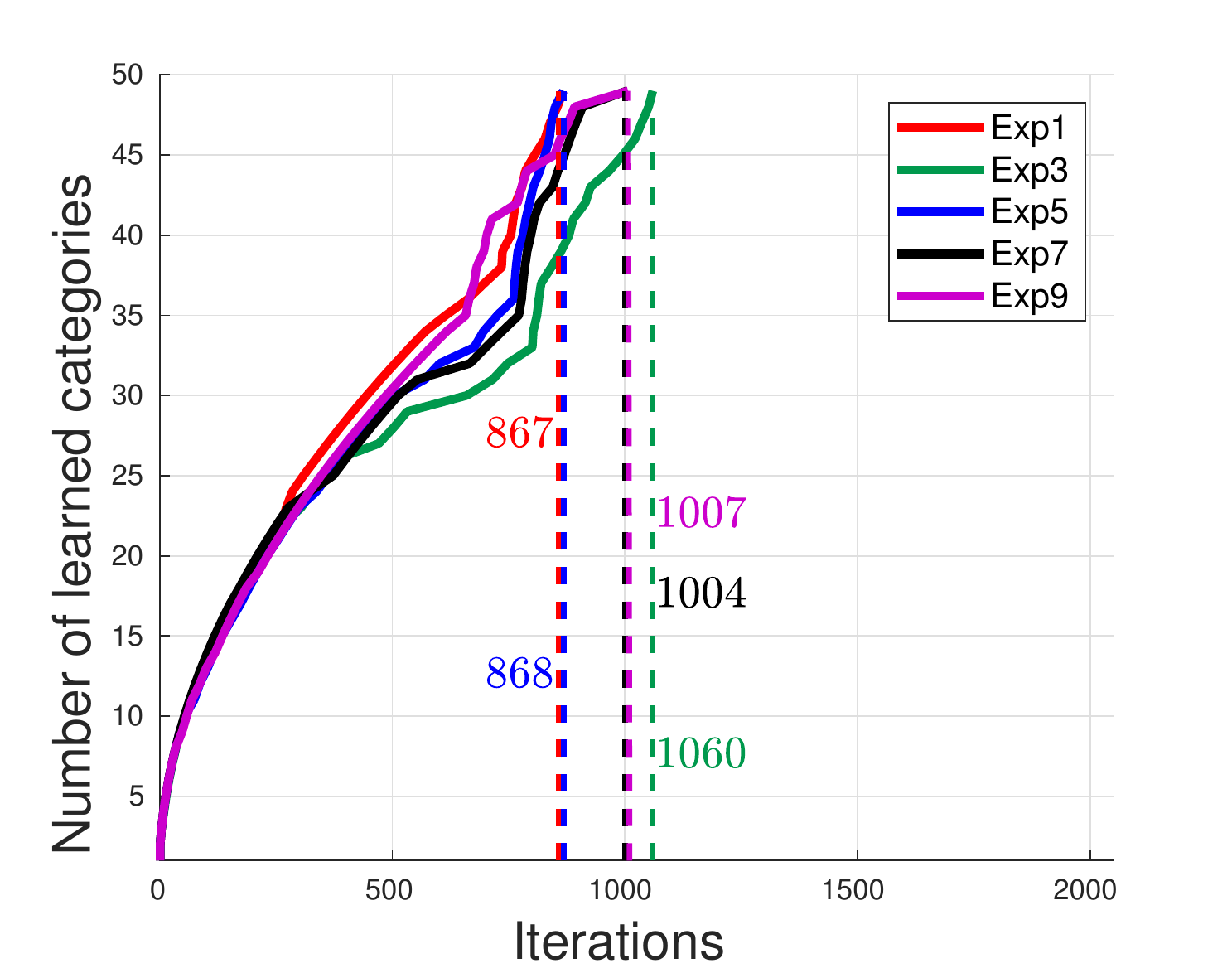}& \hspace{-0mm}
	\includegraphics[width=0.37\linewidth, trim= 0.25cm 0cm 1cm 0.65cm,clip=true]{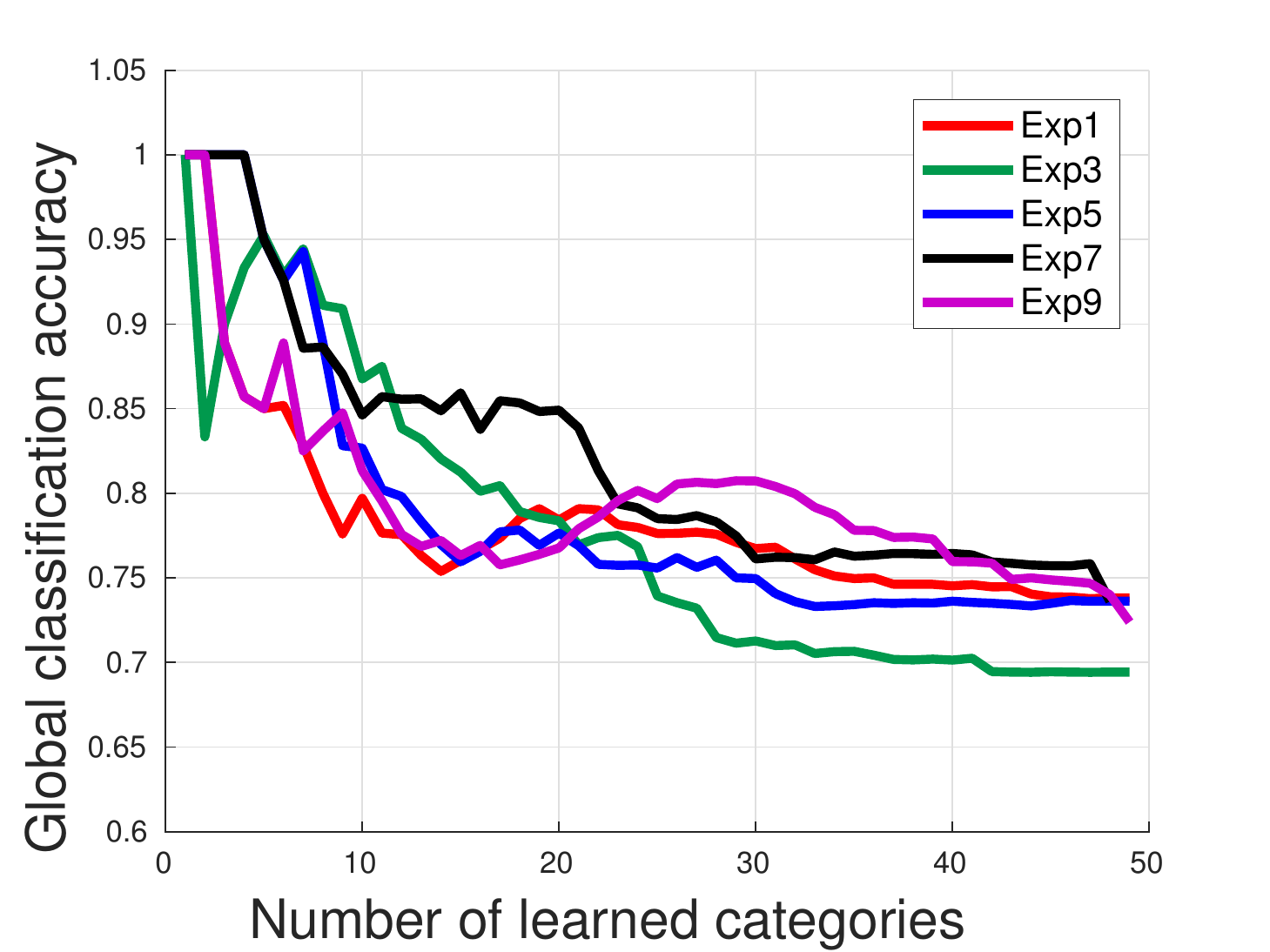}	\vspace{0mm}\\
	\multicolumn{3}{c}{(\emph{c}) Summary of experiments using Local LDA}\vspace{2mm}
	\\
	\includegraphics[width=0.45\linewidth,  trim= 0cm -1cm 0cm 0cm,clip=true]{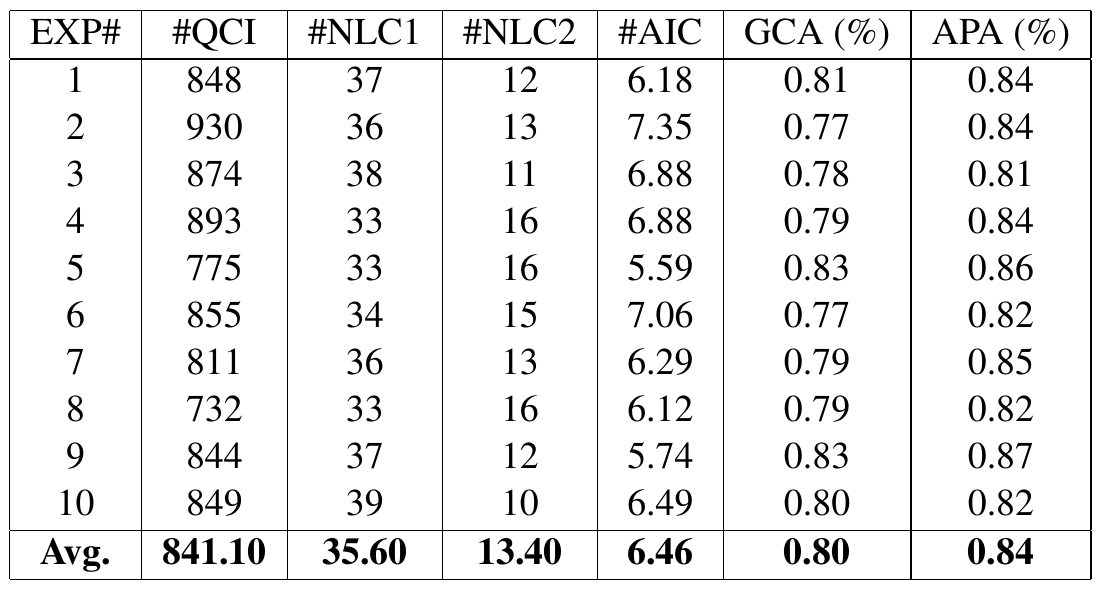}&\hspace{-0mm}
	\includegraphics[width=0.37\linewidth, trim= 0.5cm 0cm 0cm 0.65cm,clip=true]{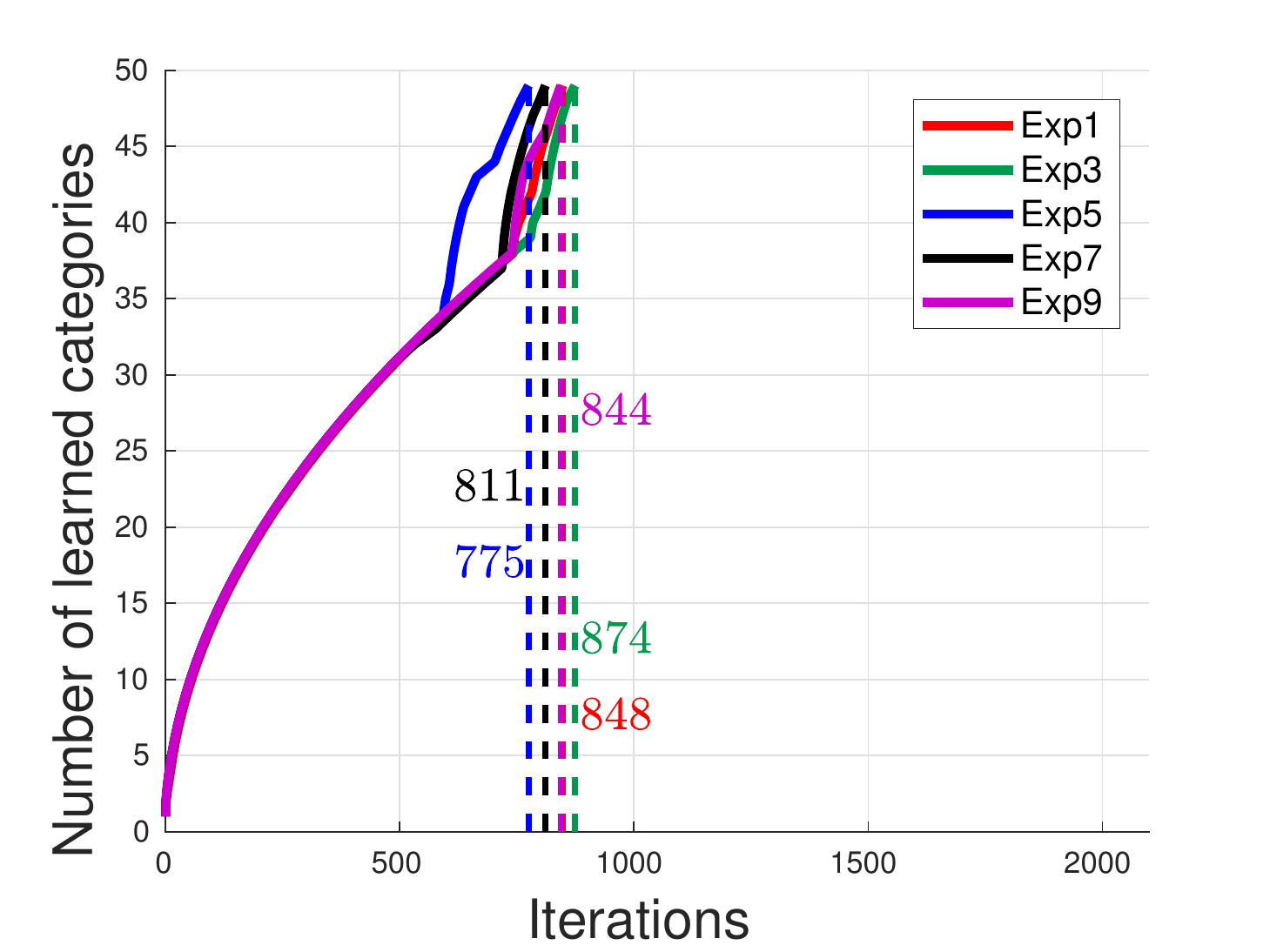}& \hspace{-0mm}
	\includegraphics[width=0.37\linewidth, trim= 0.25cm 0cm 1cm 0.65cm,clip=true]{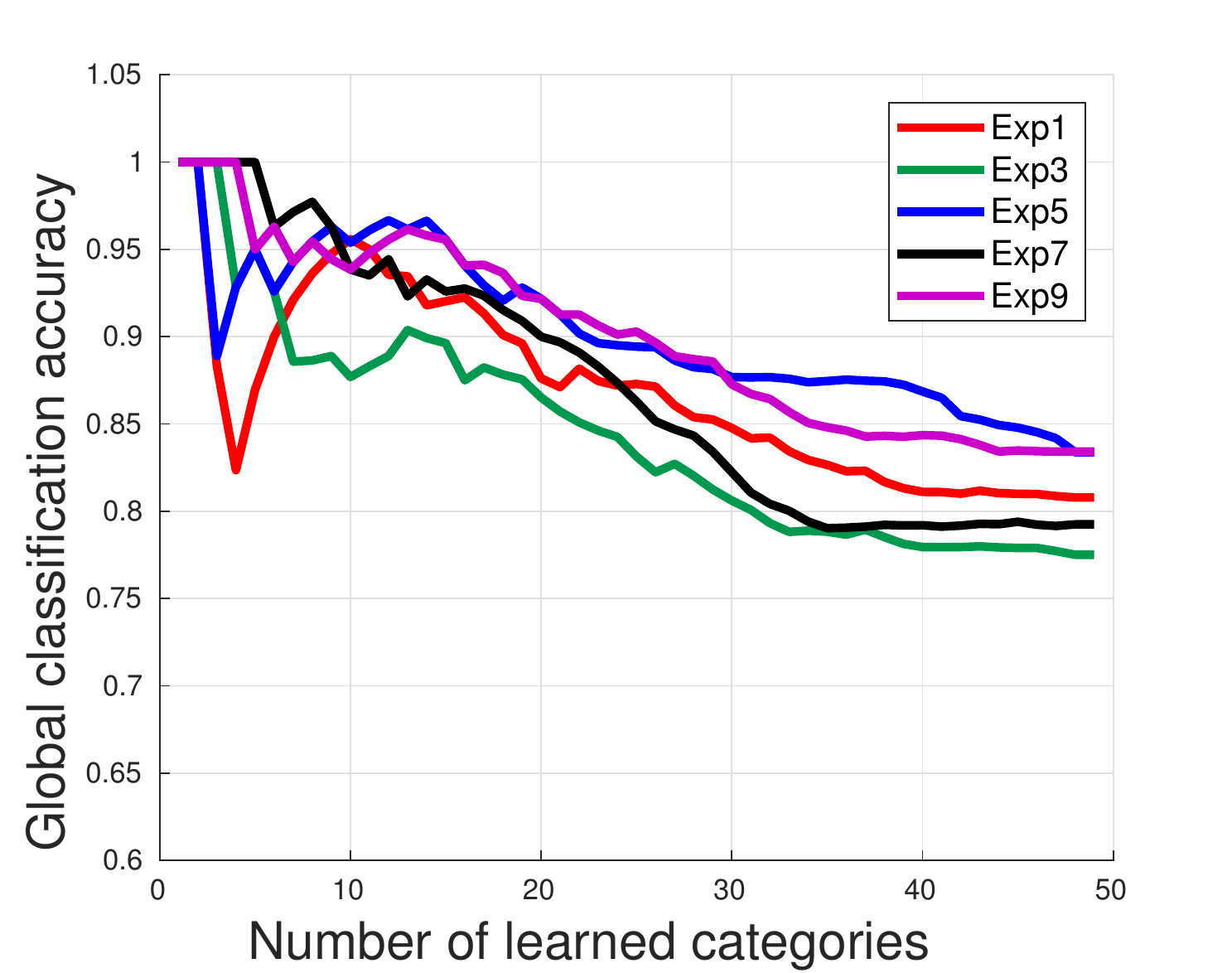}	\vspace{0mm}\\
	\multicolumn{3}{c}{(\emph{c}) Summary of experiments using GOOD}\vspace{2mm}
	\\
\end{tabular}}
\vspace{-2mm}
\caption{Summary of open-ended evaluations of model-based approaches with context change.}
\label{fig:open_ended_evaluation_with_context_change_NB}
\end{figure*}
The left column in Fig.~\ref{fig:open_ended_evaluation_with_context_change_NB} provides a detailed summary of the obtained results. Concerning ALC, in all experiments, the agent learned all the existing categories in the dataset using GOOD and Local-LDA. Therefore, experiments finished due to ``\emph{lack of data}'', and not by reaching the breakpoint. Therefore, more categories have been learned with GOOD and Local LDA if they were available in the dataset (see Fig.~\ref{fig:open_ended_evaluation_with_context_change_NB} (\emph{c} and \emph{d})). BoW achieved the third place. We observed that in 8 out of 10 experiments with BoW, all the categories in the dataset were learned by the agent. Standard LDA achieved the fourth place. The agent in none of the tests could learn all categories in the second context.

By comparing all experiments, it is clear that the number of iterations required to learn all object categories with GOOD was smaller than other approaches. The agent with GOOD, on average, learned all categories after $841.10$ iterations. Local LDA achieved the second place. It is also clear from Fig.~\ref{fig:open_ended_evaluation_with_context_change_NB} that the GOOD achieved the best accuracies (i.e., APA and GCA) with stable performance and clearly outperforms the other approaches.

As depicted in the center column of Fig.~\ref{fig:open_ended_evaluation_with_context_change_NB}, the best performance, in terms of number of learned categories and the number of iterations required to learn a certain number of categories, was obtained with the GOOD, closely followed by the Local LDA approach. The agent with BoW also showed good results. The worst one was standard LDA since the agent needs more data and time to reduce the effects of the topics learned in the first context.

The right column of Fig.~\ref{fig:open_ended_evaluation_with_context_change_NB} shows the global classification accuracy (i.e., since the beginning of the experiment) as a function of the number of learned categories. It is clear from Fig.~\ref{fig:open_ended_evaluation_with_context_change_NB} (right) that GOOD achieved the best accuracies with stable performance and clearly outperforms the other approaches. Local LDA achieved the second places also with stable performance. The worst one was standard LDA, since it uses shared topics among all categories and the agent needs more data and time to reduce the effects of the topics learned in the first context. 

Since the proposed adaptability evaluation presupposes that experiments reach the breakpoint, it was not possible to compute the adaptability of GOOD, Local LDA and BoW approaches.
In fact, the agent with GOOD and Local LDA successfully learned all categories in both contexts and all experiments concluded prematurely due to the ``\emph{Lack of data}''. Similarly, the agent with BoW, in 8 out of 10 experiments could learn all categories in both contexts. The adaptability of the standard LDA approach was 0.85, much better than the adaptability of instance-base learning with the same representation (adaptability of 0.58).

\subsection {Demonstration}
A real demonstration was carried out using Imperial College domestic environment dataset \citep{Doumanoglou2016}. For this demonstration, an instance-based learning approach with BoW has been integrated into the object perception system. The system initially had no prior knowledge. In the first context, an instructor teaches two object categories including \emph{oreo} and \emph{amita} to the system, and the system conceptualizes those categories. The system is then tested by the other scenes captured from different viewpoints. The system could recognize all objects properly (see Fig.~\ref{fig:context_pr2} \emph{top-left}). Then, the system is moved to the second context. In this context, the system gains knowledge about \emph{lipton} and \emph{softkings} categories. Similar to the first context, the knowledge of system is validated using several scenes (see Fig.\ref{fig:context_pr2} \emph{bottom-left}). Later, we moved the system to a new context, which is much more crowded and complex than the previous contexts. In this context, eight instances of the four categories exist. The robot could recognize all objects correctly by using knowledge from previous contexts (see Fig.\ref{fig:context_pr2} \emph{right}). This evaluation illustrates the process of acquiring object categories in an open-ended fashion from multi contexts. Moreover, it shows that disrupting or erasing the category models learned from the previous contexts is not a rational choice. A video of this demonstration is online at:
{\cblue{\small \href{https://youtu.be/l6q6fI5H6zY}{https://youtu.be/l6q6fI5H6zY}}}.

\section { Summary}
\label{sec:summary}
In this chapter, we defined an online evaluation approach to assess the performance of open-ended object recognition approaches regarding their ability to cope with the effects of the context change in multi-context scenarios. Two learning approaches (instance-based and model-based) with several object representations were evaluated using the proposed methodology. Two sets of experiments were carried out to assess the performance of all the proposed 3D object category learning and recognition approaches in single context and two-context scenarios. Experimental results proved that all the approaches can incrementally learn new object categories. 

In case of instance-based approaches, the overall number of categories learned with GOOD is clearly better than the best performances obtained with the other approaches. The agent with Local LDA also demonstrated an appropriate balance among all criteria. The underlying reason was that Local LDA used distribution over distribution representation for providing powerful representations. Furthermore, adaptability of all approaches was evaluated. BoW was the most adaptable approach immediately followed by the Local LDA. Therefore, the agent could adapt better to a new context when it uses BoW or Local LDA approaches. It was observed that BoW and Local LDA provided a good balance between discriminative power and number of \emph{question/correction} iterations (QCI). The discriminative power of Approach~II and standard LDA were clearly not good for such context-sensitive environments.

In case of model-based approaches, experimental results show that the overall classification performance obtained with GOOD is comparable to the best performances obtained with the other approaches. Moreover, it was observed that the agent could learn all categories in both contexts when it uses GOOD and Local LDA approaches. Similar to the instance-based experiments, it was observed that GOOD and Local LDA provided a good balance between discriminative power and number of \emph{question/correction} iterations (QCI). The discriminative power of standard LDA was not good when compared with the other approaches. Furthermore, adaptability of all approaches was evaluated. Since the proposed adaptability evaluation presupposes that experiments reach the breakpoint, it was not possible to compute the adaptability of GOOD, BoW and Local LDA. Therefore, the best adaptability was achieved by standard LDA. 

For future work we would like to investigate the possibility of improving performance by considering both external and internal contexts, since there are several evidences of a relation between the external and internal contexts \citep{kokinov1997dynamic}\citep{qian2012learning}. It would also be important to do experiments with a larger dataset in order to avoid the ``lack-of-data'' termination condition in open-ended evaluations.

\cleardoublepage
\chapter{System Demonstration and Profiling}
\label{chapter_8}

Throughout this thesis, a set of interactive open-ended learning approaches for grounding 3D object categories has been presented, enabling robots to adapt to different environments and reason out how to behave in response to the request of a complex task. In this chapter, three types of experiments were carried out to evaluate the proposed approaches. First, based on a session where users manipulate objects on a table and interact with the developed perception and perceptual learning system, we carry out a profiling analysis of the main modules of the system. We also used the recorded session to demonstrate all the characteristics of the proposed GOOD descriptor. 
Second, two scenarios namely \emph{clear\_table} and \textit{serve\_a\_meal} have been designed to show all functionalities of the object recognition and grasping. These demonstrations show that the proposed approaches have been successfully tested on a PR2 robot and a JACO arm, showing the importance of having a tight coupling between perception and manipulation. 
Finally, two demonstrations using Washington RGB-D Scenes Dataset v2 and Imperial College Domestic Environment Dataset \citep{Doumanoglou2016} have been performed. These demonstrations showed that the system is capable of using prior knowledge to recognize new objects in the scene and learn about new object categories in an open-ended fashion. All tests were performed with an i7, 2.40GHz processor and 16GB RAM.

\section{ Open-Ended 3D Object Category Learning Scenarios}

To show all the functionalities of the system, a session has been recorded, where several users interacted with the system. During this session, users presented objects to the system and provided the respective category labels. All raw data from the RGB-D sensor was recorded in a rosbag, which was then used to test different configurations of our system. Three demonstrations were performed using the recorded rosbag. In all demonstrations, we have assumed that the set of object categories to be learned is not known in advance and the training instances are extracted from actual experiences of a robot rather than being available at the beginning of the learning process. In the first and third demonstrations, when the system started, the set of categories known to the system was empty while in the second demonstration, the system initially had prior knowledge about two categories and there is no information about the other categories. In these demonstrations, we have used instance-based object category learning and recognition as discussed in chapter~\ref{chapter_6}, section~\ref{sec:instance_based_learning}.

\subsection {The object perception system in an interactive session}
A 3.5 minutes session has been recorded, where several users interacted with the object perception system. During this session, users present objects to the system and provide their respective category labels. The system detects pointing gestures of the user and detects and tracks the presented objects. Figure~\ref{fig:demo1} presents some snapshots of this session. Table~\ref{table:video_script} presents a summary of the main events. In more detail, the session progressed as follows:

% Throughout this session, the system must be able to detect, conceptualize and recognize new object categories, as well as to detect pointing gestures used for labeling them. Figure \ref{fig:demo1} and the following explanation illustrate the behavior of the main modules of the system, from user and object tracking to learning and recognition.  A video of this demonstration is available at: {\cblue{\small \href{https://youtu.be/XvnF2JMfhvc}{https://youtu.be/XvnF2JMfhvc}}}. 

\begin{figure}[!b]
\centering
\begin{centering}
\begin{tabular}{ccc}
\includegraphics[width=0.41\linewidth, trim= 0cm 0cm 5.3cm 0cm, clip=true]{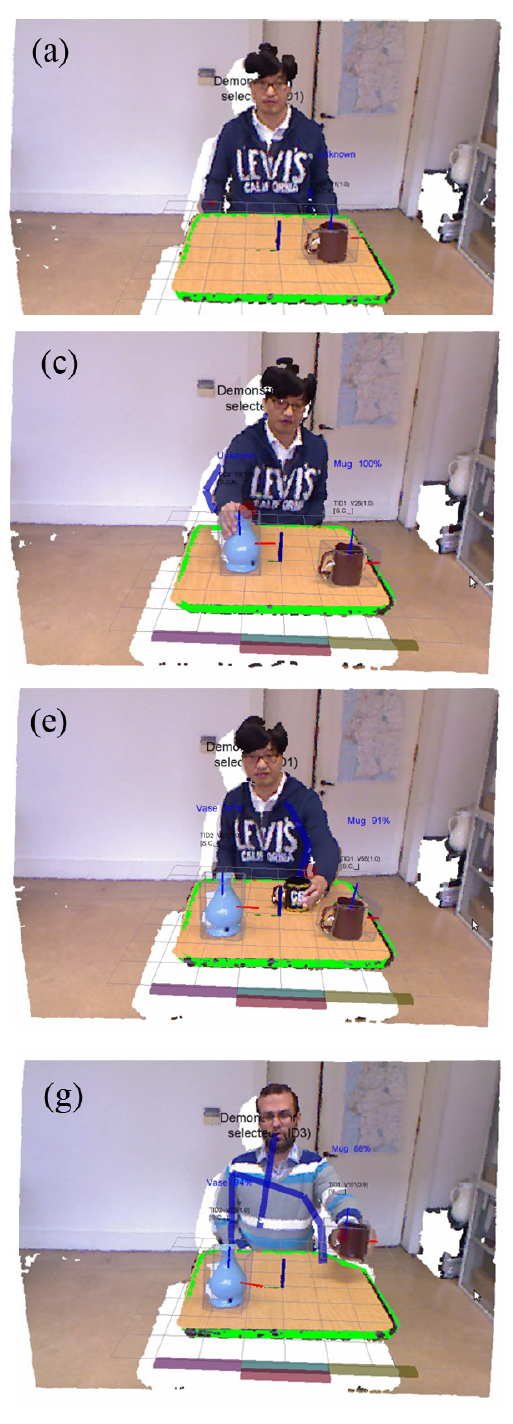}& \quad\quad\quad\quad\quad&
\includegraphics[width=0.41\linewidth, trim= 5.3cm 0cm 0cm 0cm, clip=true]{Figures/scenario_veritical.pdf}
\end{tabular}
\end{centering}
\vspace{-3mm}
\caption{Sequence of events in the experiment}
\label{fig:demo1}
\end{figure}

\begin{enumerate}[(a)]
\item The system works in scenario where a table is in front of the robot and there is no knowledge about any category. The graphical menu in front of the table is the interactive menu that enables teaching new object categories. The instructor
puts a \emph{Mug} on the table. Tracking is initialized with track ID 1 (TID1). The gray bonding box signals the pose of the object as estimated by the tracker. TID1 is classified as Unknown because mugs are not yet known to the system;
\item Instructor labels TID1 as a \emph{Mug}. The system conceptualizes the category;
\item The \emph{Mug} is correctly classified. The instructor places a \emph{Vase} on the table. Tracking is initialized with TID2. The Vase is unknown to the system; this frame shows that the system is able to detect and track multiple objects in the scene. Moreover, it demonstrates that both the tracking and recognition work when the user is holding the objects; 
\item  The instructor labels TID2 as a Vase. This labeling is done using a different interaction modality: the instructor points at track ID 2, and labels this
object as \emph{Vase};
\item The \emph{Vase} is properly recognized. An additional \emph{Mug} is placed at the center of the table. Tracking is initialized with TID3. This particular \emph{Mug} had not been previously seen, but the system can correctly recognize it, because the \emph{Mug} category was previously taught. This shows that the system is capable of using prior knowledge to recognize new objects in the scene; 
\item Another instructor arrives; once he sits on front of the robot, he will be considered as the system's instructor. This frame shows instructor detection
and tracking;
\item The instructor removes all objects from the scene; no objects will be visible;
\item A Plate enters the scene. It is detected and assigned to TID4. Because there is no prior knowledge about plates, TID4 is classified as Unknown. TID4 is labeled as a Plate and the system conceptualized the category.
\end{enumerate}

This sequence shows that the proposed architecture is capable of detecting new objects, tracking and recognizing those object in various positions. Moreover, it shows capability of human-robot interaction based on a graphical interface and pointing gesture recognition.

All raw data from the RGB-D sensor was recorded in a rosbag, which was then used to test different configurations of our system. Three demonstrations were performed using the recorded rosbag.

%% ------------------------------------------------------------------
\subsection{Profiling} \label{sec:profiling}
%% ------------------------------------------------------------------

In the first demonstration, objects are represented by sets of spin-images. The instance-based learning approach (section~\ref{sec:instance_based_learning}) is adopted. In addition to the basic object and user perception already included in the recorded rosbag, the system now conceptualizes and recognizes object categories. A video of this demonstration is available at: {\cblue{\small \href{https://youtu.be/XvnF2JMfhvc}{https://youtu.be/XvnF2JMfhvc}}} 

Based on this demonstration, different aspects of the performance of the system were
profiled. As discussed in chapter~\ref{chapter_2}, section~\ref{subsec:addressing}, nodelets can significantly improve the efficiency
since they support zero copy transport and they enable simultaneous access to LevelDB.
Figure \ref{fig:processing_time_nodes_nodelets} compares the processing time of the object perception modules. The tracker modules (Fig. \ref{fig:processing_time_nodes_nodelets} (\textit{a}) nodes and (\textit{d}) nodelets) tend to display a stable processing time shortly after their initialization. This is explained by the fact that the size of the input data is more or less stable over time. In this case, nodelets are more
efficient when compared to nodes: for example for pipelines 1, 2 and 3 in the
100 to 150 time interval, nodes display an average processing time of 45
miliseconds, compared to 25 miliseconds in the case of nodelets. Since the
trackers do not access the database, the main factor contributing to the
increase in efficiency is the zero copy transport. 
\begin{figure*}[!t]
\begin{centering}
\begin{tabular}{ccc}
	\includegraphics[width=0.45\linewidth, trim=1.3cm 0.5cm 1.8cm 1.3cm,clip=true]{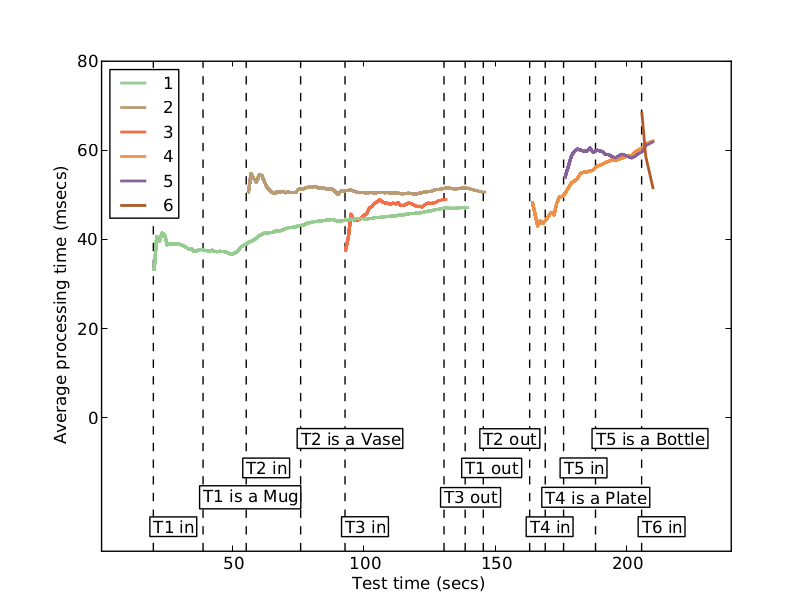}&&
	\includegraphics[width=0.45\linewidth, trim=1.3cm 0.5cm 1.8cm 1.3cm,clip=true]{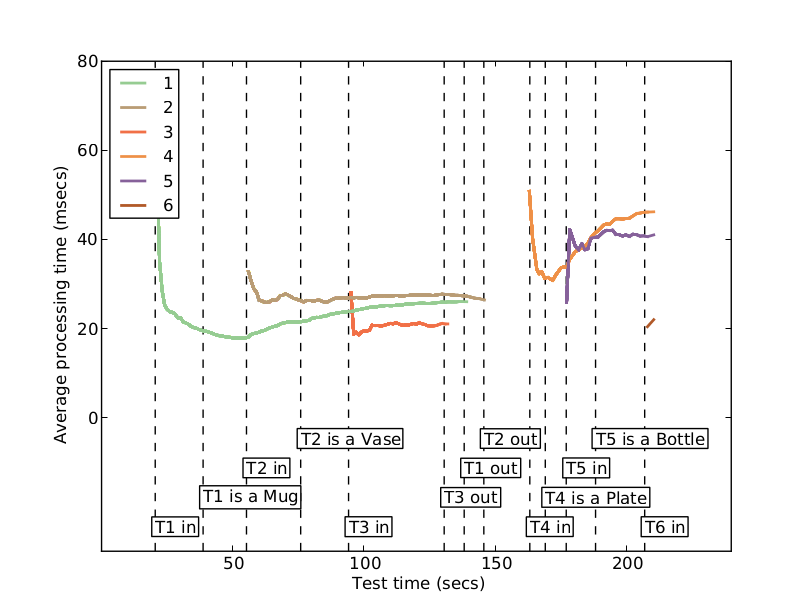}\\
	(\textit{a}) & &(\textit{d})\\
	\includegraphics[width=0.45\linewidth, trim=1.3cm 0.5cm 1.8cm 1.3cm,clip=true]{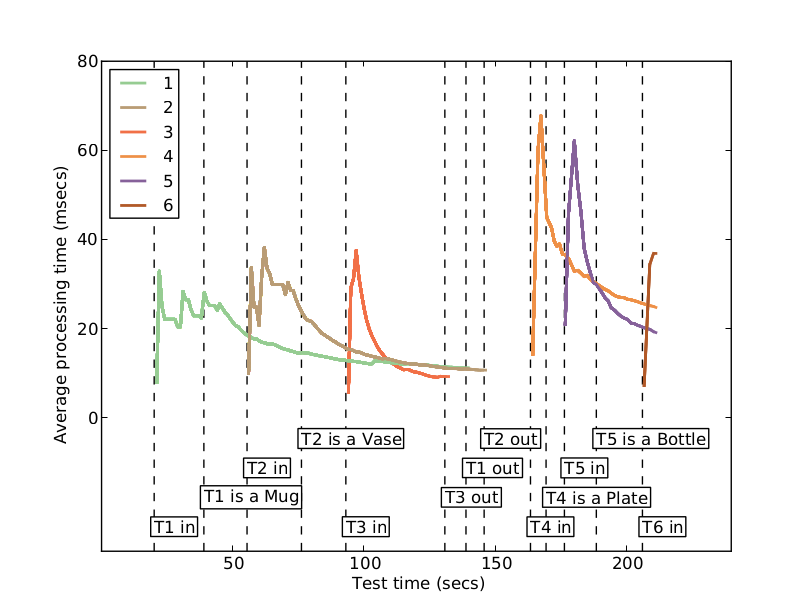}&&
	\includegraphics[width=0.45\linewidth, trim=1.3cm 0.5cm 1.8cm 1.3cm,clip=true]{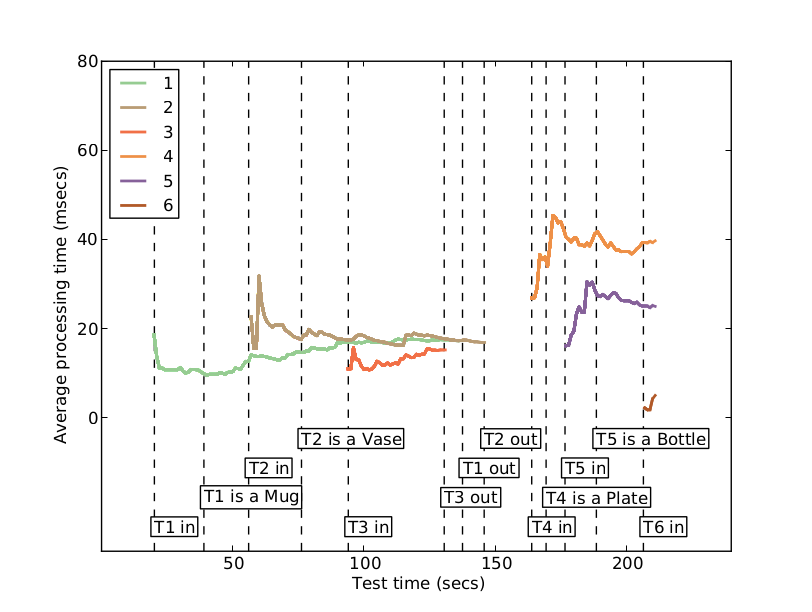}\\
		(\textit{b}) && (\textit{e})\\
	\includegraphics[width=0.45\linewidth, trim=1.3cm 0.5cm 1.8cm 1.3cm,clip=true]{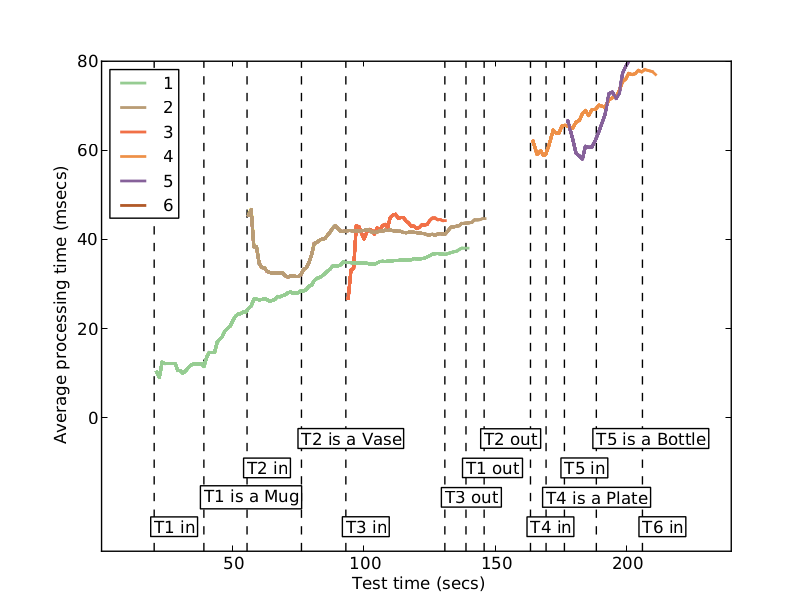}&&
	\includegraphics[width=0.45\linewidth, trim=1.3cm 0.5cm 1.8cm 1.3cm,clip=true]{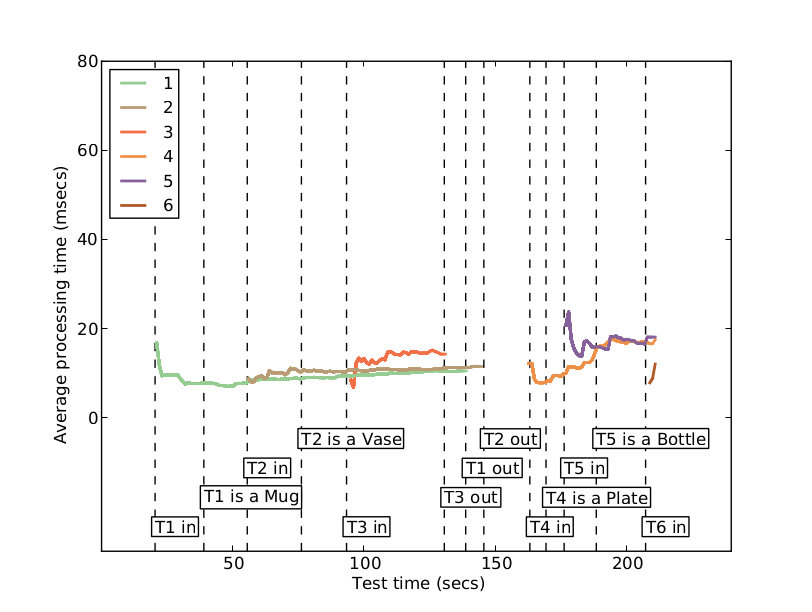}\\
(\textit{c}) & &(\textit{f})
\end{tabular}
\end{centering}
\caption{Processing time in the object perception pipelines 
in the video sequence, comparing nodes (\emph{left column}) with 
nodelets (\emph{right column}). Six objects appear in the video, corresponding 
to pipelines 1 through 6. (\textit{a}) tracker nodes; (\textit{b}) feature extraction nodes; (\textit{c}) object recognition nodes; (\textit{d}) tracker nodelets; (\textit{e}) feature extraction nodelets; (\textit{f}) object recognition nodelets.
}
\label{fig:processing_time_nodes_nodelets}
\vspace{-10pt}
\end{figure*}
The messages that are received (sensor point cloud) and sent (partial object point cloud) by the trackers are of large size, which explains why zero copy transport enables such a significant improvement. The feature extraction modules (Fig. \ref{fig:processing_time_nodes_nodelets} (\textit{b}) nodes and (\textit{e}) nodelets) show a different behaviour. These modules periodically compute the spin-image representation from the partial object point cloud. At some points, the point cloud is signaled to belong to a key view, which will trigger the writting of that representation to the perceptual memory.

%\begin{table}[!b] \small
\begin{wraptable}{r}{0.65\linewidth}\footnotesize
\vspace{-4mm}
\begin{center}
    \caption {Sequence of events in the experiment (see video)}
    \label{table:video_script}
    \centering
    \begin{tabular}{|c|c|c|}
    \hline
    \shortstack{Time}  & Event & Description\\
    \hline
    25 & T1 in & A Mug (T1) is placed on the table\\
    40 & T1 is a Mug & T1 is labelled as Mug\\
    60 & T2 in & A Vase (T2) is placed on the table\\
    75 & T2 is a Vase & T2 is labelled as a Vase\\
90 & T3 in & Another Mug (T3) is placed on the table\\
135 & T3 out & T3 is removed from the table\\
140 & T1 out & T1 is removed from the table\\
145 & T2 out & T2 is removed from the table\\
165 & T4 in & A Plate (T4) is placed on the table\\
170 & T4 is a Plate & T4 is labelled as a Plate\\
175 & T5 in & A Bottle (T5) is placed on the table\\
190 & T5 is a Bottle & T5 is labelled as a Bottle\\
210 & T6 in & A Spoon (T6) is placed on the table\\
    \hline
    \end{tabular}
\end{center}
\vspace{-7mm}
\end{wraptable}
The curves show these points in time with a rapid increase in processing time. Nodelets also display these peaks, but because access to the database is much faster, the peaks are smaller, as is the average processing time.
The object recognition modules (Fig. \ref{fig:processing_time_nodes_nodelets}
		(\textit{c}) nodes and (\textit{f}) nodelets) receive a representation
of the current object view from the feature extraction, and compare it against
the representations of all known category views. Thus, they are continuously
reading the database in the search for an update to the known categories. As a
result, the larger the size of the database, the slower the reading of the
complete set of categories.
However, in the case of nodelets, this deterioration is minor when compared with nodes, since accessing the database is much more efficient.

\begin{figure}[!t]
\begin{centering}
\begin{tabular}{cc}
\includegraphics[width=.47\linewidth, trim=1cm 0.5cm 1.8cm 1.3cm,	clip=true]{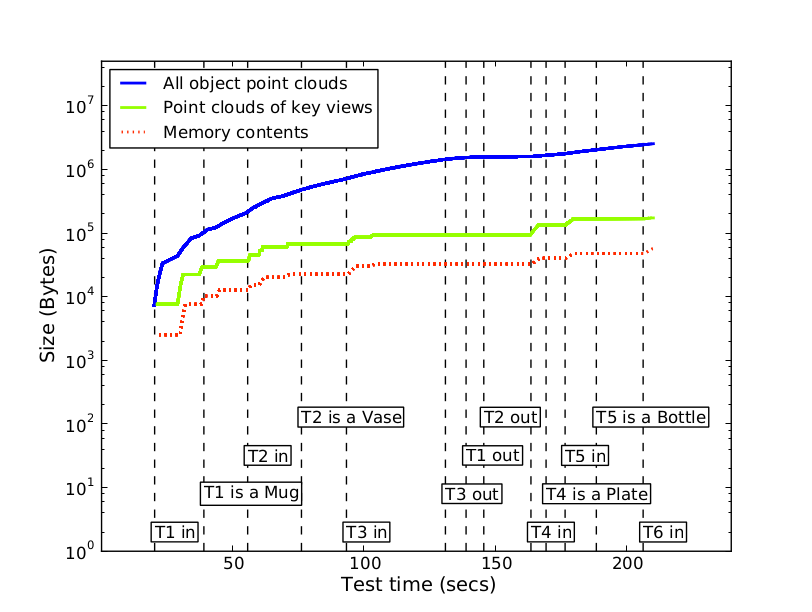}&
\includegraphics[width=.47\linewidth, trim=1cm 0.5cm 1.8cm 1.3cm,	clip=true]{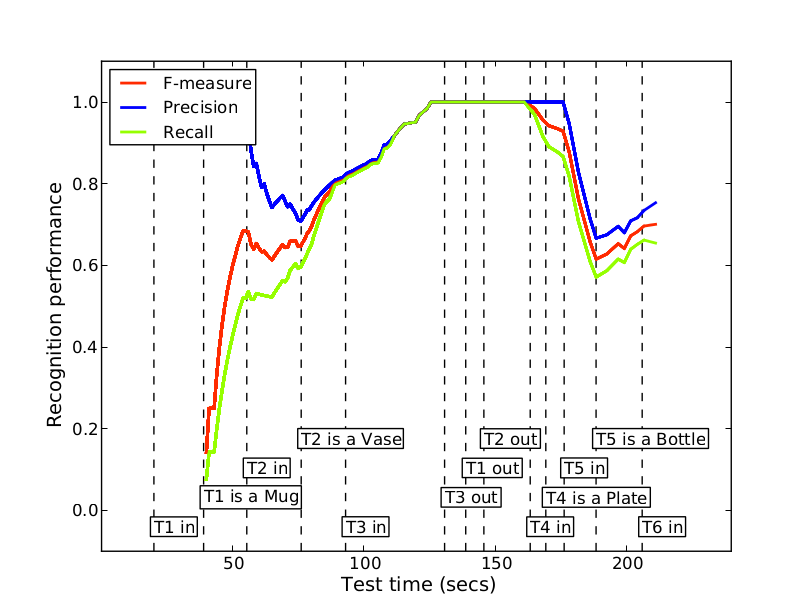}
\end{tabular}
\end{centering}
\caption{Perceptual memory usage in logarithmic scale and object recognition performance (precision, recall and F-measure) during the experiment. (\emph{left}) 
The blue (upper) curve represents the total size of all point clouds of object views extracted by the trackers. The green (middle) curve represents the total accumulated size of all point clouds of key views. The red (bottom) curve represents the actual perceptual memory content (shape-based representations of key views);  (\emph{right}) Each point in these curves is computed based on the object recognition results in the previous 20 seconds.}
\label{fig:recognition_performance}
\vspace{-10pt}
\end{figure}

Figure  \ref{fig:recognition_performance} (\emph{left}) shows the memory usage of the system.
Notice that at the end of the experiment the memory size would be above 1 MB if all object
point clouds extracted by the trackers would be stored (roughly 5 Kb / sec.). In a continously running system, this
rate of data acumulation would be hard to handle, and would not bring 
any real benefit. The total size of the point clouds of all
the selected key views is much smaller (one order of magnitude in 
this experiment). The data actually acumulated in memory
(shape representations based on spin-images) is even smaller.

Figure \ref{fig:recognition_performance} (\emph{right}) shows the evolution of object recognition performance throughout the experiment.
When the first Mug (T1) is placed on the table the system recognizes it as Unknown. After some time, the instructor labels T1 as a Mug and the system starts displaying a precision of 1.0. However, the recall score is under 0.2, because the system classified T1 as Unknown several times before the user labelled the object. 
After the labelling, the recall starts improving continuously. 
The instructor then places a Vase (T2) on the table. 
Because the category Vase has not been taught yet, the performance goes down.
After labelling T2 as Vase, performance starts going up again.
When a second Mug (T3) enters the scene, the system can correctly 
recognize it and the scores continue to increase.
Then, a Plate (T4) enters the scene, causing recall to drop. Successively, the Plate is taught, a Bottle is placed on the
table and then taught, and eventually performance starts going up again.
This illustrates the process of acquiring categories in an open-ended fashion with user mediation.

%%%%%%%%%%%%%%%%%%%%%%%%%%%%%%%%%%%%%%%%%%%%%%%%%%%%%%%%%%%%%%%%%%%%%%%%%
\subsection {Local LDA}

\begin{figure}[!b]
\begin{tabular}{cc}
	\includegraphics[width=.45\linewidth, trim= 0.1cm 0.0cm 0cm 0cm,clip=true]{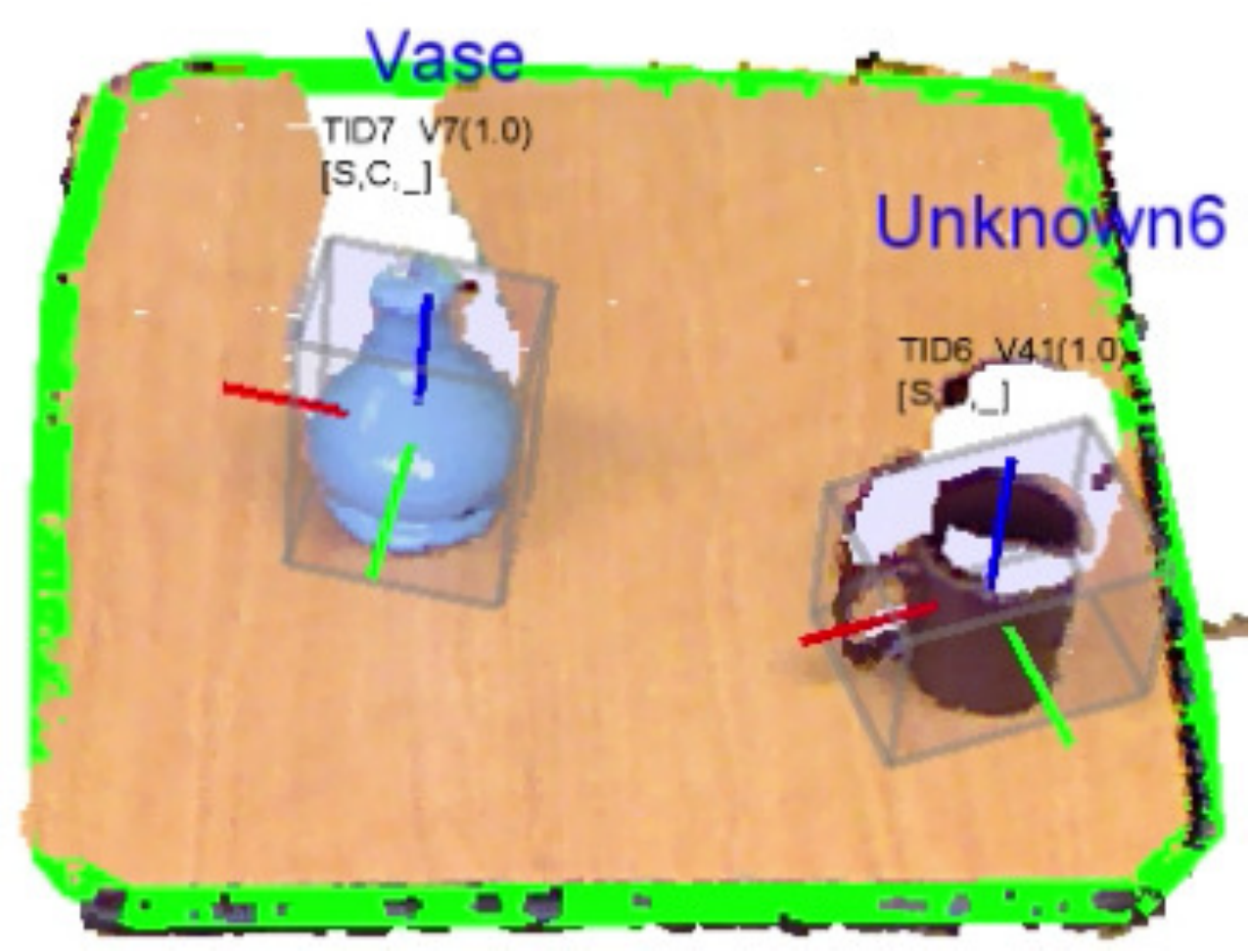}&\quad\quad
	\includegraphics[width=.45\linewidth, trim= 0.1cm 0cm 0cm 0.2cm,clip=true]{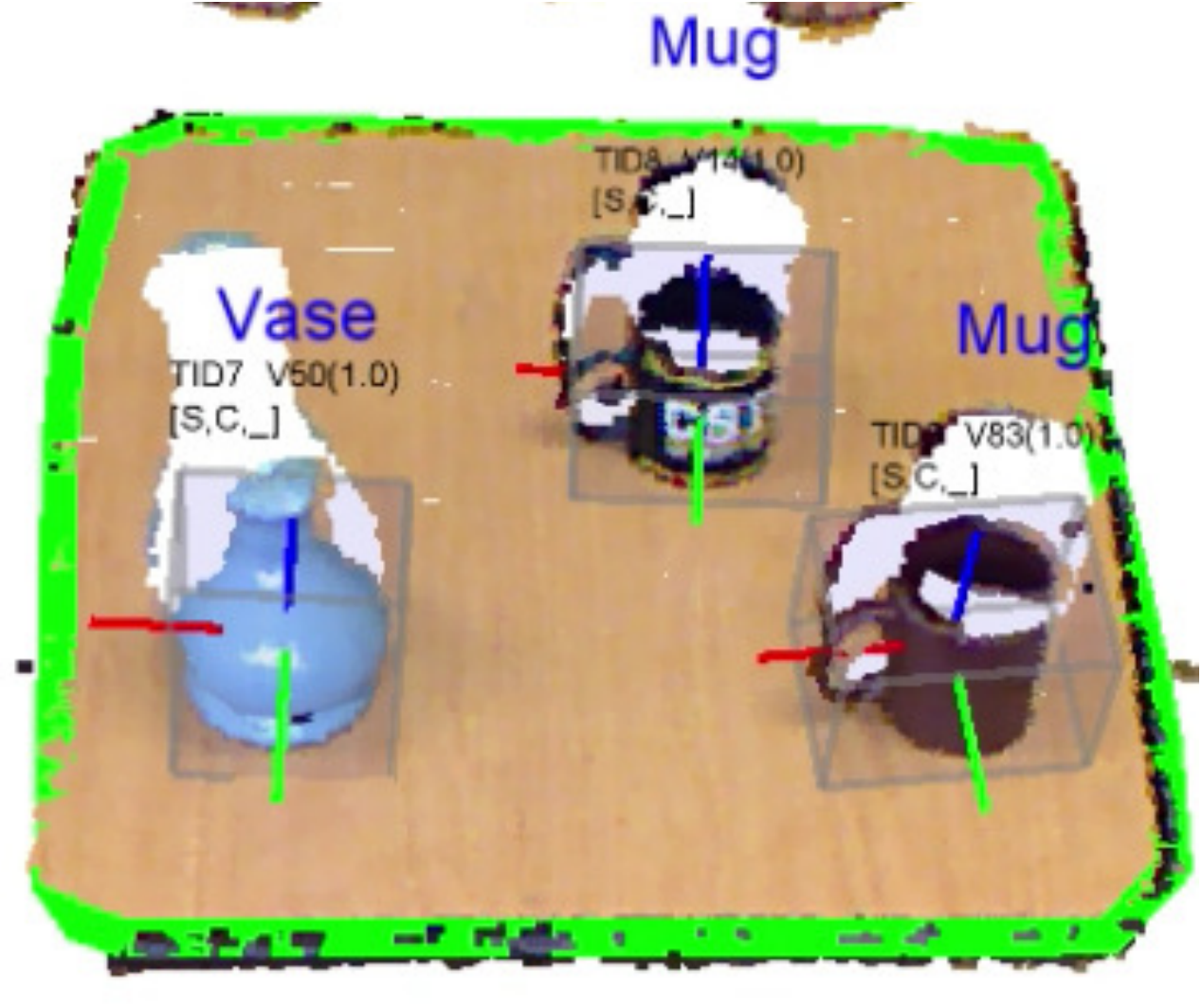}\\
\end{tabular}
\caption{\small Two snapshots showing object recognition results in the demonstration:
these snapshots show the proposed system supports (\emph{left}) classical learning from a batch of train labelled data and (\emph{right}) open-ended learning from on-line experiences. }
\label{fig:system_test}
\end{figure}
In this demonstration, we configured the system to use the Local LDA representation proposed in Chapter~\ref{chapter_4}. Initially, the system only had prior knowledge about the \emph{Vase} and \emph{Dish} categories, learned from batch data (i.e. set of observations with ground truth labels), and there is no information about other categories (i.e. \emph{Mug}, \emph{Bottle}, \emph{Spoon}). Throughout this session, the system must be able to recognize instances of learned categories and incrementally learn new object categories.
%Note, in case of training online, the instructor can be a human user, and in the case of batch learning it can be a set of ground truth labels.  
 Figure~\ref{fig:system_test} illustrates the behaviour of the system: 
\begin{enumerate}[(a)]
\item The instructor puts object TID6 (a \emph{Mug}) on the table. It is classified as \emph{Unknown} because mugs are not known to the system; Instructor labels TID6 as a \emph{Mug}. The system conceptualizes \emph{Mug} and TID6  is correctly recognized. The instructor places a \emph{Vase} on the table. The system has learned the \emph{Vase} category from batch data, therefore, the \emph{Vase} is properly recognized (Fig.\ref{fig:system_test} (\emph{left})).
 \item Later, another \emph{Mug} is placed on the table. This particular \emph{Mug} had not been previously seen, but the system can recognize it, because the Mug category was previously taught  (Fig.\ref{fig:system_test} (\emph{right})). 
\end {enumerate}

This demonstration shows that the system is capable of using prior knowledge to recognize new objects in the scene and learn about new object categories in an open-ended fashion. A video of this demonstration is available at: {\cblue{\small \href{https://youtu.be/J0QOc_Ifde4}{https://youtu.be/J0QOc\_Ifde4}}}.

%%%%%%%%%%%%%%%%%%%%%%%%%%%%%%%%%%%%%%%%%%%%%%%%%%%%%%%%%%%%%%%%%%%%%%%%%
\subsection {GOOD}
\label {System_Demonstration_GOOD}

To show all the described functionalities and properties of the GOOD descriptor, another demonstration was performed using the recorded rosbag. For this purpose, GOOD has been integrated in the RACE object perception system presented in Chapter \ref{chapter_2} (see Fig.\ref{fig:perceptionsystem}) \citep{KasaeiInteractive2015,oliveira20153d,oliveira2014perceptual}. 
%In this demonstration a table is in front of a robot and two users interact with the system. During the demonstration, users presented objects to the system and provided the respective category labels. Therefore, throughout this session, the system must be able to detect, conceptualize and recognize unknown (i.e. new) objects.
 It should be noted that a constraint has been set on the Z axis that the initial direction of $Z$ axis of objects' LRF should be similar to the $Z$ axis direction of the table LRF. There are no learned categories in memory at the beginning of the demonstration. 
It was observed that the proposed object descriptor is capable to provide distinctive global feature for recognizing different types of objects. It also estimates poses of objects and build orthographic projections for object manipulation purposes (see Fig.~\ref{fig:system_test_GOOD}). A video of this demonstration is available in: {\small \href{https://youtu.be/iEq9TAaY9u8}{\cblue{https://youtu.be/iEq9TAaY9u8}}}.

\vspace{3mm}
\begin{figure}[!b]
\centering
\begin{centering}
\begin{tabular}{cc}
	\includegraphics[width=0.4\linewidth, trim= 0.cm 0cm 0cm 0.0cm,clip=true]{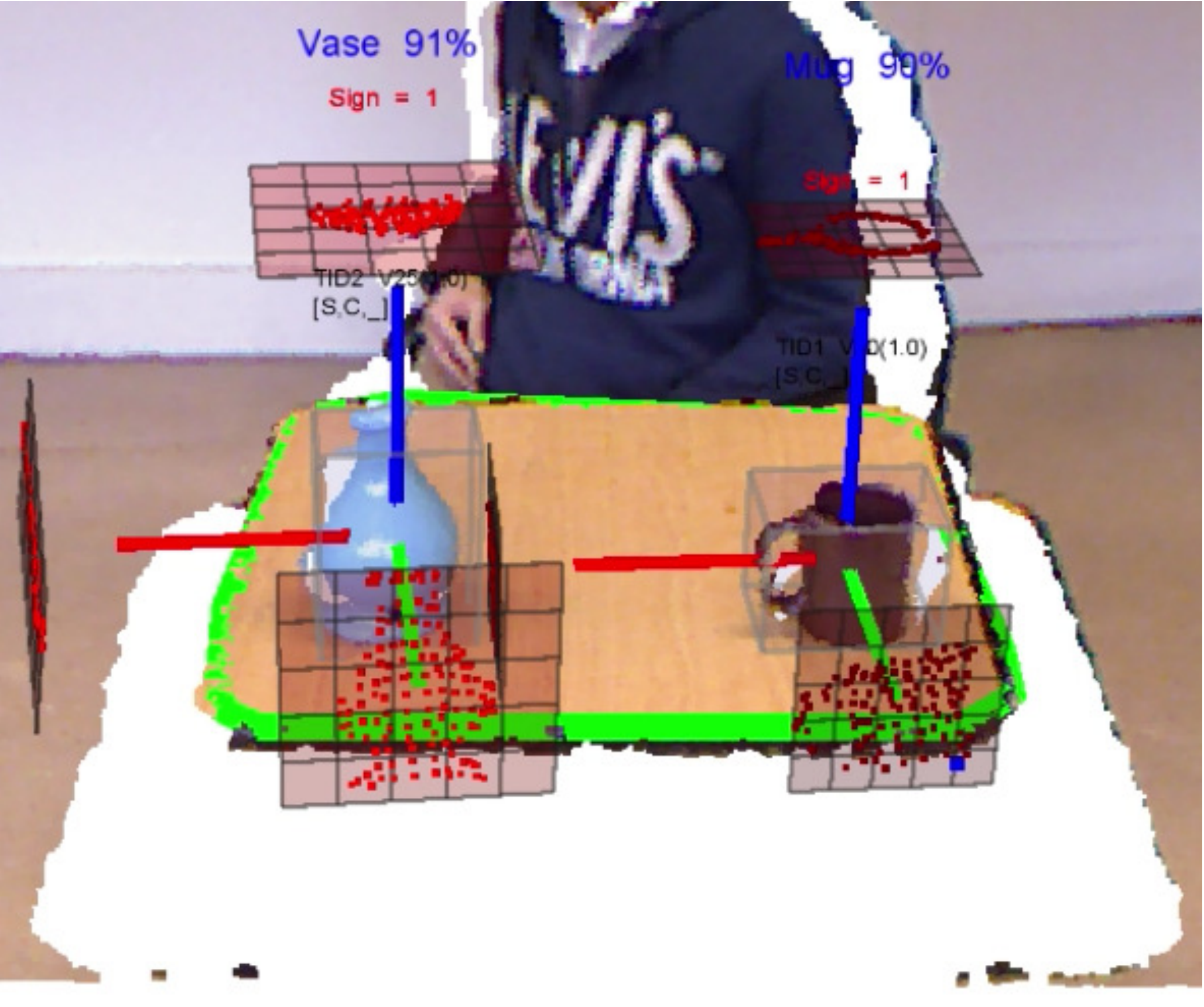}&\quad\quad\quad\quad
	\includegraphics[width=0.41\linewidth]{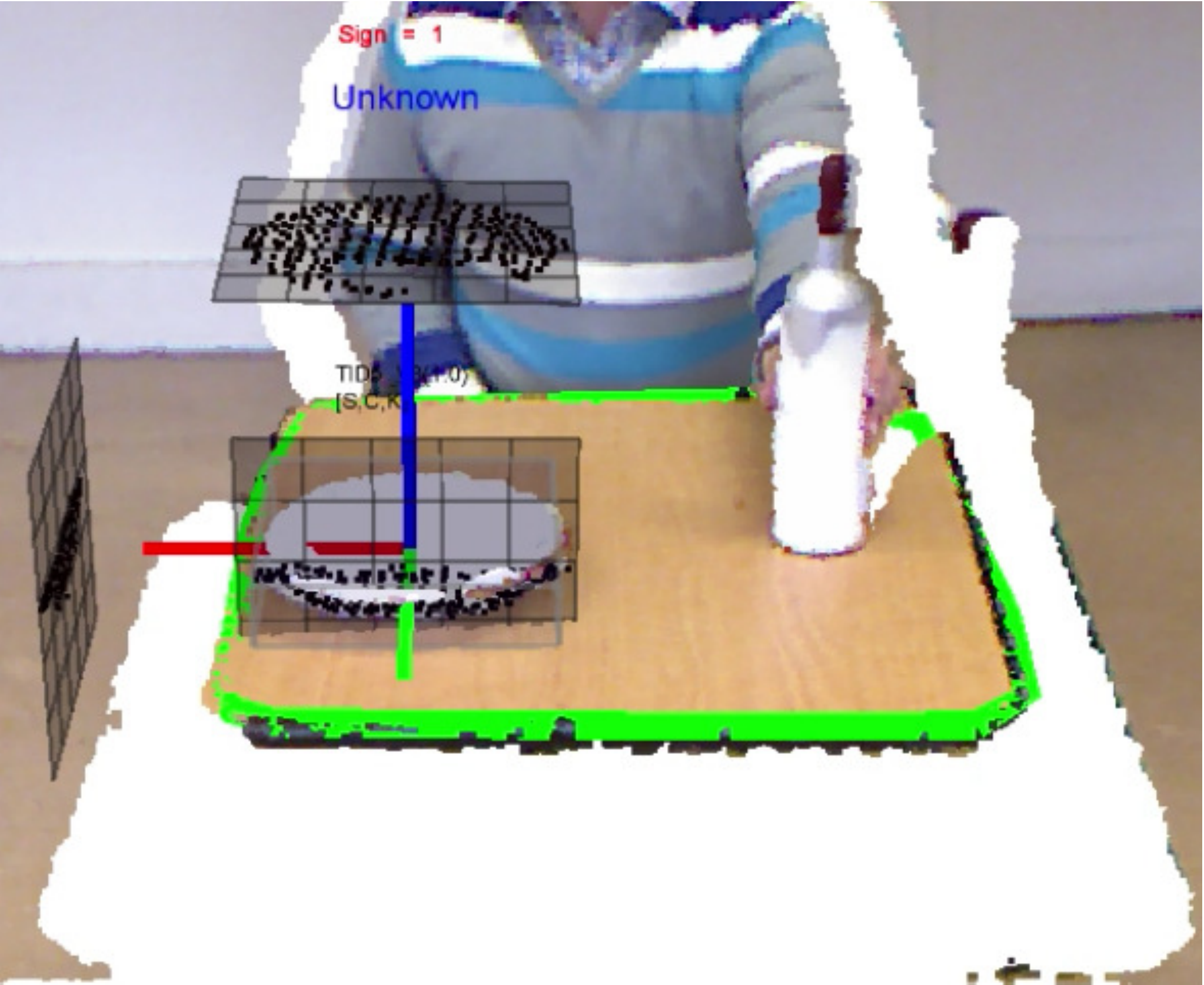}\\
\vspace{-3mm}
\end{tabular}
\end{centering}
\caption{Two snapshots showing the object perception system performing object recognition and pose estimation using the GOOD descriptor;  
The instructor puts a \emph{Mug}, a \emph{Vase} and a \emph{Plate} on the table. The grey bonding boxes and red, green and blue lines signal the pose of the object and the GOOD descriptions are visualized and computed; this frame shows that the system is able to compute the GOOD description and estimate pose of objects in the scene.}
\label{fig:system_test_GOOD}
\end{figure}

%%%%%%%%%%%%%%%%%%%%%%%%%%%%%%%%%%%%%%%%%%%%%%%%%%%%%%%%%%%%%%%%%%%%%%%%%
\section{Assistive Robotic Scenarios: Coupling Perception and Manipulation}

\begin{figure}[!b]
\begin{tabular}{ccc}
	\includegraphics[width=.4\linewidth,trim= 2cm 1cm 0cm 0cm, clip=true]{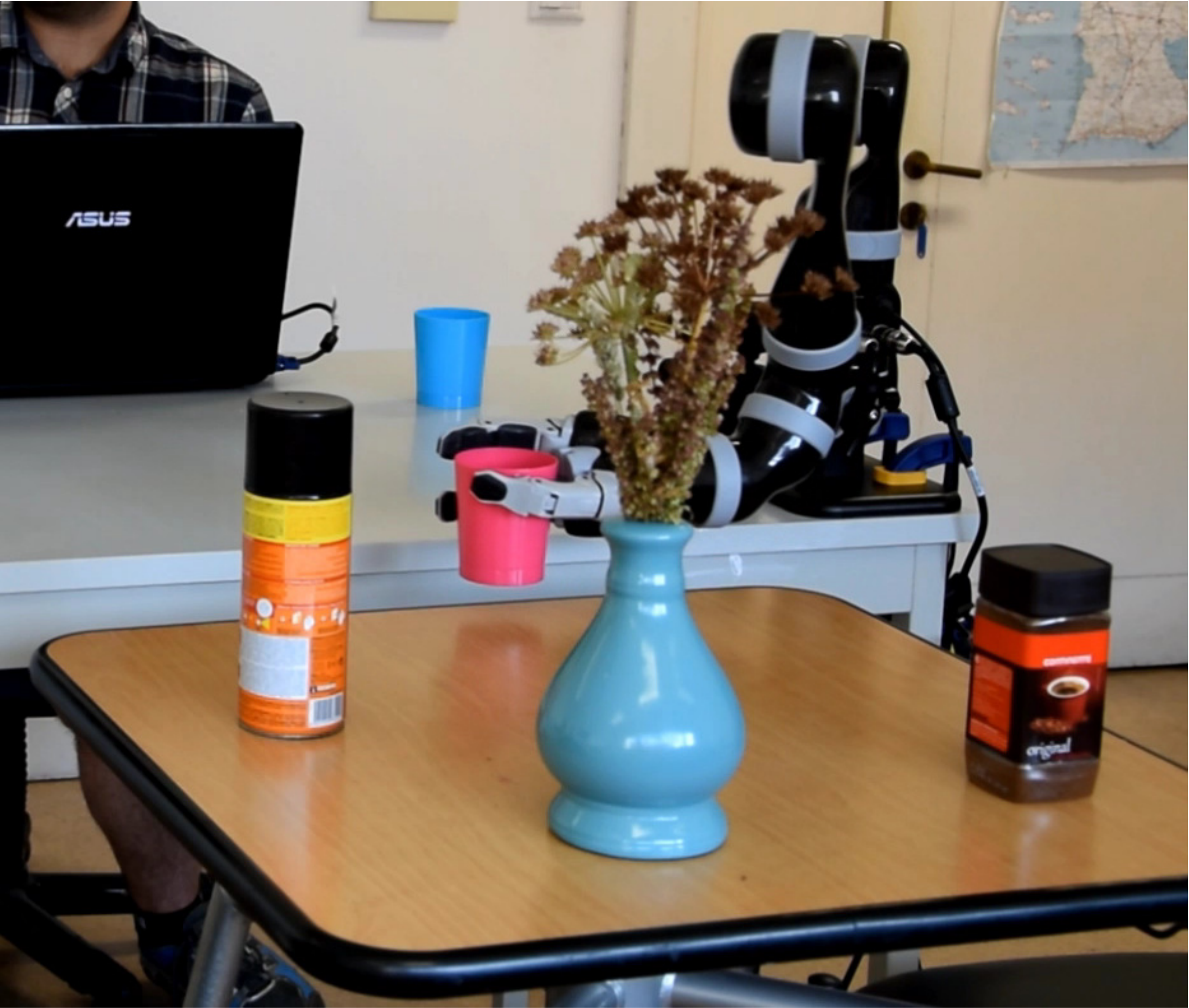}&&\quad\quad
	\includegraphics[width=.5\linewidth,trim= 0.68cm 0cm 0cm 1.5cm, clip=true]{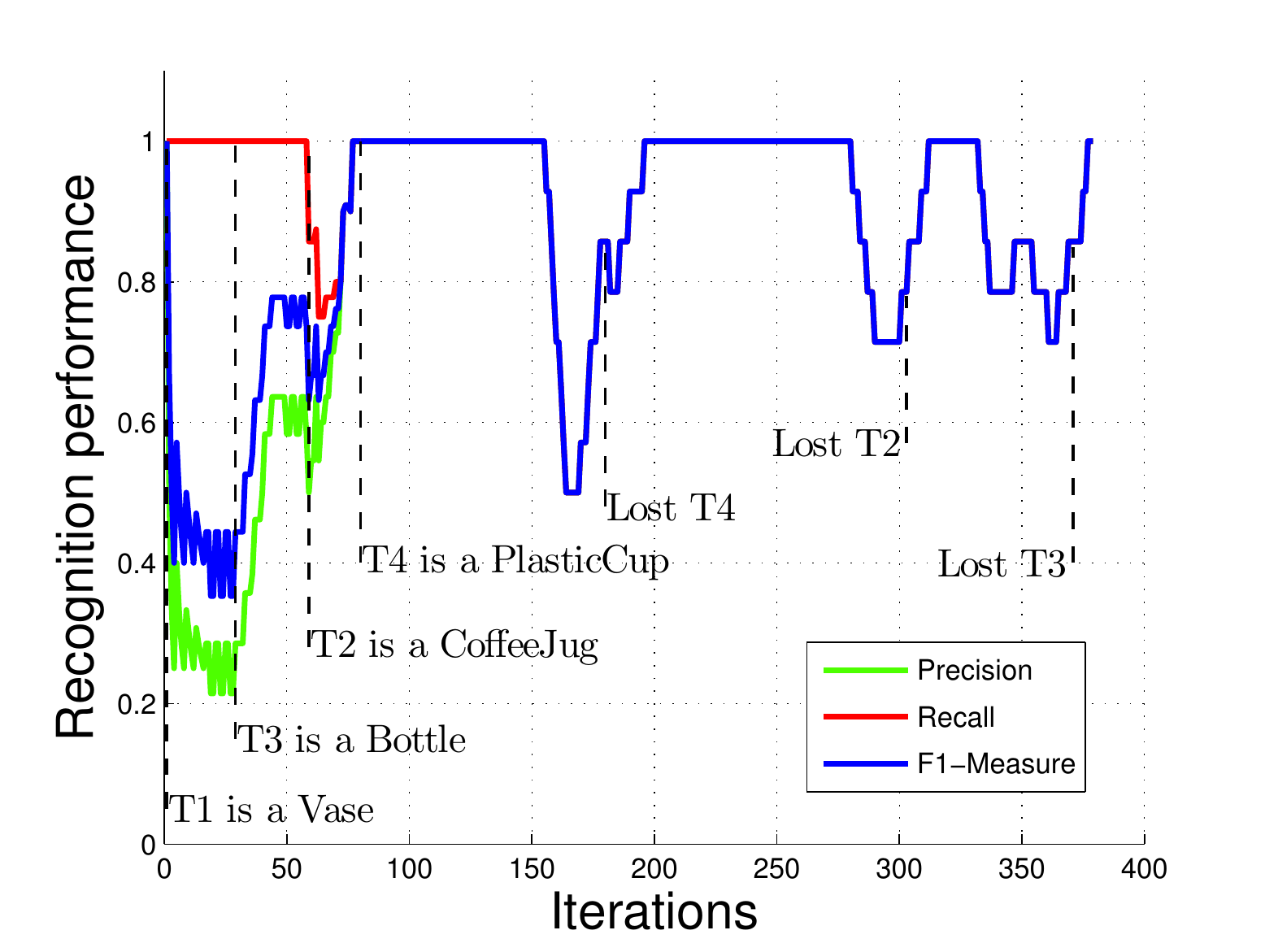}		
\end{tabular}
\caption{System performance during the \emph{clear\_table} demonstration; (\emph{left}) JACO arm manipulates a \emph{plastic cup}; (\emph{right}) Object recognition performance. Each point in these curves is computed based on the object recognition results in the previous 15 iterations.}
\label{fig:jaco}
\vspace{-4mm}
\end{figure}

Elderly, injured, and disabled people have consistently attributed a high priority to object manipulation tasks \citep{jain2010assistive}. Object manipulation tasks consist of two phases: the first is the perception of the object and the second is the planning and execution of arm or body motions for grasping the object and carrying out the manipulation task. These two phases are closely related: object perception provides information to update the model of the environment, while planning uses this world model information to generate sequences of arm movements and grasp actions for the robot. In addition, assistive robots must perform the tasks in reasonable time. It is also expected that the competence of the robot increases over time, that is, robots must robustly adapt to new environments by being capable of handling new objects.

To show the strength of the proposed perception system, two qualitative analysis of the coupling between perception and manipulation for service robots are shown and analysed in this section. In this case, two demonstrations are described, where users manipulate objects on a table and interact with the system to instruct the robot to perform a \emph{``clear\_table''} and \emph{``serve a meal''}
task as well as teach object categories to the robot. In both demonstrations, a naive Bayes learning approach with a Bag-of-Words object representation are used to acquire
and refine object category models. Moreover, a JACO robotic arm manufactured by KINOVA, as depicted in Fig. \ref{fig:serve_a_meal_setup}, is used. It has six degrees of freedom and a three fingers gripper. Since the JACO arm can carry up to 1.5kg\footnote{http://www.kinovarobotics.com}, it is ideal for manipulating everyday objects. Moreover, infinite rotation around the wrist joints allows for flexible and effective interaction in a domestic environment.  It should be noted that grasping itself is not in the scope of this thesis. Previously, we showed how to grasp household objects in different situations \citep{kasaei2016object,shafii2016learning}.

\subsection {``\emph{Clear\_Table}'' Scenario}
In this demonstration, the system works in a scenario where a table is in front of the robot and a user interacts with the system. Note that, 
at the start of the experiment, the set of categories known to the system is empty. During the session, a user presents objects to the system and provides the respective category labels. The user then instructs the robot to perform a \emph{clear\_table} task (i.e. puts the table back into a clear state). To achieve this task, the robot must be able to detect and recognize different objects and transport all objects except standard table items (e.g. table sign, flower, etc.) to predefined areas. While there are active objects on the table, the robot retrieves the world model information from the \emph{Working Memory} including label and position of all active objects. The robot then selects the object which is closer to the arm's base and clears it from the table (see Fig.~\ref{fig:pick_place} \emph{left}). 
In case the system predicts a category that is not the true object category, both a false negative (true object category not detected) and a false positive (predicted object category not correct) are accounted for. Figure~\ref{fig:jaco} (\emph{right}) shows the evolution of object recognition performance throughout the experiment. 
First, the system recognizes all table-top objects as \emph{Unknown}. After some time, the instructor labels T1 as a \emph{Vase} and the system starts displaying a recall of 1.0. However, the precision starts to decrease, because the categories \emph{Bottle}, \emph{Coffee Jug} and \emph{Plastic Cup} have not been taught yet, the performance goes down. 
After labelling the objects, the precision starts improving continuously. As it is shown in the Fig.~\ref{fig:jaco} (\emph{right}), whenever the robot grasps an object (i.e. iterations 155, 280, 332), the shape of the object is partially changed, misclassification might happen and the performance goes down. The grasped object is then transported to the placing area and the tracking of the object is lost (i.e. iterations 181, 302, 375). Afterwards, the performance starts going up again. 
A video of this session is available at: {\cblue{\small \href{https://youtu.be/cTK10iNyYXg}{https://youtu.be/cTK10iNyYXg}}}. 

We also provide another demonstration for the \emph{clear\_table} task. A video of this demonstration can be found at: {\cblue{\small \href{https://youtu.be/LZtI-s95uTk}{https://youtu.be/LZtI-s95uTk}}}

\begin{figure}[!b]
\begin{tabular}{cc}
	\includegraphics[width=.4\linewidth, trim= 0.5cm .5cm 0.8cm 1.2cm,clip=true]{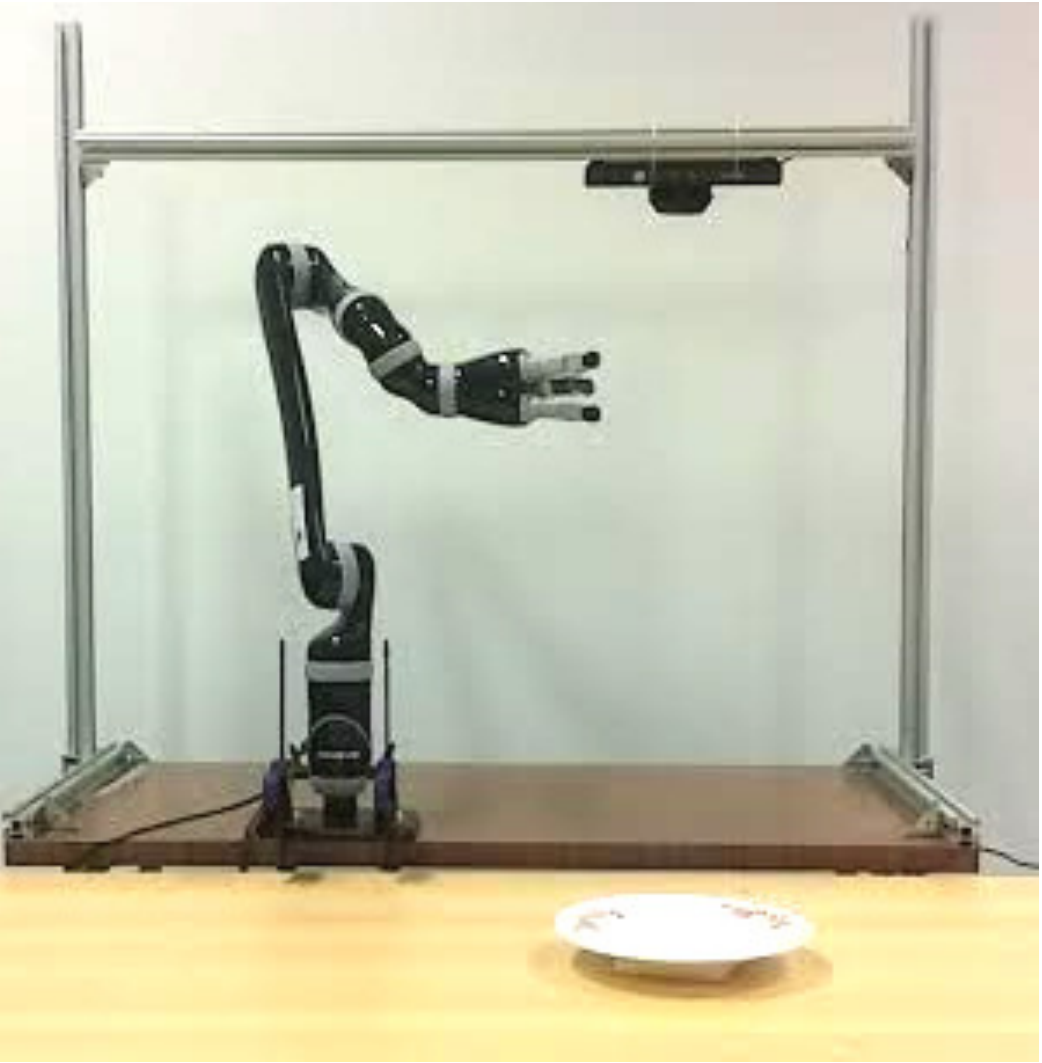}&\quad\quad
	\includegraphics[width=.5\linewidth, trim= 0.1cm 0cm 0cm 0.2cm,clip=true]{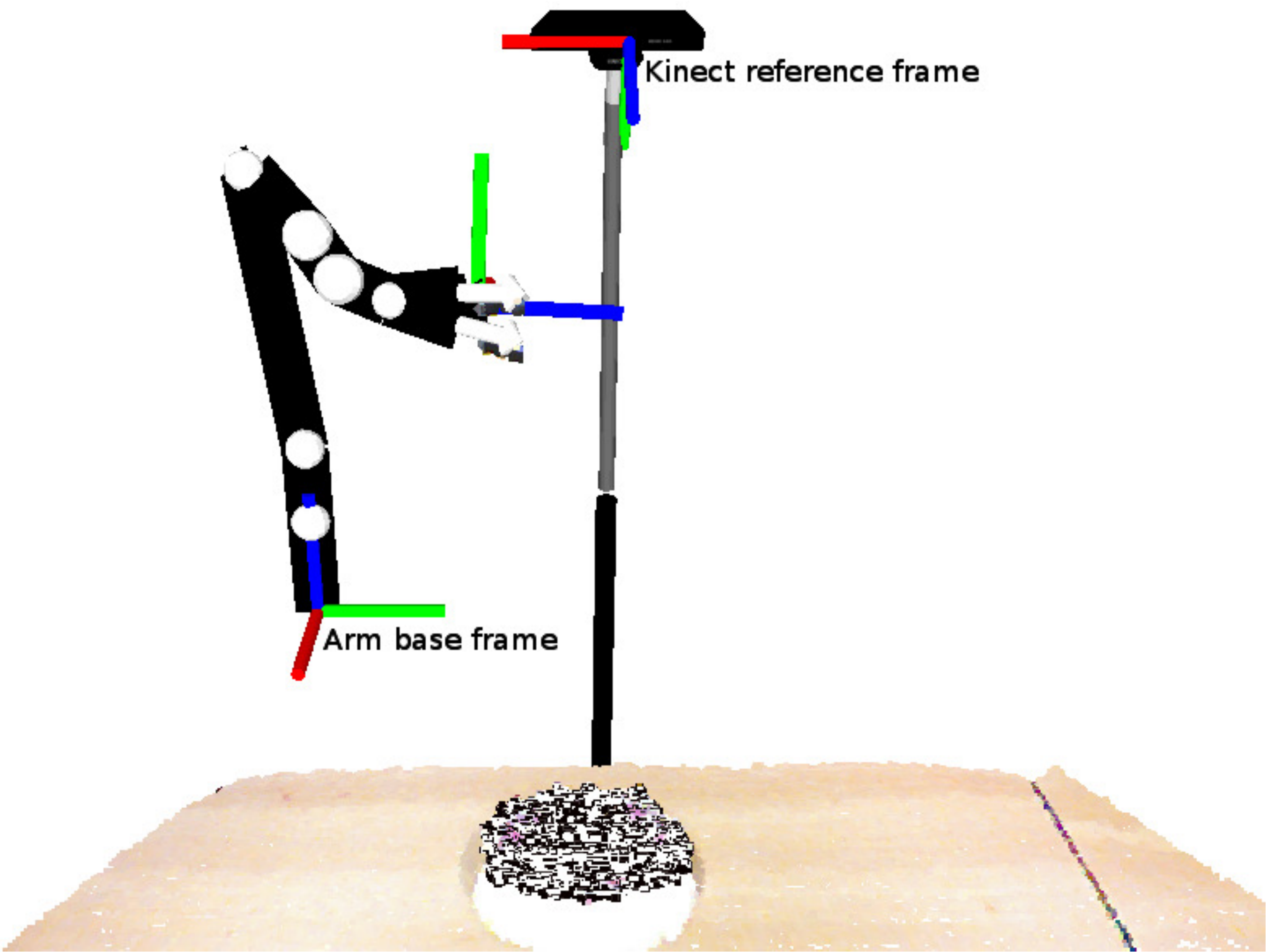}\\
\end{tabular}
\caption{Reference frames in our robotic setup for the \emph{serve\_a\_meal} scenario.}
\label{fig:serve_a_meal_setup}
\end{figure}

\subsection {``\emph{Serve\_A\_Meal}'' Scenario}
Similar to the previous demonstration, a user presents objects to the system and provides the respective category labels throughout the session. The user then instructs the robot to perform a \emph{serve a meal} task (i.e. puts different restaurant objects on the table in front of the user). The setup of our robotic system and the system reference frames for the \emph{serve a meal} scenario are shown in Fig. \ref{fig:serve_a_meal_setup}. To achieve this task, the robot must be able to detect and recognize different objects and transport the objects to the predefined areas and completely serve a meal. For this purpose, the robot retrieves the world model information from the \emph{Working Memory}, including label and position of all active objects. The robot then chooses the object  that is nearest to the \emph{arm base frame} and serves it to the user. A video of this demonstration is available at: {\cblue{\small \href{https://youtu.be/GtXBiejdccw}{https://youtu.be/GtXBiejdccw}}}. 

These small demonstrations show that the developed system is capable of detecting new objects, tracking and recognizing them, as well as manipulating objects in various positions. Moreover, it shows how human-robot interaction is currently supported.

\section{Demonstrations using Scenes Datasets}
Two demonstrations have been performed using the Washington RGB-D Scenes Dataset v2~\citep{lai2014unsupervised} and the Imperial College Domestic Environment Dataset~\citep{Doumanoglou2016}. In particular, the first demonstration was conducted using Washington RGB-D Scenes Dataset~\citep{lai2014unsupervised} to show the strength of the proposed Local LDA representation. In the second demonstration, a set of tests was executed to measure the accuracy of the proposed GOOD descriptor on the Imperial College Domestic Environment Dataset \citep{Doumanoglou2016}. In both demonstrations, an instance-based learning approach is used, i.e., object categories are represented by sets of known instances. Similarly, a simple baseline recognition mechanism in the form of an Euclidean nearest neighbor classifier is used. These demonstrations show that the proposed system supports classical learning from a batch of train labeled data and open-ended learning from actual experiences of a robot.

\subsection{Demonstration I}
 
A real demonstration was carried out using the Washington RGB-D Scenes Dataset v2~\citep{lai2014unsupervised}. This dataset consists of 14 scenes containing a subset of the objects in the RGB-D Object Dataset, including \emph{bowls}, \emph{caps}, \emph{mugs}, \emph{soda}, \emph{cans} and \emph{cereal boxes}. For this demonstration, an instance-based learning approach with the Local LDA has been integrated into the object perception system presented in chapter~\ref{chapter_2}. It is worth mentioning that in Local LDA each object view was described as a random mixture over a set of latent topics, and each topic was defined as a discrete distribution over visual words. 
The system initially had no prior knowledge. The four first objects are introduced to the system using the first scene and the system conceptualizes those categories. The system is then tested using the second scene of the dataset and it can recognize all objects except cereal boxes, because this category was not previously taught. The instructor provided corrective feedback and the system conceptualized the cereal boxes category. Afterwards, all objects are classified correctly in all 12 remaining scenes. This evaluation illustrates the process of acquiring categories in an open-ended fashion. Results are depicted in Fig.\ref{fig:Washington_scene}. A video of this demonstration is online at: {\cblue{\small \cblue{ \href{https://youtu.be/pe29DYNolBE}{https://youtu.be/pe29DYNolBE}}}}.

\begin{figure}[!t]
\begin{tabular}{ccc}
	\includegraphics[width=.3\linewidth, trim= 3cm 0cm 8cm 0cm, clip=true]{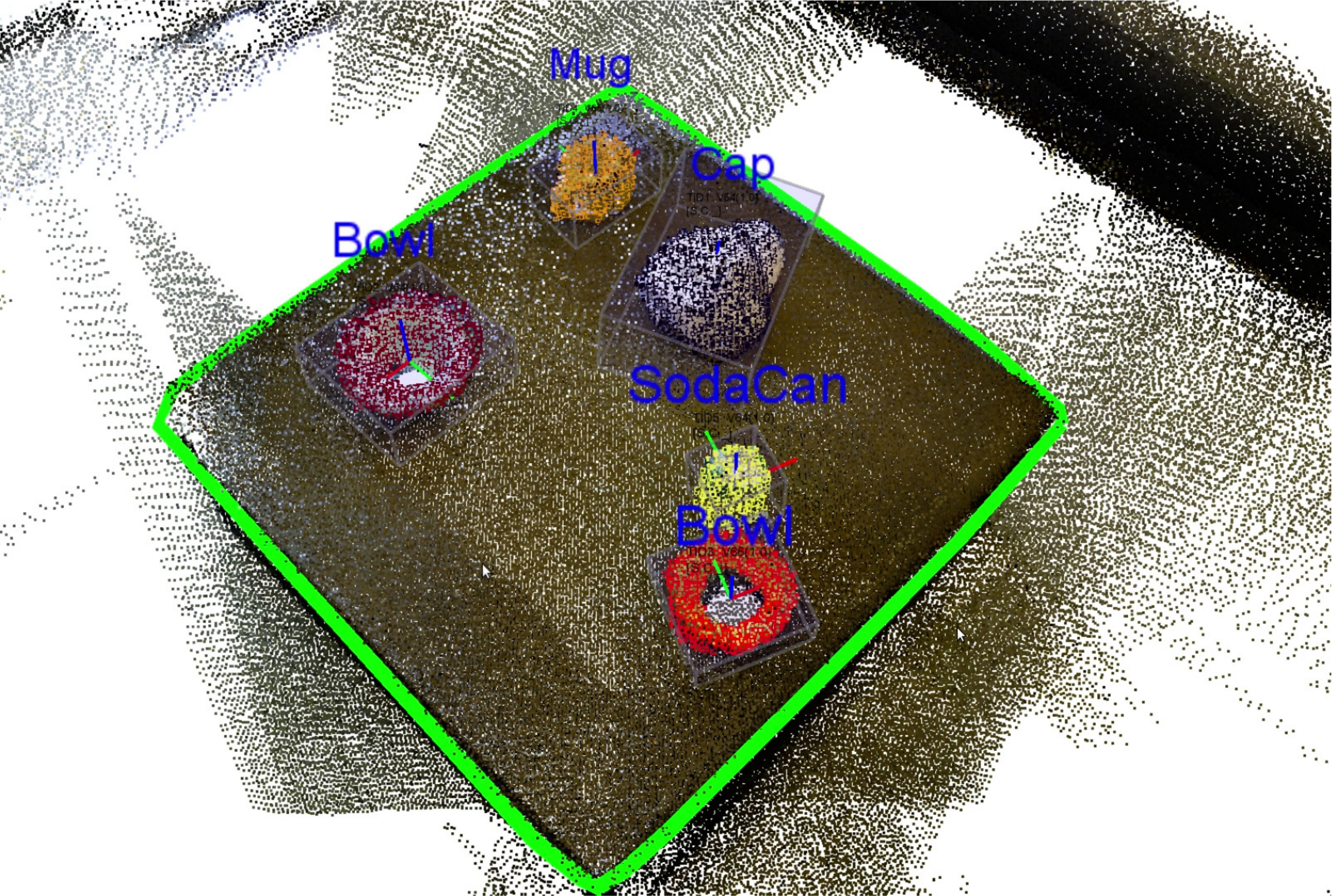}&
	\includegraphics[width=.3\linewidth, trim= 9cm 4cm 6cm 0cm, clip=true]{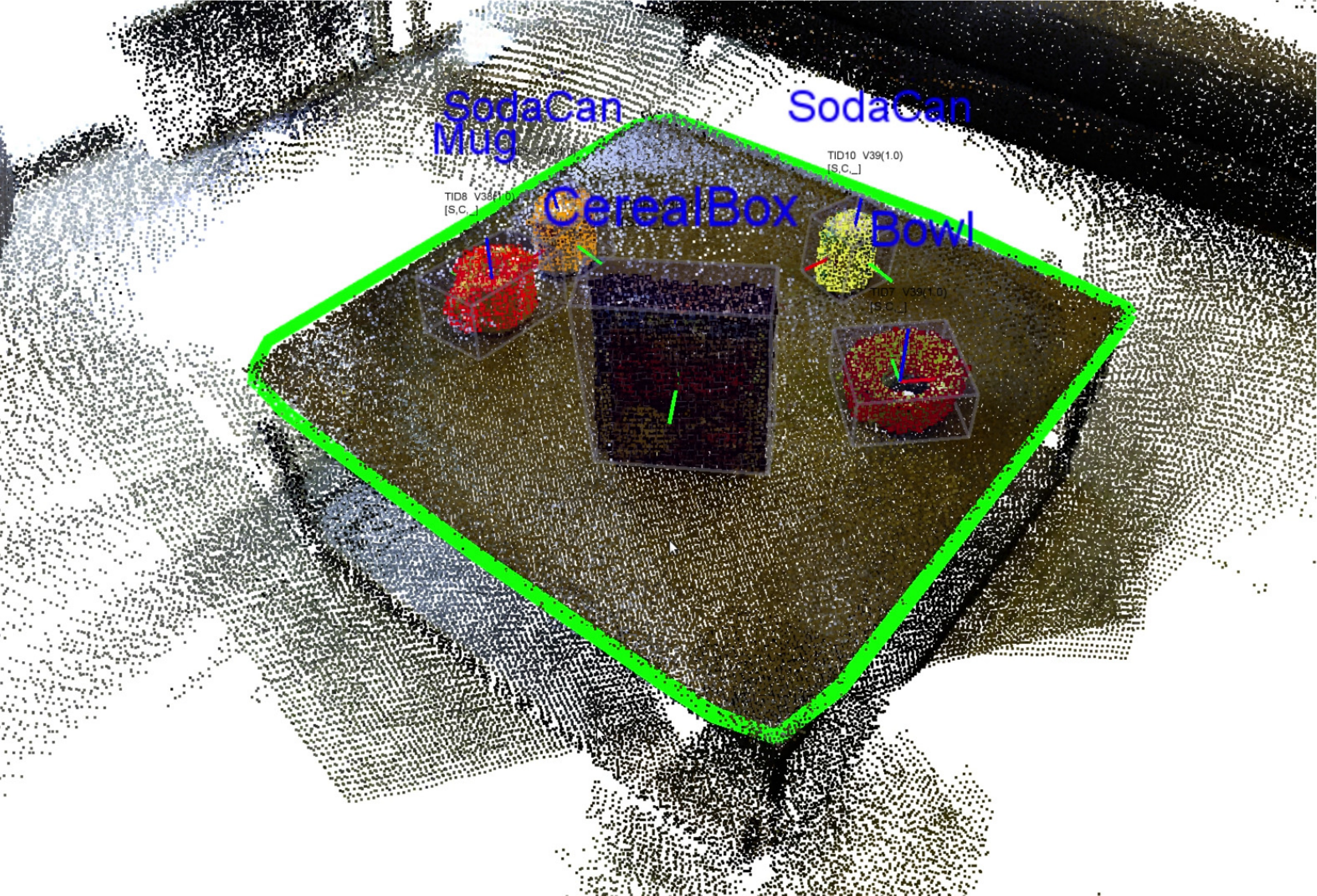}&
	\includegraphics[width=.3\linewidth, trim= 10cm 5cm 10cm 4cm, clip=true]{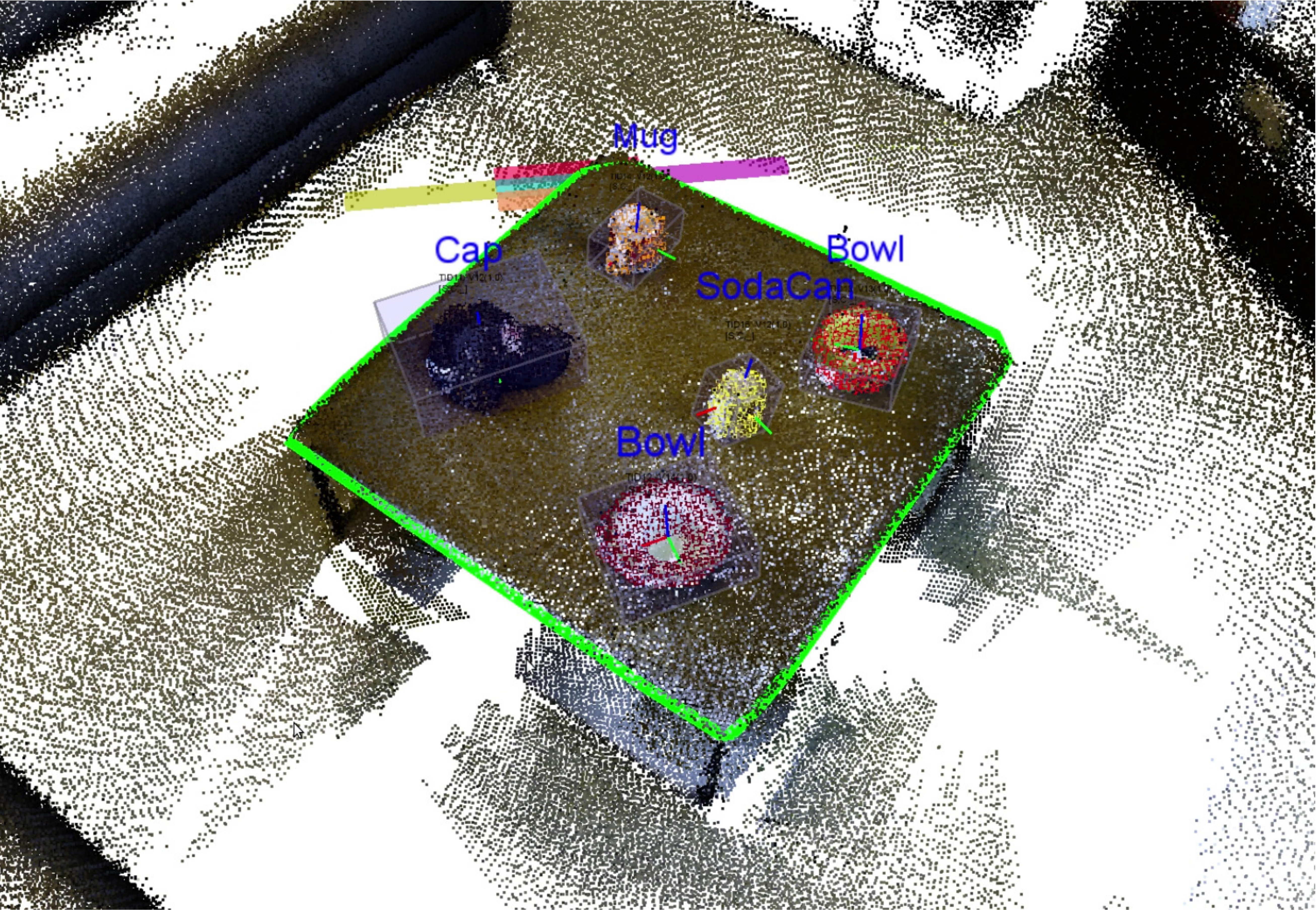}\\
	\includegraphics[width=.3\linewidth, trim= 3cm 0cm 8cm 0cm, clip=true]{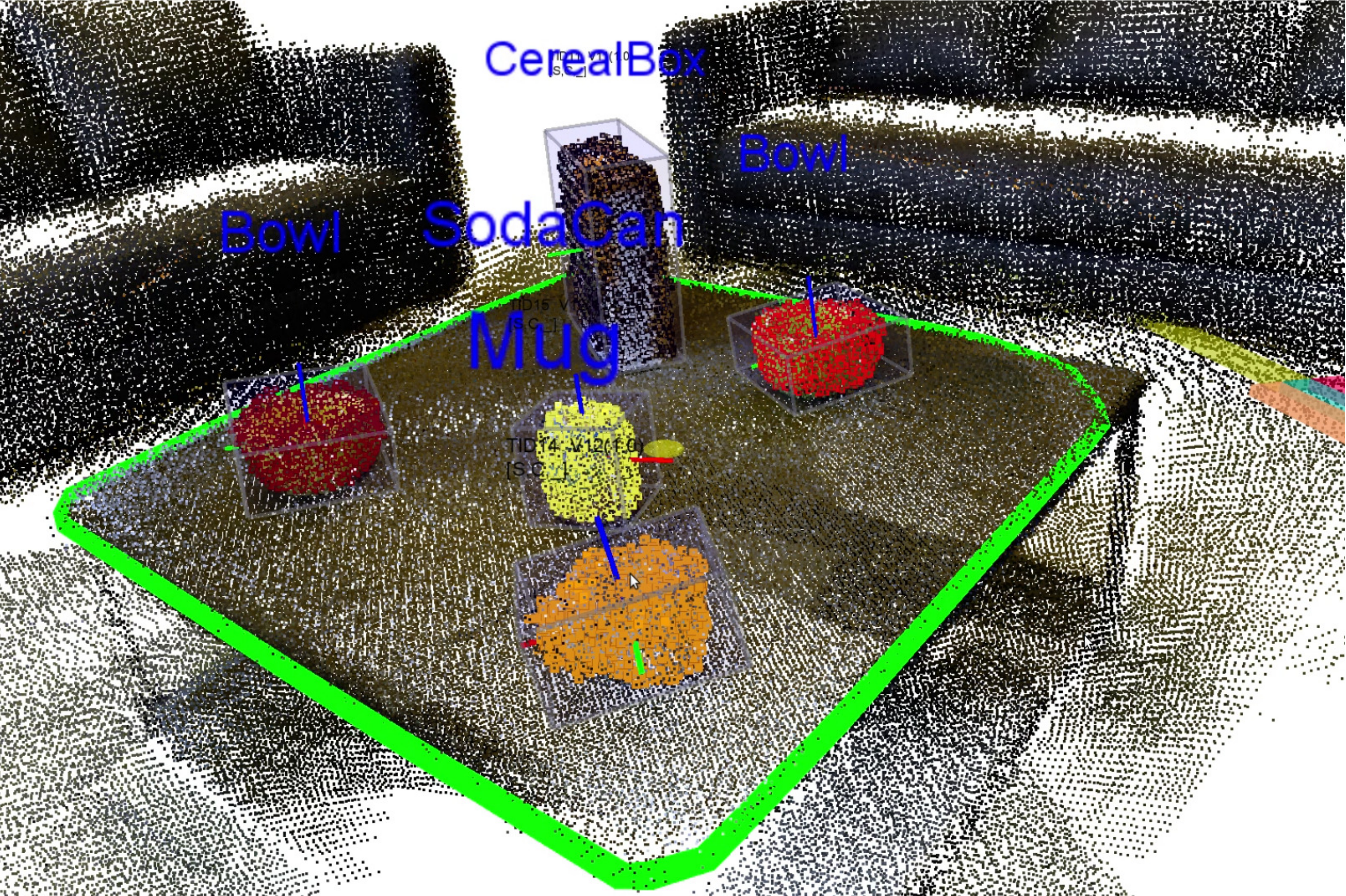}&
	\includegraphics[width=.3\linewidth, trim= 9cm 5.5cm 5cm 1cm, clip=true]{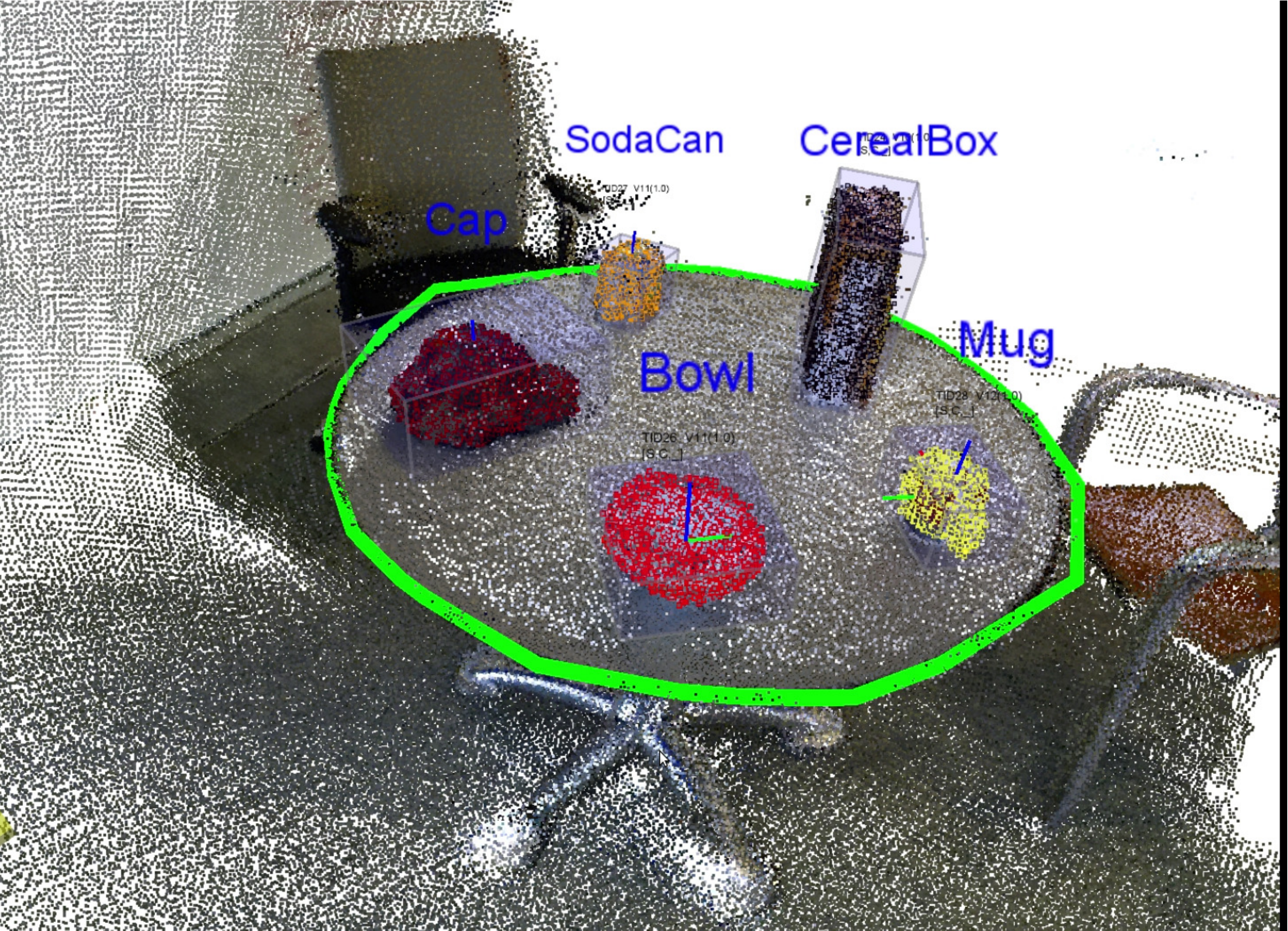}&
	\includegraphics[width=.3\linewidth, trim= 5cm 3cm 8cm 2cm, clip=true]{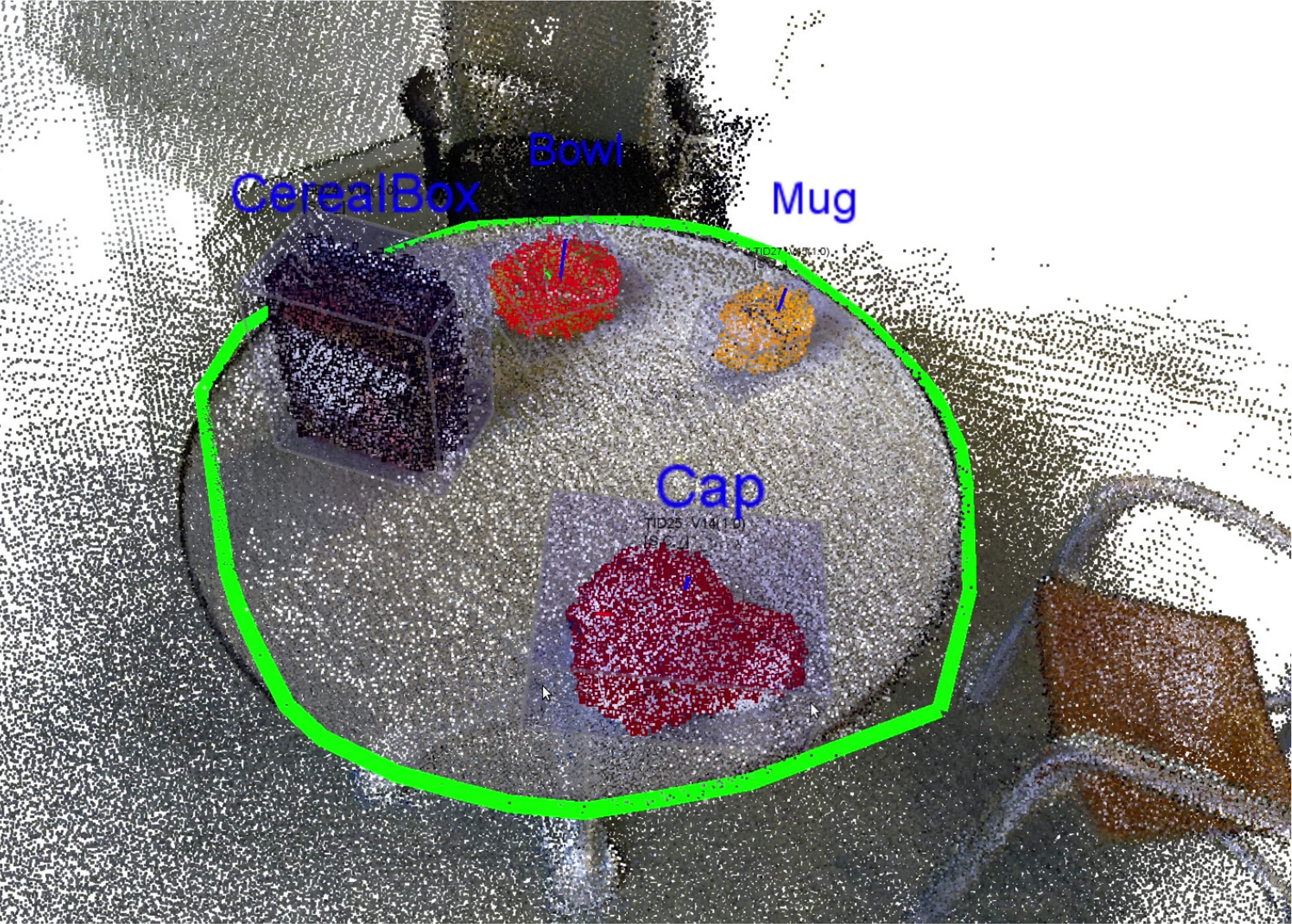}\\
	\includegraphics[width=.3\linewidth, trim= 7cm 8cm 11cm 2cm, clip=true]{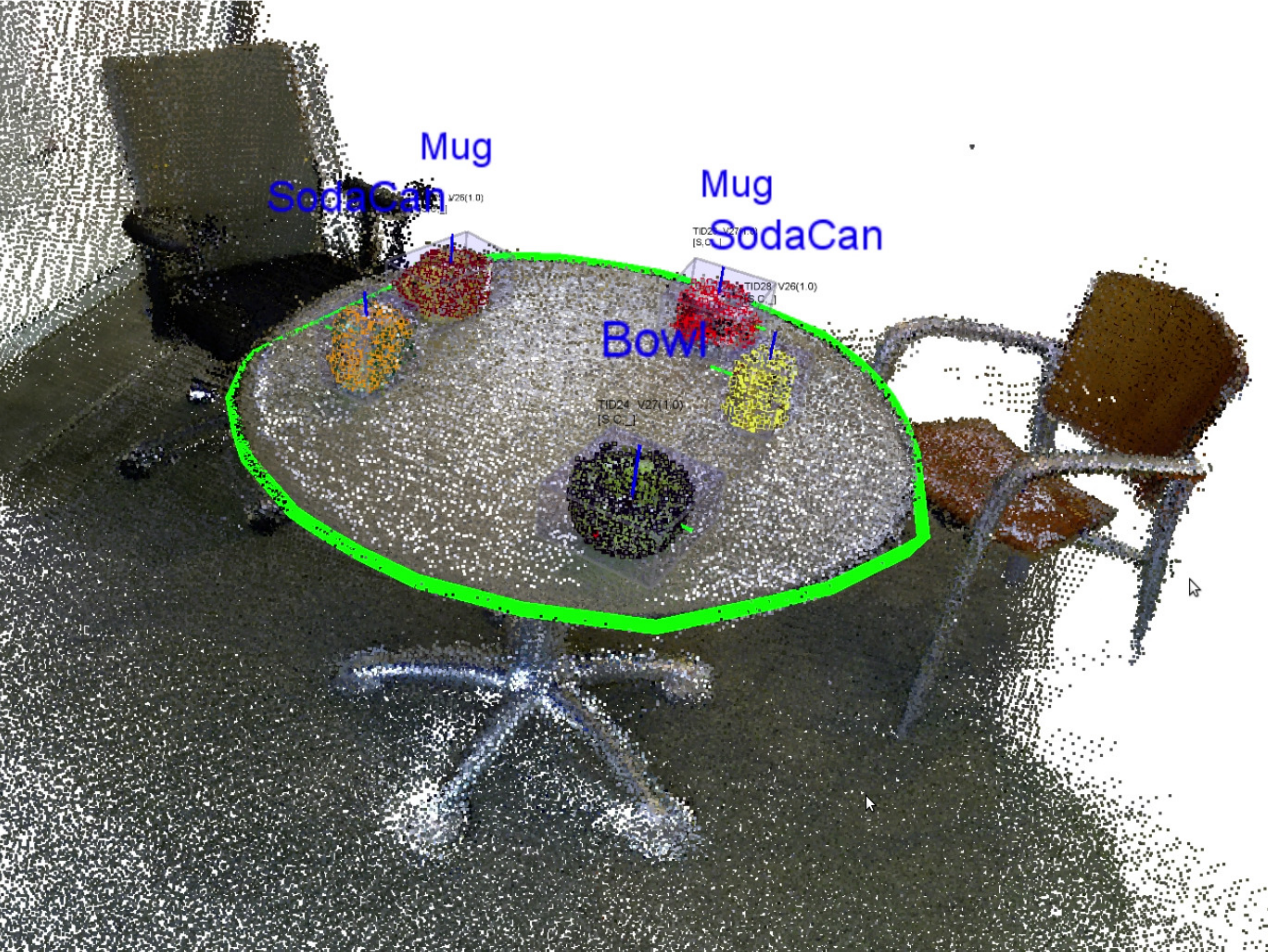}&
	\includegraphics[width=.3\linewidth, trim= 6cm 5cm 12cm 5.5cm, clip=true]{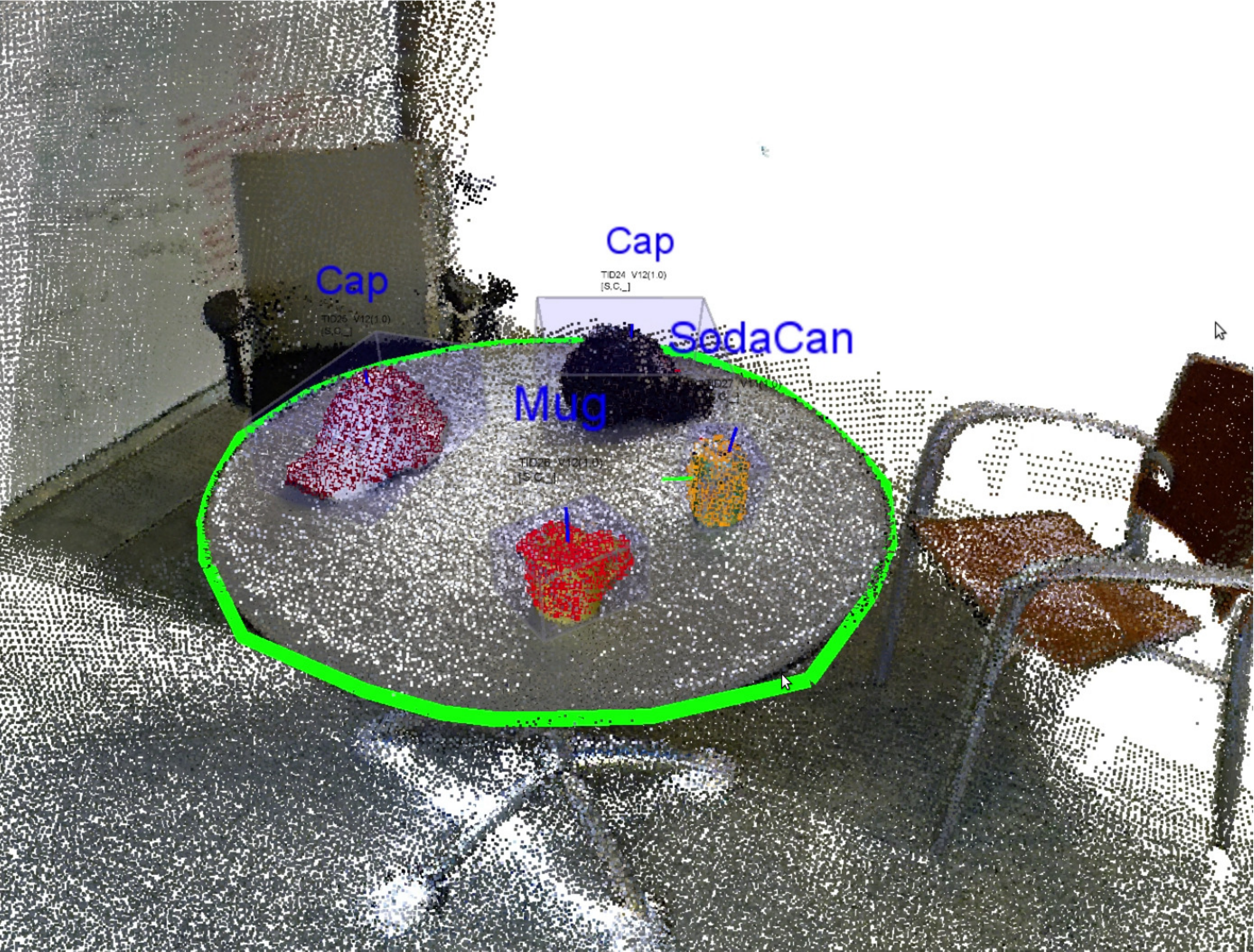}&
	\includegraphics[width=.3\linewidth, trim= 10cm 6cm 9cm 0.2cm, clip=true]{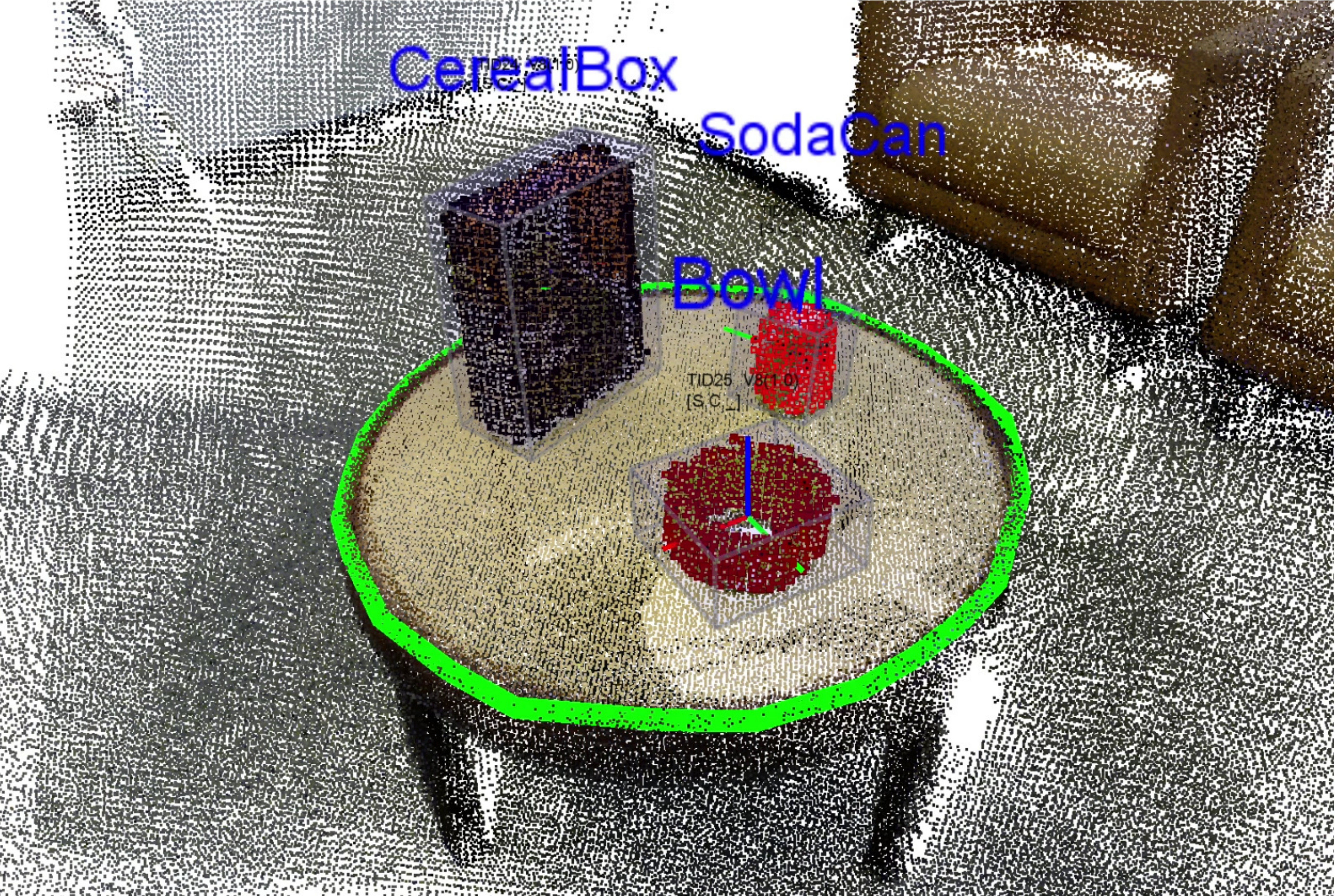}\\
	\includegraphics[width=.3\linewidth, trim= 7cm 4cm 12cm 2cm, clip=true]{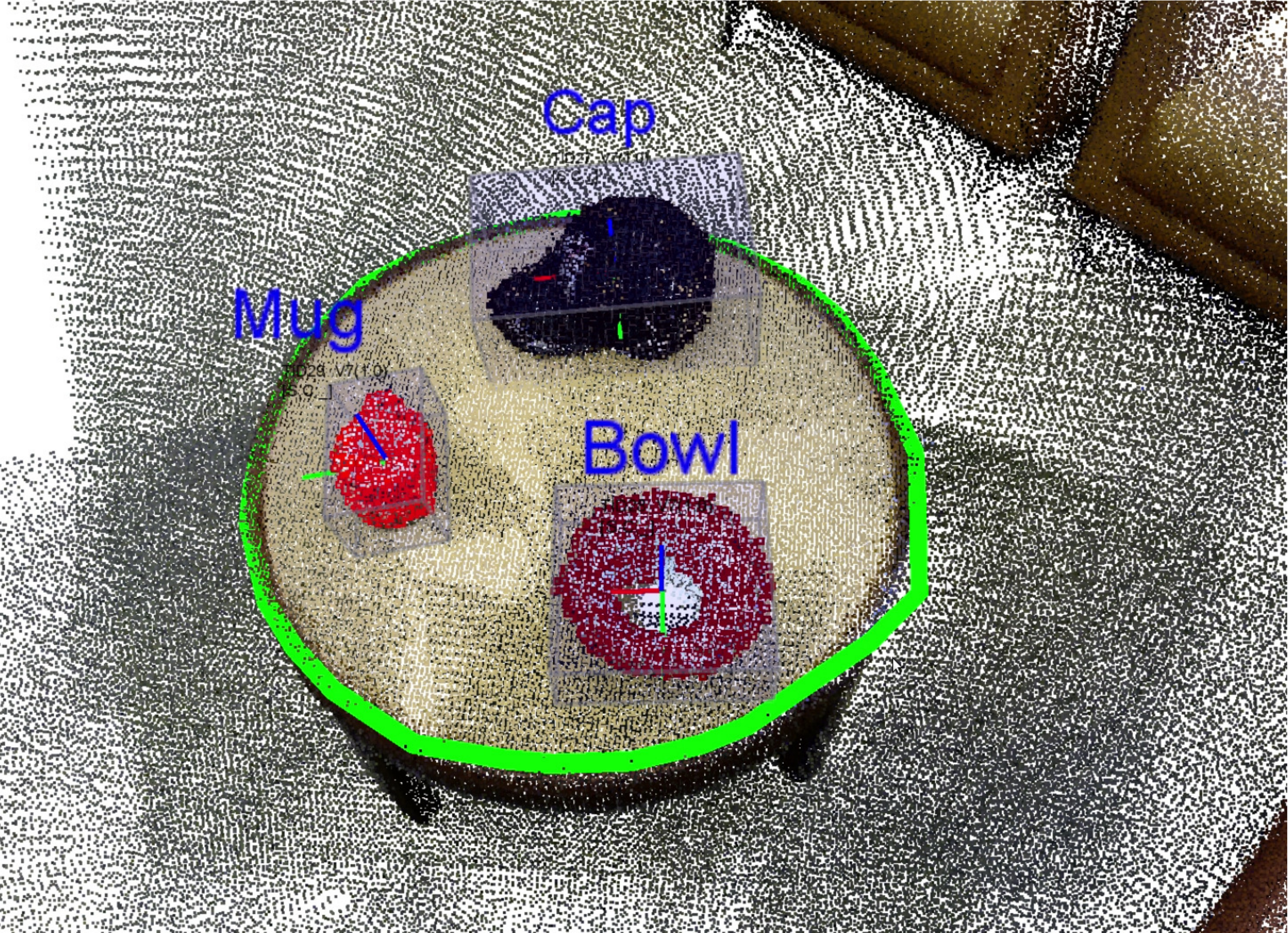}&
	\includegraphics[width=.3\linewidth, trim= 6cm 5cm 12cm 3.3cm, clip=true]{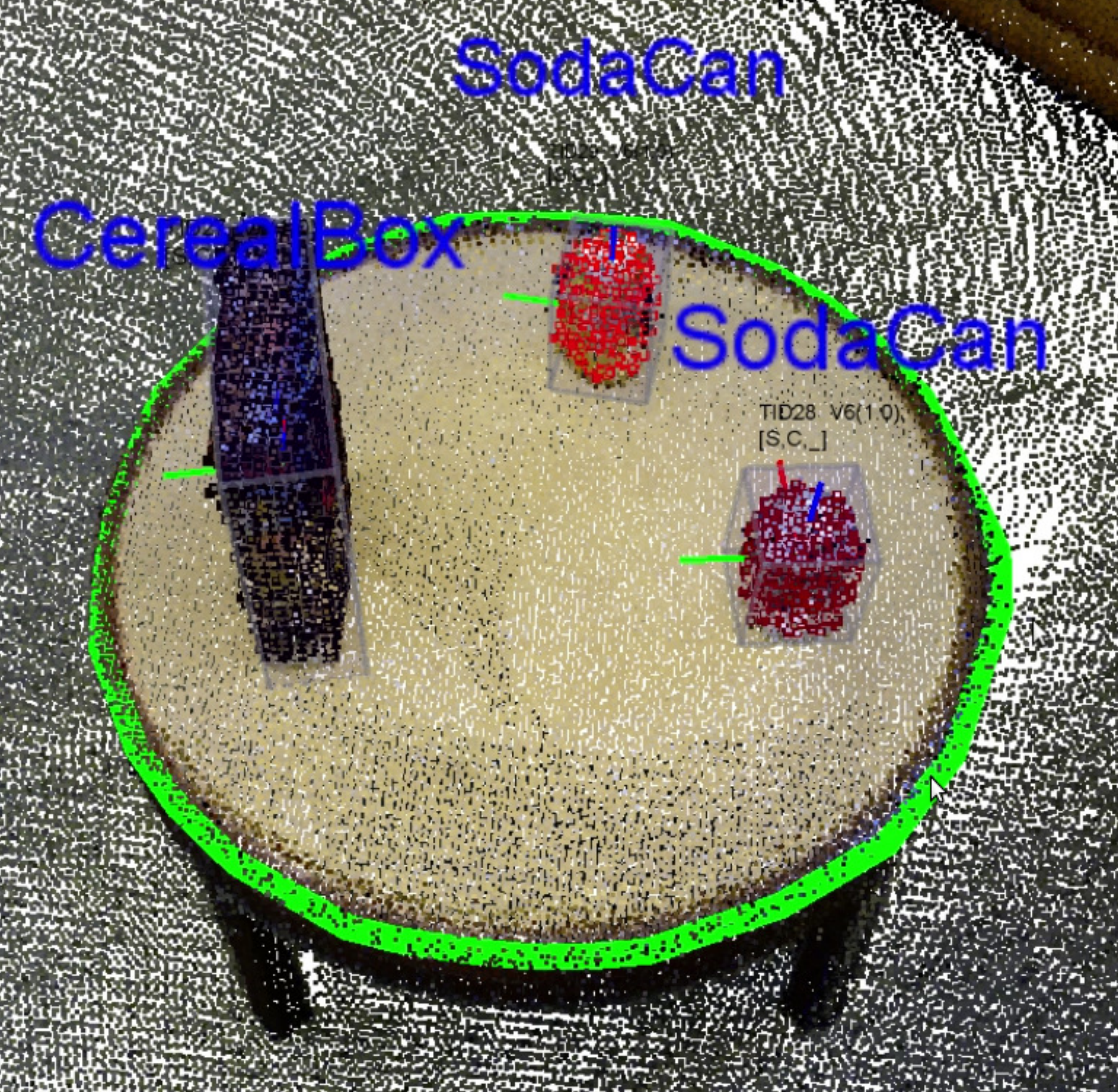}&
	\includegraphics[width=.3\linewidth, trim= 6cm 5cm 11.5cm 3.3cm, clip=true]{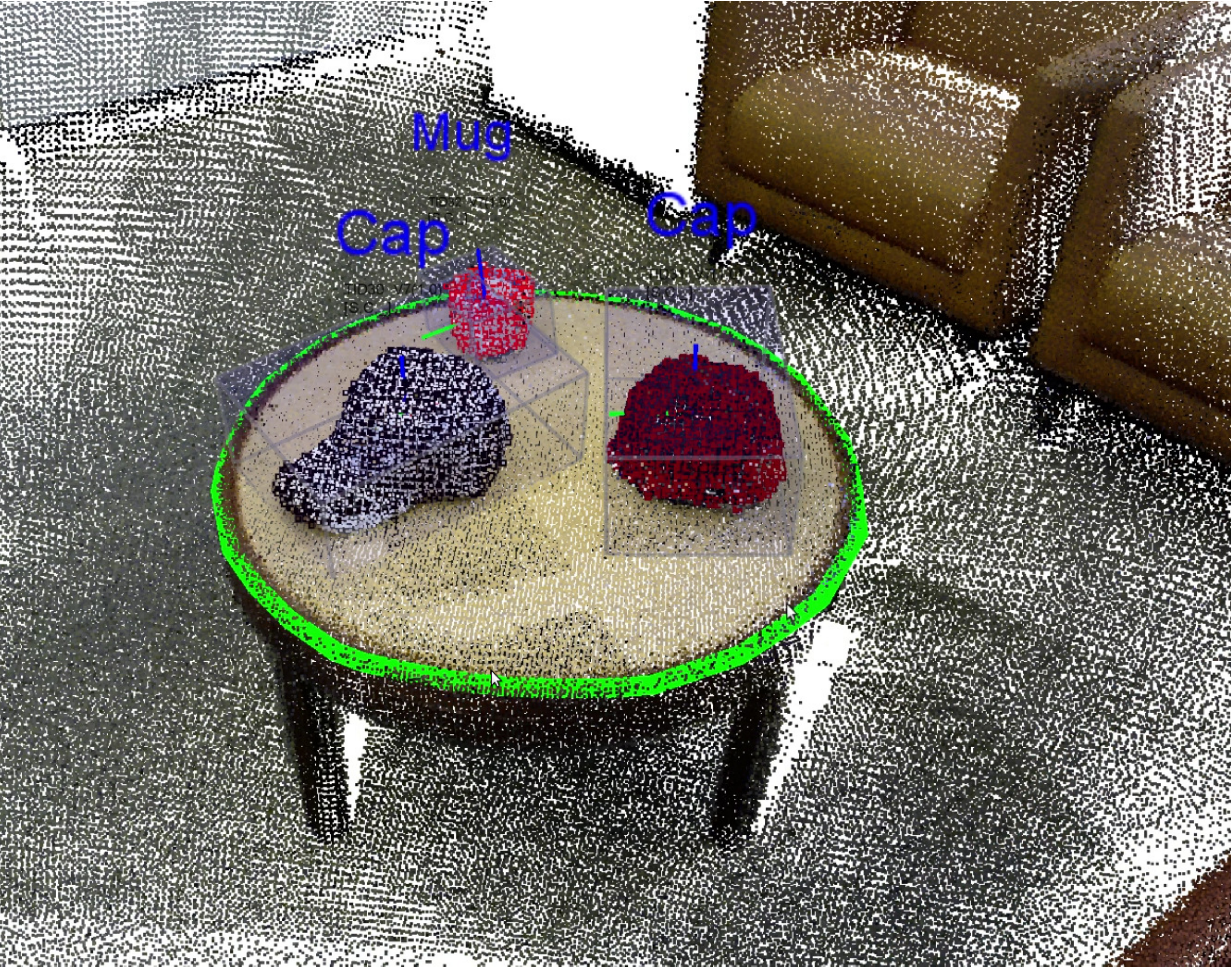}
	\\
	\includegraphics[width=.3\linewidth, trim= 7cm 13cm 12.5cm 4cm, clip=true]{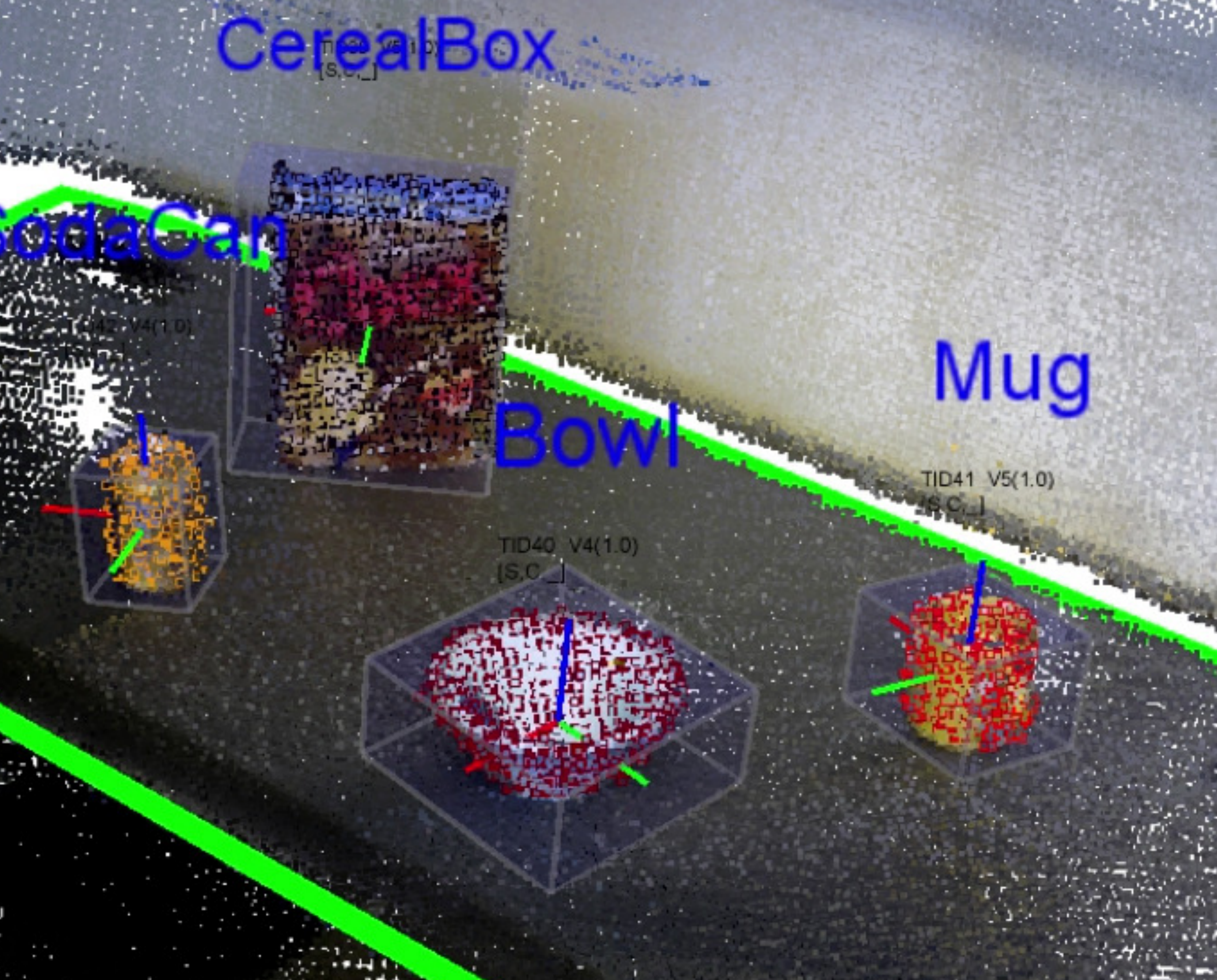}&
	\includegraphics[width=.3\linewidth, trim= 16cm 8cm 11cm 3.3cm, clip=true]{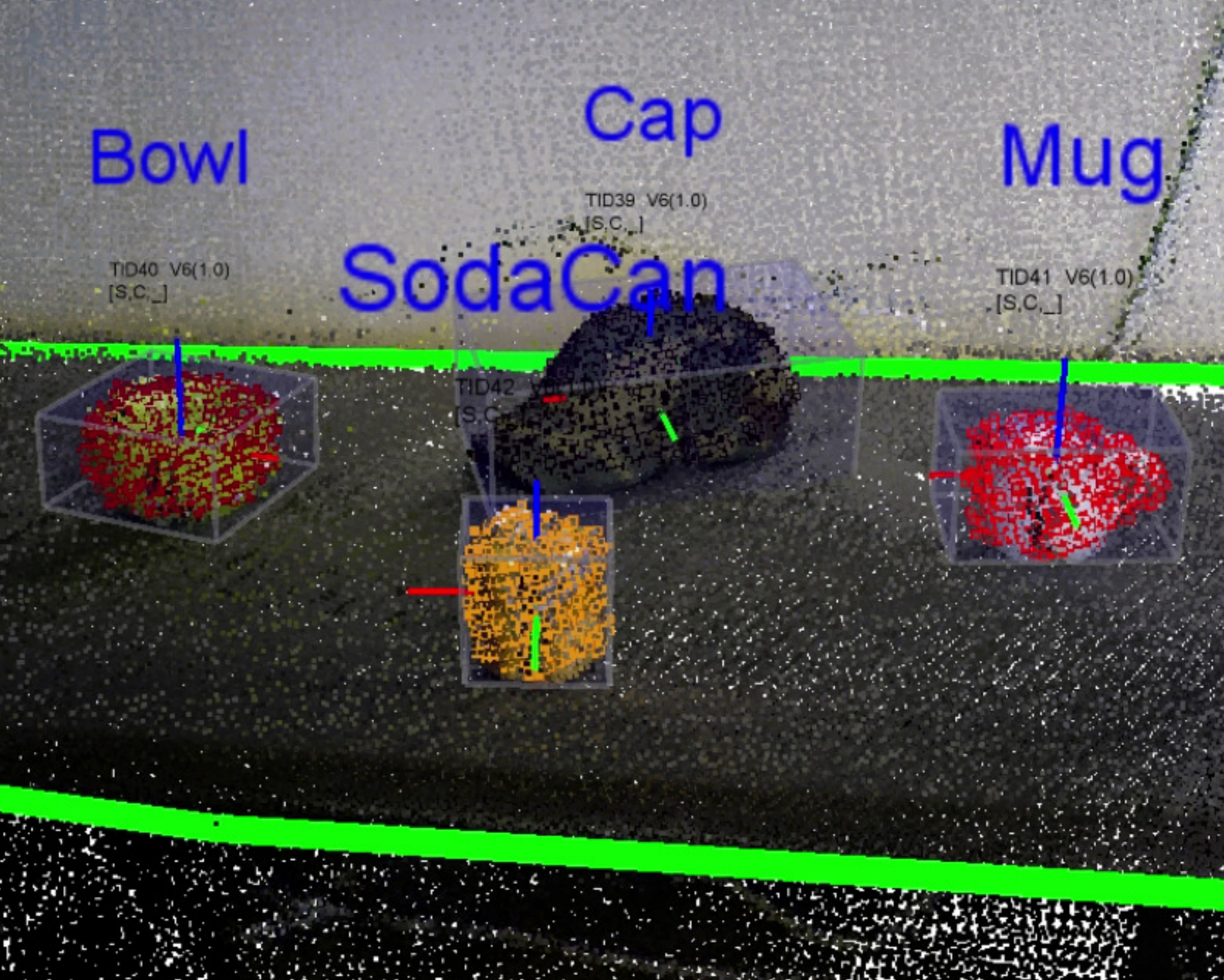}&
\end{tabular}
\caption{Object recognition results on Washington RGB-D scene dataset.}
\label{fig:Washington_scene}
\end{figure}

\subsection{Demonstration II}

Imperial College dataset is related to domestic environments, where everyday objects are placed on a kitchen table \citep{Doumanoglou2016}. It consists of variety of different scenes with a set of table top objects including \emph{amita}, \emph{colgate}, \emph{lipton}, \emph{elite} , \emph{oreo}, \emph{softkings}, \emph{mug}, \emph{shampoo} and \emph{salt-shaker}. This dataset contains $6$ different sets of table-top scenes from two different heights which has $353$ scenes in total (see Fig.~\ref{fig:IC_dataset}). Each set consists of several views of table-top scenes to cover 360 degrees view around the table. The objects were extracted from the table-top scenes by running the proposed object detection. This is an especially suitable dataset to evaluate the system since the object dataset was collected under various clutter conditions and distances. The objects were extracted from the scenes by running the object segmentation proposed in chapter~\ref{chapter_3}. All detected objects were manually labelled by the author. To examine the performance of the proposed approach, a 10-fold cross-validation has been used. The accuracy of the object recognition system was $0.92$ for the extracted objects. A set of results is visualized on Fig.~\ref{fig:ic_example}.

\begin{figure}[!t]
\centering
	\includegraphics[width=.7\linewidth, trim= 0cm 0cm 0.07cm 0cm, clip = true]{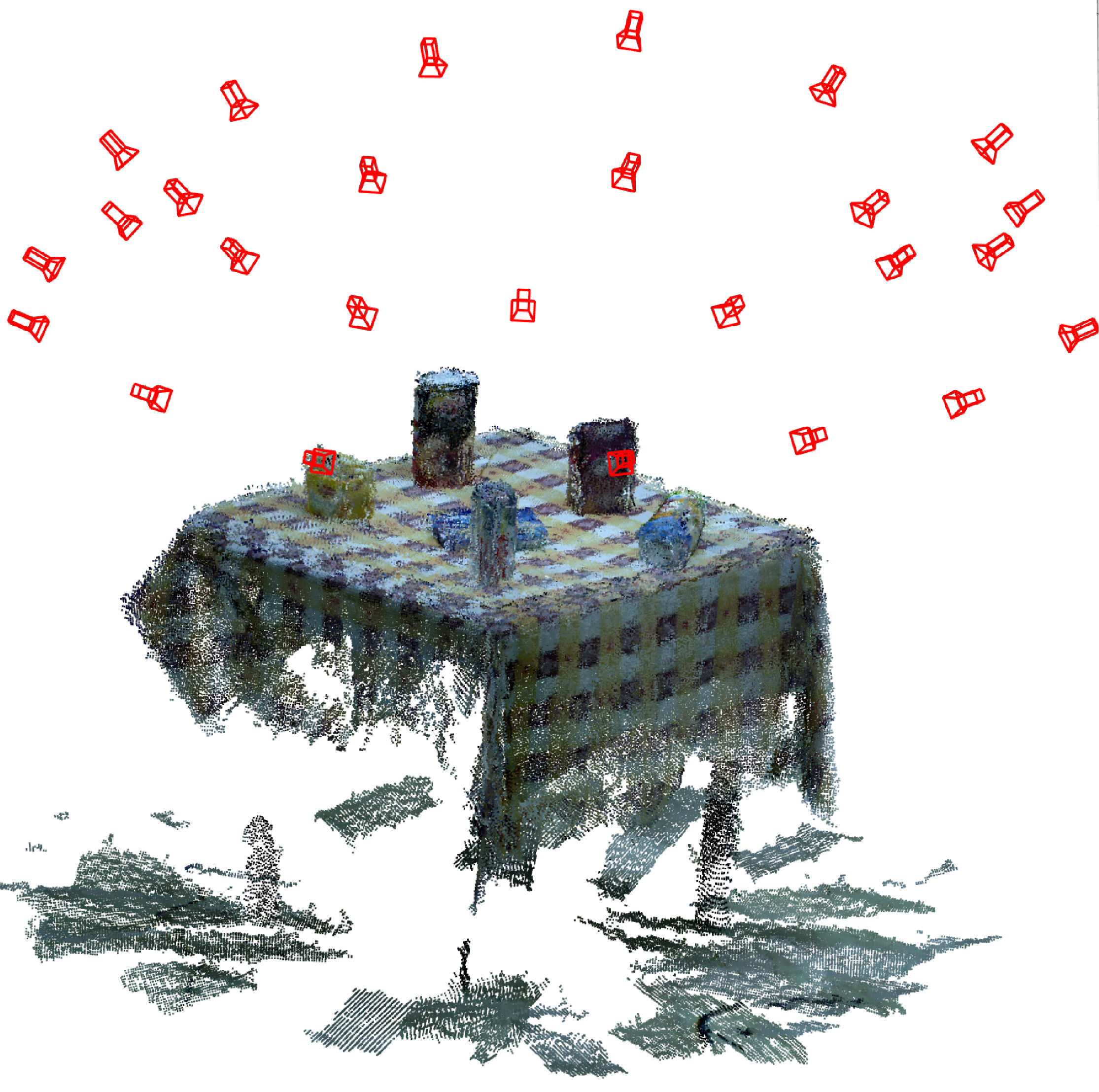}
\caption{Imperial College Dataset: accumulated view and camera pose of one of the scene sets.}
\label{fig:IC_dataset}
\end{figure}

\begin{figure}[!t]
\begin{tabular}{ccc}
	\includegraphics[width=.3\linewidth, trim= 2cm 0cm 3cm 2cm,clip=true]{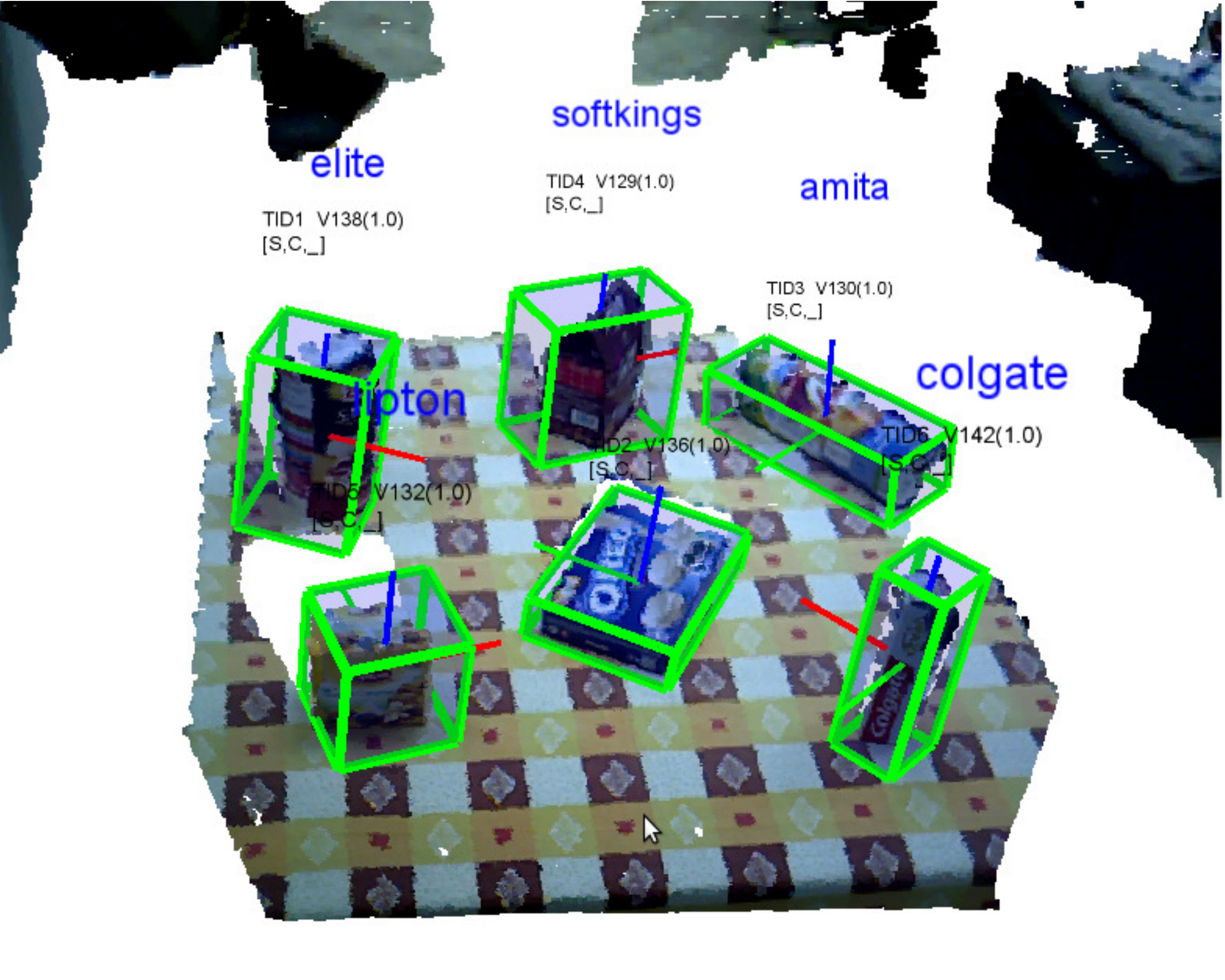}&
		\includegraphics[width=.32\linewidth, trim= 10cm 0.5cm 5cm 5cm,clip=true]{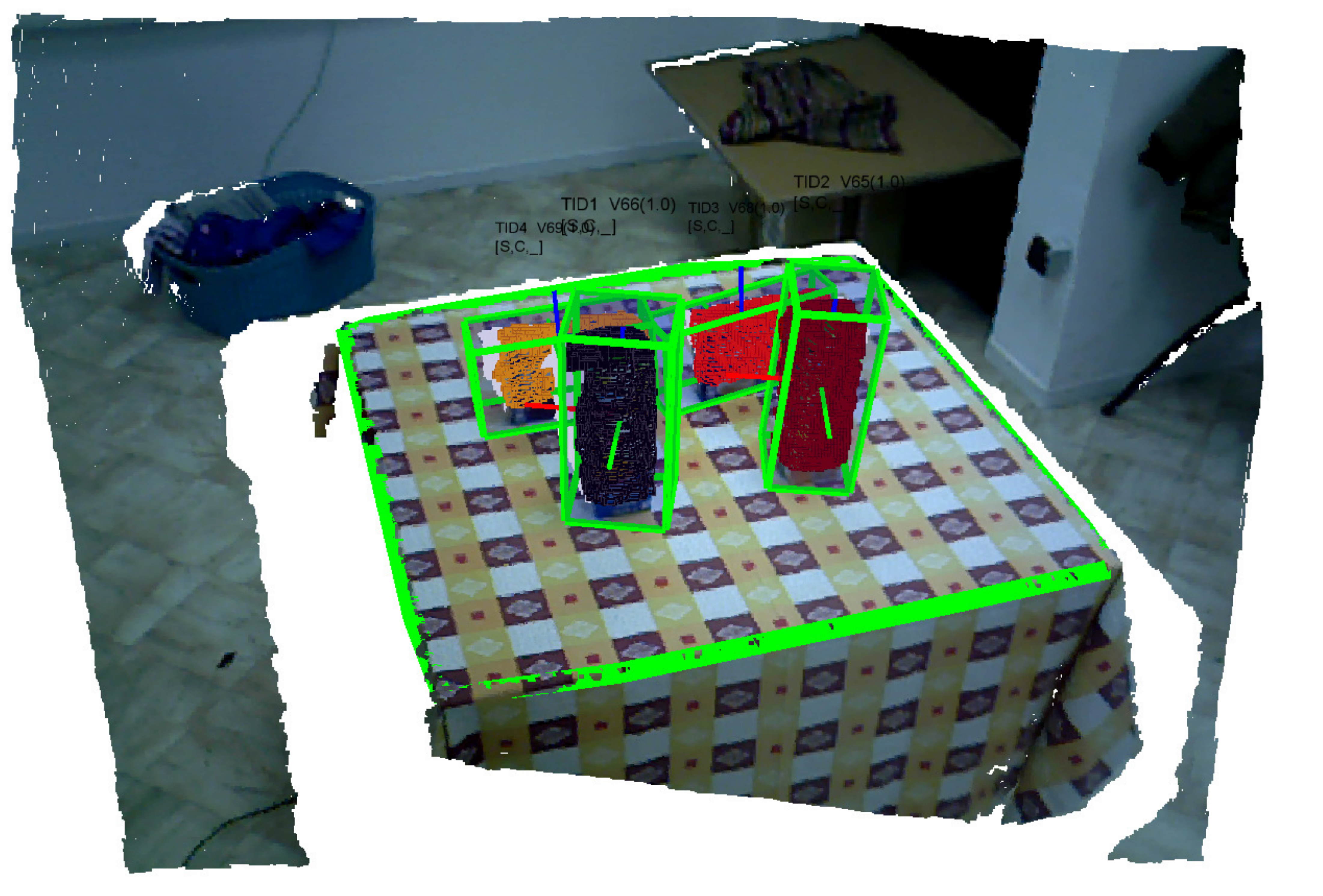}&	
		\includegraphics[width=.3\linewidth, trim= 5cm 3cm 4cm 1cm,clip=true]{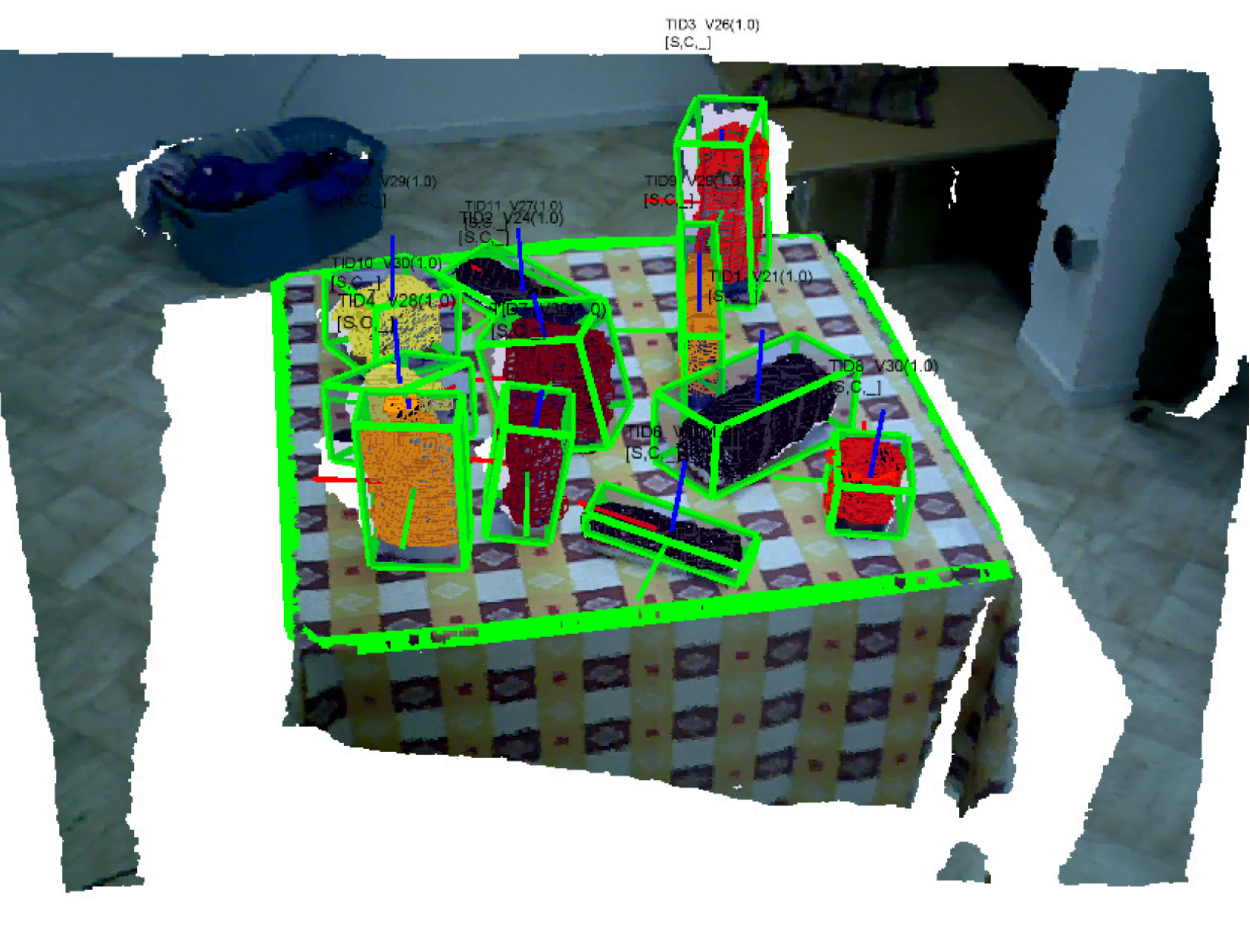}	\\
	\includegraphics[width=.3\linewidth, trim= 1cm 1cm 1cm 1cm,clip=true]{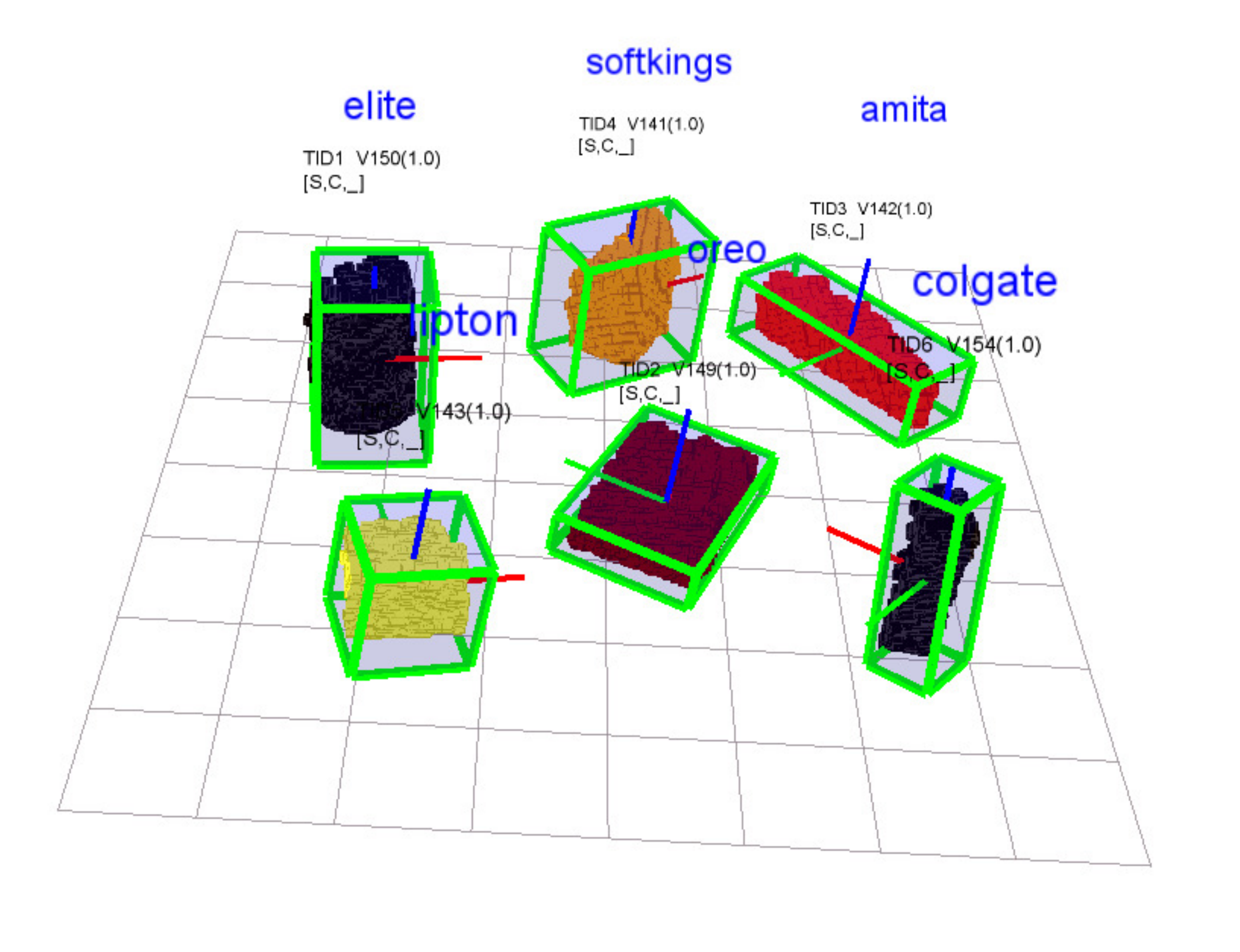} &
	\includegraphics[width=.3\linewidth, trim= 2cm 1cm 2cm 2cm,clip=true]{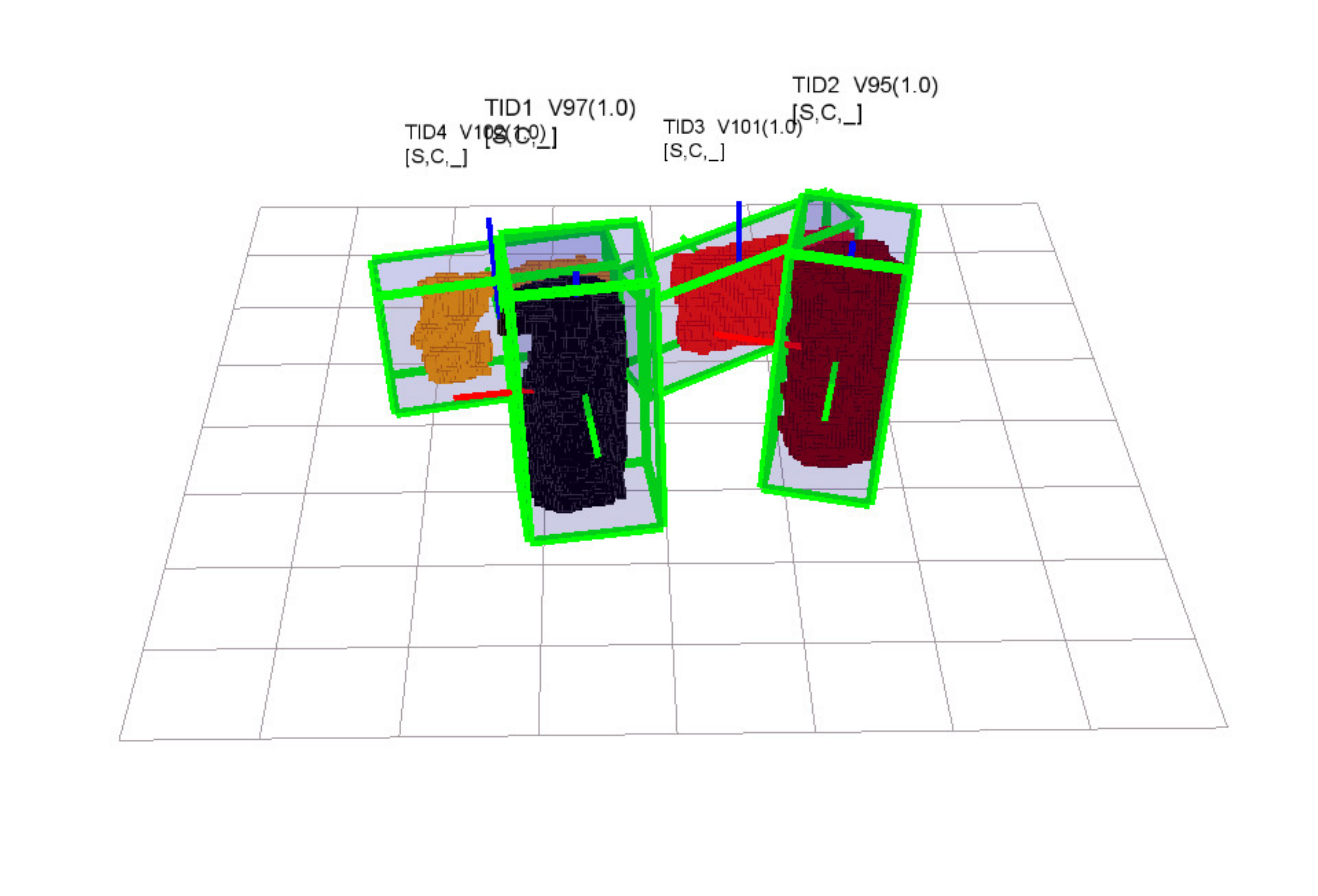}&
	\includegraphics[width=.3\linewidth, trim= 2cm 2cm 2cm 2cm,clip=true]{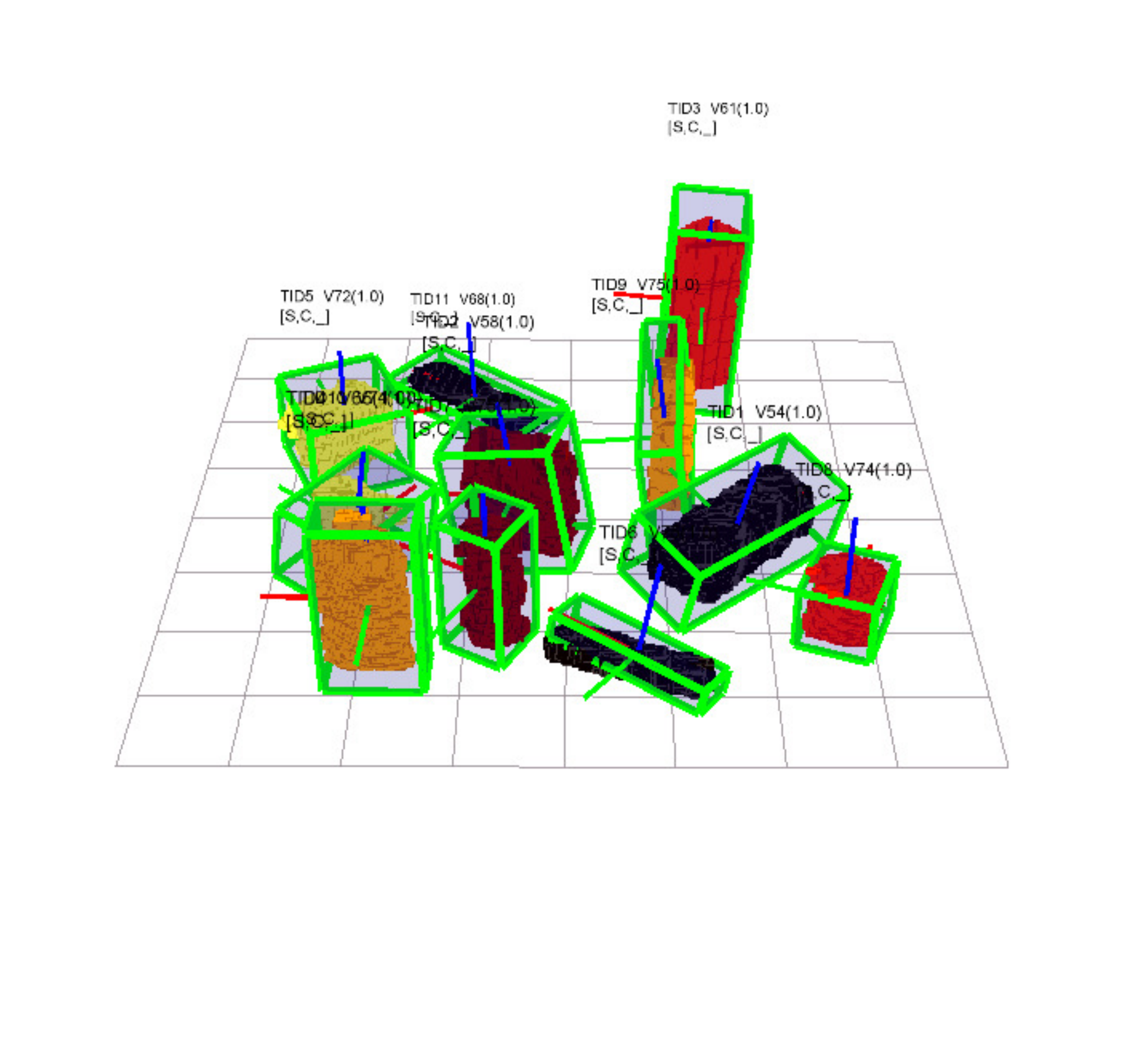}\\
	\includegraphics[width=.3\linewidth, trim= 3cm 2cm 4cm 2cm,clip=true]{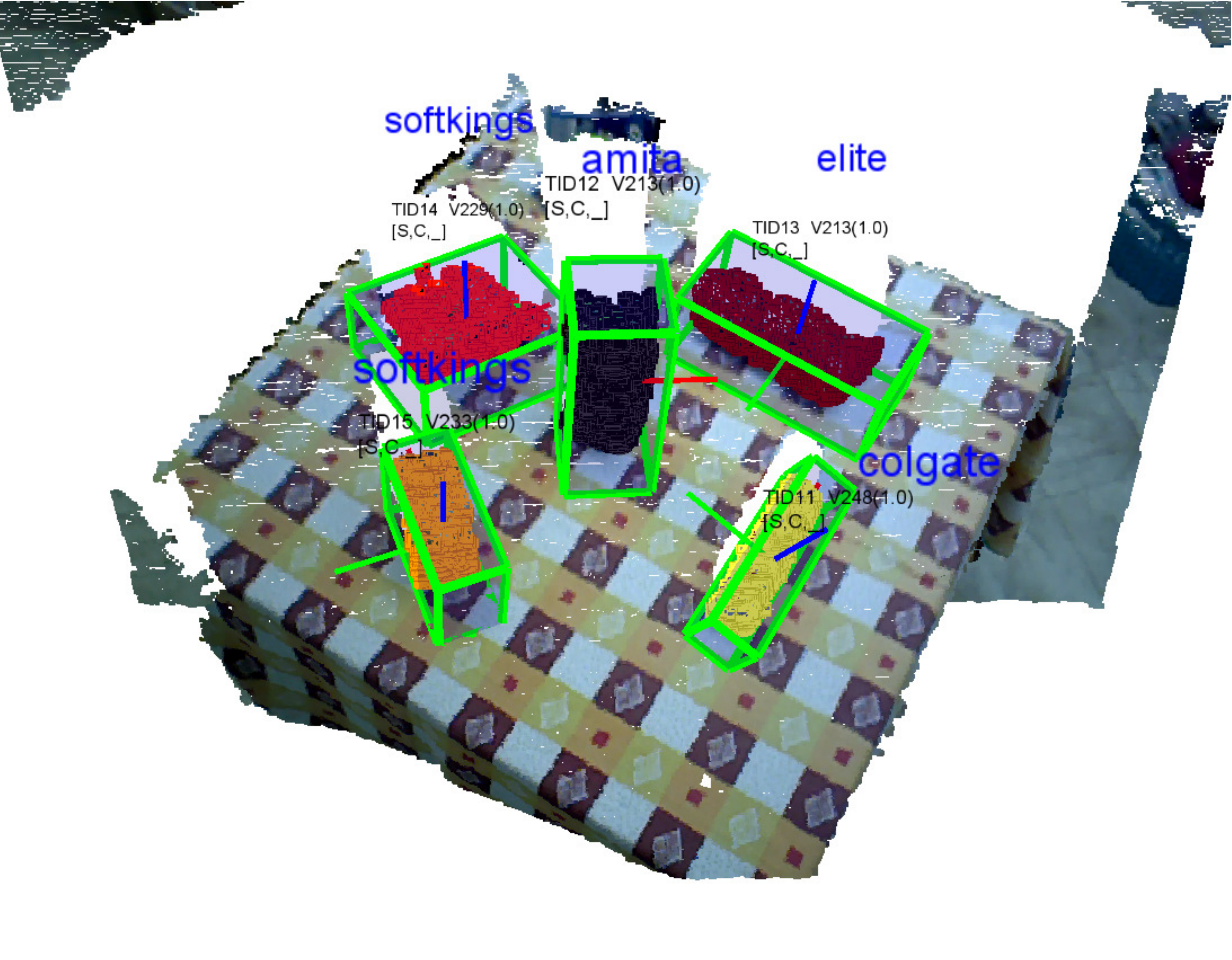}&
	\includegraphics[width=.3\linewidth, trim= 5cm 3cm 3cm 2.5cm,clip=true]{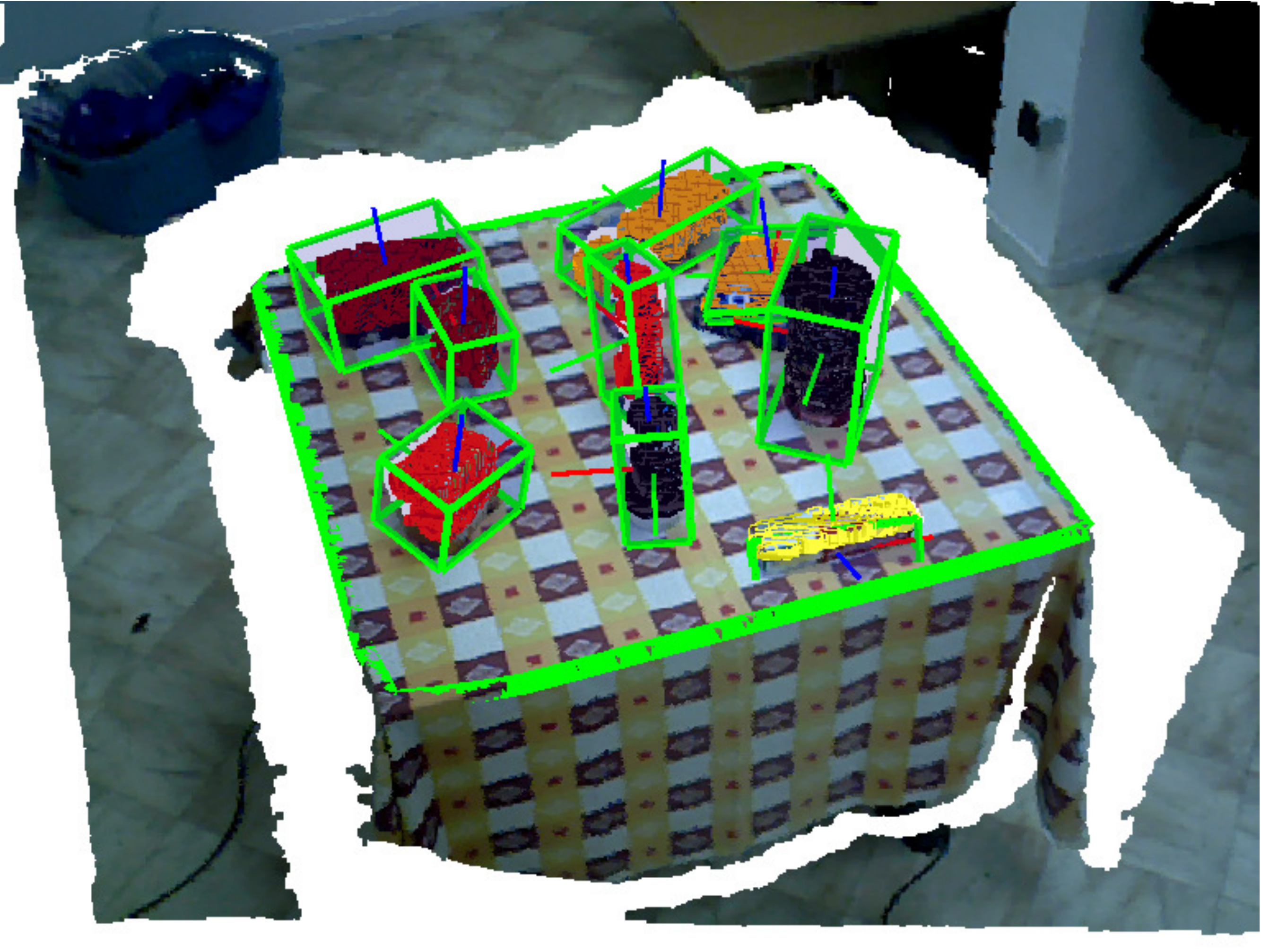}&
	\includegraphics[width=.3\linewidth, trim= 8cm 3cm 6cm 6cm,clip=true]{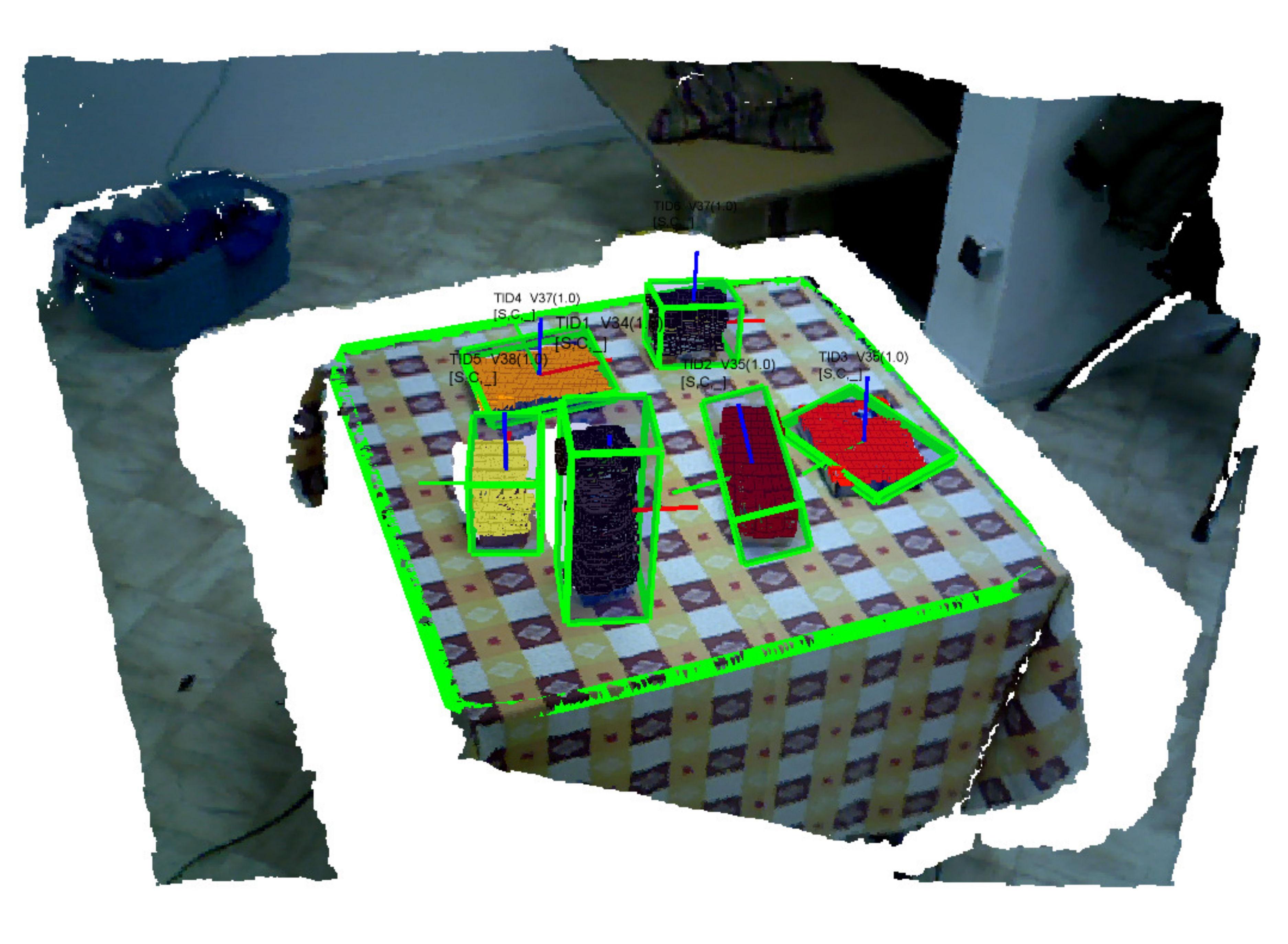}\\
	\includegraphics[width=.3\linewidth, trim= 2cm 1.5cm 3cm 2cm,clip=true]{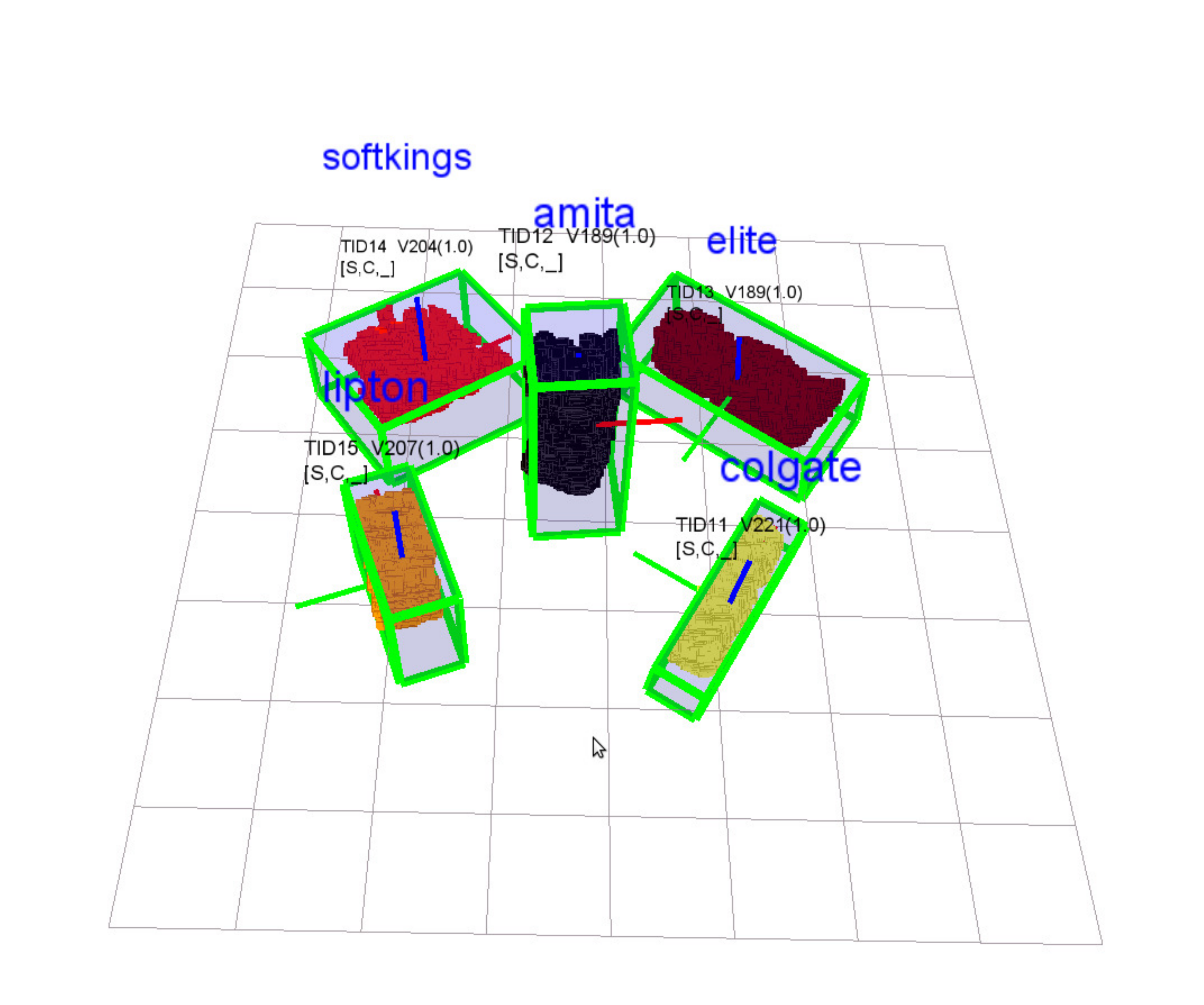}&	
	\includegraphics[width=.3\linewidth, trim= 2cm 1.5cm 3cm 2cm,clip=true]{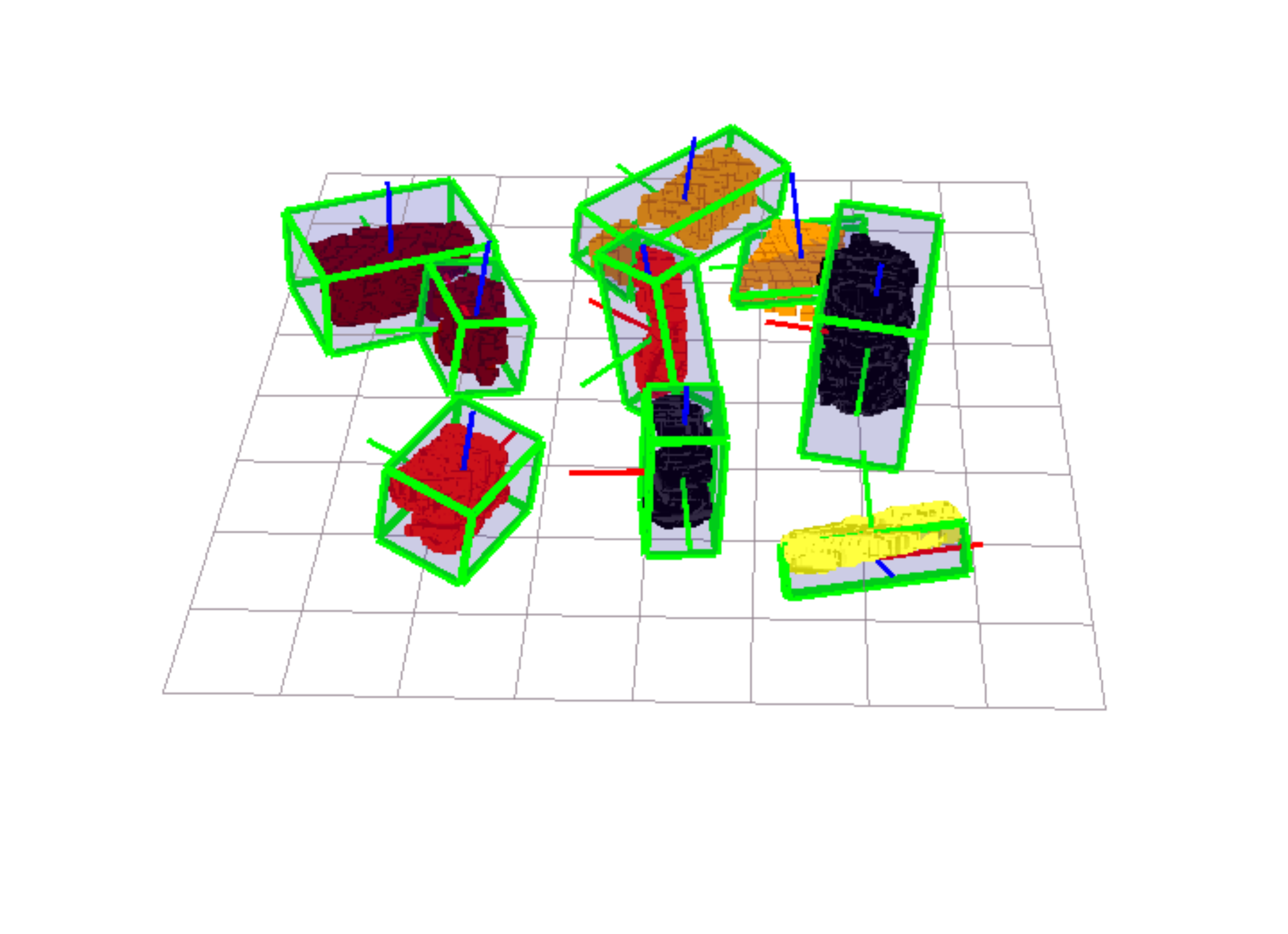}&	
	\includegraphics[width=.3\linewidth, trim= 2cm 1.5cm 3cm 2cm,clip=true]{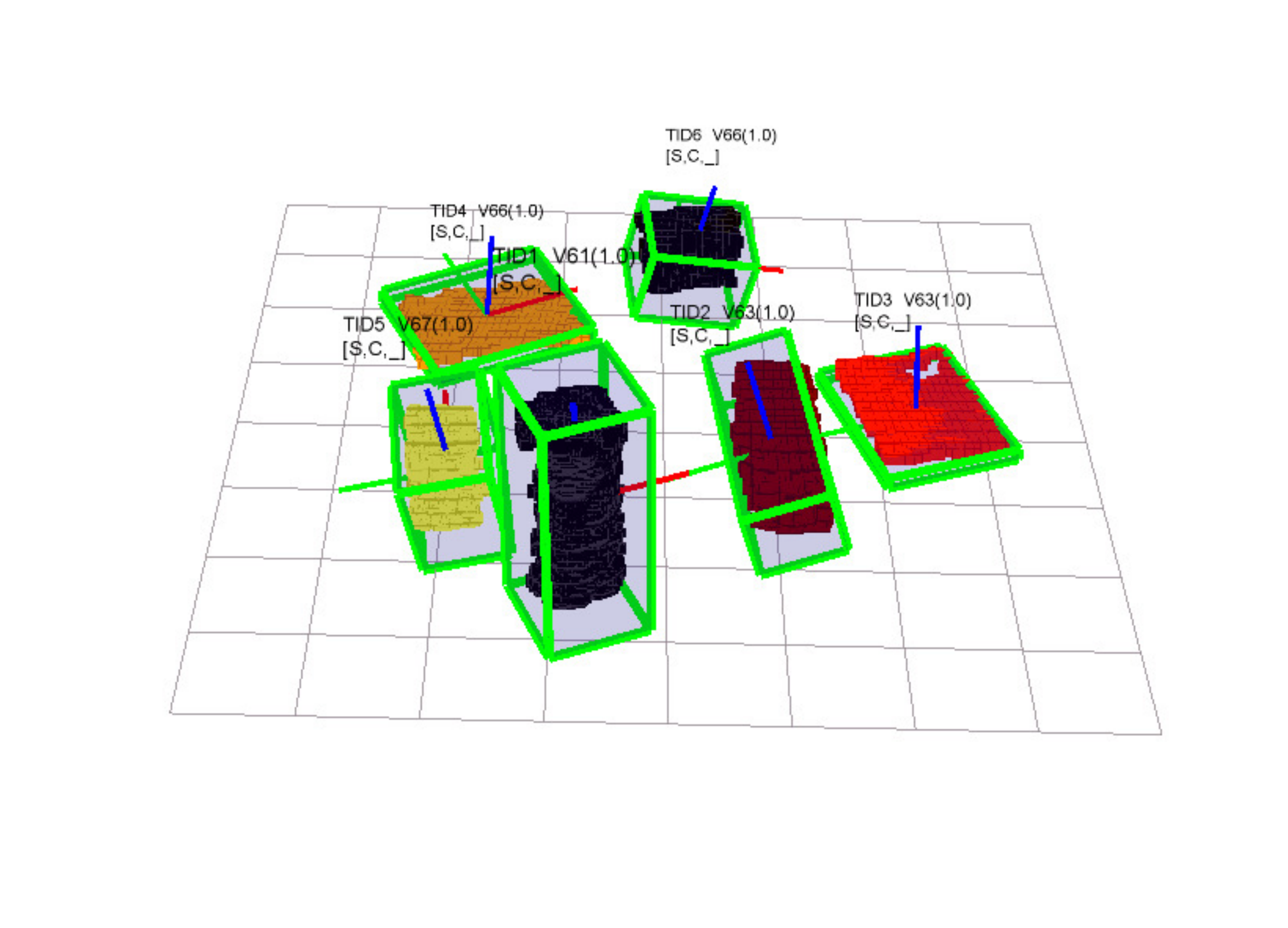}
	\\	
\end{tabular}
\caption{Qualitative results on six different table-top views from Imperial College Domestic Environment Dataset \citep{Doumanoglou2016}; (\emph{left}) Object detection and recognition on two scenes; (\emph{center} and \emph{right}) four original point clouds and corresponding segmented point clouds are shown. The red, green and blue lines represent the local reference frame of the objects.}
\label{fig:ic_example}
\end{figure}

\section{Summary}

In this chapter, a set of evaluations and demonstrations has been performed to show all the functionalities of the system for object recognition as well as its relevance for grasping. These demonstrations showed that the system can incrementally learn new object categories and perform manipulation tasks in reasonable time and appropriate manner. Our approach to object perception has been successfully tested on a JACO arm, showing the importance of having a tight coupling between perception and manipulation. We also report on two demonstrations of the system using Washington RGB-D Scenes Dataset v2 and Imperial College Dataset \citep{Doumanoglou2016}. These demonstrations showed that the system is capable of using prior knowledge to recognize new objects in the scene and learn about new object categories in an open-ended fashion.

\cleardoublepage
\chapter{Conclusions and Future Research Directions}
\label{chapter_9}
One of the primary challenges of service robotics is the adaptation of robots to new tasks in changing environments, where they interact with non-expert users. This challenge requires support from complex perception routines which can learn and recognize object categories in open-ended manner based on human-robot interaction. 
In this thesis, we assumed that versatility and competence enhancement can be obtained by learning from experiences. This thesis focused on acquiring and conceptualizing experiences about objects as a means to enhance robot competence over time thus achieving robustness. Towards this end, two perception architectures were explored that can be used by robotic agents for long-term and open-ended category learning. The problem of scaling-up to larger number of categories and adapting to new contexts were addressed using these architectures and the corresponding learning approaches. Moreover, some realistic scenarios were designed, where a human or a simulated instructor taught the robot the names of the objects present in their shared visual environment. This work has been integrated in a larger effort in the framework of the  European project RACE (Robustness by Autonomous Competence Enhancement \citep{Hertzberg2014projrep}), in which robot performance improves with accumulated experiences and conceptualizations.

\section {Contributions}
The contributions of this thesis consist in multiple theoretical formulations and practical solutions to the problem of open-ended 3D object category learning and recognition. 
They are listed as follows:

\begin{itemize}

\item The first contribution is focused on the development of a 3D object perception and perceptual learning system designed for a complex artificial cognitive agent working in a restaurant scenario. This system, developed in collaboration with other researchers in IEETA and within the scope of the European project RACE, integrates detection, tracking, learning and recognition of tabletop objects. Interaction capabilities were also developed to enable a human user to take the role of instructor and teach new object categories. Thus, the system learns in an incremental and open-ended way from user-mediated experiences. Thorough details of the agent's complete architecture have been presented in chapter~\ref{chapter_2}.

\item The second contribution is related to the gathering object experiences in both supervised and unsupervised manner. In general, gathering object experiences is a challenging task because of the dynamic nature of the world and ill-definition of the objects. In this work, a system of boolean equations was used for encoding the world and object candidates. In particular, we proposed automatic perception capabilities that will allow robots to automatically detect multiple objects in a crowded scene.
The relevant aspects have been described in chapter~\ref{chapter_3}.

\item Other contributions, presented in chapter~\ref{chapter_4}, are concerned
with object representation. Object representation is one of the most challenging tasks in robotics because it must provide reliable information in real-time to enable the robot to physically interact with the objects in its environment. We have tackled the problem of object representation, by proposing a novel global object descriptor named Global Orthographic Object Descriptor (GOOD). It has been designed to be robust, descriptive and efficient to compute and use. The overall classification performance obtained with GOOD was comparable to the best performances obtained with the state-of-the-art descriptors. Concerning memory and computation time, GOOD clearly outperformed the other descriptors. Therefore, GOOD is especially suited for real-time applications. The estimated object's pose is precise enough for real-time object manipulation tasks. %The proposed global object descriptor has been presented and discussed in chapter~\ref{chapter_4}.

We also proposed an extension of Latent Dirichlet Allocation to learn structural semantic features (i.e., topics) from low-level feature co-occurrences for each object category independently. Although the model we developed has been intended to be used for object category learning and recognition, it is a novel probabilistic model that can be used in the fields of computer vision and machine learning. %This contribution has been presented and discussed in chapter~\ref{chapter_4}.

\item The problem of open-ended learning for 3D object category recognition has been tackled in chapter~\ref{chapter_5}. We approached object category learning and recognition from a long-term perspective and with emphasis on open-endedness, i.e. not assuming a pre-defined set of categories. The major contributions are the following: (\emph{i}) defining new distance functions for estimating dissimilarity between sets of local shape features that can be used in instance-based learning approaches; (\emph{ii}) proposing a learning approach to incrementally learn probabilistic models of object categories to achieve adaptability.  

\item The last contribution of this dissertation is concerned with the evaluation of open-ended object category learning and recognition approaches in multi-context scenarios.  Off-line evaluation approaches such as cross-validation do not comply with the simultaneous nature of learning and recognition autonomous agents. An adaptability measure and a teaching protocol, supporting context change, were therefore designed and used for open-ended experimental evaluation. This contribution has been presented and discussed in chapter~\ref{chapter_7}.

\end{itemize}

%^^^^^^^^^^^^^^^^^^^^^^^^^^^^^^^^^^^^^^^^^^^^^^^^^^^^^^^^^^^^^^^^
All the algorithms and concepts presented in this thesis have been implemented and tested on data acquired in realistic restaurant environments. A set of experimental results was also carried out on two different robotic platforms, including the PR2 and the manipulation platform at the university of Aveiro (see Figure \ref{fig:platforms}). 

\section {Future Research Directions}
Despite the promising results presented in this thesis, there are a list of open issues that still remain to be tackled for future research:

\begin{itemize}
\item Though we proved the usefulness of 3D geometry in the context of learning and recognizing object categories, this thesis has not addressed the fusion of color information with geometry at all. Color and texture are two important features for particular applications which geometry alone cannot solve. For example, 
color and texture information can also be used to distinguish objects that have the same geometric properties with different texture (e.x. a Coke can from a Diet Coke can).

\item In this thesis, we already tackled the problem of environment exploration and visual word dictionary construction but mostly in terms of building the dictionary in advance by feeding a sample set of features of extracted objects to a clustering algorithm e.g., K-means. Since the proposed architectures receive a continuous stream of 3D data, we would like to consider data stream clustering methods to update the visual word dictionary as another direction of future work. We are already taking steps towards addressing this point \citep{Oliveira2015}.

\item Although the work in this thesis mainly focused on 3D object category learning, it would be interesting and relevant to extend the proposed learning architecture to other domains, including grasp learning for object manipulation. We are already taking steps towards addressing this point and some interesting results have already been published \citep{kasaei2016object,shafii2016learning}.

\item Currently, a popular approach in computer vision is deep learning. However, there are several limitations to use Deep Neural Networks (DNN) in open-ended domains. Deep networks are incremental by nature but not open-ended, since the inclusion of novel categories enforces a restructuring in the topology of the network. Overcoming such limitations is also one of the possible directions of continuation of the work in thesis. The relevant keywords for this research topic are \emph{Zero-shot learning}, \emph{Low-shot learning} and \emph{Overcoming catastrophic forgetting in neural networks}.

\item In the continuation of this work, we will also investigate the possibility of improving memory management by considering salience and forgetting mechanisms. 

\item I believe this framework has great potential for further developments. During the development of the proposed architecture, I attended several conferences to discuss my research with many experts. The feedbacks I received broadened my view and encouraged me to further develop the system. The functionalities of the developed system make it unique in the robotic community. Currently, it provides several functionalities that will allow robots to: (\emph{\textbf{i}}) detect objects in highly crowded scenes, (\emph{\textbf{ii}}) incrementally learn object categories from the set of accumulated experiences, (\emph{\textbf{iii}}) construct the full model of an unknown object in an on-line manner, (\emph{\textbf{iv}}) infer how to grasp objects in different situations, (\emph{\textbf{v}}) predict the next-best-view to improve object detection and manipulation.

\end{itemize}
%^^^^^^^^^^^^^^^^^^^^^^^^^^^^^^^^^^^^^^^^^^^^^^^^^^^^^^^^^^^^^^^^

\cleardoublepage
%\input{}
%\cleardoublepage
%%%%%%%%%%%%%%%%%%%%%%%%%%%%%%%%%%%%%%%%%%%%%%%%%%%%%%%%%%%%%%%%%%%%%%%%%%%%%%%%

\value{chapter}=0
\bibliographystyle{apalike}
\bibliography{clean_thesis_refs}

\begin{thebibliography}{}

\bibitem[Akata et~al., 2014]{akata2014good}
Akata, Z., Perronnin, F., Harchaoui, Z., and Schmid, C. (2014).
\newblock Good practice in large-scale learning for image classification.
\newblock {\em IEEE Transactions on Pattern Analysis and Machine Intelligence},
  36(3).

\bibitem[Aldoma et~al., 2012a]{aldoma2012point}
Aldoma, A., Marton, Z.-C., Tombari, F., Wohlkinger, W., Potthast, C., Zeisl,
  B., Rusu, R.~B., Gedikli, S., and Vincze, M. (2012a).
\newblock Point cloud library.
\newblock {\em IEEE Robotics \& Automation Magazine}, 1070(9932/12).

\bibitem[Aldoma et~al., 2012b]{aldoma2012}
Aldoma, A., Marton, Z.-C., Tombari, F., Wohlkinger, W., Potthast, C., Zeisl,
  B., Rusu, R.~B., Gedikli, S., and Vincze, M. (2012b).
\newblock Point cloud library: Three-dimensional object recognition and 6 dof
  pose estimation.
\newblock {\em IEEE Robotics and Automation Magazine}, 19(3):80--91.

\bibitem[Anderson et~al., 1997]{anderson1997act}
Anderson, J.~R., Matessa, M., and Lebiere, C. (1997).
\newblock Act-r: A theory of higher level cognition and its relation to visual
  attention.
\newblock {\em Human-Computer Interaction}, 12(4):439--462.

\bibitem[Ando et~al., 2013]{AndoTocic}
Ando, Y., Nakamura, T., Araki, T., and Nagai, T. (2013).
\newblock Formation of hierarchical object concept using hierarchical latent
  dirichlet allocation.
\newblock In {\em 2013 IEEE/RSJ International Conference on Intelligent Robots
  and Systems}, pages 2272--2279.

\bibitem[Andreopoulos and Tsotsos, 2013]{andreopoulos201350}
Andreopoulos, A. and Tsotsos, J.~K. (2013).
\newblock 50 years of object recognition: Directions forward.
\newblock {\em Computer Vision and Image Understanding}, 117(8):827--891.

\bibitem[Banerjee and Basu, 2007]{banerjee2007topic}
Banerjee, A. and Basu, S. (2007).
\newblock Topic models over text streams: A study of batch and online
  unsupervised learning.
\newblock In {\em SDM}, volume~7, pages 437--442. SIAM.

\bibitem[Bariya et~al., 2012]{bariya20123d}
Bariya, P., Novatnack, J., Schwartz, G., and Nishino, K. (2012).
\newblock 3d geometric scale variability in range images: Features and
  descriptors.
\newblock {\em International journal of computer vision}, 99(2):232--255.

\bibitem[Beetz et~al., 2011]{beetz2011robotic}
Beetz, M., Klank, U., Kresse, I., Maldonado, A., Mosenlechner, L., Pangercic,
  D., Ruhr, T., and Tenorth, M. (2011).
\newblock Robotic roommates making pancakes.
\newblock In {\em Humanoid Robots (Humanoids), 2011 11th IEEE-RAS International
  Conference on}, pages 529--536. IEEE.

\bibitem[Beis and Lowe, 1997]{Beis}
Beis, J.~S. and Lowe, D.~G. (1997).
\newblock Shape indexing using approximate nearest-neighbour search in
  high-dimensional spaces.
\newblock In {\em Proceedings of the Conference on Computer Vision and Pattern
  Recognition (CVPR 1997)}, Washington, DC, USA. IEEE Computer Society.

\bibitem[Biasotti et~al., 2013]{biasotti2013sketch}
Biasotti, S., Pratikakis, I., Castellani, U., Schreck, T., Godil, A., and
  Veltkamp, R. (2013).
\newblock Sketch-based 3d model retrieval by viewpoint entropy-based adaptive
  view clustering.

\bibitem[Bifet et~al., 2010]{bifet2010moa}
Bifet, A., Holmes, G., Kirkby, R., and Pfahringer, B. (2010).
\newblock Moa: Massive online analysis.
\newblock {\em Journal of Machine Learning Research}, 11(May):1601--1604.

\bibitem[Blei, 2012]{blei2012probabilistic}
Blei, D.~M. (2012).
\newblock Probabilistic topic models.
\newblock {\em Communications of the ACM}, 55(4):77--84.

\bibitem[Blei et~al., 2003]{blei2003latent}
Blei, D.~M., Ng, A.~Y., and Jordan, M.~I. (2003).
\newblock Latent dirichlet allocation.
\newblock {\em the Journal of machine Learning research}, 3:993--1022.

\bibitem[Bo et~al., 2011]{bo2011object}
Bo, L., Lai, K., Ren, X., and Fox, D. (2011).
\newblock Object recognition with hierarchical kernel descriptors.
\newblock In {\em Computer Vision and Pattern Recognition (CVPR), 2011 IEEE
  Conference on}, pages 1729--1736. IEEE.

\bibitem[Bro et~al., 2008]{bro2008resolving}
Bro, R., Acar, E., and Kolda, T.~G. (2008).
\newblock Resolving the sign ambiguity in the singular value decomposition.
\newblock {\em Journal of Chemometrics}, 22(2):135--140.

\bibitem[Canini et~al., 2009]{canini2009online}
Canini, K.~R., Shi, L., and Griffiths, T.~L. (2009).
\newblock Online inference of topics with latent dirichlet allocation.
\newblock In {\em International conference on artificial intelligence and
  statistics}, pages 65--72.

\bibitem[Celikkanat et~al., 2016]{celikkanat2016learning}
Celikkanat, H., Orhan, G., Pugeault, N., Guerin, F., {\c{S}}ahin, E., and
  Kalkan, S. (2016).
\newblock Learning context on a humanoid robot using incremental latent
  dirichlet allocation.
\newblock {\em IEEE Transactions on Cognitive and Developmental Systems},
  8(1):42--59.

\bibitem[Cha, 2007]{cha2007comprehensive}
Cha, S.-H. (2007).
\newblock Comprehensive survey on distance/similarity measures between
  probability density functions.
\newblock {\em City}, 1(2):1.

\bibitem[Chauhan, 2014]{chauhan2014grounding}
Chauhan, A. (2014).
\newblock Grounding human vocabulary in robot perception through interaction.

\bibitem[Chauhan et~al., 2013]{chauhan2013towards}
Chauhan, A., Lopes, L.~S., Tom{\'e}, A.~M., and Pinho, A. (2013).
\newblock Towards supervised acquisition of robot activity experiences: an
  ontology-based approach.
\newblock In {\em 16th Portuguese Conference on Artificial Intelligence-EPIA},
  volume 2013.

\bibitem[Chauhan and Seabra~Lopes, 2011]{chauhan2011}
Chauhan, A. and Seabra~Lopes, L. (2011).
\newblock Using spoken words to guide open-ended category formation.
\newblock {\em Cognitive Processing}, 12:341--354.

\bibitem[Chauhan and Seabra~Lopes, 2015]{chauhan2015experimental}
Chauhan, A. and Seabra~Lopes, L. (2015).
\newblock An experimental protocol for the evaluation of open-ended category
  learning algorithms.
\newblock In {\em Evolving and Adaptive Intelligent Systems (EAIS), 2015 IEEE
  International Conference on}, pages 1--8. IEEE.

\bibitem[Chen and Bhanu, 2007]{chen20073d}
Chen, H. and Bhanu, B. (2007).
\newblock 3d free-form object recognition in range images using local surface
  patches.
\newblock {\em Pattern Recognition Letters}, 28(10):1252--1262.

\bibitem[Ciocarlie et~al., 2014]{ciocarlie2014towards}
Ciocarlie, M., Hsiao, K., Jones, E.~G., Chitta, S., Rusu, R.~B., and
  {\c{S}}ucan, I.~A. (2014).
\newblock Towards reliable grasping and manipulation in household environments.
\newblock In {\em Experimental Robotics}, pages 241--252. Springer.

\bibitem[Collet et~al., 2014]{collet2014herbdisc}
Collet, A., Xiong, B., Gurau, C., Hebert, M., and Srinivasa, S.~S. (2014).
\newblock Herbdisc: Towards lifelong robotic object discovery.
\newblock {\em The International Journal of Robotics Research}.

\bibitem[{Collet Romea} et~al., 2009]{Collet}
{Collet Romea}, A., Berenson, D., Srinivasa, S., and {Ferguson }, D. (2009).
\newblock Object recognition and full pose registration from a single image for
  robotic manipulation.
\newblock In {\em Robotics and Automation, (ICRA 2009) IEEE International
  Conference on}.

\bibitem[Cousins et~al., 2010]{Cousins2010a}
Cousins, S., Gerkey, B., Conley, K., and Garage, W. (2010).
\newblock Welcome to {ROS} topics.
\newblock {\em Robotics Automation Magazine, IEEE}, 17(1):12 --14.

\bibitem[Cover and Thomas, 2012]{cover2012elements}
Cover, T.~M. and Thomas, J.~A. (2012).
\newblock {\em Elements of information theory}.
\newblock John Wiley \& Sons.

\bibitem[Csurka et~al., 2004]{csurka2004visual}
Csurka, G., Dance, C., Fan, L., Willamowski, J., and Bray, C. (2004).
\newblock Visual categorization with bags of keypoints.
\newblock In {\em Workshop on statistical learning in computer vision, ECCV},
  volume~1, pages 1--22. Prague.

\bibitem[Daelemans and Van~den Bosch, 2005]{daelemans2005memory}
Daelemans, W. and Van~den Bosch, A. (2005).
\newblock {\em Memory-based language processing}.
\newblock Cambridge University Press.

\bibitem[Deng et~al., 2009]{deng2009imagenet}
Deng, J., Dong, W., Socher, R., Li, L.-J., Li, K., and Fei-Fei, L. (2009).
\newblock Imagenet: A large-scale hierarchical image database.
\newblock In {\em Computer Vision and Pattern Recognition, 2009. CVPR 2009.
  IEEE Conference on}, pages 248--255. IEEE.

\bibitem[Dinh and Kropac, 2006]{dinh2006multi}
Dinh, H.~Q. and Kropac, S. (2006).
\newblock Multi-resolution spin-images.
\newblock In {\em Computer Vision and Pattern Recognition, 2006 IEEE Computer
  Society Conference on}, volume~1, pages 863--870. IEEE.

\bibitem[Doumanoglou et~al., 2016]{Doumanoglou2016}
Doumanoglou, A., Kouskouridas, R., Malassiotis, S., and Kim, T.-K. (2016).
\newblock {Recovering 6D Object Pose and Predicting Next-Best-View in the
  Crowd}.
\newblock In {\em 2016 IEEE Conference on Computer Vision and Pattern
  Recognition (CVPR)}, pages 3583--3592. IEEE.

\bibitem[Dubba et~al., 2014]{dubba2014grounding}
Dubba, K.~S., De~Oliveira, M.~R., Lim, G.~H., Kasaei, H., Lopes, L.~S.,
  Tom{\'e}, A., and Cohn, A.~G. (2014).
\newblock Grounding language in perception for scene conceptualization in
  autonomous robots.
\newblock In {\em Proceedings of AAAI 2014 spring symposium on qualitative
  representations for robots}.

\bibitem[Evans, 2008]{Evans2008}
Evans, J.~S. (2008).
\newblock {Dual-processing accounts of reasoning, judgment, and social
  cognition}.
\newblock {\em Annual Review of Psychology}, 59(1):255--278.

\bibitem[F{\"a}ulhammer et~al., 2017]{faulhammer2017autonomous}
F{\"a}ulhammer, T., Ambru{\c{s}}, R., Burbridge, C., Zillich, M., Folkesson,
  J., Hawes, N., Jensfelt, P., and Vincze, M. (2017).
\newblock Autonomous learning of object models on a mobile robot.
\newblock {\em IEEE Robotics and Automation Letters}, 2(1):26--33.

\bibitem[Fei-Fei and Perona, 2005]{fei2005bayesian}
Fei-Fei, L. and Perona, P. (2005).
\newblock A bayesian hierarchical model for learning natural scene categories.
\newblock In {\em Computer Vision and Pattern Recognition, 2005. CVPR 2005.
  IEEE Computer Society Conference on}, volume~2, pages 524--531. IEEE.

\bibitem[Fischler and Bolles, 1981]{fischler}
Fischler, M.~A. and Bolles, R.~C. (1981).
\newblock Random sample consensus: A paradigm for model fitting with
  applications to image analysis and automated cartography.
\newblock {\em Commun. ACM}, 24(6):381--395.

\bibitem[Frome et~al., 2004]{frome2004recognizing}
Frome, A., Huber, D., Kolluri, R., B{u}low, T., and Malik, J. (2004).
\newblock Recognizing objects in range data using regional point descriptors.
\newblock In {\em Computer Vision-ECCV 2004}, pages 224--237. Springer.

\bibitem[Galleguillos and Belongie, 2010]{galleguillos2010context}
Galleguillos, C. and Belongie, S. (2010).
\newblock Context based object categorization: A critical survey.
\newblock {\em Computer Vision and Image Understanding}, 114(6):712--722.

\bibitem[Gama et~al., 2009]{gama2009issues}
Gama, J., Sebasti{\~a}o, R., and Rodrigues, P.~P. (2009).
\newblock Issues in evaluation of stream learning algorithms.
\newblock In {\em Proceedings of the 15th ACM SIGKDD international conference
  on Knowledge discovery and data mining}, pages 329--338. ACM.

\bibitem[Gama et~al., 2013]{gama2013evaluating}
Gama, J., Sebasti{\~a}o, R., and Rodrigues, P.~P. (2013).
\newblock On evaluating stream learning algorithms.
\newblock {\em Machine learning}, 90(3):317--346.

\bibitem[Gao et~al., 2014]{gao2014learning}
Gao, S., Tsang, I. W.-H., and Ma, Y. (2014).
\newblock Learning category-specific dictionary and shared dictionary for
  fine-grained image categorization.
\newblock {\em IEEE Transactions on Image Processing}, 23(2):623--634.

\bibitem[Griffiths and Steyvers, 2004]{griffiths2004finding}
Griffiths, T.~L. and Steyvers, M. (2004).
\newblock Finding scientific topics.
\newblock {\em Proceedings of the National academy of Sciences}, 101(suppl
  1):5228--5235.

\bibitem[Guo et~al., 2013]{guo2013trisi}
Guo, Y., Sohel, F.~A., Bennamoun, M., Lu, M., and Wan, J. (2013).
\newblock Trisi: A distinctive local surface descriptor for 3d modeling and
  object recognition.
\newblock In {\em GRAPP-IVAPP}, pages 86--93.

\bibitem[Gupta and Sukhatme, 2012]{gupta2012using}
Gupta, M. and Sukhatme, G.~S. (2012).
\newblock Using manipulation primitives for brick sorting in clutter.
\newblock In {\em Robotics and Automation (ICRA), 2012 IEEE International
  Conference on}, pages 3883--3889. IEEE.

\bibitem[Hamidreza~Kasaei et~al., 2014]{Kasaei2014}
Hamidreza~Kasaei, S., Oliveira, M., Lim, G.~H., Seabra~Lopes, L., and Tom{\'e},
  A. (2014).
\newblock An interactive open-ended learning approach for {3D} object
  recognition.
\newblock In {\em Autonomous Robot Systems and Competitions (ICARSC), 2014 IEEE
  International Conference on}, pages 47--52.

\bibitem[Hariharan and Girshick, 2017]{hariharan2017low}
Hariharan, B. and Girshick, R. (2017).
\newblock Low-shot visual recognition by shrinking and hallucinating features.
\newblock In {\em Proc. of IEEE Int. Conf. on Computer Vision (ICCV), Venice,
  Italy}.

\bibitem[Hartigan and Wong, 1979]{hartigan1979algorithm}
Hartigan, J.~A. and Wong, M.~A. (1979).
\newblock Algorithm as 136: A k-means clustering algorithm.
\newblock {\em Applied statistics}, pages 100--108.

\bibitem[He and Chen, 2008]{he2008imorl}
He, H. and Chen, S. (2008).
\newblock Imorl: Incremental multiple-object recognition and localization.
\newblock {\em Neural Networks, IEEE Transactions on}, 19(10):1727--1738.

\bibitem[Hertzberg et~al., 2014]{Hertzberg2014projrep}
Hertzberg, J., Zhang, J., Zhang, L., Rockel, S., Neumann, B., Lehmann, J.,
  Dubba, K. S.~R., Cohn, A.~G., Saffiotti, A., Pecora, F., Mansouri, M.,
  Kone\u{c}n\'{y}, {\v{S}}., G{\"u}nther, M., Stock, S., {Seabra Lopes}, L.,
  Oliveira, M., Lim, G.~H., Kasaei, H., Mokhtari, V., Hotz, L., and Bohlken, W.
  (2014).
\newblock The {RACE} project.
\newblock {\em KI - K{\"u}nstliche Intelligenz}, 28(4):297--304.

\bibitem[Hinterstoisser et~al., 2011]{Hinterstoisser2011}
Hinterstoisser, S., Holzer, S., Cagniart, C., Ilic, S., Konolige, K., Navab,
  N., and Lepetit, V. (2011).
\newblock {Multimodal templates for real-time detection of texture-less objects
  in heavily cluttered scenes}.
\newblock In {\em Proceedings of the IEEE International Conference on Computer
  Vision}, pages 858--865.

\bibitem[Hofmann, 1999]{hofmann1999probabilistic}
Hofmann, T. (1999).
\newblock Probabilistic latent semantic indexing.
\newblock In {\em Proceedings of the 22nd annual international ACM SIGIR
  conference on Research and development in information retrieval}, pages
  50--57. ACM.

\bibitem[Horn, 1984]{horn1984extended}
Horn, B.~K. (1984).
\newblock Extended gaussian images.
\newblock {\em Proceedings of the IEEE}, 72(12):1671--1686.

\bibitem[Islam et~al., 2011]{Islam2011}
Islam, M., Jahan, F., Min, J.-H., and hwan Baek, J. (2011).
\newblock Object classification based on visual and extended features for video
  surveillance application.
\newblock In {\em Control Conference (ASCC 2011), 8th Asian}.

\bibitem[Iwahashi et~al., 2010]{Iwahashi2010}
Iwahashi, N., Sugiura, K., Taguchi, R., Nagai, T., and Taniguchi, T. (2010).
\newblock Robots that learn to communicate: A developmental approach to
  personally and physically situated human-robot conversations.
\newblock In {\em AAAI Fall Symposium, Dialog with Robots(FS-10-05)}, pages
  38--43.

\bibitem[Jain and Kemp, 2010]{jain2010assistive}
Jain, A. and Kemp, C.~C. (2010).
\newblock El-e: an assistive mobile manipulator that autonomously fetches
  objects from flat surfaces.
\newblock {\em Autonomous Robots}, 28(1):45--64.

\bibitem[Jeong and Lee, 2012]{Jeong2012}
Jeong, S. and Lee, M. (2012).
\newblock Adaptive object recognition model using incremental feature
  representation and hierarchical classification.
\newblock {\em Neural Networks}, 25(0):130 -- 140.

\bibitem[Johns et~al., 2016]{Johns2016}
Johns, E., Leutenegger, S., and Davison, A.~J. (2016).
\newblock {Pairwise Decomposition of Image Sequences for Active Multi-view
  Recognition}.
\newblock In {\em 2016 IEEE Conference on Computer Vision and Pattern
  Recognition (CVPR)}, pages 3813--3822. IEEE.

\bibitem[Johnson and Hebert, 1999]{johnson1999using}
Johnson, A.~E. and Hebert, M. (1999).
\newblock Using spin images for efficient object recognition in cluttered 3d
  scenes.
\newblock {\em Pattern Analysis and Machine Intelligence, IEEE Transactions
  on}, 21(5):433--449.

\bibitem[Kang et~al., 2011]{kang2011discovering}
Kang, H., Hebert, M., and Kanade, T. (2011).
\newblock Discovering object instances from scenes of daily living.
\newblock In {\em Computer Vision (ICCV), 2011 IEEE International Conference
  on}, pages 762--769. IEEE.

\bibitem[Karpathy et~al., 2013]{karpathy2013object}
Karpathy, A., Miller, S., and Fei-Fei, L. (2013).
\newblock Object discovery in 3{D} scenes via shape analysis.
\newblock In {\em Robotics and Automation (ICRA), 2013 IEEE International
  Conference on}, pages 2088--2095. IEEE.

\bibitem[Kasaei et~al., 2016a]{kasaeiNips2016}
Kasaei, H., Tome, A.~M., and Seabra~Lopes, L. (2016a).
\newblock Hierarchical object representation for open-ended object category
  learning and recognition.
\newblock In Lee, D.~D., Sugiyama, M., Luxburg, U.~V., Guyon, I., and Garnett,
  R., editors, {\em Advances in Neural Information Processing Systems (NIPS
  2016) 29}, pages 1948--1956. Curran Associates, Inc.

\bibitem[Kasaei et~al., 2015a]{KasaeiInteractive2015}
Kasaei, S., Oliveira, M., Lim, G., Seabra~Lopes, L., and Tome, A.~M. (2015a).
\newblock Interactive open-ended learning for 3d object recognition: An
  approach and experiments.
\newblock {\em Journal of Intelligent \& Robotic Systems}, 80(3-4):537--553.

\bibitem[Kasaei et~al., 2016b]{kasaei2016concurrent}
Kasaei, S.~H., Lopes, L.~S., and Tom{\'e}, A.~M. (2016b).
\newblock Concurrent 3d object category learning and recognition based on topic
  modelling and human feedback.
\newblock In {\em Autonomous Robot Systems and Competitions (ICARSC), 2016
  International Conference on}, pages 329--334. IEEE.

\bibitem[Kasaei et~al., 2018a]{kasaeiIROS2018}
Kasaei, S.~H., Lopes, L.~S., and Tom{\'e}, A.~M. (2018a).
\newblock Coping with context change in open-ended object recognition without
  explicit context information.
\newblock In {\em 2018 IEEE/RSJ International Conference on Intelligent Robots
  and Systems (IROS2018)}.

\bibitem[Kasaei et~al., 2015b]{kasaei2015adaptive}
Kasaei, S.~H., Oliveira, M., Lim, G.~H., Lopes, L.~S., and Tom{\'e}, A.~M.
  (2015b).
\newblock An adaptive object perception system based on environment exploration
  and bayesian learning.
\newblock In {\em Autonomous Robot Systems and Competitions (ICARSC), 2015 IEEE
  International Conference on}, pages 221--226. IEEE.

\bibitem[Kasaei et~al., 2015c]{kasaei2015interactive}
Kasaei, S.~H., Oliveira, M., Lim, G.~H., Lopes, L.~S., and Tom{\'e}, A.~M.
  (2015c).
\newblock Interactive open-ended learning for 3d object recognition: An
  approach and experiments.
\newblock {\em Journal of Intelligent and Robotic Systems}, 80(3):537--553.

\bibitem[Kasaei et~al., 2018b]{kasaei2017Neurocomputing}
Kasaei, S.~H., Oliveira, M., Lim, G.~H., {Seabra Lopes}, L., and Tom{\'e},
  A.~M. (2018b).
\newblock Towards lifelong assistive robotics: A tight coupling between object
  perception and manipulation.
\newblock {\em Neurocomputing}, 291:151 -- 166.

\bibitem[Kasaei et~al., 2016c]{kasaeiobjectNIPS}
Kasaei, S.~H., Seabra~Lopes, L., and Tom{\'e}, A.~M. (2016c).
\newblock An object perception framework for open-ended object
  conceptualization from experiences.
\newblock In {\em Workshop on Continual Learning and Deep Networks, Advances in
  Neural Information Processing Systems (NIPS), Spain}.

\bibitem[Kasaei et~al., 2013]{kasaeiRSSinstance}
Kasaei, S.~H., Seabra~Lopes, L., Tom{\'e}, A.~M., Chauhan, A., and Mokhtari, V.
  (2013).
\newblock An instance-based approach to 3d object recognition in open-ended
  robotic domains.
\newblock In {\em 4{th} Workshop on RGB-D: Advanced Reasoning with Depth
  Cameras, Robotics: Science and Systems (RSS), 2013, Germany}.

\bibitem[Kasaei et~al., 2016d]{kasaei2016orthographic}
Kasaei, S.~H., Seabra~Lopes, L., Tom{\'e}, A.~M., and Oliveira, M. (2016d).
\newblock An orthographic descriptor for 3d object learning and recognition.
\newblock In {\em Intelligent Robots and Systems (IROS), 2016 IEEE/RSJ
  International Conference on}, pages 4158--4163. IEEE.

\bibitem[Kasaei et~al., 2016e]{kasaei2016object}
Kasaei, S.~H., Shafii, N., Seabra~Lopes, L., and Tom{\'e}, A.~M. (2016e).
\newblock Object learning and grasping capabilities for robotic home
  assistants.
\newblock {\em LectureNotes in Computer Science. Springer}, 9776.

\bibitem[Kasaei et~al., 2018c]{kasaeiAAAI2018}
Kasaei, S.~H., Sock, J., Lopes, L.~S., Tom{\'e}, A.~M., and Kim, T.-K. (2018c).
\newblock Perceiving, learning, and recognizing 3d objects: An approach to
  cognitive service robots.
\newblock In {\em Thirty-Second Conference on Artificial Intelligence
  (AAAI-18), New Orleans, Louisiana, USA}.

\bibitem[Kasaei et~al., 2016f]{kasaei2016hierarchical}
Kasaei, S.~H., Tom{\'e}, A.~M., and Lopes, L.~S. (2016f).
\newblock Hierarchical object representation for open-ended object category
  learning and recognition.
\newblock In {\em Advances in Neural Information Processing Systems}, pages
  1948--1956.

\bibitem[Kasaei et~al., 2016g]{GOODKasaei2016}
Kasaei, S.~H., Tomé, A.~M., Lopes, L.~S., and Oliveira, M. (2016g).
\newblock Good: A global orthographic object descriptor for 3d object
  recognition and manipulation.
\newblock {\em Pattern Recognition Letters}, pages~--.

\bibitem[Katz et~al., 2013]{katz2013clearing}
Katz, D., Kazemi, M., Bagnell, J.~A., and Stentz, A. (2013).
\newblock Clearing a pile of unknown objects using interactive perception.
\newblock In {\em Robotics and Automation (ICRA), 2013 IEEE International
  Conference on}, pages 154--161. IEEE.

\bibitem[Katz et~al., 2014]{katz2014perceiving}
Katz, D., Venkatraman, A., Kazemi, M., Bagnell, J.~A., and Stentz, A. (2014).
\newblock Perceiving, learning, and exploiting object affordances for
  autonomous pile manipulation.
\newblock {\em Autonomous Robots}, 37(4):369--382.

\bibitem[Khoshelham, 2011]{khoshelham2011accuracy}
Khoshelham, K. (2011).
\newblock Accuracy analysis of kinect depth data.
\newblock In {\em ISPRS workshop laser scanning}, volume~38, page W12.

\bibitem[Kim et~al., 2009]{kim2009adaptation}
Kim, J.~G., Biederman, I., Lescroart, M.~D., and Hayworth, K.~J. (2009).
\newblock Adaptation to objects in the lateral occipital complex (loc): shape
  or semantics?
\newblock {\em Vision research}, 49(18):2297--2305.

\bibitem[Kirstein et~al., 2012a]{kirstein2012}
Kirstein, S., Wersing, H., Gross, H.-M., and K{\"o}rner, E. (2012a).
\newblock A life-long learning vector quantization approach for interactive
  learning of multiple categories.
\newblock {\em Neural Networks}, 28:90 -- 105.

\bibitem[Kirstein et~al., 2012b]{Kirstein}
Kirstein, S., Wersing, H., Gross, H.-M., and K{\"o}rner, E. (2012b).
\newblock A life-long learning vector quantization approach for interactive
  learning of multiple categories.
\newblock {\em Neural Networks}, 28:90 -- 105.

\bibitem[Kokinov, 1995]{kokinov1995dynamic}
Kokinov, B. (1995).
\newblock A dynamic approach to context modeling.
\newblock 95(11):199--209.

\bibitem[Kokinov, 1997]{kokinov1997dynamic}
Kokinov, B. (1997).
\newblock A dynamic theory of implicit context.
\newblock In {\em Proceedings of the 2nd European conference on cognitive
  science}, pages 9--11.

\bibitem[Kootstra et~al., 2008]{Kootstra}
Kootstra, G., Ypma, J., and De~Boer, B. (2008).
\newblock Active exploration and keypoint clustering for object recognition.
\newblock In {\em Robotics and Automation, (ICRA 2008) IEEE International
  Conference on}, pages 1005--1010.

\bibitem[Krainin et~al., 2010]{krainin2010manipulator}
Krainin, M., Henry, P., Ren, X., and Fox, D. (2010).
\newblock Manipulator and object tracking for in hand model acquisition.
\newblock In {\em Proceedings, IEEE International Conference on Robots and
  Automation}.

\bibitem[Lai et~al., 2014]{lai2014unsupervised}
Lai, K., Bo, L., and Fox, D. (2014).
\newblock Unsupervised feature learning for 3d scene labeling.
\newblock In {\em Robotics and Automation (ICRA), 2014 IEEE International
  Conference on}, pages 3050--3057. IEEE.

\bibitem[Lai et~al., 2011a]{Lai2011}
Lai, K., Bo, L., Ren, X., and Fox, D. (2011a).
\newblock A large-scale hierarchical multi-view rgb-d object dataset.
\newblock In {\em Robotics and Automation (ICRA), 2011 IEEE International
  Conference on}, pages 1817--1824.

\bibitem[Lai et~al., 2011b]{lai2011large}
Lai, K., Bo, L., Ren, X., and Fox, D. (2011b).
\newblock A large-scale hierarchical multi-view rgb-d object dataset.
\newblock In {\em Robotics and Automation (ICRA), 2011 IEEE International
  Conference on}, pages 1817--1824. IEEE.

\bibitem[Laird, 2012]{laird2012soar}
Laird, J.~E. (2012).
\newblock {\em The Soar cognitive architecture}.
\newblock MIT Press.

\bibitem[Laird et~al., 2012]{laird2012cognitive}
Laird, J.~E., Kinkade, K.~R., Mohan, S., and Xu, J.~Z. (2012).
\newblock Cognitive robotics using the soar cognitive architecture.
\newblock {\em Cognitive Robotics AAAI Technical Report WS-12-06. Accessed},
  pages 46--54.

\bibitem[Langley et~al., 2009]{langley2009cognitive}
Langley, P., Laird, J.~E., and Rogers, S. (2009).
\newblock Cognitive architectures: Research issues and challenges.
\newblock {\em Cognitive Systems Research}, 10(2):141--160.

\bibitem[Leroux and Lebec, 2013]{leroux2013armen}
Leroux, C. and Lebec, A. (2013).
\newblock Armen: Assistive robotics to maintain elderly people in natural
  environment.
\newblock {\em IRBM}, 34(2):101--107.

\bibitem[Li et~al., 2016]{li2016fpnn}
Li, Y., Pirk, S., Su, H., Qi, C.~R., and Guibas, L.~J. (2016).
\newblock Fpnn: Field probing neural networks for 3d data.
\newblock {\em arXiv preprint arXiv:1605.06240}.

\bibitem[Lim et~al., 2014a]{GiHyunLim}
Lim, G.~H., Oliveira, M., Mokhtari, V., Hamidreza~Kasaei, S., Chauhan, A.,
  Seabra~Lopes, L., and Tom{\'e}, A. (2014a).
\newblock Interactive teaching and experience extraction for learning about
  objects and robot activities.
\newblock In {\em Robot and Human Interactive Communication, 2014 RO-MAN: The
  23rd IEEE International Symposium on}, pages 153--160.

\bibitem[Lim et~al., 2014b]{lim2014interactive}
Lim, G.~H., Oliveira, M., Mokhtari, V., Hamidreza~Kasaei, S., Chauhan, A.,
  {Seabra Lopes}, L., and Tom{\'e}, A.~M. (2014b).
\newblock Interactive teaching and experience extraction for learning about
  objects and robot activities.
\newblock In {\em Robot and Human Interactive Communication, 2014 RO-MAN: The
  23rd IEEE International Symposium on}, pages 153--160. IEEE.

\bibitem[Liu et~al., 2006]{Liu}
Liu, Y., Zha, H., and Qin, H. (2006).
\newblock Shape topics: A compact representation and new algorithms for 3d
  partial shape retrieval.
\newblock In {\em Computer Vision and Pattern Recognition, IEEE Computer
  Society Conference on}, volume~2, pages 2025--2032.

\bibitem[Lopes and Chauhan, 2007]{lopes2007many}
Lopes, L.~S. and Chauhan, A. (2007).
\newblock How many words can my robot learn?: An approach and experiments with
  one-class learning.
\newblock {\em Interaction Studies}, 8(1):53--81.

\bibitem[{Martinez Torres} et~al., 2010]{Martinez2010}
{Martinez Torres}, M., {Collet Romea}, A., and Srinivasa, S. (2010).
\newblock Moped: A scalable and low latency object recognition and pose
  estimation system.
\newblock In {\em Robotics and Automation, (ICRA 2010) IEEE International
  Conference on}.

\bibitem[Marton et~al., 2010]{marton2010hierarchical}
Marton, Z.-C., Pangercic, D., Rusu, R.~B., Holzbach, A., and Beetz, M. (2010).
\newblock Hierarchical object geometric categorization and appearance
  classification for mobile manipulation.
\newblock In {\em Humanoid Robots (Humanoids), 2010 10th IEEE-RAS International
  Conference on}, pages 365--370. IEEE.

\bibitem[Mason et~al., 2014]{mason2014unsupervised}
Mason, J., Marthi, B., and Parr, R. (2014).
\newblock Unsupervised discovery of object classes with a mobile robot.
\newblock In {\em Robotics and Automation (ICRA), 2014 IEEE International
  Conference on}, pages 3074--3081. IEEE.

\bibitem[Mauro et~al., 2014]{mauro2014unified}
Mauro, M., Riemenschneider, H., Signoroni, A., Leonardi, R., and Van~Gool, L.
  (2014).
\newblock A unified framework for content-aware view selection and planning
  through view importance.
\newblock In {\em Proceedings BMVC 2014}, pages 1--11.

\bibitem[Mcauliffe and Blei, 2008]{mcauliffe2008supervised}
Mcauliffe, J.~D. and Blei, D.~M. (2008).
\newblock Supervised topic models.
\newblock In {\em Advances in neural information processing systems}, pages
  121--128.

\bibitem[Mian et~al., 2010]{mian2010repeatability}
Mian, A., Bennamoun, M., and Owens, R. (2010).
\newblock On the repeatability and quality of keypoints for local feature-based
  3d object retrieval from cluttered scenes.
\newblock {\em International Journal of Computer Vision}, 89(2-3):348--361.

\bibitem[Mian et~al., 2006]{mian2006three}
Mian, A.~S., Bennamoun, M., and Owens, R. (2006).
\newblock Three-dimensional model-based object recognition and segmentation in
  cluttered scenes.
\newblock {\em IEEE transactions on pattern analysis and machine intelligence},
  28(10):1584--1601.

\bibitem[Miller, 1995]{miller1995wordnet}
Miller, G.~A. (1995).
\newblock Wordnet: a lexical database for english.
\newblock {\em Communications of the ACM}, 38(11):39--41.

\bibitem[Mokhtari et~al., 2016]{mokhtari2016experience}
Mokhtari, V., Lopes, L.~S., and Pinho, A.~J. (2016).
\newblock Experience-based robot task learning and planning with goal
  inference.
\newblock In {\em Twenty-Sixth International Conference on Automated Planning
  and Scheduling}.

\bibitem[Mokhtari et~al., 2017]{mokhtari2017learning}
Mokhtari, V., Lopes, L.~S., and Pinho, A.~J. (2017).
\newblock Learning robot tasks with loops from experiences to enhance robot
  adaptability.
\newblock {\em Pattern Recognition Letters}, 99:57--66.

\bibitem[Mottaghi et~al., 2014]{mottaghi2014role}
Mottaghi, R., Chen, X., Liu, X., Cho, N.-G., Lee, S.-W., Fidler, S., Urtasun,
  R., and Yuille, A. (2014).
\newblock The role of context for object detection and semantic segmentation in
  the wild.
\newblock In {\em Proceedings of the IEEE Conference on Computer Vision and
  Pattern Recognition}, pages 891--898.

\bibitem[Munaro et~al., 2013]{Munaro2013}
Munaro, M., Basso, F., Michieletto, S., Pagello, E., and Menegatti, E. (2013).
\newblock A software architecture for {RGB-D} people tracking based on {ROS}
  framework for a mobile robot.
\newblock In {\em Frontiers of Intelligent Autonomous Systems}, volume 466 of
  {\em Studies in Computational Intelligence}, pages 53--68.

\bibitem[Nau et~al., 2003]{Nau2003}
Nau, D.~S., Au, T.-C., Ilghami, O., Kuter, U., Murdock, J.~W., Wu, D., and
  Yaman, F. (2003).
\newblock Shop2: An htn planning system.
\newblock {\em Journal of Artificial Intelligence Research (JAIR)},
  20:379--404.

\bibitem[Nigam and Riek, 2015]{nigam2015social}
Nigam, A. and Riek, L.~D. (2015).
\newblock Social context perception for mobile robots.
\newblock In {\em Intelligent Robots and Systems (IROS), 2015 IEEE/RSJ
  International Conference on}, pages 3621--3627. IEEE.

\bibitem[Oliva and Torralba, 2007]{oliva2007role}
Oliva, A. and Torralba, A. (2007).
\newblock The role of context in object recognition.
\newblock {\em Trends in cognitive sciences}, 11(12):520--527.

\bibitem[Oliveira et~al., 2014a]{oliveira2014perceptual}
Oliveira, M., Lim, G.~H., Seabra~Lopes, L., Hamidreza~Kasaei, S., Tome, A.~M.,
  and Chauhan, A. (2014a).
\newblock A perceptual memory system for grounding semantic representations in
  intelligent service robots.
\newblock In {\em Intelligent Robots and Systems (IROS 2014), 2014 IEEE/RSJ
  International Conference on}, pages 2216--2223. IEEE.

\bibitem[Oliveira et~al., 2014b]{Oliveira2014}
Oliveira, M., Lim, G.~H., {Seabra Lopes}, L., Kasaei, H., Tome, A., and
  Chauhan, A. (2014b).
\newblock A perceptual memory system for grounding semantic representations in
  intelligent service robots.
\newblock In {\em Proceedings of the IEEE/RSJ International Conference on
  Intelligent Robots and Systems (IROS)}, Chicago, Illinois. IEEE.

\bibitem[Oliveira et~al., 2015a]{Oliveira2015}
Oliveira, M., Lopes, L.~S., Lim, G.~H., Kasaei, S.~H., Sappa, A.~D., and
  Tom{\'e}, A.~M. (2015a).
\newblock Concurrent learning of visual codebooks and object categories in
  open-ended domains.
\newblock In {\em Intelligent Robots and Systems (IROS), 2015 IEEE/RSJ
  International Conference on}, pages 2488--2495. IEEE.

\bibitem[Oliveira et~al., 2016]{Oliveira2016614}
Oliveira, M., Lopes, L.~S., Lim, G.~H., Kasaei, S.~H., Tomé, A.~M., and
  Chauhan, A. (2016).
\newblock 3{D} object perception and perceptual learning in the {RACE} project.
\newblock {\em Robotics and Autonomous Systems}, 75, Part B:614 -- 626.

\bibitem[Oliveira et~al., 2015b]{oliveira20153d}
Oliveira, M., {Seabra Lopes}, L., Lim, G.~H., Kasaei, S.~H., Tom{\'e}, A.~M.,
  and Chauhan, A. (2015b).
\newblock {3D} object perception and perceptual learning in the race project.
\newblock {\em Robotics and Autonomous Systems}.

\bibitem[Osada et~al., 2002]{osada2002shape}
Osada, R., Funkhouser, T., Chazelle, B., and Dobkin, D. (2002).
\newblock Shape distributions.
\newblock {\em ACM Transactions on Graphics (TOG)}, 21(4):807--832.

\bibitem[Pang and Neumann, 2015]{pang2015fast}
Pang, G. and Neumann, U. (2015).
\newblock Fast and robust multi-view 3d object recognition in point clouds.
\newblock In {\em 3D Vision (3DV), 2015 International Conference on}, pages
  171--179. IEEE.

\bibitem[Pasqualotto et~al., 2013]{pasqualotto2013combining}
Pasqualotto, G., Zanuttigh, P., and Cortelazzo, G.~M. (2013).
\newblock Combining color and shape descriptors for 3d model retrieval.
\newblock {\em Signal Processing: Image Communication}, 28(6):608--623.

\bibitem[Pele, 2011]{pele2011distance}
Pele, O. (2011).
\newblock {\em Distance functions: Theory, algorithms and applications}.
\newblock Citeseer.

\bibitem[Pele and Werman, 2008]{pele2008linear}
Pele, O. and Werman, M. (2008).
\newblock A linear time histogram metric for improved sift matching.
\newblock In {\em European conference on computer vision}, pages 495--508.
  Springer.

\bibitem[Pele and Werman, 2009]{pele2009fast}
Pele, O. and Werman, M. (2009).
\newblock Fast and robust earth mover's distances.
\newblock In {\em Computer vision, 2009 IEEE 12th international conference on},
  pages 460--467. IEEE.

\bibitem[Philipona et~al., 2003]{Philipona6789645}
Philipona, D., O'Regan, J.~K., and Nadal, J.~P. (2003).
\newblock Is there something out there? inferring space from sensorimotor
  dependencies.
\newblock {\em Neural Computation}, 15(9):2029--2049.

\bibitem[Pomerleau et~al., 2013]{pomerleau2013comparing}
Pomerleau, F., Colas, F., Siegwart, R., and Magnenat, S. (2013).
\newblock Comparing icp variants on real-world data sets.
\newblock {\em Autonomous Robots}, 34(3):133--148.

\bibitem[Porteous et~al., 2008]{porteous2008fast}
Porteous, I., Newman, D., Ihler, A., Asuncion, A., Smyth, P., and Welling, M.
  (2008).
\newblock Fast collapsed gibbs sampling for latent dirichlet allocation.
\newblock In {\em Proceedings of the 14th ACM SIGKDD international conference
  on Knowledge discovery and data mining}, pages 569--577. ACM.

\bibitem[Powers, 2011]{powers2011evaluation}
Powers, D.~M. (2011).
\newblock Evaluation: from precision, recall and f-measure to roc,
  informedness, markedness and correlation.

\bibitem[Qian et~al., 2012]{qian2012learning}
Qian, T., Jaeger, T.~F., and Aslin, R.~N. (2012).
\newblock Learning to represent a multi-context environment: more than
  detecting changes.
\newblock {\em Frontiers in Psychology}, 3:228.

\bibitem[Quigley et~al., 2009]{quigley2009ros}
Quigley, M., Conley, K., Gerkey, B., Faust, J., Foote, T., Leibs, J., Wheeler,
  R., and Ng, A.~Y. (2009).
\newblock Ros: an open-source robot operating system.
\newblock 3(3.2):5.

\bibitem[Ramage et~al., 2009]{ramage2009labeled}
Ramage, D., Hall, D., Nallapati, R., and Manning, C.~D. (2009).
\newblock Labeled lda: A supervised topic model for credit attribution in
  multi-labeled corpora.
\newblock In {\em Proceedings of the 2009 Conference on Empirical Methods in
  Natural Language Processing: Volume 1-Volume 1}, pages 248--256.

\bibitem[Regazzoni et~al., 2014]{regazzoni2014rgb}
Regazzoni, D., de~Vecchi, G., and Rizzi, C. (2014).
\newblock Rgb cams vs rgb-d sensors: low cost motion capture technologies
  performances and limitations.
\newblock {\em Journal of Manufacturing Systems}, 33(4):719--728.

\bibitem[Riesenhuber and Poggio, 1999]{riesenhuber1999hierarchical}
Riesenhuber, M. and Poggio, T. (1999).
\newblock Hierarchical models of object recognition in cortex.
\newblock {\em Nature neuroscience}, 2(11):1019--1025.

\bibitem[Rockel and et~al., 2013]{Rockel20132}
Rockel, S. and et~al. (2013).
\newblock An ontology-based multi-level robot architecture for learning from
  experiences.
\newblock In {\em Designing Intelligent Robots: Reintegrating AI II, AAAI
  Spring Symposium}, Stanford (USA).

\bibitem[Rockel et~al., 2013]{RACE2013}
Rockel, S., Neumann, B., Zhang, J., Dubba, K. S.~R., Cohn, A.~G., \u{S}.
  Kone\u{c}n\'{y}, Mansouri, M., Pecora, F., Saffiotti, A., G\"{u}nther, M.,
  Stock, S., Hertzberg, J., Tom\'{e}, A.~M., Pinho, A.~J., {Seabra Lopes}, L.,
  von Riegen, S., and Hotz, L. (2013).
\newblock An ontology-based multi-level robot architecture for learning from
  experiences.
\newblock In {\em Designing Intelligent Robots: Reintegrating AI II, AAAI
  Spring Symposium on}, Stanford (USA).

\bibitem[Rosas and Callejas-Aguilera, 2006]{rosas2006context}
Rosas, J.~M. and Callejas-Aguilera, J.~E. (2006).
\newblock Context switch effects on acquisition and extinction in human
  predictive learning.
\newblock {\em Journal of Experimental Psychology: learning, Memory, and
  cognition}, 32(3):461.

\bibitem[Rosas et~al., 2013]{rosas2013context}
Rosas, J.~M., Todd, T.~P., and Bouton, M.~E. (2013).
\newblock Context change and associative learning.
\newblock {\em Wiley Interdisciplinary Reviews: Cognitive Science},
  4(3):237--244.

\bibitem[Ruiz-Sarmiento et~al., 2015a]{ruiz2015joint}
Ruiz-Sarmiento, J.-R., Galindo, C., and Gonz{\'a}lez-Jim{\'e}nez, J. (2015a).
\newblock Joint categorization of objects and rooms for mobile robots.
\newblock In {\em Intelligent Robots and Systems (IROS), 2015 IEEE/RSJ
  International Conference on}, pages 2523--2528. IEEE.

\bibitem[Ruiz-Sarmiento et~al., 2015b]{RuizSarmiento20158805}
Ruiz-Sarmiento, J.-R., Galindo, C., and Gonzalez-Jimenez, J. (2015b).
\newblock Scene object recognition for mobile robots through semantic knowledge
  and probabilistic graphical models.
\newblock {\em Expert Systems with Applications}, 42(22):8805 -- 8816.

\bibitem[Rusu et~al., 2009a]{rusu2009fast}
Rusu, R.~B., Blodow, N., and Beetz, M. (2009a).
\newblock Fast point feature histograms (fpfh) for {3D} registration.
\newblock In {\em Robotics and Automation, 2009. ICRA'09. IEEE International
  Conference on}, pages 3212--3217. IEEE.

\bibitem[Rusu et~al., 2010]{rusu2010fast}
Rusu, R.~B., Bradski, G., Thibaux, R., and Hsu, J. (2010).
\newblock Fast {3D} recognition and pose using the viewpoint feature histogram.
\newblock In {\em Intelligent Robots and Systems (IROS), 2010 IEEE/RSJ
  International Conference on}, pages 2155--2162. IEEE.

\bibitem[Rusu et~al., 2009b]{rusu2009detecting}
Rusu, R.~B., Holzbach, A., Beetz, M., and Bradski, G. (2009b).
\newblock Detecting and segmenting objects for mobile manipulation.
\newblock In {\em Computer Vision Workshops (ICCV Workshops), 2009 IEEE 12th
  International Conference on}, pages 47--54. IEEE.

\bibitem[Rusu et~al., 2008]{rusu2008towards}
Rusu, R.~B., Marton, Z.~C., Blodow, N., Dolha, M., and Beetz, M. (2008).
\newblock Towards 3d point cloud based object maps for household environments.
\newblock {\em Robotics and Autonomous Systems}, 56(11):927--941.

\bibitem[Sahib, 2013]{sahib2013}
Sahib, S. G.~G. (2013).
\newblock A review of non relational databases, their types, advantages and
  disadvantages.
\newblock {\em International Journal of Engineering \& Technology}, 2(2).

\bibitem[Scheutz et~al., 2013]{scheutz2013novel}
Scheutz, M., Briggs, G., Cantrell, R., Krause, E., Williams, T., and Veale, R.
  (2013).
\newblock Novel mechanisms for natural human-robot interactions in the diarc
  architecture.
\newblock In {\em Proceedings of AAAI Workshop on Intelligent Robotic Systems}.

\bibitem[Schulz et~al., 2001]{Schulz}
Schulz, D., Burgard, W., Fox, D., and Cremers, A. (2001).
\newblock Tracking multiple moving targets with a mobile robot using particle
  filters and statistical data association.
\newblock In {\em Robotics and Automation, (ICRA 2001) IEEE International
  Conference on}, volume~2, pages 1665--1670 vol.2.

\bibitem[Seabra~Lopes and Chauhan, 2007]{Seabra2007}
Seabra~Lopes, L. and Chauhan, A. (2007).
\newblock How many words can my robot learn?: An approach and experiments with
  one-class learning.
\newblock {\em Interaction Studies}, 8(1):53 -- 81.

\bibitem[{Seabra Lopes} and Chauhan, 2008]{Lopes2008}
{Seabra Lopes}, L. and Chauhan, A. (2008).
\newblock Open-ended category learning for language acquisition.
\newblock {\em Connect. Sci}, 20(4):277--297.

\bibitem[Seabra~Lopes and Connell, 2001]{Lopes2001}
Seabra~Lopes, L. and Connell, J. (2001).
\newblock Semisentient robots: routes to integrated intelligence.
\newblock {\em Intelligent Systems, IEEE}, 16(5):10--14.

\bibitem[Seabra~Lopes and Wang, 2002]{lopes2002towards}
Seabra~Lopes, L. and Wang, Q. (2002).
\newblock Towards grounded human-robot communication.
\newblock In {\em Robot and Human Interactive Communication, 2002. Proceedings.
  11th IEEE International Workshop on}, pages 312--318. IEEE.

\bibitem[Shafii et~al., 2016]{shafii2016learning}
Shafii, N., Kasaei, S.~H., and Lopes, L.~S. (2016).
\newblock Learning to grasp familiar objects using object view recognition and
  template matching.
\newblock In {\em Intelligent Robots and Systems (IROS), 2016 IEEE/RSJ
  International Conference on}, pages 2895--2900. IEEE.

\bibitem[Shelhamer et~al., 2016]{shelhamer2016fully}
Shelhamer, E., Long, J., and Darrell, T. (2016).
\newblock Fully convolutional networks for semantic segmentation.
\newblock {\em IEEE Transactions on Pattern Analysis and Machine Intelligence}.

\bibitem[Sissons and Miller, 2009]{sissons2009spontaneous}
Sissons, H.~T. and Miller, R.~R. (2009).
\newblock Spontaneous recovery of excitation and inhibition.
\newblock {\em Journal of Experimental Psychology: Animal Behavior Processes},
  35(3):419.

\bibitem[Sivic et~al., 2005]{sivic2005discovering}
Sivic, J., Russell, B.~C., Efros, A., Zisserman, A., Freeman, W.~T., et~al.
  (2005).
\newblock Discovering objects and their location in images.
\newblock In {\em Computer Vision, 2005. ICCV 2005. Tenth IEEE International
  Conference on}, volume~1, pages 370--377. IEEE.

\bibitem[Sko{\v{c}}aj et~al., 2016]{skovcaj2016integrated}
Sko{\v{c}}aj, D., Vre{\v{c}}ko, A., Mahni{\v{c}}, M., et~al. (2016).
\newblock An integrated system for interactive continuous learning of
  categorical knowledge.
\newblock {\em Journal of Experimental \& Theoretical Artificial Intelligence},
  28(5):823--848.

\bibitem[Smith and Gasser, 2005]{Smith2005}
Smith, L. and Gasser, M. (2005).
\newblock The development of embodied cognition: Six lessons from babies.
\newblock {\em Artificial life}, 11(1-2):13--29.

\bibitem[Snidaro et~al., 2015]{snidaro2015context}
Snidaro, L., Garc{\'\i}a, J., and Llinas, J. (2015).
\newblock Context-based information fusion: a survey and discussion.
\newblock {\em Information Fusion}, 25:16--31.

\bibitem[Sock et~al., 2017]{JuilICCVW2017}
Sock, J., Kasaei, S., {Seabra Lopes}, L., and Kim, T.-K. (2017).
\newblock Multi-view 6{D} object pose estimation and camera motion planning
  using rgbd images.
\newblock In {\em International Conference on Computer Vision (ICCV), 3rd
  International Workshop on Recovering 6D Object Pose}. IEEE.

\bibitem[Srinivasa et~al., 2008]{srinivasa2008robotic}
Srinivasa, S., Ferguson, D.~I., Vande~Weghe, M., Diankov, R., Berenson, D.,
  Helfrich, C., and Strasdat, H. (2008).
\newblock The robotic busboy: Steps towards developing a mobile robotic home
  assistant.
\newblock In {\em International Conference on Intelligent Autonomous Systems},
  pages 2155--2162.

\bibitem[Srinivasa et~al., 2010]{srinivasa2010herb}
Srinivasa, S.~S., Ferguson, D., Helfrich, C.~J., Berenson, D., Collet, A.,
  Diankov, R., Gallagher, G., Hollinger, G., Kuffner, J., and Weghe, M.~V.
  (2010).
\newblock Herb: a home exploring robotic butler.
\newblock {\em Autonomous Robots}, 28(1):5--20.

\bibitem[Steels and Kaplan, 2000]{steels2000aibo}
Steels, L. and Kaplan, F. (2000).
\newblock Aibo’s first words: The social learning of language and meaning.
\newblock {\em Evolution of communication}, 4(1):3--32.

\bibitem[Steels and Kaplan, 2002]{Steels2002}
Steels, L. and Kaplan, F. (2002).
\newblock {AIBO}'s first words: The social learning of language and meaning.
\newblock {\em Evolution of Communication}, 4(1):3--32.

\bibitem[St{\"u}ckler et~al., 2013]{stuckler2013efficient}
St{\"u}ckler, J., Steffens, R., Holz, D., and Behnke, S. (2013).
\newblock Efficient {3D} object perception and grasp planning for mobile
  manipulation in domestic environments.
\newblock {\em Robotics and Autonomous Systems}, 61(10):1106--1115.

\bibitem[Su et~al., 2015]{su2015multi}
Su, H., Maji, S., Kalogerakis, E., and Learned-Miller, E. (2015).
\newblock Multi-view convolutional neural networks for 3d shape recognition.
\newblock In {\em Proceedings of the IEEE International Conference on Computer
  Vision}, pages 945--953.

\bibitem[Tejani et~al., 2014]{Tejani2014}
Tejani, A., Tang, D., Kouskouridas, R., and Kim, T.~K. (2014).
\newblock {Latent-class Hough forests for 3D object detection and pose
  estimation}.
\newblock In {\em Lecture Notes in Computer Science (including subseries
  Lecture Notes in Artificial Intelligence and Lecture Notes in
  Bioinformatics)}, volume 8694 LNCS, pages 462--477. Springer Verlag.

\bibitem[Todorov et~al., 2012]{todorov2012mujoco}
Todorov, E., Erez, T., and Tassa, Y. (2012).
\newblock Mujoco: A physics engine for model-based control.
\newblock In {\em Intelligent Robots and Systems (IROS), 2012 IEEE/RSJ
  International Conference on}, pages 5026--5033. IEEE.

\bibitem[Tombari et~al., 2010]{tombari2010unique}
Tombari, F., Salti, S., and Di~Stefano, L. (2010).
\newblock Unique signatures of histograms for local surface description.
\newblock In {\em Computer Vision--ECCV 2010}, pages 356--369. Springer.

\bibitem[Tulving, 1991]{Tulving1991}
Tulving, E. (1991).
\newblock Concepts of human memory.
\newblock In Squire, L., Lynch, G., Weinberger, N., and McGaugh, J., editors,
  {\em Memory: Organization and locus of change}, pages 3--32. Oxford Univ.
  Press.

\bibitem[Tulving, 2005]{Tulving2005}
Tulving, E. (2005).
\newblock Episodic memory and autonoesis: Uniquely human?
\newblock In H.~S.~Terrace, . J.~M., editor, {\em The Missing Link in
  Cognition}, pages 4--56. Oxford Univ. Press, NewYork, NY.

\bibitem[Vahrenkamp et~al., 2010]{vahrenkamp2010integrated}
Vahrenkamp, N., Do, M., Asfour, T., and Dillmann, R. (2010).
\newblock Integrated grasp and motion planning.
\newblock In {\em Robotics and Automation (ICRA), 2010 IEEE International
  Conference on}, pages 2883--2888. IEEE.

\bibitem[Van~Hoof et~al., 2014]{van2014probabilistic}
Van~Hoof, H., Kroemer, O., and Peters, J. (2014).
\newblock Probabilistic segmentation and targeted exploration of objects in
  cluttered environments.
\newblock {\em IEEE Transactions on Robotics}, 30(5):1198--1209.

\bibitem[Wang et~al., 2009]{wang2009simultaneous}
Wang, C., Blei, D., and Li, F.-F. (2009).
\newblock Simultaneous image classification and annotation.
\newblock In {\em Computer Vision and Pattern Recognition, 2009. CVPR 2009.
  IEEE Conference on}, pages 1903--1910. IEEE.

\bibitem[Wang et~al., 2007]{wang2007semi}
Wang, Y., Sabzmeydani, P., and Mori, G. (2007).
\newblock Semi-latent dirichlet allocation: A hierarchical model for human
  action recognition.
\newblock In {\em Human Motion--Understanding, Modeling, Capture and
  Animation}, pages 240--254. Springer.

\bibitem[Wilson and Martinez, 2000]{wilson2000reduction}
Wilson, D.~R. and Martinez, T.~R. (2000).
\newblock Reduction techniques for instance-based learning algorithms.
\newblock {\em Machine learning}, 38(3):257--286.

\bibitem[Wohlkinger and Vincze, 2011]{wohlkinger2011ensemble}
Wohlkinger, W. and Vincze, M. (2011).
\newblock Ensemble of shape functions for 3d object classification.
\newblock In {\em Robotics and Biomimetics (ROBIO), 2011 IEEE International
  Conference on}, pages 2987--2992. IEEE.

\bibitem[Wood et~al., 2011]{Wood2011}
Wood, R., Baxter, P., and Belpaeme, T. (2011).
\newblock A review of long-term memory in natural and synthetic systems.
\newblock {\em Adaptive Behavior}, 20:81--103.

\bibitem[Wu et~al., 2015]{wu20153d}
Wu, Z., Song, S., Khosla, A., Yu, F., Zhang, L., Tang, X., and Xiao, J. (2015).
\newblock 3d shapenets: A deep representation for volumetric shapes.
\newblock In {\em Proceedings of the IEEE Conference on Computer Vision and
  Pattern Recognition}, pages 1912--1920.

\bibitem[Yang et~al., 2015]{yang2015chi}
Yang, W., Xu, L., Chen, X., Zheng, F., and Liu, Y. (2015).
\newblock Chi-squared distance metric learning for histogram data.
\newblock {\em Mathematical Problems in Engineering}, 2015.

\bibitem[Yeh and Darrell, 2008]{Yeh2008}
Yeh, T. and Darrell, T. (2008).
\newblock Dynamic visual category learning.
\newblock In {\em Computer Vision and Pattern Recognition, (CVPR 2008). IEEE
  Conference on}, pages 1--8.

\bibitem[Yeh et~al., 2009]{yeh2009fast}
Yeh, T., Lee, J.~J., and Darrell, T. (2009).
\newblock Fast concurrent object localization and recognition.
\newblock In {\em Computer Vision and Pattern Recognition, 2009. CVPR 2009.
  IEEE Conference on}, pages 280--287. IEEE.

\bibitem[Yeh and Barsalou, 2006]{yeh2006situated}
Yeh, W. and Barsalou, L.~W. (2006).
\newblock The situated nature of concepts.
\newblock {\em The American journal of psychology}, pages 349--384.

\bibitem[Zaman et~al., 2013]{Zaman2013}
Zaman, S., Steinbauer, G., Maurer, J., Lepej, P., and Uran, S. (2013).
\newblock An integrated model-based diagnosis and repair architecture for
  {ROS}-based robot systems.
\newblock In {\em {IEEE} International Conference on Robotics and Automation},
  Karlsruhe, Germany.

\bibitem[Zhang et~al., 2016]{zhang2016weakly}
Zhang, Y., Wei, X.-S., Wu, J., Cai, J., Lu, J., Nguyen, V.-A., and Do, M.~N.
  (2016).
\newblock Weakly supervised fine-grained categorization with part-based image
  representation.
\newblock {\em IEEE Transactions on Image Processing}, 25(4):1713--1725.

\bibitem[Zhong, 2009]{zhong2009intrinsic}
Zhong, Y. (2009).
\newblock Intrinsic shape signatures: A shape descriptor for 3d object
  recognition.
\newblock In {\em Computer Vision Workshops (ICCV Workshops), 2009 IEEE 12th
  International Conference on}, pages 689--696. IEEE.

\end{thebibliography}

\end{document}